\documentclass[a4paper,12pt]{report}
\usepackage[margin=1in,headheight=14.6pt,top=1.5in]{geometry}

\usepackage{blindtext}
\usepackage{fancyhdr}
\usepackage{amsmath}
\usepackage{tikz}
\usepackage{indentfirst}
\usepackage{lipsum}  
\usepackage{booktabs}
\usepackage[export]{adjustbox}
\usepackage{graphicx}
\graphicspath{ {images/} }
\usepackage{subcaption}
\usepackage{array}
\usepackage{pifont}
\usepackage{multirow}
\usepackage{wrapfig}
\usepackage{tabularx}
\usepackage{comment} 
\usepackage{xspace}
\usepackage{amsfonts}
\usepackage{xhfill}
\usepackage{setspace}
\usepackage{makecell}
\usepackage[utf8]{inputenc}
\usepackage[maxnames=99]{biblatex}
\usepackage{footnote}
\usepackage{titlesec}
\usepackage[T1]{fontenc}
\usepackage{blindtext, color}
\definecolor{gray75}{gray}{0.75}
\newcommand{\hsp}{\hspace{20pt}}
\titleformat{\chapter}[hang]{\Huge\bfseries}{\thechapter\hsp\textcolor{gray75}{|}\hsp}{0pt}{\Huge\bfseries}

\usepackage[nottoc]{tocbibind}

\usepackage[hidelinks]{hyperref}
\usepackage{cleveref}
\usepackage{etoolbox}
\makeatletter
\patchcmd{\tableofcontents}{\@starttoc{toc}}{\vspace*{-0.9cm}\@starttoc{toc}}{}{}
\makeatother

\title{Leveraging Large Vision-Language Models for Visual Communication}
\author{
Yael Vinker
\thanks{School of Mathematical Sciences, Raymond and Beverly Sackler Faculty of Exact Sciences, Tel Aviv University, Tel Aviv 6997801, Israel.
Email: {\tt yaelvinker@mail.tau.ac.il}.}
}
\date{June 2024}

\newcommand{\revision}[1]{\textcolor{black}{{#1}}}

\newcommand{\ap}[1]{``#1''}
\newcommand{\cliploss}[2]{CLIP_{\ell_{#1}}(\mathcal{#2})}
\DeclareMathOperator*{\argmin}{arg\,min}
\DeclareMathOperator*{\argmax}{arg\,max}

\newcommand{\sketch}{\mathcal{S}}
\newcommand{\image}{\mathcal{I}}
\newcommand{\renderer}{\mathcal{R}}

\newcommand{\widthabs}{0.15}
\newcommand{\widthteapot}{0.12}
\newcommand{\widthgiraffe}{0.099}
\newcommand{\widthface}{0.12}
\newcommand{\widthcats}{0.09}
\newcommand{\widthclasses}{0.09}

\makeatletter
\DeclareRobustCommand\onedot{\futurelet\@let@token\@onedot}
\def\@onedot{\ifx\@let@token.\else.\null\fi\xspace}
\def\eg{\emph{e.g}\onedot}

\def\ie{\emph{i.e}\onedot}

\def\etal{\emph{et al}\onedot}

\def\Bezier{B\'{e}zier\xspace}

\def\gtransform{\mathcal{T}}

\definecolor{limegreen}{HTML}{32CD32}
\newcommand{\cmark}{\textcolor{limegreen}{\ding{51}}}%
\newcommand{\xmark}{\textcolor{red}{\ding{55}}}%

\begin{document}
\pagestyle{fancy}
\fancyhf{}
\fancyhead[RO, LE]{\scshape\nouppercase{\leftmark}}
\fancyfoot[CO, CE]{\thepage}

\begin{titlepage}
    \begin{center}
        \includegraphics[width=0.3\linewidth]{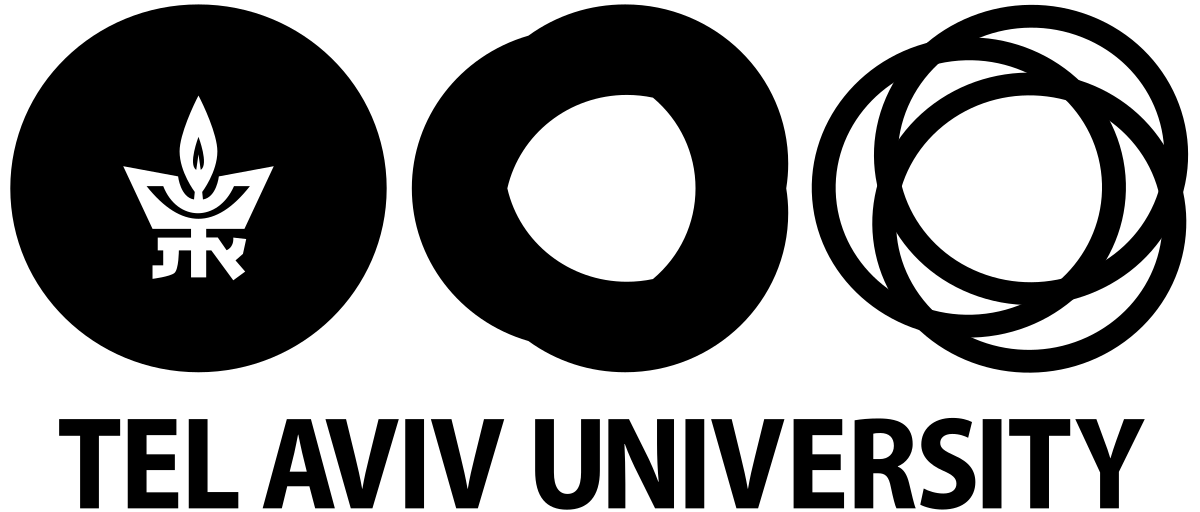}\\
        \vspace{0.7cm}
        {\Large Raymond and Beverly Sackler Faculty of Exact Sciences
Blavatnik School of Computer Science}\\
        \vspace{1.2cm}
        {\huge Generative Visual Communication in the Era of Vision-Language Models\par}
        \vspace{0.8cm}
        {\Large A dissertation submitted for the degree of Doctor of Philosophy in Computer Science in the School of Computer Science at Tel Aviv University\par}
        \vspace{0.8cm}
        {\Large By}\\
        \vspace{0.3cm}
        {\LARGE \textbf{Yael Vinker}}\\
        \vspace{1.2cm}
        {\Large The research work for the thesis has been carried out}\\
        \vspace{0.3cm}
        {\Large under the supervision of}\\
        \vspace{0.3cm}
        {\Large \textbf{Prof. Daniel Cohen-Or and Prof. Ariel Shamir}}\\
        \vspace{1cm}
        {\Large Submitted to the Senate of Tel Aviv University on June 2024}\\
        
    \end{center}
\end{titlepage}

\begin{center}
\begin{minipage}{0.8\textwidth}
\section*{\vspace{1cm}Acknowledgments}
\vspace{0.8cm}
\textit{I would like to thank my husband, Assaf, my family, and my friends for their love and support throughout this journey.
I am grateful to my advisors, Ariel Shamir, Daniel Cohen-Or, and Amit Bermano, for their guidance, mentorship, and belief in my work. Their support has been a key part of this process. \\}

\textit{A special acknowledgment goes to my grandfather, Alexander Pervin, who sparked my love for science at a young age and taught me to be curious and brave. Although he is no longer here to see this milestone, his influence has been an important part of my journey.}
\end{minipage}
\end{center}
\hfill
\pagebreak

\renewcommand{\abstractname}{\vspace{-2.68cm}Abstract}

\begin{abstract}
\vspace{-0.05cm}
Visual communication entails the use of visual elements to convey ideas and information. It is considered a design discipline, often synonymous with graphic design. This encompasses various visual mediums, including drawings, signs, icons, posters, typography, illustrations, advertising, animation, and more.
Effective visual communication designs should be concise and simple so that they convey messages clearly to a large audience, transcending geographical and cultural barriers. 
In addition, the demand for compelling and memorable designs in a world saturated with visual stimuli is becoming more and more difficult to meet.

\vspace{-0.09cm}
In this dissertation, we pose the question: Can computers automatically produce effective visual communication designs that are clear and concise across various mediums?
Recent advancements in large vision-language models have brought this question to the forefront. These models have demonstrated unprecedented capabilities in generating high-quality, realistic images from textual descriptions, and have rapidly gained popularity among professional and novice designers. 
Our aim is to demonstrate how pretrained large vision-language models can be utilized to address challenging design-related tasks within visual communication, which require rich visual knowledge and deep understanding of complex concepts. This has the potential to support designers in the challenging process of creating compelling graphic designs that deliver messages effectively. 
We examine this objective from multiple perspectives.

\vspace{-0.09cm}
In a first research direction we focus on sketches and visual abstractions -- fundamental elements of visual expression and creativity. 
We introduce \revision{two} optimization-based generative tools\revision{, CLIPasso and CLIPascene,} to generate sketches at different levels of abstraction from images.
\revision{CLIPasso focuses on object sketching, adjusting the number of vector strokes to control abstraction, while CLIPascene extends this approach to scene sketching, and broadens the notion of sketch abstraction.}
\revision{Both tools utilize} the prior of a pretrained vision-language model to guide the generation process, reducing the reliance on human-drawn sketch datasets.

\vspace{-0.09cm}
The second research direction examines typography, a central tool in visual communication. We showcase the application of automatically generating word-as-image illustrations by utilizing the prior of a text-to-image diffusion model. These illustrations aim to visually represent the meaning of a given word by manipulating the appearance of its letters.
Expanding upon this work, we explore the use of pretrained text-to-video models to animate still input sketches based on a provided text prompt. By incorporating domain-specific regularizations, we achieve success in these challenging tasks. The highly editable nature of vector representations makes these results valuable as initial solutions for complex design problems, offering designers the flexibility to refine and adapt the generated outputs. 

\vspace{-0.09cm}
In another research direction, we explore how the prior of pretrained large models can be leveraged to provide visual inspiration within the design process. \revision{In design, visual inspiration often involves decomposing a source of inspiration into its key aspects and reassembling them to meet design goals. For example, the Lotus Temple in India was inspired by the structure of the lotus flower, with the architect focusing on its form while disregarding its color.}
We utilize text-to-image personalization to decompose a visual concept into distinct aspects, organized in a hierarchical tree structure. 
These individual aspects can then be combined with elements from other visual references to foster the generation and exploration of novel visual concepts and designs.

\end{abstract}

\pagebreak

\tableofcontents

\listoffigures
\listoftables

\hfill
\pagebreak

\titlespacing*{\chapter}{0pt}{20pt}{40pt} %
\chapter{Introduction}
\titlespacing*{\chapter}{0pt}{50pt}{40pt} %

Visual communication is the conveyance of information and ideas through visual means. Visual communication is a vital aspect of human interaction that has evolved alongside our capacity to create and interpret visual stimuli \cite{meggs2016meggs,TverskyVisualizingThought2011}. Historically, visual communication traces back to prehistoric cave paintings, ancient Egyptian hieroglyphics, and medieval manuscripts (see \Cref{fig:historical} below), where visuals were used to communicate stories, convey messages, and record important events.

\begin{figure}[h]
     \centering
     \begin{subfigure}[b]{0.29\textwidth}
         \centering
         \includegraphics[height=0.7\textwidth]{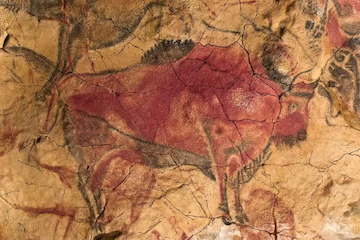}
         \caption{Cave paintings}
     \end{subfigure}
     \hfill
     \begin{subfigure}[b]{0.29\textwidth}
         \centering
         \includegraphics[height=0.7\textwidth]{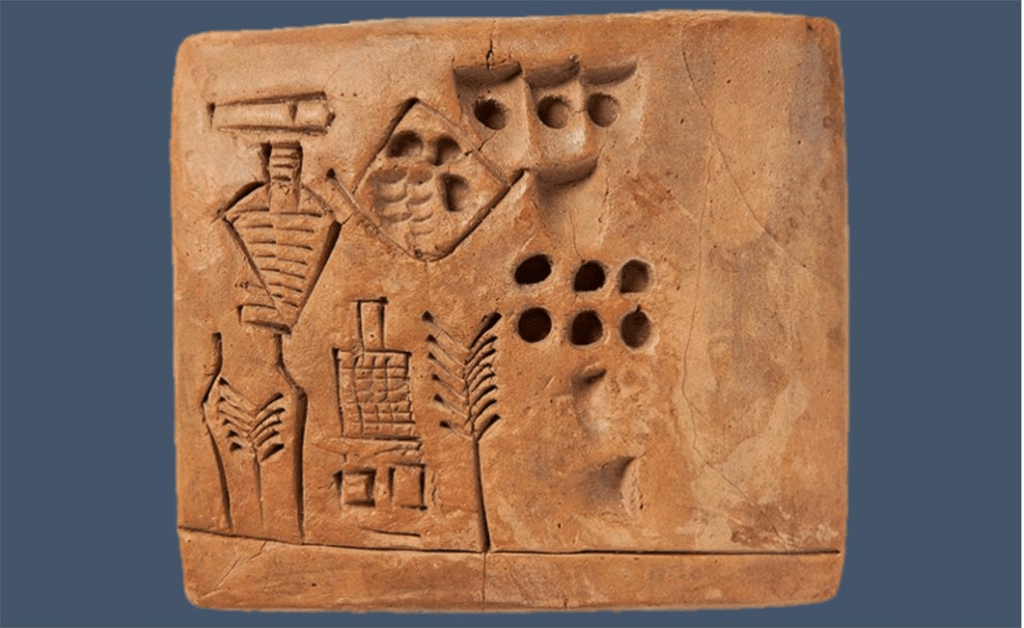}
         \caption{Sumerian tablet}
     \end{subfigure}
     \hfill
     \hfill
     \begin{subfigure}[b]{0.29\textwidth}
         \centering
         \includegraphics[height=0.7\textwidth]{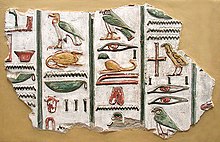}
         \caption{Egyptian hieroglyphics}
     \end{subfigure}
     \hfill
     \caption{\small Historical visual communication. (a) Cave paintings (Cantabia Spain) were the first form of visual communication. They originate to around 40,000 years ago. (b) An ancient Sumerian tablet with one of the first personal signatures in the world. (c) Egyptian hieroglyphics developed from Sumerian Cuneiform script, where images and symbols create a visual language.}
    \label{fig:historical}
\end{figure}

\begin{figure}
    \centering
    \includegraphics[width=1\linewidth]{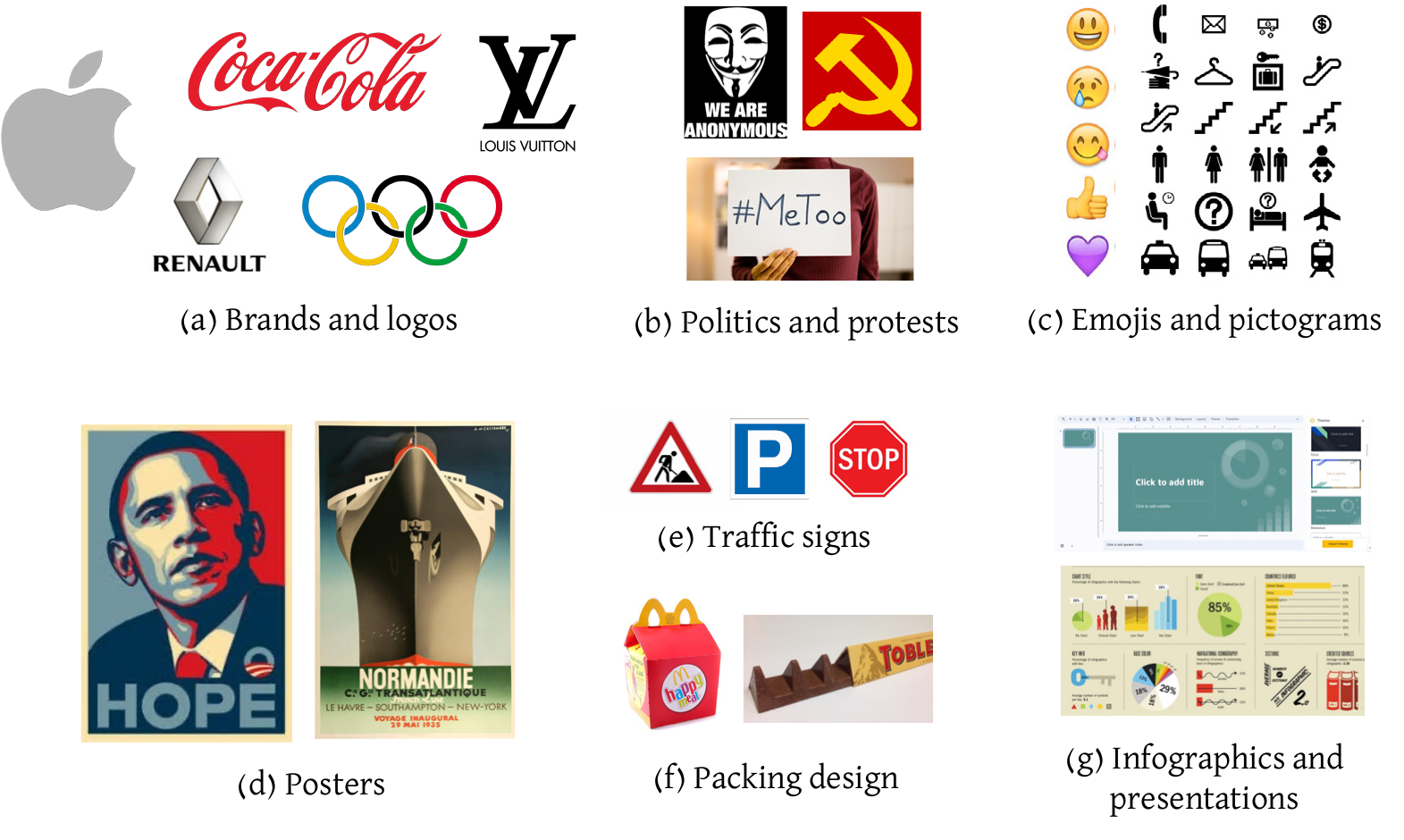}
    \vspace{-0.5cm}
    \caption{\small Examples of modern visual communication designs across many mediums.}
    \label{fig:modern-visual-com}
\end{figure}

While the term 'visual communication' encompasses a broad spectrum of visual stimuli, 'visual communication design' typically refer to a specific design discipline, often centered on graphic design. It includes a wide range of domains such as images, symbols, typography, infographics, illustrations, presentations, animation, logo design, commercials, and more.
Examples of such designs are presented in \Cref{fig:modern-visual-com}. 

Some visual communication designs are so ingrained in our daily lives that we use them without consciously recognizing them as designed objects. For example, a stop sign (\cref{fig:modern-visual-com}.e) is a red octagonal sign with the word "STOP" written in white. It's a simple yet highly effective visual cue used in traffic control. The color red signals danger or caution, the octagonal shape is highly distinguishable, and the word "STOP" clearly convey the instruction for drivers.
Another example of an effective visual communication design is the Apple logo, designed by Rob Janoff (\cref{fig:modern-visual-com}.a). The logo is a simple yet iconic image of an apple with a bite taken out of it. This minimalist design has become one of the most recognizable symbols in the world, conveying a sense of innovation and human-centric design philosophy that has come to be associated with Apple products. 
It has transcended its original purpose as a corporate logo to become a symbol of modernity and technological expertise.

Motivated by the fundamental role of visual communication in our society, in this dissertation we aim to explore the ability of computers to automatically produce effective visual communication designs across different mediums, especially in light of recent advancements in large vision-language models and generative models (also called \ap{Generative Artificial Intelligence}). 
Our primary objectives are twofold: (1) to shed light on how the hidden representations of modern \ap{Generative AI} systems can be leveraged to tackle challenging cognitive-visual tasks from a design standpoint; and (2) to develop practical generative tools that can serve and empower designers in their creative endeavors, taking into account the methodologies and mediums commonly employed by graphic designers.

We note that our goal is not to build systems or applications that would \emph{replace} graphic designers. Instead, we aim to leverage recent advancements in \ap{Generative AI} to develop tools that more synergistically support the creative process of designers.

Creating effective visual-communication designs presents a number of challenges.
First, there’s the delicate balance between clarity and complexity, requiring careful consideration in simplifying complex ideas while retaining essential details. For instance, the iconic Apple logo exemplifies simplicity yet effectively conveys a strong message of technology and prestige. Second, communicating effectively across diverse cultures presents challenges due to varying interpretations a certain visual might have. Third, designers must possess a rich interdisciplinary knowledge of concepts in fields such as history, politics, and art to ensure designs are sophisticated, up-to-date, and capable of conveying complex messages. 

\vspace{0.5cm}
Until recent years, generative models were quite limited in their ability to assist designers with complex visual communication tasks. These models primarily relied on task-specific datasets \cite{SketchRNN, CycleGAN2017, pix2pixHD,pix2pix,wangDeepVecFontSynthesizingHighquality2021,tendulkarTrickTReATThematic2019,Karras2021}, which were often limited in scope, failing to incorporate necessary social knowledge and visual understanding. Additionally, the quality of the generated content was relatively low, and the user interfaces were not intuitive, making these models more suitable for developers than for designers.

In recent years, significant advancements have been made in several parallel fields, including diffusion models \cite{Ho2020DenoisingDP,Song2020DenoisingDI}, contrastive learning \cite{Radfordclip, Chen2020ASF}, and transformers \cite{Dosovitskiy2020AnII}. Coupled with increasing computational power that facilitates the handling of extensive datasets, these developments have led to the creation of strong vision-language models (VLMs) \cite{ramesh2022hierarchical, sdxl2023, midjourney, lai2023minidalle3} that have significantly transformed the generative landscape.

\begin{figure}[h]
    \centering
    \vspace{0.5cm}
    \includegraphics[width=0.7\linewidth]{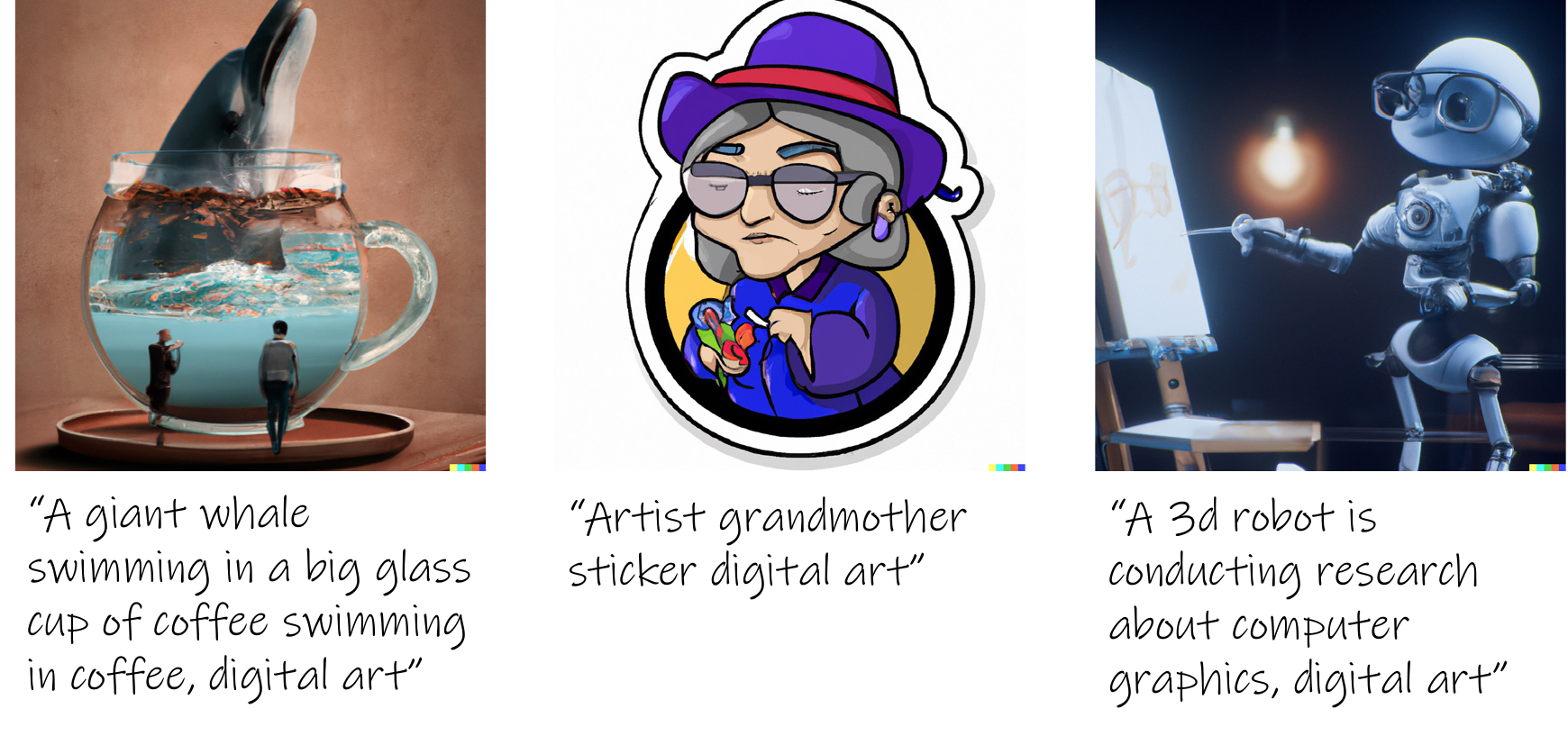}
    \caption{\small Text-to-image generation examples. These examples were created completely automatically by feeding the written text into DallE2 \cite{ramesh2022hierarchical}.}
    \label{fig:text2image}
\end{figure}

VLMs are multimodal deep neural networks that learn rich vision-language correlations from images and text \cite{Zhang2023VisionLanguageMF}.
Having been trained on millions (and recently, billions \cite{schuhmann2022laion5b}) of image-text pairs, these models possess extensive knowledge about visual and semantic concepts and have demonstrated remarkable generative abilities. High-quality visual content can be created automatically, by anyone, simply by typing a descriptive sentence \cite{ramesh2022hierarchical,sdxl2023,rombach2022highresolution}. These models can generate very realistic images depicting surprising and creative combinations of endless concepts and in numerous styles (examples are shown in \Cref{fig:text2image}). These capabilities, combined with the ease of use, facilitated their transition from the academic domain to different applications and widespread adoption, gaining popularity among professionals and amateurs alike across various disciplines \cite{cosmopolitanMagazine, DrewHarwellAI, jiang2023motiongpt, civitai, Epstein2023ArtAT}.

A significant research effort in the computer vision and graphics communities aims at enhancing VLMs in terms of generation quality \cite{sdxl2023, Saharia2022PhotorealisticTD}, speed \cite{Song2020DenoisingDI, Song2023ConsistencyM, Luo2023LatentCM}, and better user control \cite{Zhang2023AddingCC, Chefer2023AttendandExciteAS, Couairon2022DiffEditDS, Mou2023T2IAdapterLA}. 
In addition, it has been discussed that such models hold immense potential in assisting designers in their creative processes \cite{Zhang2023GenerativeIA, Smith2023TrashTT, Paananen2023UsingTG}.
We believe that these models also posses substantial potential in enhancing visual communication, reshaping how messages are conveyed.
This aspect has not yet been a large part of computational research, a gap we wish to narrow in this thesis.

In this dissertation we develop novel tools that leverage the extensive knowledge provided by pretrained VLMs to support designers in the process of concisely conveying complex messages via visual designs.
Many of the tasks we define are inspired by the author's own experience as a non-professional artist and a past visual communication student.
To this end, we explore various aspects of visual communication, namely, sketches and visual abstraction, typography, animation, and visual inspiration as described next.

\vspace{0.2cm}
\section{Visual Abstractions Through Sketches}

\begin{figure}[h]
  \begin{center}
    \vspace{0.5cm}
    \includegraphics[width=0.8\linewidth]{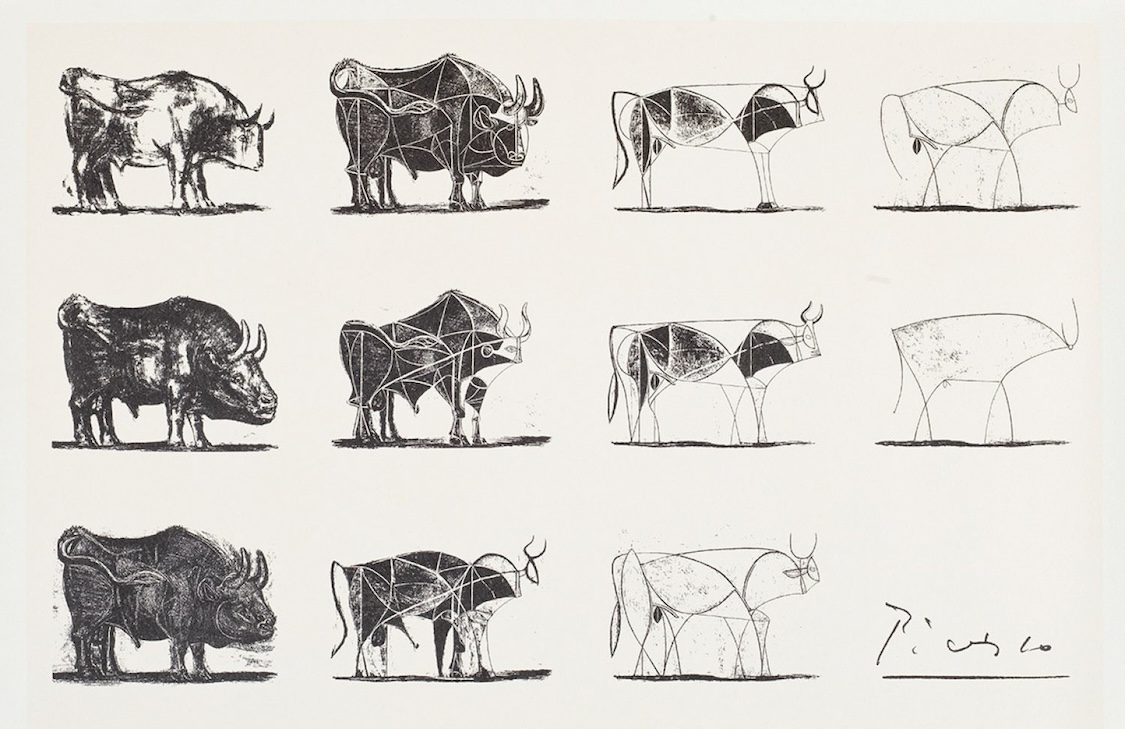}
    \end{center}
    \caption{\small \ap{Le Taureau} by Picasso — note how the abstraction process is achieved by gradually \emph{removing} elements while the bull's essence is preserved.}
    \vspace{0.5cm}
    \label{fig:picasso1}
\end{figure}

Visual abstraction refers to the process of reducing visual information to its essential features or underlying structure \cite{AbstractionViola2018, Manning1993UnderstandingCT,Fan2019PragmaticIA}\footnote{This is the definition of \ap{visual abstraction} in the context of visual communication and in the context of this dissertation. A fundamental research challenge in cognitive science, psychology, and art is to understand what visual abstraction really means.}. 
Abstraction is not merely a stylistic choice in art but a fundamental aspect of human cognition, shaping how we understand and interact with the world around us \cite{langer_feeling_1953,Arnheim1969Visual,TverskyVisualizingThought2011}.
Artists and viewers alike engage in abstraction when perceiving and creating visual representations \cite{Arnheim1969Visual}.
For example, in the famous ”Le Taureau” series (Figure \ref{fig:picasso1}), Picasso depicts the progressive abstraction of a bull. In this series of lithographs, the artist transforms a bull from a concrete, anatomical drawing, into a sketch composition of a few lines that still manages to capture the \textit{essence} of a bull.

The process of visual abstraction is central to design and to visual communication in particular \cite{Arnheim1969Visual,SimplicityMaeda,Tufte1986,lidwell_universal_2010,TverskyVisualizingThought2011}.
By distilling information to its essential elements, abstraction directs attention to key points and encourages viewer interpretation, making visuals more memorable, clear, and versatile across different mediums. 
Achieving a plausible visual abstraction that is both accurate and comprehensible is a highly challenging task, as it requires cognitive and artistic decisions about what to include and omit while ensuring coherence, clarity, and avoidance of ambiguity in the final representation.

In a first research direction, we explore the ability of large pretrained VLMs to understand and generate visual abstractions.
Since visual abstractions encompass a very wide range of representations, we narrow our investigation into a specific and fundamental domain -- \textbf{sketches}. 

Sketches are one of the most accessible and enduring technique employed by individuals across various cultures to visually express and explore ideas \cite{Fan2023DrawingAA,Hertzmann2021TheRO,Tversky2002WhatDS}.
The term `sketch' refers to the result of a rough, preliminary mark-making activity.
A sketch is an excellent example of an expressive visual representation that is highly simplified.
The act of sketching has been described as ``a conversation with ourselves in which we communicate with sketches'' \cite{laseau2000graphic, schon1986reflective,Tversky2002WhatDS}.
Sketches consist of only strokes, and often only a limited number of strokes. Hence, the process of \emph{abstraction} is central to sketching~\cite{Chamberlain2016TheGO, Fan2019PragmaticIA, Yang2021VisualCO}. 

\begin{figure}[h]
    \centering
    \vspace{0.5cm}
    \includegraphics[width=1\linewidth]{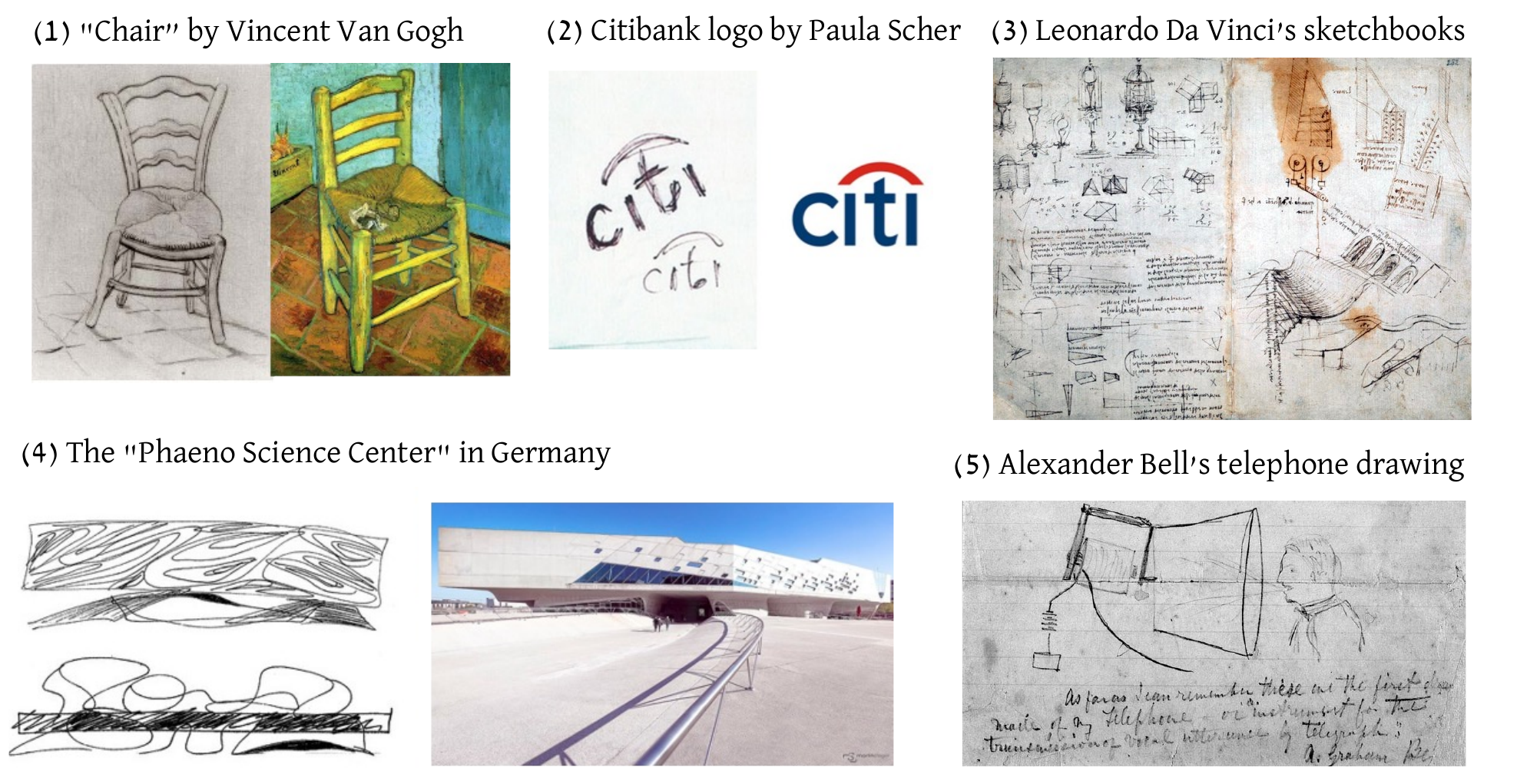}
    \vspace{-0.5cm}
    \caption{\small (1) A sketch by Vincent Van Gogh for his famous piece \ap{Chair}. (2) The Citibank logo, designed by Paula Scher, started with a sketch on a napkin during one of her first meetings with the company. (3) Drawings by Leonardo Da Vinici on the mechanical powers and forces. (4) The "Phaeno Science Center" in Germany, designed by the architect Zaha Hadid, known for her unusual approach to traditional architecture. (5) Alexander Bell's drawing of the telephone.}
    \vspace{0.9cm}
    \label{fig:sketches_of_designers}
\end{figure}

Designers use sketches to think, plan, and develop ideas \cite{Goel1995SketchesOT,VERSTIJNEN1998519, cross1982designerly,Tversky2003}.
Figure \ref{fig:sketches_of_designers} provides examples of freehand sketches used by artists, designers, and inventors, leading to some well-known designs. For instance, Van Gogh's famous artwork, the \ap{Chair} began as a simple sketch, allowing the artist to plan and contemplate the chair's form, pose, and composition before creating the detailed colored oil painting. 

While computational sketch generation has been an active research field in computer graphics and vision \cite{SketchRNN, tong2021sketch, human-like-sketches, gao2020sketchycoco, Sketchy-Database, Learning-to-Sketch-with-Deep-Q, mo2021virtualsketching, Hertzmann2003-survey, xu2020deep}, the notion of abstraction which is core to sketches is often ignored in previous works.
In \Cref{chap:clipasso,chap:clipascene}, we propose a new approach for automatic sketch generation while focusing on the importance of abstractions in the sketching process. In addition, our work was the first in the sketch domain to utilize the semantic-visual prior embedded within pre-trained VLMs instead of reliance on human-drawn sketch datasets for generation.

Our first work in this domain is CLIPasso, which converts an image of an object into a sketch.
We emphasize the importance of abstraction in sketches and focus on generating sketches with different levels of abstraction. Our method is built upon three key components: (1) Representation: We represent a sketch as a set of strokes, using vector representation, allowing us to control the level of abstraction by altering the number of strokes.
(2) We utilize a differentiable rasterizer \cite{diffvg} to compute pixel-based losses between the rasterized sketch and the input image. This enables us to backpropagate gradients and modify the parameters of strokes accordingly.
(3) We incorporate CLIP \cite{Radfordclip}, a pre-trained language-vision model, to bridge the semantic gap and guide the optimization of strokes parameters with respect to the input image.
We demonstrate the robustness of our method across various domains and levels of abstraction. Through both human evaluation and quantitative metrics, we validate that our sketches accurately depict the input object in both semantic and geometric aspects.

In our subsequent work, CLIPascene, we expand upon the foundations laid by CLIPasso to address the challenging task of \emph{scene} sketching. Furthermore, we introduce an extended notion of sketch abstraction inspired by recognizable patterns observed in human-drawn sketches.
We distinguish between two axes of abstraction. The first axis defines the \textit{fidelity} of the sketch, varying its representation from a more precise portrayal of the input to a looser, more semantic, depiction. The second axis defines the visual \textit{simplicity} of the sketch, which progresses from detailed to sparse by selectively removing strokes from a detailed sketch.

While CLIPasso relies on a predefined number of strokes to define abstraction levels, which proves sub-optimal as images vary in complexity, in CLIPascene we introduce a method to implicitly learn the optimal number of strokes per image. This allows us to generate sketches of complex scenes, including those featuring intricate backgrounds (such as natural and urban settings) and diverse subjects (such as animals and people), while progressively abstracting the input scene in terms of fidelity and simplicity.

\section{Communicative Illustrations in Typography}
In a second research direction, we investigate how recent advancements in vision-language models can be leveraged to support visual communication from the perspective of typography.

\begin{figure}[h]
     \centering
     \begin{subfigure}[b]{0.29\textwidth}
         \centering
         \includegraphics[height=1.2\textwidth]{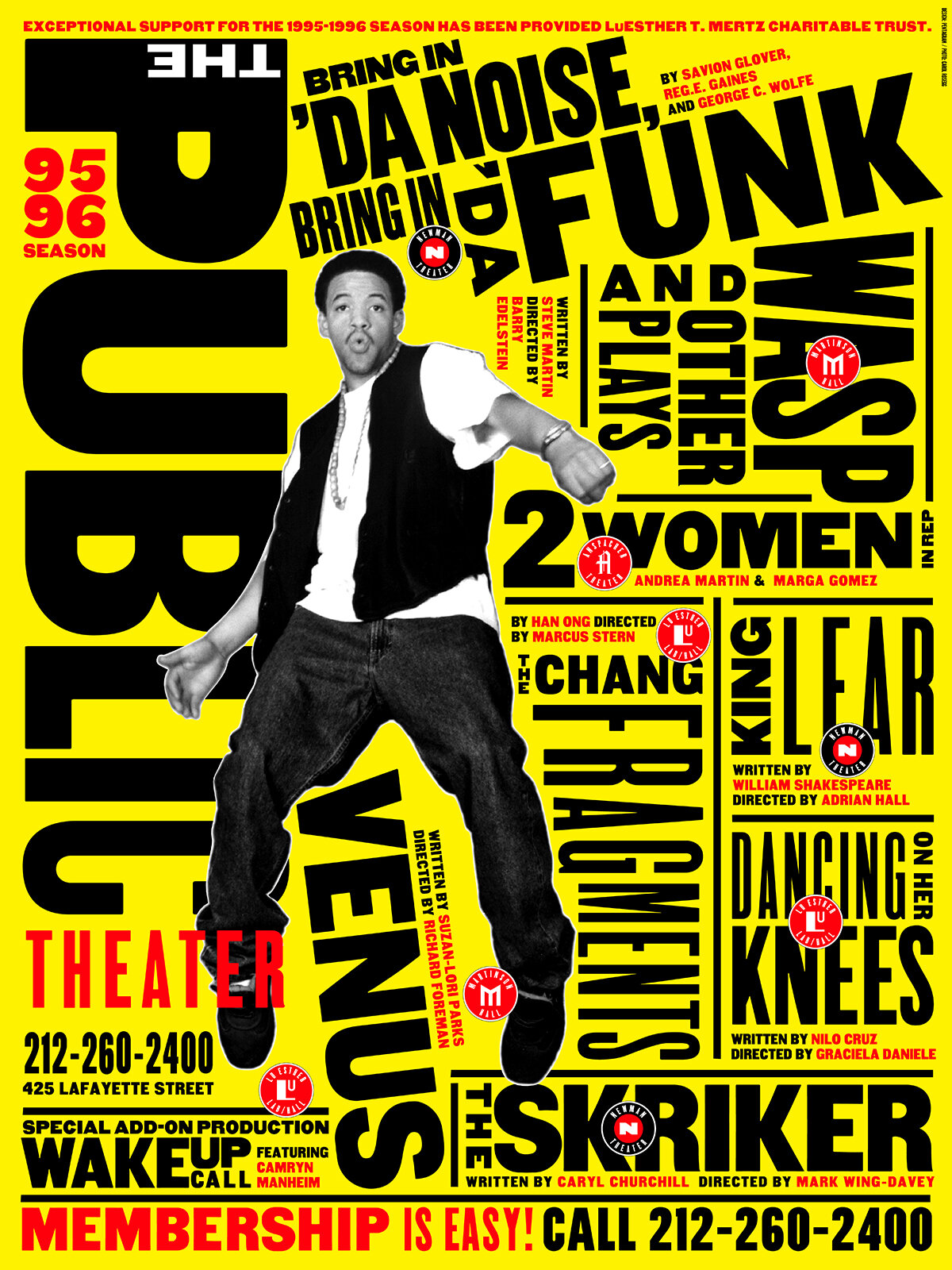}
         \caption{}
     \end{subfigure}
     \hfill
     \begin{subfigure}[b]{0.29\textwidth}
         \centering
         \includegraphics[height=1.2\textwidth]{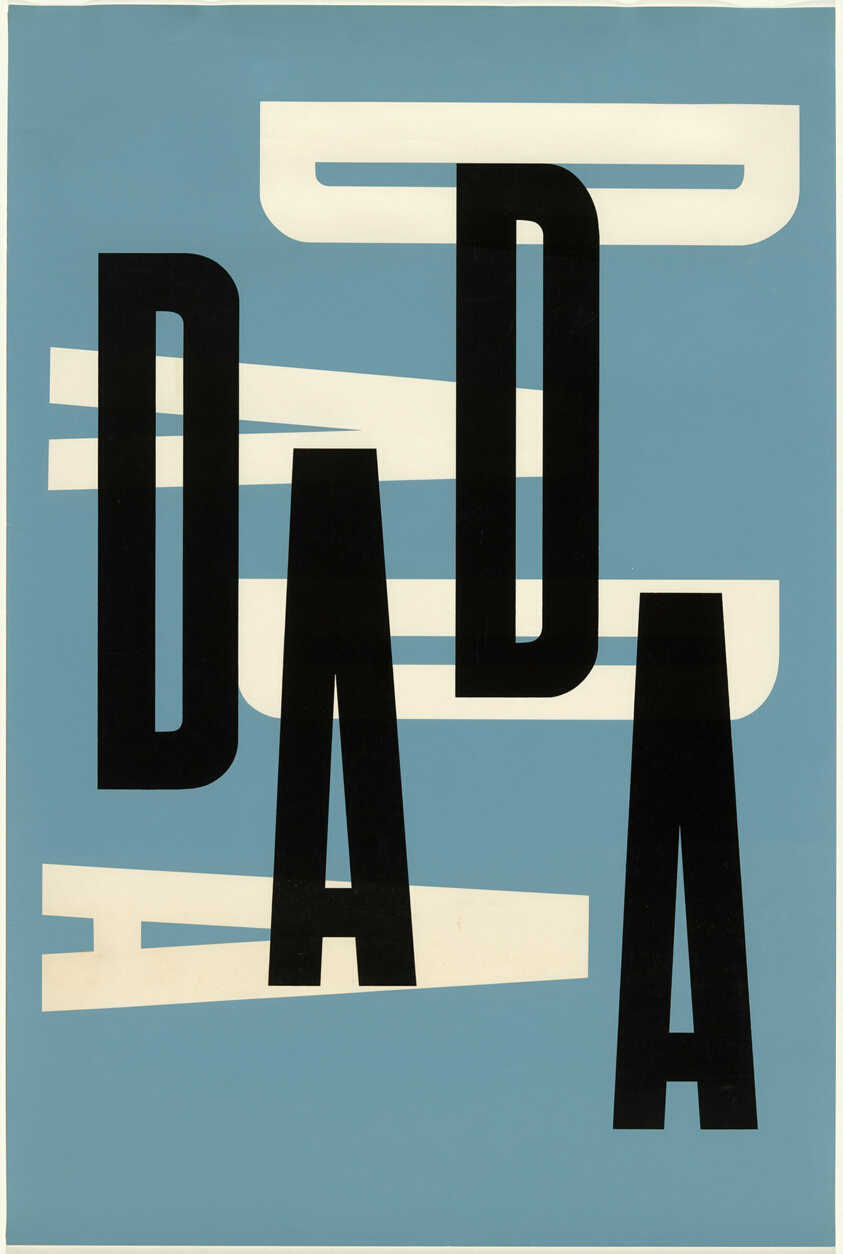}
         \caption{}
     \end{subfigure}
     \hfill
     \begin{subfigure}[b]{0.29\textwidth}
         \centering
         \includegraphics[height=1.3\textwidth]{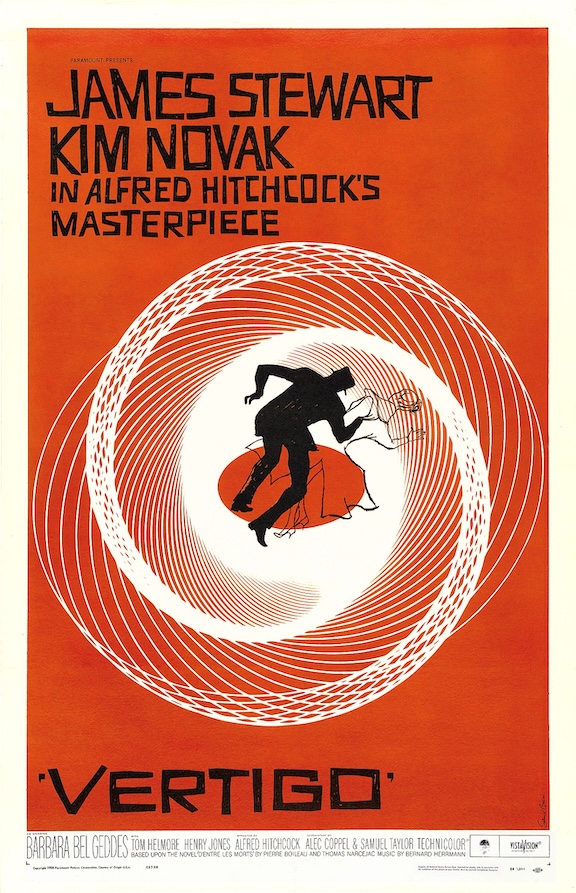}
         \caption{}
     \end{subfigure}
     \hfill
     \caption{\small Examples of iconic visual communication designs that leverage typographical elements in creative and strategic ways to convey messages effectively. (a) The Public Theatre advertisement posters, designed by Scher, 1994. (b) DADA book cover by Paul Rand, 1951. (c) Poster for Alfred Hitchcock’s Vertigo, by Saul Bass, 1958.}
     \label{fig:typography-in-visual-com}
\end{figure}

Typography encompasses the design and arrangement of type, including letters, numbers, and symbols, to create visually appealing and communicative text \cite{typography}.
Typography plays a significant role in visual communication by facilitating clarity, readability, and emotional resonance in design \cite{bringhurst2004elements, VisualDesignSolutions2015}.  
Moreover, typography contributes to establishing and reinforcing brand identity, as consistent font usage across various platforms aids in brand recognition. In \Cref{fig:typography-in-visual-com} we show examples of how typography is used to enhance visual communication by pioneering graphic designers in various ways.

\begin{figure}
    \centering
    \includegraphics[width=0.23\linewidth]{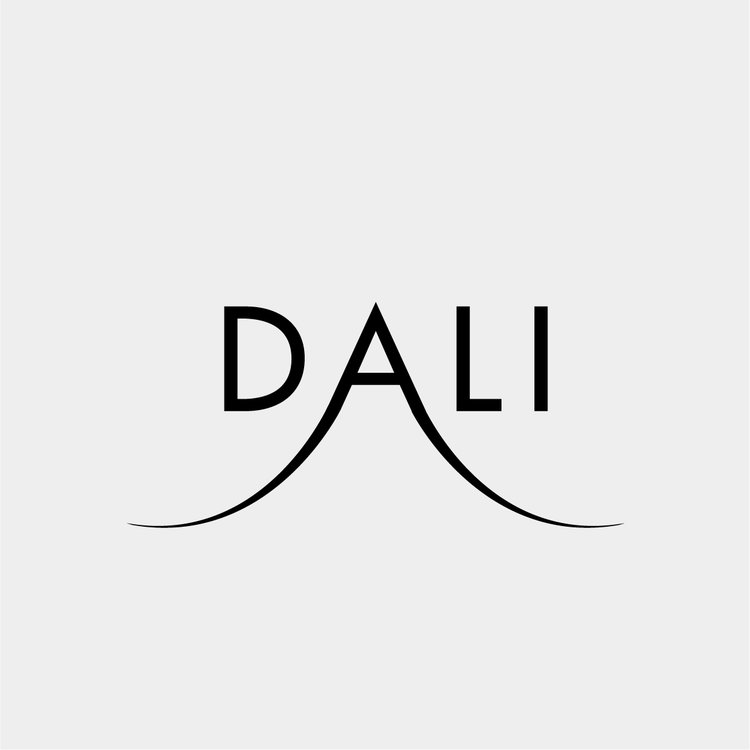}
    \includegraphics[width=0.23\linewidth]{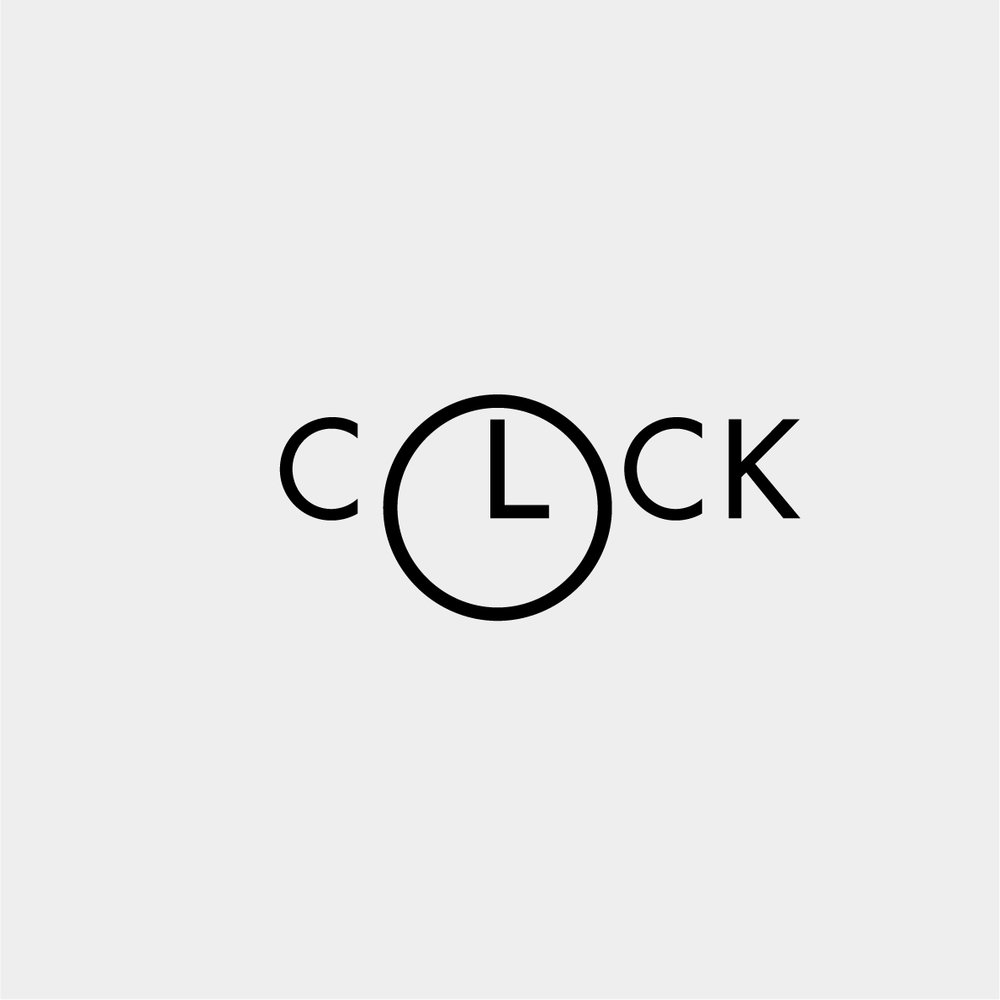}
    \includegraphics[width=0.23\linewidth]{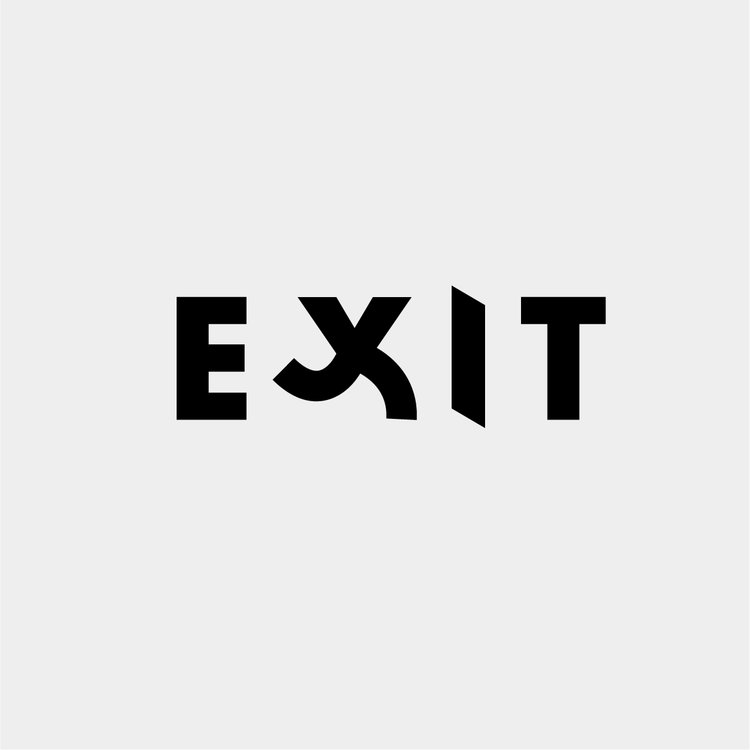}
    \includegraphics[width=0.23\linewidth]{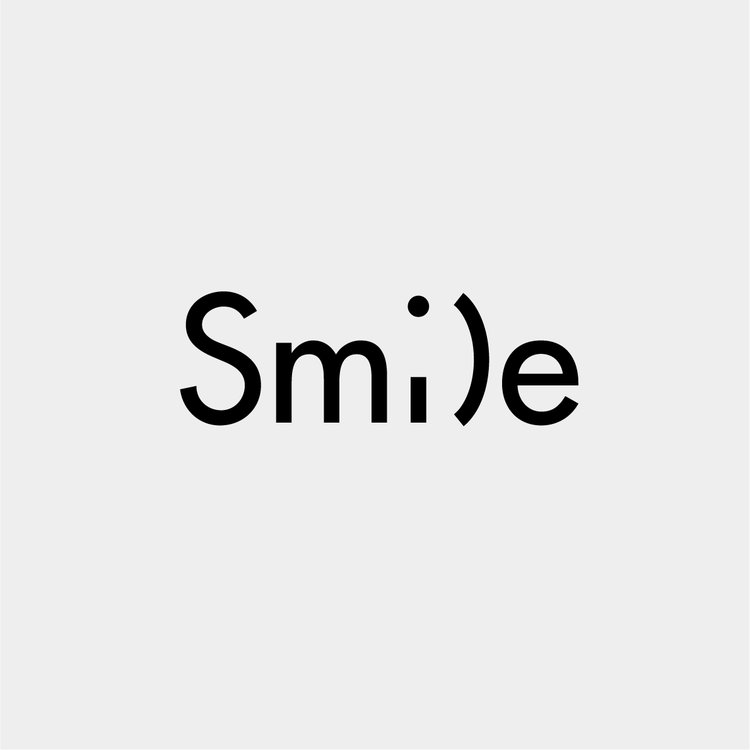}
    \caption{\small Examples of word-as-image illustrations made by the artist Ji Lee \cite{WordAsImage}.}
    \label{fig:word-as-image-real}
\end{figure}

In our research, we investigate how typographical elements can be automatically created with generative models to enhance visual expression. Specifically, in \Cref{chap:word-as-image} we focus on the automatic generation of word-as-image illustrations \cite{WordAsImage}, also known as typographic illustrations or text art.
Word-as-image illustrations are visual representations where the textual elements of a word or phrase are manipulated to create a visual image that embodies the meaning of the word itself. 
In \Cref{fig:word-as-image-real} we show examples of such illustration made by the artist Ji Lee.
Word-as-image illustrations blur the lines between text and image, engaging viewers with their inventive and playful use of typography. They can be found in various contexts, including advertising, graphic design, digital art, and typographic posters.
While traditional typography primarily focuses on legibility and readability, word-as-image typography goes beyond mere text to convey meaning through visual imagery. Therefore, word-as-image can be seen as a specialized and expressive subset of typography, where the emphasis is on the visual impact and artistic interpretation of text.

Designing word-as-image illustrations presents several challenges, primarily due to the need to integrate typography with visual imagery effectively, while maintaining legibility. In addition, conceptualization of the semantic concept requires visual knowledge and the ability to generate innovative ideas that effectively communicate the desired message or theme. 
Overcoming these challenges requires a combination of rich visual knowledge, creativity, and technical skill, to create visually compelling and communicative illustrations.

In \Cref{chap:word-as-image}, we suggest an algorithm for the automatic generation of word-as-image illustrations. 
We rely on the remarkable ability of recent large pretrained vision-language models (specifically we use Stable Diffusion \cite{rombach2022highresolution}) to distill textual concepts visually.
We target simple, concise, black-and-white designs that convey the semantics clearly. 
We use vector representation for the letters, and similar to CLIPasso \cite{vinker2022clipasso} we optimize the control points of each letter directly to depict the desired semantic concept. We use a differentiable rasterizer \cite{diffvg} to transfer the vector representation into pixels, and utilize a pretrained text-to-image diffusion model \cite{rombach2022highresolution} to guide the optimization to fit the desired semantic concept. To extract the training signal we rely on the score-distillation sampling (SDS) loss \cite{jain2022vectorfusion, poole2022dreamfusion}. This loss was designed to leverage pretrained text-to-image diffusion models to optimize non-pixel representations to meet given text-based constraints. We incorporate regularization terms to ensure the legibility of the text and the preservation of the font's style.

Our technique highlights the potential of pretrained VLMs in assisting designers by offering initial solutions in editable formats (vector graphics), facilitating further refinement and development by human users.

\section{Text Guided Generation of Short Animations}

Short animations play a significant role in visual communication by conveying messages, stories, and emotions through visual \emph{dynamic} sequences. They are highly effective for capturing viewers' attention, conveying complex ideas in a concise format, and evoking emotional responses through movement, sound, and visual effects. Short animations are utilized across various platforms, including social media, advertising, education, and entertainment, to engage audiences and communicate messages effectively. 

\begin{figure}[h]
    \centering
    \includegraphics[width=0.2\linewidth]{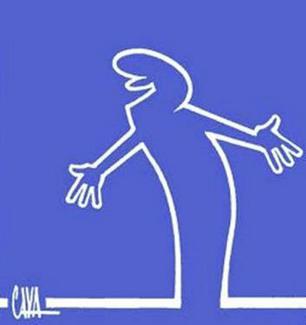}
    \caption{\small "La Linea", a classic Italian animated series created by Osvaldo Cavandoli in the 1960s.}
    \label{fig:La-linea}
\end{figure}

An iconic example of a simple yet expressive short animation is "La Linea", a classic Italian animated series created by Osvaldo Cavandoli in the 1960s (see \cref{fig:La-linea}). It features a character simply known as "The Line," a simple outline character who encounters various obstacles and situations drawn by an unseen animator. The series is characterized by its minimalistic animation style and humorous, silent comedy. "La Linea" became internationally popular and is often cited as an iconic example of visual storytelling through simple animation. The simplicity of its concept and the clever use of the line character make it a memorable and influential series in the world of animation.

In \Cref{chap:videosketch} we investigate the automatic generation of simplified short animations, utilizing the prior of large pretrained text-to-video models. Specifically, we propose a method to automatically animate (breath life into) a static sketch, based on textual input.
Animating sketches is a laborious process, requiring extensive experience and professional design skills.
We present a method that automatically adds motion to a single-subject sketch, based on a textual prompt, without the need for any human annotations or explicit reference motions. 
The output is a short animation provided in vector representation, which can be easily edited.
We do so by leveraging a pretrained text-to-video diffusion model~\cite{videocomposer2023} using a score-distillation loss \cite{jain2022vectorfusion, poole2022dreamfusion} to guide the placement of strokes. 
To promote natural and smooth motion and to better preserve the sketch's appearance, we model the learned motion through two components. The first governs small local deformations and the second controls global affine transformations.
Surprisingly, we find that even models that struggle to generate sketch videos on their own can still serve as a useful backbone for animating abstract representations.

\section{Generative Models for Visual Inspiration}

Following our goal to empower designers and artists, in a parallel research direction, we examine how VLMs can be leveraged to support effective visual communication by providing visual inspiration.
Such inspiration can stimulate creativity by exposing designers to new ideas and techniques, which is essential for solving communication challenges. Furthermore, drawing inspiration from diverse sources helps in creating designs that stand out in a crowded visual landscape.

\begin{figure}[h]
    \centering
    \includegraphics[width=0.7\linewidth]{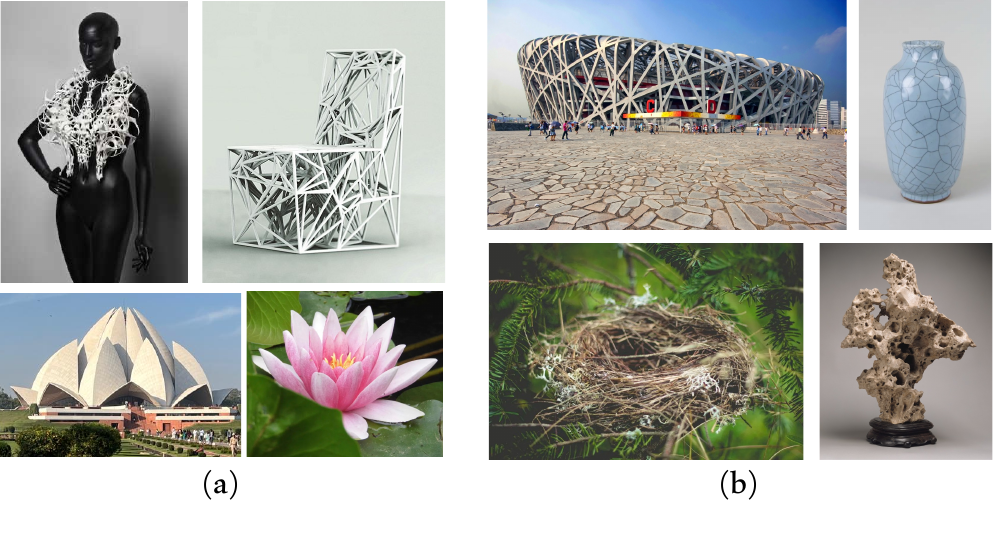}
    \vspace{-0.5cm}
    \caption{\small Examples of design inspired by visual concepts taken from other concepts.
    (a) top left - fashion design by Iris Van Herpen and Chair by Emmanuel Touraine inspired by nature patterns, bottom left - the Lotus Temple in India, inspired by the lotus flower (b) Beijing National Stadium is inspired by a combination of local Chinese art forms - the crackle glazed pottery that is local to Beijing, and the heavily veined Chinese scholar stones.}
    \label{fig:design_inpiration_examples1}
\end{figure}

A creative idea is often born from transforming, combining, and modifying ideas from existing visual examples capturing various concepts \cite{henderson1999line, MULLER198912, ECKERT2000523}.
Large vision-language models posses a huge prior knowledge about visual and semantic concepts, this knowledge has the potential to support designers and provide visual inspiration \cite{Zhang2023GenerativeIA, Smith2023TrashTT, Paananen2023UsingTG}. 
The core idea behind our proposed approach is that rather than simply replicating previous designs, the ability to extract only certain aspects of a given concept is essential to generating original ideas (as demonstrated in \Cref{fig:design_inpiration_examples1}). Additionally, by combining multiple aspects from various concepts, designers are often able to create something new. 

In \Cref{chap:inspiraitontree}, we propose a method to decompose a visual concept, represented as a set of images, into different visual \textit{aspects} encoded in a hierarchical tree structure. We utilize VLMs and their rich latent space for concept decomposition and generation. 
Each node in the tree represents a sub-concept using a learned vector embedding injected into the latent space of a pretrained text-to-image model. We use a set of regularizations to guide the optimization of the embedding vectors encoded in the nodes to follow the hierarchical structure of the tree.
Our method allows to explore and discover new concepts derived from the original one. The tree provides the possibility of endless visual sampling at each node, allowing the user to explore the hidden sub-concepts of the object of interest.
The learned aspects in each node can be combined within and across trees to create new visual ideas, and can be used in natural language sentences to apply such aspects to new designs.

\clearpage
\section{Thesis Overview}
We begin with an overview of fundamental concepts in generative models (\Cref{chap:background}), covering the necessary background information and notations essential for understanding the ideas presented in this dissertation. The relevant publications comprising this dissertation are then organized into chapters divided into three parts:

\begin{itemize}
    \item \Cref{part:one} - Sketch Generation and Abstraction
    \begin{itemize}
        \item \Cref{chap:clipasso} - We present CLIPasso, a method to convert an image of an object into a sketch, allowing for varying levels of abstraction, while preserving the key visual features of the input object. 
        \item \Cref{chap:clipascene} - In a follow-up work, CLIPascene, we extend CLIPasso to scene sketching. We also propose an extended definition to sketch abstraction by disentangling abstraction into two axes of control: \textit{fidelity} and \textit{simplicity}. 
    \end{itemize}
    \item \Cref{part:two} - Communicative Illustrations in Typography
    \begin{itemize}
        \item \Cref{chap:word-as-image} - We suggest an algorithm for the automatic generation of word-as-image illustrations, where a the appearance of a given word is manipulated to present a visualization of its meaning.
    \end{itemize}
    \item \Cref{part:three} - Text Guided Generation of Short Animations
    \begin{itemize}
        \item \Cref{chap:videosketch} - We present a method that automatically adds motion to a single-subject sketch (hence, ``breathing life into it''), merely by providing a text prompt indicating the desired motion. The output is a short animation provided in vector representation, which can be easily edited.
    \end{itemize}
    \item \Cref{part:four} - Generative Models for Visual Inspiration
    \begin{itemize}
        \item \Cref{chap:inspiraitontree} - We leverage the prior knowledge of pretrained text-to-image model to build a tree-structured visual exploration space based on a given visual concept. The nodes of the tree encode different aspects of the subject. Through examining combinations within and across trees, the different aspects can inspire the creation of new designs and concepts.
    \end{itemize}
\end{itemize}
  
\section{List of Publications}

The following publications are discussed in this dissertation:

\begin{itemize}
    \item Yael Vinker, Ehsan Pajouheshgar, Jessica Y. Bo, Roman C. Bachmann, Amit Bermano, Daniel Cohen-Or, Amir Zamir, Ariel Shamir. \ap{CLIPasso: Semantically-Aware Object Sketching}. ACM Transactions on Graphics 41, no. 4 (SIGGRAPH 2022). \textbf{Best Paper Award}.
    
    \item Yael Vinker, Yuval Alaluf, Daniel Cohen-Or, Ariel Shamir. \ap{CLIPascene: Scene Sketching with Different Types and Levels of Abstraction}. IEEE International Conference on Computer Vision (ICCV 2023). Oral Presentation.
    
    \item Shir Iluz*, Yael Vinker*, Amir Hertz, Daniel Berio, Daniel Cohen-Or, Ariel Shamir. \ap{Word-as-Image for Semantic Typography}. ACM Transactions on Graphics 42, no. 4 (SIGGRAPH 2023). \textbf{Honorable Mention Award}.
    
    \item Yael Vinker, Andrey Voynov, Daniel Cohen-Or, Ariel Shamir. \ap{Concept Decomposition for Visual Exploration and Inspiration}. ACM Transactions on Graphics 42, no. 6 (SIGGRAPH Asia 2023). \textbf{Best Paper Award}.
    
    \item Rinon Gal*, Yael Vinker*, Yuval Alaluf, Amit Bermano, Daniel Cohen-Or, Ariel Shamir, Gal Chechik. \ap{Breathing Life Into Sketches Using Text-to-Video Priors}. IEEE/CVF Conference on Computer Vision and Pattern Recognition (CVPR 2024). Poster Highlight.
\end{itemize}

\chapter{Background}
\label{chap:background}

The works presented in this dissertation are all based on leveraging the priors of pretrained large vision-language models to facilitate design-related generative tasks. Some of the works also employ a vector representation for images.
This chapter provides a review of fundamental concepts in vector graphics, generative models, and vision-language models. The notations used in subsequent chapters are introduced below.

\section{Vector Representation for Images}
\begin{figure}[h!]
    \centering
    \includegraphics[width=0.5\linewidth]{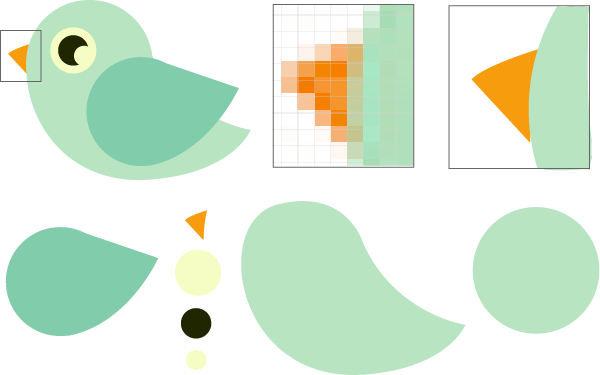}
    \caption{\small Illustration of a vector image. The bird is composed of a set of shapes (shown at the bottom). Additionally, we illustrate the difference between a raster representation on a pixel grid, and a vector representation which is resolution free.}
    \label{fig:vector_im}
\end{figure}

Vector graphics allow us to create visual images directly from geometric shapes such as points, lines, curves, and polygons. The vector image can therefore be represented by a set of parameters defining the geometric shapes. Unlike raster images (represented with pixels), vector representation is resolution-free, more compact, and easier to modify. An illustration is presented in \Cref{fig:vector_im}.
This quality makes vector images the preferred choice for various design applications, such as logo design, prints, animation, CAD, typography, web design, infographics, and illustrations.

Scalable Vector Graphics (SVG) stands out as a popular vector image format due to its excellent support for interactivity and animation. 
SVG is a XML-based format for describing two-dimensional vector graphics. An SVG is a plain text file that describes geometric shapes and paths using XML markup. The structure of an SVG document consists of elements such as \verb|Rect|, \verb|Circle|, and \verb|Path| each defining different graphical elements and their attributes.
For example, the Rect command creates a rectangular shape defined by the starting point, width, and height arguments, like \verb|<Rect x="40" y="100" width="50" height="50"/>|. 
SVG data presents many challenges due to its highly irregular nature. Each sample can be composed of any number of shapes, such as ellipses, polygons, circles, splines, etc., and each shape can be defined by a varying number of parameters. For example, a polygon can be constructed from any number of 2D points, making it difficult to process and analyze. Additionally, two visually identical shapes can be defined in completely different ways. For instance, a rectangle can be defined by four parameters (starting point, width, and height) or with four strait lines, or with a polygon shape.
Thus, in the works described in this dissertation that utilize vector image representation, cubic \Bezier curves are used. 
A cubic \Bezier curve is a smooth parametric curve which is formulated by a set of $4$ ordered control points $\{p_0, p_1, p_2, p_3\}$, and is defined as follows: 
\begin{equation}
    B(t) = (1 - t)^3p_0 + 3(1-t)^2tp_1 + 3(1-t)t^2p_2 + t^3p_3, t \in [0,1]
    \label{eq:cubic_bezier}
\end{equation}

This definition is derived recursively from \Bezier curves of lower degree, where an order 1 \Bezier curve is defined by 2 control points $p_0, p_1$ and is simply a linear curve:
\begin{equation}
    B(t) = (1-t)p_0+t p_1, t\in [0,1]
\end{equation}

In a similar way, a quadratic \Bezier curve is defined by three control points $p_0, p_1, p_2$ with a linear interpolation of corresponding points on the linear \Bezier curves from $p_0$ to $p_1$ and from $p_1$ to $p_2$:
\begin{equation}
    B(t) = (1-t)[(1-t)p_0 + t p_1] + t[(1-t)p_1 + t p_2], t\in [0,1]
\end{equation}
This can be rearranged into:
\begin{equation}
    B(t) = (1-t)^2 p_0 + 2(1-t) t p_1 + t^2 p_2, t\in [0,1]
\end{equation}

\begin{figure}[h!]
    \centering\includegraphics[width=0.8\linewidth]{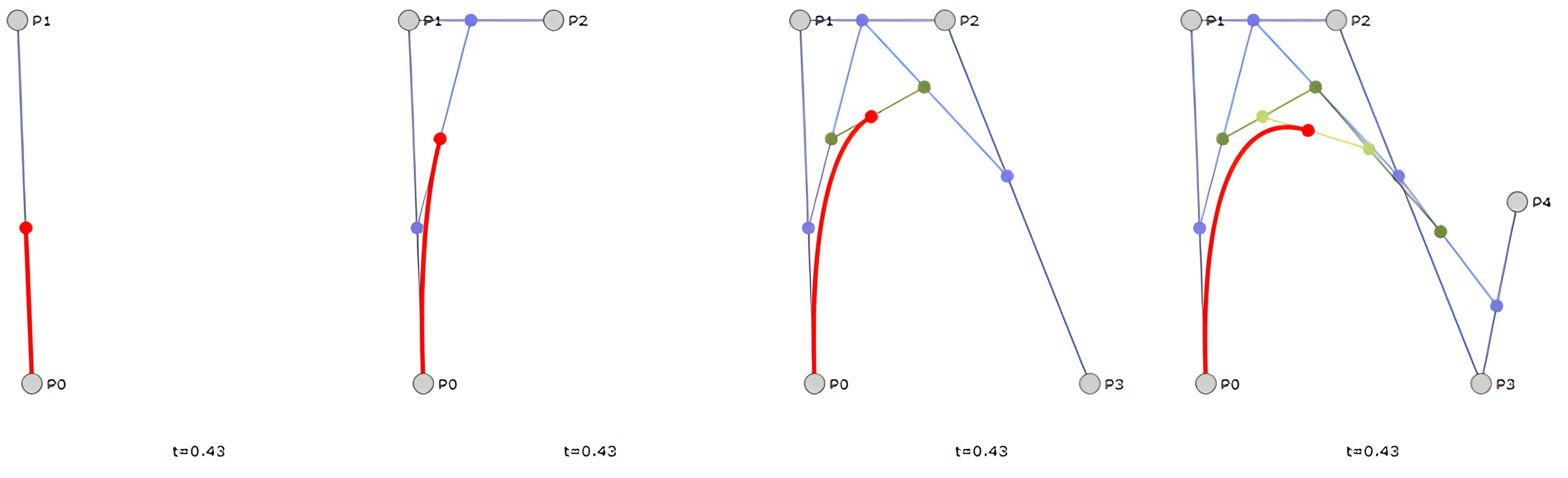}
    \caption{\small Illustration of \Bezier curves of degrees 1 to 4 (left to right) and how they are constructed recursively based on $t\in [0,1]$.}
    \label{fig:bezier_def}
\end{figure}

In a similar way, a cubic \Bezier curve can be defined recursively to receive the definition shown in \cref{eq:cubic_bezier}. This concept is illustrated in \Cref{fig:bezier_def}.

The choice of cubic \Bezier curves is common in the field of generative models for vector graphics \cite{iconshop, deepsvg, wangDeepVecFontSynthesizingHighquality2021} as many other complex shapes such as rectangle, circle, ellipse, and more can be approximated by combinations of cubic \Bezier curves with negligible visual differences. For example, we can concatenate four \Bezier curves to form a circle.

The recent development of differentiable rendering algorithms \cite{Zheng2019StrokeNetAN, Mihai2021DifferentiableDA, diffvg} makes it possible to manipulate or synthesize vector content by using raster-based loss functions. In all our works, we utilize the method of Li et al. \cite{diffvg}, as it can handle a wide range of curves and strokes, including Bézier curves.
In the context of sketch generation and non-photorealistic rendering, there is a substantial literature summarized in the surveys by Hertzmann \cite{Hertzmann2003-survey}, and by Bénard and Hertzmann \cite{Bnard2019LineDF}.
In addition, vector representations are widely used for a variety of sketching tasks and applications, employing a number of deep learning models including RNN \cite{SketchRNN}, BERT \cite{Lin2020SketchBERTLS}, Transformers \cite{Bhunia2020PixelorAC, Ribeiro2020SketchformerTR}, CNNs \cite{Chen2017Sketchpix2seqAM} GANs \cite{Varshaneya2021TeachingGT} and reinforcement learning algorithms \cite{Zhou2018LearningTS, spiralpp, Ganin2018SynthesizingPF}.

\section{Generative Models}
Generative models are a class of computational data-driven models aiming at automatically generating new content, based on a given dataset. In the field of computer vision and graphics, we focus on generative models that learn how to generate visual content, such as images, 3D data, videos, and more.

\begin{figure}[h]
    \centering
    \includegraphics[width=0.7\linewidth]{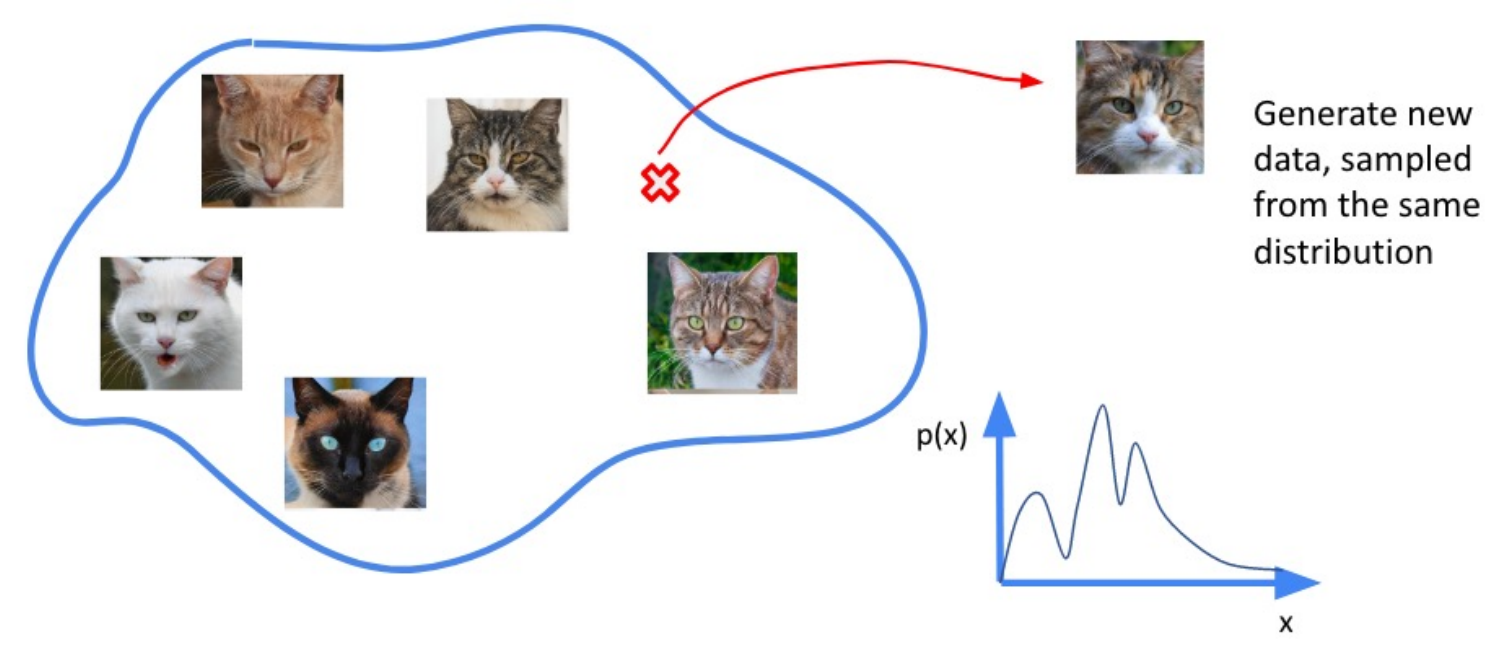}
    \caption{\small An example of a dataset composed of images of cats. This group of images can be thought of as discrete samples from a distribution over the pixel space that defines all possible cat images. A generative model is tasked with generating new images that belong to this distribution with a high probability.}
    \label{fig:cat-sampling}
\end{figure}

Generally, the idea behind generative models is to learn a function that estimates the underlying distribution of a given dataset, and by that it could generate a new sample from this distribution with high probability.
For example, let's assume we have a large dataset of images of cats. This data defines a distribution over the space of pixels, where sampling an image of a cat means that the pixel's value in that image will be structured in a specific way so that the image they compose would look like a cat (see \Cref{fig:cat-sampling}).

Formally, let us denote the distribution $\mathcal{X}$ over a space $\mathbb{R}^n$ (for example, $\mathcal{X}$ could be the distribution of cat images and $\mathbb{R}^n$ is the space of all possible images with $n$ pixels). We are given a finite set of independent and identically distributed (i.i.d) samples $x_1,...x_n \sim \mathcal{X}$ that we refer to as the training data. Our goal is to be able to generate new samples $\hat{x} \sim \hat{\mathcal{X}}$ where $\hat{\mathcal{X}}$ is an approximation of $\mathcal{X}$.
Note that the real distribution $\mathcal{X}$ of the data is intractable, mathematically we will need an infinite set of images to define the data distribution in a complete way.
To generate the samples $\hat{x}$ we will learn a function $g:\mathbb{R}^q\rightarrow \mathbb{R}^n$, that maps samples $z \sim \mathcal{Z}$ such that $\mathcal{Z}$ is a tractable distribution supported in $\mathbb{R}^q$, to points in $\mathbb{R}^n$ that match the given data.
Thus, we assume that for each sample $x\sim \mathcal{X}$ there exists a point $z \sim \mathcal{Z}$ such that $g(z) = \hat{x} \approx x$.
$\mathcal{Z}$ is usually a known distribution such as a Normal distribution. The function $g$ is often referred to as the \ap{generator}, and the space $\mathbb{R}^q$ is often referred to as the \ap{latent space}.
With the generator $g$ we can generate new samples by sampling $z$ from our known distribution and compute $g(z)$.

Image generation is a challenging task, since there are a multitude of ways to assign values to individual pixels, and relatively few of those combinations constitute an image of the desired concept.
Therefore in generative models, we learn the parameters of a model (that can be thought of as a huge complex function) and this model estimates the data distribution based on the given samples.
Many different types of generative models exist, including VAEs \cite{VAE2014}, GANs \cite{goodfellow2016deep}, normalizing flows \cite{pmlr-v37-rezende15}, auto-regressive \cite{NIPS2016_b1301141}, diffusion \cite{ddpm}, and others.
For a thorough survey on the evolution of generative models and their different formulations we refer the reader to \cite{RuthottoGenerative,TaylorGenerative,GM2020100285}.

\section{CLIP (Contrastive Language-Image Pretraining)}
CLIP \cite{Radfordclip} is a large vision-language model which is based on a neural network trained on image-text pairs with the objective of creating a joint latent space of text and images.
CLIP was trained on a large data of 400 million image-text pairs collected from the internet \cite{schuhmann2021laion400m}, and was one of the first large-scale open-source models, therefore significantly advancing scientific research.

\begin{figure}
    \centering
    \includegraphics[width=0.6\linewidth]{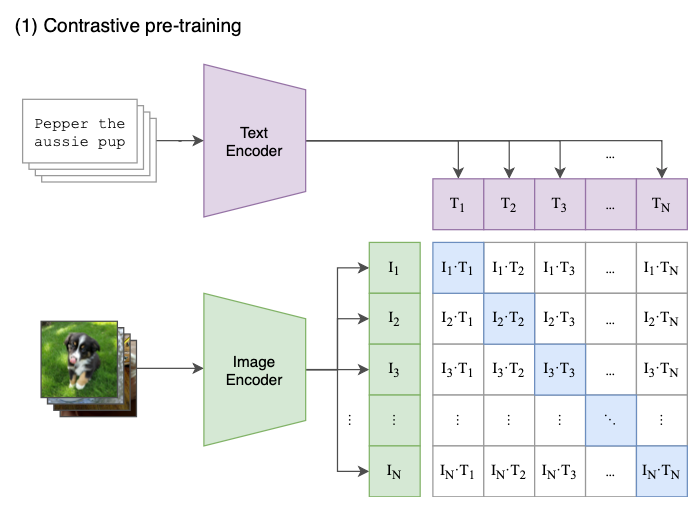}
    \caption{\small CLIP jointly trains an image encoder and a text encoder to predict the correct pairings of a batch of (image, text) training examples. The illustration was taken from the CLIP paper \cite{Radfordclip}.}
    \label{fig:clip_train}
\end{figure}

In CLIP, two encoders are being trained -- a text encoder $E_T$ and an image encoder $E_I$. During training, at each step, a large batch of $N$ paired images and text $\{x_I, x_T\}$ are sampled from the training dataset, where the paired images and text form the positive pair and the unpaired ones are considered as negative samples.
The training loss is based on a contrastive approach, and can be formulated with:

\begin{equation}
    L_{clip} = -\Sigma_{i=1}^N \log \frac{exp(sim(E_I(x_I^i), E_T(x_T^i))/\tau)}{\Sigma_{k=1}^N exp(sim(E_I(x_I^i),E_T(x_T^k))/\tau)},
\end{equation}

where ${x_I^i, x_T^i}$ are the i'th image-text pair in a batch, and $sim(\dot,\dot)$ is the cosine similarity (a dot product between the input vectors). The temperture $\tau$ is a learned parameter. 
This loss simply means that in this shared latent space, the representation of positive text-image pairs should be close to each other (under the cosine distance metric), while negative pairs should be far away from each other. This procedure is illustrated in \Cref{fig:clip_train} (the illustration was taken from the CLIP paper \cite{Radfordclip}).

Being trained on a wide variety of image domains along with lingual concepts, CLIP models are found to be very useful for a wide range of zero-shot tasks.
It was demonstrated by the authors of CLIP that the pretrained model can be used for image classification, and they show that the model is robust to distribution shift. As an example, CLIP achieved high classification rates for sketches and cartoons despite not having been trained with this objective. 

Due to these impressive capabilities, CLIP was later used in many zero-shot tasks. In our context, CLIP was also used for more creative generative tasks as described next.

The first work to leverage CLIP's prior with vector data was CLIPDraw \cite{CLIPDraw}. CLIPDraw optimizes a set of random \Bezier curves to create a drawing that maximizes the CLIP similarity for a given text prompt. 
Later, Tian and Ha \cite{Evolution-Strategies-for-Creativity} employ evolutionary algorithms combined with CLIP, to produce creative abstract concepts represented by colored triangles guided by text or shape. Their results are limited to either fully semantic (using CLIP's text encoder) or entirely geometric (using the L2 loss).

\section{Diffusion Models}
\begin{figure}[h!]
    \centering
    \vspace{-0.2cm}
    \includegraphics[width=0.9\linewidth]{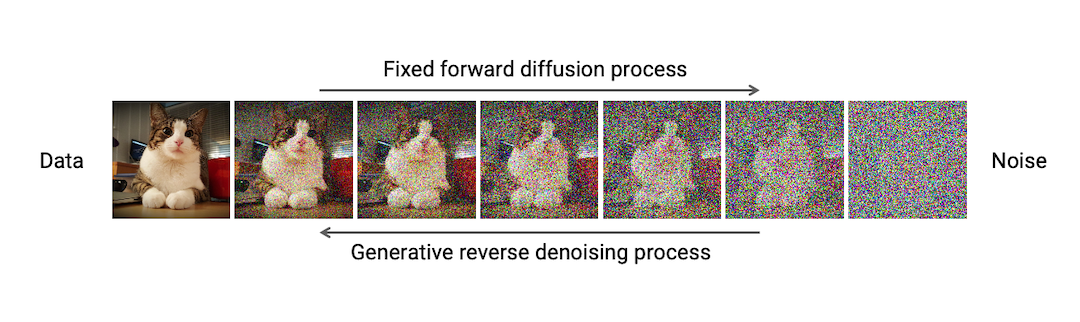}
    \vspace{-0.3cm}
    \caption{\small A visualization of forward diffusion process and reverse denoising process of an image.}
    \label{fig:cat_fusion}
\end{figure}

Diffusion models are generative models trained to learn a data distribution $p(x)$ by gradually denoising a variable sampled from a Gaussian distribution. 
Denoising diffusion models consist of two processes: Forward diffusion process that gradually adds noise to the input, and reverse denoising process that learns to generate data by denoising.
A visualization of such a forward and backward operation in the context of image generation is shown in \Cref{fig:cat_fusion}.

The denoising process corresponds to learning
the reverse process of a fixed Markov Chain of length T.
For image synthesis, the most successful models rely on a reweighted variant of the variational lower bound on $p(x)$, which mirrors denoising score-matching [85].
Most image synthesis approaches \cite{NEURIPS2021_49ad23d1,ddpm, Refinement23} rely on a reweighted version of the variational lower bound on $p(x)$, which is similar to denoising scoring-matching \cite{song2021scorebased}.
In general, the idea is to train a neural network that is tasked with cleaning a noised signal $x_t$ in small steps. Then this network can be used for generation by gradually removing noise from a given input as shown in \Cref{fig:cat_fusion}.
Specifically this network is often defined with $\epsilon_{\theta}(x_t,t)$, where $t$ is a time step, and $x_t$ is the noised image at time step $t$, defined as:
\begin{equation}
    x_t = \alpha_t x + \sigma_t \epsilon,
    \label{eq:xt-sample}
\end{equation}
where $\alpha_t$, $\sigma_t$ are parameters dependant on the noising schedule of the pretrained diffusion model, and $\epsilon \in \mathbb{N}\left(0, 1\right)$ is a noise sample. 
During training, at each iteration, an image $x_0 \sim p(x_0)$, a timestep $t \sim Uniform(\{1,..,T\})$, and a noise $\epsilon \sim \mathcal{N}(0,I)$ are sampled to define $x_t$, and the network $\epsilon_{\theta}$ is trained with the objective of predicting the noise $\epsilon$:
\begin{equation}
    \mathcal{L}_{DM} = 
    \mathbb{E}_{x_0,\epsilon\sim\mathcal{N}(0,1),t} \left [ || \epsilon - \epsilon_\theta(x_t, t) ||_2^2 \right ]
    \label{eq:diffusion_loss}
\end{equation}

Once the network $\epsilon_{\theta}$ is trained, it can be used at inference time to generate a new image by gradually denoising a sample $x_T \sim \mathcal{N}(0,I)$ for $T$ steps.

\paragraph{Latent Diffusion Models}
Most academic computer vision researchers use the Stable Diffusion model \cite{rombach2022highresolution} (as of the time of writing this dissertation), since it is relatively lightweight and was the first large-scale text-to-image diffusion model made available publicly. 
Stable Diffusion is a type of a latent diffusion model (LDM), where the diffusion process is applied on the latent space of a pretrained image autoencoder. The encoder $\mathcal{E}$ maps an input image $x$ into a latent vector $z$, and the decoder $\mathcal{D}$ is trained to decode $z$ such that $\mathcal{D}(z)\approx x$.
As a second stage, a denoising diffusion probabilistic model (DDPM) \cite{ddpm} is trained to generate codes within the learned latent space.
At each step during training, a scalar $t\in\{1,2,...T\}$ is uniformly sampled and used to define a noised latent code $z_t = \alpha_t z + \sigma_t \epsilon$, where $\epsilon \sim \mathcal{N}(0, I)$ and $\alpha_t, \sigma_t$ are terms that control the noise schedule, and are functions of the diffusion process time $t$.
The denoising network $\epsilon_{\theta}$ which is based on a UNet architecture \cite{Unet}, receives as input the noised code $z_t$, the timestep $t$, and an optional condition vector $c(y)$, and is tasked with predicting the added noise $\epsilon$.
The LDM loss is defined by:
\begin{equation}
    \mathcal{L}_{LDM} = 
    \mathbb{E}_{z\sim\mathcal{E}(x),y,\epsilon\sim\mathcal{N}(0,1),t} \left [ || \epsilon - \epsilon_\theta(z_t, t, c(y)) ||_2^2 \right ]
    \label{eq:ldm_loss}
\end{equation}
For text-to-image generation the condition $y$ is a text input and $c(y)$ represents the text embedding.
At inference time, a random latent code $z_T \sim \mathcal{N}(0, I)$ is sampled, and iteratively denoised by the trained $\epsilon_{\theta}$ until producing a clean $z_0$ latent code, which is passed through the decoder $D$ to produce the image $x$. 

The text conditioning is often performed through the cross-attention layers of the network, following the attention mechanism \cite{Attention2017}.
Specifically, in each layer, the deep spatial features $x$ are projected to a query matrix $Q = l_Q(x)$, and the textual embedding is projected to a key matrix $K = l_K(c)$ and a value matrix $V = l_V(c)$ via learned linear projections $l_Q, l_K, l_V$.
The attention maps are then defined by:
\begin{equation}
    A_t = Softmax(\frac{QK^T}{\sqrt{d}}),
\end{equation}
where $d$ is the latent projection dimension of the keys and queries.
For more details about diffusion models we refer the reader to the comprehensive survey by Yang et al. \cite{Diffusion-survey}, and the survey by Cao et al. \cite{Cao2022ASO}.

\section{Score Distillation Sampling}
One of our goals in this work was to utilize the strong prior of pretrained large text-image models for the generation of modalities beyond rasterized images.
In Stable Diffusion, text conditioning is performed via the cross-attention layers defined at different resolutions in the UNet network.
Thus, it is not trivial to guide an optimization process using the conditioned diffusion model. 

The score-distillation sampling (SDS) loss, first proposed in Poole \etal~\cite{poole2022dreamfusion}, serves as a means for extracting a signal from a pretrained text-to-image diffusion model.
In their seminal work, Poole~\etal proposed a way to use the diffusion loss to optimize the parameters of a NeRF~\cite{mildenhall2021nerf} for text-to-3D generation.
At each iteration, the radiance field is rendered from a random angle, forming the image $x$, which is then noised to some intermediate diffusion time step $t$: $x_t = \alpha_t x + \sigma_t \epsilon$ (as described in \cref{eq:xt-sample}).

The noised image is then passed through the diffusion model, conditioned on a text-prompt $c$ describing some desired scene. The diffusion model's output, $\epsilon_\theta(x_t,t,c)$, the prediction of the noise added to the image. The deviation of this prediction from the true noise, $\epsilon$, can serve as a measure of the difference between the input image and one that better matches the prompt. This measure can then be used to approximate the gradients to the initial image synthesis model's parameters, $\phi$, that would better align its outputs with the prompt. Specifically, 
\begin{equation}\label{eq:sds_loss}
    \nabla_\phi \mathcal{L}_{SDS} = \left[ w(t)(\epsilon_\theta(x_t,t,y) - \epsilon) \frac{\partial x}{\partial \phi} \right] ,
\end{equation}
where $w(t)$ is a constant that depends on $\alpha_t$. This optimization process is repeated, with the parametric model converging toward outputs that match the conditioning prompt.

Note that the gradients of the UNet are skipped, and the gradients to modify the Nerf's parameters are derived directly from the LDM loss.

Later, VectorFusion \cite{jain2022vectorfusion} utilized the SDS loss for the task of text-to-SVG generation. The proposed generation pipeline involves two stages. Given a text prompt, first, an image is generated using Stable Diffusion (with an added suffix to the prompt), and is then vectorized automatically using LIVE \cite{live}.
This defines an initial set of parameters to be optimized in the second stage using the SDS loss.
At each iteration, a differentiable rasterizer \cite{diffvg} is used to produce a $600 \times 600$ image, which is then augmented as suggested in CLIPDraw \cite{CLIPDraw} to get a $512 \times 512$ image $x_{aug}$. 
Then $x_{aug}$ is fed into the pretrained encoder $\mathcal{E}$ of Stable Diffusion to produce the corresponding latent code $z=\mathcal{E}(x_{aug})$.
The SDS loss is then applied in this latent space, in a similar way to the one defined in DreamFusion:
\begin{equation}
    \nabla_\theta \mathcal{L}_\text{LSDS} = 
    \quad\mathbb{E}_{t, \epsilon} \left[ w(t) \Big(\hat{\epsilon}_\phi(\alpha_t z_t + \sigma_t \epsilon, y)  - \epsilon\Big) \frac{\partial z}{\partial z_{aug}} \frac{\partial x_{aug}}{\partial \theta} \right]   
\end{equation}

In \Cref{chap:word-as-image} we employ this loss for the generation of word-as-image illustrations.

\section{Textual Inversion}
Textual inversion \cite{gal2022textual} is a method for performing personalization of text-to-image generation results. In such method, a pretrained text-to-image model is adapted such that it can generate images of novel, user-provided concepts.
These concepts are typically unseen during training, and may represent specific personal objects (such as the user's pet) or more abstract categories (new artistic style).
There are multiple approaches for performing text-to-image personalization. Existing methods employ either token optimization techniques \cite{gal2022textual, Prospect23, Voynov2023PET}, fine-tuning the model's weights \cite{ruiz2023dreambooth,hu2021lora,kumari2022customdiffusion}, or a combination of both \cite{NeTIf23,arar2024Palp, avrahami2023bas}. 
We focus in this part on the details of Textual Inversion \cite{gal2022textual} as we rely on its settings in \Cref{chap:inspiraitontree}.

Given a text prompt $y$, for example ``A photo of a cat'', the sentence is first converted into tokens, which are indexed into a pre-defined dictionary of vector embeddings.
The dictionary is a lookup table that connects each token to a unique embedding vector.
After retrieving the vectors for a given sentence from the table, they are passed to a text transformer, which processes the connections between the individual words in the sentence and outputs $c(y)$. 
The output encoding $c(y)$ is then used as a condition to the UNet in the denoising process: $\epsilon_\theta(z_t, t, c(y))$.
Let us denote words with $S$, and the vector embeddings from the lookup table with $V$.

In \cite{gal2022textual} the embedding space of $V$ is chosen as the target for inversion. They formulate the task of inversion as fitting a new word $s^*$ to represent a personal concept, depicted by a small set of input images provided by the user.
They extend the predefined lookup table with a new embedding vector $v_*$ that is linked to $s^*$. 
The vector $v_*$ is often initialized with the embedding of an existing word from the dictionary that has some relation to the given concept, and then optimized to represent the desired personal concept. 
This process can be thought of as ``injecting'' the new concept into the vocabulary.
The vector $v_*$ is optimized w.r.t. the LDM loss in \Cref{eq:ldm_loss} over images sampled from the input set.
At each step of optimization, a random image $x$ is sampled from the set, along with a neutral context text $y$, derived from the CLIP ImageNet templates~\cite{Radfordclip} (such as \ap{A photo of $s^*$}).
Then, the image $x$ is encoded to $z=\mathcal{E}(x)$ and noised w.r.t.\ a randomly sampled timestep $t$ and noise $\epsilon$: $z_t = \alpha_t z + \sigma_t\epsilon$.
The noisy latent image $z_t$, timestep $t$, and text embedding $c(y)$ are then fed into a pretrained UNet model which is trained to predict the noise $\epsilon$ applied w.r.t.\ the conditioned text and timestep.
This way, $v_*$ is optimized to describe the object depicted in the small training set of images.

\part{Generative Models for Abstract Sketches}
\label{part:one}
\chapter{CLIPasso: Semantically-Aware Object Sketching}~\label{chap:clipasso}

\begin{figure*}[h!]
\centering
  \includegraphics[width=1\textwidth]{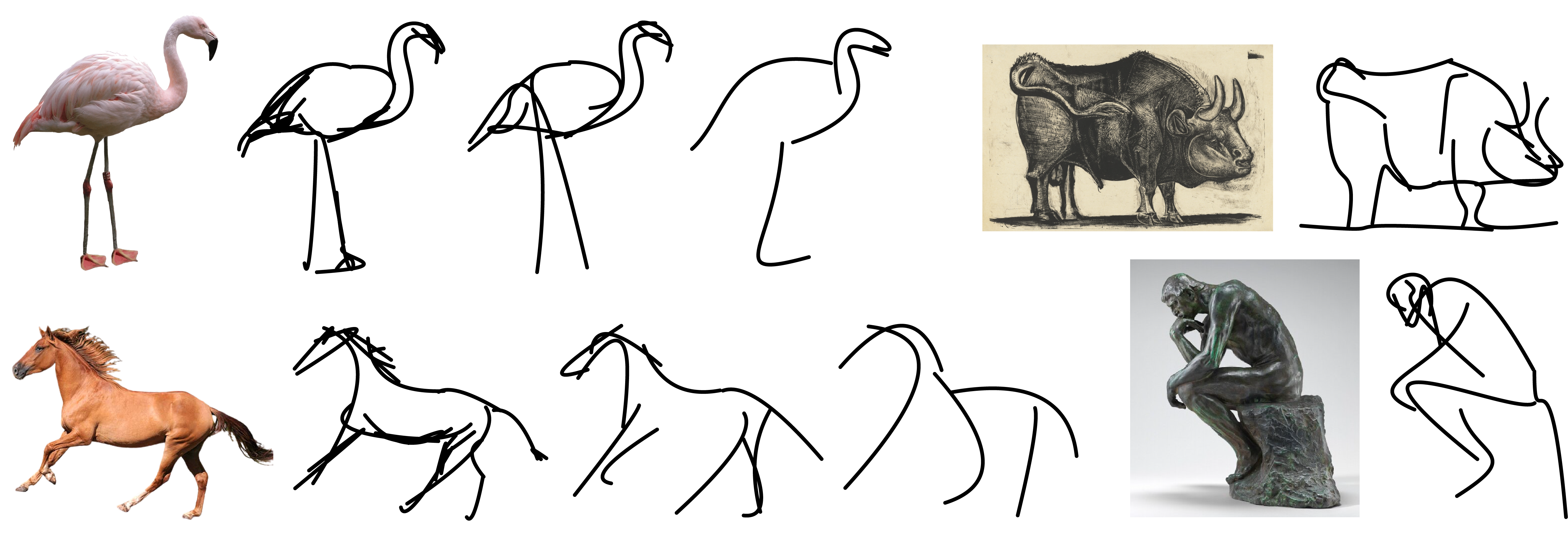}
  \caption[]{\small Our work converts an image of an object to a sketch, allowing for varying levels of abstraction, while preserving its key visual features. Even with a very minimal representation (the rightmost flamingo and horse are drawn with only a few strokes), one can recognize both the semantics and the structure of the subject depicted.\footnotemark}
  \label{fig:teaser_clipasso}
\end{figure*}

\footnotetext{Project page: \url{https://clipasso.github.io/clipasso/}}

Free-hand sketching is a valuable visual tool for expressing ideas, concepts, and actions \cite{Fan2018CommonOR, Hertzmann2020WhyDL, xu2020deep, Gryaditskaya2019OpenSketchAR, Tversky2002WhatDS}. As sketches consist of only strokes, and often only a limited number of strokes, the process of \emph{abstraction} is central to sketching. 
An artist must make representational decisions to choose key visual features of the subject drawn to capture the relevant information she wishes to express, while omitting (many) others~\cite{Chamberlain2016TheGO, Fan2019PragmaticIA, Yang2021VisualCO}.

For example, in the famous "Le Taureau" series (Figure \ref{fig:picasso1}), Picasso depicts the progressive abstraction of a bull. In this series of lithographs, the artist transforms a bull from a concrete, fully rendered, anatomical drawing, into a sketch composition of a few lines that still manages to capture the essence of the bull.

In this paper, we pose the question: Can computers imitate such a process of sketching abstraction, converting an image from a concrete depiction to an abstract one?

Today, machines can render realistic sketches simply by applying mathematical and geometric operations to an input photograph \cite{Winnemller2012XDoGAE, canny1986computational}. However, creating abstractions is more difficult for machines to achieve. The abstraction process suggests that the artist selects visual features that capture the underlying structure and semantic meaning of the object or scene, to produce a minimal, yet descriptive rendering. 
This demands semantic understanding of the subject, which is more complex than applying simple geometric operations to the image. To fill this semantic gap, we use CLIP \cite{Radfordclip}, a neural network trained on various styles of images paired with text. CLIP is exceptional at encoding the semantic meaning of visual depictions, regardless of their style \cite{goh2021multimodal}.

Previous works that attempt to replicate human-like sketching often use sketch datasets of the desired level of abstraction to guide the form and style of the generated sketch \cite{Berger2013, Deformable_Stroke, Deep-Sketch-Abstraction}. While such data-driven approach can imitate the final rendering of human artwork, it requires the existence and availability of relevant datasets, and it restricts the output style to match this data. %
In contrast, we present an optimization-based photo-to-sketch generation technique that achieves different levels of abstraction without requiring an explicit sketch dataset. 
Our method uses the CLIP image encoder to guide the process of converting a photograph to an abstract sketch. CLIP encoding provides the semantic understanding of the concept depicted, while the photograph itself provides the geometric grounding of the sketch to the concrete subject.

Our sketches are defined using a set of thin, black strokes (Bézier curves) placed on a white background, and the level of abstraction is dictated by the number of strokes used.
Given the target image to be drawn, we use a differentiable rasterizer \cite{diffvg} to directly optimize the strokes’ parameters (control points positions) with respect to a CLIP-based loss.
We combine the final and intermediate activations of a pre-trained CLIP model to achieve both geometric and semantic simplifications.
For improved robustness, we propose a saliency-guided initialization process, based on the local attention maps of a pretrained vision transformer model.

The resulting sketches (see Figure~\ref{fig:teaser_clipasso}) demonstrate a combination of the semantic and visual features that capture the essence of the input object, while still being minimal and providing good category and instance level object recognition clues.

\begin{figure}[h!]
\centering
\begin{tabular}{@{\hskip2pt}c@{\hskip2pt}c@{\hskip2pt}c@{\hskip2pt}c}

    \includegraphics[width=\widthabs\linewidth]{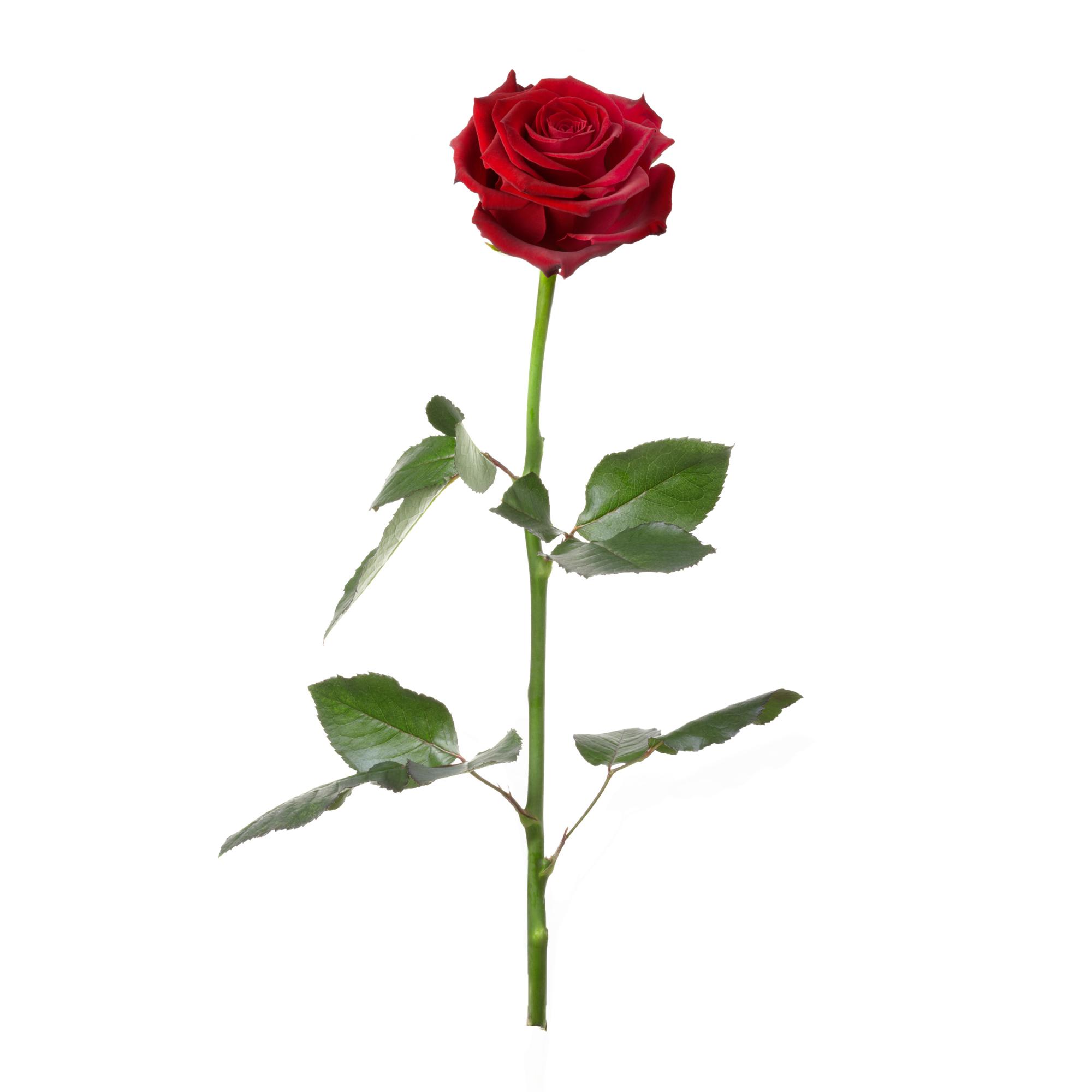} &
    \includegraphics[width=\widthabs\linewidth]{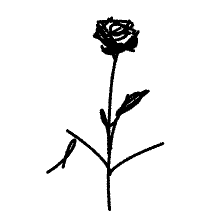} &
    \includegraphics[width=\widthabs\linewidth]{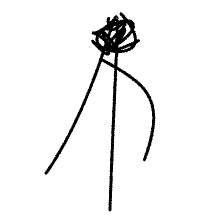} &
    \includegraphics[width=\widthabs\linewidth]{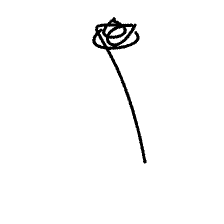} \\
    
    \midrule
    \includegraphics[width=\widthabs\linewidth]{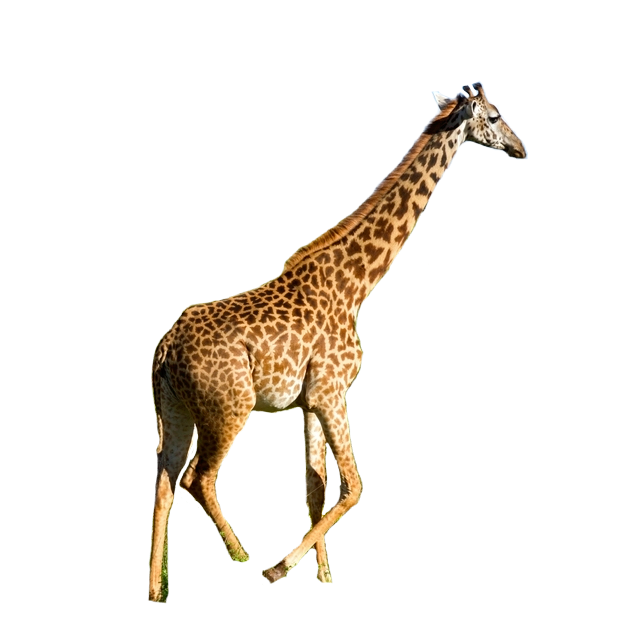} &
    \includegraphics[width=\widthabs\linewidth]{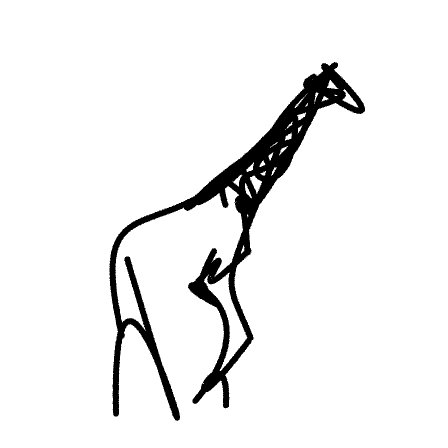} &
    \includegraphics[width=\widthabs\linewidth]{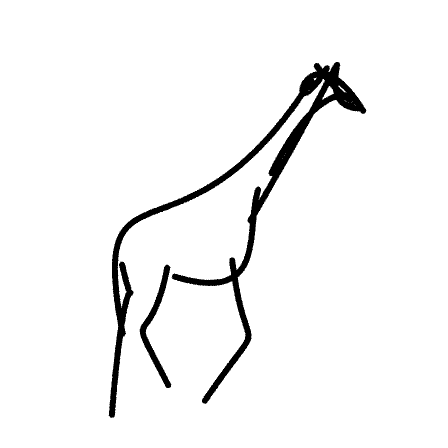} &
    \includegraphics[width=\widthabs\linewidth]{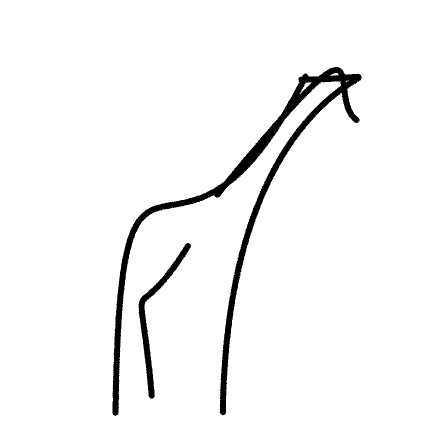} \\
    
    \midrule
    \includegraphics[width=\widthabs\linewidth]{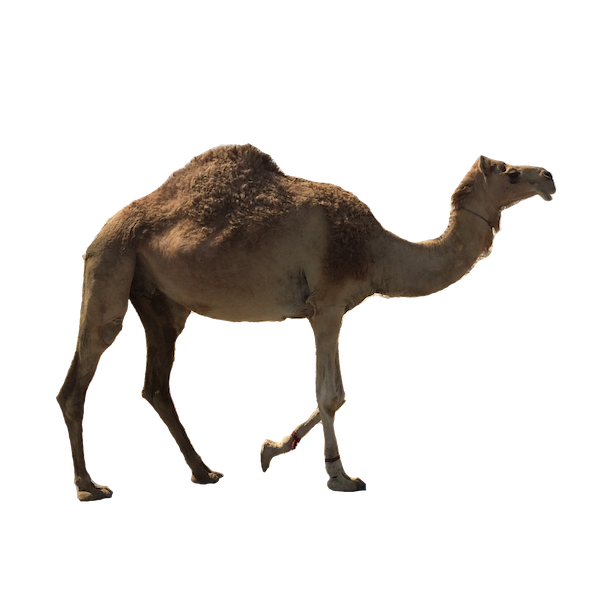} &
    \includegraphics[width=\widthabs\linewidth]{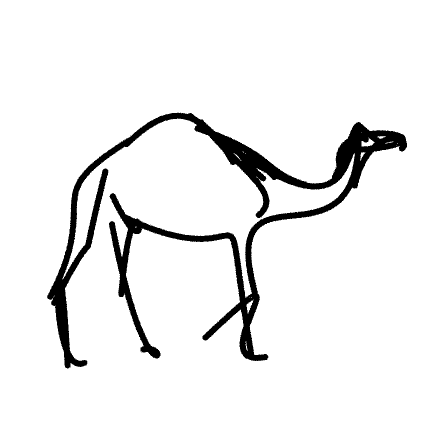} &
    \includegraphics[width=\widthabs\linewidth]{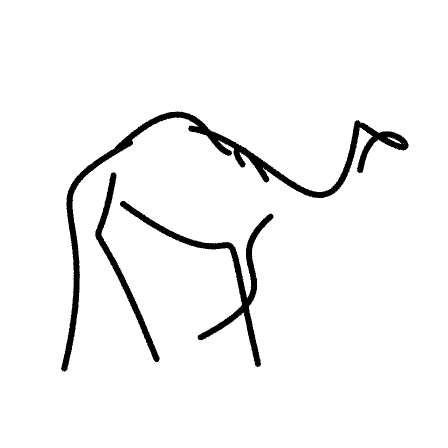} &
    \includegraphics[width=\widthabs\linewidth]{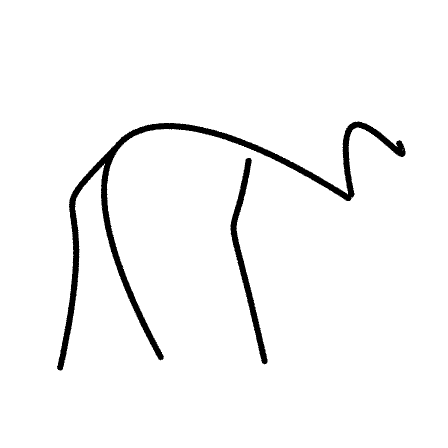} \\
    
    \midrule
    \includegraphics[width=\widthabs\linewidth]{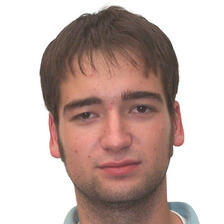} &
  \includegraphics[width=\widthabs\linewidth]{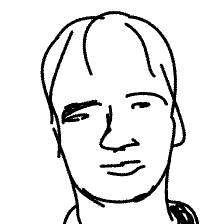} &
  \includegraphics[width=\widthabs\linewidth]{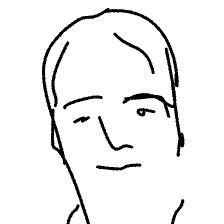} &
  \includegraphics[width=\widthabs\linewidth]{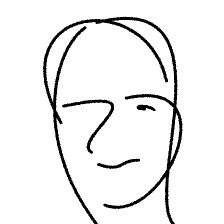} \\

\end{tabular}
 \caption{\small Different levels of abstraction produced by our method. Left to right: input images and increased level of abstraction. The top three sketches were produced using 16, 8, and 4 strokes in columns 2, 3, and 4, respectively, and the man's sketch was produced using 32, 16, and 8 strokes. }
\label{fig:abstraction_levels}
\end{figure}

\section{Related Work}

\begin{figure}[ht!]
    \centering
    \includegraphics[width=1\linewidth]{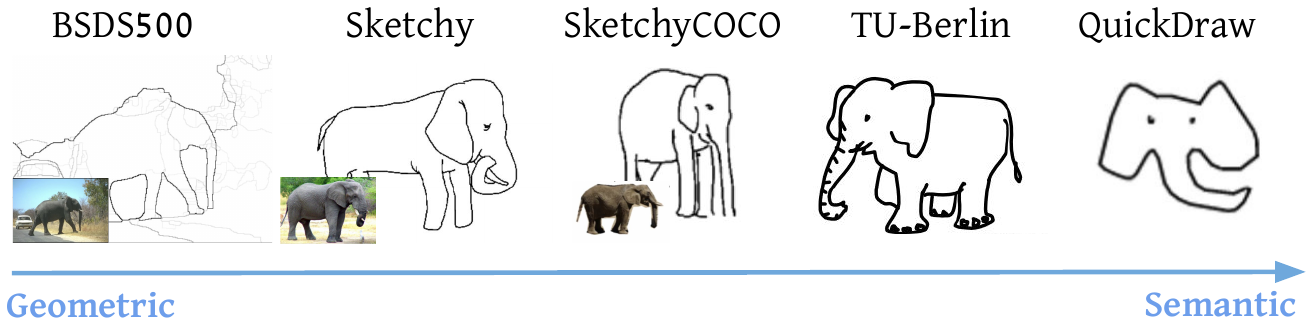}
    \caption{Variations in style and abstraction among sketch datasets — examples are arranged from left to right by the degree of abstraction: from edge-based to category-based sketches. For datasets that have image references, the input image is placed alongside the sketch; otherwise, the input is just the category label (e.g., ``elephant'').}
    \label{fig:datasets}
\end{figure}

Unlike edge-map extraction methods \cite{Winnemller2012XDoGAE, canny1986computational} which are purely based on geometry, free-hand sketch generation aims to produce sketches that are abstract in terms of structure and semantic interpretation so as to mimic a human-like style.
This high-level goal varies among different works, as there are many styles and levels of abstraction that can be produced. Consequently, existing works tend to choose the desired output style based on a given dataset: from highly abstract — guided only by a category-based text prompt \cite{SketchRNN}, 
to more concrete \cite{BSDS500data}, which is guided by contour detection. 
Figure \ref{fig:datasets} illustrates this spectrum.
While methods that rely on sketch datasets are limited to the abstraction levels present, our method is optimization based. Hence, it is capable of producing multiple levels of abstraction without relying on the existence of suitable sketch datasets or requiring a lengthy new training phase.

We provide a brief review of existing relevant photo-sketch synthesis works, which all rely on sketch-specific datasets.
Table \ref{tab:sketch_synth_comparison} summarizes the high-level characteristics that differentiate these methods.

\paragraph{Photo-Sketch Synthesis}
Early methods learn explicit models to synthesize facial sketches \cite{2001faces, Berger2013}. To generalise to categories beyond faces, Li et al. \cite{Deformable_Stroke} learn a deformable stroke model based on perceptual grouping.

In the deep learning era, it is intuitive to think of photo-sketch generation as a domain translation task.
However, the highly sparse and abstract nature of sketches introduces challenges for trivial methods \cite{pix2pix, pix2pixHD} to adhere to the sketch domain, and therefore sketch-specific adjustments must be made.

Song et al. \cite{song2018learning} propose a hybrid supervised-unsupervised multi-task learning approach with a shortcut cycle consistency constraint. 
Li et al. \cite{li2019photosketching} present a learning-based contour generation algorithm to resolve the diversity of the human drawings in the dataset.
Kampelmuhler and Pinz \cite{human-like-sketches} propose an encoder-decoder architecture, where the loss is guided by a pretrained sketch classifier network.
Qi and Su et al. \cite{Qi2021SketchLatticeLR} propose a lattice representation for sketches, employing LSTM and graph models to generate a vector sketch from points sampled from the edge map. The density of points determines the abstraction level of the sketch.

A different approach for image-sketch synthesis formulates the sketching task as a multi-agent referential game in which two reinforcement learning agents must communicate visual concepts to each other through sketches~\cite{Qiu2021EmergentGC, Mihai2021}. Similarly to these works, we were inspired by cognitive processes to identify and formulate our problem and objective. However, our primary focus is on abstractions as an integral part of sketches, and our objective is to produce sketches that are interpertable by humans. Their primary focus is on building a visual communication channel between agents, using sketches as a tool, where drawing is done in context.

\newcommand\RotText[1]{\rotatebox{90}{\parbox{2cm}{\centering#1}}}

\begin{wraptable}{R}{0.5\linewidth}
\caption{\small Comparison of sketch synthesis algorithms. \textbf{(A)} Is not restricted to categories from training dataset, \textbf{(B)} Can produce different levels of abstractions, \textbf{(C)} Is not limited to abstractions in the dataset, \textbf{(D)} Can produce  vector sketches, \textbf{(E)} Can produce a sequential sketch \textbf{(F)} Is not directly relying on the edge map.}
\resizebox{\linewidth}{!}{%
\begin{tabular}{ccccccc}
\toprule
\begin{tabular}[c]{@{}c@{}}{Method}\end{tabular} &
\begin{tabular}[c]{@{}c@{}}{A}\end{tabular} &
\begin{tabular}[c]{@{}c@{}}{B}\end{tabular} 
&\begin{tabular}[c]{@{}c@{}}{C}\end{tabular} & \begin{tabular}[c]{@{}c@{}}{D}\end{tabular} & \begin{tabular}[c]{@{}c@{}}{E}\end{tabular} & \begin{tabular}[c]{@{}c@{}}{F}\end{tabular} \\

\midrule 
Berger et al. \cite{Berger2013} & \xmark & \cmark & \xmark & \xmark & \cmark & \xmark \\

Li et al. \cite{Deformable_Stroke} &\xmark & \xmark & \xmark & \cmark & \xmark & \xmark \\

Muhammad et al. \cite{Deep-Sketch-Abstraction} &\xmark & \cmark & \cmark & \cmark & \cmark & \xmark \\

Song et al. \cite{song2018learning} &\xmark & \xmark & \xmark & \cmark & \cmark & \cmark \\

Li et al. \cite{li2019photosketching} &\cmark & \xmark & \xmark & \xmark & \xmark & \cmark \\

Kampelm{\"{u}}hler and Pinz \cite{human-like-sketches} & \xmark & \xmark & \xmark & \xmark & \xmark & \cmark \\

Qi and Su et al. \cite{Qi2021SketchLatticeLR} & \xmark & \cmark & \cmark & \cmark & \cmark & \xmark \\
\midrule
\textbf{Ours} & \cmark & \cmark & \cmark & \cmark & \xmark & \cmark \\
\bottomrule
\end{tabular}
}

\label{tab:sketch_synth_comparison}
\end{wraptable}

\paragraph{Sketches Abstraction}
Only two previous works propose a unified model to produce sketches of a given image at different levels of abstraction.
Berger et al. \cite{Berger2013} collected a dataset of portraits drawn by professional artists at different levels of abstraction. 
For each artist, a library of strokes is created indexed by shape, curvature, and length, and these are used to replace curves extracted from the image edge map.
Their method is limited to portraits and requires a new dataset for each level of abstraction.

Muhammad et al. \cite{Deep-Sketch-Abstraction} propose a stroke-level sketch abstraction model. A reinforcement learning agent is trained to select which strokes can be removed from an edge map representation of the input image without affecting its recognizability.
The recognition signal is provided by a sketch classifier trained on 9 classes from the QuickDraw dataset \cite{SketchRNN}, and hence to operate on new classes, a fine-tuning stage is required.

\paragraph{CLIP-based Image Abstraction} 
CLIP \cite{Radfordclip} is a neural network trained on 400 million image-text pairs collected from the internet with the objective of creating a joint latent space using contrastive learning.
Being trained on a wide variety of image domains along with lingual concepts, CLIP models are found to be very useful for a wide range of zero-shot tasks. 
The most relevant works within our context are by Frans et al.~\cite{CLIPDraw} (CLIPDraw), and Tian and Ha \cite{Evolution-Strategies-for-Creativity}.
CLIPDraw optimizes a set of random Bezier curves to create a drawing that maximizes the CLIP similarity for a given text prompt. Likewise, we also use a differentiable rasterizer \cite{diffvg} and a CLIP-based loss. However, while CLIPDraw is purely text-driven, we allow control over the output appearance, conditioned on the input image. For this purpose, we introduce a new geometric loss term and a saliency-guided initialization procedure.

Tian and Ha \cite{Evolution-Strategies-for-Creativity} employ evolutionary algorithms combined with CLIP, to produce creative abstract concepts represented by colored triangles guided by text or shape. Their results are limited to either fully semantic (using CLIP's text encoder) or entirely geometric (using L2), whereas we are able to integrate both.

\begin{figure*}[t]
  \centering
  \includegraphics[width=1\linewidth]{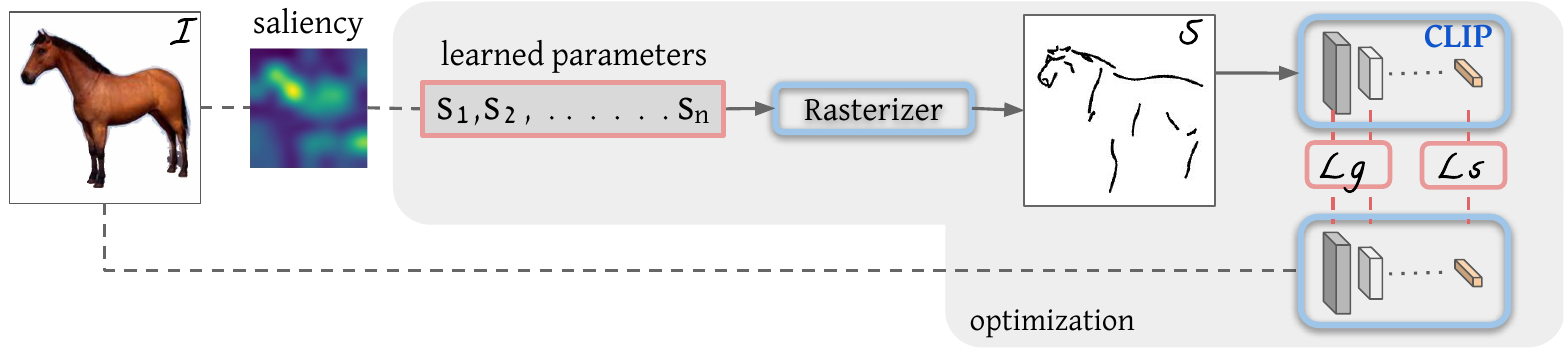}
  \caption{\small Method overview – Given a target image $\image$, and the number of strokes $n$, a saliency map is used as the distribution to sample the initial strokes locations $\{s_1, .. s_n\}$. A differentiable rasterizer $\renderer$ is used to create a resterized sketch $\sketch$. Both the sketch and the image are fed into a pretrained CLIP model to evaluate the geometric distance $L_g$ and semantic distance $L_s$ between the two. The loss is backpropagated through $\renderer$ to optimize the strokes parameters until convergence.
  The learned parameters and loss terms are highlighted in red, while the blue components are frozen during the entire optimization process, solid arrows are used to mark the backpropagation path.}
\label{fig:pipeline}
\end{figure*}

\section{Method}
We define a sketch as a set of $n$ black strokes $\{s_1, .. s_n\}$ placed on a white background. We use a two-dimensional Bézier curve with four control points $s_i = \{p_i^j\}_{j=1}^4 = \{(x_i,y_i)^j\}_{j=1}^4$ to represent each stroke. For simplicity, we only optimize the position of control points and choose to keep the degree, width, and opacity of the strokes fixed. However, these parameters can later be used to achieve variations in style (see Figure~\ref{fig:n_control_points}).
The parameters of the strokes are fed to a differentiable rasterizer $\renderer$, which forms the rasterized sketch $\sketch = \renderer(\{p_1^j\}_{j=1}^4, ... \{p_n^j\}_{j=1}^4) = \renderer(s_1, .. s_n)$. As is often conventional \cite{Evolution-Strategies-for-Creativity, spiralpp, Deep-Sketch-Abstraction}, we vary the number of strokes $n$ to create different levels of abstraction. 

An overview of our method can be seen in Figure \ref{fig:pipeline}.
Given a target image $\image$ of the desired subject, our goal is to synthesize the corresponding sketch $\sketch$ while maintaining both the semantic and geometric attributes of the subject.
We begin by extracting the salient regions of the input image to define the initial locations of the strokes.
Next, in each step of the optimization we feed the stroke parameters to a differentiable rasterizer $\renderer$ to produce the rasterized sketch. The resulting sketch, as well as the original image are then fed into CLIP to define a CLIP-based perceptual loss. We back-propagate the loss through the 
differentiable rasterizer and update the strokes' control points directly at each step until convergence of the loss function.

\begin{figure}[h]
    \centering
    \begin{tabular}{@{\hskip2pt}c@{\hskip2pt}c@{\hskip2pt}c@{\hskip2pt}c@{\hskip2pt}c}\\
    \toprule
    Input & XDoG & L2 & LPIPS & Ours \\
        \hline  
        \includegraphics[width=0.14\linewidth]{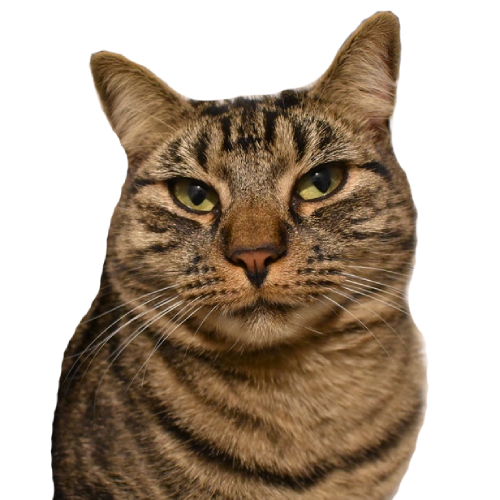} &
        \includegraphics[width=0.14\linewidth]{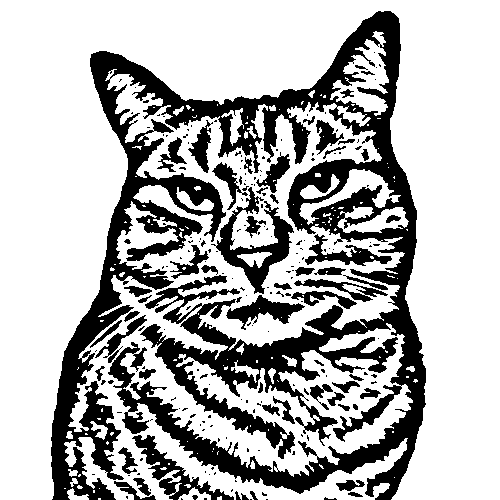}&
        \includegraphics[width=0.14\linewidth]{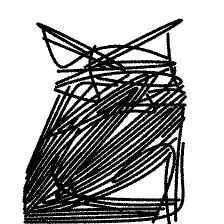} &
        \includegraphics[width=0.14\linewidth]{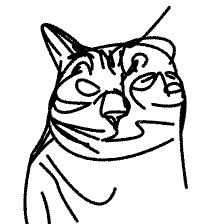} &
        \includegraphics[width=0.14\linewidth]{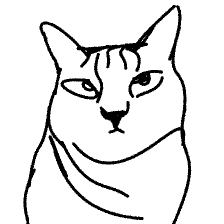}
    \end{tabular}
    \caption{\small Loss functions comparison — we optimize the strokes by minimizing different losses: L2 loss simply encourages the filling of colored pixels, LPIPS is more semantically aware, but the resulting sketch is still close to the edge map (see the XDog edges for comparison). In contrast, our CLIP-based loss allows better semantic depiction while preserving the morphology of the subject. }
    \label{fig:loss_compare}
\end{figure}

\paragraph{Loss Function}
As sketches are highly sparse and abstract, pixel-wise metrics are not sufficient to measure the distance between a sketch and an image. Additionally, even though perceptual losses such as LPIPS \cite{Zhang2018TheUE} can encode semantic information from images, they may not be suitable to encode abstract sketches, as illustrated in Figure \ref{fig:loss_compare} (for further analysis, please refer to the supplementary material).
One solution is to train task-specific encoders to learn a shared embedding space of images and sketches under which the distance between the two modalities can be computed~\cite{human-like-sketches, song2018learning}. This approach depends on the availability of such datasets, and requires additional effort for training the models. 

Instead, we utilize the pretrained image encoder model of CLIP, which was trained on various image modalities so that it can encode information from both natural images and sketches without the need for further training.
CLIP encodes high-level semantic attributes in the last layer since it was trained on both images and text.
We therefore define the distance between the embeddings of the sketch $CLIP(\renderer(\{s_i\}_{i=1}^n)$ and image $CLIP(\image)$ as:

\begin{equation}
\label{eq:semnticloss}
    L_{semantic} = dist\big(CLIP(\image),  CLIP(\renderer(\{s_i\}_{i=1}^n)\big),
\end{equation}

where $dist(x,y)=1 - \frac{x\cdot y}{||x||\cdot ||y||}$ is the cosine distance.
However, the final encoding of the network is agnostic to low-level spatial features such as pose and structure.
To measure the geometric similarity between the image and the sketch, and consequently, allow some control over the appearance of the output, we compute the $L2$ distance between intermediate level activations of CLIP:
\begin{equation}
\label{eq:geometric}
    L_{geometric} =\ \sum_{l}{\big{\|}\ CLIP_{l}(\image) - \    CLIP_{l}(\renderer(\{s_i\}_{i=1}^n))\ \big{\|}_2^2},
\end{equation}

where $CLIP_{l}$ is the $CLIP$ encoder activation at layer $l$. Specifically, we use layers 3 and 4 of the ResNet101 CLIP model.
The final objective of the optimization is then defined as:
\begin{equation}
\label{eq:finalloss}
\min_{\{s_i\}_{i=1}^n} {L_{geometry} + w_{s} \cdot L_{semantic}},
\end{equation}
with $w_s = 0.1$.
We analyze the contribution of different layers and weights, as well as the results of using different CLIP models in the supplementary material.

\paragraph{Optimization}

\begin{figure}[ht]
\centering
\begin{tabular}{@{\hskip2pt}c@{\hskip2pt}c@{\hskip2pt}c@{\hskip2pt}c@{\hskip2pt}c@{\hskip2pt}c}
    \midrule
    Input & 0 & 100 & 300 & 1000 & 2000 \\
    
    \midrule
    \includegraphics[width=0.12\linewidth]{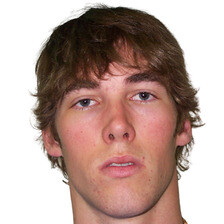} &
    \includegraphics[width=0.12\linewidth]{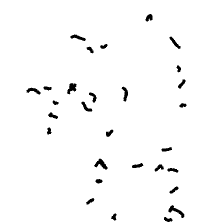} &
    \includegraphics[width=0.12\linewidth]{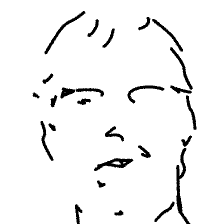} &
    \includegraphics[width=0.12\linewidth]{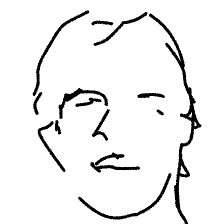} &
    \includegraphics[width=0.12\linewidth]{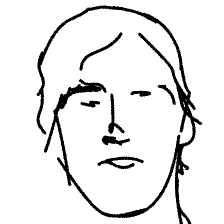} &
    \includegraphics[width=0.12\linewidth]{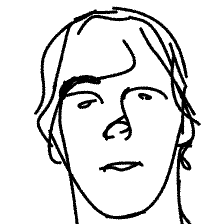}\\
\end{tabular}
\caption{\small The sketch appearance throughout the optimization iterations. \copyright Face image from \cite{Minear2004ALD}, used with permission.}
\label{fig:optimization_stages}
\end{figure}

Our goal is to optimize the set of parameters 
$\{s_i\}_{i=1}^{n} = \{\{p_i^j\}_{j=1}^4\}_{i=1}^n$
to define a sketch that closely resembles the target image $\image$ in terms of both geometry and semantics.
At each step of the optimization, we use the Adam optimizer \cite{Kingma2015AdamAM} to compute the gradients of the loss with respect to the strokes' parameters $\{s_i\}_{i=1}^n$.
We follow the same data augmentation scheme suggested in CLIPDraw \cite{CLIPDraw} and apply random affine augmentations to both the sketch and the target image before feed-forwarding into CLIP. The transformations we use are RandomPerspective and RandomResizedCrop.
These augmentations prevent the generation of adversarial sketches, which minimize the objective but are not meaningful to humans.
We repeat this process until convergence, (when the difference in loss between two successive evaluations is less than 0.00001). This typically takes around 2000 iterations. 
Figure \ref{fig:optimization_stages} illustrates the progression of the generated sketch as the optimization evolves.
The learning rate is set to 1 and we evaluate the output sketch every 10 iterations. Evaluation is done by computing the loss without random augmentations.
It takes 6 minutes to run 2000 iterations on a single Tesla V100 GPU.

\begin{figure}
    \centering
    \begin{subfigure}[b]{0.6\linewidth}
        \begin{tabular}{@{\hskip2pt}c@{\hskip2pt}c@{\hskip2pt}c@{\hskip2pt}c@{\hskip2pt}c}
            \midrule
            Input & Attention & Distribution & Proposed & Random \\
            \midrule
            \includegraphics[width=0.18\linewidth]{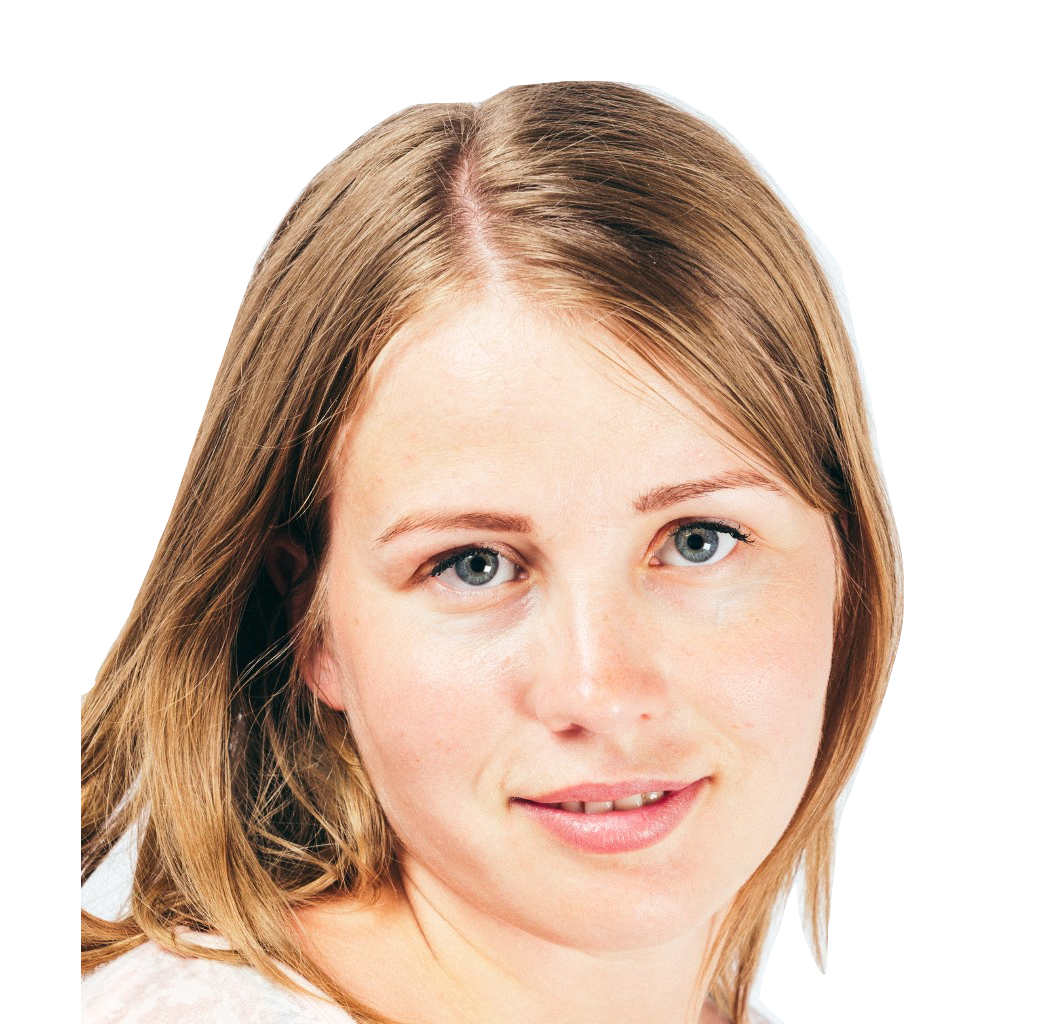} &
            \includegraphics[width=0.18\linewidth]{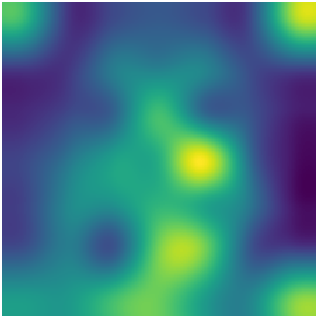} &
            \includegraphics[width=0.18\linewidth]{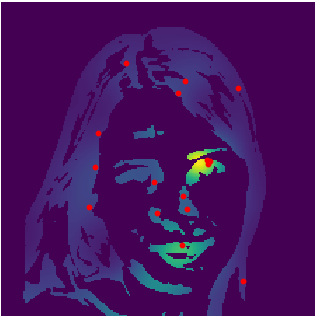} &
            \includegraphics[width=0.18\linewidth]{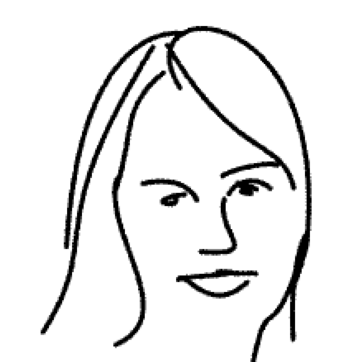} &
            \includegraphics[width=0.18\linewidth]{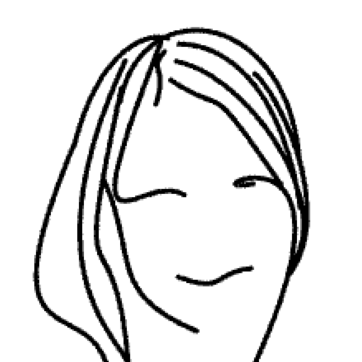}
        \end{tabular}\subcaption{Saliency-guided initialization }
        \label{fig:initialisation_random_comp}
    \end{subfigure}
    \\[3ex]
    \begin{subfigure}[b]{0.6\linewidth}
    \centering
        \begin{tabular}{@{\hskip2pt}c@{\hskip2pt}c@{\hskip5pt}c@{\hskip2pt}c@{\hskip2pt}c@{\hskip2pt}c@{\hskip2pt}c}
        \toprule
            \includegraphics[width=0.13\linewidth,cfbox=cyan 0.5pt 0.5pt]{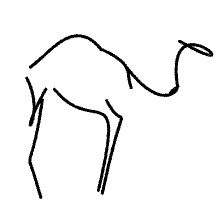} &
            \includegraphics[width=0.13\linewidth]{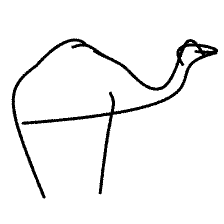} &
            \includegraphics[width=0.13\linewidth]{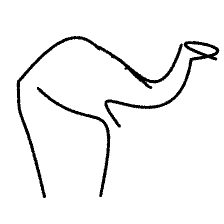} & &
            
            \includegraphics[width=0.13\linewidth,cfbox=cyan 0.5pt 0.5pt]{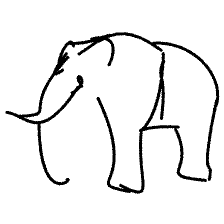} &
            \includegraphics[width=0.13\linewidth]{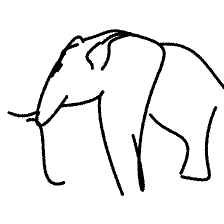} &
            \includegraphics[width=0.13\linewidth]{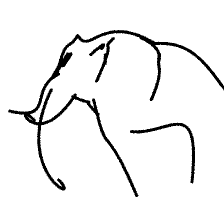} \\
            \bottomrule
        \end{tabular}
        \subcaption{Automatic selection procedure of the final sketch}
        \label{fig:different_seeds}
    \end{subfigure}
    \caption{\small Strokes Initialization. (a) Left to right: input, the saliency map produced from CLIP ViT activations, final distribution map (adjusted to adhere to image edges) with sampled initial stroke locations (in red), the sketch produced using the proposed initialization procedure, and the sketch produced when using random initialization. (b) Results of three different initializations with the same number of strokes. The sketches marked in blue produced the lowest loss value, and would thus be used as the final output. \copyright Face image from \cite{Minear2004ALD}, used with permission.}
    \label{fig:initialisation_examples}
\end{figure}

\newpage
\paragraph{Strokes Initialization}
Our objective function is highly non-convex. Therefore, the optimization process is susceptible to the initialization (i.e., the initial location of the strokes). 
This is especially significant at higher levels of abstraction – where very few strokes must be wisely placed to emphasize semantic components.
For example, in Figure \ref{fig:initialisation_random_comp}, the sketches in the last two columns were produced using the same number of strokes, however, in the ``Random'' initialization case, more strokes were devoted to the hair while the eyes, nose, and mouth are more salient and critical features of the face.

To improve convergence towards semantic depictions, we place the initial strokes based on the salient regions of the target image. To find these regions, we use the pretrained vision transformer \cite{Kolesnikov2021ViT} ViT-B/32 model of CLIP, that performs global context modeling using self-attention between patches of a given image to capture meaningful features.
We use the recent transformer interpretability method by Chefer et al. \cite{Chefer_2021_ICCV} to extract a relevancy map from the self-attention heads, without any text supervision.
Next, we multiply the relevancy map with an edge map of the image extracted using XDoG method~\cite{Winnemller2012XDoGAE}. Multiplying with XDoG is used to strengthen the morphological positioning of the strokes, motivated by the hypothesis that edges are effective in predicting where people draw lines \cite{Hertzmann2021TheRO}.
Finally, we normalize the final relevancy map using softmax and use it as a distribution map so that pixels in salient regions are assigned a higher probability. We sample $n$ positions (pixels), and use them as the position of the first control point $p_i^1$ of each Bezier curve. 
The other 3 additional points ($p_i^2, p_i^3, p_i^4$) are sampled within a small radius (0.05 of image size) around $p_i^1$ to define the initial set of Bezier curves $\{\{p_i^j\}_{j=1}^4\}_{i=1}^n$.

Figure \ref{fig:initialisation_random_comp} illustrates this procedure.
It can be seen that our saliency-based initialization contributes significantly to the quality of the final sketch compared to random initialization. 

This sampling-based approach also lends itself to providing variability in the results. In all our examples we use 3 initializations and automatically choose the one that yields the lowest loss (see Figure \ref{fig:different_seeds}). Only for the highly abstract cases of Figure~\ref{fig:abstraction_levels} (rightmost column) we used 5 initializations.
We further analyze the initialization procedure and variability in the supplementary material.

\section{Results}
\label{sec:results_clipasso}
Section \ref{sec:qualitative} provide qualitative evaluations.
In section \ref{sec:comparison} we compare our method with existing image-to-sketch methods, which were all trained on sketch-specific datasets.
In Section \ref{sec:quantitative}, we supply a quantitative evaluation of our method's ability to produce recognizable sketches testing both category and instance recognition.
For images with background, we use an automatic method (U2-Net \cite{Qin_2020_PR}) to mask out their background.
We provide further analysis of our method, extra results, and extended comparison with other methods in the supplemental file.

\begin{figure}[h]
    \centering
    \includegraphics[width=0.7\linewidth]{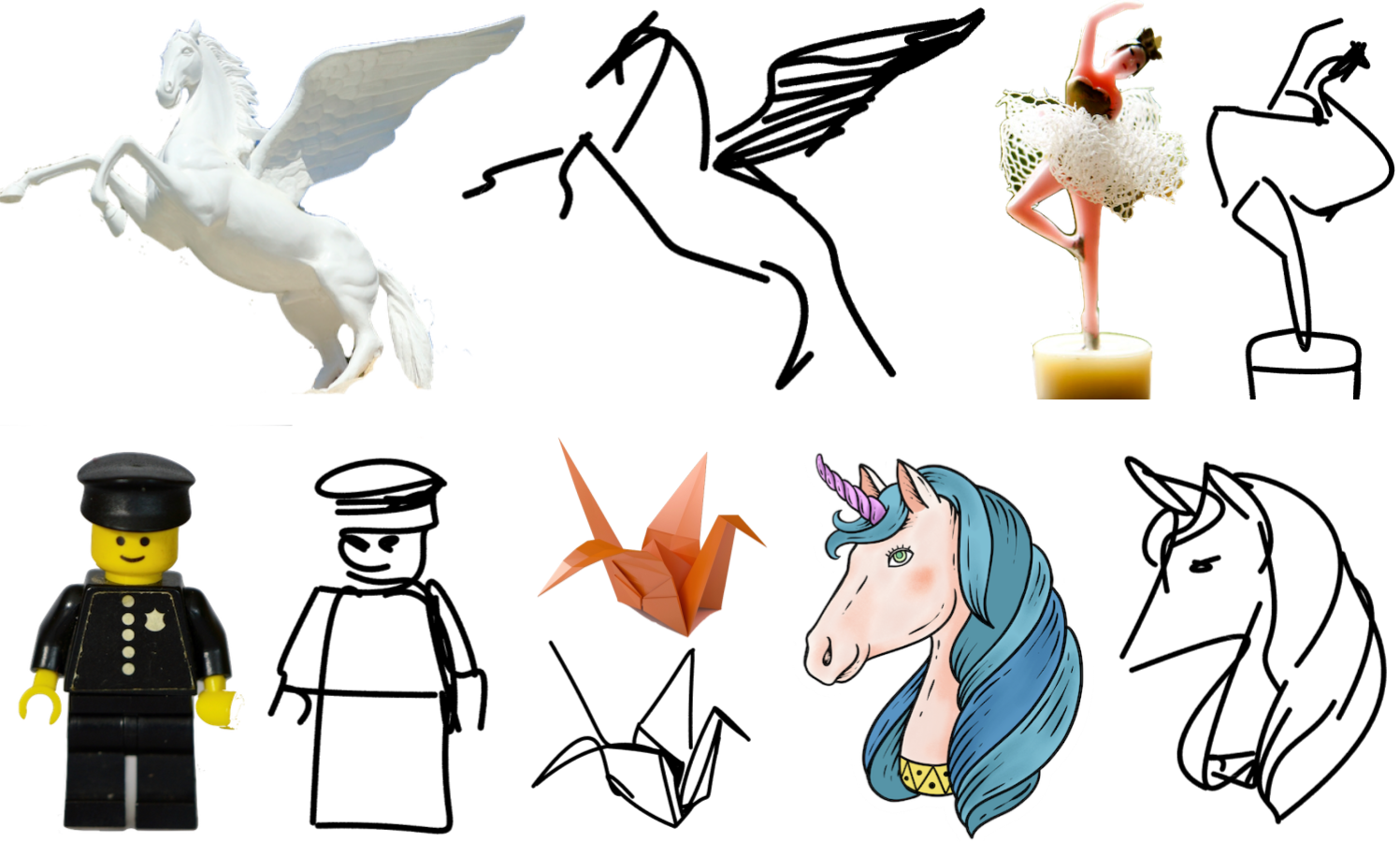}
    \caption{\small Sketches produced by our method for infrequent categories.}
    \label{fig:robust}
\end{figure}

\subsection{Qualitative Evaluation}
\label{sec:qualitative}
Our approach is different from conventional sketching methods in that it does not utilize a sketch dataset for training, rather it is optimized under the guidance of CLIP. Thus, our method is not limited to specific categories observed during training, as no category definition was introduced at any stage. This makes our method robust to various inputs, as shown in Figures~\ref{fig:teaser_clipasso} and \ref{fig:robust}.

In Figures \ref{fig:teaser_clipasso} and ~\ref{fig:abstraction_levels} we demonstrate the ability of our method to produce sketches at different levels of abstraction.
As the number of strokes decreases, the task of minimizing the loss becomes more challenging, forcing the strokes to capture the essence of the object. For example, in the abstraction process of the flamingo in Figure \ref{fig:teaser_clipasso}, the transition from 16 to 4 strokes led to the removal of details such as the eyes, feathers, and feet, while maintaining the important visual features such as the general pose, the neck and legs which are iconic characteristics of a flamingo. 

\begin{figure}[t]
    \begin{subfigure}[b]{0.49\linewidth}
    \centering
            \begin{tabular}{@{\hskip2pt}c@{\hskip2pt}c@{\hskip2pt}c@{\hskip2pt}c}
            4 cp & 3 cp & 2 cp  \\
            \includegraphics[width=0.29\linewidth]{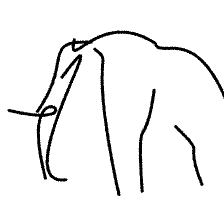} & 
            \includegraphics[width=0.29\linewidth]{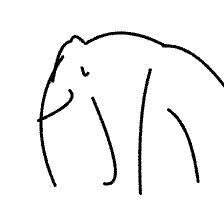} &
            \includegraphics[width=0.29\linewidth]{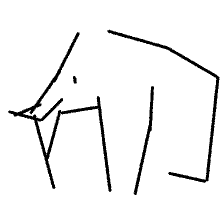}
            \end{tabular}
             \subcaption{Changing the degree of the curves}\label{fig:n_control_points}
        \end{subfigure}
        \begin{subfigure}[b]{0.5\linewidth}
            \includegraphics[width=0.45\linewidth]{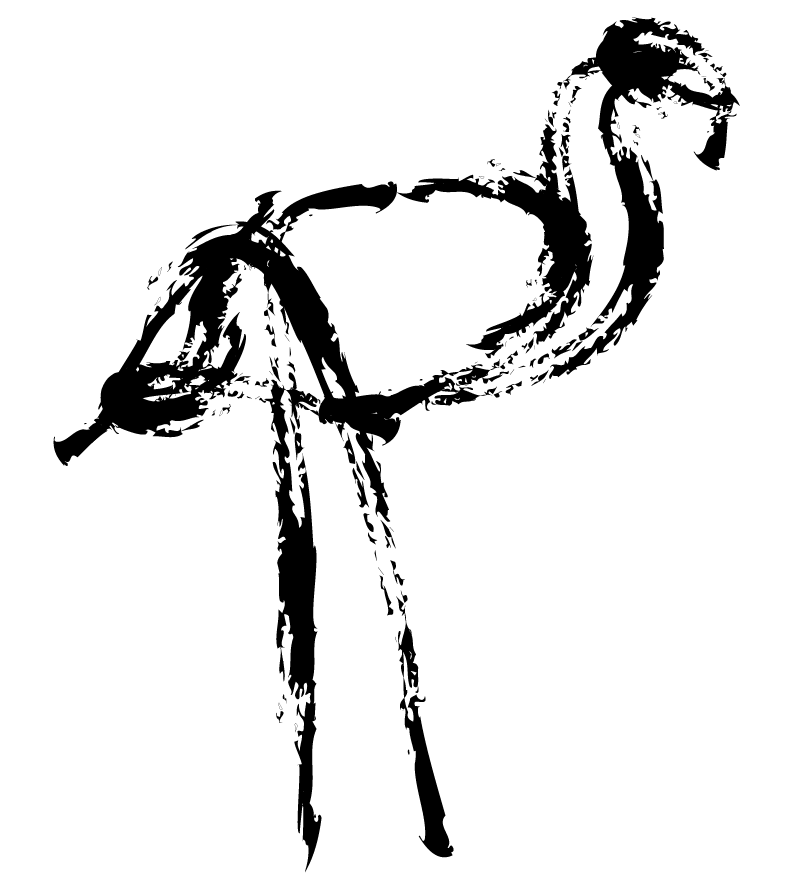}
            \includegraphics[width=0.53\linewidth]{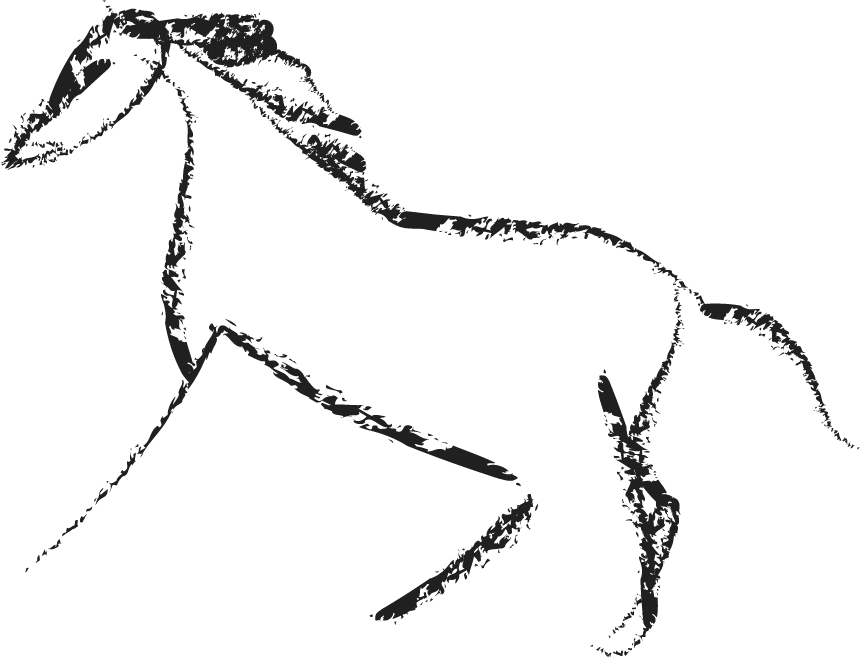}
            \subcaption{Editing the brush style on SVGs}
            \label{fig:pencil_style}
        \end{subfigure}
        
    \caption{\small Changing sketch style. (a) From left to right are the results produced by our method when using Bézier curves with 4, 3, and 2 control points (cp), respectively. We can see how this affects the style of the output sketch. (b) Using Adobe Illustrator, horse — pencil feather, flamingo — dry brush. }
    \label{fig:style_change}
\end{figure}

Besides changing the number of strokes, different sketch styles can be achieved by varying the degree of the strokes (Figure~\ref{fig:n_control_points}) or using a brush style on top of the vector strokes (Figure \ref{fig:pencil_style}).

\subsection{Comparison with Existing Methods}
\label{sec:comparison}
\begin{figure}
    \centering
    \includegraphics[width=0.75\linewidth]{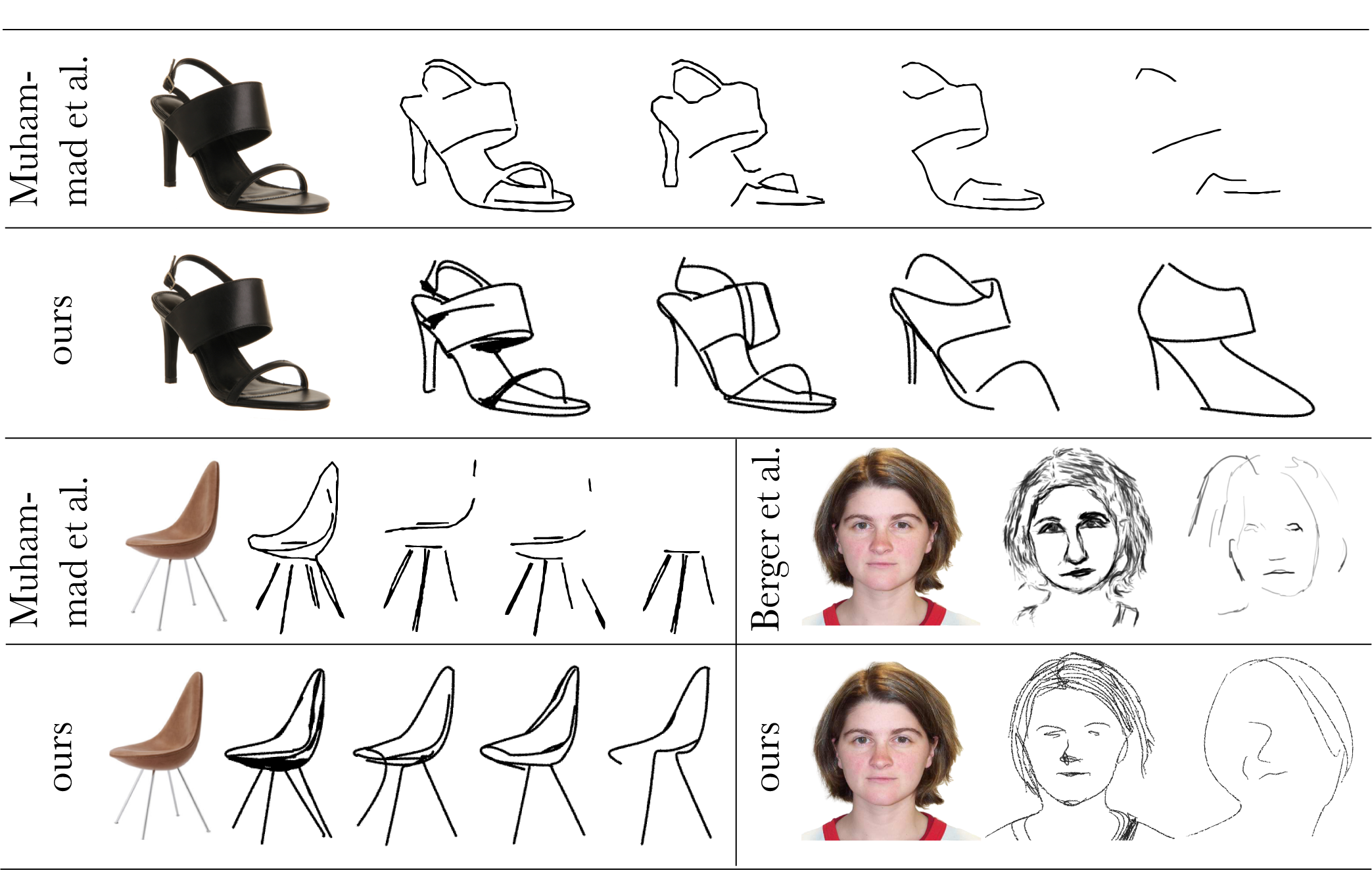}
    \caption{\small Levels of Abstraction Comparison — in the top and left part are comparisons to Muhammad et al. \cite{Deep-Sketch-Abstraction}. The leftmost column shows the input image, and the next four columns show different levels of abstraction. For the shoe and chair our results were produced using 32, 16, 8, and 4 curves (from left to right).
    In the right bottom part is a comparison to Berger et al. \cite{Berger2013}, we use 64 and 8 strokes to generate our sketches.}
    \label{fig:comp_shoes}
\end{figure}

\paragraph{Sketches with different levels of abstraction.}
Only a few works have attempted to sketch objects at different levels of abstraction. In Figure \ref{fig:comp_shoes} we compare with Muhammad et al. \cite{Deep-Sketch-Abstraction} and Berger et al. \cite{Berger2013}.
The results by Muhammad et al. demonstrate four levels of abstraction on two simple inputs — a shoe and a chair (in the absence of their code, the results were taken directly from the paper).
We produce sketches at four levels of abstraction using 32, 16, 8, and 4 strokes.
The sketches by Muhammad et al. are coherent with the geometry of the image; but to achieve higher levels of abstraction, they only remove strokes from the generated sketch without changing the remaining ones. This can result in losing class-level recognizability at higher levels of abstraction (rightmost sketches). 
Such an approach is sub-optimal, since a better arrangement may be possible for fewer strokes.
Our method successfully produces a recognizable rendition of the subject while preserving its geometry, even in the challenging 4-stroke case (rightmost sketch). 

In the right bottom part of Figure \ref{fig:comp_shoes} we compare with the method of Berger et al. \cite{Berger2013}. Their results were provided by the authors and demonstrate two levels of abstraction generated based on the style of a particular artist. 
We use 64 and 8 strokes, respectively, to achieve two comparable levels of abstraction and place a pencil style on top of the generated sketch to better fit the artist's style. As can be seen, our approach is more geometrically coherent while still allowing abstraction. Their results fit better to a specific style, but can only work with faces and are limited to the dataset gathered.

\begin{figure}
\centering
\begin{tabular}{@{\hskip1pt}c|@{\hskip1pt}c@{\hskip1pt}c@{\hskip1pt}c|@{\hskip1pt}c@{\hskip1pt}c}
     Input & \multicolumn{3}{c}{Data-Driven} & \multicolumn{2}{|c}{Ours} \\
     & A & B & C & 16 & 8 \\
     \hline
    \includegraphics[width=\widthteapot\linewidth]{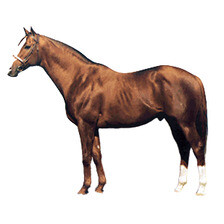} &
    \includegraphics[width=\widthteapot\linewidth]{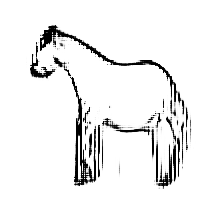} &
    \raisebox{.2\height}{\includegraphics[width=\widthteapot\linewidth]{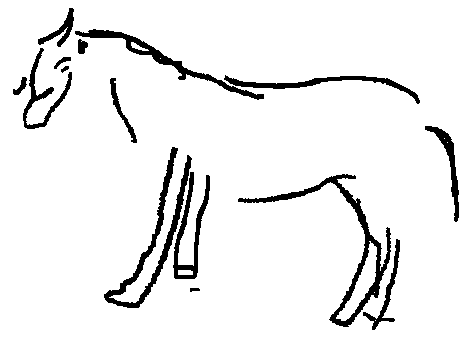}} &
    \includegraphics[width=\widthteapot\linewidth]{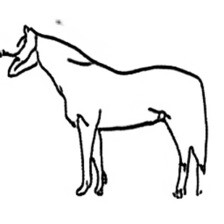} &
    \includegraphics[width=\widthteapot\linewidth]{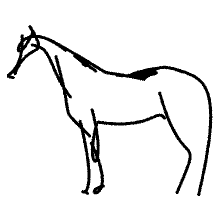} &
    \includegraphics[width=\widthteapot\linewidth]{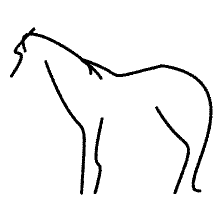} \\
   
    \includegraphics[trim={0 1.5cm 0  1.5cm},clip,width=\widthteapot\linewidth]{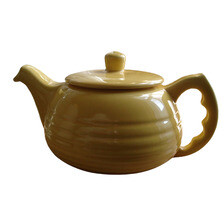} &
    \includegraphics[trim={0 1cm 0  1cm},clip,width=\widthteapot\linewidth]{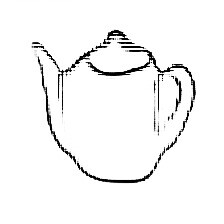} &
    \raisebox{.2\height}{\includegraphics[width=\widthteapot\linewidth]{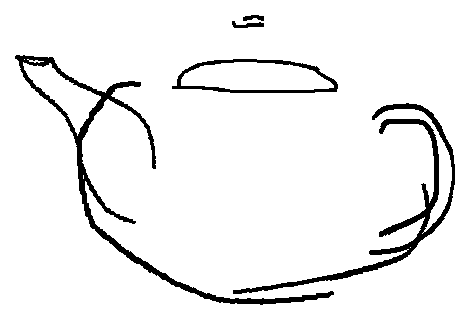}} &
    \includegraphics[trim={0 1.5cm 0  2cm},clip,width=\widthteapot\linewidth]{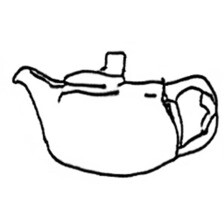} &
    \includegraphics[trim={0 1cm 0  1cm},clip,width=\widthteapot\linewidth]{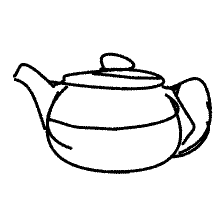} &
    \includegraphics[trim={0 1cm 0  1cm},clip,width=\widthteapot\linewidth]{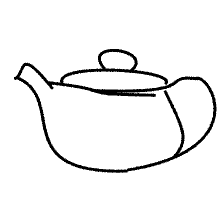} \\
    
    \includegraphics[width=\widthteapot\linewidth]{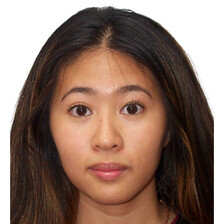} & 
    &
    \includegraphics[width=0.12\linewidth]{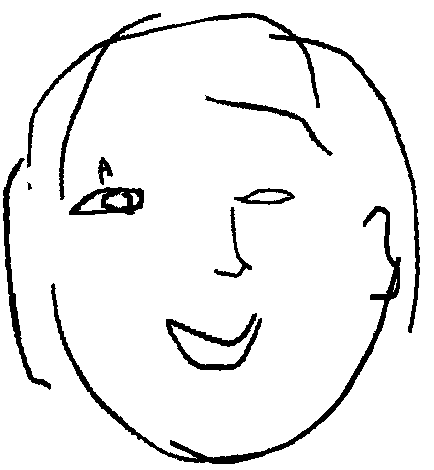} &
    \includegraphics[width=\widthteapot\linewidth]{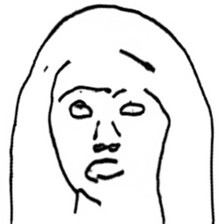} &
    \includegraphics[width=\widthteapot\linewidth]{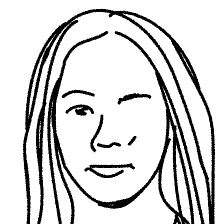} &
    \includegraphics[width=\widthteapot\linewidth]{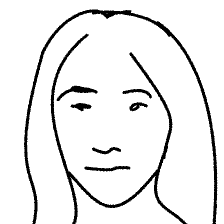}
    \\

     \midrule
      & D & E &  & 16 & 4 \\
    \includegraphics[trim={0 1cm 0  1cm},clip,width=\widthteapot\linewidth]{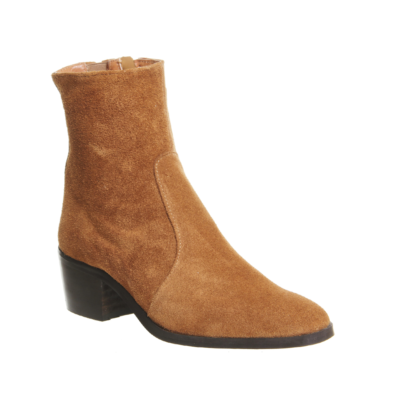} & 
    \includegraphics[trim={0 1cm 0  1cm},clip,width=\widthteapot\linewidth]{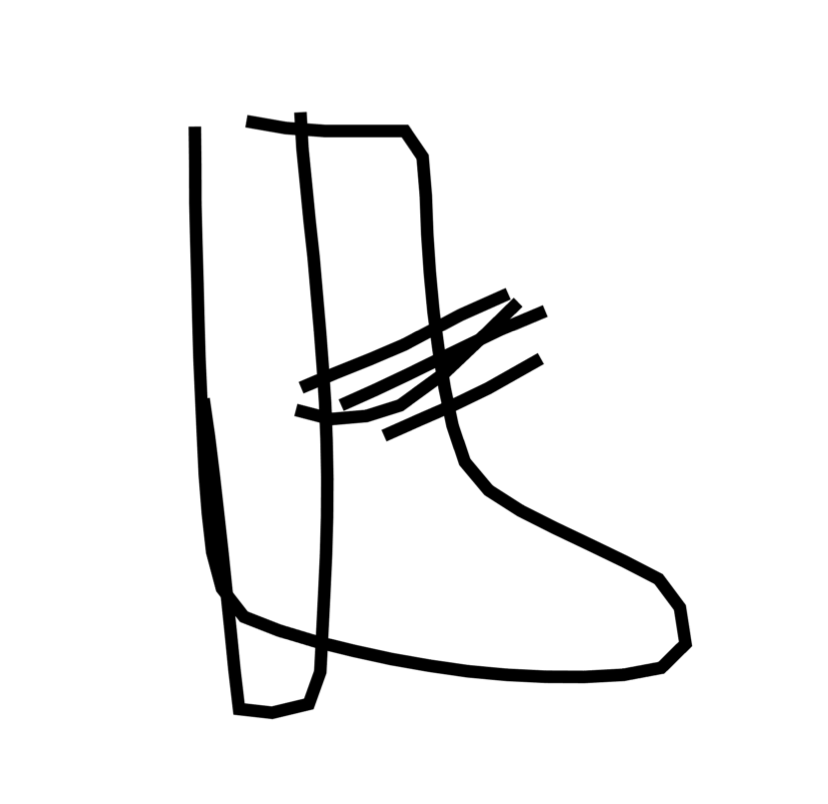} & 
    \includegraphics[trim={0 1cm 0  1cm},clip,width=\widthteapot\linewidth]{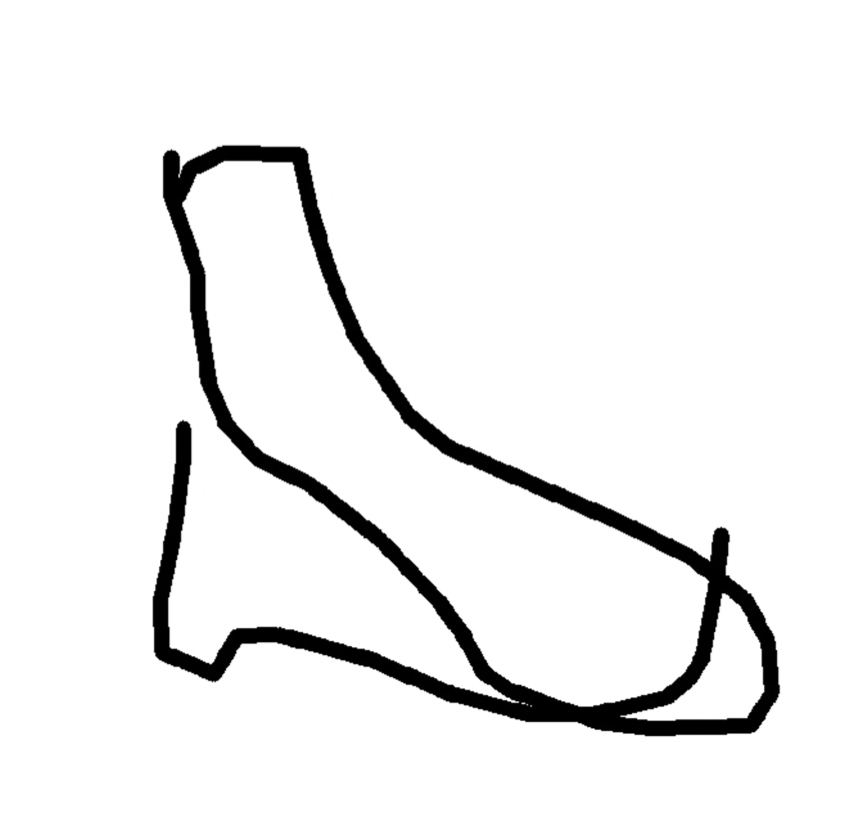} &
    &
    \includegraphics[trim={0 1cm 0  1cm},clip,width=\widthteapot\linewidth]{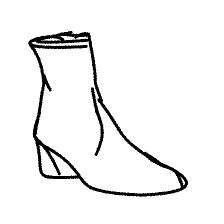} &
    \includegraphics[trim={0 1cm 0  1cm},clip,width=\widthteapot\linewidth]{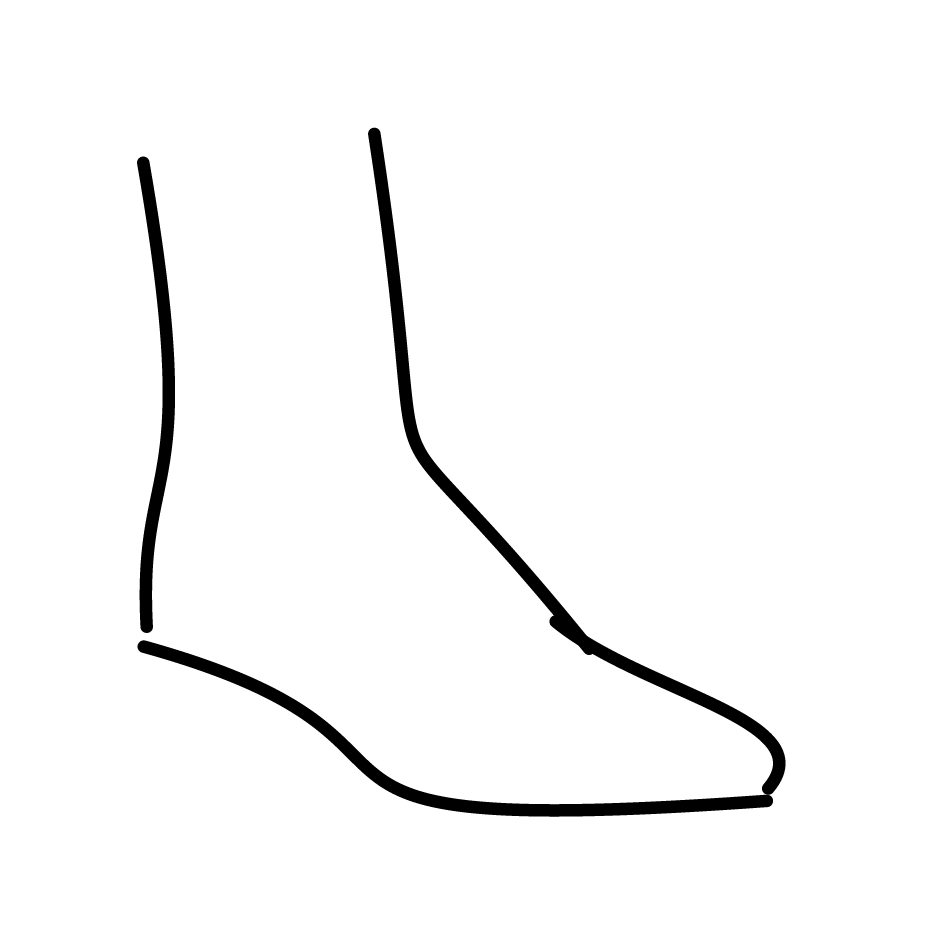}

\end{tabular}
 \caption{\small Comparison to Existing Image-to-Sketch Works — the leftmost column shows the input images. The methods presented are (A) Kampelmühler and Pinz \cite{human-like-sketches}, (B) Li et al. \cite{Deformable_Stroke}, (C) Li et al. \cite{li2019photosketching} (D) Song et al. \cite{song2018learning}, (E) SketchLattice \cite{Qi2021SketchLatticeLR}.}
 
\label{fig:comp1}
\end{figure}

\paragraph{Photo-Sketch Synthesis.}
In Figure \ref{fig:comp1} we present a comparison with the five works outlined in Table \ref{tab:sketch_synth_comparison}. 
The results by Kampelmühler and Pinz \cite{human-like-sketches} (A), Li et al. \cite{Deformable_Stroke} (B) and Li et al. \cite{li2019photosketching} (C) were generated based on the authors' implementation and best practice.
Due to the lack of a publicly available implementation of SketchLattice \cite{Qi2021SketchLatticeLR} (E), their results are taken directly from the paper.
We present the sketches of Song et al. \cite{song2018learning} (D) on shoe images, since their method only works with shoes and chairs. 

Each of these methods define a specific objective which influences their dataset selection and final output style.
Li et al. \cite{li2019photosketching} (C) aim for boundary-like drawings, and indeed, geometric coherence is achieved with the input image, capturing salient outlines.
The other methods are designed to produce human-like sketches of non-experts, and indeed, the synthesized sketches exhibit a ``doodle-like'' style. Furthermore, the methods learn highly abstract concepts (such as highlighting the eyes) while maintaining some relation to the geometry of the input object.

Each method accomplishes their respective objective, but we wish to emphasize the particular benefits of our method.
First, all of the above methods are sketch-data dependent, meaning they can only be used with the style and level of abstraction observed during training. With our framework, we can handle images of all categories and produce sketches of various levels of abstraction simply by changing the number of strokes. Although every method can be retrained with new datasets, this is neither convenient nor practical, and depends on the availability of such datasets.
Second, while each method leans towards a more semantic or a more geometric sketching style, our method can provide both. For example, our method did not produce a perfect alignment of the legs of the horse, as in (C), but it captured the movement of the horse in a minimal way.

\begin{figure}
\centering
    \begin{tabular}{@{\hskip2pt}c@{\hskip2pt}c@{\hskip2pt}c|@{\hskip2pt}c@{\hskip2pt}c@{\hskip2pt}c}
         Input & CLIPDraw & Ours & Input & CLIPDraw & Ours  \\
         \hline
         \includegraphics[width=0.15\linewidth]{clipasso/figs/comp_objects/input/face_7_edge.jpg} &
         \includegraphics[width=0.15\linewidth]{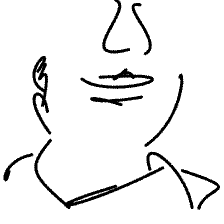} &
         \includegraphics[width=0.15\linewidth]{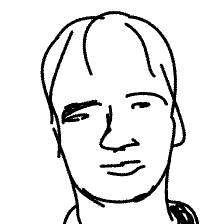} &
         
         \includegraphics[width=0.15\linewidth]{clipasso/figs/comp_objects/input/horse_458355343_aeb34d03e8.jpg} &
        \includegraphics[width=\widthteapot\linewidth]{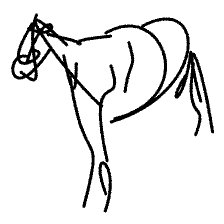} &
        \includegraphics[width=\widthteapot\linewidth]{clipasso/figs/comp_objects/ours/horse_458355343_aeb34d03e8.png}
    \end{tabular}
    \caption{\small Comparison to CLIPDraw \cite{CLIPDraw}.
    All sketches were produced using 16 strokes. }
    \label{fig:clipdraw_comp}
\end{figure}

In Figure \ref{fig:clipdraw_comp} we provide a comparison with CLIPDraw \cite{CLIPDraw}. The text input for CLIPDraw is replaced with the target image. This was made possible since CLIP encodes both text and images to the same latent space. To provide a comparable visualization, we constrain the output primitives of CLIPDraw in the same manner as we defined our strokes.
As can be seen, although the parts of the subject can be recognizable using CLIPDraw, since there is no geometric grounding to the image, the overall structure is destroyed. Further comparisons of CLIPDraw incorporating text and color can be found in the supplemental file.
\setlength{\columnsep}{20pt}

\newpage
\subsection{Quantitative Evaluation}
\label{sec:quantitative}
We conduct a perceptual study to assess both the category-level and instance-level recognizability of the sketches generated by our method at different levels of abstraction.
Additionally, similar to previous image-to-sketch works \cite{human-like-sketches, song2018learning, Deep-Sketch-Abstraction}, we also use pretrained classifier networks to evaluate the category-level recognizability of the sketches generated by our method. 

\paragraph{Perceptual Study}
We choose five popular animal classes (cat, dog, elephant, giraffe, and horse) from the SketchyCOCO dataset \cite{gao2020sketchycoco} and randomly sample five images per class.
We synthesize sketches at four levels of abstraction for each image, with 4, 8, 16, and 32 strokes. In total, we generate 100 sketches. We compare the recognition rates with the two recent photo-sketch synthesis methods by Kampelmühler and Pinz \cite{human-like-sketches} and Li et al. \cite{li2019photosketching}. The sketches were produced with one level of abstraction, suited to the capabilities of the methods. 
We had 121 participants in total, out of which 60 were assigned to evaluate our sketches at different levels of abstraction, 38 to the sketches by Kampelmühler and Pinz \cite{human-like-sketches}, and 24 to the sketches by Li et al. \cite{li2019photosketching}.

Each participant was presented with randomly selected sketches of a single method. To examine category-level recognition, participants were asked to choose the correct category text description alongside four confound categories and the option 'None'. For the instance-level recognition experiment, the distractors are images from the same object category. 
Table \ref{tab:user_study_animal} shows the average recognition rates attained from the perceptual study.
Both the category-level and instance-level recognizability are inversely correlated with the level of abstraction. At four strokes, we can see that our sketches are hardly recognizable at the category level (36\%), which illustrates a ``breaking point'' of our method. At eight strokes and above, both instance and class level rates are high. Li et al. \cite{li2019photosketching} demonstrate full recognition at the instance level; this is understandable given that the sketches are contour-based and not abstract. At 16 and 32 strokes, we achieve comparable rates, and even with the high level of abstraction when using only eight strokes, we achieve 95\% instance level recognizability. The sketches by Kampelmühler and Pinz are more abstract, which explains their low accuracy at the instance and class level.

\begin{table}[ht!]
\centering
  \caption{\small Perceptual study results – average recognition rates. (A) Kampelmühler and Pinz \cite{human-like-sketches}, (B) Li et al. \cite{li2019photosketching}.}
  \label{tab:user_study_animal}
  \begin{tabular}{cccccccl}
    \toprule
     & A & B & Ours4 & Ours8 & Ours16 & Ours32\\
    \midrule
    \begin{tabular}[c]{@{}c@{}}Category-\\ Level\end{tabular} & 
    \begin{tabular}[c]{@{}c@{}}65\% \\ $\pm
    2\%$\end{tabular} & 
    \begin{tabular}[c]{@{}c@{}}96.9\% \\ $\pm 0.7\%$\end{tabular} & 
    
    \begin{tabular}[c]{@{}c@{}} 36\%    \\ $\pm 3\%$\end{tabular} & 
    \begin{tabular}[c]{@{}c@{}} 87\% \\ $\pm 2\%$\end{tabular} & 
    \begin{tabular}[c]{@{}c@{}} 97.9\% \\ $\pm 0.8\%$\end{tabular} & 
    \begin{tabular}[c]{@{}c@{}} 99.3\% \\ $\pm 0.5\%$\end{tabular} \\
    \midrule
    
    \begin{tabular}[c]{@{}c@{}}Instance-\\ Level\end{tabular} & 
    \begin{tabular}[c]{@{}c@{}}65\% \\ $\pm 2\%$\end{tabular} & 
    \begin{tabular}[c]{@{}c@{}}99.1\% \\ $\pm 0.4\%$\end{tabular} & 
    
    \begin{tabular}[c]{@{}c@{}}72\% \\ $\pm 3\%$\end{tabular} & 
    \begin{tabular}[c]{@{}c@{}}95\% \\ $\pm 1\%$\end{tabular} & 
    \begin{tabular}[c]{@{}c@{}}96\% \\ $\pm 1\%$\end{tabular} & 
    \begin{tabular}[c]{@{}c@{}}97\% \\ $\pm 1\%$\end{tabular} \\

  \bottomrule
\end{tabular}

\end{table}

\begin{figure}
    \centering
    \includegraphics[width=0.9\linewidth]{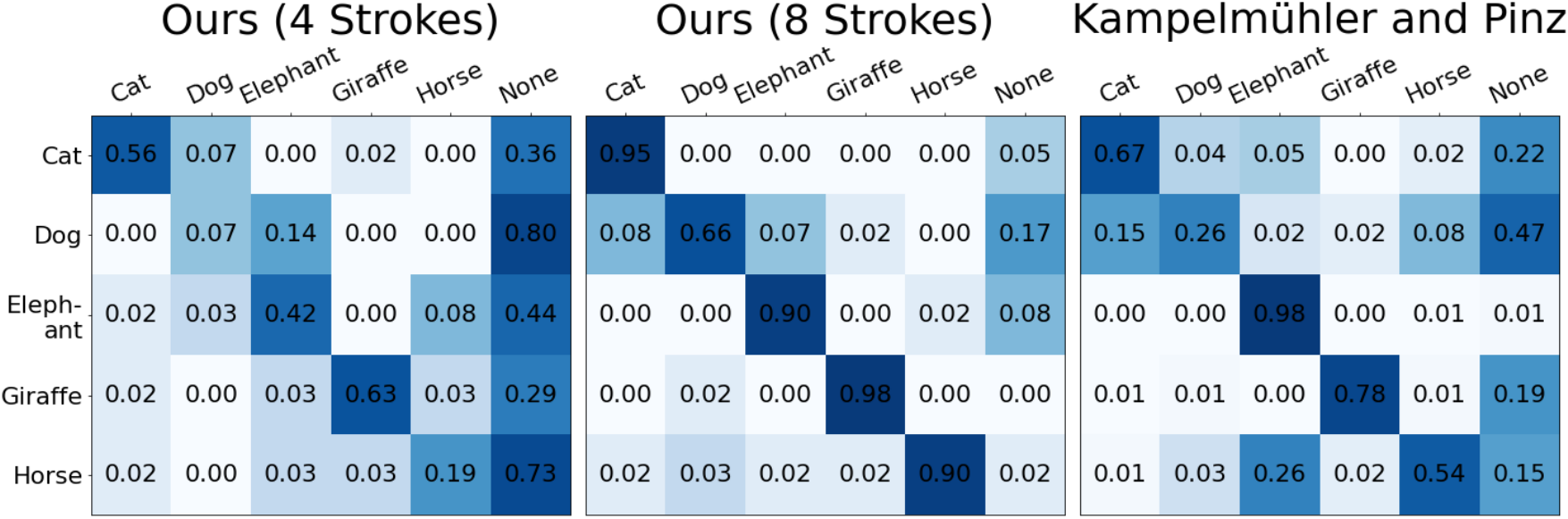}
    \caption{\small Confusion matrices of category-level recognition of the perceptual study of our method with four and eight strokes (left and middle matrices) and the method of Kampelmühler and Pinz \cite{human-like-sketches}.}
    \label{fig:conf_mat}
\end{figure}

In Figure \ref{fig:conf_mat} we provide the confusion matrices for analyzing the sources of errors made in the category recognition task at the abstraction levels with higher error rates. Specifically, our sketches produced with 4 and 8 strokes, and those of Kampelmühler and Pinz \cite{human-like-sketches}. The three matrices show that the majority of the classification errors can be attributed to insufficient confidence (selecting `None'). Overall, the `dog' class achieved the lowest rates of correct answers. In the four strokes case dogs were mostly confused with elephants, whereas in the case of Kampelmühler and Pinz, the errors beside `None' resulted from a confusion with the cat class. The `horse' class achieved the second lowest scores. In the four strokes case, most of the errors were due to insufficient confidence, while in Kampelmühler and Pinz, besides the `None' answers, horses were mostly miscategorized as elephants.

\paragraph{Sketch classifier}
Two different classifiers are used for evaluating the synthesized sketches; a ResNet34 classifier from Kampelm{\"{u}}hler and Pinz \cite{human-like-sketches} trained on the Sketchy-Database \cite{Sketchy-Database} with 125 categories, and a CLIP ViT-B/32 zero-shot classifier using text prompts defined as "A sketch of a(n) \textit{class-name}". Note that this is not the CLIP model we use for training. 

Table \ref{tab:classification_res} compares the sketch-classifier recognition accuracy of our sketches to that of Kampelm{\"{u}}hler and Pinz \cite{human-like-sketches} and Li et al. \cite{li2019photosketching} based upon 200 randomly selected images of 10 categories from the SketchyCOCO dataset \cite{gao2020sketchycoco}.
The recognition accuracy on human sketches from the SketchyCOCO dataset is also calculated as a baseline.
The method by Kampelm{\"{u}}hler and Pinz achieves the highest scores for ResNet34 classifier, possibly since they use the same model and dataset during training.
Despite the distribution differences, our method still achieves good recognition rates under this classifier.
With CLIP classifier, our method achieves very high accuracy levels of 78\% with 16 strokes and 91\% with 32 strokes. For more details and analysis please refer to the supplementary material.

\begin{table}[h]
\caption{\small Top-1 and Top-3 sketch recognition accuracy computed with ResNet34 and CLIP ViT-B/32 on 200 sketches from 10 categories. (A) Kampelmühler and Pinz \cite{human-like-sketches}, (B) Li et al. 
\cite{li2019photosketching}}
\label{tab:classification_res}
\centering
\begin{tabular}{@{\hskip3pt}c@{\hskip3pt}c|ccccc}
\toprule
                        Classifier  &      & \begin{tabular}[c]{@{}c@{}}Human \\ Sketches\end{tabular} & A & B & Ours16              & Ours32 \\ 
\hline
\multirow{2}{*}{ResNet34} & Top1 & 98\%  & \textbf{\textbf{67\%}}  & 61\%   & 54\%  &  63\%  \\
                          & Top3 & 99\%  & \textbf{\textbf{82\%}}  & 78\%   & 75\%  &  77\%  \\ 
\hline
\multirow{2}{*}{CLIP ViT-B/32}  & Top1 & 75\%  & 49\%  & 60\%  & \textbf{\textbf{78\%}} & 91\%  \\
                                & Top3 & 93\%  & 65\%  & 77\%  & \textbf{\textbf{93\%}} & 97\%  \\
\hline
\end{tabular}

\end{table}

\paragraph{Sketch diversity}
We conduct an analysis to quantify the diversity of the sketches produced at different levels of abstraction. The input images for this study are identical to those used in the perceptual study (i.e., five random samples from five different animals classes). In total, we produce 1000 sketches – ten sketches per image, for each level of abstraction, using 10 different initializations.
We measure the diversity of each set of sketches $\{\mathcal{S}_{i}\}_{i=1}^{n}, n=10$ derived from the same image and at the same abstraction level $t$ as the normalized average variance of the sketches:

\begin{equation}
\label{eq:diversity}
    D_{t} = \left\Vert\frac{1}{\left\Vert\mu_{s}\right\Vert_{1}\cdot n} \sum_{i=1}^{n} (\mathcal{S}_{i} - \mu_{s})^2\right\Vert_{1}
\end{equation}

\begin{figure}[h]
\centering
 \setlength{\belowcaptionskip}{-2.5pt}
    \setlength{\tabcolsep}{1pt}
    \begin{tabular}{@{\hskip1pt}c@{\hskip1pt}c@{\hskip1pt}c@{\hskip1pt}c@{\hskip3pt}|c@{\hskip1pt}c@{\hskip1pt}c@{\hskip1pt}c}
    
        \raisebox{0.16in}{\multirow{2}{*}{\raisebox{100pt}{\includegraphics[width=0.18\linewidth]{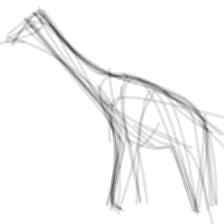}}}} & 
        \includegraphics[width=\widthgiraffe\linewidth]{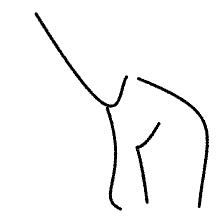} & 
        \includegraphics[width=\widthgiraffe\linewidth]{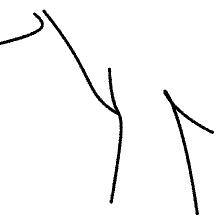} & 
        \includegraphics[width=\widthgiraffe\linewidth]{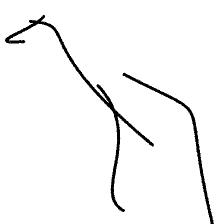} &
        \raisebox{0.16in}{\multirow{2}{*}{\includegraphics[width=0.18\linewidth]{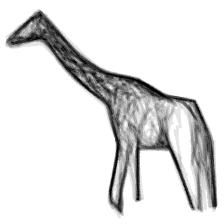}}} & 
         \includegraphics[width=\widthgiraffe\linewidth]{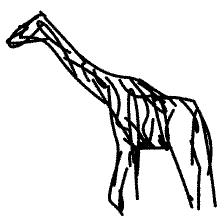} & 
        \includegraphics[width=\widthgiraffe\linewidth]{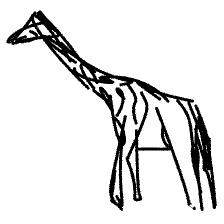} & 
        \includegraphics[width=\widthgiraffe\linewidth]{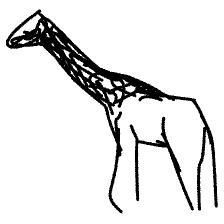}
        \\
        & 
        \includegraphics[width=\widthgiraffe\linewidth]{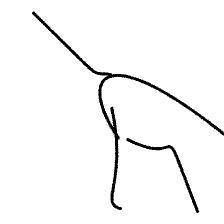} &  
        \includegraphics[width=\widthgiraffe\linewidth]{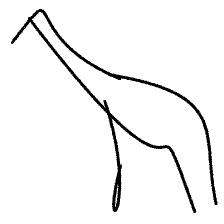} &  
        \includegraphics[width=\widthgiraffe\linewidth]{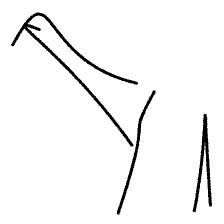} &
        
        & 
        \includegraphics[width=\widthgiraffe\linewidth]{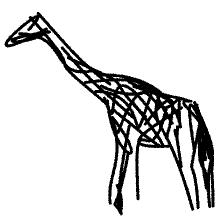} &  
        \includegraphics[width=\widthgiraffe\linewidth]{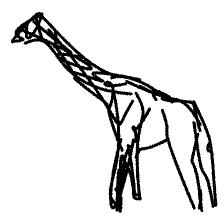} &  
        \includegraphics[width=\widthgiraffe\linewidth]{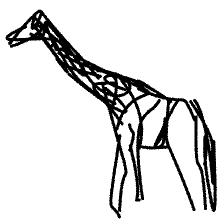} 
        \\
        \end{tabular}
    \caption{\small A visualisation of the appearance of the mean sketch $\mu_s$ (the larger giraffes) at two levels of abstraction, next to six samples of distinct sketches from the corresponding set.}
    \label{fig:diversity_animal_example}
\end{figure}

\begin{figure}[h]
    \begin{subfigure}[t]{0.49\linewidth}
        \includegraphics[width=\linewidth]{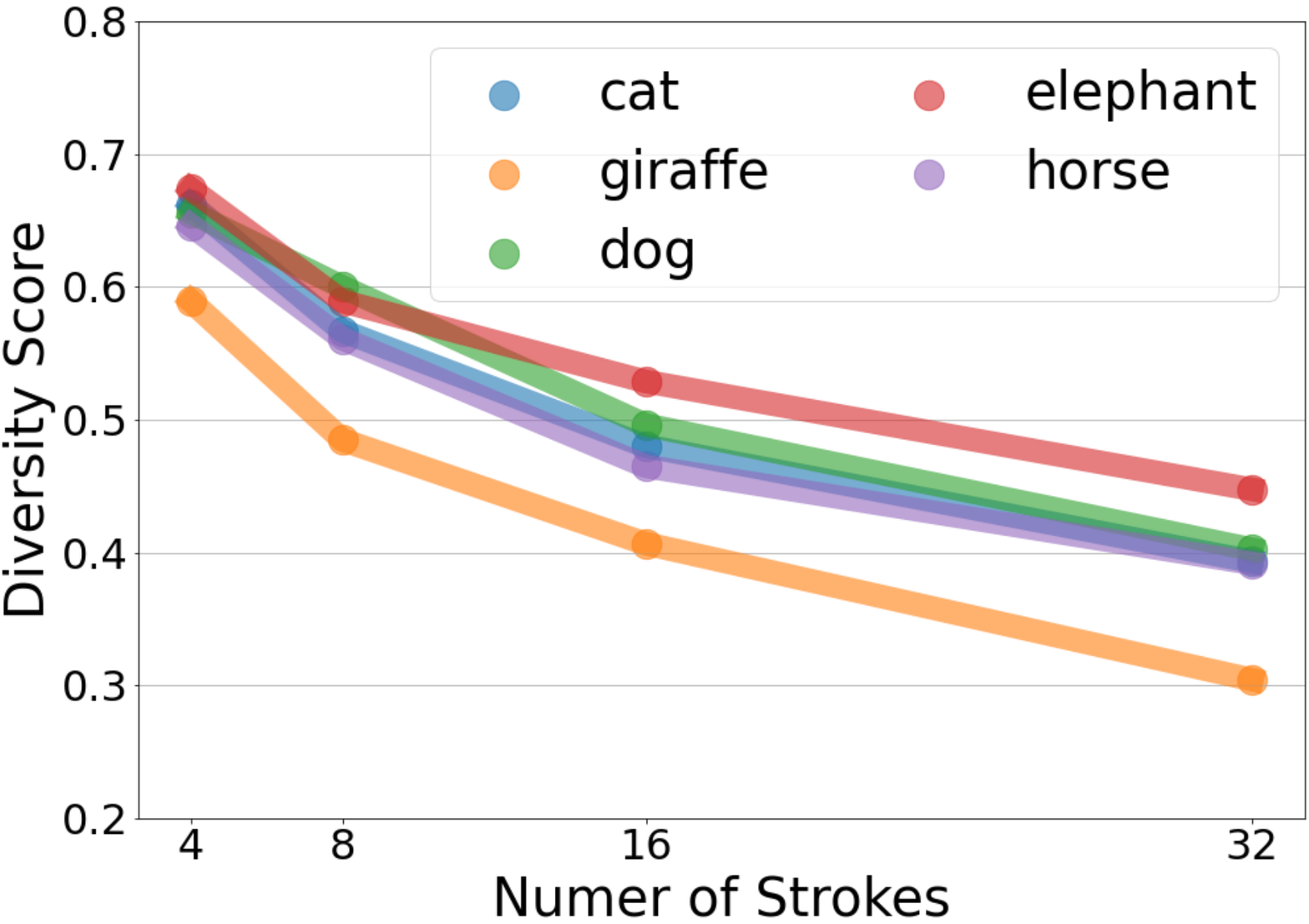}
         \subcaption{Our method's diversity score increases with the level of abstraction.}\label{fig:animals_diversity_res}
    \end{subfigure}\hspace{2mm}%
    \begin{subfigure}[t]{0.47\linewidth}
        \includegraphics[width=1\linewidth]{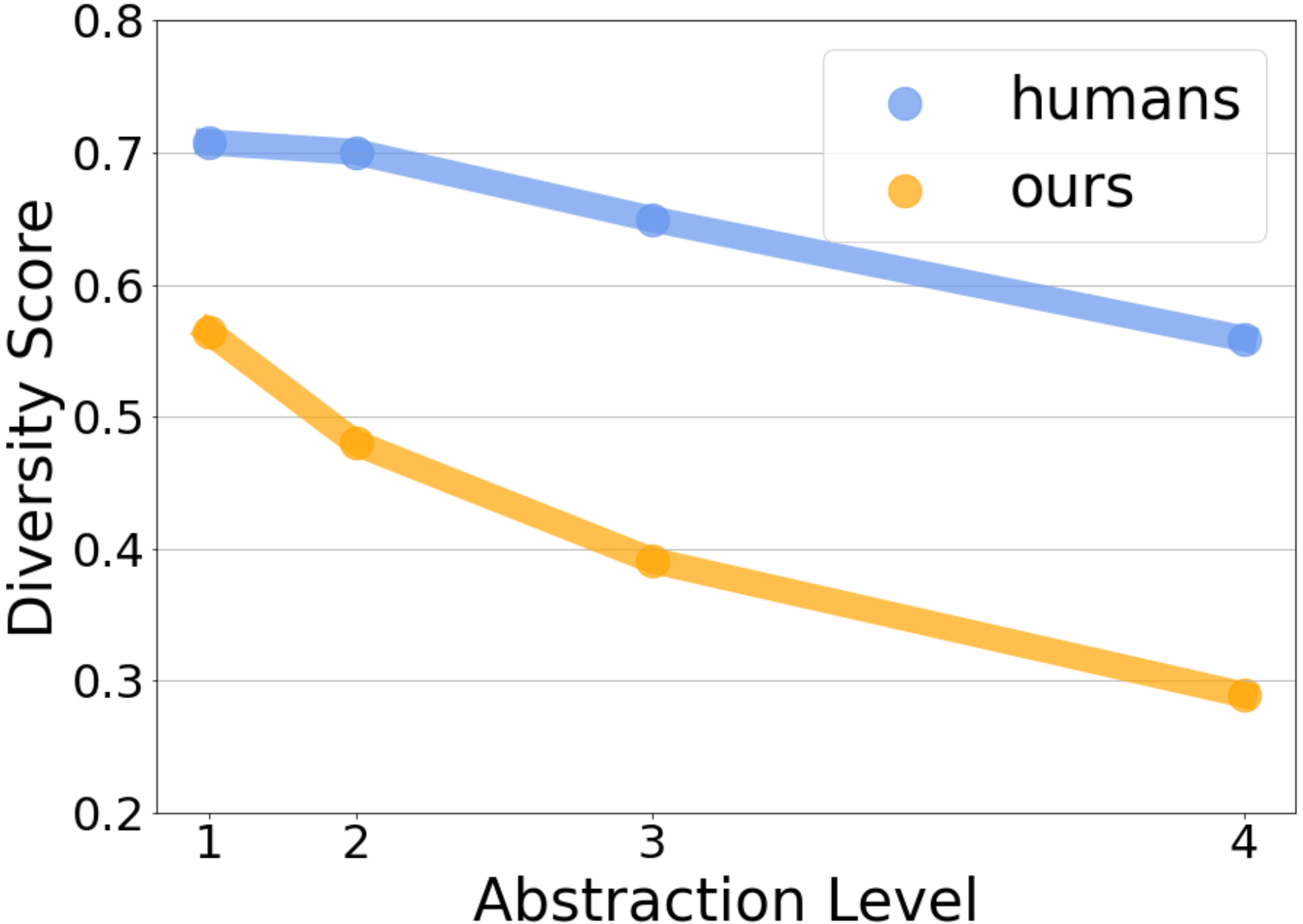}
         \subcaption{Diversity of human sketches produced by different artists is larger than the diversity of our method in all levels of abstraction.}\label{fig:faces_div_res}
    \end{subfigure}
 \caption{\small Diversity score as a function of the abstraction level.}
\label{fig:diversity_res}
\end{figure}

Where $\mu_{s}=\frac{1}{n}\sum_{i=1}^{n}\mathcal{S}_{i}$ is the pixel-wise mean sketch. To be able to compare different abstraction levels, we normalize the variance with respect to the magnitude of the average sketch $||\mu_{s}||_{1}$ at a specific level of abstraction. This normalization is needed because as the level of abstraction increases, the number of strokes decrease so the average magnitude is smaller. 
Figure \ref{fig:diversity_animal_example} shows an example of a few sketches from the highest and lowest levels of abstraction along with the mean sketch at each level. Figure \ref{fig:animals_diversity_res} shows the average diversity score for each class and level of abstraction. We can see that the diversity increases with the abstraction level and that similar patterns were observed among different classes.

We also compare the diversity of the sketches generated by our method to the ones drawn by humans. For this study, we use the dataset collected by Berger et al. \cite{Berger2013}, which contains 672 portraits sketches drawn by 7 different artists at 4 levels of abstraction, of 24 faces from the face database of the Center for Vital Longevity \cite{Minear2004ALD}. The abstractions were created by limiting the amount of time available for the artists to produce the drawings, which is a common exercise in drawing classes, and therefore, a more natural way for people to produce abstractions. To compare visually similar level of abstraction, we use 8 strokes for the highest abstraction level and 64 for the lowest one. We generate 7 sketches using our method for each face and level of abstraction using 7 different seeds, each seed imitates a different artist in this case.

\begin{figure}[h!]
        \begin{tabular}{@{\hskip1pt}c@{\hskip1pt}c@{\hskip1pt}c@{\hskip1pt}c@{\hskip1pt}c@{\hskip1pt}c@{\hskip1pt}c@{\hskip1pt}c}
        \includegraphics[trim={1cm 0cm 1cm  0cm},clip,width=\widthface\linewidth]{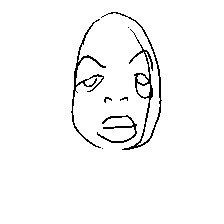} & 
        \includegraphics[trim={1cm 0cm 1cm  0cm},clip,width=\widthface\linewidth]{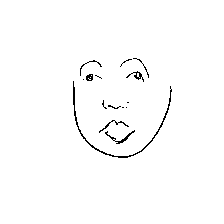} & 
        \includegraphics[trim={1cm 0cm 1cm  0cm},clip,width=\widthface\linewidth]{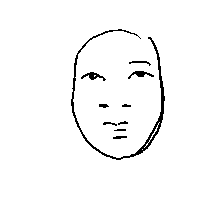} & 
        \includegraphics[trim={1cm 0cm 1cm  0cm},clip,width=\widthface\linewidth]{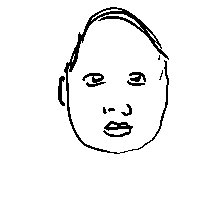} & 
        \includegraphics[trim={1cm 0cm 1cm  0cm},clip,width=\widthface\linewidth]{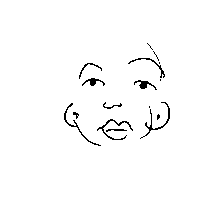} & 
        \includegraphics[trim={1cm 0cm 1cm  0cm},clip,width=\widthface\linewidth]{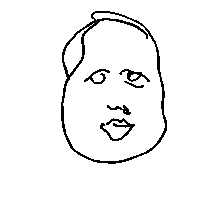} & 
        \includegraphics[trim={1cm 0cm 1cm  0cm},clip,width=\widthface\linewidth]{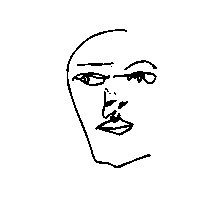} & 
        \includegraphics[trim={1cm 0cm 1cm  0cm},clip,width=\widthface\linewidth]{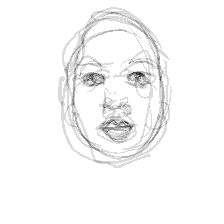}
        \\
        \includegraphics[trim={1cm 0cm 1cm  0cm},clip,width=\widthface\linewidth]{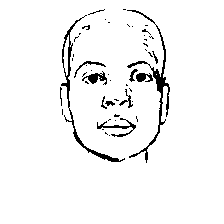} & 
        \includegraphics[trim={1cm 0cm 1cm  0cm},clip,width=\widthface\linewidth]{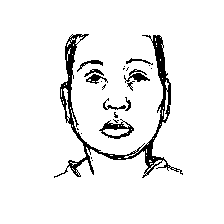} & 
        \includegraphics[trim={1cm 0cm 1cm  0cm},clip,width=\widthface\linewidth]{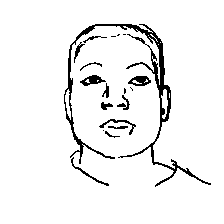} & 
        \includegraphics[trim={1cm 0cm 1cm  0cm},clip,width=\widthface\linewidth]{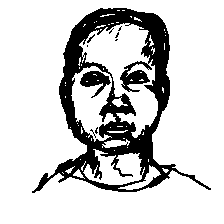} & 
        \includegraphics[trim={1cm 0cm 1cm  0cm},clip,width=\widthface\linewidth]{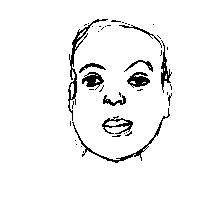} & 
        \includegraphics[trim={1cm 0cm 1cm  0cm},clip,width=\widthface\linewidth]{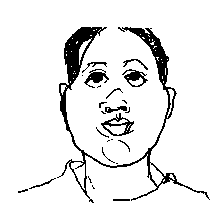} & 
        \includegraphics[trim={1cm 0cm 1cm  0cm},clip,width=\widthface\linewidth]{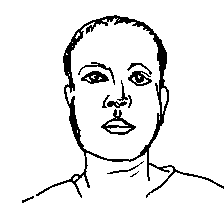} & 
        \includegraphics[trim={1cm 0cm 1cm  0cm},clip,width=\widthface\linewidth]{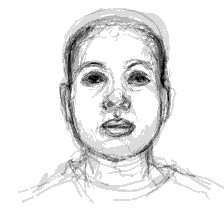}
        \\
        \midrule
        \includegraphics[width=\widthface\linewidth]{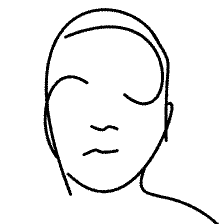} & 
        \includegraphics[width=\widthface\linewidth]{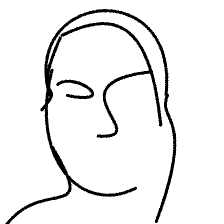} & 
        \includegraphics[width=\widthface\linewidth]{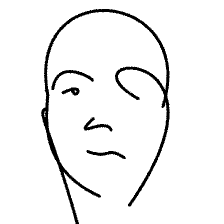} & 
        \includegraphics[width=\widthface\linewidth]{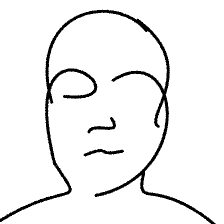} & 
        \includegraphics[width=\widthface\linewidth]{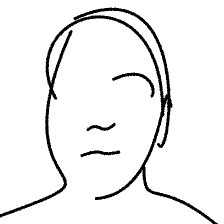} & 
        \includegraphics[width=\widthface\linewidth]{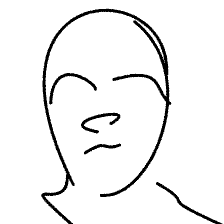} & 
        \includegraphics[width=\widthface\linewidth]{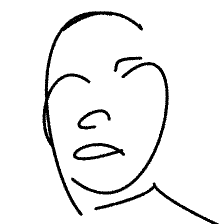} & 
        \includegraphics[width=\widthface\linewidth]{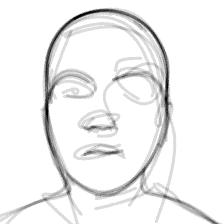}
        \\
        \includegraphics[width=\widthface\linewidth]{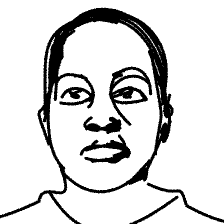} & 
        \includegraphics[width=\widthface\linewidth]{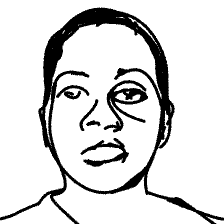} & 
        \includegraphics[width=\widthface\linewidth]{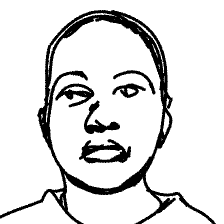} & 
        \includegraphics[width=\widthface\linewidth]{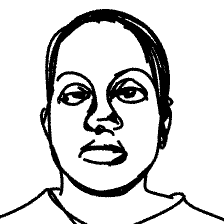} & 
        \includegraphics[width=\widthface\linewidth]{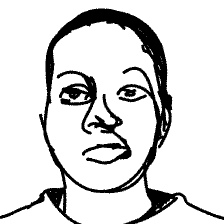} & 
        \includegraphics[width=\widthface\linewidth]{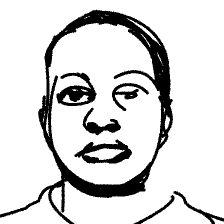} & 
        \includegraphics[width=\widthface\linewidth]{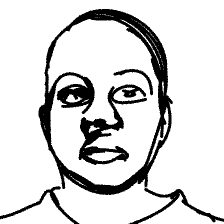} & 
        \includegraphics[width=\widthface\linewidth]{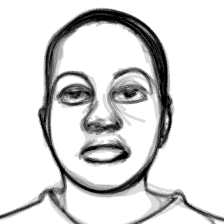}
        \\
        \end{tabular}
        \caption{\small An example of the sketches used in the diversity study. In each row, we show seven distinct sketches generated from a single input face, along with the average sketch that corresponds to this set in the rightmost column. The first two rows show the sketches drawn by the seven artists at 15 seconds in the first row and 270 seconds in the second row. The third and fourth rows show the sketches produced by our method with 8 and 64 strokes respectively. \copyright The portraits in the first two rows are from Berger et al. \cite{Berger2013}, used with permission.}
        \label{fig:face_example}
\label{fig:diversity_face}
\end{figure}

Figure \ref{fig:face_example} illustrates examples of the sketches produced by the seven artists as well as examples of the sketches generated by our method on a single input face at the lowest and highest abstraction levels. As can be seen, human sketches are more varied in style and semantic choices, which is understandable considering that they were created by different people, while our sketches appear to be less diverse (but still distinct). This visual observation is also expressed through our diversity measure presented in Figure \ref{fig:faces_div_res}, showing the diversity score of sketches produced by the artists (blue) is larger than our method (orange). However, as can be seen, the graphs follow a similar pattern (diversity decreases with the level of abstraction).

Finally, we show outputs from a large set of diverse input classes in Figure \ref{fig:diverse_sketchy}. We present 108 sketches of different classes sampled randomly from the SketchyDatabase dataset \cite{Sketchy-Database}. We also present sketches of 100 random cats from the SketchyCOCO dataset \cite{gao2020sketchycoco} in Figure \ref{fig:many_from_same_class}. We use a default number of 16 strokes to produce the sketches presented in both figures.

\section{Limitations and Future Work}
For images with background, our method's performance is reduced at higher abstraction levels.
This limitation can be addressed by using an automatic mask.
However, a potential future development would be to include such a remedial term within the loss function.
In addition, our sketches are not created sequentially and all strokes are optimized simultaneously, which differs from the conventional way of sketching.
Furthermore, the number of strokes must be determined in advance to achieve the desired level of abstraction. Another possible extension could be to make this a learned parameter, as different images may require different numbers of strokes to reach similar levels of abstraction.

Future potential uses of our method could include leveraging its ability to generalize to a variety of new categories in order to build datasets containing corresponding pairs of images and sketches that could be applied to the inverse problem as well.

\begin{figure}[h!]
    \begin{subfigure}[b]{0.5\linewidth}
        \begin{tabular}{@{\hskip2pt}c@{\hskip2pt}c@{\hskip2pt}c}
        Input & 4s & 16s \\
        \includegraphics[width=0.3\linewidth]{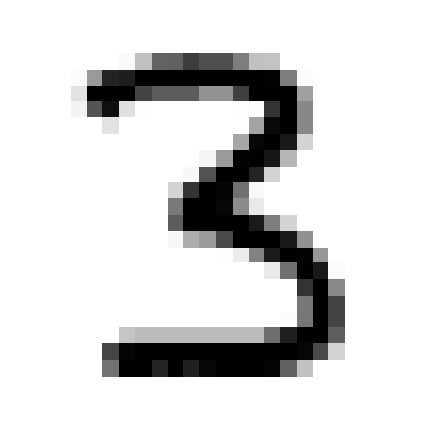} &
         \includegraphics[width=0.3\linewidth]{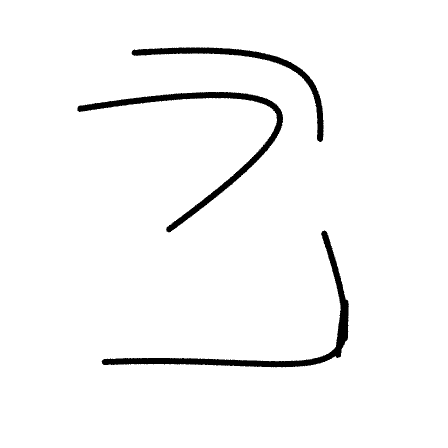} & 
         \includegraphics[width=0.3\linewidth]{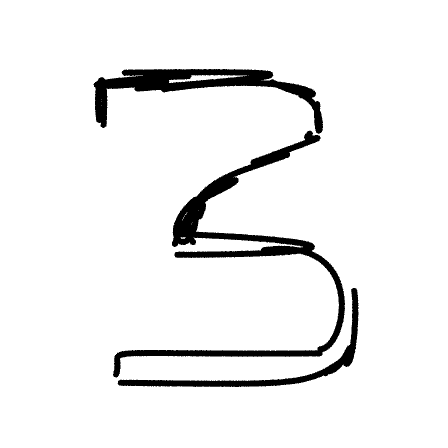}\\
        \end{tabular}
        \subcaption{}
        \label{fig:mnist}
    \end{subfigure}\hspace{4mm}%
    \begin{subfigure}[b]{0.4\linewidth}
    \centering
        \begin{tabular}{@{\hskip2pt}c@{\hskip2pt}c}
         Input & 16s  \\  
         \includegraphics[width=0.49\linewidth]{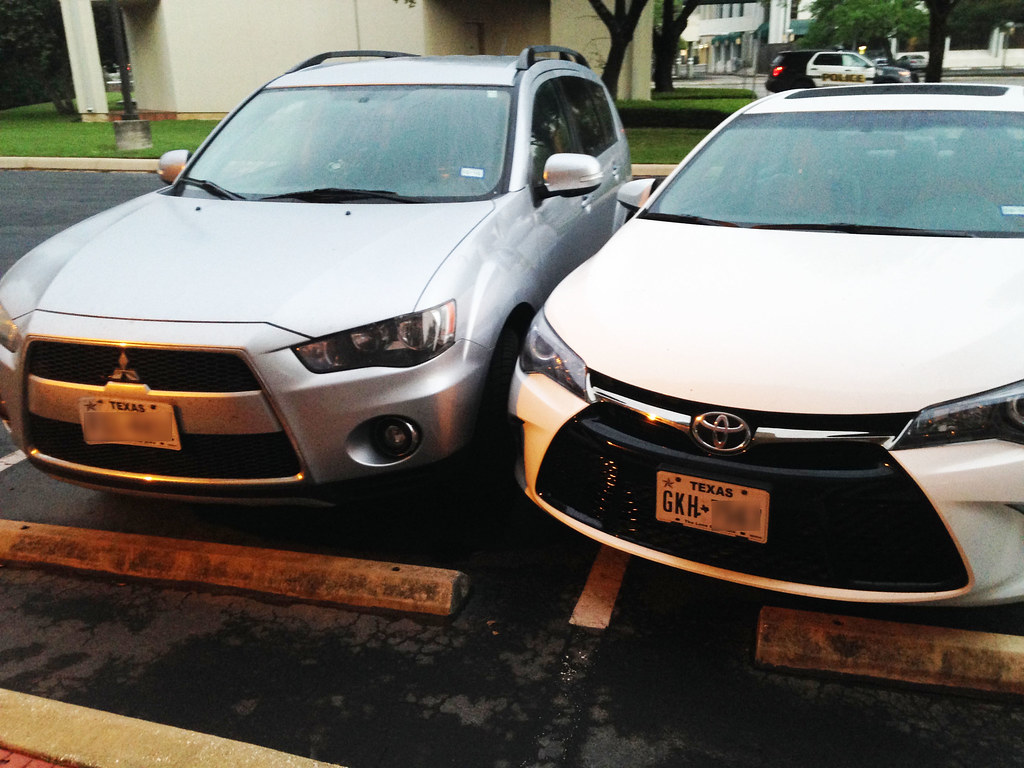} &
        \includegraphics[width=0.49\linewidth]{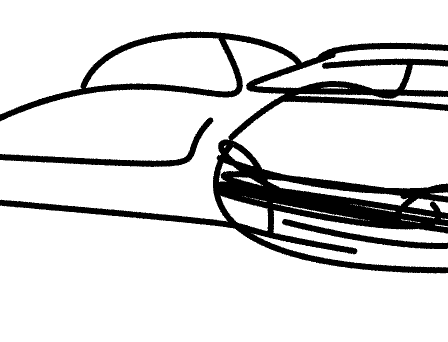} \\
        \end{tabular}
        \subcaption{}
        \label{fig:cars}
    \end{subfigure}
 \caption{\small Limitations inherited from CLIP. Figure (a) illustrates that the semantics of the input image (e.g. the digit three) are missing. This results in our method being unable to convey this meaning, which could potentially be expressed with only four strokes. Rather, as we can see, the strokes lie close to the edges of the digit. This conclusion also applies to the 16 stroke case. As shown in figure (b), CLIP's difficulty distinguishing fine-grain attributes is also present in our method. When semantics are considered, the two cars are distinguished by their brands, and this could be shown potentially by using a few strokes in the sketch; however, the optimization focuses on class-level depictions.}
\label{fig:limitations}
\end{figure}

\begin{figure}[ht]
\centering
\small
\begin{tabular}{@{\hskip2pt}c@{\hskip2pt}c@{\hskip2pt}c@{\hskip2pt}c@{\hskip2pt}c@{\hskip2pt}c@{\hskip2pt}c@{\hskip2pt}c@{\hskip2pt}c@{\hskip2pt}c}
zebra & rifle & couch & church & hat & sailboat & racket & songbird & shoe & knife\\
\includegraphics[width=\widthclasses\linewidth]{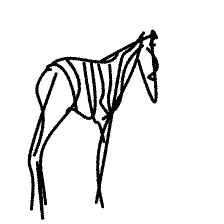} &
\includegraphics[width=\widthclasses\linewidth]{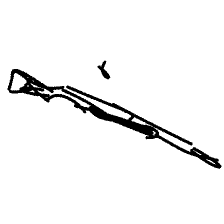} &
\includegraphics[width=\widthclasses\linewidth]{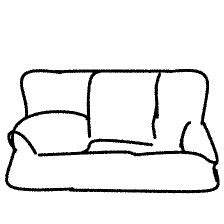} &
\includegraphics[width=\widthclasses\linewidth]{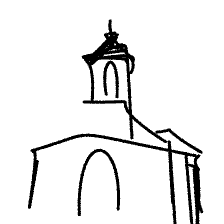} &
\includegraphics[width=\widthclasses\linewidth]{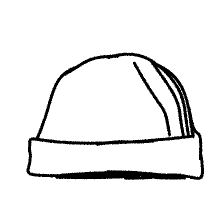} &
\includegraphics[width=\widthclasses\linewidth]{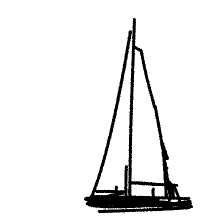} &
\includegraphics[width=\widthclasses\linewidth]{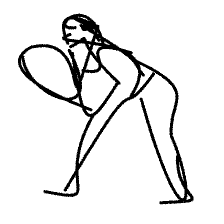} &
\includegraphics[width=\widthclasses\linewidth]{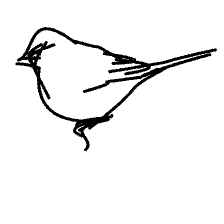} &
\includegraphics[width=\widthclasses\linewidth]{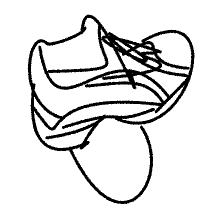} &
\includegraphics[width=\widthclasses\linewidth]{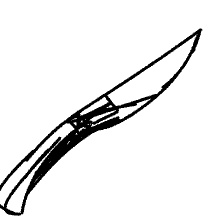}
\\
\begin{tabular}[c]{@{}c@{}}sea \\ turtle\end{tabular} & banana & tank & swan & mouse & snail & blimp & pizza & bench & sword \\
\includegraphics[width=\widthclasses\linewidth]{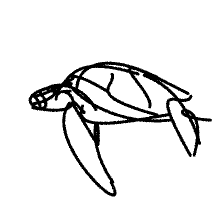} &
\includegraphics[width=\widthclasses\linewidth]{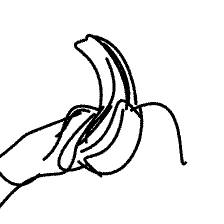} &
\includegraphics[width=\widthclasses\linewidth]{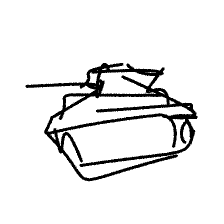} &
\includegraphics[width=\widthclasses\linewidth]{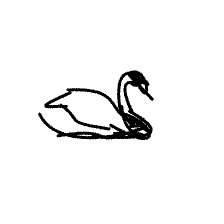} &
\includegraphics[width=\widthclasses\linewidth]{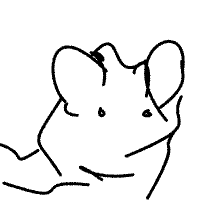} &
\includegraphics[width=\widthclasses\linewidth]{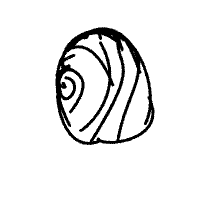} &
\includegraphics[width=\widthclasses\linewidth]{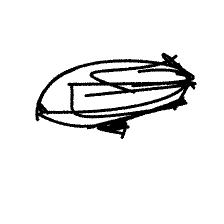} &
\includegraphics[width=\widthclasses\linewidth]{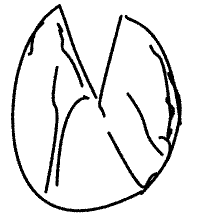} &
\includegraphics[width=\widthclasses\linewidth]{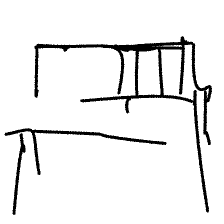} &
\includegraphics[width=\widthclasses\linewidth]{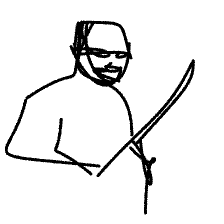}
\\
trumpet & camel & fish & horse & pear & cannon & bat & \begin{tabular}[c]{@{}c@{}}teddy \\ bear\end{tabular} & starfish & lion \\
\includegraphics[width=\widthclasses\linewidth]{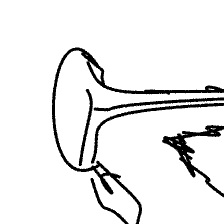} &
\includegraphics[width=\widthclasses\linewidth]{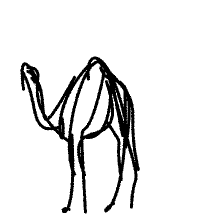} &
\includegraphics[width=\widthclasses\linewidth]{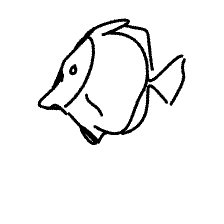} &
\includegraphics[width=\widthclasses\linewidth]{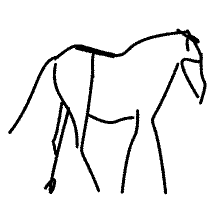} &
\includegraphics[width=\widthclasses\linewidth]{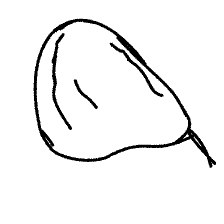} &
\includegraphics[width=\widthclasses\linewidth]{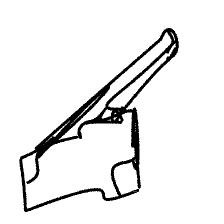} &
\includegraphics[width=\widthclasses\linewidth]{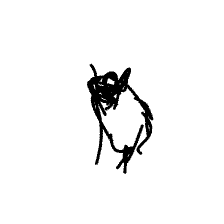} &
\includegraphics[width=\widthclasses\linewidth]{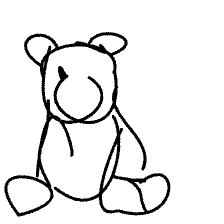} &
\includegraphics[width=\widthclasses\linewidth]{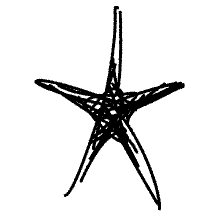} &
\includegraphics[width=\widthclasses\linewidth]{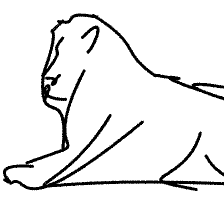}
\\
tree & window & frog & piano & lizard & bear & \begin{tabular}[c]{@{}c@{}}scor- \\ pion\end{tabular} & \begin{tabular}[c]{@{}c@{}}saxo- \\ phone\end{tabular} & door & \begin{tabular}[c]{@{}c@{}}alarm \\ clock\end{tabular} \\
\includegraphics[width=\widthclasses\linewidth]{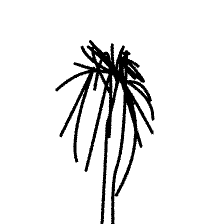} &
\includegraphics[width=\widthclasses\linewidth]{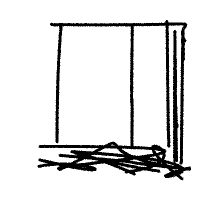} &
\includegraphics[width=\widthclasses\linewidth]{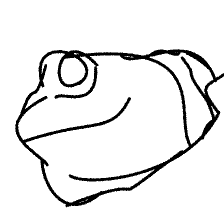} &
\includegraphics[width=\widthclasses\linewidth]{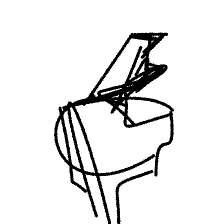} &
\includegraphics[width=\widthclasses\linewidth]{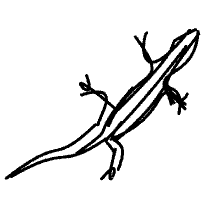} &
\includegraphics[width=\widthclasses\linewidth]{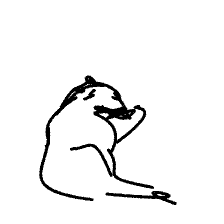} &
\includegraphics[width=\widthclasses\linewidth]{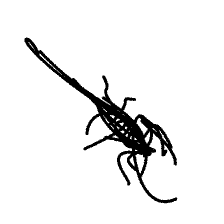} &
\includegraphics[width=\widthclasses\linewidth]{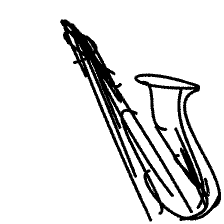} &
\includegraphics[width=\widthclasses\linewidth]{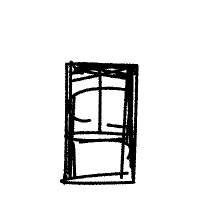} &
\includegraphics[width=\widthclasses\linewidth]{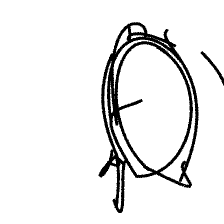}
\\
turtle & crab & \begin{tabular}[c]{@{}c@{}}straw- \\ berry\end{tabular} & spider & snake & flower & \begin{tabular}[c]{@{}c@{}}hour- \\ glass\end{tabular} & shark & pig & \begin{tabular}[c]{@{}c@{}}umbre- \\ lla\end{tabular}\\
\includegraphics[width=\widthclasses\linewidth]{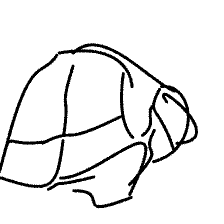} &
\includegraphics[width=\widthclasses\linewidth]{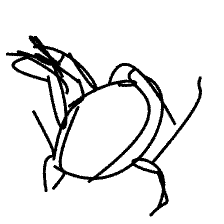} &
\includegraphics[width=\widthclasses\linewidth]{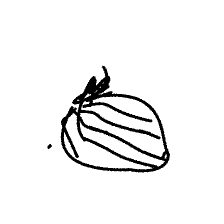} &
\includegraphics[width=\widthclasses\linewidth]{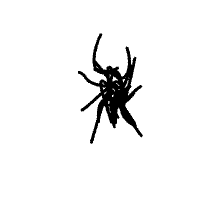} &
\includegraphics[width=\widthclasses\linewidth]{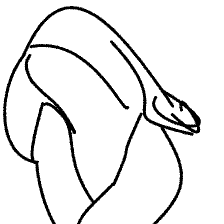} &
\includegraphics[width=\widthclasses\linewidth]{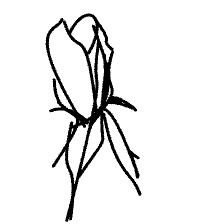} &
\includegraphics[width=\widthclasses\linewidth]{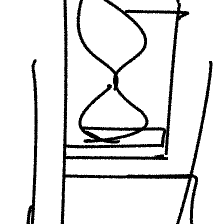} &
\includegraphics[width=\widthclasses\linewidth]{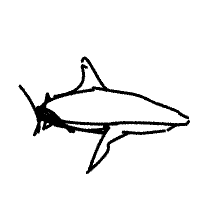} &
\includegraphics[width=\widthclasses\linewidth]{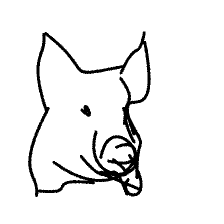} &
\includegraphics[width=\widthclasses\linewidth]{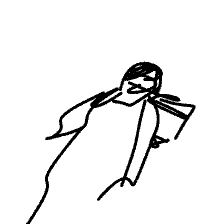}
\\
rocket & chicken & \begin{tabular}[c]{@{}c@{}}wading \\ bird\end{tabular} & apple & \begin{tabular}[c]{@{}c@{}}sky \\ scraper\end{tabular} & \begin{tabular}[c]{@{}c@{}}hot-air \\ balloon\end{tabular} & armor & \begin{tabular}[c]{@{}c@{}}jack-o- \\ lantern\end{tabular} & cow & table \\
\includegraphics[width=\widthclasses\linewidth]{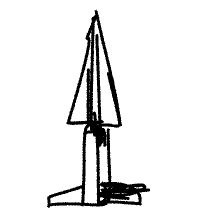} &
\includegraphics[width=\widthclasses\linewidth]{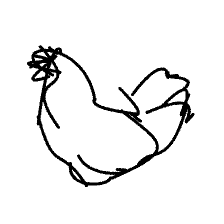} &
\includegraphics[width=\widthclasses\linewidth]{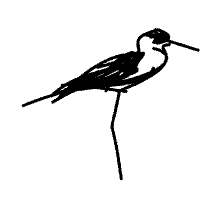} &
\includegraphics[width=\widthclasses\linewidth]{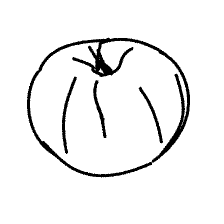} &
\includegraphics[width=\widthclasses\linewidth]{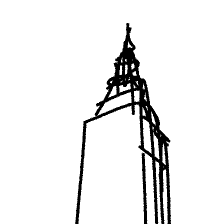} &
\includegraphics[width=\widthclasses\linewidth]{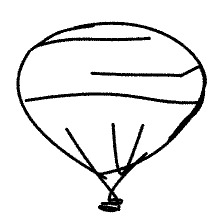} &
\includegraphics[width=\widthclasses\linewidth]{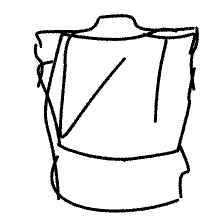} &
\includegraphics[width=\widthclasses\linewidth]{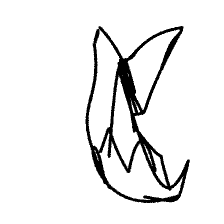} &
\includegraphics[width=\widthclasses\linewidth]{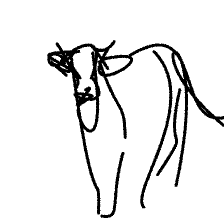} &
\includegraphics[width=\widthclasses\linewidth]{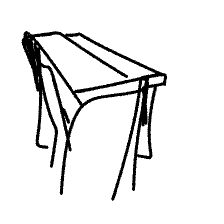}
\\
seagull & hotdog & \begin{tabular}[c]{@{}c@{}}eye- \\ glasses\end{tabular} & \begin{tabular}[c]{@{}c@{}}heli- \\ copter\end{tabular} & axe & \begin{tabular}[c]{@{}c@{}}mush- \\ room\end{tabular} & deer & bread & \begin{tabular}[c]{@{}c@{}}pine- \\ apple\end{tabular} & \begin{tabular}[c]{@{}c@{}}pickup \\ truck\end{tabular} \\
\includegraphics[width=\widthclasses\linewidth]{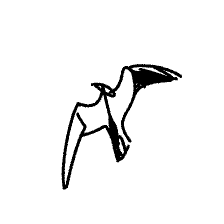} &
\includegraphics[width=\widthclasses\linewidth]{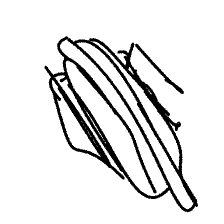} &
\includegraphics[width=\widthclasses\linewidth]{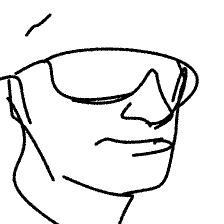} &
\includegraphics[width=\widthclasses\linewidth]{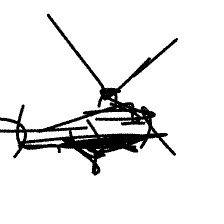} &
\includegraphics[width=\widthclasses\linewidth]{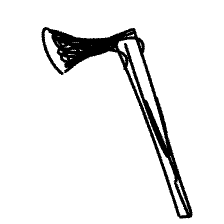} &
\includegraphics[width=\widthclasses\linewidth]{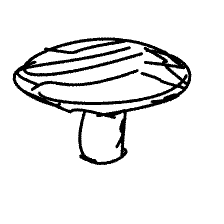} &
\includegraphics[width=\widthclasses\linewidth]{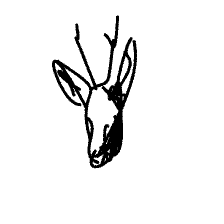} &
\includegraphics[width=\widthclasses\linewidth]{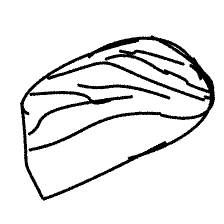} &
\includegraphics[width=\widthclasses\linewidth]{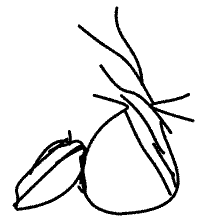} &
\includegraphics[width=\widthclasses\linewidth]{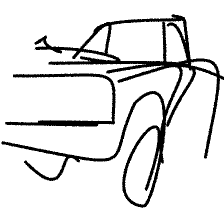}
\\
squirrel & fan & duck & \begin{tabular}[c]{@{}c@{}}racc- \\ oon\end{tabular} & spoon & pretzel & penguin & \begin{tabular}[c]{@{}c@{}}rhino- \\ ceros\end{tabular} & seal & rabbit \\
\includegraphics[width=\widthclasses\linewidth]{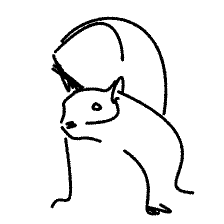} &
\includegraphics[width=\widthclasses\linewidth]{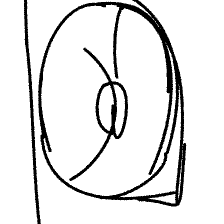} &
\includegraphics[width=\widthclasses\linewidth]{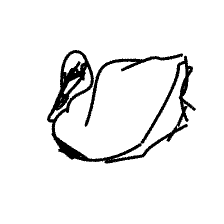} &
\includegraphics[width=\widthclasses\linewidth]{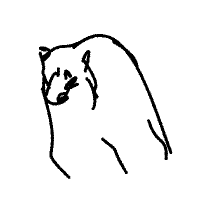} &
\includegraphics[width=\widthclasses\linewidth]{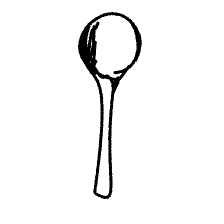} &
\includegraphics[width=\widthclasses\linewidth]{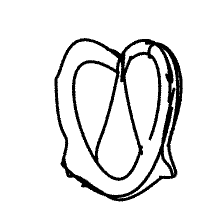} &
\includegraphics[width=\widthclasses\linewidth]{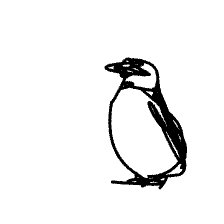} &
\includegraphics[width=\widthclasses\linewidth]{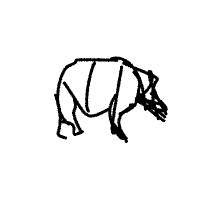} &
\includegraphics[width=\widthclasses\linewidth]{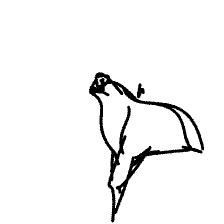} & 
\includegraphics[width=\widthclasses\linewidth]{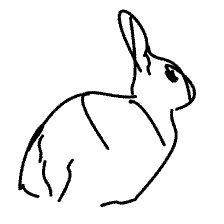}
\\

\begin{tabular}[c]{@{}c@{}}butter- \\ fly\end{tabular} & parrot & tiger & cup & pistol & \begin{tabular}[c]{@{}c@{}}kan- \\ garoo\end{tabular} & owl & scissors & chair & teapot \\
\includegraphics[width=\widthclasses\linewidth]{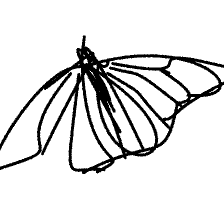} &
\includegraphics[width=\widthclasses\linewidth]{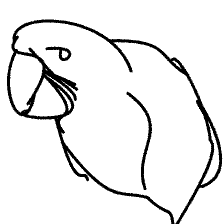} &
\includegraphics[width=\widthclasses\linewidth]{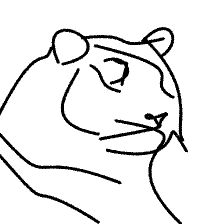} &
\includegraphics[width=\widthclasses\linewidth]{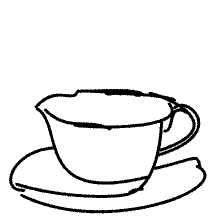} &
\includegraphics[width=\widthclasses\linewidth]{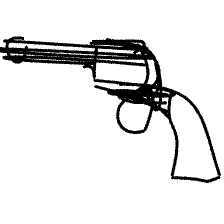} &
\includegraphics[width=\widthclasses\linewidth]{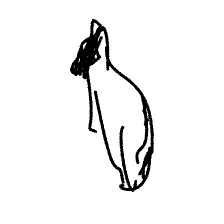} &
\includegraphics[width=\widthclasses\linewidth]{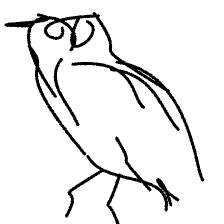} &
\includegraphics[width=\widthclasses\linewidth]{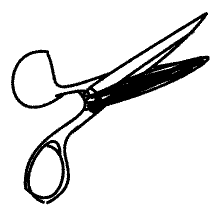} &
\includegraphics[width=\widthclasses\linewidth]{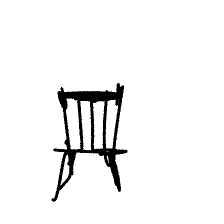} & 
\includegraphics[width=\widthclasses\linewidth]{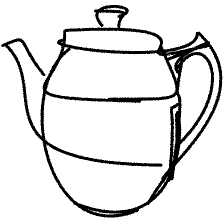}
\\

\end{tabular}
 \caption{Results of 90 random images from 90 classes from the SketchyDatabase \cite{Sketchy-Database}. The class name is presented on top of each sketch.}
\label{fig:diverse_sketchy}
\end{figure}

\begin{figure}[ht]
\centering
\begin{tabular}{@{\hskip2pt}c@{\hskip2pt}c@{\hskip2pt}c@{\hskip2pt}c@{\hskip2pt}c@{\hskip2pt}c@{\hskip2pt}c@{\hskip2pt}c@{\hskip2pt}c@{\hskip2pt}c}
   
    \includegraphics[width=\widthcats\linewidth]{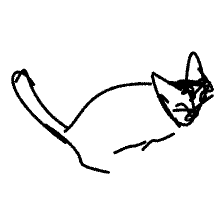} &
    \includegraphics[width=\widthcats\linewidth]{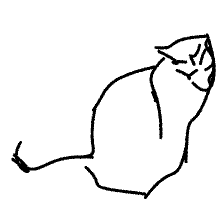} &
    \includegraphics[width=\widthcats\linewidth]{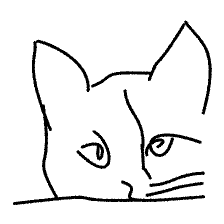} &
    \includegraphics[width=\widthcats\linewidth]{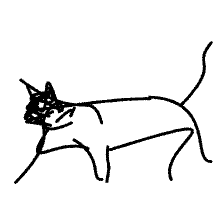} &
    \includegraphics[width=\widthcats\linewidth]{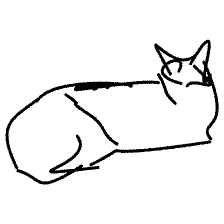} &
    \includegraphics[width=\widthcats\linewidth]{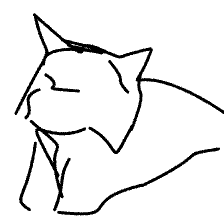} &
    \includegraphics[width=\widthcats\linewidth]{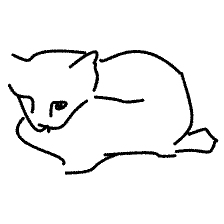} &
    \includegraphics[width=\widthcats\linewidth]{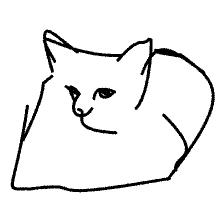} &
    \includegraphics[width=\widthcats\linewidth]{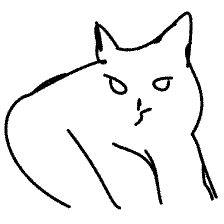} &
    \includegraphics[width=\widthcats\linewidth]{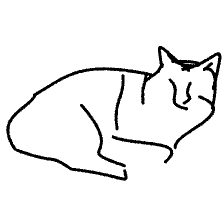} \\
    
    \includegraphics[width=\widthcats\linewidth]{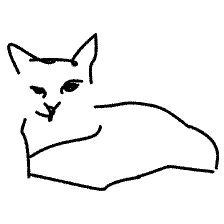} &
    \includegraphics[width=\widthcats\linewidth]{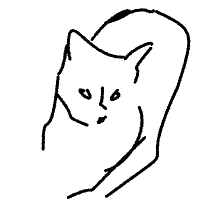} &
    \includegraphics[width=\widthcats\linewidth]{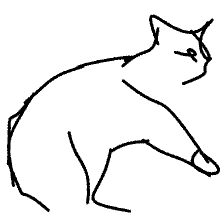} &
    \includegraphics[width=\widthcats\linewidth]{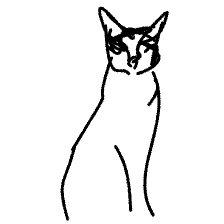} &
    \includegraphics[width=\widthcats\linewidth]{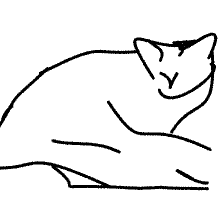} &
    \includegraphics[width=\widthcats\linewidth]{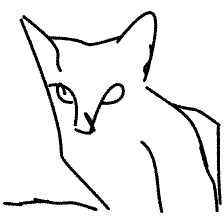} &
    \includegraphics[width=\widthcats\linewidth]{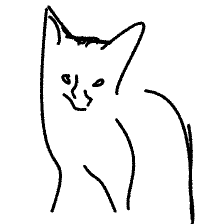} &
    \includegraphics[width=\widthcats\linewidth]{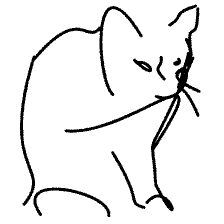} &
    \includegraphics[width=\widthcats\linewidth]{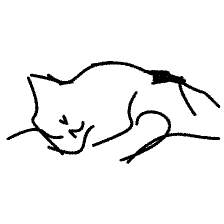} &
    \includegraphics[width=\widthcats\linewidth]{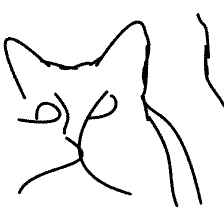} \\
    
    \includegraphics[width=\widthcats\linewidth]{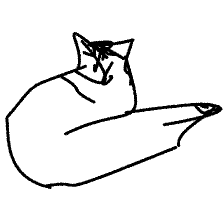} &
    \includegraphics[width=\widthcats\linewidth]{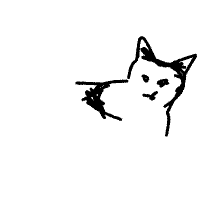} &
    \includegraphics[width=\widthcats\linewidth]{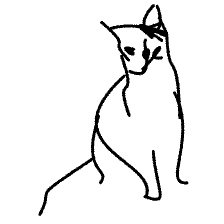} &
    \includegraphics[width=\widthcats\linewidth]{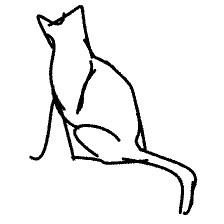} &
    \includegraphics[width=\widthcats\linewidth]{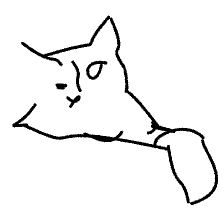} &
    \includegraphics[width=\widthcats\linewidth]{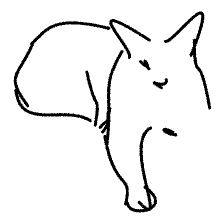} &
    \includegraphics[width=\widthcats\linewidth]{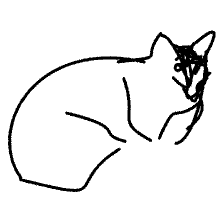} &
    \includegraphics[width=\widthcats\linewidth]{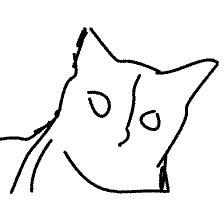} &
    \includegraphics[width=\widthcats\linewidth]{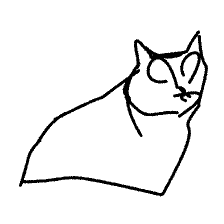} &
    \includegraphics[width=\widthcats\linewidth]{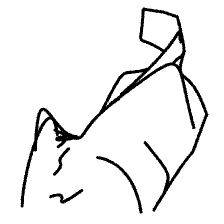}  \\
    
    \includegraphics[width=\widthcats\linewidth]{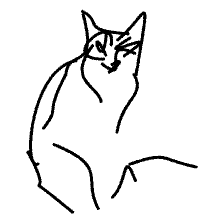} &
    \includegraphics[width=\widthcats\linewidth]{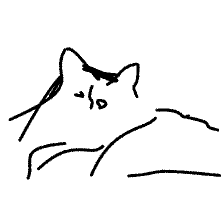} &
    \includegraphics[width=\widthcats\linewidth]{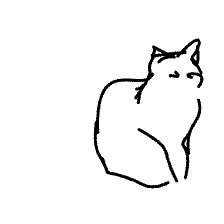} &
    \includegraphics[width=\widthcats\linewidth]{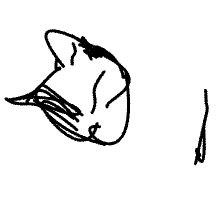} &
    \includegraphics[width=\widthcats\linewidth]{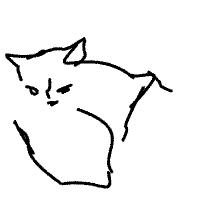} &
    \includegraphics[width=\widthcats\linewidth]{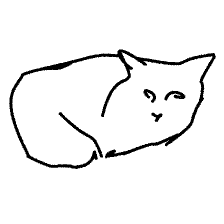} &
    \includegraphics[width=\widthcats\linewidth]{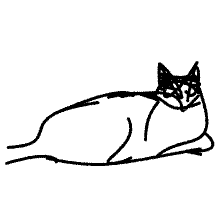} &
    \includegraphics[width=\widthcats\linewidth]{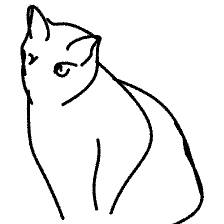} &
    \includegraphics[width=\widthcats\linewidth]{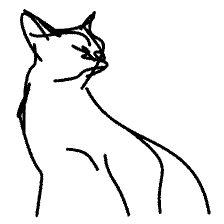} &
    \includegraphics[width=\widthcats\linewidth]{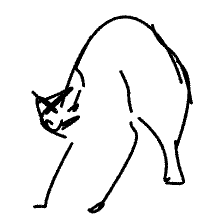} \tabularnewline

    \includegraphics[width=\widthcats\linewidth]{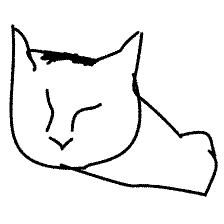} &
    \includegraphics[width=\widthcats\linewidth]{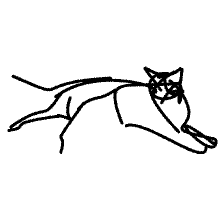} &
    \includegraphics[width=\widthcats\linewidth]{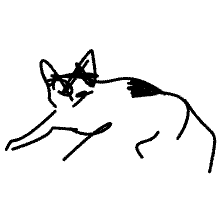} &
    \includegraphics[width=\widthcats\linewidth]{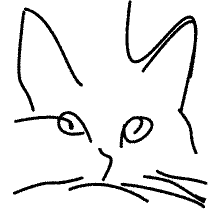} &
    \includegraphics[width=\widthcats\linewidth]{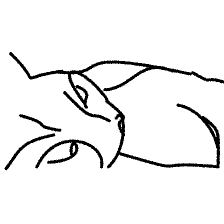} &
    \includegraphics[width=\widthcats\linewidth]{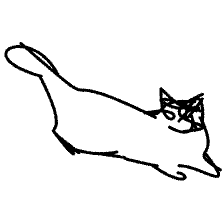} &
    \includegraphics[width=\widthcats\linewidth]{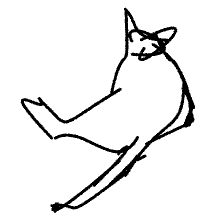} &
    \includegraphics[width=\widthcats\linewidth]{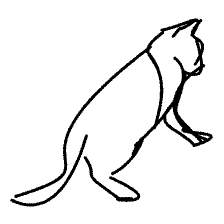} &
    \includegraphics[width=\widthcats\linewidth]{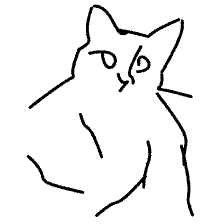} &
    \includegraphics[width=\widthcats\linewidth]{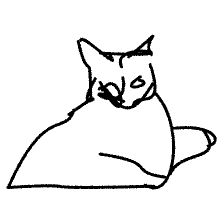} \tabularnewline
    
    \includegraphics[width=\widthcats\linewidth]{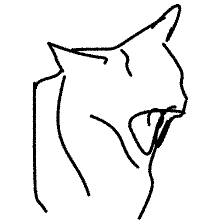} &
    \includegraphics[width=\widthcats\linewidth]{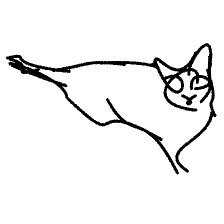} &
    \includegraphics[width=\widthcats\linewidth]{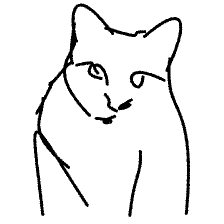} &
    \includegraphics[width=\widthcats\linewidth]{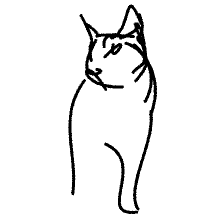} &
    \includegraphics[width=\widthcats\linewidth]{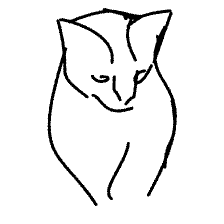} &
    \includegraphics[width=\widthcats\linewidth]{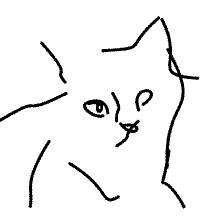} &
    \includegraphics[width=\widthcats\linewidth]{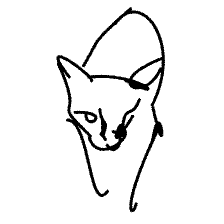} &
    \includegraphics[width=\widthcats\linewidth]{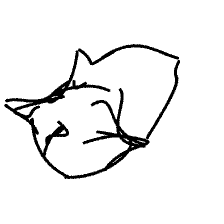} &
    \includegraphics[width=\widthcats\linewidth]{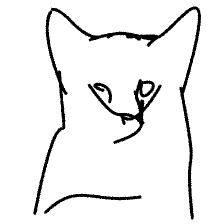} &
    \includegraphics[width=\widthcats\linewidth]{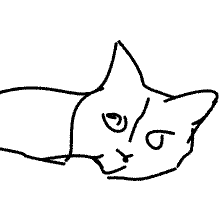} \\
    
    \includegraphics[width=\widthcats\linewidth]{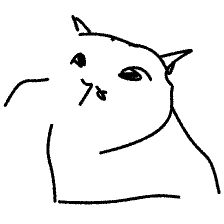} &
    \includegraphics[width=\widthcats\linewidth]{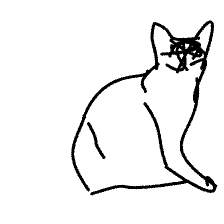} &
    \includegraphics[width=\widthcats\linewidth]{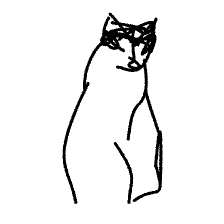} &
    \includegraphics[width=\widthcats\linewidth]{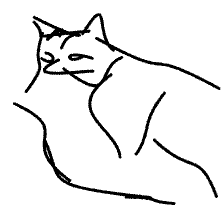} &
    \includegraphics[width=\widthcats\linewidth]{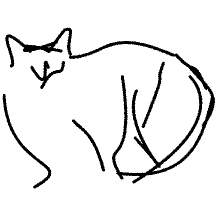} &
    \includegraphics[width=\widthcats\linewidth]{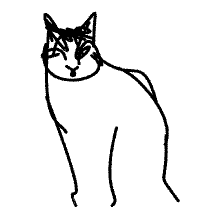} &
    \includegraphics[width=\widthcats\linewidth]{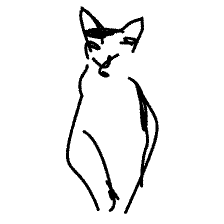} &
    \includegraphics[width=\widthcats\linewidth]{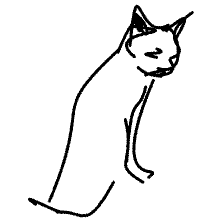} &
    \includegraphics[width=\widthcats\linewidth]{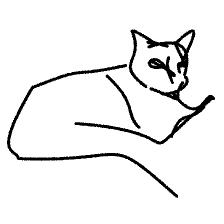} &
    \includegraphics[width=\widthcats\linewidth]{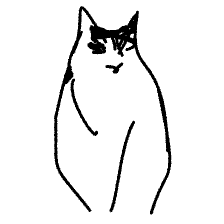}
    \\
    \includegraphics[width=\widthcats\linewidth]{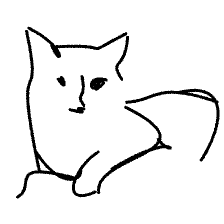} &
    \includegraphics[width=\widthcats\linewidth]{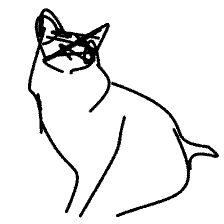} &
    \includegraphics[width=\widthcats\linewidth]{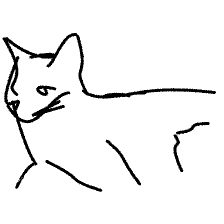} &
    \includegraphics[width=\widthcats\linewidth]{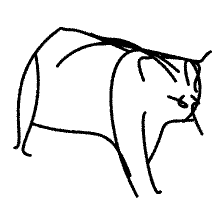} &
    \includegraphics[width=\widthcats\linewidth]{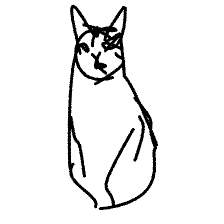} &
    \includegraphics[width=\widthcats\linewidth]{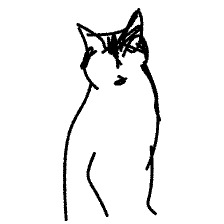} &
    \includegraphics[width=\widthcats\linewidth]{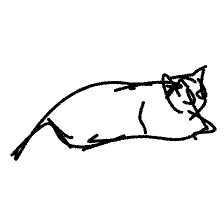} &
    \includegraphics[width=\widthcats\linewidth]{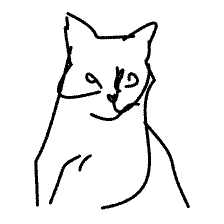} &
    \includegraphics[width=\widthcats\linewidth]{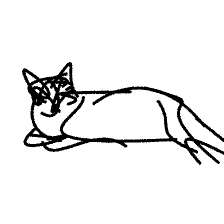} &
    \includegraphics[width=\widthcats\linewidth]{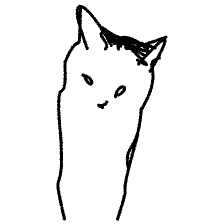}
    \\
    \includegraphics[width=\widthcats\linewidth]{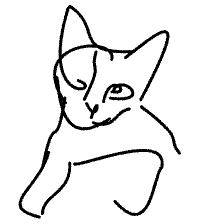} &
    \includegraphics[width=\widthcats\linewidth]{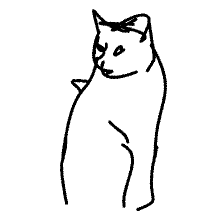} &
    \includegraphics[width=\widthcats\linewidth]{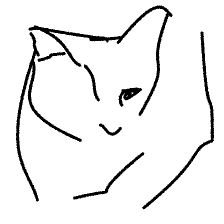} &
    \includegraphics[width=\widthcats\linewidth]{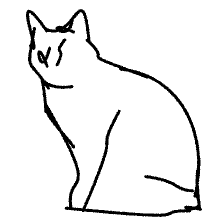} &
    \includegraphics[width=\widthcats\linewidth]{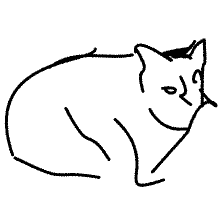} &
    
    \includegraphics[width=\widthcats\linewidth]{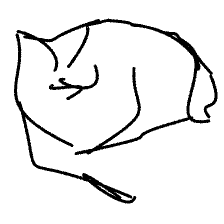} &
    \includegraphics[width=\widthcats\linewidth]{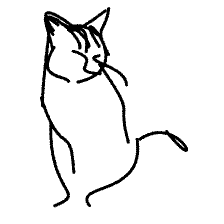} &
    \includegraphics[width=\widthcats\linewidth]{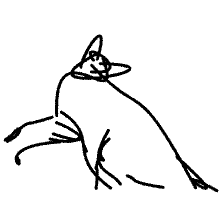} &
    \includegraphics[width=\widthcats\linewidth]{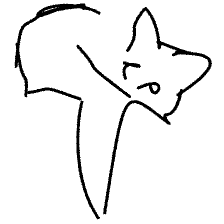} &
    \includegraphics[width=\widthcats\linewidth]{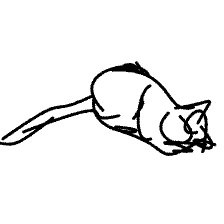}
    \\
    \includegraphics[width=\widthcats\linewidth]{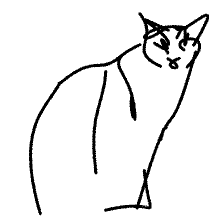} &
    \includegraphics[width=\widthcats\linewidth]{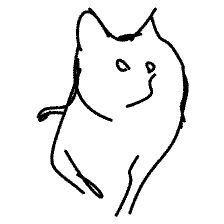} &
    \includegraphics[width=\widthcats\linewidth]{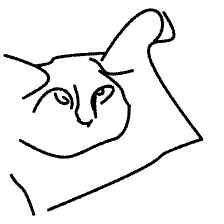} &
    \includegraphics[width=\widthcats\linewidth]{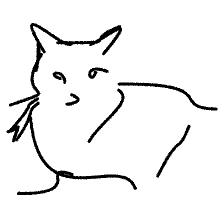} &
    \includegraphics[width=\widthcats\linewidth]{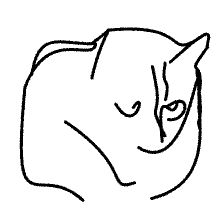} &
    
    \includegraphics[width=\widthcats\linewidth]{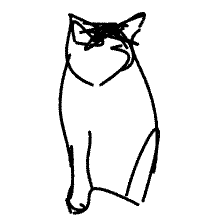} &
    \includegraphics[width=\widthcats\linewidth]{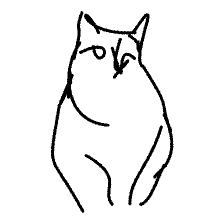} &
    \includegraphics[width=\widthcats\linewidth]{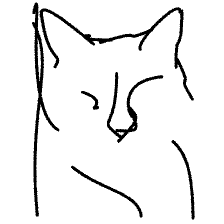} &
    \includegraphics[width=\widthcats\linewidth]{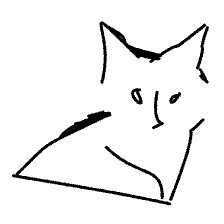} &
    \includegraphics[width=\widthcats\linewidth]{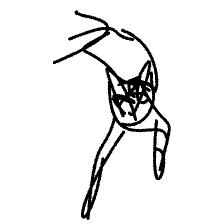} \\
    
\end{tabular}
 \caption{Sketching "in the wild": results of 100 random images of cats from SketchyCOCO \cite{gao2020sketchycoco}.}
\label{fig:many_from_same_class}
\end{figure}

Finally, as our framework is based on CLIP and its latent encoding, there are limitations of CLIP that carry over to our technique. As noted by the authors of CLIP, one example is CLIP's poor performance on a simple dataset such as MNIST \cite{LeCun2005TheMD}, which is caused by the lack of similar images in CLIP's training dataset. Figure \ref{fig:mnist} illustrates that our approach does in fact fail to draw such a simple subject, and the semantic gap may account for this failure. Another example reported in the CLIP paper is the poor performance on several types of fine-grained classification, such as distinguishing types of cars. Figure \ref{fig:cars} illustrates this point, as the abstraction focuses on painting the cars, whereas the semantics should focus on the brands of the cars.

\section{Conclusions}
We presented a method for photo-sketch synthesis, producing sketches with different levels of abstraction, without the need to train on specific sketch datasets.
Our method can generalize to various categories and cope with challenging levels of abstraction, while maintaining the semantic visual clues that allow for instance-level and class-level recognition.

\chapter{CLIPascene: Scene Sketching with Different Types and Levels of Abstraction}
\label{chap:clipascene}
 \begin{figure*}[h!]
     \centering
     \includegraphics[width=1\textwidth]{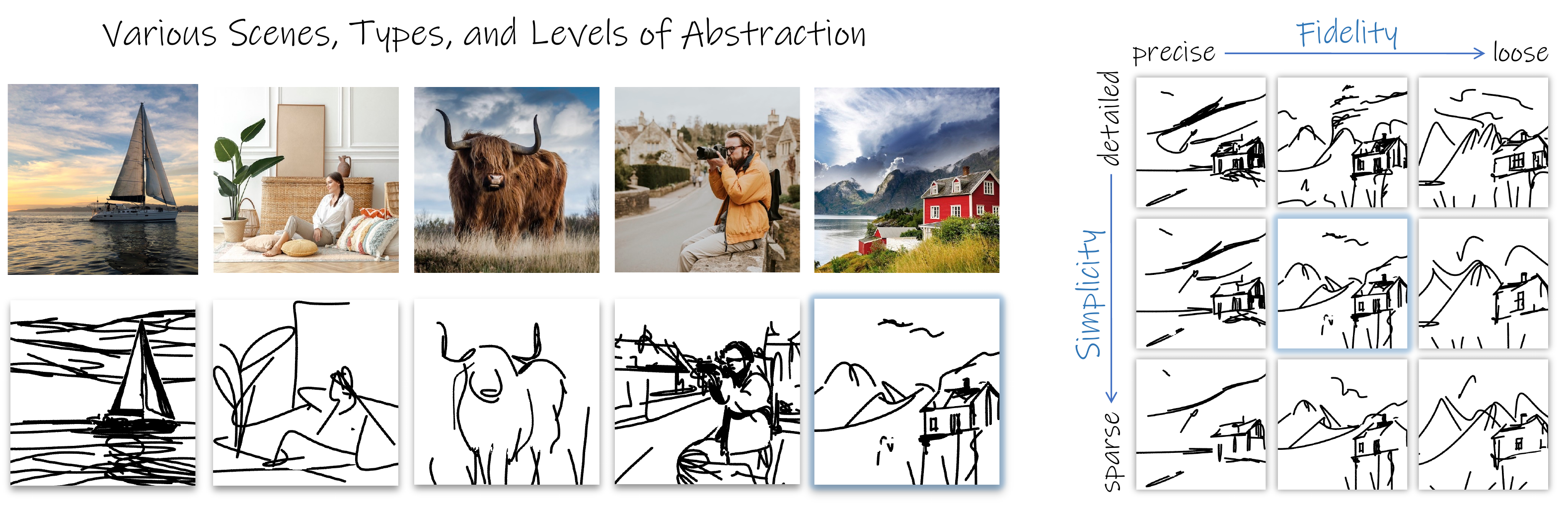}
     \caption[]{\small Our method converts a \textit{scene} image into a sketch  with different types and levels of abstraction by disentangling abstraction into two axes of control: \textit{fidelity} and \textit{simplicity}. 
        The sketches on the left were selected from a complete \textit{matrix} generated by our method (an example is shown on the right), encompassing a broad range of possible sketch abstractions for a given image. Our sketches are generated in vector form, which can be easily used by designers for further editing.\footnotemark}
     \label{fig:clipascene_teaser}
 \end{figure*}

\footnotetext{Project page: \url{https://clipascene.github.io/CLIPascene/}}
Several studies have demonstrated that abstract, minimal representations are not only visually pleasing but also helpful in conveying an idea more effectively by emphasizing the essence of the subject~\cite{Hertzmann_2020,biederman1988surface}. In this paper, we concentrate on converting photographs of natural scenes to sketches as a prominent minimal representation. 

\begin{figure}[t]
    \centering
    \setlength{\tabcolsep}{1.5pt}
    {\small
    \begin{tabular}{c c c}
        \includegraphics[width=0.2\linewidth]{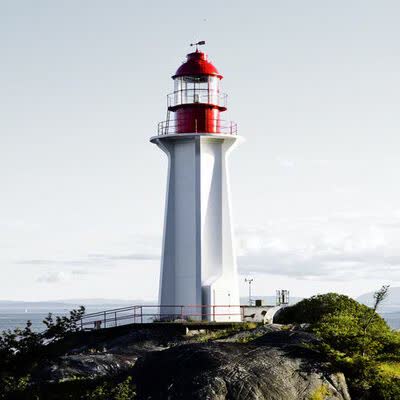} &
        \includegraphics[width=0.2\linewidth]{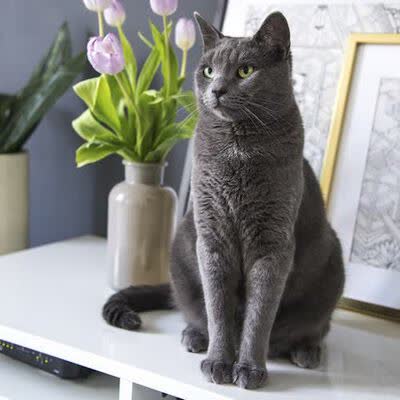} &
        \includegraphics[width=0.2\linewidth]{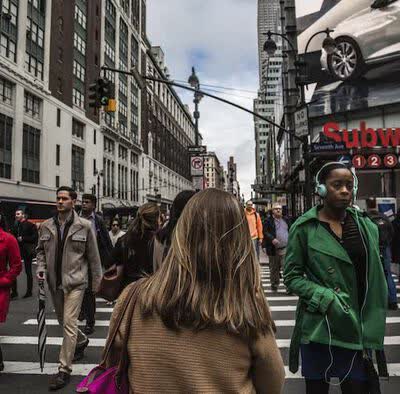} \\
        A & B & C
    \end{tabular}
    }
    \vspace{0.05cm}
    \caption{\small Scene complexity. (A) contains a single, central object with a simple background, (B) contains multiple objects (the cat and vase) with a slightly more complicated background, and (C) contains both foreground and background that include many details. Our work tackles all types of scenes.}
    \label{fig:im_complex}
\end{figure}

\begin{figure}
  \centering
  \includegraphics[width=0.9\linewidth]{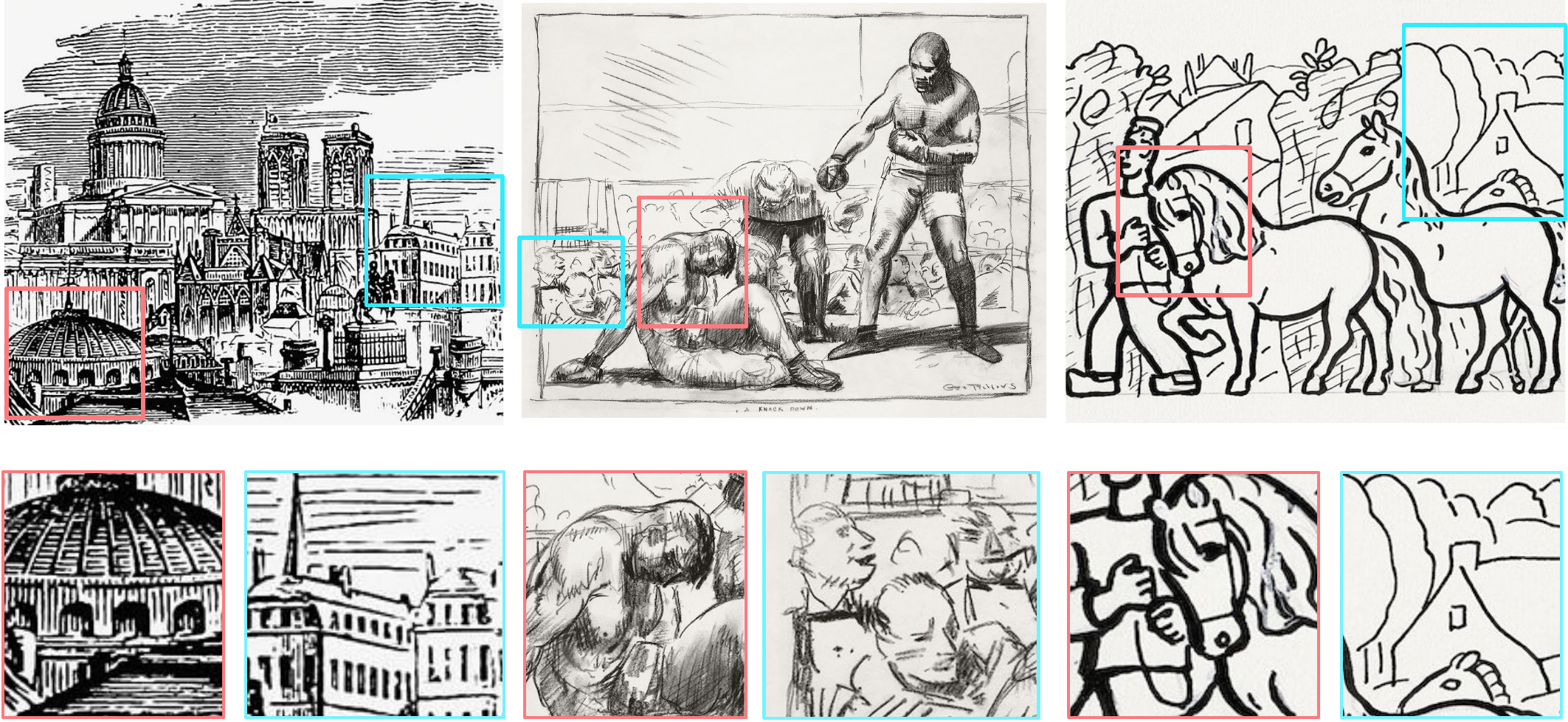}
  \caption{\small Drawings of different scenes by different artists. Notice the significant differences in style and level of abstraction between the drawings --- moving from more detailed and precise (left) to more abstract (right).
  The second row shows how the level of abstraction not only varies \textit{between} drawings, but also \textit{within} the \textit{same} drawing. Where each drawing contains areas that are relatively more detailed (red) and more abstract (blue).}
 \label{fig:drawing_style_lob}
\end{figure}

Converting a photograph to a sketch involves abstraction, which requires the ability to understand, analyze, and interpret the complexity of the visual scene. 
A scene consists of multiple objects of varying complexity, as well as relationships between the foreground and background (see~\Cref{fig:im_complex}). 
Therefore, when sketching a scene, the artist has many options regarding how to express the various components and the relations between them (see~\Cref{fig:drawing_style_lob}).

\newpage
In a similar manner, computational sketching methods must deal with scene complexity and consider a variety of abstraction levels. Our work focuses on the challenging task of \emph{scene} sketching while doing so using different \emph{types} and multiple \emph{levels} of abstraction. 
Only a few previous works attempted to produce sketches with multiple levels of abstraction.
However, these works focus specifically on the task of \emph{object} sketching~\cite{vinker2022clipasso,Deep-Sketch-Abstraction} or \emph{portrait} sketching~\cite{Berger2013}, and often simply use the number of strokes to define the level of abstraction. 
We are not aware of any previous work that attempts to separate different \emph{types} of abstractions.
Moreover, existing works for \emph{scene} sketching often focus on producing sketches based on certain styles, without taking into account the abstraction level, which is an essential concept in sketching. Lastly, most existing methods for scene sketching do not produce sketches in vector format. Providing vector-based sketches is a natural choice for sketches as it allows further editing by designers (such as in~\cref{fig:svg_editing}).

\begin{figure}[t]
  \centering
  \includegraphics[width=0.7\linewidth]{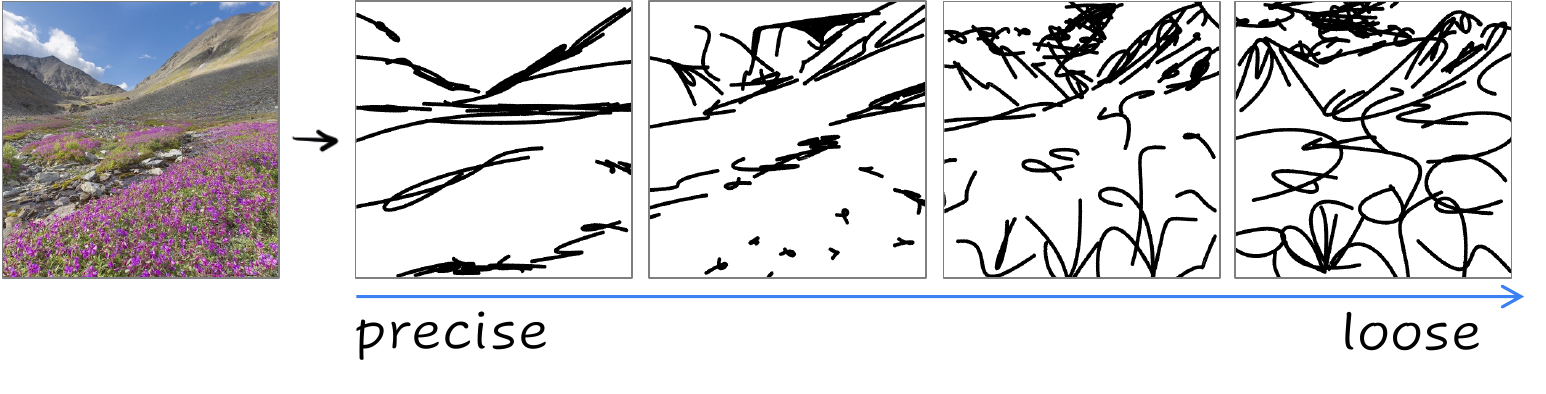}
  \caption{\small The \emph{fidelity} axis. From left to right, using the same number of strokes the sketches gradually depart from the geometry of the input image, but still convey the semantics of the scene.}
 \label{fig:semantic_axis}
\end{figure}

\begin{figure}[t]
  \centering
  \includegraphics[width=0.7\linewidth]{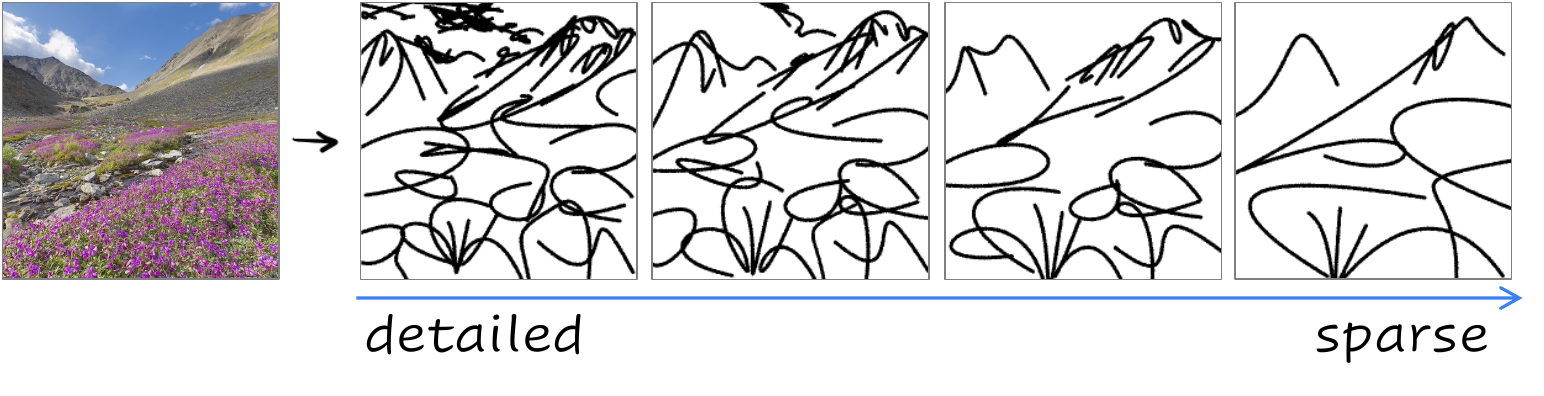}
  \caption{\small The \emph{simplicity} axis. On the left, we start with a more detailed sketch and as we move to the right the sketch is gradually simplified while still remaining consistent with the overall appearance of the initial sketch.}
 \label{fig:sparse_axis}
\end{figure}

\begin{figure}
    \centering
    \includegraphics[width=0.6\linewidth]{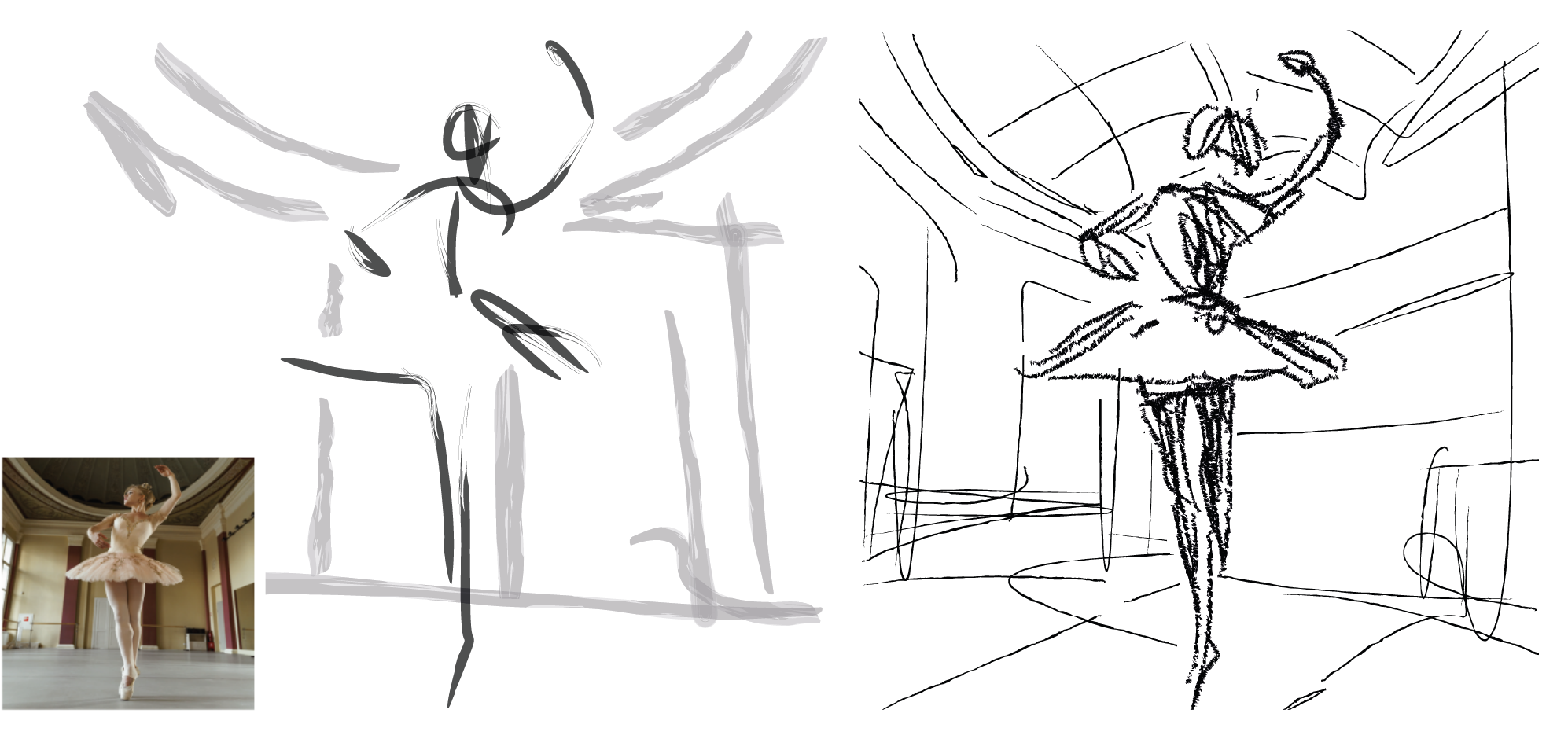}
    \caption{\small Artistic stylization of the strokes using Adobe Illustrator.}
    \label{fig:svg_editing}
\end{figure}

We define two axes representing two types of abstractions and produce sketches by gradually moving along these axes.
The first axis governs the \emph{fidelity} of the sketch. This axis moves from more precise sketches, where the sketch composition follows the geometry and structure of the photograph to more loose sketches, where the composition relies more on the semantics of the scene. An example is shown in~\Cref{fig:semantic_axis}, where the leftmost sketch follows the contours of the mountains on the horizon, and as we move right, the mountains and the flowers in the front gradually deviate from the edges present in the input, but still convey the correct semantics of the scene.
The second axis governs the level of details of the sketch and moves from detailed to sparse depictions, which appear more abstract. Hence, we refer to this axis as the \textit{simplicity} axis.
\eg
An example can be seen in \Cref{fig:sparse_axis}, where the same general characteristics of the scene (\eg the mountains and flowers) are captured in all sketches, but with gradually fewer details.

To deal with scene complexity, we separate the foreground and background elements and sketch each of them separately. This explicit separation and the disentanglement into two abstraction axes provide a more flexible framework for computational sketching, where users can choose the desired sketch from a range of possibilities, according to their goals and personal taste.

We define a sketch as a set of B\'{e}zier\xspace curves, and train a simple multi-layer perceptron (MLP) network to learn the stroke parameters.
Training is performed per image (\eg without an external dataset) and is guided by a pre-trained CLIP-ViT model~\cite{Radfordclip,dosovitskiy2020image}, leveraging its powerful ability to capture the semantics and global context of the entire scene.

To realize the \emph{fidelity} axis, we utilize different intermediate layers of CLIP-ViT to guide the training process, where shallow layers preserve the geometry of the image and deeper layers encourage the creation of looser sketches that emphasize the scene's semantics. 

To realize the \emph{simplicity} axis, we jointly train an additional MLP network that learns how to best discard strokes gradually and smoothly, without harming the recognizability of the sketch.
As shall be discussed, the use of the networks over a direct optimization-based approach allows us to define the level of details \textit{implicitly} in a learnable fashion, as opposed to explicitly determining the number of strokes \revision{(as was done in CLIPasso)}.

The resulting sketches demonstrate our ability to cope with various scenes and to capture their core characteristics while providing gradual abstraction along both the fidelity and simplicity axes, as shown in~\Cref{fig:clipascene_teaser}.
We compare our results with existing methods for scene sketching. 
We additionally evaluate our results quantitatively and demonstrate that the generated sketches, although abstract, successfully preserve the geometry and semantics of the input scene.

\section{Related Work}
Free-hand sketch generation differs from edge-map extraction \cite{canny1986computational, Winnemller2012XDoGAE} in that it attempts to produce sketches that are representative of the style of human drawings. Yet, there are significant differences in drawing styles among individuals depending on their goals, skill levels, and more (see~\Cref{fig:drawing_style_lob}). As such, computational sketching methods must consider a wide range of sketch representations.

This ranges from methods aiming to produce sketches that are grounded in the edge map of the input image~\cite{li2019photo,xie2015holistically,tong2021sketch,Deformable_Stroke}, to those that aim to produce sketches that are more abstract \cite{bhunia2021doodleformer,CLIPDraw,vinker2022clipasso,ha2017neural,qi2021sketchlattice,ge2021creative,qiu2021emergent,mihai2021learning,Zhou2018LearningTS}. Several works have attempted to develop a unified algorithm that can output sketches with a variety of styles \cite{chan2022learning,yi2020unpaired,liu2021neural}. There are, however, only a few works that attempt to provide various levels of abstraction~\cite{Berger2013,Deep-Sketch-Abstraction,vinker2022clipasso}.
In the following, we focus on scene-sketching approaches, and we refer the reader to \cite{xu2022deep} for a comprehensive survey on computational sketching techniques.

\vspace{-0.35cm}
\paragraph{Photo-Sketch Synthesis}
Various works formulate this task as an image-to-image translation task using paired data of corresponding images and sketches~\cite{li2019im2pencil,yi2019apdrawinggan,li2019photo,artline}. 
Others approach the translation task via unpaired data, often relying on a cycle consistency constraint~\cite{yi2020unpaired,song2018learning,chan2022learning}.
Li~\etal~\cite{li2019photo} introduce a GAN-based contour generation algorithm and utilize multiple ground truth sketches to guide the training process.
Yi~\etal~\cite{yi2020unpaired} generate portrait drawings with unpaired data by employing a cycle-consistency objective and a discriminator trained to learn a specific style.

Recently, Chan~\etal~\cite{chan2022learning} propose an unpaired GAN-based approach. They train a generator to map a given image into a sketch with multiple styles defined explicitly from four existing sketch datasets with a dedicated model trained for each desired style. 
They utilize a CLIP-based loss to achieve semantically-aware sketches.
As these works rely on curated datasets, they require training a new model for each desired style while supporting a single level of sketch abstraction.
In contrast, our approach does not rely on any explicit dataset and is not limited to a pre-defined set of styles. Instead, we leverage the powerful semantics captured by a pre-trained CLIP model~\cite{Radfordclip}.
Additionally, our work is the only one among the alternative scene sketching approaches that provides sketches with multiple levels of abstraction and in vector form, which allows for a wider range of editing and manipulation.

\vspace{-0.3cm}
\paragraph{CLIPasso} \revision{Our work rely on the general sketch generation framework and representation defined in CLIPasso~\cite{vinker2022clipasso}.}
\revision{As was introduced in \Cref{chap:clipasso}, CLIPasso} was designed for \textit{object} sketching at multiple levels of abstraction.
In CLIPasso, we define a sketch as a set of \Bezier curves and optimize the stroke parameters with respect to a CLIP-based \cite{Radfordclip} similarity loss between the input image and generated sketch. Multiple levels of abstraction are realized by reducing the number of strokes used to compose the sketch.
In contrast to CLIPasso, CLIPacene is not restricted to objects and can handle the challenging task of \textit{scene} sketching. 
Additionally, while in CLIPasso we examine only a single form of abstraction, in CLIPascene we disentangle abstraction into two distinct axes controlling both the simplicity and the fidelity of the sketch.
Moreover, in CLIPasso, the user is required to \textit{explicitly} define the number of strokes needed to obtain the desired abstraction. However, different images require a different number of strokes, which is difficult to determine in advance. In contrast, we \textit{implicitly} learn the desired number of strokes by training two MLP networks to achieve a desired trade-off between simplicity and fidelity with respect to the input image.

\setlength{\abovedisplayskip}{6pt}
\setlength{\belowdisplayskip}{6pt}

\section{Method}~\label{sec:method}
Given an input image $\mathcal{I}$ of a scene, our goal is to produce a set of corresponding sketches at $n$ levels of \textit{fidelity} and $m$ levels of \textit{simplicity},  forming a sketch abstraction matrix of size $m \times n$.
We begin by producing a set of sketches along the \textit{fidelity} axis (\Cref{sec:training_scheme,sec:sem_abs}) with no simplification, thus forming the top row in the abstraction matrix. Next, for each sketch at a given level of fidelity, we perform an iterative visual \textit{simplification} by learning how to best remove select strokes and adjust the locations of the remaining strokes (\Cref{sec:sketch_simp}).
For clarity, in the following we describe our method taking into account the entire scene as a whole. 
However, to allow for greater control over the appearance of the output sketches, and to tackle the high complexity presented in a whole scene, our final scheme splits the image into two regions -- the salient foreground object(s), and the background. We apply our 2-axes abstraction method to each region separately, and then combine them to form the matrix of sketches (details in \Cref{sec:split_for_back}).

\subsection{Training Scheme}~\label{sec:training_scheme}
\revision{Following the sketch representation introduced in \Cref{chap:clipasso},} we define a sketch as a set of $n$ strokes placed over a white background, where each stroke is a two-dimensional \Bezier curve with four control points.
We mark the $i$-th stroke by its set of control points $z_i = \{(x_i,y_i)^j\}_{j=1}^4$, and denote the set of the $n$ strokes by $Z=\{z_i\}_{i=1}^n$. Our goal is to find the set of stroke parameters that produces a sketch adequately depicting the input scene image.

\begin{figure}
    \centering
    \includegraphics[width=0.7\linewidth]{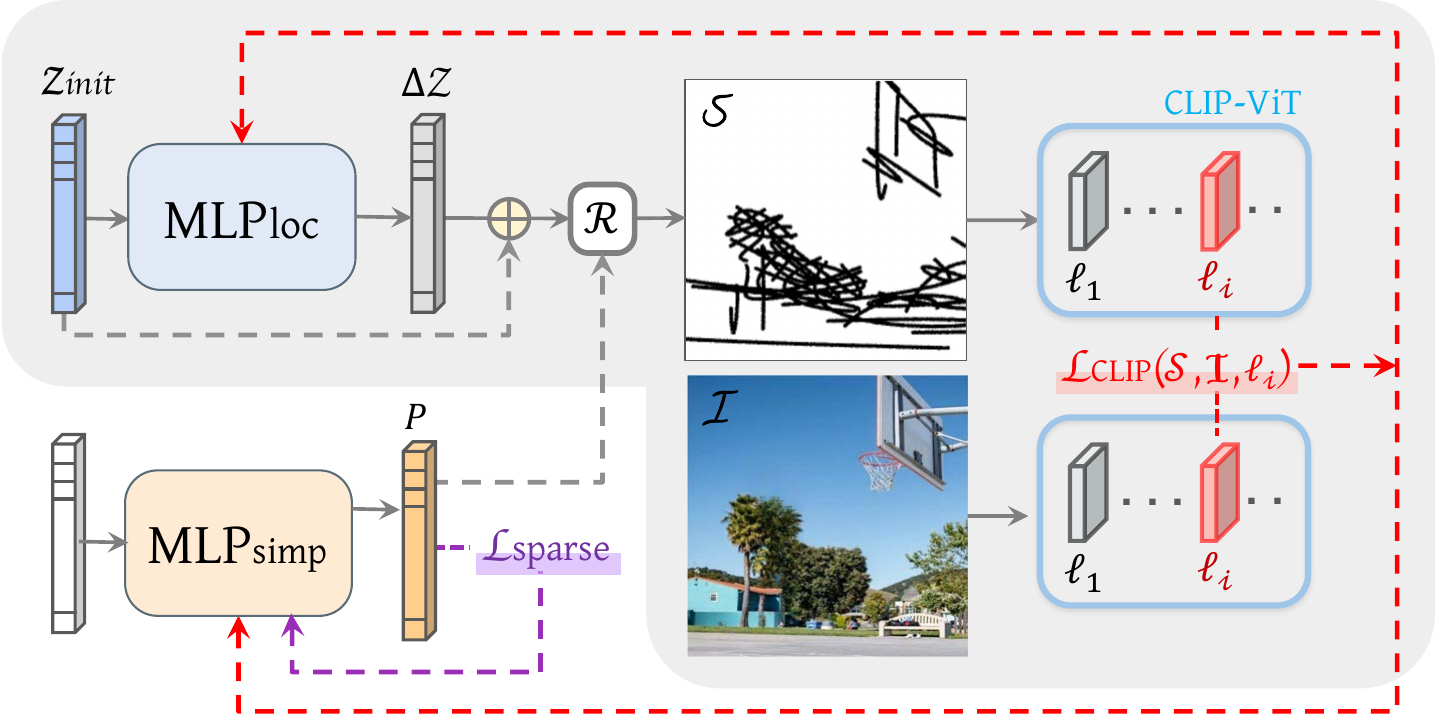}
    \caption{\small Single sketch generation scheme. 
    In gray, we show our training scheme for producing a single sketch image at a single level of fidelity. 
    In the bottom left we show the additional components used to generate a single sketch at a single level of simplicity.}
    \label{fig:pipeline_mlp}
\end{figure}

An overview of our training scheme used to produce a single sketch image is presented in the gray area of~\Cref{fig:pipeline_mlp}.
We train an MLP network, denoted by $MLP_{loc}$, that receives an initial set of control points $Z_{init} \in R^{n \times 4 \times 2}$ (marked in blue) and returns a vector of offsets $MLP_{loc}(Z_{init}) = \Delta Z \in \mathbb{R}^{n \times 4 \times 2}$ with respect to the initial stroke locations. The final set of control points are then given by $Z = Z_{init} + \Delta Z$, which are then passed to a differentiable rasterizer $\mathcal{R}$~\cite{diffvg} that outputs the rasterized sketch,

\begin{equation}
    \mathcal{S} = \renderer(Z_{init} + \Delta Z).
\end{equation}

For initializing the locations of the $n$ strokes, we follow the saliency-based initialization introduced in \Cref{chap:clipasso}, in which, strokes are initialized in salient regions based on a relevancy map extracted automatically~\cite{Chefer_2021_ICCV}.

To guide the training process, we leverage a pre-trained CLIP model due to its capabilities of encoding shared information from both sketches and natural images.
As opposed to CLIPasso that use the ResNet-based~\cite{he2016deep} CLIP model for the sketching process (and struggles with depicting a scene image),
we find that the ViT-based~\cite{dosovitskiy2020image} CLIP model is able to capture the global context required for generating a coherent sketch of a whole scene, including both foreground and background. This also follows the observation of Raghu~\etal~\cite{raghu2021vision} that ViT models better capture more global information at lower layers compared to ResNet-based models. We further analyze this design choice in the supplementary material.

The loss function is then defined as the L2 distance between the activations of CLIP on the image $\mathcal{I}$ and sketch $\mathcal{S}$ at a layer $\ell_k$:

\begin{equation}~\label{eq:clip_loss}
    \mathcal{L}_{CLIP}(\sketch,\image,\ell_k) = \big{\|}\ \cliploss{k}{\sketch} - \cliploss{k}{\image} \big{\|}_2^2. 
\end{equation}

At each step during training, we back-propagate the loss through the CLIP model and the differentiable rasterizer $\renderer$ whose weights are frozen, and only update the weights of $MLP_{loc}$. This process is repeated iteratively until convergence. Observe that no external dataset is needed for guiding the training process, as we rely solely on the expressiveness and semantics captured by the pre-trained CLIP model.
This training scheme produces a \textit{single} sketch image at a \textit{single} level of fidelity and simplicity. Below, we describe how to control these two axes of abstraction.

\subsection{Fidelity Axis}
\label{sec:sem_abs}
To achieve different levels of fidelity, as illustrated by a single row in our abstraction matrix, we select different activation layers of the CLIP-ViT model for computing the loss defined in~\Cref{eq:clip_loss}. 
Optimizing via deeper layers leads to sketches that are more semantic in nature and do not necessarily confine to the precise geometry of the input. 
Specifically, in all our examples we train a separate $MLP_{loc}$ using layers $\{\ell_2,\ell_7,\ell_8,\ell_{11}\}$ of CLIP-ViT and set the number of strokes to $n=64$.
Note that it is possible to use the remaining layers to achieve additional fidelity levels (see the supplementary material).

\subsection{Simplicity Axis}
\label{sec:sketch_simp}
Given a sketch $\sketch_k$ at fidelity level $k$, our goal is to find a set of sketches $\{\sketch_k^1, ..., \sketch_k^m\}$ that are visually and conceptually similar to $\sketch_k$ but have a gradually simplified appearance. In practice, we would like to learn how to best remove select strokes from a given sketch and refine the locations of the remaining strokes without harming the overall recognizability of the sketch. 

We illustrate our sketch simplification scheme for generating a single simplified sketch $\sketch_k^j$ in the bottom left region of ~\Cref{fig:pipeline_mlp}. We train an additional network, denoted as $MLP_{simp}$ (marked in orange), that receives a random-valued vector and is tasked with learning an $n$-dimensional vector $P = \{p_i\}_{i=1}^n$, where $p_i\in[0,1]$ represents the probability of the $i$-th stroke appearing in the rendered sketch.
$P$ is passed as an additional input to $\renderer$ which outputs the simplified sketch $\sketch_k^j$ in accordance.

\newpage
To implement the probabilistic-based removal or addition of strokes (which are discrete operations) into our learning framework, we multiply the width of each stroke $z_i$ by $p_i$. When rendering the sketch, strokes with a very low probability will be ``hidden'' due to their small width.

Similar to Mo~\etal~\cite{mo2021virtualsketching}, to encourage a sparse representation of the sketch (\ie one with fewer strokes) we minimize the normalized L1 norm of $P$:

\begin{equation}~\label{eq:struct_loss}
    \mathcal{L}_{sparse}(P) = \frac{\|P\|_1}{n}.
\end{equation}

To ensure that the resulting sketch still resembles the original input image, we additionally minimize the $\mathcal{L}_{CLIP}$ loss presented in~\Cref{eq:clip_loss}, and continue to fine-tune $MLP_{loc}$ during the training of $MLP_{simp}$. 
Formally, we minimize the sum:

\begin{equation}
    \mathcal{L}_{CLIP}(\sketch_k^j,\image,\ell_k) + \mathcal{L}_{sparse}(P).
\end{equation}

We back-propagate the gradients from $\mathcal{L}_{CLIP}$ to both $MLP_{loc}$ and $MLP_{simp}$ while $\mathcal{L}_{sparse}$ is used only for training $MLP_{simp}$ (as indicated by the red and purple dashed arrows in~\Cref{fig:pipeline_mlp}).

\begin{figure}
\centering
        \includegraphics[width=0.7\linewidth]{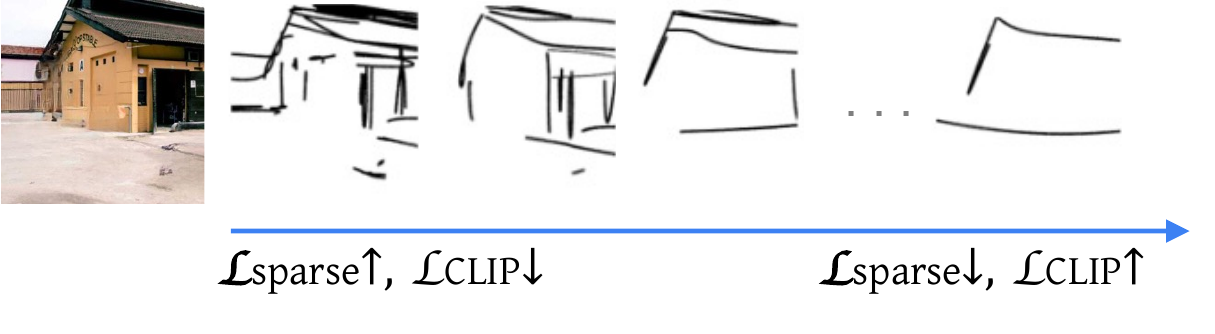}
 \caption{\small Trade-off between $\mathcal{L}_{sparse}$ and $\mathcal{L}_{CLIP}$. As the sketch becomes sparser, $\mathcal{L}_{sparse}$ obtains lower score. However, the sketch also becomes less recognizable with respect to the input image, resulting in a higher penalty for $\mathcal{L}_{CLIP}$.}
         \label{fig:loss_tradeoff}
\end{figure}

Note that using the MLP network rather than performing a direct optimization over the stroke parameters (as is done in Vinker~\etal) is crucial as it allows the optimization to restore strokes that may have been previously removed. If we were to use direct optimization, the gradients of deleted strokes would remain removed since they were multiplied by a probability of $0$.

Here, both MLP networks are simple $3$-layer networks with SeLU~\cite{klambauer2017self} activations. For $MLP_{simp}$ we append a Sigmoid activation to convert the outputs to probabilities.

\vspace{-0.4cm}
\paragraph{\textbf{Balancing the Losses.}}
Naturally, there is a trade-off between $\mathcal{L}_{CLIP}$ and $\mathcal{L}_{sparse}$, which affects the appearance of the simplified sketch (see ~\Cref{fig:loss_tradeoff}).
We utilize this trade-off to gradually alter the level of simplicity. 

Finding a balance between $\mathcal{L}_{sparse}$ and $\mathcal{L}_{CLIP}$ is essential for achieving recognizable sketches with varying degrees of abstraction.
Thus, we define the following loss:
\begin{equation}~\label{eq:clip_loss_simp}
    \mathcal{L}_{ratio} = \big{\|}\ \frac{\mathcal{L}_{sparse}}{\mathcal{L}_{CLIP}} - r 
 \big{\|}_2^2,
\end{equation}
where the scalar factor $r$ (denoting the \textit{ratio} of the two losses) controls the strength of simplification.
As we decrease $r$, we encourage the network to output a sparser sketch and vice-versa.
The final objective for generating a single simplified sketch $\sketch_k^j$, is then given by:
\begin{equation}~\label{eq:clip_loss_simp_sum}
   \mathcal{L}_{simp} = \mathcal{L}_{CLIP} + \mathcal{L}_{sparse} + \mathcal{L}_{ratio}.
\end{equation}

To achieve the set of gradually simplified sketches $\{\sketch_k^1, ..., \sketch_k^m\}$, we define a set of corresponding factors $\{r_k^1, ..., r_k^m\}$ to be applied in~\Cref{eq:clip_loss_simp}. 
The first factor $r_k^1$, is derived directly from~\Cref{eq:clip_loss_simp}, aiming to reproduce the strength of simplification present in $\sketch_k$:
\begin{equation}
    r_k^1 = \frac{1}{\mathcal{L}_{CLIP}(\sketch_k,\image,\ell_k)},
\end{equation}
where $\mathcal{L}_{sparse}$ equal to $1$ means that no simplification is performed.
The derivation of the remaining factors $r_k^j$ is described next. 

\vspace{-0.4cm}
\paragraph{\textbf{Perceptually Smooth Simplification}}
\begin{figure}
    \setlength{\belowcaptionskip}{-4pt}
    \centering
    \includegraphics[width=0.7\linewidth]{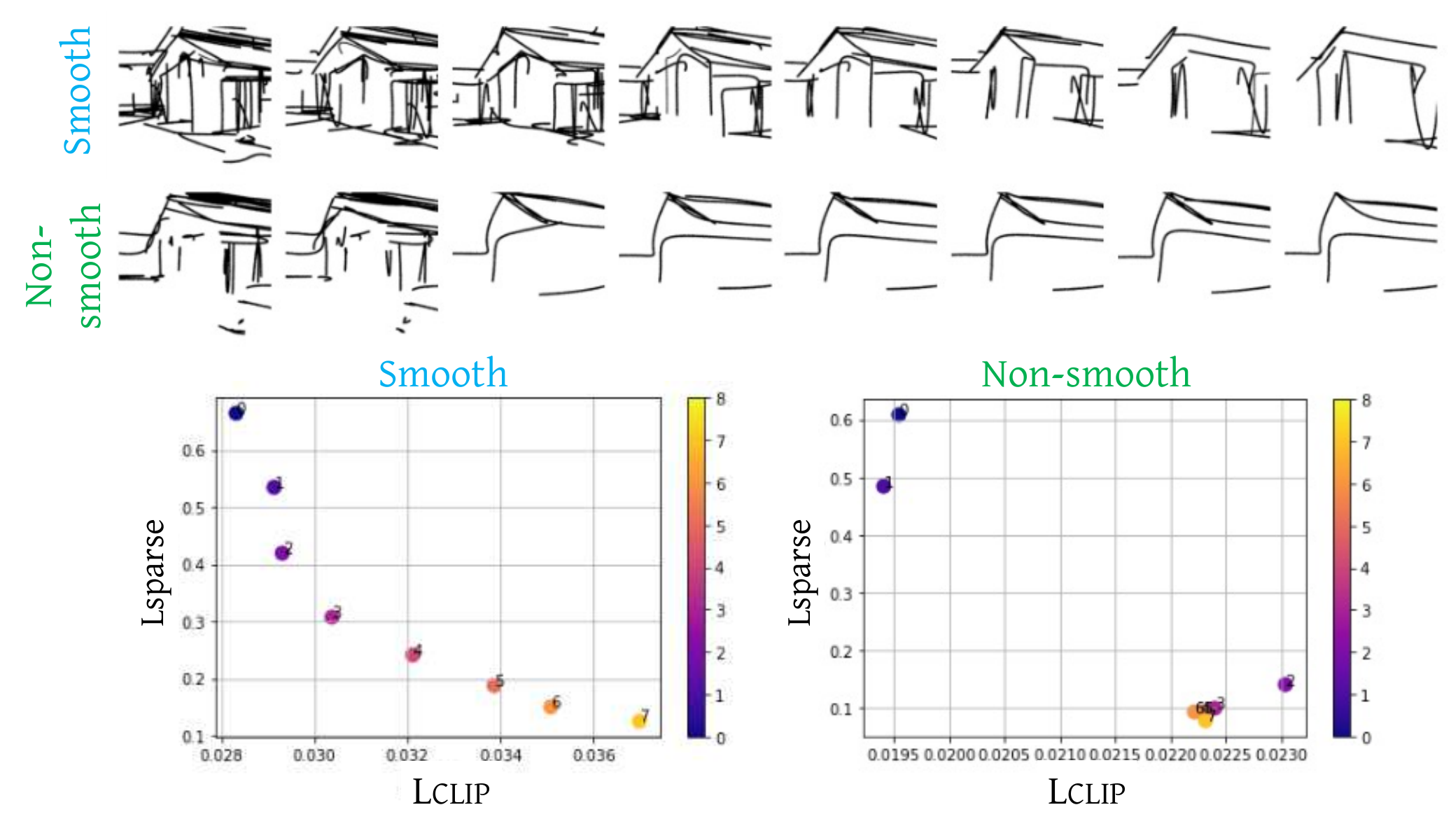}
    \caption{\small Smooth v.s. non-smooth simplification.
    In the first row, the simplification appears perceptually smooth, where a consistent change in the degree of abstraction is performed. 
    The second row demonstrates a non-smooth simplification, as there is a visible ``jump'' between the second and third sketches. These visual patterns are illustrated quantitatively in the corresponding graphs, where each dot in the graph represents a single sketch.}
    \label{fig:exponent}
\end{figure}

As introduced above, the set of factors determines the strength of the visual simplification. When defining the set of factors $r_k^j$, we aim to achieve a \textit{smooth} simplification. By \textit{smooth} we mean that there is no large change perceptually between two consecutive steps.
This is illustrated in~\Cref{fig:exponent}, where the first row provides an example of \emph{smooth} transitions, and the second row demonstrates a non-smooth transition, where there is a large perceptual ``jump'' in the abstraction level between the second and third sketches, and almost no perceptual change in the following levels.

We find that the simplification appears more smooth when $\mathcal{L}_{sparse}$ is exponential with respect to $\mathcal{L}_{CLIP}$.
The two graphs at the bottom of~\Cref{fig:exponent} describe this observation quantitatively, illustrating the trade-off between $\mathcal{L}_{sparse}$ and $\mathcal{L}_{CLIP}$ for each sketch. 
The smooth transition in the first row forms an exponential relation between $\mathcal{L}_{sparse}$ and $\mathcal{L}_{CLIP}$, while the large ``jump'' in the second row is clearly shown in the right graph.

Given this, we define an exponential function recursively by $f(j) = f(j-1)/2$. The initial value of the function is defined differently for each fidelity level $k$ as $f_k(1) = r_k^1$. To define the following set of factors $\{r_k^2, .. r_k^m\}$ we sample the function $f_k$, where for each $k$, the sampling step size is set proportional to the strength of the $\mathcal{L}_{CLIP}$ loss at level $k$. 
Hence, layers that incur a large $\mathcal{L}_{CLIP}$ value are sampled with a larger step size.
We found this procedure achieves simplifications that are perceptually smooth. 
This observation aligns well with the Weber-Fechner law~\cite{Weber1834-ms,fechnerlaw} which states that human perception is linear with respect to an exponentially-changing signal. 
An analysis of the factors and additional details regarding our design choices are provided in the supplementary material.

\begin{figure}
    \centering
    \includegraphics[width=0.6\linewidth]{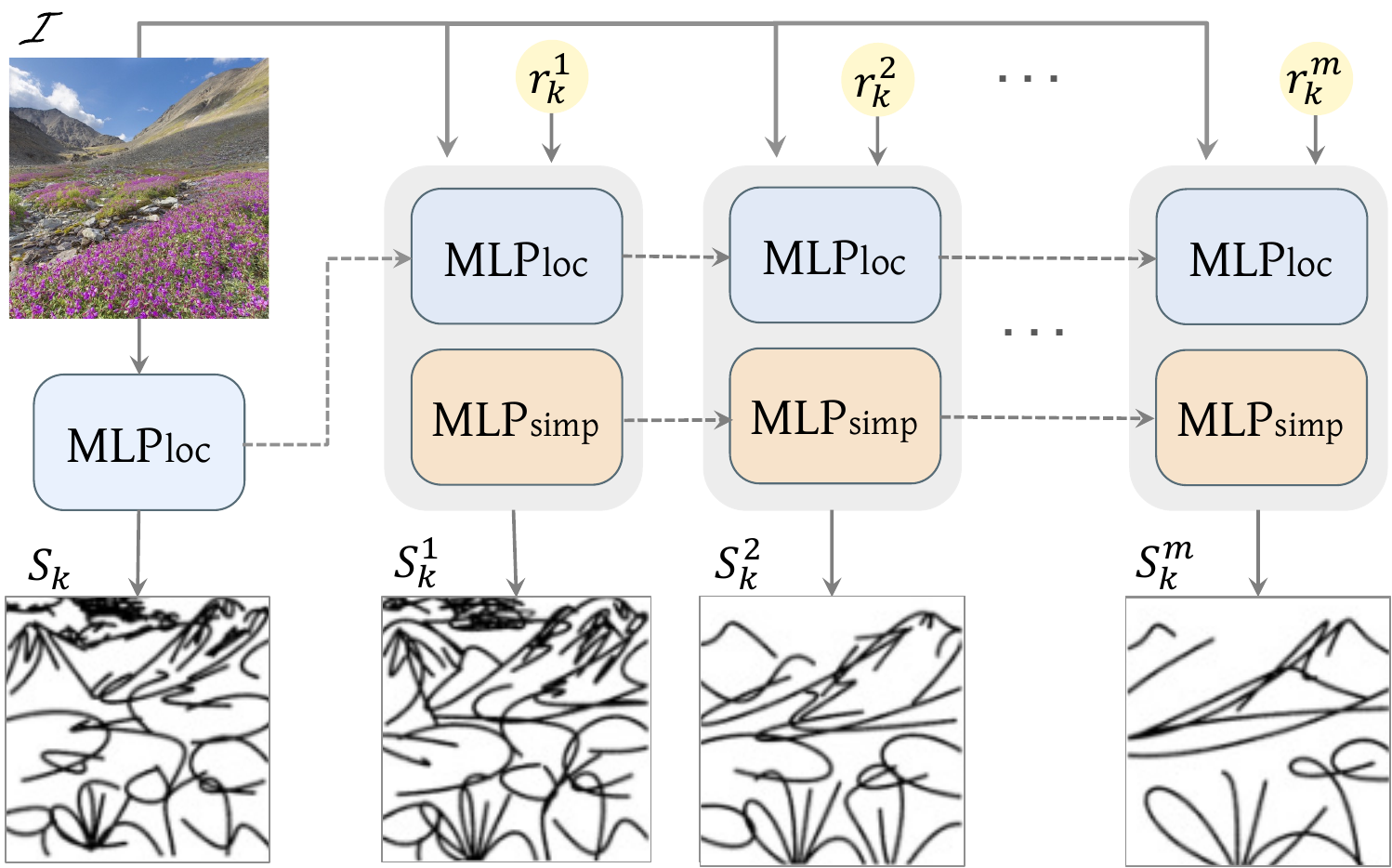}
    \caption{\small Iterative simplification of the sketch $\sketch_{k}$. 
    To produce a simplified sketch $\sketch_k^j$ we iteratively fine-tune $MLP_{loc}$ (blue) and $MLP_{simp}$ (orange) w.r.t $\mathcal{L}_{ratio}$ loss defined by each $r_k^j$.}
    \label{fig:iterative_simplification}
\end{figure}

\vspace{-0.4cm}
\paragraph{\textbf{Generating the Simplified Sketches}}
To generate the set of simplified sketches $\{\sketch_k^1 ... \sketch_k^m\}$, we apply the training procedure iteratively, as illustrated in~\Cref{fig:iterative_simplification}.
We begin with generating $\sketch_k^1$ w.r.t $r_k^1$, by fine-tuning $MLP_{loc}$ and training $MLP_{simp}$ from scratch.
After generating $\sketch_k^1$, we sequentially generate each $\sketch_k^j$ for $2\leq j \leq m$ by continuing training both networks for 500 steps and applying $\mathcal{L}_{ratio}$ with the corresponding factor $r_k^j$.
\begin{figure}[h!]
    \centering
    \includegraphics[width=0.7\linewidth]{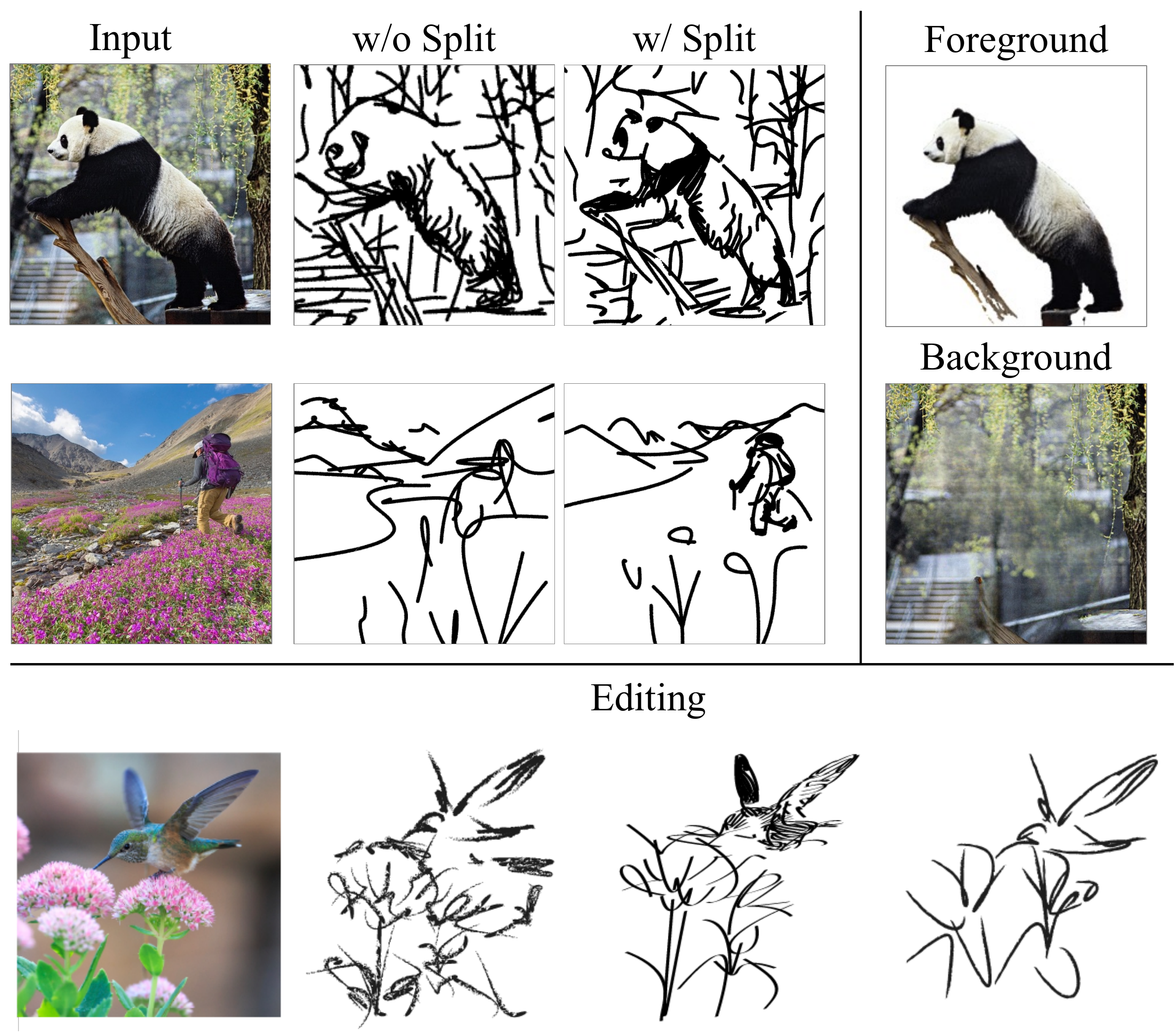}
    \caption{\small Scene decomposition. Top right -- an example of the separation technique. Left -- scene sketching results obtained with and without decomposing the scene. 
    Bottom -- examples of sketch  editing by modifying the style of strokes.}
     \vspace{-0.2cm}
    \label{fig:additional_control}
\end{figure}

\subsection{Decomposing the Scene}~\label{sec:split_for_back}
The process described above takes the entire scene as a whole. However, in practice, we separate the scene's foreground subject from the background and sketch each of them independently. We use a pretrained U$^2$-Net~\cite{qin2020u2} to extract the salient object(s), and then apply a pretrained LaMa~\cite{suvorov2022resolution} inpainting model to recover the missing regions (see~\Cref{fig:additional_control}, top right).

We find that this separation helps in producing more visually pleasing and stable results. When performing object sketching, we additionally compute $\mathcal{L}_{CLIP}$ over layer $l_4$. This helps in preserving the object's geometry and finer details. On the left part of~\Cref{fig:additional_control} we demonstrate the artifacts that may occur when scene separation is not applied. For example, over-exaggeration of features of the subject at a low fidelity level, such as the panda's face, or, the object might ``blend'' into the background.
Additionally, this explicit separation provides users with more control over the appearance of the final sketches (\Cref{fig:additional_control}, bottom right). For example, users can easily edit the vector file by modifying the brush's style or combine the foreground and background sketches at different levels of abstraction.

\vspace{-0.1cm}
\section{Results}~\label{sec:results_clipascene}
In the following, we demonstrate the performances of our scene sketching technique qualitatively and quantitatively, and provide comparisons to state-of-the-art sketching methods.
Further analysis, results, and a user study are provided in the supplementary material.

\vspace{-0.1cm}
\subsection{Qualitative Evaluation}
In~\Cref{fig:qualitative_pairs,fig:clipascene_teaser} we show sketches at different levels of abstraction on various scenes generated by our method. Notice how it is easy to recognize that the sketches are depicting the same scene even though they vary significantly in their abstraction level.

\begin{figure}[h]
    \centering
    \includegraphics[width=0.7\linewidth]{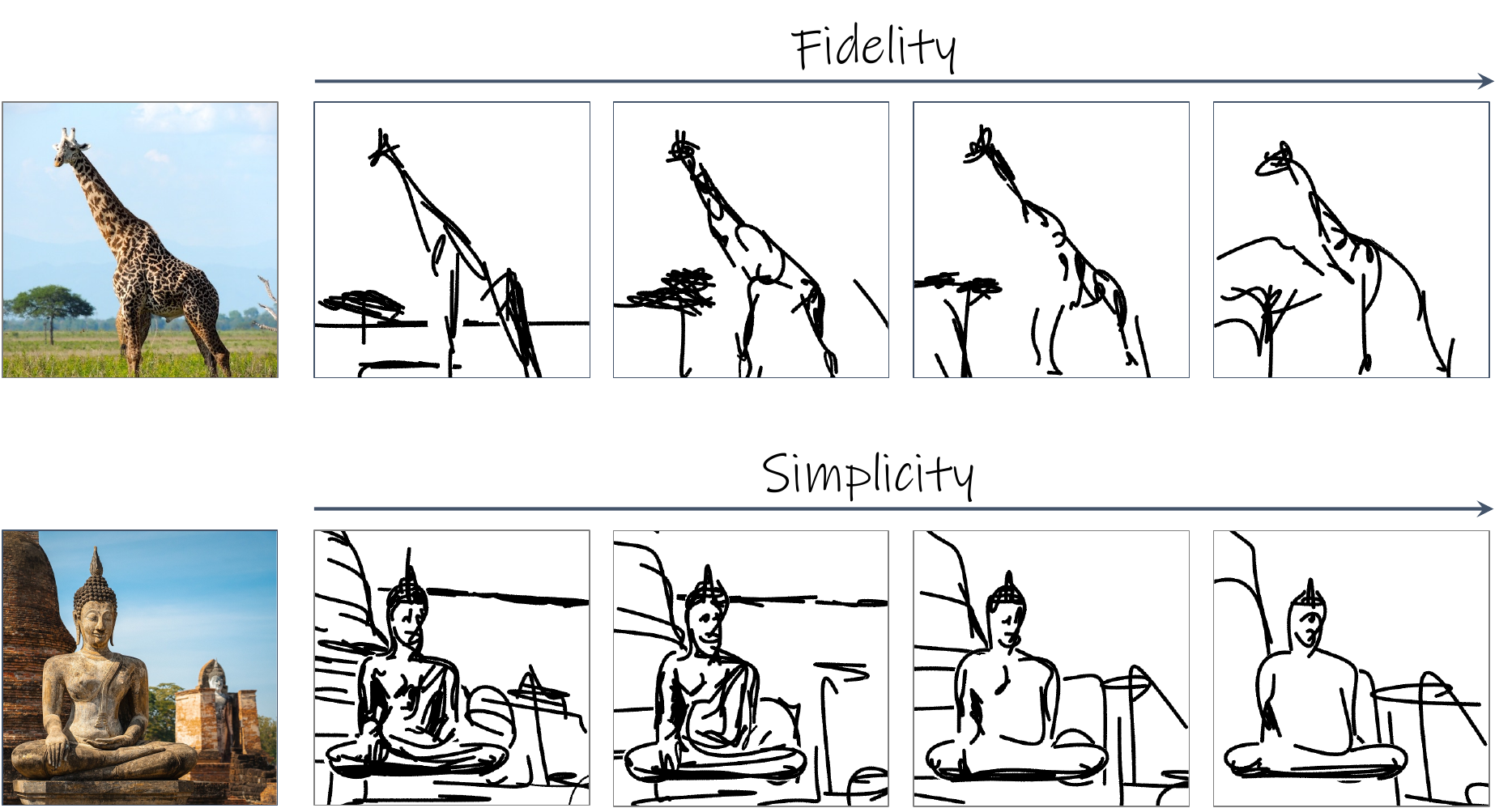}
    \caption{\small Sketches along the two abstraction axes.}
    \label{fig:two_axes}
\end{figure}

In~\Cref{fig:clipascene_teaser,fig:semantic_axis,fig:two_axes} (top) we show sketch abstractions along the \textit{fidelity} axis, where the sketches become less precise as we move from left to right, while still conveying the semantics of the images (for example the mountains in the background in~\Cref{fig:clipascene_teaser} and the tree and giraffe's body in~\Cref{fig:two_axes}). 
In~\Cref{fig:clipascene_teaser,fig:sparse_axis,fig:two_axes} (bottom) we show sketch abstractions along the \textit{simplicity} axis. Our method successfully simplifies the sketches in a smooth manner, while still capturing the core characteristics of the scene. For example, notice how the shape of the Buddha sculpture is preserved across all levels.
Observe that these simplifications are achieved \textit{implicitly} using our iterative sketch simplification technique. 
Please refer to the supplemental file for many more results.

\vspace{-0.01cm}
\subsection{Comparison with Existing Methods}
\label{sec:compareMethods}
\begin{figure}[h]
    \centering
    \includegraphics[width=0.7\linewidth]{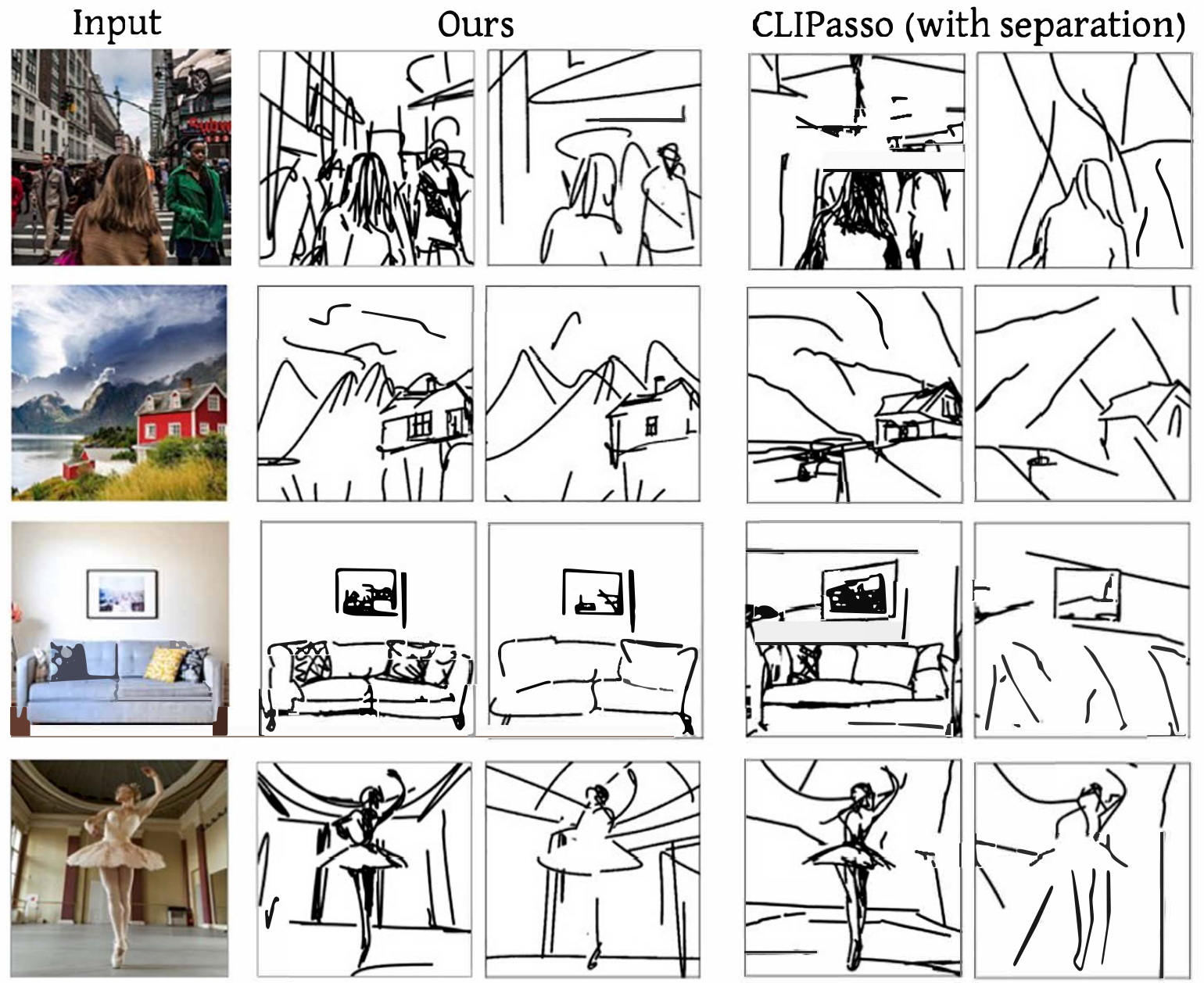}
    \caption{\small Comparison to CLIPasso~\cite{vinker2022clipasso}. Note how CLIPasso fails to capture the background in most cases, especially at higher abstraction levels, despite having the same stroke budget.}
    \label{fig:qualitative_pairs}
\end{figure}

In~\Cref{fig:qualitative_pairs} we present a comparison to our object sketching technique  CLIPasso which was introduced in \Cref{chap:clipasso}. For a fair comparison, we use our scene decomposition technique to separate the input images into foreground and background and use CLIPasso to sketch each part separately before combining them. Also, since CLIPasso requires a predefined number of strokes as input, we set the number of strokes in CLIPasso to be the same as that learned implicitly by our method.
For each image we show two sketches with two different levels of abstraction. 
As expected, CLIPasso is able to portray objects accurately, as it was designed for this purpose. However, in some cases, such as the sofa, CLIPasso fails to depict the object at a higher abstraction level. This drawback may result from the abstraction being learned from scratch per a given number of strokes, rather than gradually. Additionally, in most cases CLIPasso completely fails to capture the background, even when using many strokes (\eg in the first and fourth rows). Our method captures both the foreground and the background in a visually pleasing manner, thanks to our \textit{learned} simplification approach. For example, our method is able to convey the notion of the buildings in the first row or mountains in the second row with only a small number of simple scribbles. Similarly, our approach successfully depicts the subjects across all scenes.

\newpage
Note that the main advantage of CLIPascene over CLIPasso does not stem only from using the CLIP-ViT architecture, but is rooted in our neural simplification approach that gradually removes strokes by considering the complexity of each image, as well as the tradeoff between fidelity and simplicity in the abstraction process. In Figures 6 and 8 of the supplementary, we show simplifications obtained when using CLIP-ViT with a pre-defined number of strokes on background images (which is equivalent to applying CLIPasso with ViT instead of ResNet). The  simplification using CLIPasso does not succeed in the more abstract levels of  $16$ or fewer strokes. 
We provide additional sketch results comparing our method and CLIPasso at different abstraction levels in the supplementary.

In~\Cref{fig:scene_sketching_comparisons} we present a comparison with three state-of-the-art methods for scene sketching~\cite{yi2020unpaired, li2019photo, chan2022learning}.
On the left, as a simple baseline, we present the edge maps of the input images obtained using XDoG~\cite{Winnemller2012XDoGAE}. 
On the right, we present three sketches produced by our method depicting three representative levels of abstraction.

The sketches produced by UPDG~\cite{yi2020unpaired} and Chan~\etal~\cite{chan2022learning} are detailed, closely following the edge maps of the input images (such as the buildings in row $2$). 
These sketches are most similar to the sketches shown in the leftmost column of our set of results, which also align well with the input scene structure.
The sketches produced by Photo-Sketching~\cite{li2019photo} are less detailed and may lack the semantic meaning of the input scene. For example, in the first row, it is difficult to identify the sketch as being that of a person.
Importantly, none of the alternative \textit{scene} sketching approaches can produce sketches with varying abstraction levels, nor can they produce sketches in vector format.
We note that in contrast to the methods considered in~\Cref{fig:scene_sketching_comparisons}, our method operates per image and requires no training data. However, this comes with the disadvantage of longer running time, taking 6 minutes to produce one sketch on a commercial GPU.

\begin{figure*}[ht]
    \centering
    \setlength{\tabcolsep}{1.5pt}
    {\small
    \begin{tabular}{c @{\hspace{0.2cm}} | c c c c @{\hspace{0.2cm}} | c c c}

        \includegraphics[width=0.11\textwidth]{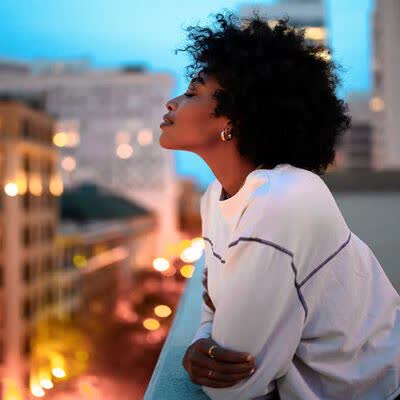} &
        \hspace{0.11cm}
        \includegraphics[width=0.11\textwidth]{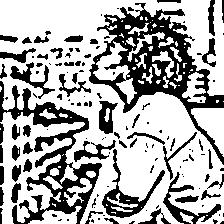} &
        \includegraphics[width=0.11\textwidth]{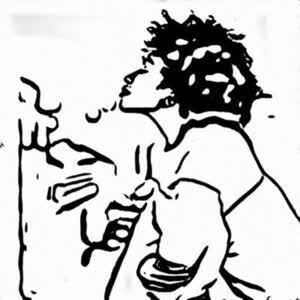} &
        \includegraphics[width=0.11\textwidth]{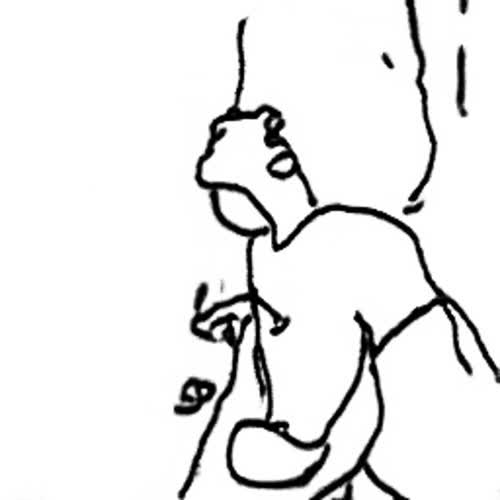} &
        \includegraphics[width=0.11\textwidth]{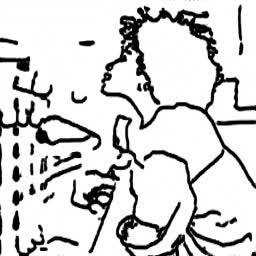} &
        \hspace{0.11cm}
        \includegraphics[width=0.11\textwidth]{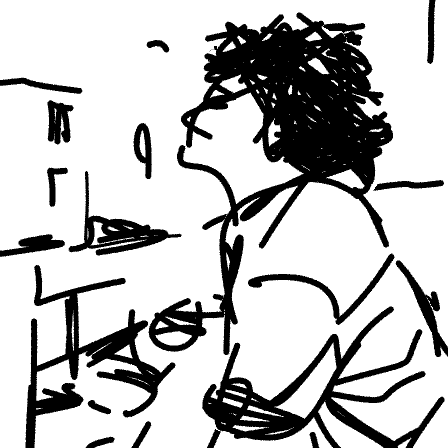} &
        \includegraphics[width=0.11\textwidth]{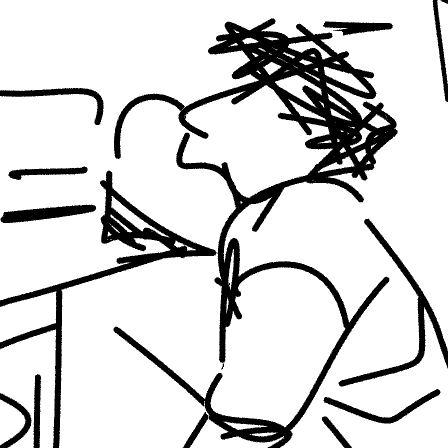} &
        \includegraphics[width=0.11\textwidth]{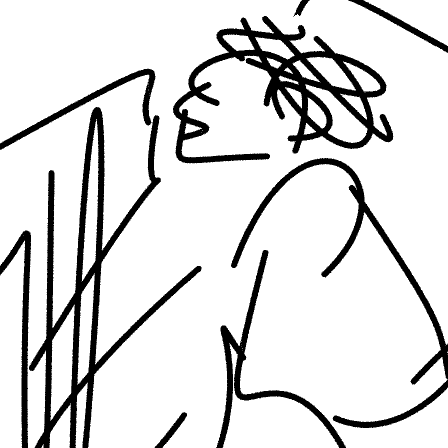}\\

        \includegraphics[width=0.11\textwidth]{clipascene/figs/inputs/woman_city.jpg} &
        \hspace{0.11cm}
        \includegraphics[width=0.11\textwidth]{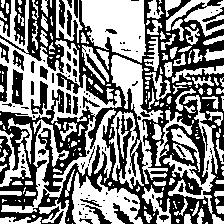} &
        \includegraphics[width=0.11\textwidth]{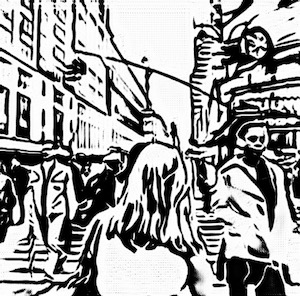} &
        \includegraphics[width=0.11\textwidth]{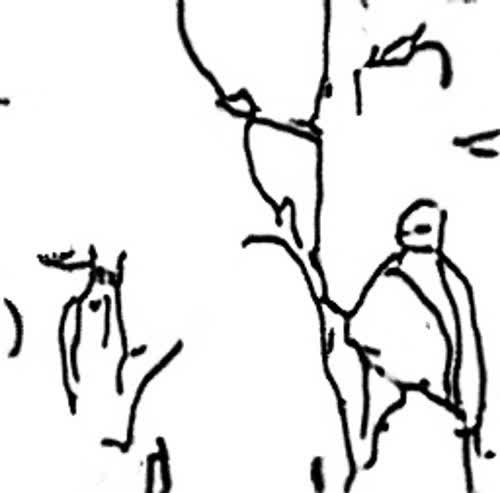} &
        \includegraphics[width=0.11\textwidth]{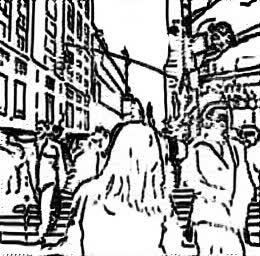} &
        \hspace{0.11cm}
        \includegraphics[width=0.11\textwidth]{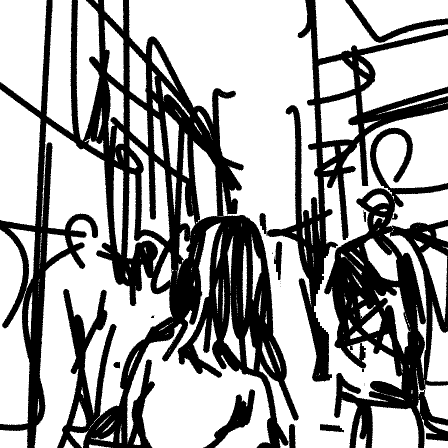} &
        \includegraphics[width=0.11\textwidth]{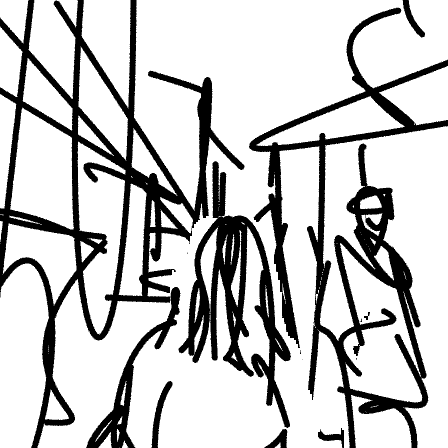} &
        \includegraphics[width=0.11\textwidth]{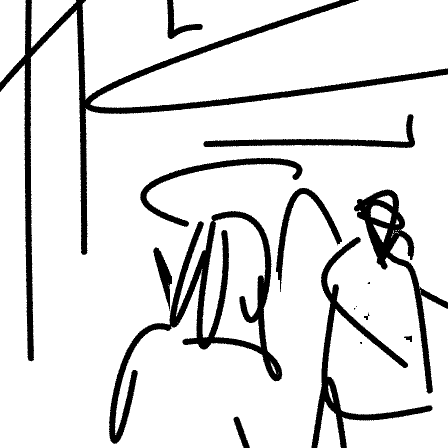} \\
        
        \includegraphics[width=0.11\textwidth]{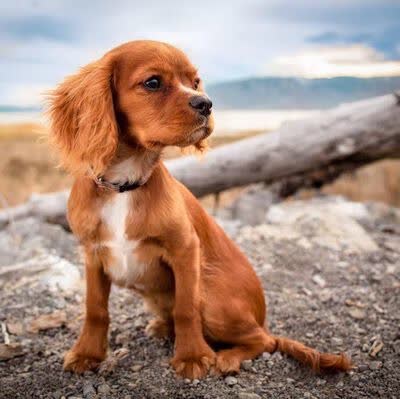} &
        \hspace{0.11cm}
        \includegraphics[width=0.11\textwidth]{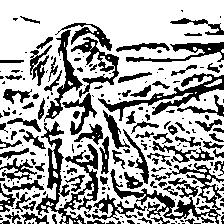} &
        \includegraphics[width=0.11\textwidth]{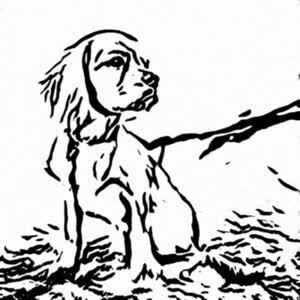} &
        \includegraphics[width=0.11\textwidth]{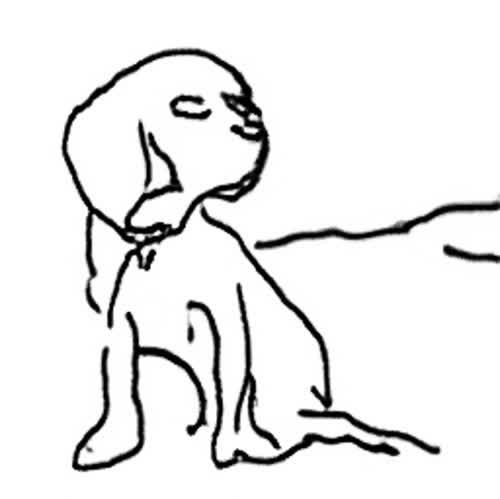} &
        \includegraphics[width=0.11\textwidth]{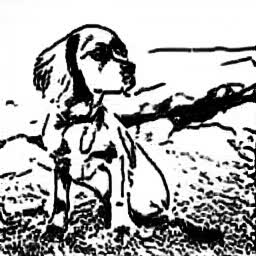} &
        \hspace{0.11cm}
        \includegraphics[width=0.11\textwidth]{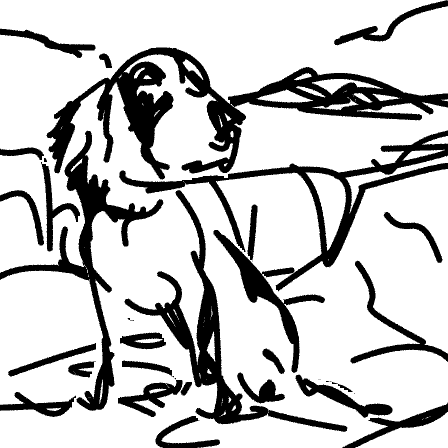} &
        \includegraphics[width=0.11\textwidth]{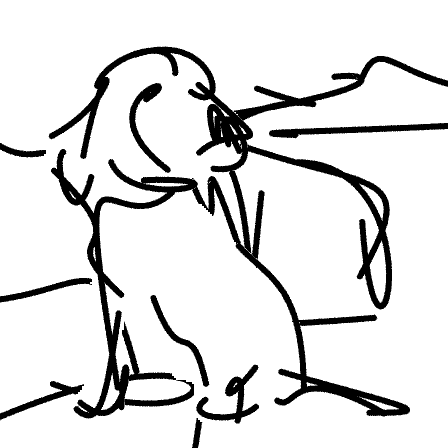} &
        \includegraphics[width=0.111\textwidth]{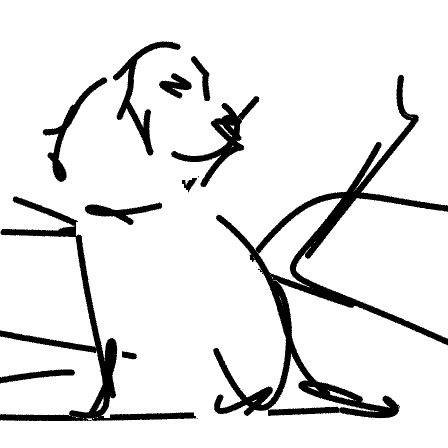} \\

        Input & XDoG & UPDG & \begin{tabular}[c]{@{}c@{}} Photo \\ Sketching \end{tabular} & Chan~\etal &
        \multicolumn{3}{c}{\hspace{0.111cm} \xrfill[0.5ex]{0.5pt}\quad Ours \quad \xrfill[0.5ex]{0.5pt}}

    \end{tabular}
    
    }
    \vspace{0.11cm}
    \caption{\small Scene sketching results and comparisons. From left to right are the sketches obtained using XDoG~\cite{Winnemller2012XDoGAE}, UPDG \cite{yi2020unpaired}, Photo-Sketching \cite{li2019photo}, and Chan~\etal~\cite{chan2022learning}. On the right, are three representative sketches produced by our method depicting three levels of abstraction. Note that UPDG and Chan~\etal can produce sketches with three different styles, however all the sketches represent a similar level of abstraction. We choose one representative style but provide more style comparisons in the supplementary material.}
    \label{fig:scene_sketching_comparisons}
\end{figure*}

\subsection{Quantitative Evaluation}
In this section, we provide a quantitative evaluation of our method's ability to produce sketch abstractions along both the simplicity and fidelity axes.
To this end, we collected a variety of images spanning five classes of scene imagery: people, urban, nature, indoor, and animals, with seven images for each class. For each image, we created the $4\times 4$ sketch abstraction matrix -- resulting in a total of $560$ sketches, and 
created sketches using the different methods presented in Section~\ref{sec:compareMethods}.
To make a fair comparison with CLIPasso we generated sketches with four levels of abstraction, using the average number of strokes obtained by our method at the four simplicity levels. For UPDG and Chan~\etal, we obtained sketches with three different styles, and averaged the quantitative scores across the three styles, as they represent the same abstraction level. For Photo-Sketching only one level of abstraction and one style is supported.

\vspace{-0.2cm}
\paragraph{\textbf{Fidelity Changes}}
To measure the fidelity level of the generated sketches, we compute the MS-SSIM~\cite{wang2003multiscale} score between the edge map of each input image (extracted using XDoG) and the corresponding sketch.
In~\Cref{tb:geometry_metrics_by_fidelity} we show the average resulting scores among all categories, where a higher score indicates a higher similarity.
Examining the results matrix of our method, as we move right along the fidelity axis, the scores gradually decrease.  This indicates that the sketches become ``looser'' with respect to the input geometry, as desired.
The sketches by UPDG and Chan~\etal obtained high scores, which is consistent with our observation that their method produces sketches that follow the edges of the input image.
The scores for CLIPasso show that the fidelity level of their sketches does not change much across simplification levels and is similar to the fidelity of sketches of our method at the last two levels (the two rightmost columns). This suggests that CLIPasso is not capable of producing large variations of fidelity abstractions.

\begin{table}
\small
\centering
\setlength{\tabcolsep}{4pt}
\caption{\small Comparison of the average MS-SSIM score, computed  between the edge map of the input images and generated sketches.}
\begin{tabular}{c@{\hskip6pt}c@{\hskip6pt}c@{\hskip6pt}c@{\hskip6pt}ccccc} 
\toprule
& \multicolumn{4}{c}{Ours} & \multirow{2}{*}{CLIPasso} & \multirow{2}{*}{UPDG} & \multirow{2}{*}{\begin{tabular}[c]{@{}c@{}}Chan\\ et al. \end{tabular}} & \multirow{2}{*}{\begin{tabular}[c]{@{}c@{}}Photo-\\Sketch\end{tabular}}  \\ 
\cmidrule(l{2pt}r{2pt}){2-5}
& \multicolumn{4}{c}{Fidelity} & & & & \\ 
\cmidrule(l{2pt}r{2pt}){6-6}\cmidrule(l{2pt}r{2pt}){7-7}\cmidrule(l{2pt}r{2pt}){8-8}\cmidrule(l{2pt}r{2pt}){9-9}
\multirow{4}{*}{\rotatebox{90}{Simplicity}}
& 0.39 & 0.23 & 0.22 & 0.17 & 0.21 & \multirow{4}{*}{0.57} & \multirow{4}{*}{0.55} & \multirow{4}{*}{0.27} \\ 
& 0.37 & 0.23 & 0.21 & 0.19 & 0.21 &   &   & \\
& 0.36 & 0.22 & 0.20 & 0.18 & 0.15 &   &   & \\
& 0.34 & 0.22 & 0.18 & 0.14 & 0.13 &   &   & \\  
\bottomrule
\end{tabular}

\label{tb:geometry_metrics_by_fidelity}
\end{table}

\vspace{-0.35cm}
\paragraph{Sketch Recognizability} 
A key requirement for successful abstraction is that the input scene will remain recognizable in the sketches across different levels of abstraction. 
To evaluate this, we devise the following recognition experiment on the set of images described above. Using a pre-trained ViT-B/16 CLIP model (different than the one used for training), we performed zero-shot image classification over each input image and the corresponding resulting sketches from the different methods. We use a set of $200$ class names taken from commonly used image classification and object detection datasets~\cite{lin2014microsoft,krizhevsky2009learning} and compute the percent of sketches where at least $2$ of the top $5$ classes predicted for the input image were also present in the sketch's top $5$ classes.
We consider the top $5$ predicted classes since a scene image naturally contains multiple objects.

~\Cref{tb:recognizability_metrics} shows the average recognition rates across all images for each of the described methods.
The recognizability of the sketches produced by our method remains relatively consistent across different simplicity and fidelity levels, with a naturally slight decrease as we increase the simplicity level. 
We do observe a large decrease in the recognition score in the first column. This discrepancy can be attributed to the first fidelity level following the image structure closely, which makes it more difficult to depict the scene with fewer strokes.
CLIPasso's fidelity level is most similar to our two rightmost columns (as shown in \Cref{tb:geometry_metrics_by_fidelity}). When comparing our recognition rates along these columns to the results of CLIPasso, one can observe that at higher simplicity levels, their method looses the scene's semantics.

\begin{table}
\small
\centering
\setlength{\tabcolsep}{4pt}
\caption{\small Recognizability scores, using a CLIP ViT-B/16 model for zero-shot classification on the input image and generated sketches.}
\begin{tabular}{c@{\hskip6pt}c@{\hskip6pt}c@{\hskip6pt}c@{\hskip6pt}ccccc} 
\toprule
& \multicolumn{4}{c}{Ours} & \multirow{2}{*}{CLIPasso} & \multirow{2}{*}{UPDG} & \multirow{2}{*}{\begin{tabular}[c]{@{}c@{}}Chan\\et al.\end{tabular}} & \multirow{2}{*}{\begin{tabular}[c]{@{}c@{}}Photo-\\Sketch\end{tabular}}  \\ 
\cmidrule(l{2pt}r{2pt}){2-5}
& \multicolumn{4}{c}{Fidelity} & & & & \\ 
\cmidrule(l{2pt}r{2pt}){6-6}\cmidrule(l{2pt}r{2pt}){7-7}\cmidrule(l{2pt}r{2pt}){8-8}\cmidrule(l{2pt}r{2pt}){9-9}
\multirow{4}{*}{\rotatebox{90}{Simplicity}} 

& 0.92 & 1.00 & 0.95 & 0.97 & 0.92 & \multirow{4}{*}{0.87} & \multirow{4}{*}{0.91} & \multirow{4}{*}{0.62} \\
& 0.54 & 0.97 & 1.00 & 0.91 & 0.83 &   &  & \\
& 0.54 & 0.93 & 0.94 & 0.89 & 0.70 &   &  & \\
& 0.44 & 0.79 & 0.91 & 0.85 & 0.43 &   &  & \\
\bottomrule
\end{tabular}
\label{tb:recognizability_metrics}

\end{table}

\paragraph{User Study} 
As opposed to the fidelity measure, which can be determined by measuring the distance from the edge map of the input scene, validating the recognizability is more challenging.
To this end, we also conducted a user study to further validate the findings presented by the CLIP zero-shot classification approach.

\begin{wrapfigure}{R}{0.5\textwidth}
      \begin{center}
    \includegraphics[width=0.8\linewidth]{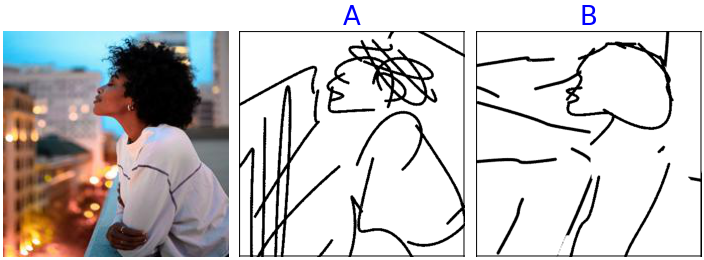}
    \end{center}
    \caption{\small Example of how images were presented to participants in the user study.}
    \vspace{-0.2cm}
    \label{fig:user_study_example}
\end{wrapfigure}

The user study examines how well the sketches depict the input scene, considering both the foreground and background.
Using $30$ images from the set described in Section 4.3, we compared our sketches with three alternative methods: CLIPasso~\cite{vinker2022clipasso}, Chan~\etal~\cite{chan2022learning}, and Photo-Sketching~\cite{li2019photo}. 
The participants were presented with the input image along with two sketches, one produced by our method and the other by the alternative method (with the sketches presented in random order). An example is provided in~\Cref{fig:user_study_example}.
The following question was posed to participants:
\textit{Which sketch better depicts the image content?}
\textit{In your answer, please relate to:}
\textit{(1) Preservation of both foreground and background.}
\textit{(2) Semantic preservation - i.e., reflecting the meaning of the elements.}
Participants could choose between three options: ``A", ``B", and ``A and B at a similar level``.

\begin{table}
\setlength{\tabcolsep}{3pt} 
 \renewcommand{\arraystretch}{1.2} 
\begin{center}
\caption{\small Results of our user study. We compare $30$ sketches produced by our method to three alternative methods: CLIPasso~\cite{vinker2022clipasso}, Chan~\etal~\cite{chan2022learning}, and Photo-Sketching~\cite{li2019photo}. For each method, we specify the percent of responses that preferred our sketch, the sketch of the alternative method, or found the sketches to be similar in their ability to capture the scene semantics.}
\begin{tabular}{c|c||c|c||c|c}
    \hline
    \multicolumn{2}{c||}{\begin{tabular}[c]{@{}c@{}}Ours v.s. \\ CLIPasso\end{tabular}} & \multicolumn{2}{c||}{\begin{tabular}[c]{@{}c@{}}Ours v.s. \\ Chan~\etal \end{tabular}} & \multicolumn{2}{c}{\begin{tabular}[c]{@{}c@{}}Ours v.s. \\ Photo-Sketching\end{tabular}} \\
    \toprule
    Ours & 84.8\% & Ours & 52.5\% & Ours & 75.6\% \\
    \midrule
     CLIPasso & 7.0\% & Chan~\etal & 27.3\% & \begin{tabular}[c]{@{}c@{}}Photo-\\ Sketching\end{tabular} & 15.8\% \\
     \midrule
     Equal & 8.2\% & Equal & 20.2\% & Equal & 8.6\%\\
    \bottomrule
\end{tabular}
\end{center}
\vspace{-0.3cm}

\label{tab:user_study}
\end{table}

To make a fair comparison, we compared the methods which produce abstract sketches (CLIPasso~\cite{vinker2022clipasso} and Photo-Sketching~\cite{li2019photo}) with our more abstract sketches (highest abstraction level of our abstraction matrix).
Conversely, we compared the sketches of Chan~\etal~\cite{chan2022learning}, which are more detailed and have greater fidelity, to our most detailed results (top left corner of our matrix).
We applied CLIPasso using our scene decomposition technique, and the same number of strokes as learned by our method, and for Chan~\etal we used the contour style sketches.
We collected responses from $25$ participants for the survey, which contained a total of $90$ questions ($2,250$ responses in total).

The average resulting scores among all participants and images are shown in~\Cref{tab:user_study}.
In the first, second, and third columns, we show the scores obtained when comparing to CLIPasso~\cite{vinker2022clipasso}, Chan~\etal~\cite{chan2022learning}, and Photo-Sketching~\cite{li2019photo} respectively.
Compared to CLIPasso and Photo-Sketching, our method achieved significantly higher rates ($84.8\%$ and $75.6\%$ of responses favored our method over the respective alternatives). Conversely, only $7\%$ and $15.8\%$ of responses preferred the results of the alternative method, respectively.
Although sketches produced by Chan~\etal~\cite{chan2022learning} are highly detailed, $52.5\%$ of the responses preferred our sketches, while only $20\%$ considered our sketches and Chan \etal's sketches to be similar.

The results of the user study demonstrate that sketches produced by our method faithfully capture both the foreground and background elements in the scene among varying abstraction levels.

\section{Limitations and Future Work}
There are several limitations to our method. 
First, since we separate foreground and background, and generate many sketches per one scene, generating a $4x4$ matrix of abstractions requires three hours on a single commercial GPU. 
Although further optimization is possible, sketches along the simplicity axis depends on the previous step.
Second, we defined two axes for sketch abstraction, however, other axes of abstraction could be defined. Third, our resulting sketches cover only a portion of each of the axes. A future extension could focus on further extending these axes. 

Finally, future potential uses of our proposed method could focus on generating image-sketch paired datasets containing a range of styles and abstractions. Such data could potentially be used to train multi-modal machine learning algorithms.

\section{Conclusions}
We presented a method for performing scene sketching with different types and multiple levels of abstraction.
We disentangled the concept of sketch abstraction into two axes: \textit{fidelity} and \textit{simplicity}. 
We demonstrated the ability to cover a wide range of abstractions across various challenging scene images and the advantage of using vector representation and scene decomposition to allow for greater artistic control.
It is our hope that our work will open the door for further research in the emerging area of computational generation of visual abstractions. Future research could focus on further extending these axes and formulate innovative ideas for controlling visual abstractions.

\begin{figure*}
    \centering
    \begin{tabular}{c c c c c c c c}\includegraphics[width=0.098\textwidth,height=0.098\textwidth]{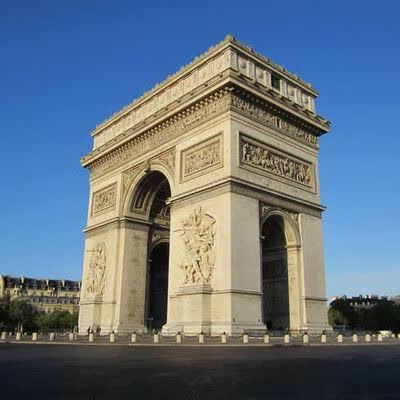} & & & &
    \hspace{0.5cm}
    \includegraphics[width=0.098\textwidth,height=0.098\textwidth]{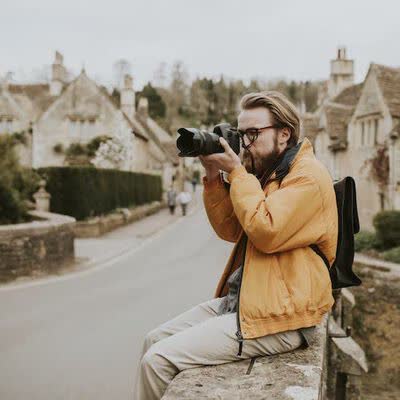} & & & \\
    \frame{\includegraphics[width=0.098\textwidth,height=0.098\textwidth]{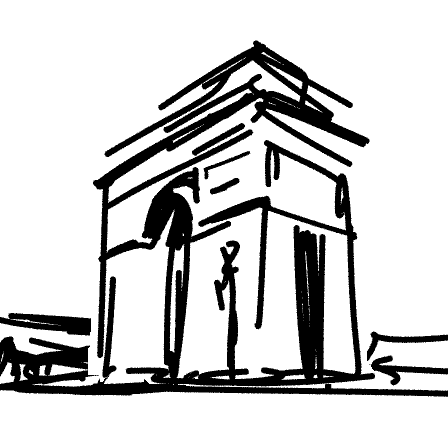}} &
    \frame{\includegraphics[width=0.098\textwidth,height=0.098\textwidth]{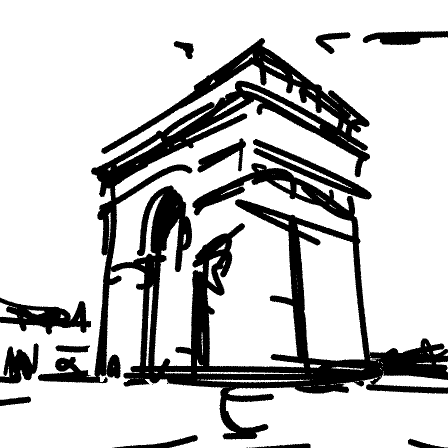}} &
    \frame{\includegraphics[width=0.098\textwidth,height=0.098\textwidth]{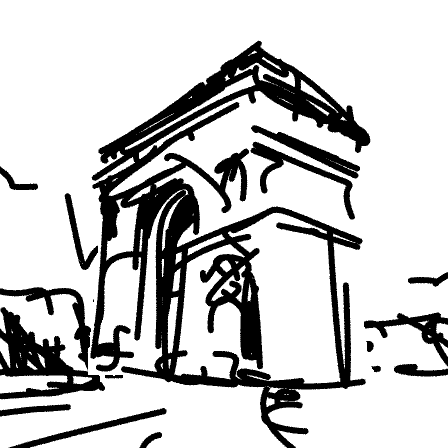}} &
    \frame{\includegraphics[width=0.098\textwidth,height=0.098\textwidth]{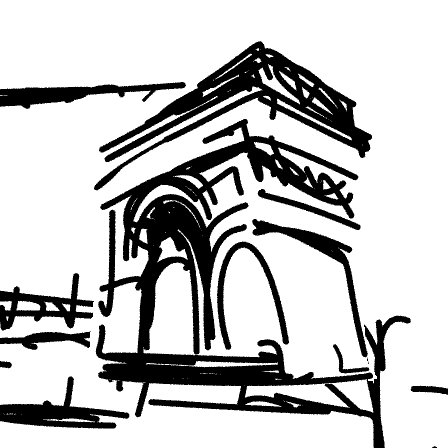}} &
    \hspace{0.5cm}
    \frame{\includegraphics[width=0.098\textwidth,height=0.098\textwidth]{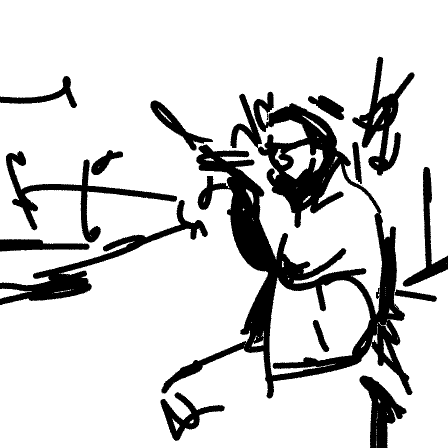}} &
    \frame{\includegraphics[width=0.098\textwidth,height=0.098\textwidth]{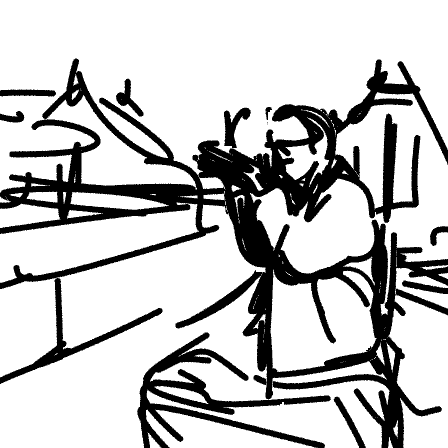}} &
    \frame{\includegraphics[width=0.098\textwidth,height=0.098\textwidth]{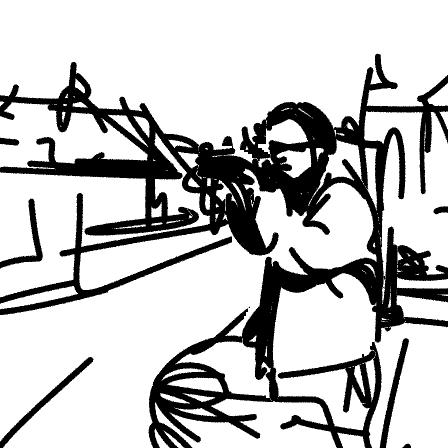}} &
    \frame{\includegraphics[width=0.098\textwidth,height=0.098\textwidth]{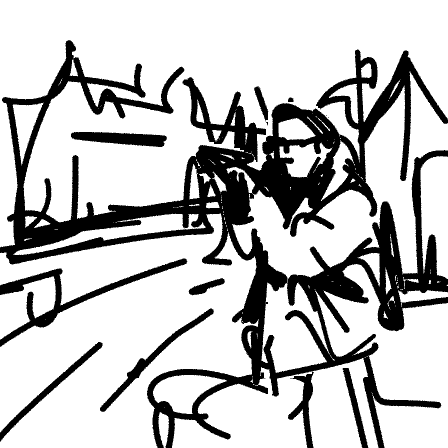}} \\
    
    \frame{\includegraphics[width=0.098\textwidth,height=0.098\textwidth]{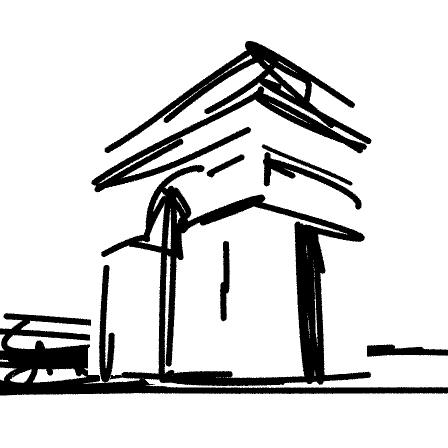}} &
    \frame{\includegraphics[width=0.098\textwidth,height=0.098\textwidth]{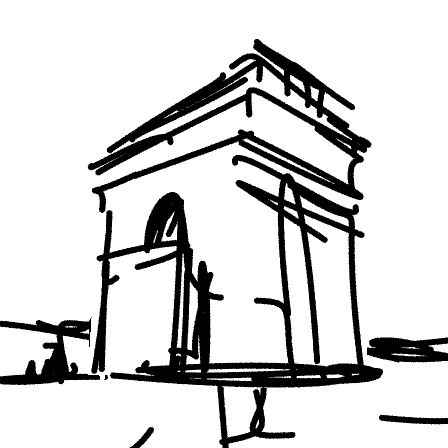}} &
    \frame{\includegraphics[width=0.098\textwidth,height=0.098\textwidth]{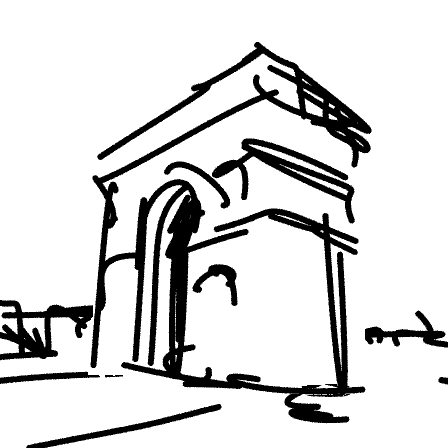}} &
    \frame{\includegraphics[width=0.098\textwidth,height=0.098\textwidth]{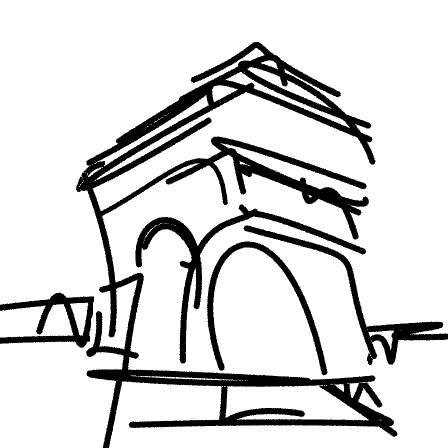}} &
    \hspace{0.5cm}
    \frame{\includegraphics[width=0.098\textwidth,height=0.098\textwidth]{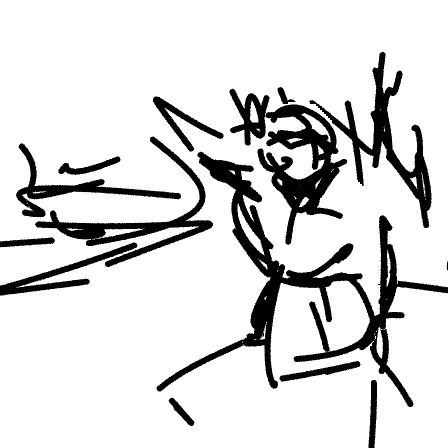}} &
    \frame{\includegraphics[width=0.098\textwidth,height=0.098\textwidth]{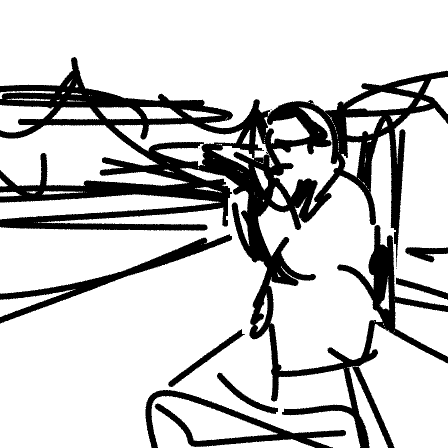}} &
    \frame{\includegraphics[width=0.098\textwidth,height=0.098\textwidth]{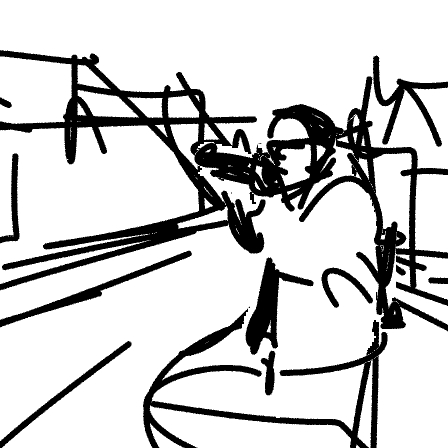}} &
    \frame{\includegraphics[width=0.098\textwidth,height=0.098\textwidth]{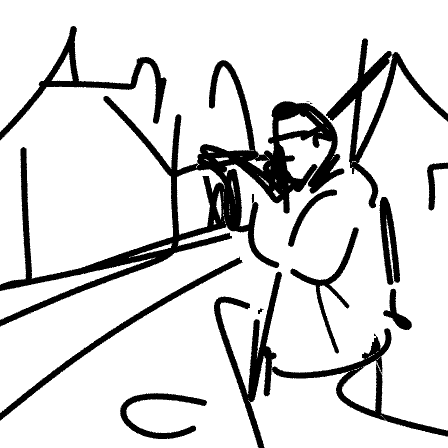}} \\
    
    \frame{\includegraphics[width=0.098\textwidth,height=0.098\textwidth]{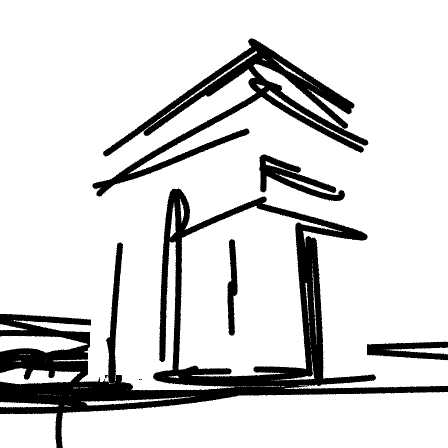}} &
    \frame{\includegraphics[width=0.098\textwidth,height=0.098\textwidth]{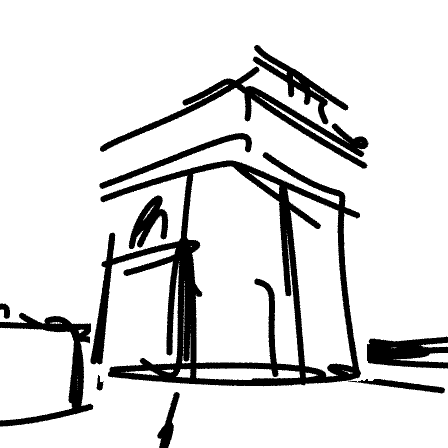}} &
    \frame{\includegraphics[width=0.098\textwidth,height=0.098\textwidth]{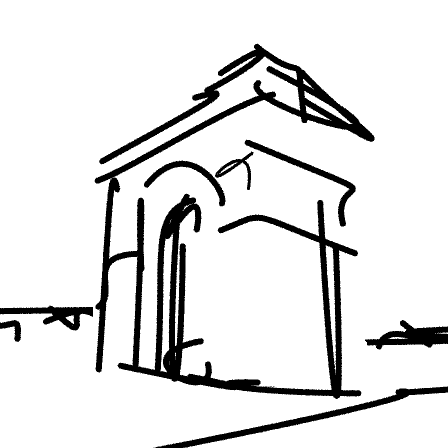}} &
    \frame{\includegraphics[width=0.098\textwidth,height=0.098\textwidth]{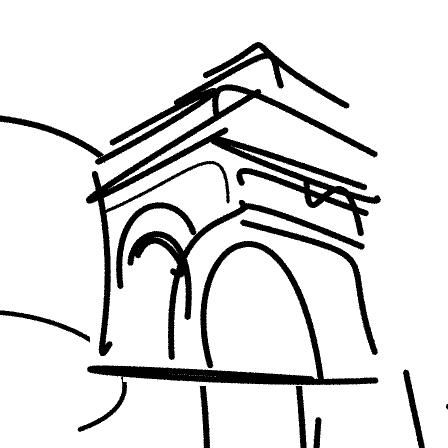}} &
    \hspace{0.5cm}
    \frame{\includegraphics[width=0.098\textwidth,height=0.098\textwidth]{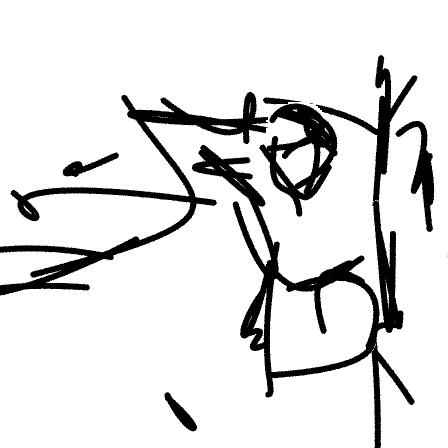}} &
    \frame{\includegraphics[width=0.098\textwidth,height=0.098\textwidth]{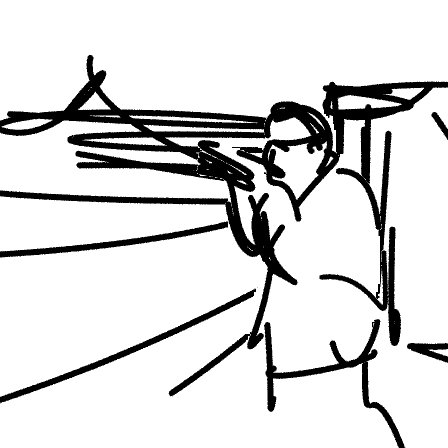}} &
    \frame{\includegraphics[width=0.098\textwidth,height=0.098\textwidth]{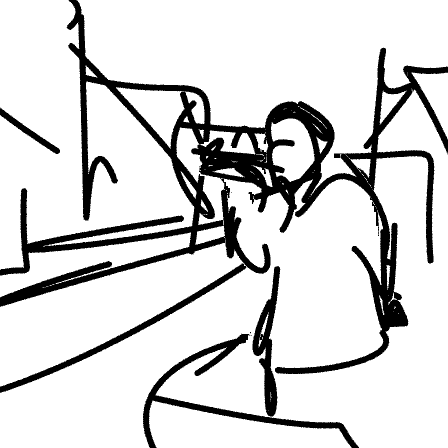}} &
    \frame{\includegraphics[width=0.098\textwidth,height=0.098\textwidth]{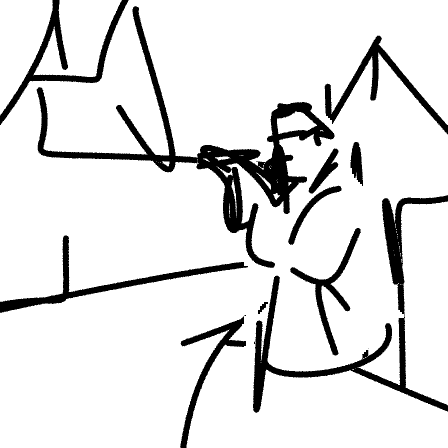}} \\
    
    \frame{\includegraphics[width=0.098\textwidth,height=0.098\textwidth]{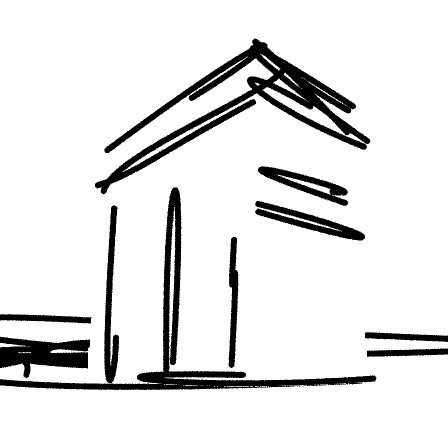}} &
    \frame{\includegraphics[width=0.098\textwidth,height=0.098\textwidth]{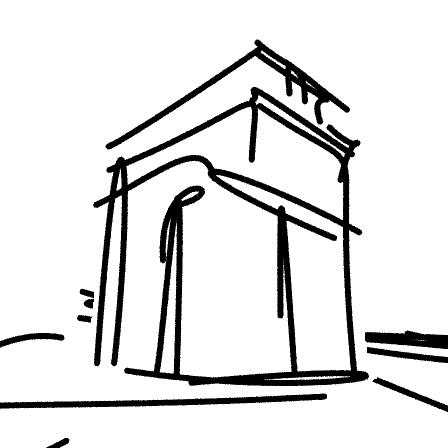}} &
    \frame{\includegraphics[width=0.098\textwidth,height=0.098\textwidth]{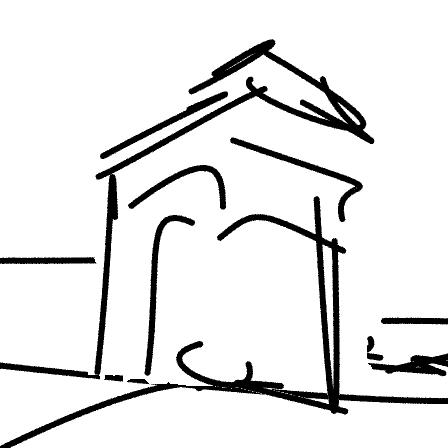}} &
    \frame{\includegraphics[width=0.098\textwidth,height=0.098\textwidth]{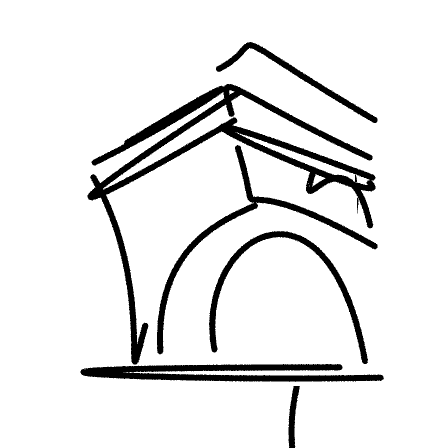}} &
    \hspace{0.5cm}
    \frame{\includegraphics[width=0.098\textwidth,height=0.098\textwidth]{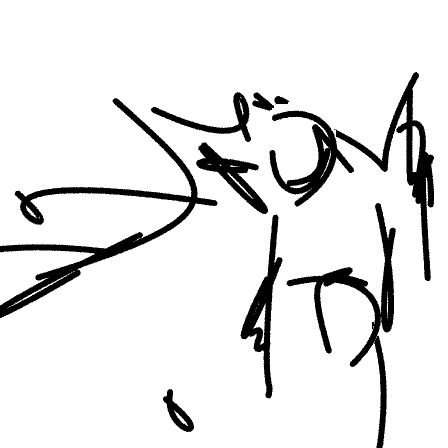}} &
    \frame{\includegraphics[width=0.098\textwidth,height=0.098\textwidth]{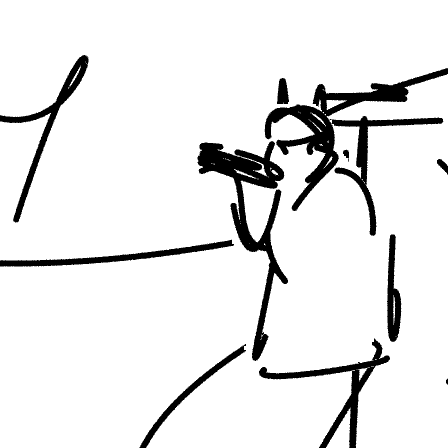}} &
    \frame{\includegraphics[width=0.098\textwidth,height=0.098\textwidth]{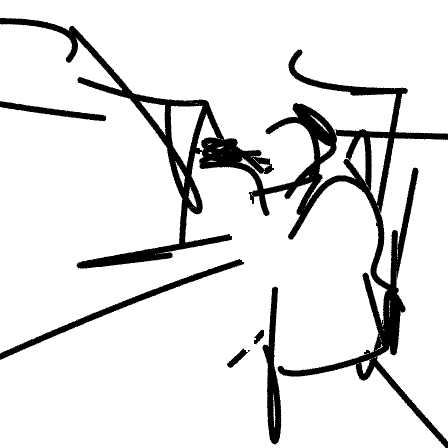}} &
    \frame{\includegraphics[width=0.098\textwidth,height=0.098\textwidth]{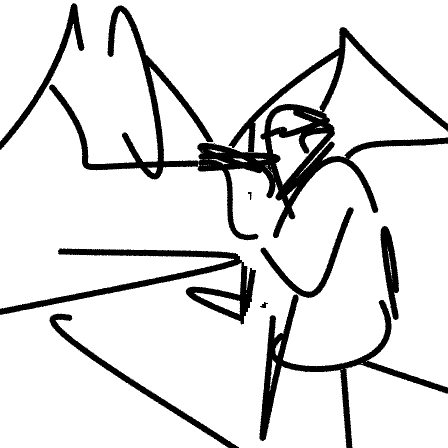}} \\

    \\
    \\
    
    \includegraphics[width=0.098\textwidth,height=0.098\textwidth]{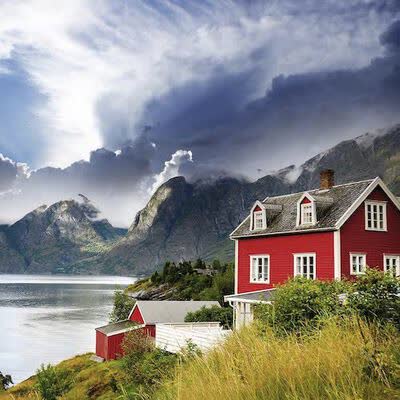} & & & &
    \hspace{0.5cm}
    \includegraphics[width=0.098\textwidth,height=0.098\textwidth]{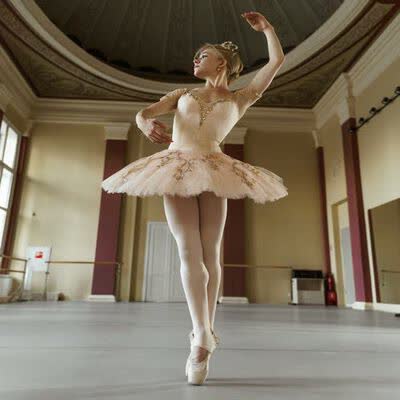} & & & \\

    \frame{\includegraphics[width=0.098\textwidth,height=0.098\textwidth]{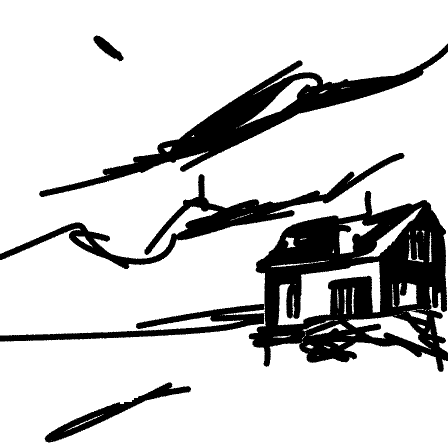}} &
    \frame{\includegraphics[width=0.098\textwidth,height=0.098\textwidth]{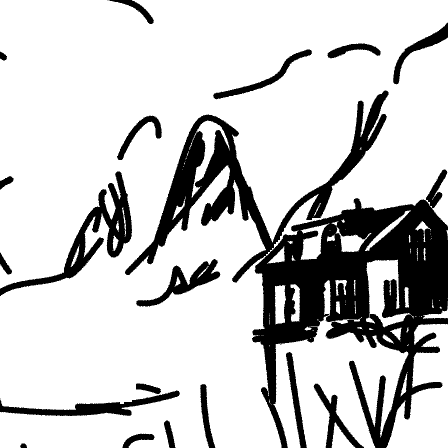}} &
    \frame{\includegraphics[width=0.098\textwidth,height=0.098\textwidth]{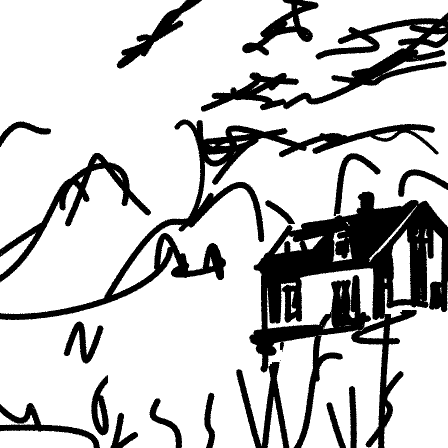}} &
    \frame{\includegraphics[width=0.098\textwidth,height=0.098\textwidth]{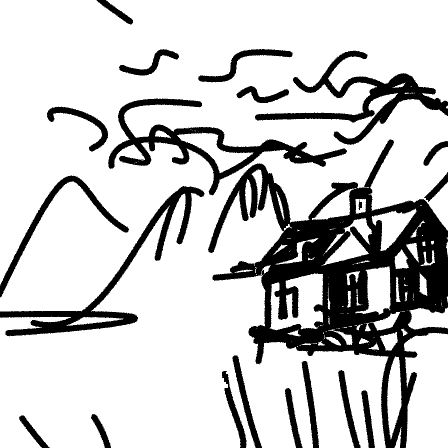}} &
    \hspace{0.5cm}
    \frame{\includegraphics[width=0.098\textwidth,height=0.098\textwidth]{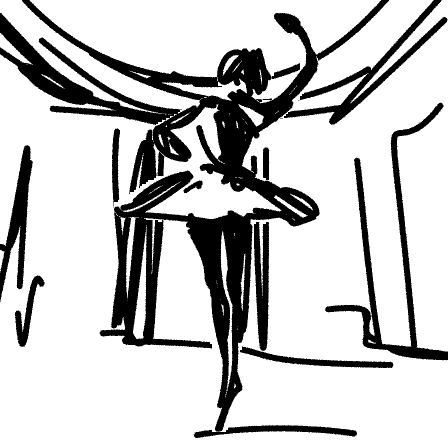}} &
    \frame{\includegraphics[width=0.098\textwidth,height=0.098\textwidth]{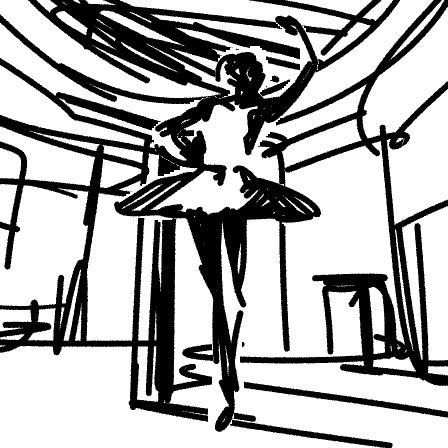}} &
    \frame{\includegraphics[width=0.098\textwidth,height=0.098\textwidth]{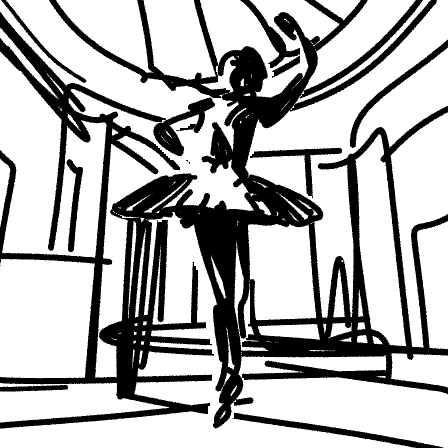}} &
    \frame{\includegraphics[width=0.098\textwidth,height=0.098\textwidth]{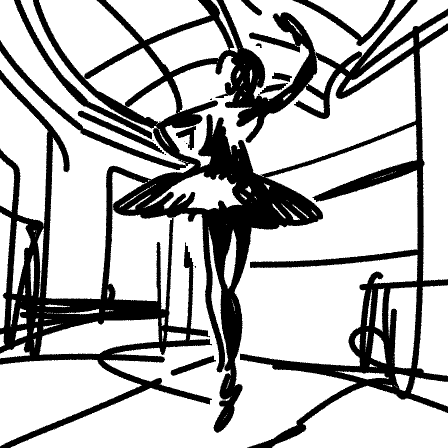}} \\
    
    \frame{\includegraphics[width=0.098\textwidth,height=0.098\textwidth]{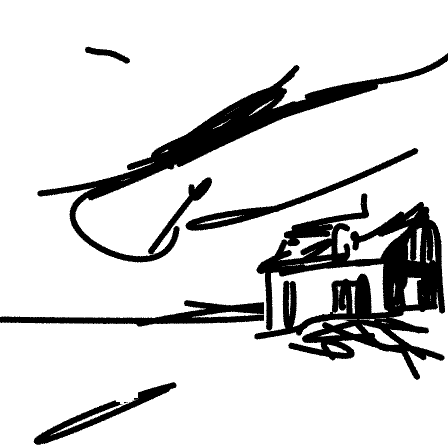}} &
    \frame{\includegraphics[width=0.098\textwidth,height=0.098\textwidth]{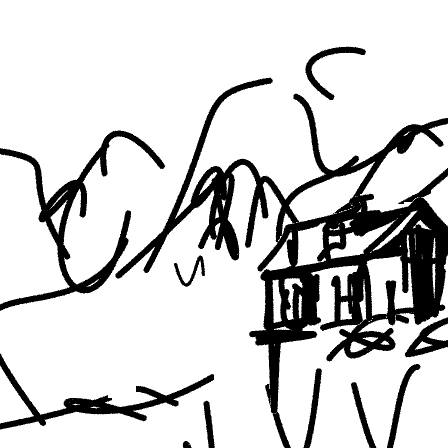}} &
    \frame{\includegraphics[width=0.098\textwidth,height=0.098\textwidth]{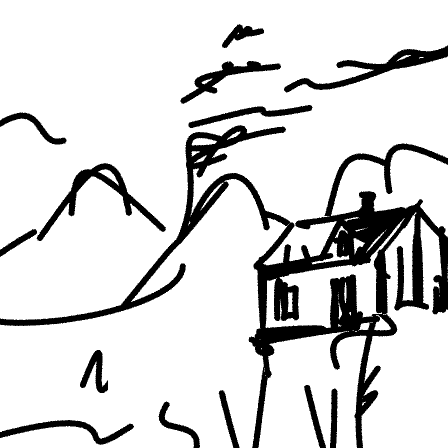}} &
    \frame{\includegraphics[width=0.098\textwidth,height=0.098\textwidth]{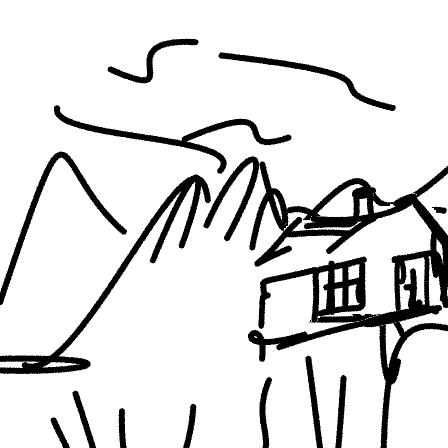}} &
    \hspace{0.5cm}
    \frame{\includegraphics[width=0.098\textwidth,height=0.098\textwidth]{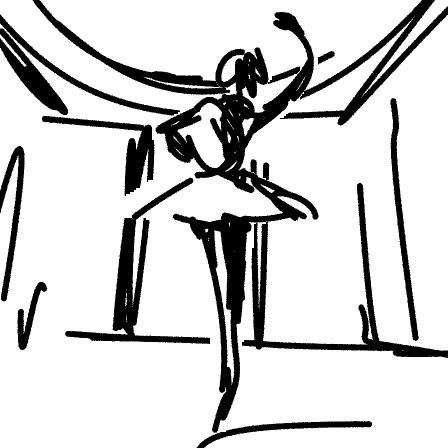}} &
    \frame{\includegraphics[width=0.098\textwidth,height=0.098\textwidth]{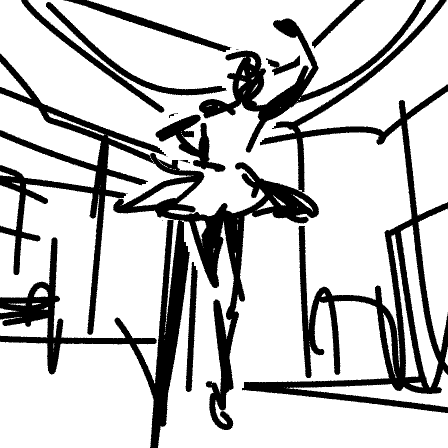}} &
    \frame{\includegraphics[width=0.098\textwidth,height=0.098\textwidth]{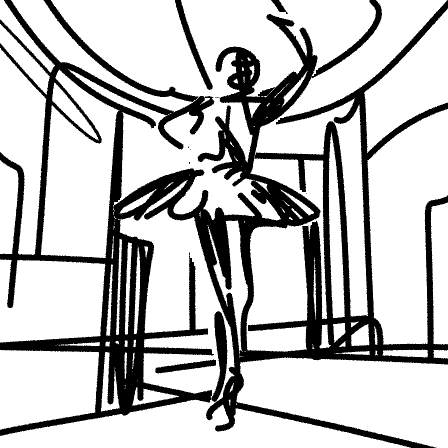}} &
    \frame{\includegraphics[width=0.098\textwidth,height=0.098\textwidth]{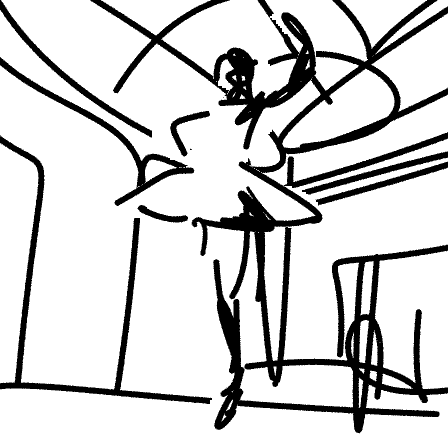}} \\
    
    \frame{\includegraphics[width=0.098\textwidth,height=0.098\textwidth]{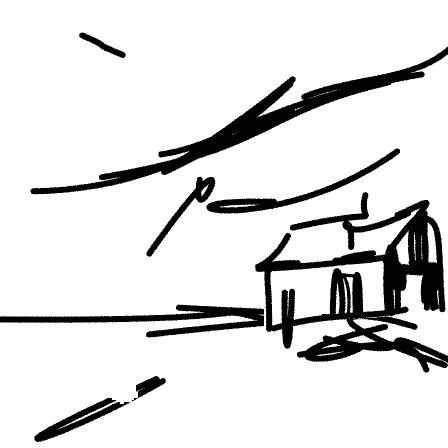}} &
    \frame{\includegraphics[width=0.098\textwidth,height=0.098\textwidth]{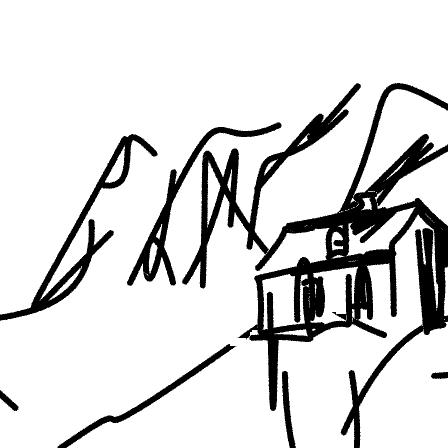}} &
    \frame{\includegraphics[width=0.098\textwidth,height=0.098\textwidth]{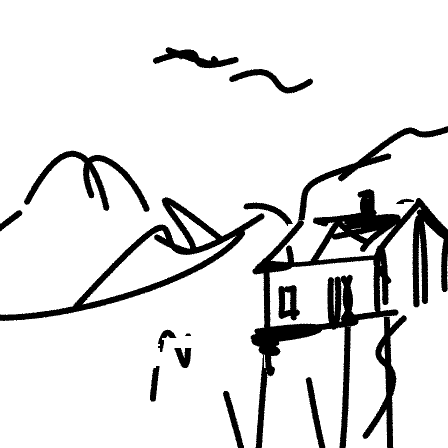}} &
    \frame{\includegraphics[width=0.098\textwidth,height=0.098\textwidth]{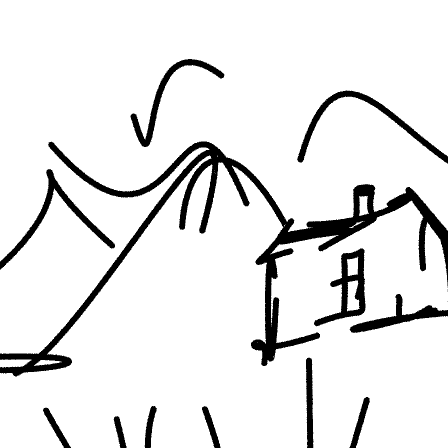}} &
    \hspace{0.5cm}
    \frame{\includegraphics[width=0.098\textwidth,height=0.098\textwidth]{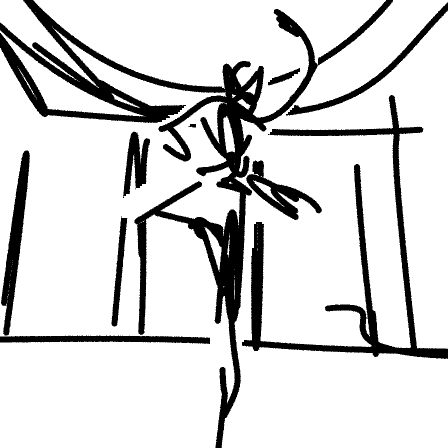}} &
    \frame{\includegraphics[width=0.098\textwidth,height=0.098\textwidth]{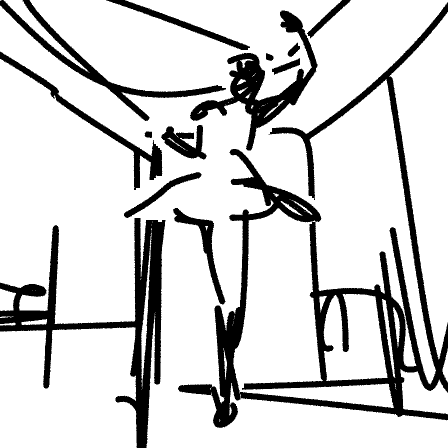}} &
    \frame{\includegraphics[width=0.098\textwidth,height=0.098\textwidth]{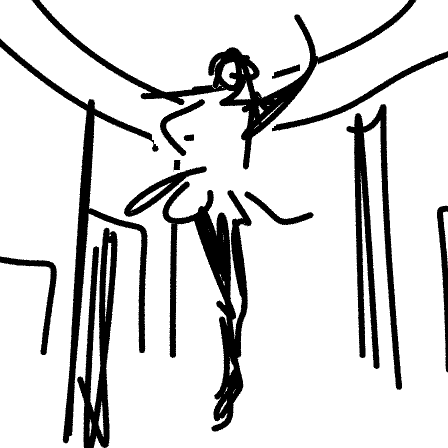}} &
    \frame{\includegraphics[width=0.098\textwidth,height=0.098\textwidth]{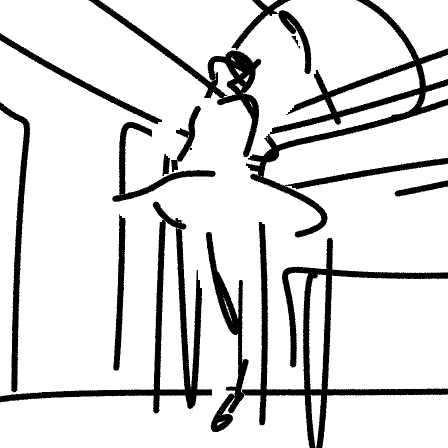}} \\
    
    \frame{\includegraphics[width=0.098\textwidth,height=0.098\textwidth]{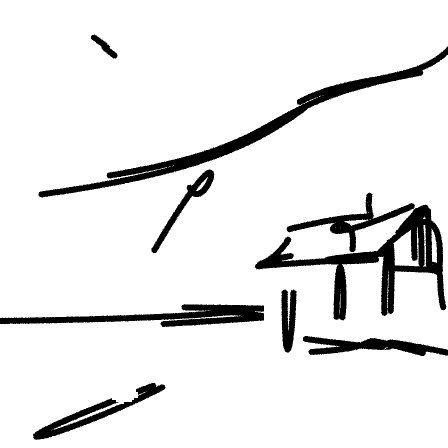}} &
    \frame{\includegraphics[width=0.098\textwidth,height=0.098\textwidth]{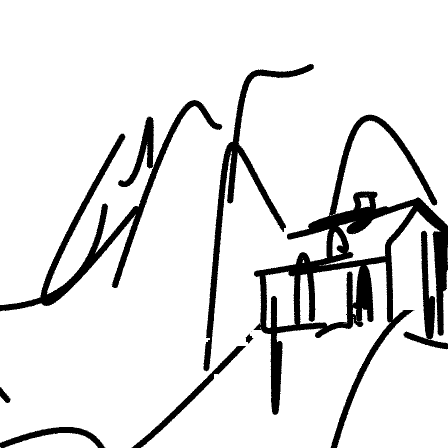}} &
    \frame{\includegraphics[width=0.098\textwidth,height=0.098\textwidth]{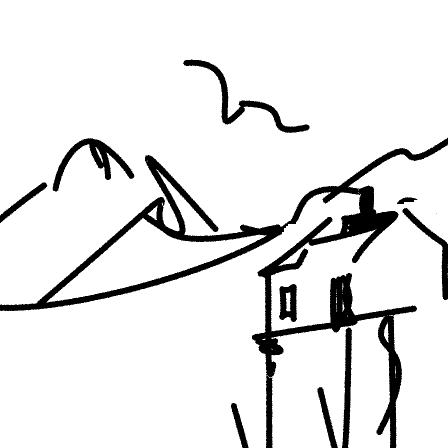}} &
    \frame{\includegraphics[width=0.098\textwidth,height=0.098\textwidth]{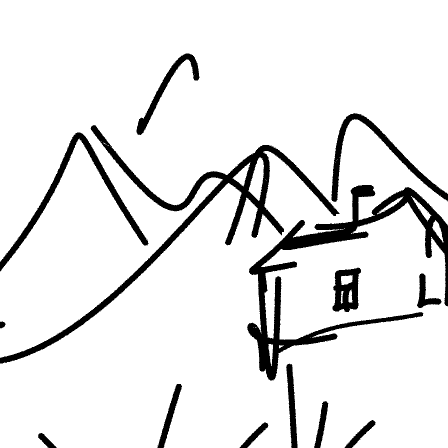}} &
    \hspace{0.5cm}
    \frame{\includegraphics[width=0.098\textwidth,height=0.098\textwidth]{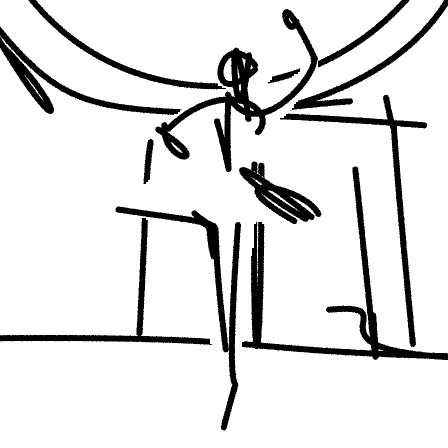}} &
    \frame{\includegraphics[width=0.098\textwidth,height=0.098\textwidth]{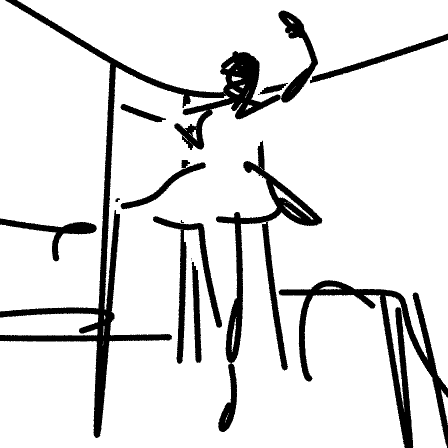}} &
    \frame{\includegraphics[width=0.098\textwidth,height=0.098\textwidth]{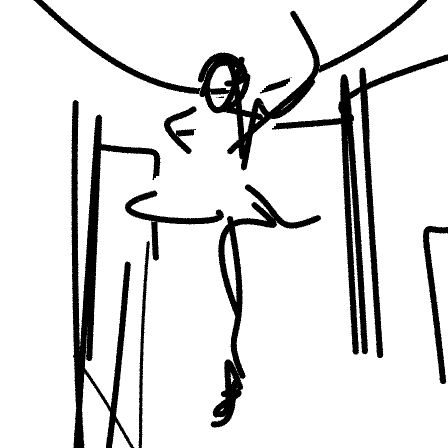}} &
    \frame{\includegraphics[width=0.098\textwidth,height=0.098\textwidth]{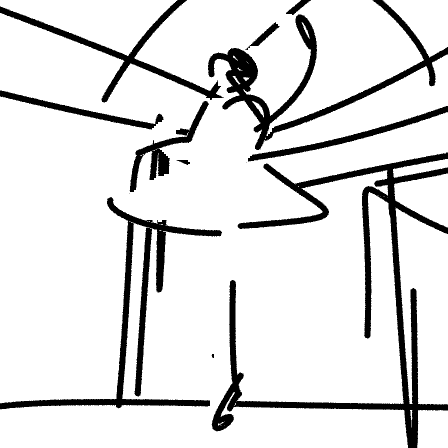}} \\

    \end{tabular}
    \caption{The $4\times4$ matrix of sketches produced by our method. Columns from left to right illustrate the change in fidelity, from precise to loose, and rows from top to bottom illustrate the visual simplification.}
    
    \label{fig:matrix1}
\end{figure*}

\begin{figure*}
    \centering
    
    \begin{tabular}{c c c c c c c c}

    \includegraphics[width=0.098\textwidth,height=0.098\textwidth]{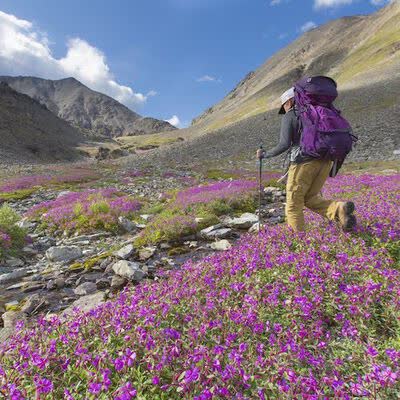} & & & &
    \hspace{0.5cm}
    \includegraphics[width=0.098\textwidth,height=0.098\textwidth]{clipascene/figs/inputs/dog.jpg} & & & \\
    
    \frame{\includegraphics[width=0.098\textwidth,height=0.098\textwidth]{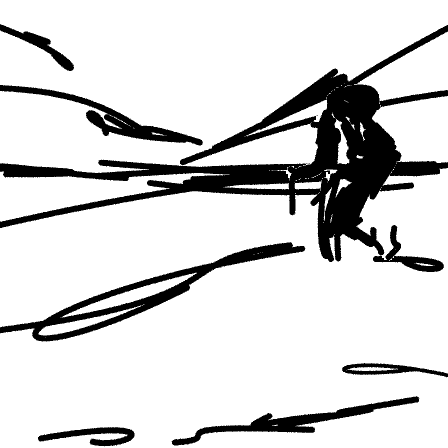}} &
    \frame{\includegraphics[width=0.098\textwidth,height=0.098\textwidth]{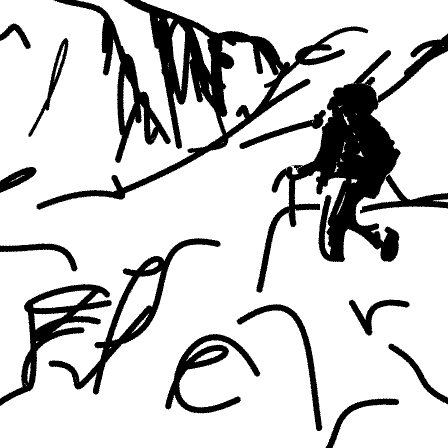}} &
    \frame{\includegraphics[width=0.098\textwidth,height=0.098\textwidth]{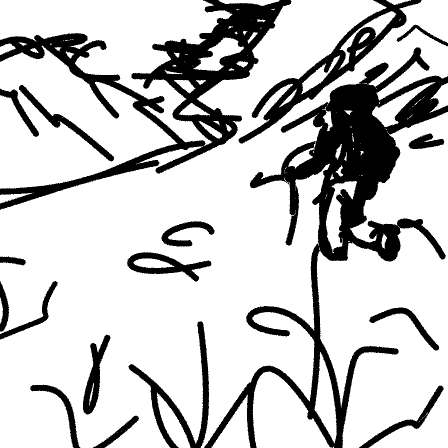}} &
    \frame{\includegraphics[width=0.098\textwidth,height=0.098\textwidth]{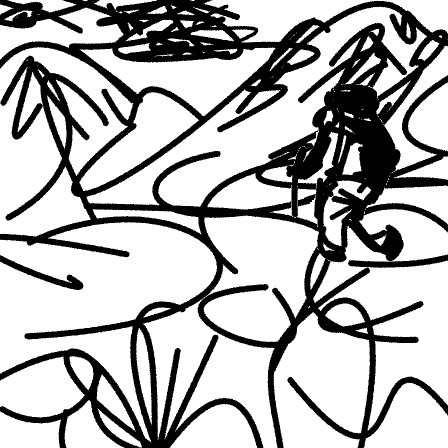}} &
    \hspace{0.5cm}
    \frame{\includegraphics[width=0.098\textwidth,height=0.098\textwidth]{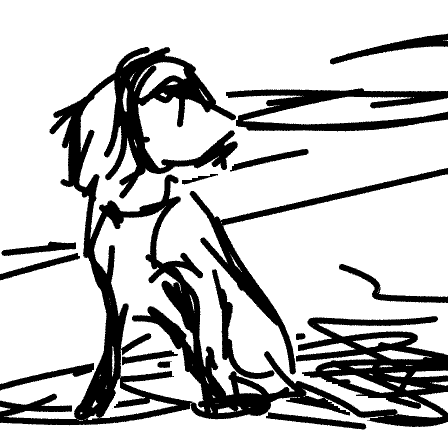}} &
    \frame{\includegraphics[width=0.098\textwidth,height=0.098\textwidth]{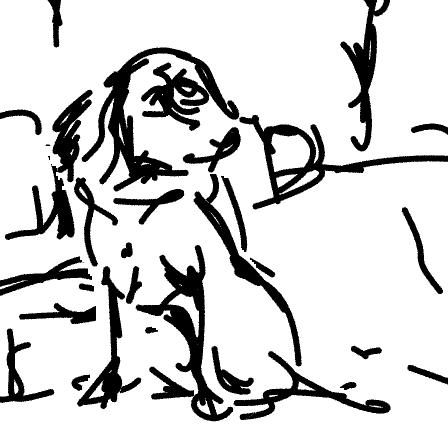}} &
    \frame{\includegraphics[width=0.098\textwidth,height=0.098\textwidth]{clipascene/figs/matrices_black/dog_row0col2_black.png}} &
    \frame{\includegraphics[width=0.098\textwidth,height=0.098\textwidth]{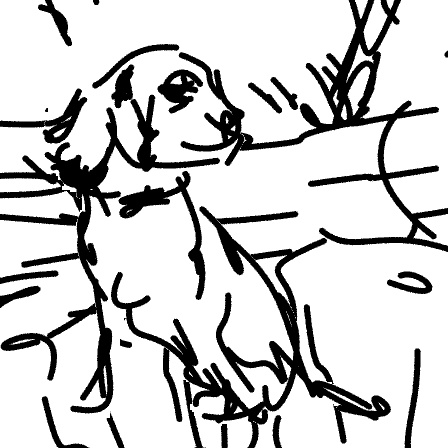}} \\
    
    \frame{\includegraphics[width=0.098\textwidth,height=0.098\textwidth]{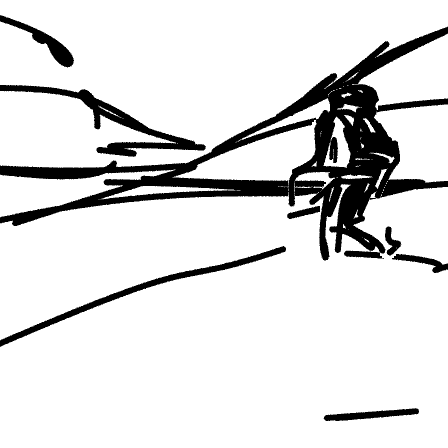}} &
    \frame{\includegraphics[width=0.098\textwidth,height=0.098\textwidth]{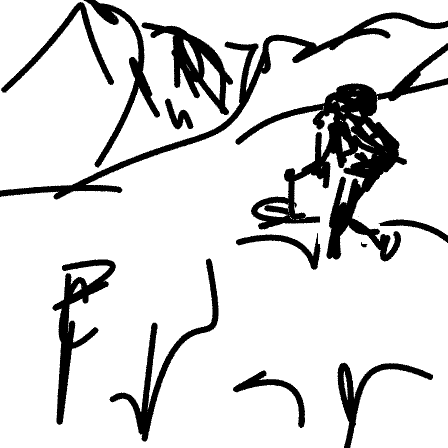}} &
    \frame{\includegraphics[width=0.098\textwidth,height=0.098\textwidth]{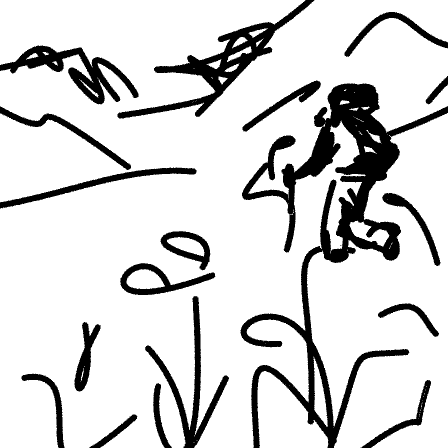}} &
    \frame{\includegraphics[width=0.098\textwidth,height=0.098\textwidth]{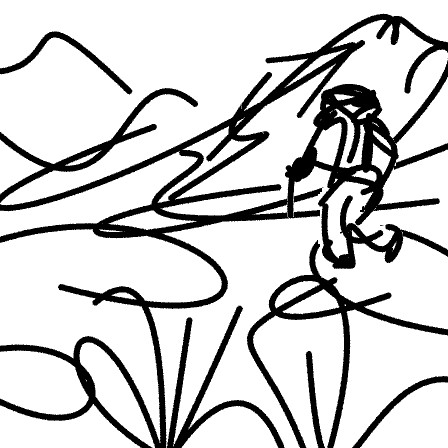}} &
    \hspace{0.5cm}
    \frame{\includegraphics[width=0.098\textwidth,height=0.098\textwidth]{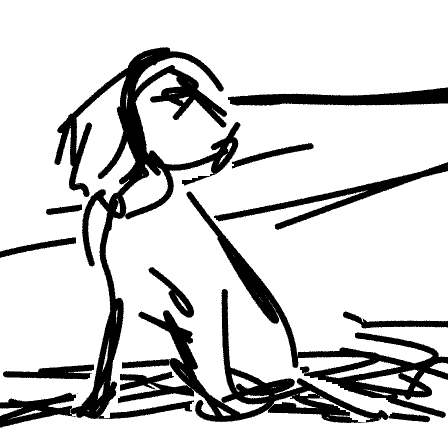}} &
    \frame{\includegraphics[width=0.098\textwidth,height=0.098\textwidth]{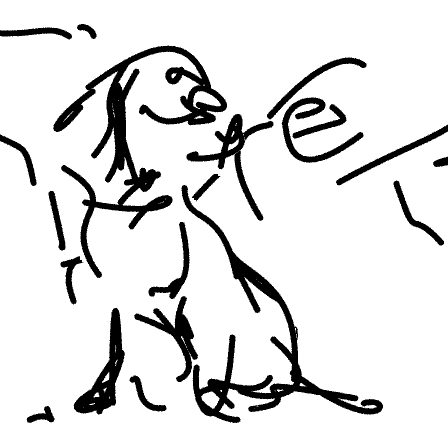}} &
    \frame{\includegraphics[width=0.098\textwidth,height=0.098\textwidth]{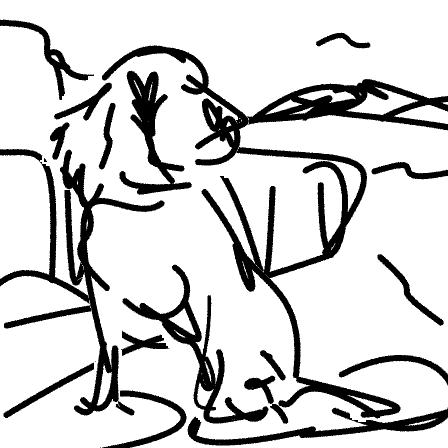}} &
    \frame{\includegraphics[width=0.098\textwidth,height=0.098\textwidth]{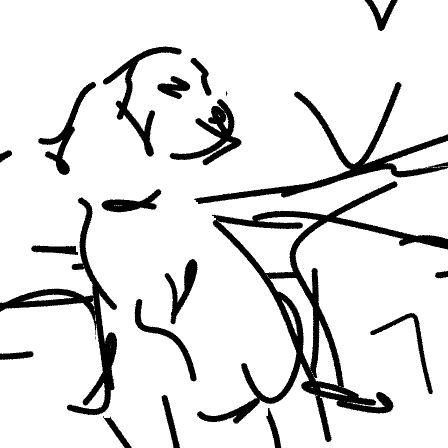}} \\
    
    \frame{\includegraphics[width=0.098\textwidth,height=0.098\textwidth]{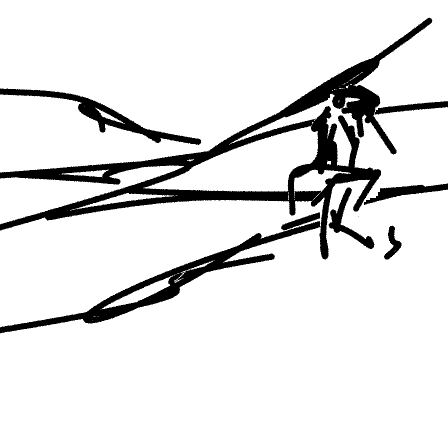}} &
    \frame{\includegraphics[width=0.098\textwidth,height=0.098\textwidth]{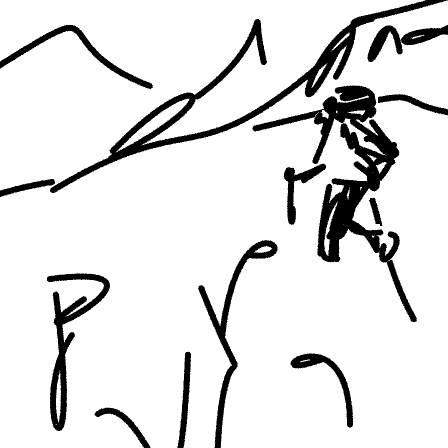}} &
    \frame{\includegraphics[width=0.098\textwidth,height=0.098\textwidth]{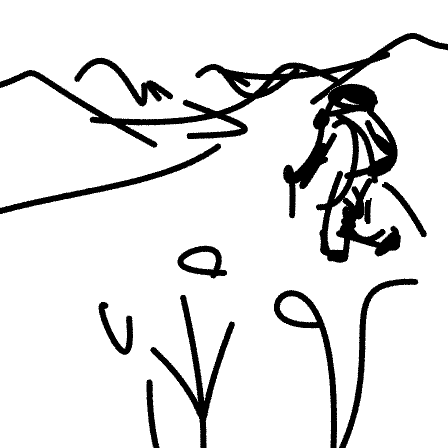}} &
    \frame{\includegraphics[width=0.098\textwidth,height=0.098\textwidth]{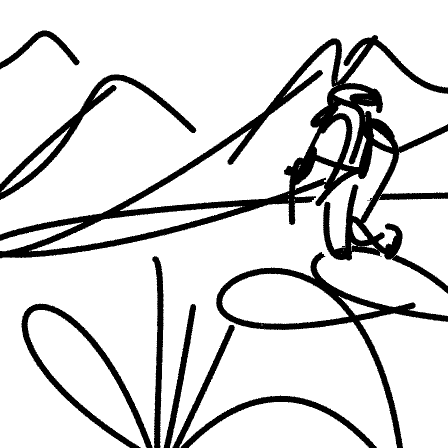}} &
    \hspace{0.5cm}
    \frame{\includegraphics[width=0.098\textwidth,height=0.098\textwidth]{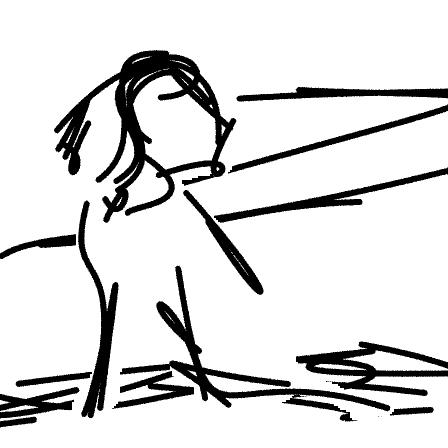}} &
    \frame{\includegraphics[width=0.098\textwidth,height=0.098\textwidth]{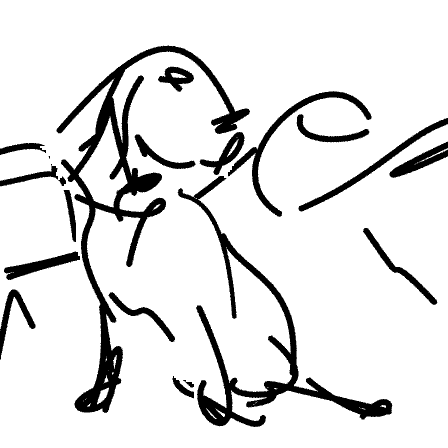}} &
    \frame{\includegraphics[width=0.098\textwidth,height=0.098\textwidth]{clipascene/figs/matrices_black/dog_row2col2_black.png}} &
    \frame{\includegraphics[width=0.098\textwidth,height=0.098\textwidth]{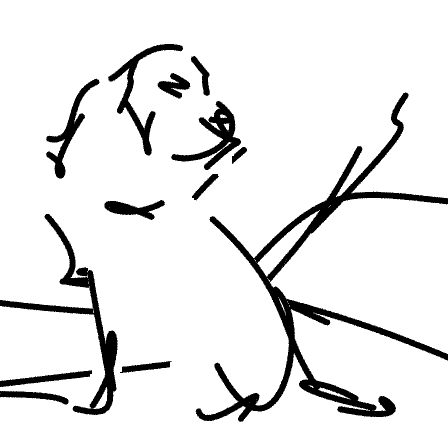}} \\
    
    \frame{\includegraphics[width=0.098\textwidth,height=0.098\textwidth]{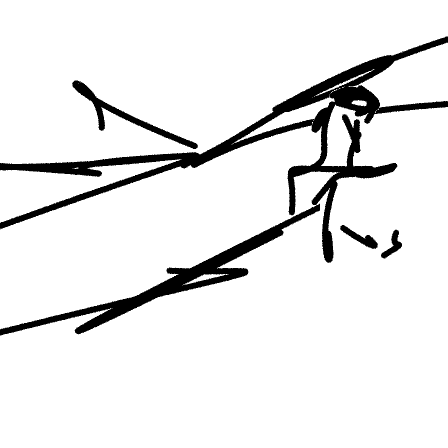}} &
    \frame{\includegraphics[width=0.098\textwidth,height=0.098\textwidth]{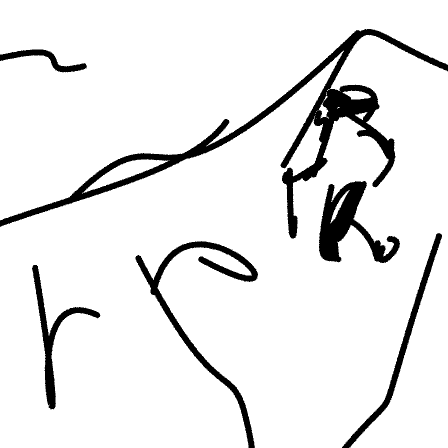}} &
    \frame{\includegraphics[width=0.098\textwidth,height=0.098\textwidth]{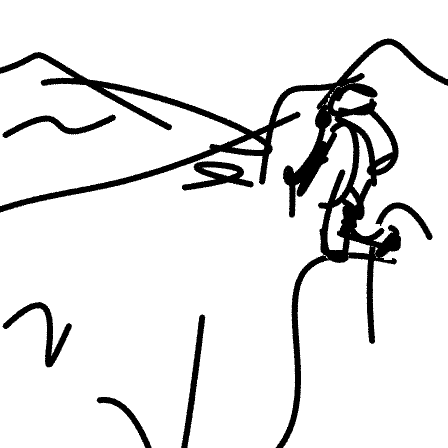}} &
    \frame{\includegraphics[width=0.098\textwidth,height=0.098\textwidth]{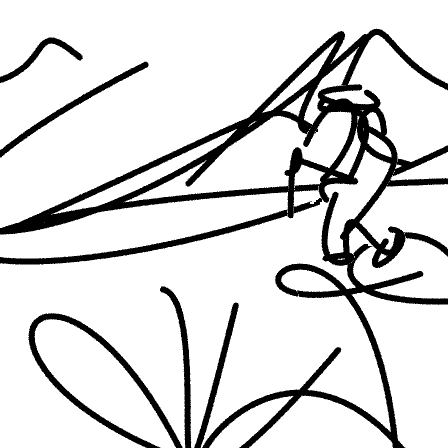}} &
    \hspace{0.5cm}
    \frame{\includegraphics[width=0.098\textwidth,height=0.098\textwidth]{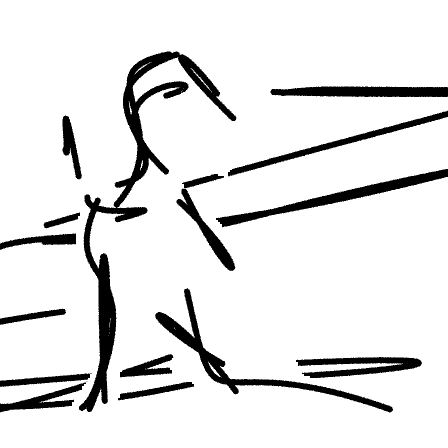}} &
    \frame{\includegraphics[width=0.098\textwidth,height=0.098\textwidth]{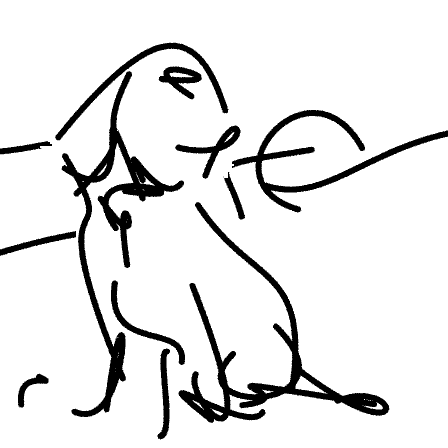}} &
    \frame{\includegraphics[width=0.098\textwidth,height=0.098\textwidth]{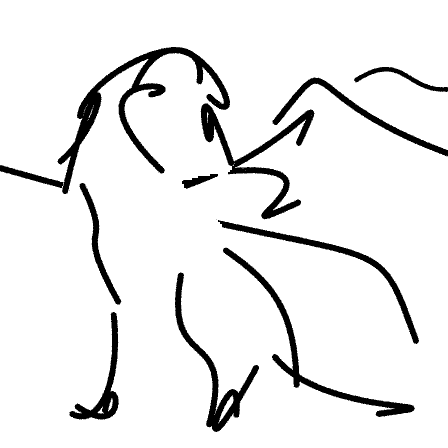}} &
    \frame{\includegraphics[width=0.098\textwidth,height=0.098\textwidth]{clipascene/figs/matrices_black/dog_row3col3_black.png}} \\

    \\
    \\
    
    \includegraphics[width=0.098\textwidth,height=0.098\textwidth]{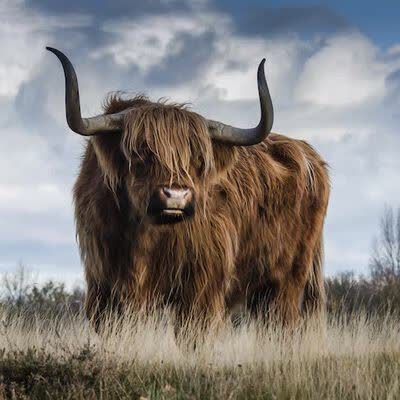} & & & &
    \hspace{0.5cm}
    \includegraphics[width=0.098\textwidth,height=0.098\textwidth]{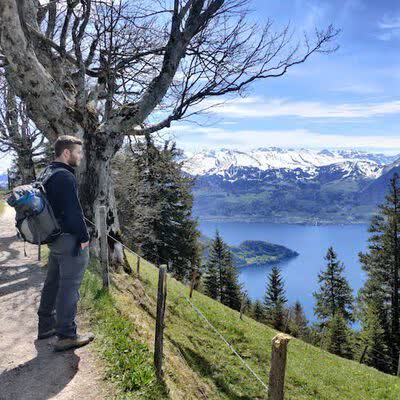} & & & \\

    \frame{\includegraphics[width=0.098\textwidth,height=0.098\textwidth]{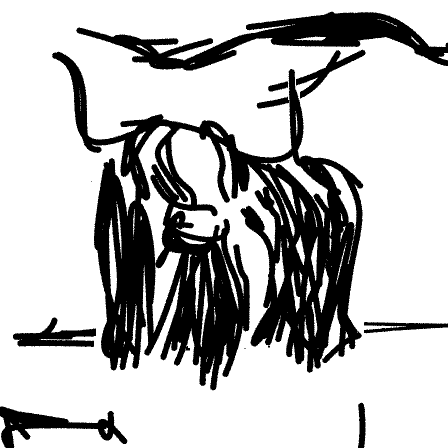}} &
    \frame{\includegraphics[width=0.098\textwidth,height=0.098\textwidth]{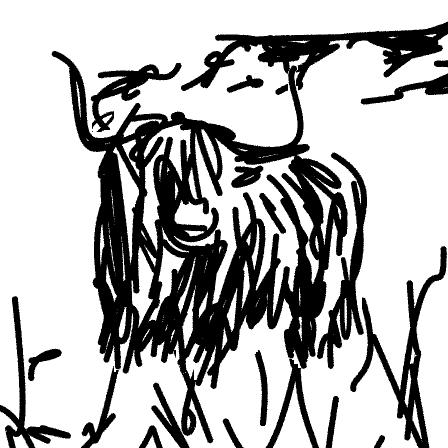}} &
    \frame{\includegraphics[width=0.098\textwidth,height=0.098\textwidth]{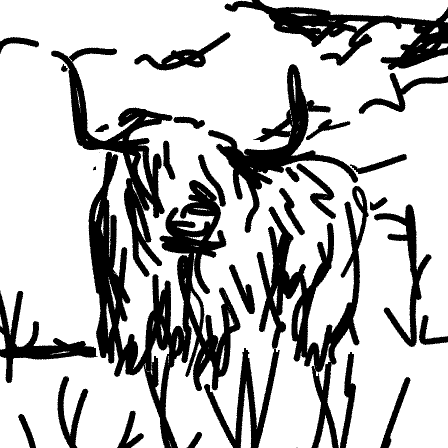}} &
    \frame{\includegraphics[width=0.098\textwidth,height=0.098\textwidth]{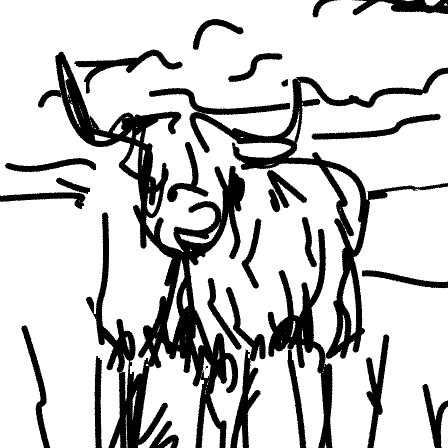}} &
    \hspace{0.5cm}
    \frame{\includegraphics[width=0.098\textwidth,height=0.098\textwidth]{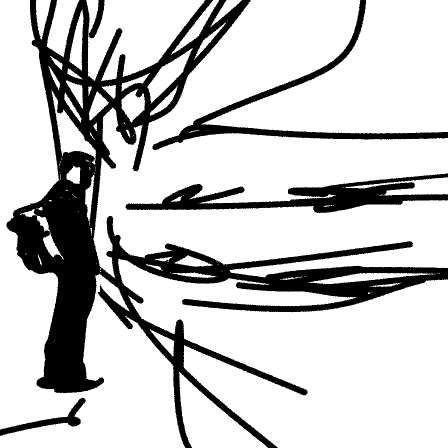}} &
    \frame{\includegraphics[width=0.098\textwidth,height=0.098\textwidth]{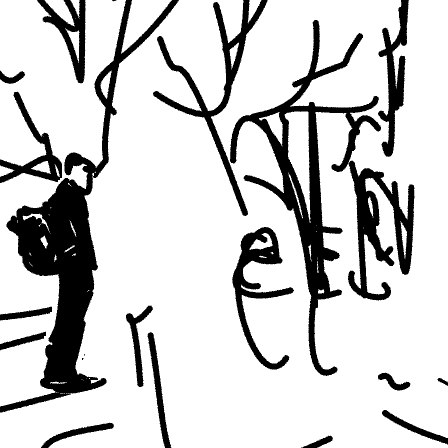}} &
    \frame{\includegraphics[width=0.098\textwidth,height=0.098\textwidth]{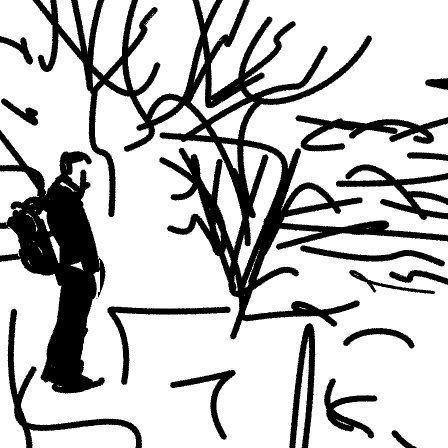}} &
    \frame{\includegraphics[width=0.098\textwidth,height=0.098\textwidth]{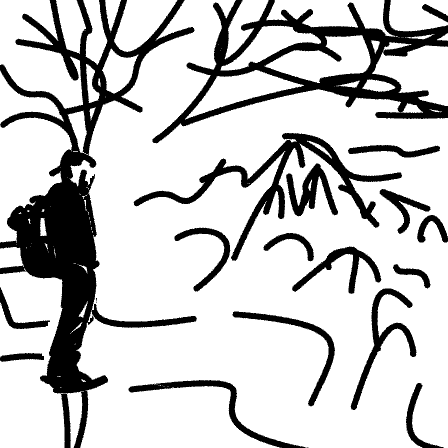}} \\
    
    \frame{\includegraphics[width=0.098\textwidth,height=0.098\textwidth]{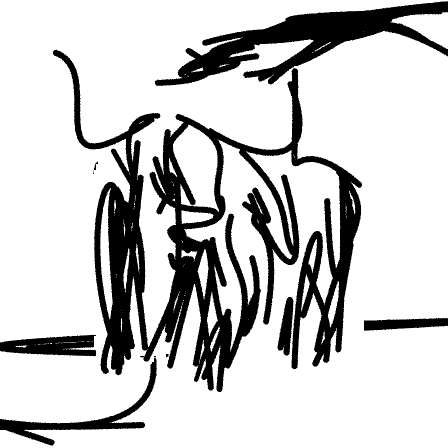}} &
    \frame{\includegraphics[width=0.098\textwidth,height=0.098\textwidth]{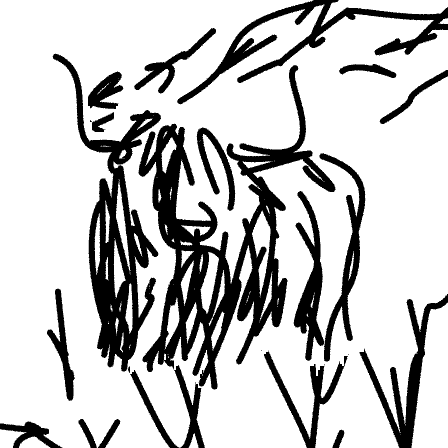}} &
    \frame{\includegraphics[width=0.098\textwidth,height=0.098\textwidth]{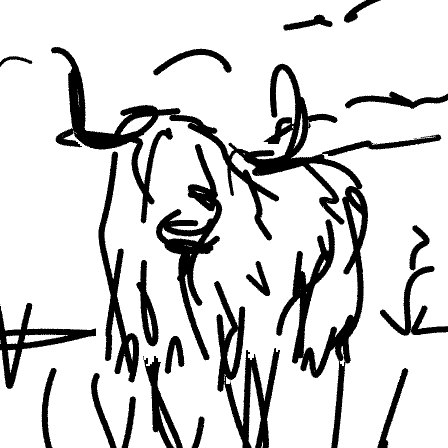}} &
    \frame{\includegraphics[width=0.098\textwidth,height=0.098\textwidth]{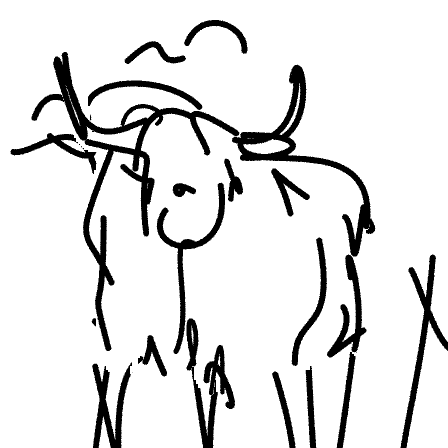}} &
    \hspace{0.5cm}
    \frame{\includegraphics[width=0.098\textwidth,height=0.098\textwidth]{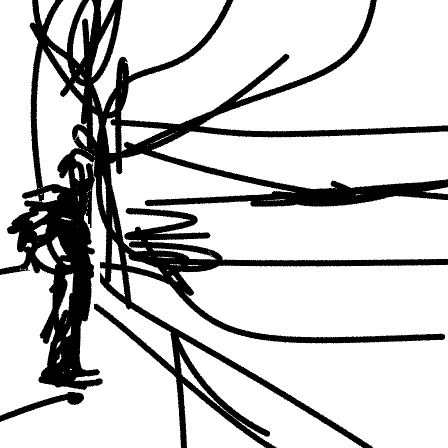}} &
    \frame{\includegraphics[width=0.098\textwidth,height=0.098\textwidth]{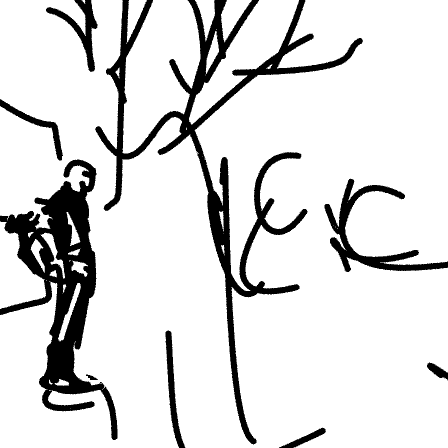}} &
    \frame{\includegraphics[width=0.098\textwidth,height=0.098\textwidth]{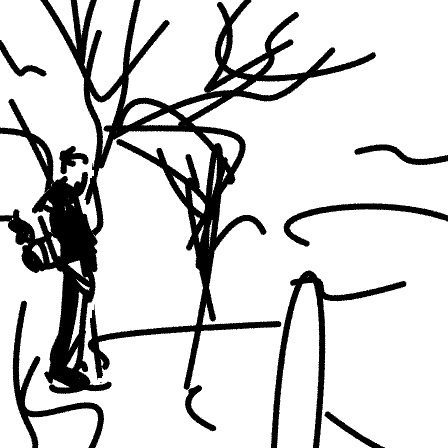}} &
    \frame{\includegraphics[width=0.098\textwidth,height=0.098\textwidth]{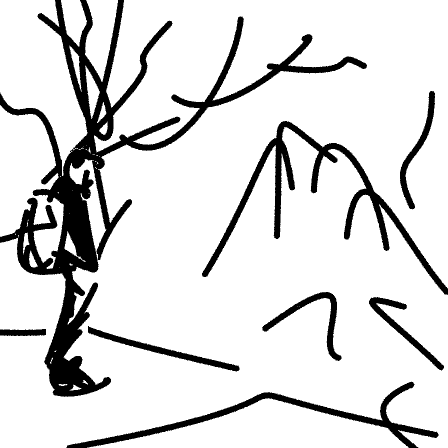}} \\
    
    \frame{\includegraphics[width=0.098\textwidth,height=0.098\textwidth]{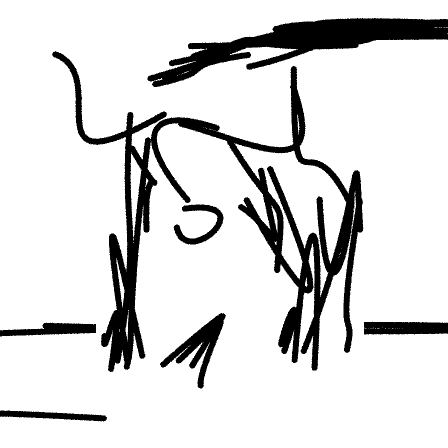}} &
    \frame{\includegraphics[width=0.098\textwidth,height=0.098\textwidth]{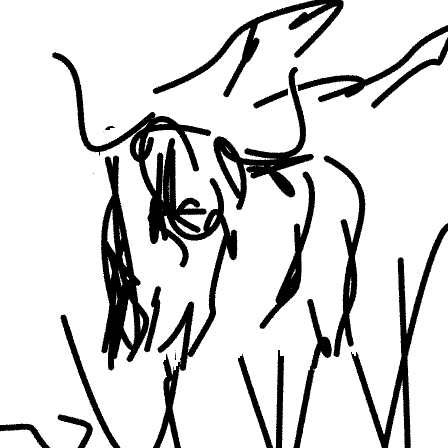}} &
    \frame{\includegraphics[width=0.098\textwidth,height=0.098\textwidth]{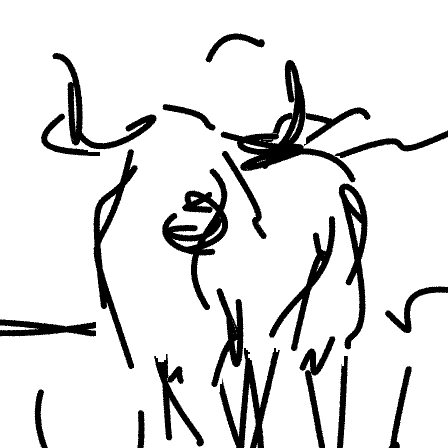}} &
    \frame{\includegraphics[width=0.098\textwidth,height=0.098\textwidth]{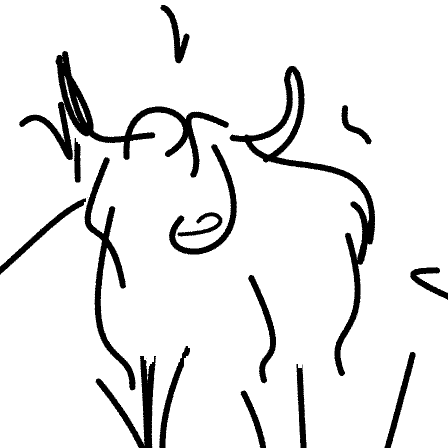}} &
    \hspace{0.5cm}
    \frame{\includegraphics[width=0.098\textwidth,height=0.098\textwidth]{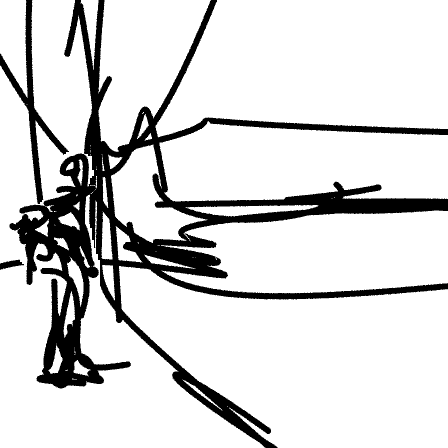}} &
    \frame{\includegraphics[width=0.098\textwidth,height=0.098\textwidth]{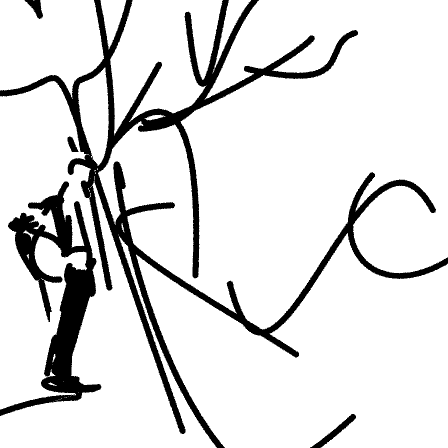}} &
    \frame{\includegraphics[width=0.098\textwidth,height=0.098\textwidth]{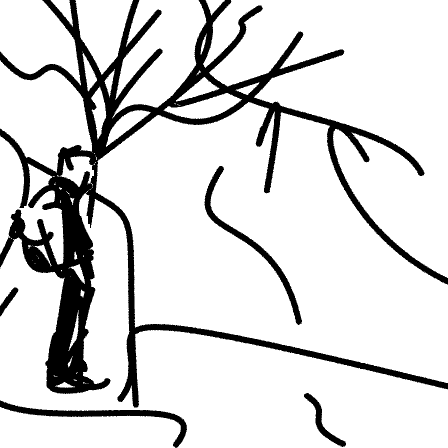}} &
    \frame{\includegraphics[width=0.098\textwidth,height=0.098\textwidth]{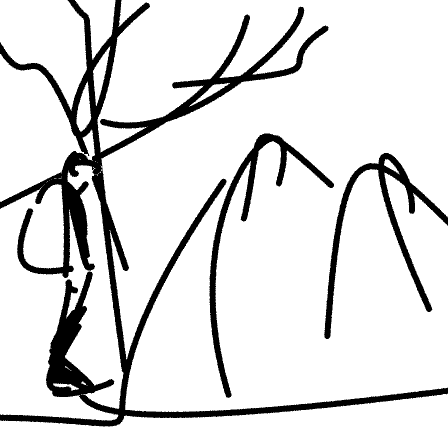}} \\
    
    \frame{\includegraphics[width=0.098\textwidth,height=0.098\textwidth]{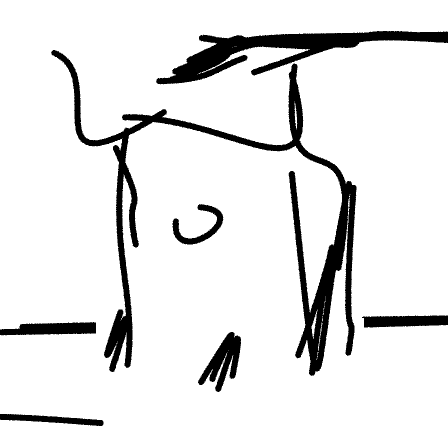}} &
    \frame{\includegraphics[width=0.098\textwidth,height=0.098\textwidth]{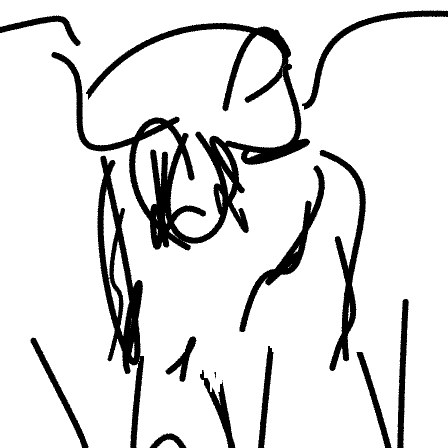}} &
    \frame{\includegraphics[width=0.098\textwidth,height=0.098\textwidth]{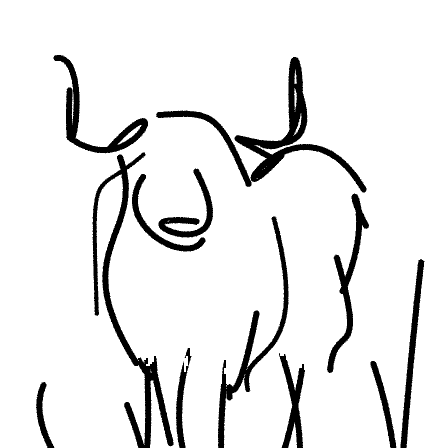}} &
    \frame{\includegraphics[width=0.098\textwidth,height=0.098\textwidth]{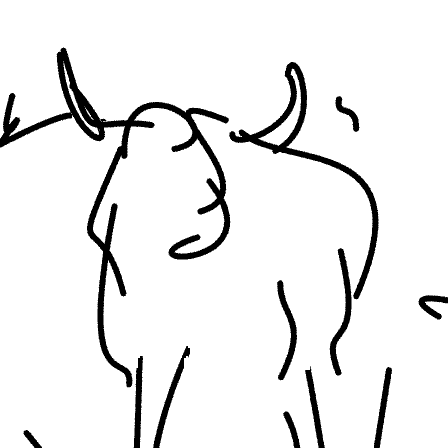}} &
    \hspace{0.5cm}
    \frame{\includegraphics[width=0.098\textwidth,height=0.098\textwidth]{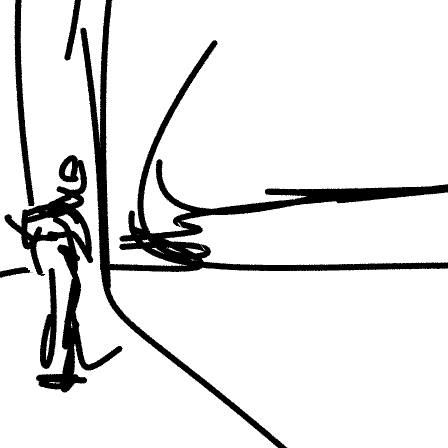}} &
    \frame{\includegraphics[width=0.098\textwidth,height=0.098\textwidth]{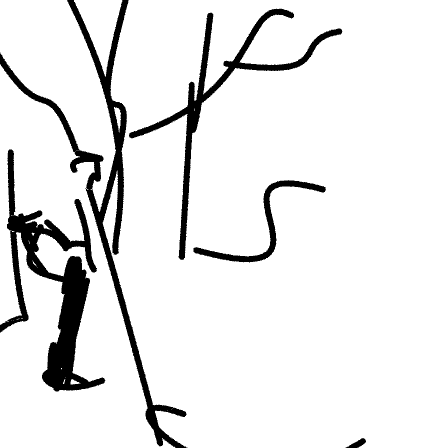}} &
    \frame{\includegraphics[width=0.098\textwidth,height=0.098\textwidth]{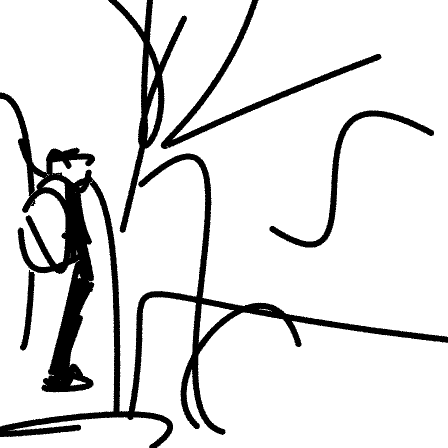}} &
    \frame{\includegraphics[width=0.098\textwidth,height=0.098\textwidth]{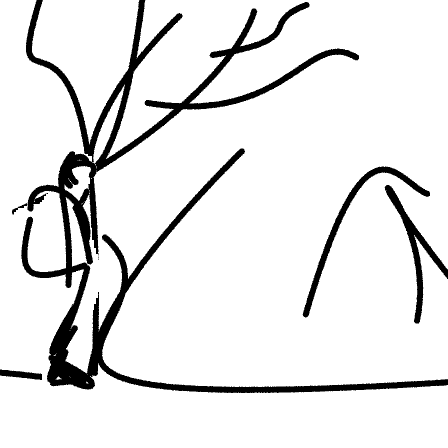}} \\

    \end{tabular}
    \caption{The $4\times4$ matrix of sketches produced by our method. Columns from left to right illustrate the change in fidelity, from precise to loose, and rows from top to bottom illustrate the visual simplification.}
    
    \label{fig:matrix2}
\end{figure*}

\begin{figure*}
    \centering
    
    \begin{tabular}{c c c c c c c c}

    \includegraphics[width=0.098\textwidth,height=0.098\textwidth]{clipascene/figs/inputs/woman_city.jpg} & & & &
    \hspace{0.5cm}
    \includegraphics[width=0.098\textwidth,height=0.098\textwidth]{clipascene/figs/inputs/black_woman.jpg} & & & \\
    
    \frame{\includegraphics[width=0.098\textwidth,height=0.098\textwidth]{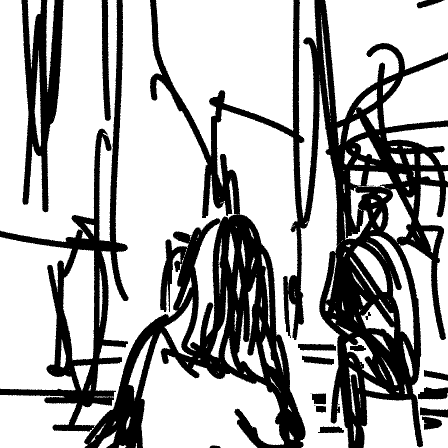}} &
    \frame{\includegraphics[width=0.098\textwidth,height=0.098\textwidth]{clipascene/figs/matrices_black/woman_city_row0col1_black.png}} &
    \frame{\includegraphics[width=0.098\textwidth,height=0.098\textwidth]{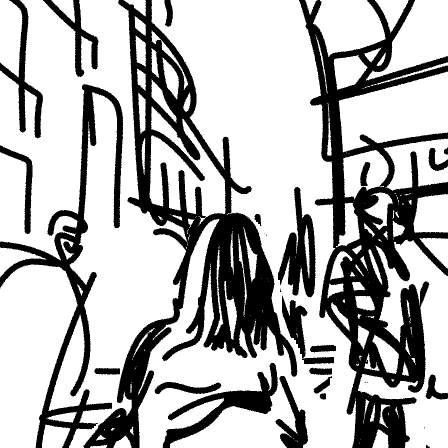}} &
    \frame{\includegraphics[width=0.098\textwidth,height=0.098\textwidth]{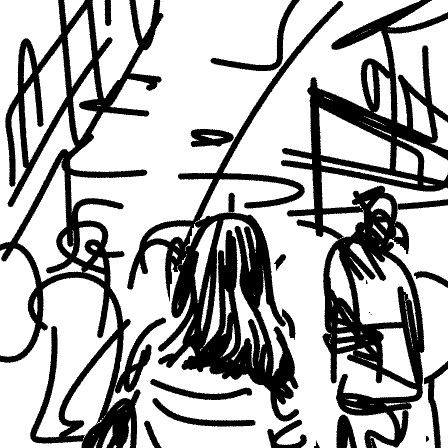}} &
    \hspace{0.5cm}
    \frame{\includegraphics[width=0.098\textwidth,height=0.098\textwidth]{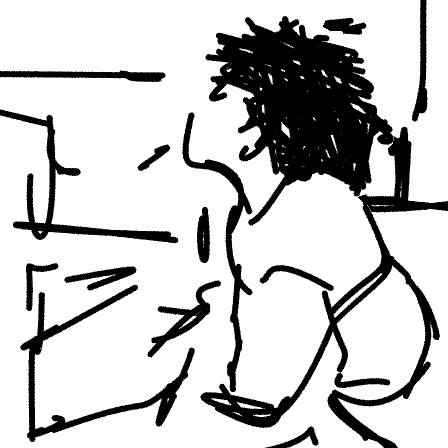}} &
    \frame{\includegraphics[width=0.098\textwidth,height=0.098\textwidth]{clipascene/figs/matrices_black/black_woman_4.png}} &
    \frame{\includegraphics[width=0.098\textwidth,height=0.098\textwidth]{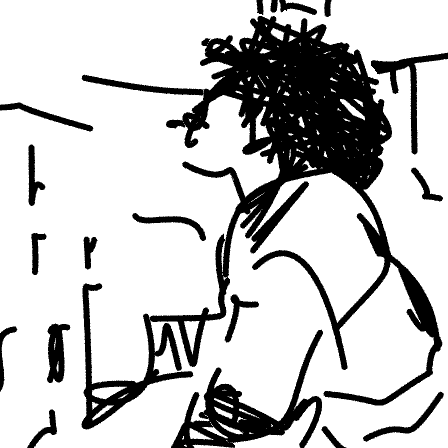}} &
    \frame{\includegraphics[width=0.098\textwidth,height=0.098\textwidth]{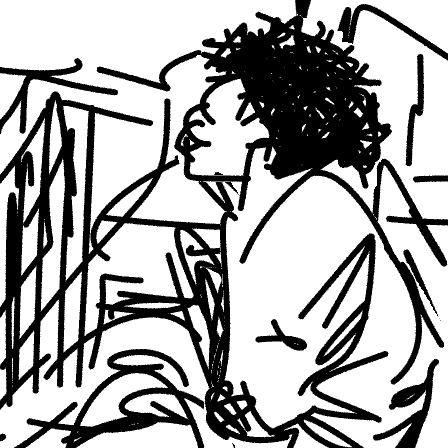}} \\
    
    \frame{\includegraphics[width=0.098\textwidth,height=0.098\textwidth]{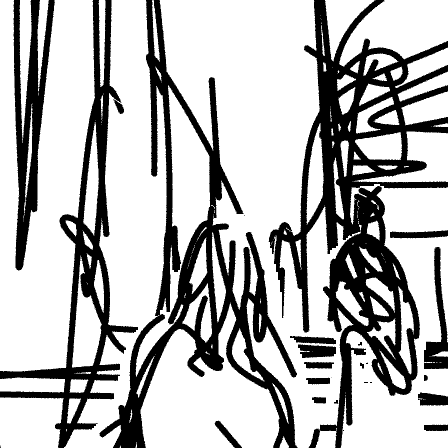}} &
    \frame{\includegraphics[width=0.098\textwidth,height=0.098\textwidth]{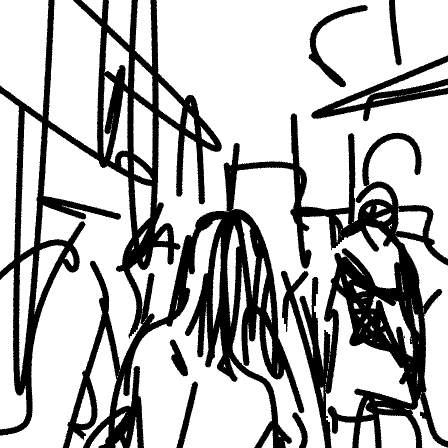}} &
    \frame{\includegraphics[width=0.098\textwidth,height=0.098\textwidth]{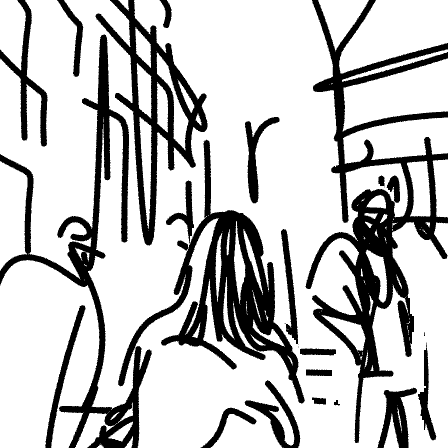}} &
    \frame{\includegraphics[width=0.098\textwidth,height=0.098\textwidth]{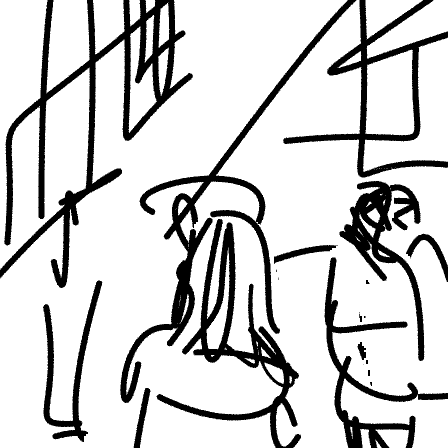}} &
    \hspace{0.5cm}
    \frame{\includegraphics[width=0.098\textwidth,height=0.098\textwidth]{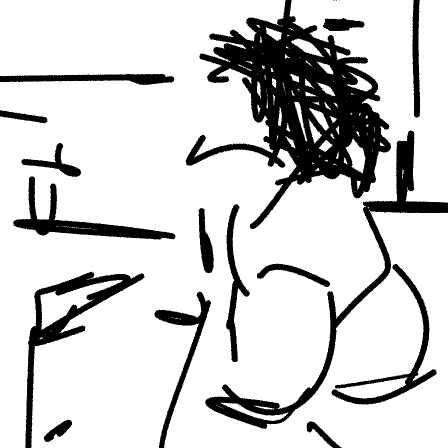}} &
    \frame{\includegraphics[width=0.098\textwidth,height=0.098\textwidth]{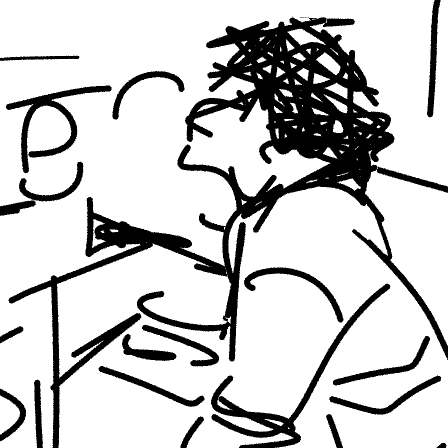}} &
    \frame{\includegraphics[width=0.098\textwidth,height=0.098\textwidth]{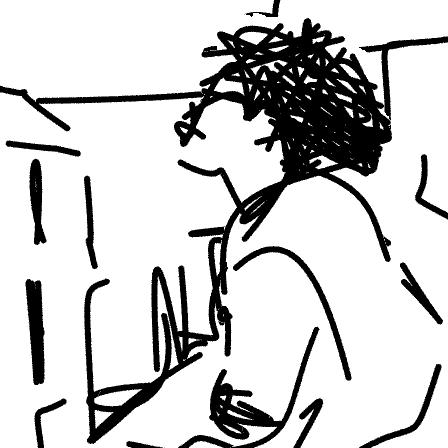}} &
    \frame{\includegraphics[width=0.098\textwidth,height=0.098\textwidth]{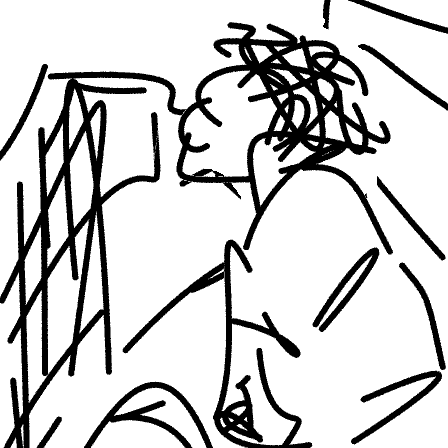}} \\
    
    \frame{\includegraphics[width=0.098\textwidth,height=0.098\textwidth]{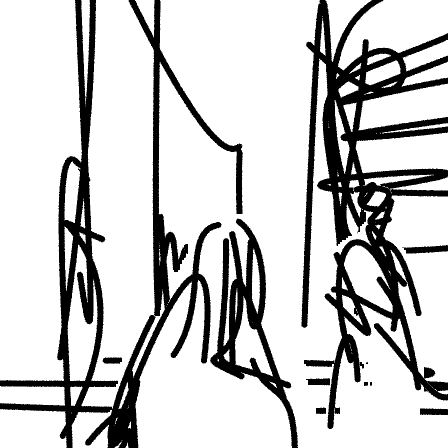}} &
    \frame{\includegraphics[width=0.098\textwidth,height=0.098\textwidth]{clipascene/figs/matrices_black/woman_city_row2col1_black.png}} &
    \frame{\includegraphics[width=0.098\textwidth,height=0.098\textwidth]{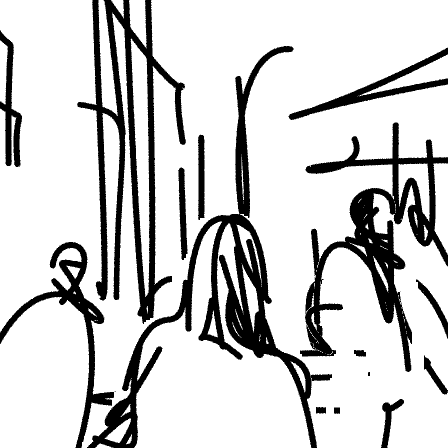}} &
    \frame{\includegraphics[width=0.098\textwidth,height=0.098\textwidth]{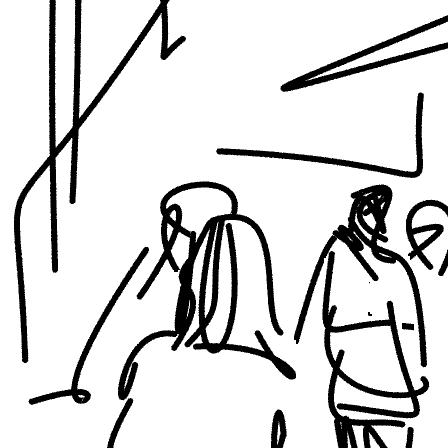}} &
    \hspace{0.5cm}
    \frame{\includegraphics[width=0.098\textwidth,height=0.098\textwidth]{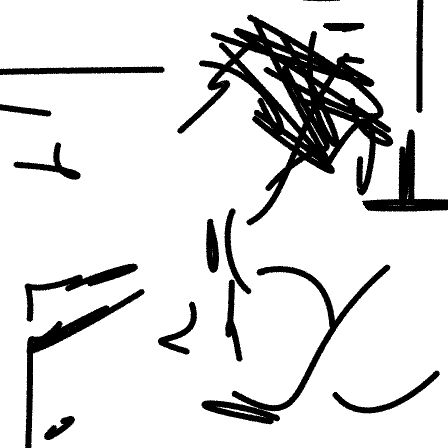}} &
    \frame{\includegraphics[width=0.098\textwidth,height=0.098\textwidth]{clipascene/figs/matrices_black/black_woman_6.png}} &
    \frame{\includegraphics[width=0.098\textwidth,height=0.098\textwidth]{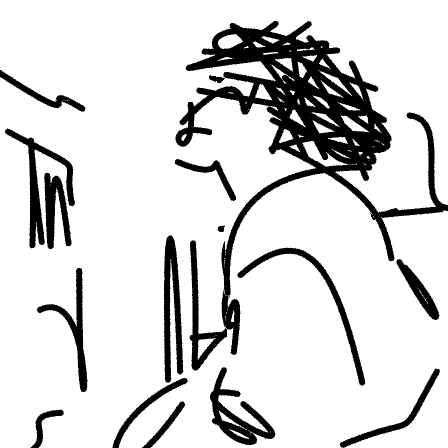}} &
    \frame{\includegraphics[width=0.098\textwidth,height=0.098\textwidth]{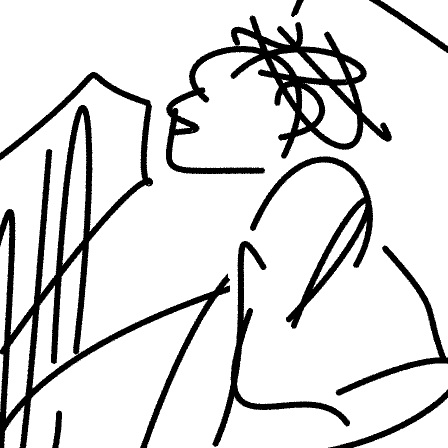}} \\
    
    \frame{\includegraphics[width=0.098\textwidth,height=0.098\textwidth]{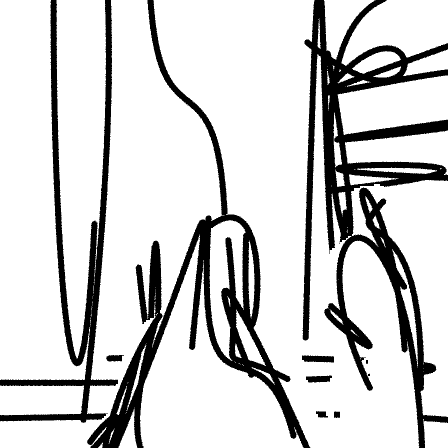}} &
    \frame{\includegraphics[width=0.098\textwidth,height=0.098\textwidth]{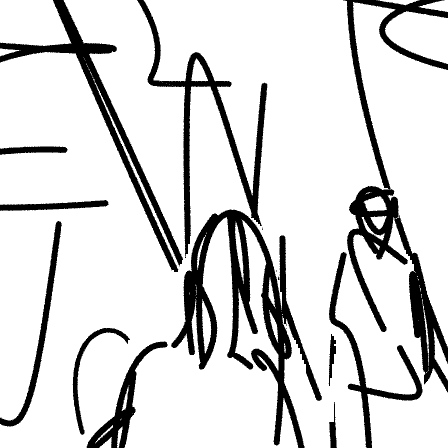}} &
    \frame{\includegraphics[width=0.098\textwidth,height=0.098\textwidth]{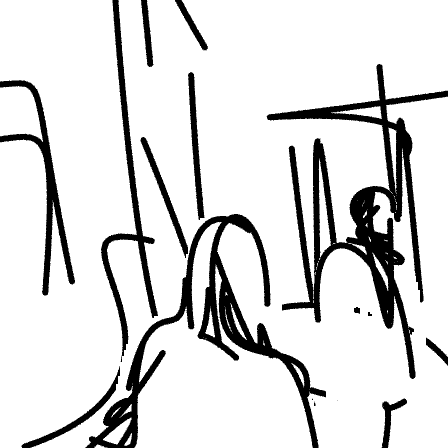}} &
    \frame{\includegraphics[width=0.098\textwidth,height=0.098\textwidth]{clipascene/figs/matrices_black/woman_city_row3col3_black.png}} &
    \hspace{0.5cm}
    \frame{\includegraphics[width=0.098\textwidth,height=0.098\textwidth]{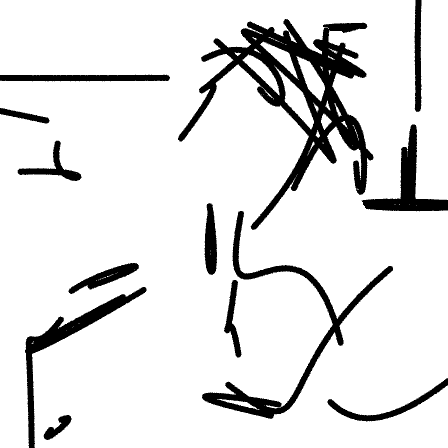}} &
    \frame{\includegraphics[width=0.098\textwidth,height=0.098\textwidth]{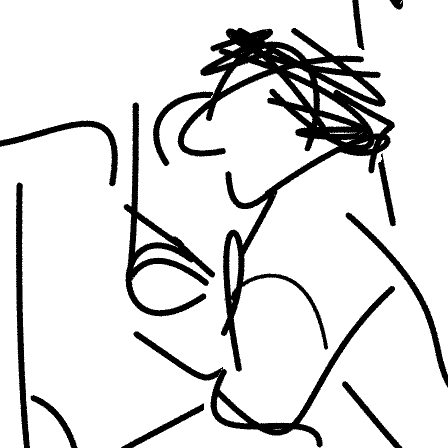}} &
    \frame{\includegraphics[width=0.098\textwidth,height=0.098\textwidth]{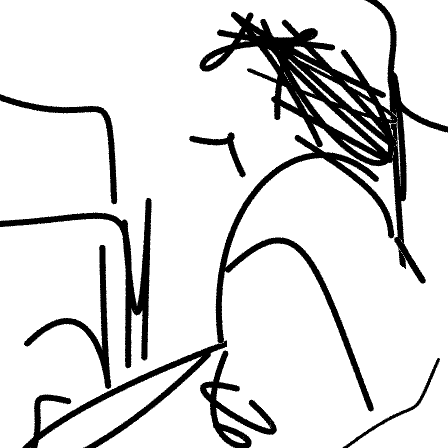}} &
    \frame{\includegraphics[width=0.098\textwidth,height=0.098\textwidth]{clipascene/figs/matrices_black/black_woman_15.png}} \\

    \\
    \\
    
    \includegraphics[width=0.098\textwidth,height=0.098\textwidth]{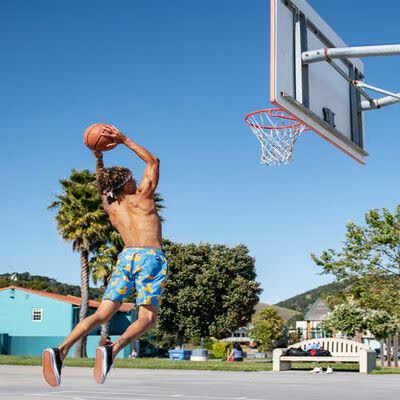} & & & &
    \hspace{0.5cm}
    \includegraphics[width=0.098\textwidth,height=0.098\textwidth]{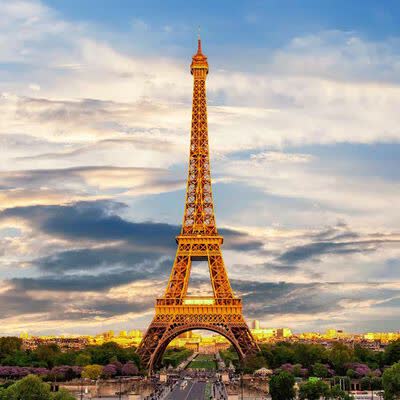} & & & \\

    \frame{\includegraphics[width=0.098\textwidth,height=0.098\textwidth]{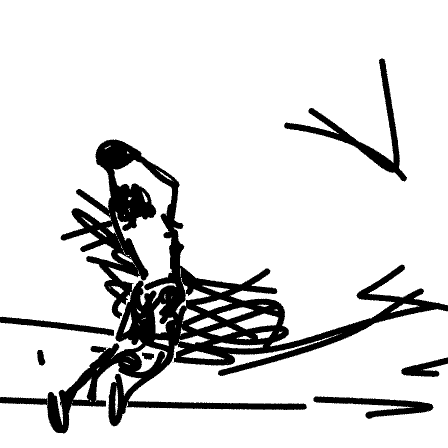}} &
    \frame{\includegraphics[width=0.098\textwidth,height=0.098\textwidth]{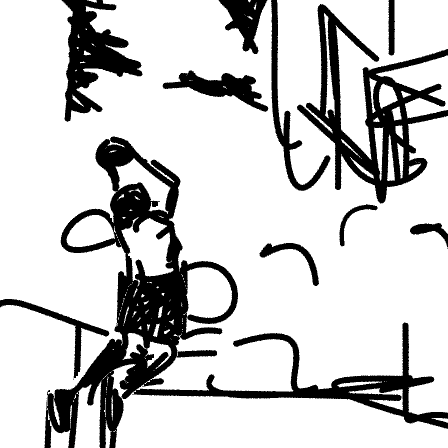}} &
    \frame{\includegraphics[width=0.098\textwidth,height=0.098\textwidth]{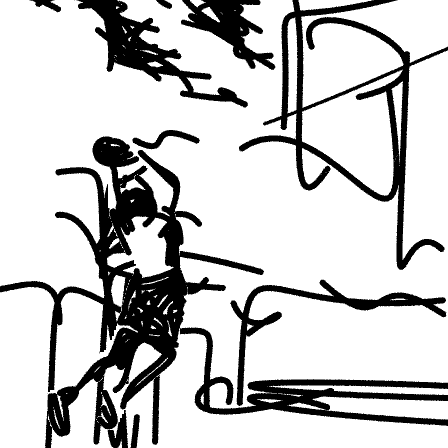}} &
    \frame{\includegraphics[width=0.098\textwidth,height=0.098\textwidth]{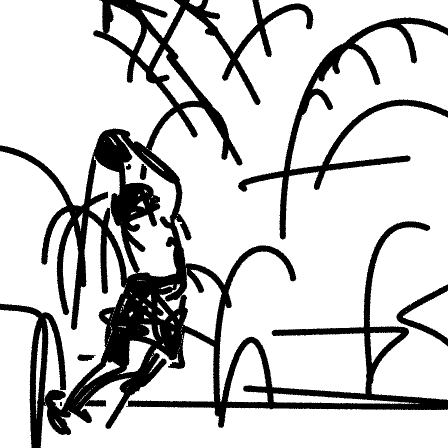}} &
    \hspace{0.5cm}
    \frame{\includegraphics[width=0.098\textwidth,height=0.098\textwidth]{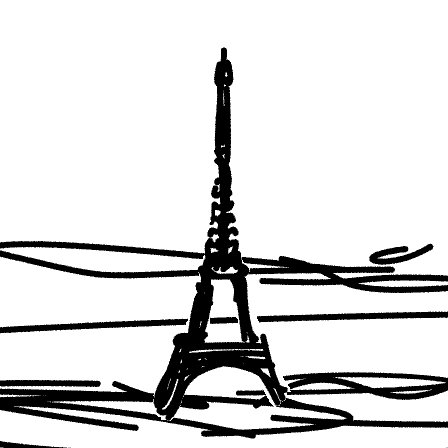}} &
    \frame{\includegraphics[width=0.098\textwidth,height=0.098\textwidth]{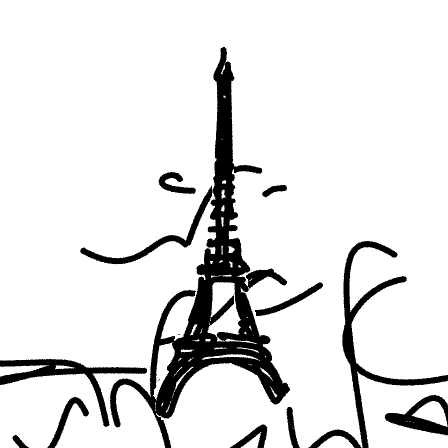}} &
    \frame{\includegraphics[width=0.098\textwidth,height=0.098\textwidth]{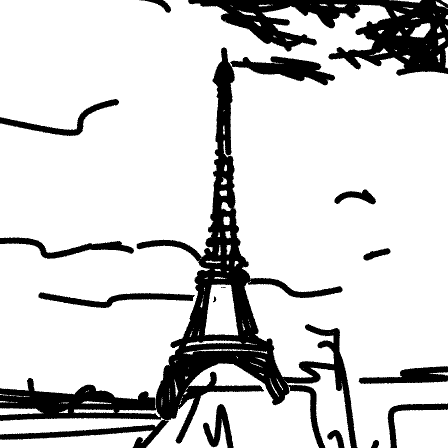}} &
    \frame{\includegraphics[width=0.098\textwidth,height=0.098\textwidth]{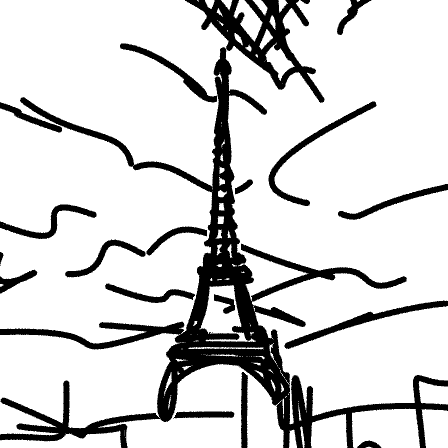}} \\
    
    \frame{\includegraphics[width=0.098\textwidth,height=0.098\textwidth]{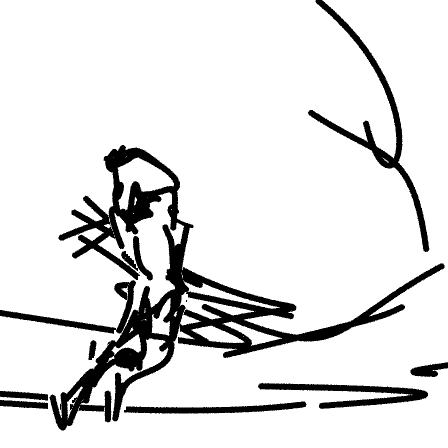}} &
    \frame{\includegraphics[width=0.098\textwidth,height=0.098\textwidth]{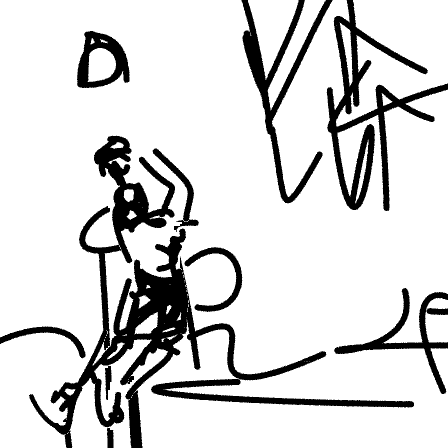}} &
    \frame{\includegraphics[width=0.098\textwidth,height=0.098\textwidth]{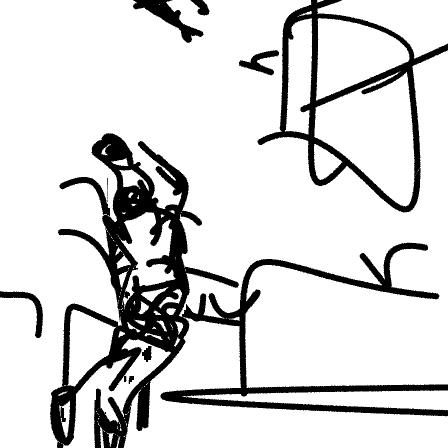}} &
    \frame{\includegraphics[width=0.098\textwidth,height=0.098\textwidth]{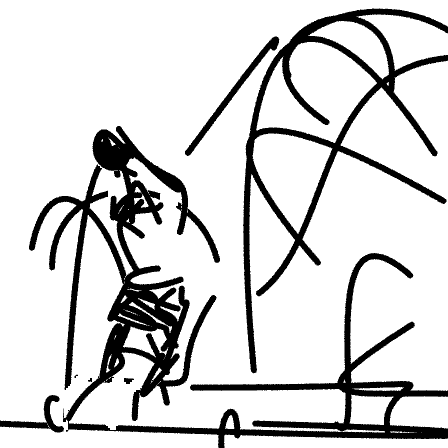}} &
    \hspace{0.5cm}
    \frame{\includegraphics[width=0.098\textwidth,height=0.098\textwidth]{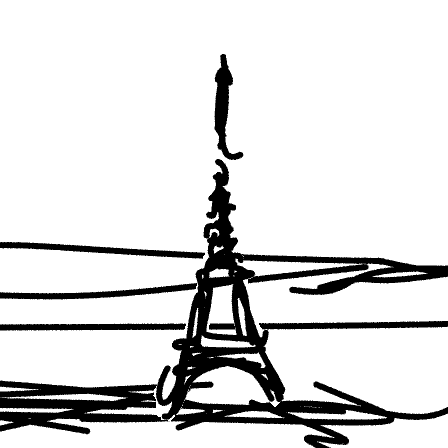}} &
    \frame{\includegraphics[width=0.098\textwidth,height=0.098\textwidth]{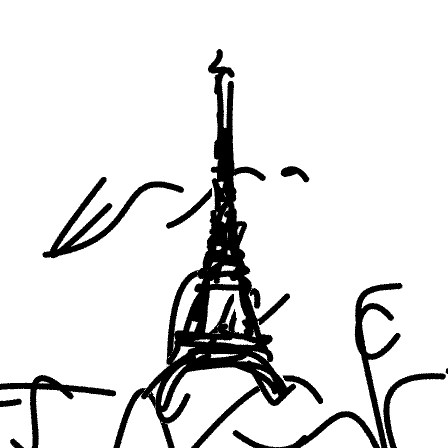}} &
    \frame{\includegraphics[width=0.098\textwidth,height=0.098\textwidth]{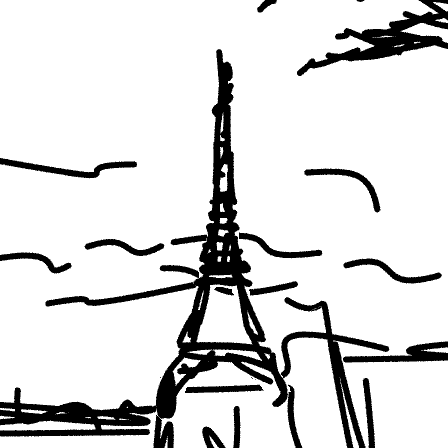}} &
    \frame{\includegraphics[width=0.098\textwidth,height=0.098\textwidth]{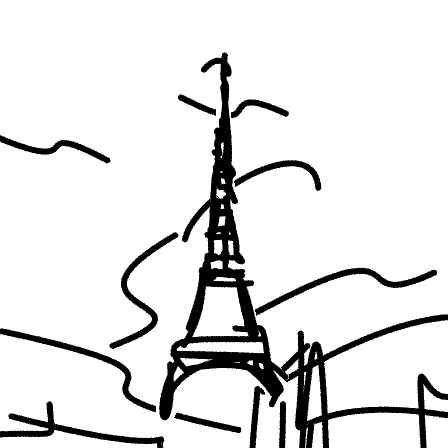}} \\
    
    \frame{\includegraphics[width=0.098\textwidth,height=0.098\textwidth]{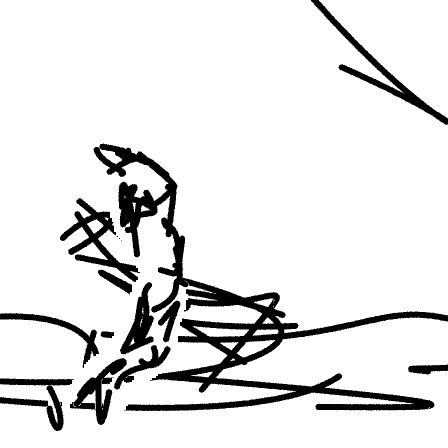}} &
    \frame{\includegraphics[width=0.098\textwidth,height=0.098\textwidth]{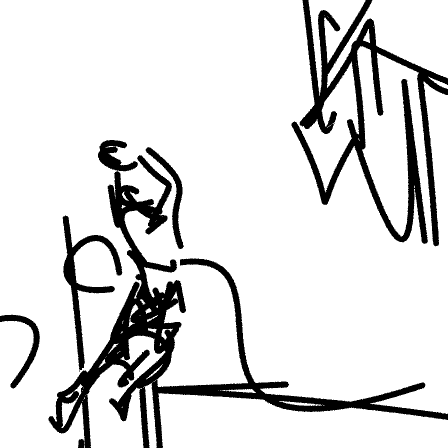}} &
    \frame{\includegraphics[width=0.098\textwidth,height=0.098\textwidth]{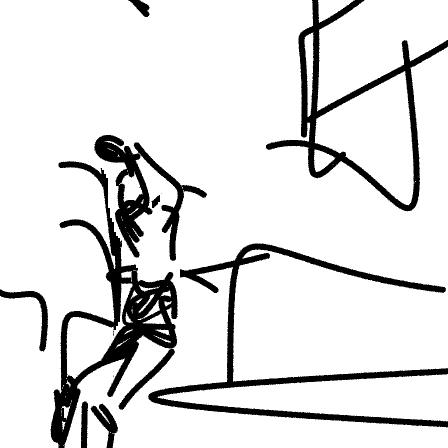}} &
    \frame{\includegraphics[width=0.098\textwidth,height=0.098\textwidth]{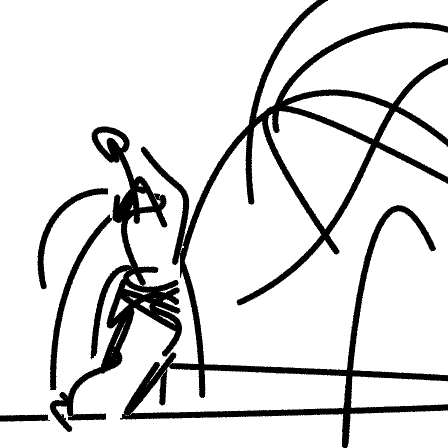}} &
    \hspace{0.5cm}
    \frame{\includegraphics[width=0.098\textwidth,height=0.098\textwidth]{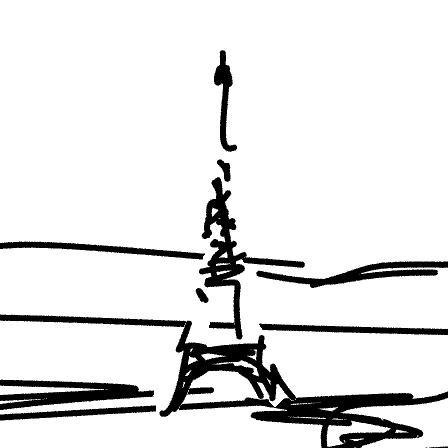}} &
    \frame{\includegraphics[width=0.098\textwidth,height=0.098\textwidth]{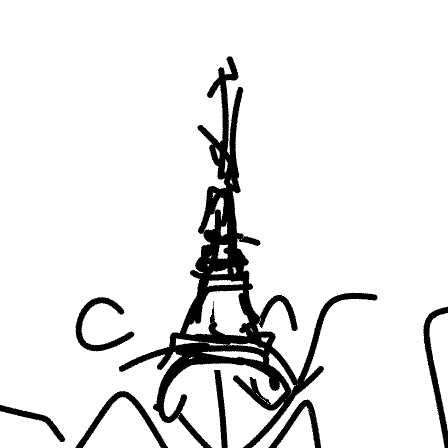}} &
    \frame{\includegraphics[width=0.098\textwidth,height=0.098\textwidth]{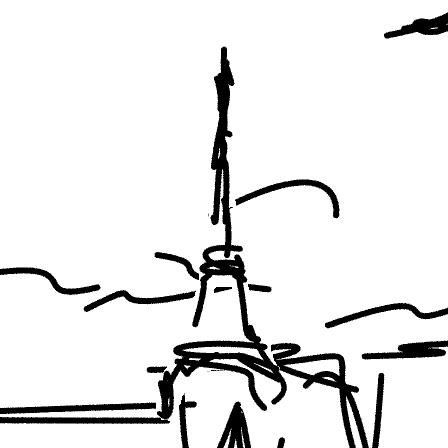}} &
    \frame{\includegraphics[width=0.098\textwidth,height=0.098\textwidth]{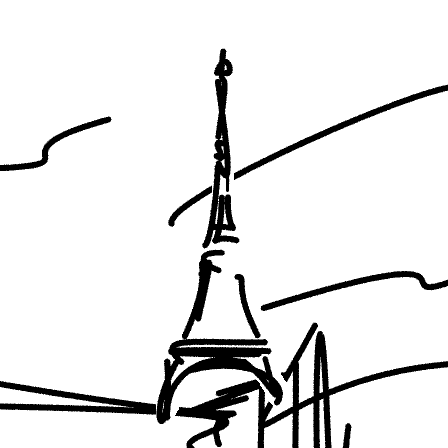}} \\
    
    \frame{\includegraphics[width=0.098\textwidth,height=0.098\textwidth]{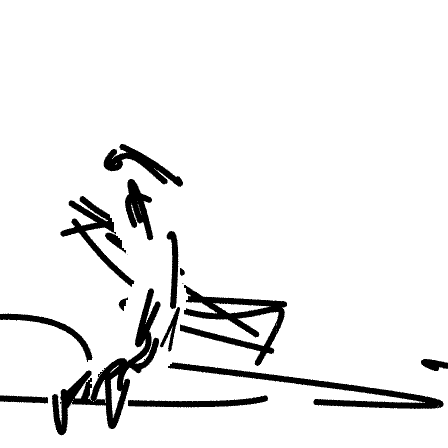}} &
    \frame{\includegraphics[width=0.098\textwidth,height=0.098\textwidth]{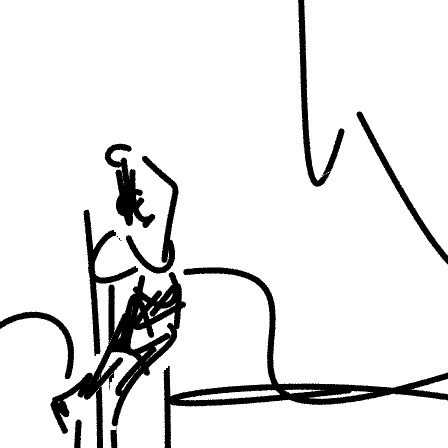}} &
    \frame{\includegraphics[width=0.098\textwidth,height=0.098\textwidth]{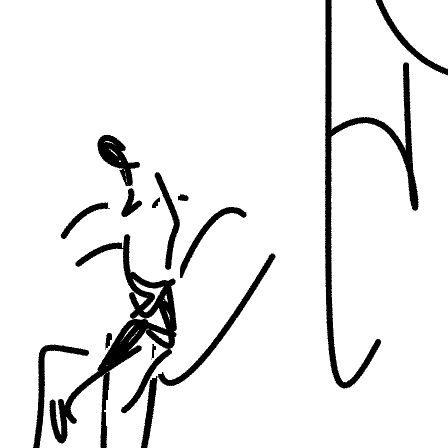}} &
    \frame{\includegraphics[width=0.098\textwidth,height=0.098\textwidth]{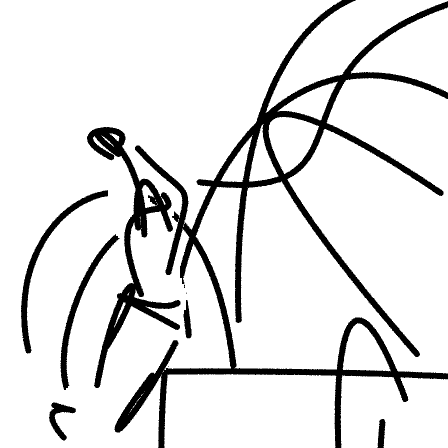}} &
    \hspace{0.5cm}
    \frame{\includegraphics[width=0.098\textwidth,height=0.098\textwidth]{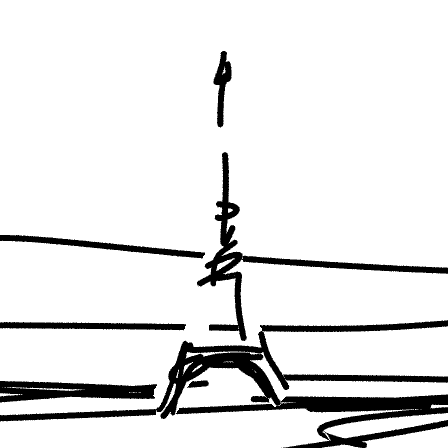}} &
    \frame{\includegraphics[width=0.098\textwidth,height=0.098\textwidth]{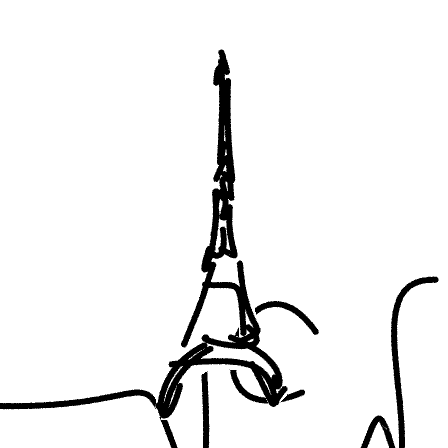}} &
    \frame{\includegraphics[width=0.098\textwidth,height=0.098\textwidth]{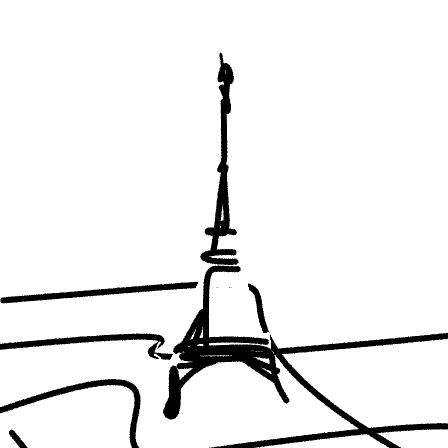}} &
    \frame{\includegraphics[width=0.098\textwidth,height=0.098\textwidth]{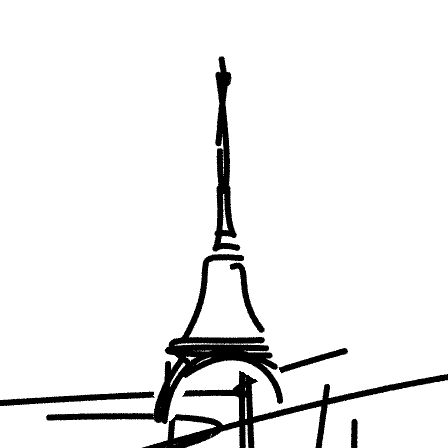}} \\

    \end{tabular}
    \caption{The $4\times4$ matrix of sketches produced by our method. Columns from left to right illustrate the change in fidelity, from precise to loose, and rows from top to bottom illustrate the visual simplification.}
    
    \label{fig:matrix3}
\end{figure*}

\begin{figure*}
    \centering
    
    \begin{tabular}{c c c c c c c c}

    \includegraphics[width=0.098\textwidth,height=0.098\textwidth]{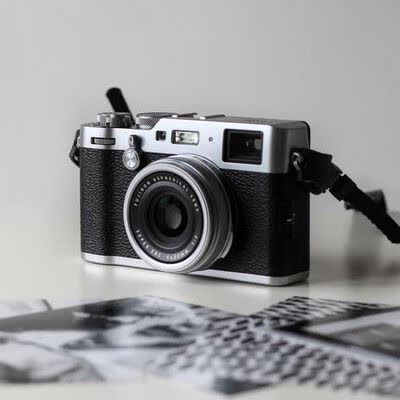} & & & &
    \hspace{0.5cm}
    \includegraphics[width=0.098\textwidth,height=0.098\textwidth]{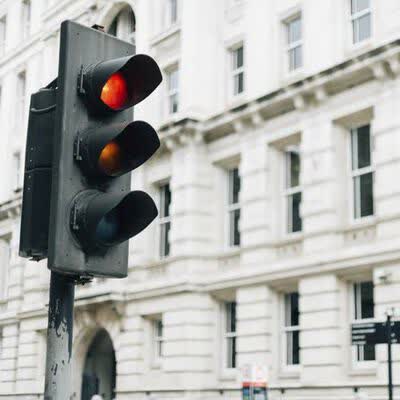} & & & \\
    
    \frame{\includegraphics[width=0.098\textwidth,height=0.098\textwidth]{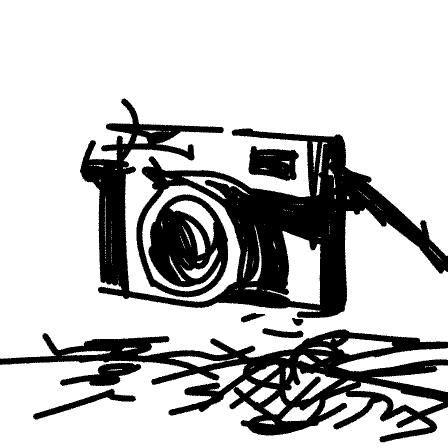}} &
    \frame{\includegraphics[width=0.098\textwidth,height=0.098\textwidth]{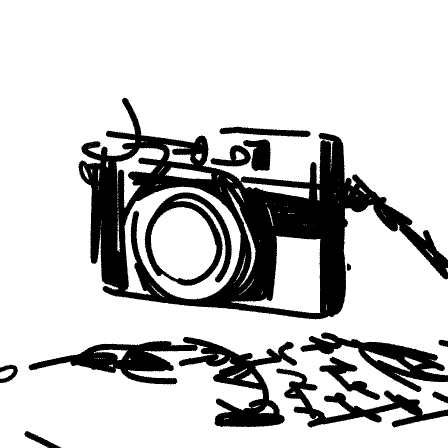}} &
    \frame{\includegraphics[width=0.098\textwidth,height=0.098\textwidth]{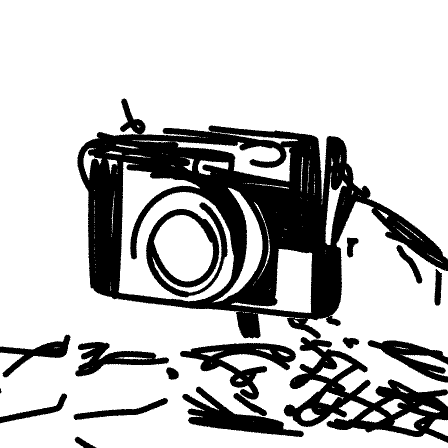}} &
    \frame{\includegraphics[width=0.098\textwidth,height=0.098\textwidth]{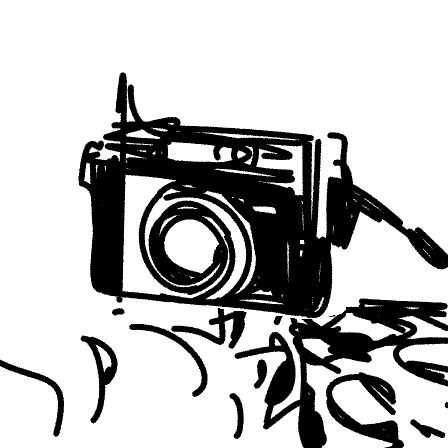}} &
    \hspace{0.5cm}
    \frame{\includegraphics[width=0.098\textwidth,height=0.098\textwidth]{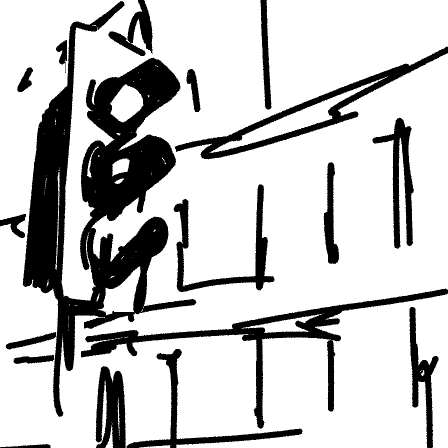}} &
    \frame{\includegraphics[width=0.098\textwidth,height=0.098\textwidth]{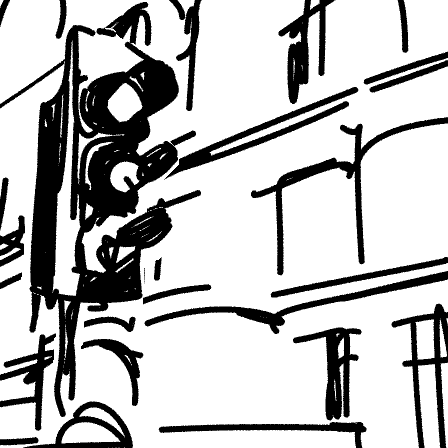}} &
    \frame{\includegraphics[width=0.098\textwidth,height=0.098\textwidth]{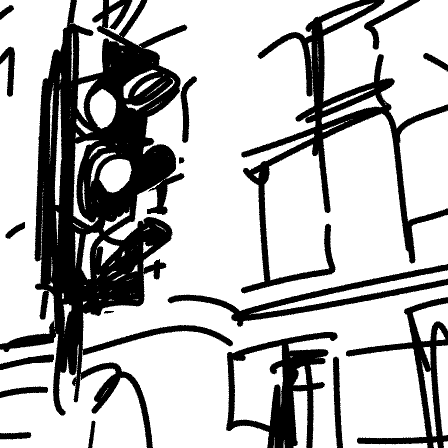}} &
    \frame{\includegraphics[width=0.098\textwidth,height=0.098\textwidth]{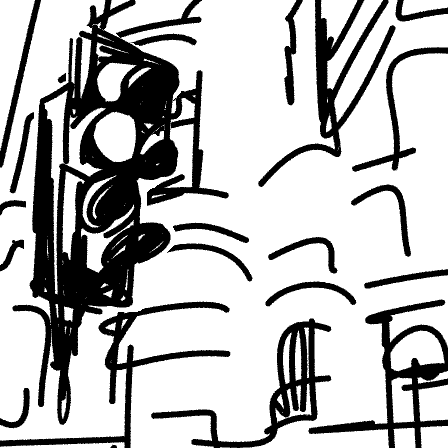}} \\
    
    \frame{\includegraphics[width=0.098\textwidth,height=0.098\textwidth]{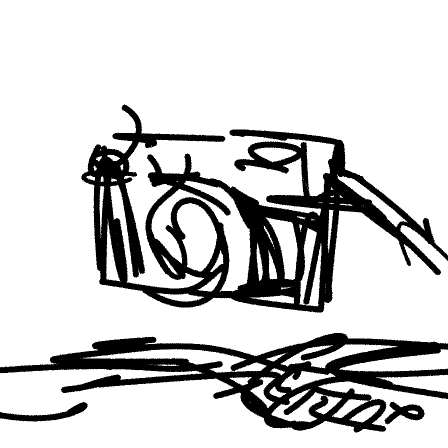}} &
    \frame{\includegraphics[width=0.098\textwidth,height=0.098\textwidth]{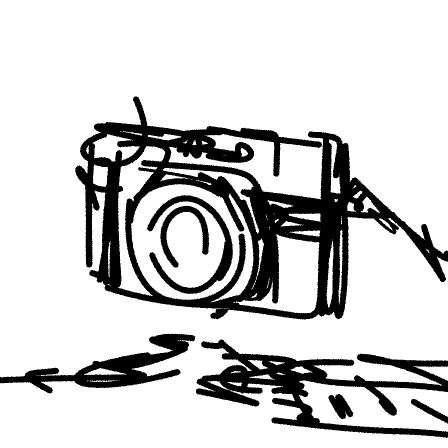}} &
    \frame{\includegraphics[width=0.098\textwidth,height=0.098\textwidth]{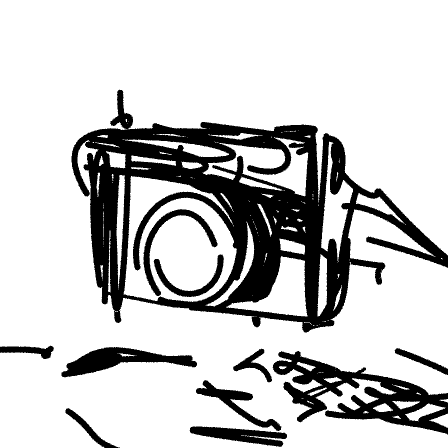}} &
    \frame{\includegraphics[width=0.098\textwidth,height=0.098\textwidth]{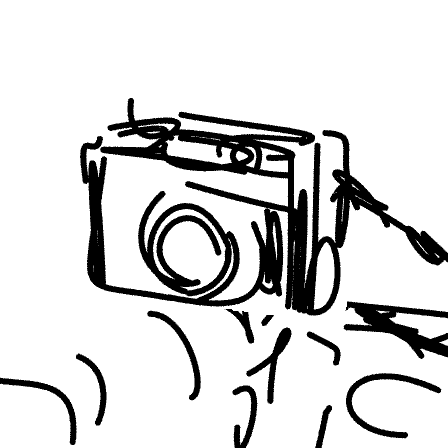}} &
    \hspace{0.5cm}
    \frame{\includegraphics[width=0.098\textwidth,height=0.098\textwidth]{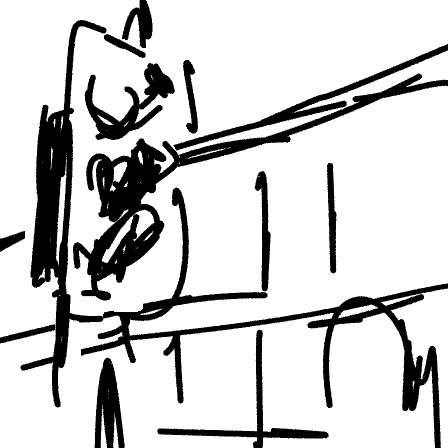}} &
    \frame{\includegraphics[width=0.098\textwidth,height=0.098\textwidth]{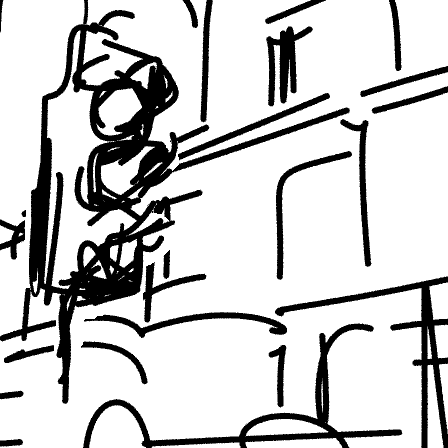}} &
    \frame{\includegraphics[width=0.098\textwidth,height=0.098\textwidth]{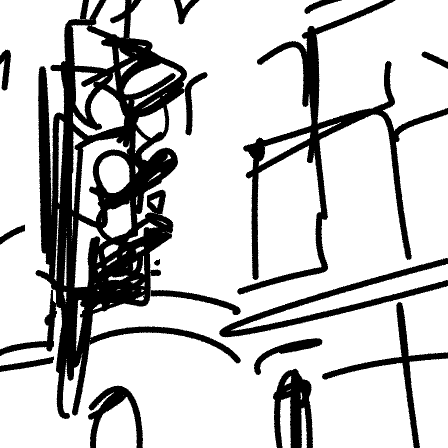}} &
    \frame{\includegraphics[width=0.098\textwidth,height=0.098\textwidth]{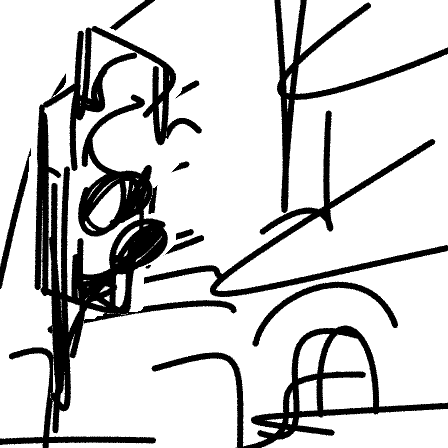}} \\
    
    \frame{\includegraphics[width=0.098\textwidth,height=0.098\textwidth]{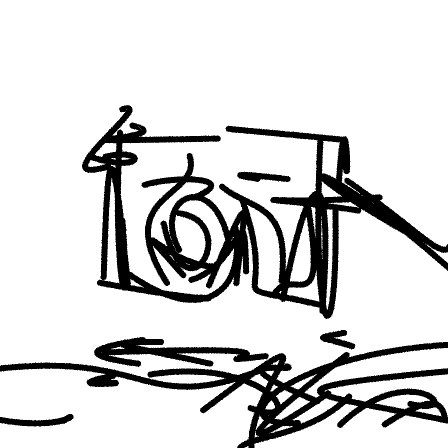}} &
    \frame{\includegraphics[width=0.098\textwidth,height=0.098\textwidth]{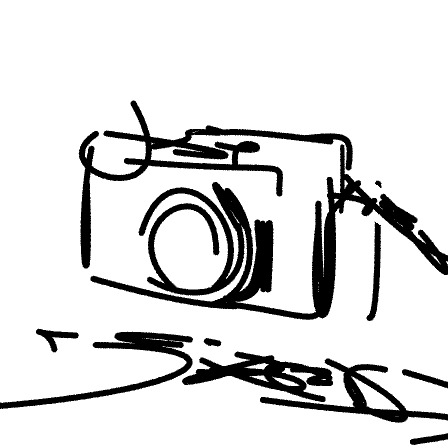}} &
    \frame{\includegraphics[width=0.098\textwidth,height=0.098\textwidth]{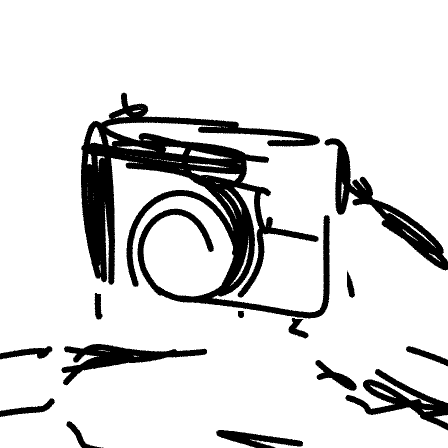}} &
    \frame{\includegraphics[width=0.098\textwidth,height=0.098\textwidth]{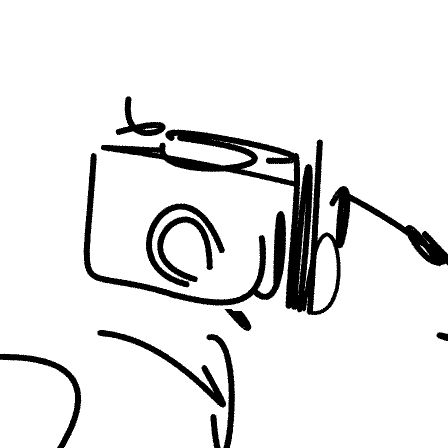}} &
    \hspace{0.5cm}
    \frame{\includegraphics[width=0.098\textwidth,height=0.098\textwidth]{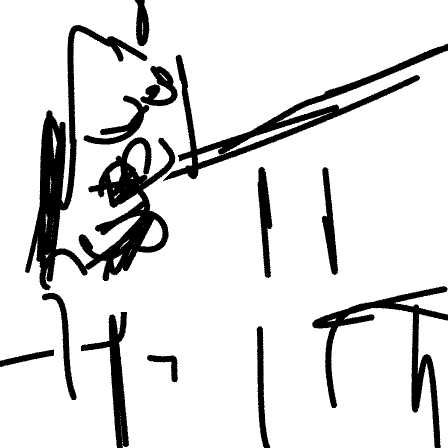}} &
    \frame{\includegraphics[width=0.098\textwidth,height=0.098\textwidth]{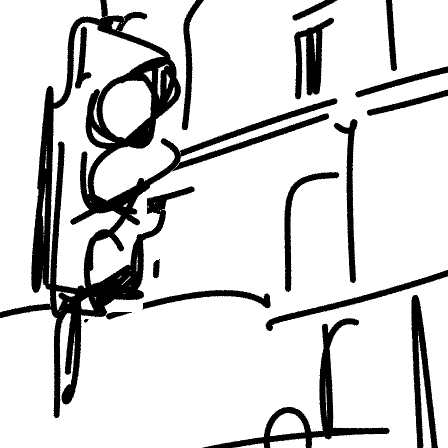}} &
    \frame{\includegraphics[width=0.098\textwidth,height=0.098\textwidth]{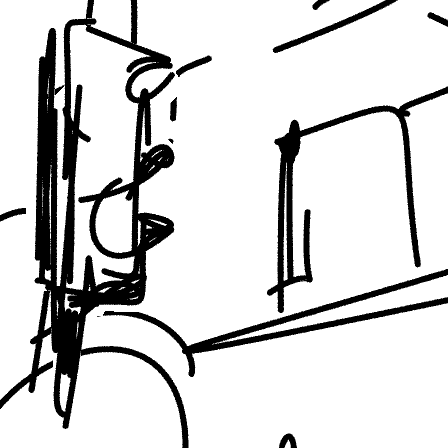}} &
    \frame{\includegraphics[width=0.098\textwidth,height=0.098\textwidth]{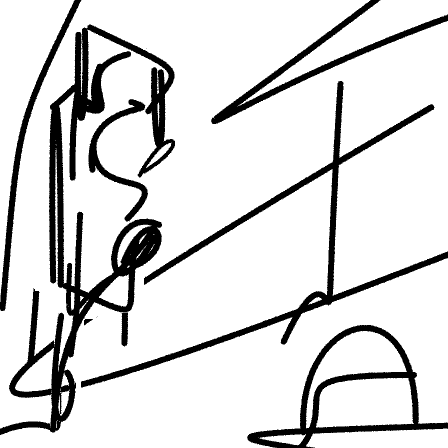}} \\
    
    \frame{\includegraphics[width=0.098\textwidth,height=0.098\textwidth]{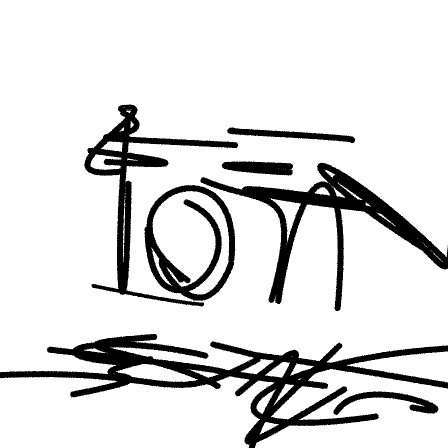}} &
    \frame{\includegraphics[width=0.098\textwidth,height=0.098\textwidth]{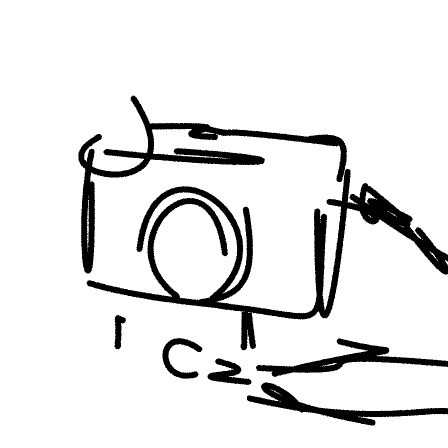}} &
    \frame{\includegraphics[width=0.098\textwidth,height=0.098\textwidth]{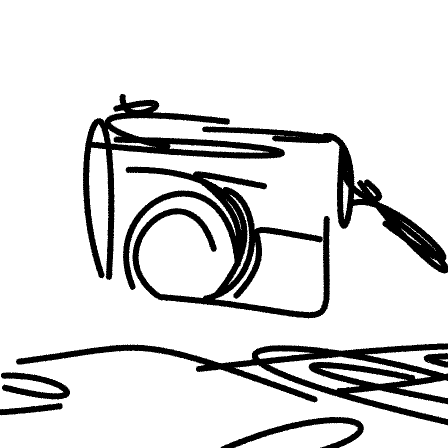}} &
    \frame{\includegraphics[width=0.098\textwidth,height=0.098\textwidth]{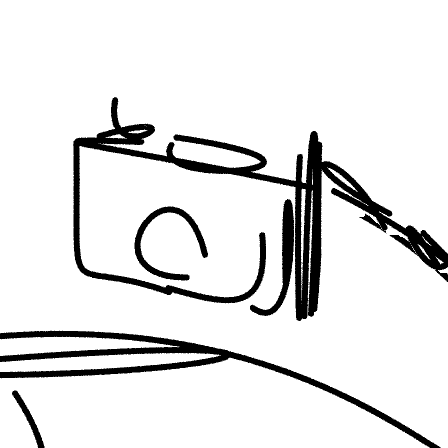}} &
    \hspace{0.5cm}
    \frame{\includegraphics[width=0.098\textwidth,height=0.098\textwidth]{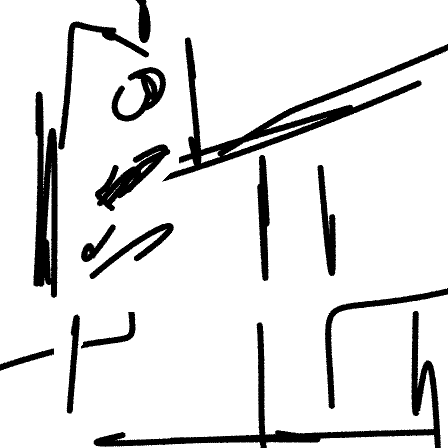}} &
    \frame{\includegraphics[width=0.098\textwidth,height=0.098\textwidth]{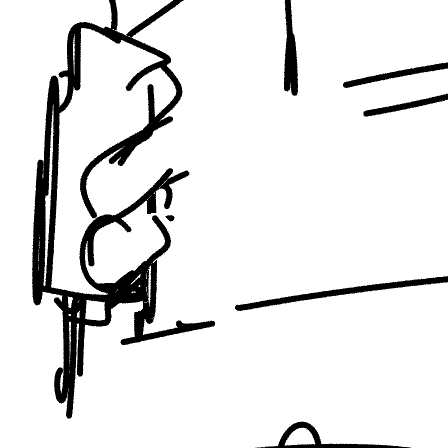}} &
    \frame{\includegraphics[width=0.098\textwidth,height=0.098\textwidth]{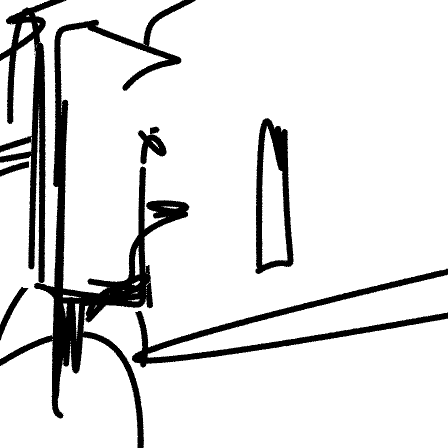}} &
    \frame{\includegraphics[width=0.098\textwidth,height=0.098\textwidth]{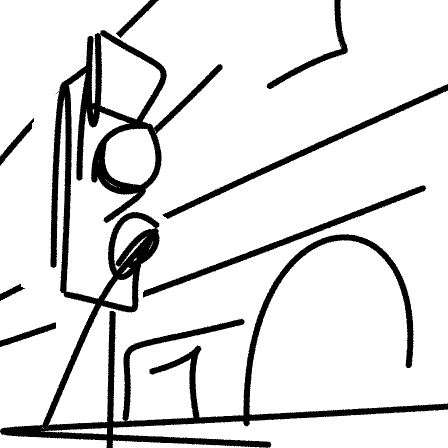}} \\

    \\
    \\
    
    \includegraphics[width=0.098\textwidth,height=0.098\textwidth]{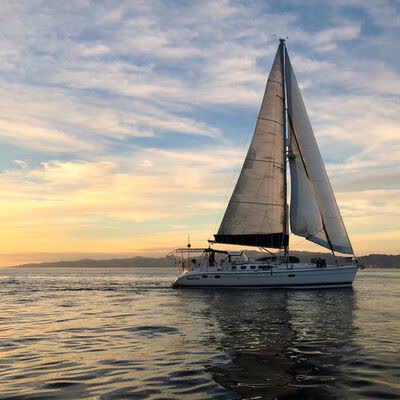} & & & &
    \hspace{0.5cm}
    \includegraphics[width=0.098\textwidth,height=0.098\textwidth]{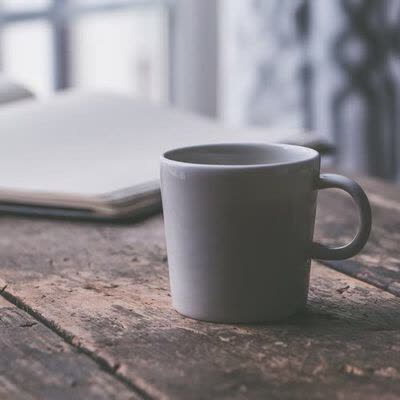} & & & \\

    \frame{\includegraphics[width=0.098\textwidth,height=0.098\textwidth]{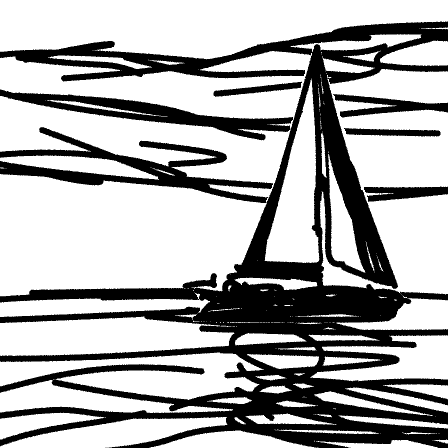}} &
    \frame{\includegraphics[width=0.098\textwidth,height=0.098\textwidth]{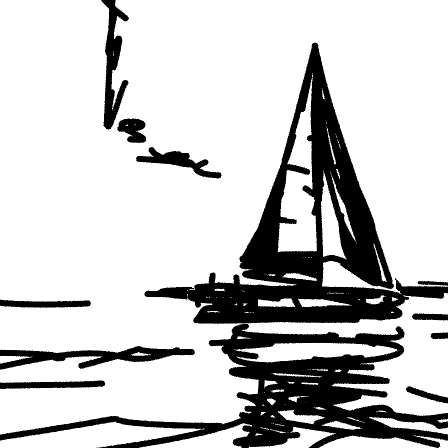}} &
    \frame{\includegraphics[width=0.098\textwidth,height=0.098\textwidth]{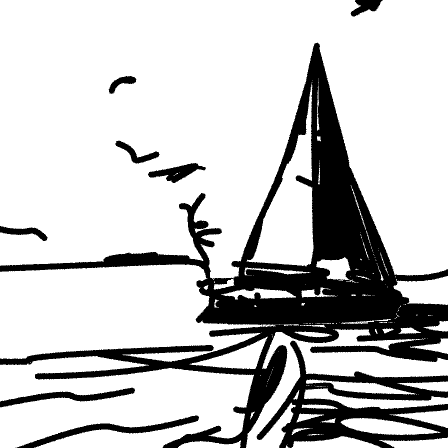}} &
    \frame{\includegraphics[width=0.098\textwidth,height=0.098\textwidth]{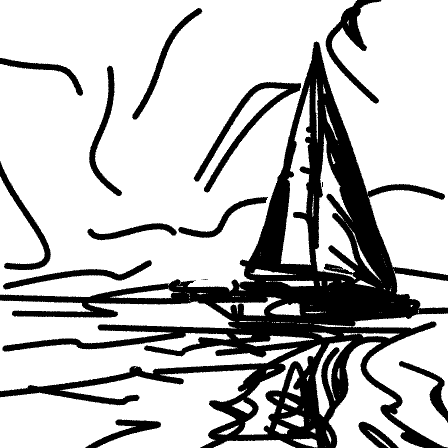}} &
    \hspace{0.5cm}
    \frame{\includegraphics[width=0.098\textwidth,height=0.098\textwidth]{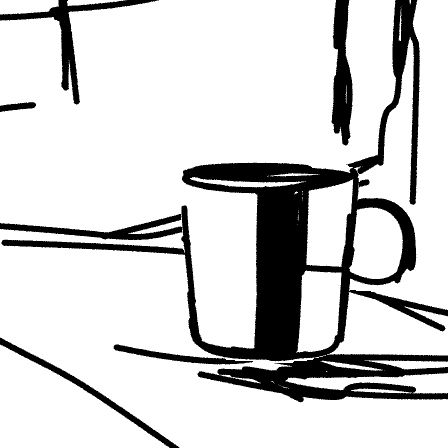}} &
    \frame{\includegraphics[width=0.098\textwidth,height=0.098\textwidth]{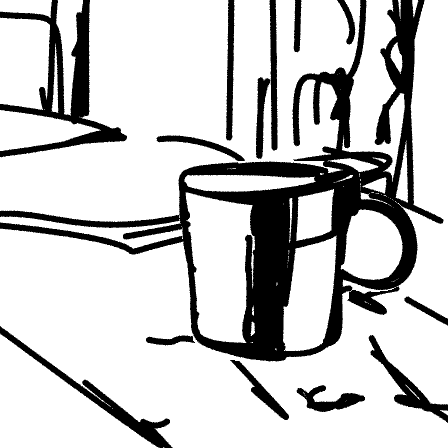}} &
    \frame{\includegraphics[width=0.098\textwidth,height=0.098\textwidth]{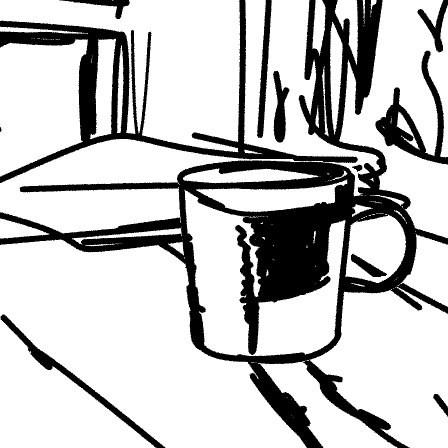}} &
    \frame{\includegraphics[width=0.098\textwidth,height=0.098\textwidth]{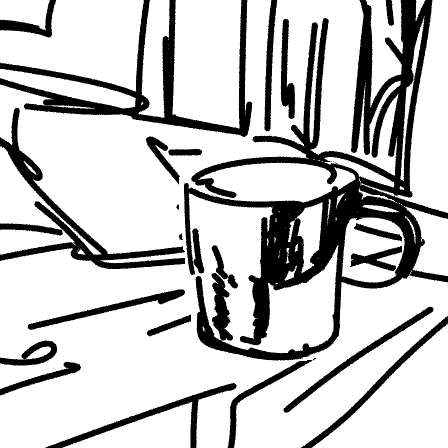}} \\
    
    \frame{\includegraphics[width=0.098\textwidth,height=0.098\textwidth]{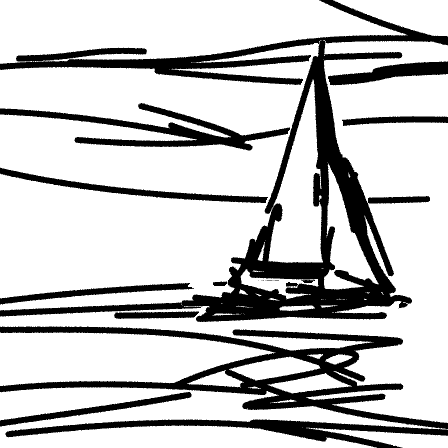}} &
    \frame{\includegraphics[width=0.098\textwidth,height=0.098\textwidth]{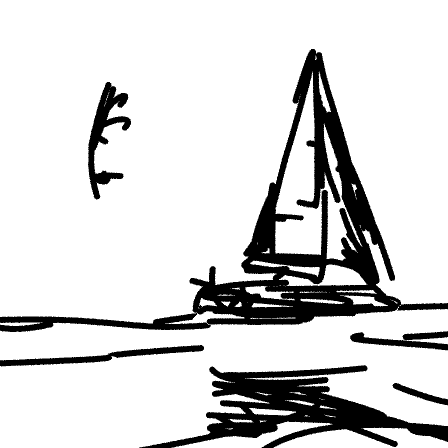}} &
    \frame{\includegraphics[width=0.098\textwidth,height=0.098\textwidth]{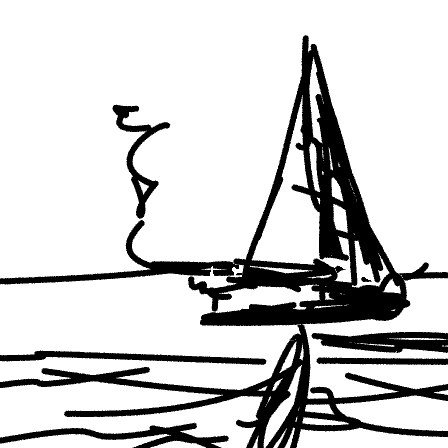}} &
    \frame{\includegraphics[width=0.098\textwidth,height=0.098\textwidth]{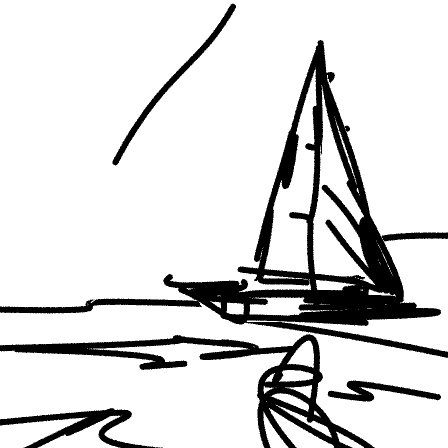}} &
    \hspace{0.5cm}
    \frame{\includegraphics[width=0.098\textwidth,height=0.098\textwidth]{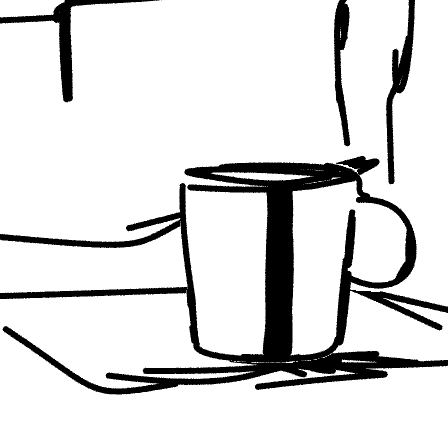}} &
    \frame{\includegraphics[width=0.098\textwidth,height=0.098\textwidth]{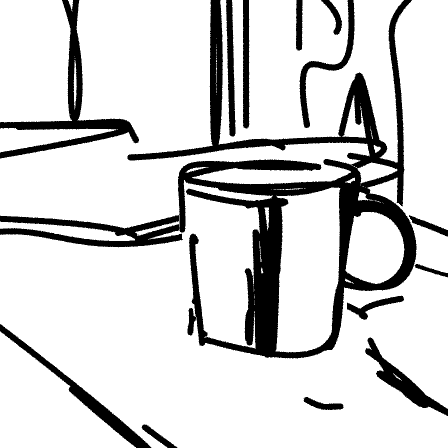}} &
    \frame{\includegraphics[width=0.098\textwidth,height=0.098\textwidth]{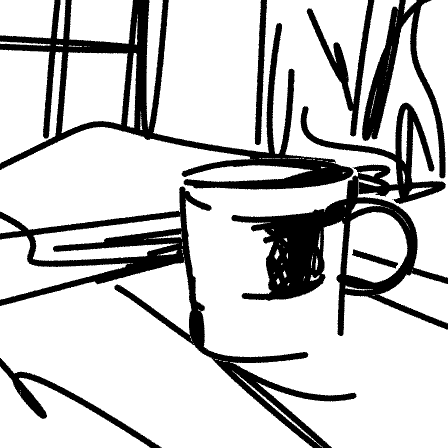}} &
    \frame{\includegraphics[width=0.098\textwidth,height=0.098\textwidth]{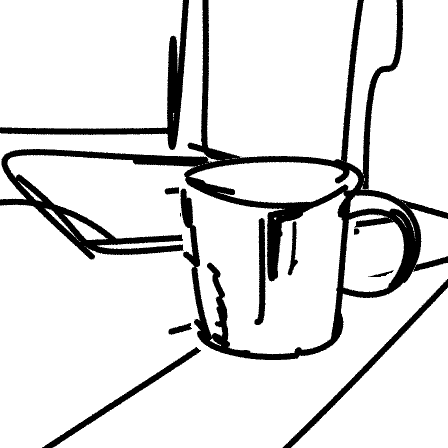}} \\
    
    \frame{\includegraphics[width=0.098\textwidth,height=0.098\textwidth]{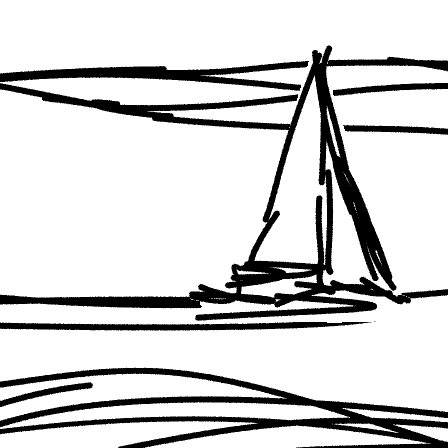}} &
    \frame{\includegraphics[width=0.098\textwidth,height=0.098\textwidth]{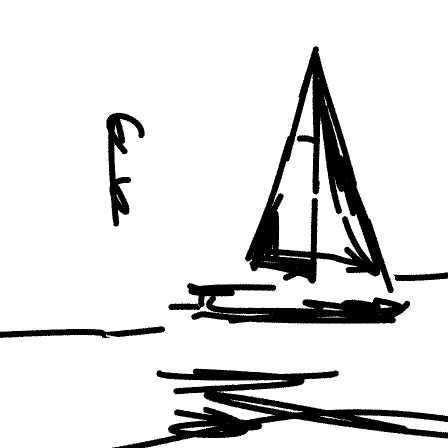}} &
    \frame{\includegraphics[width=0.098\textwidth,height=0.098\textwidth]{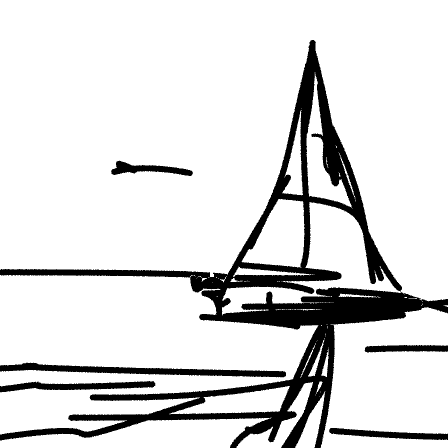}} &
    \frame{\includegraphics[width=0.098\textwidth,height=0.098\textwidth]{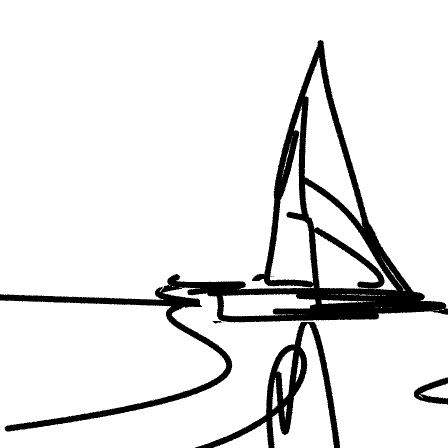}} &
    \hspace{0.5cm}
    \frame{\includegraphics[width=0.098\textwidth,height=0.098\textwidth]{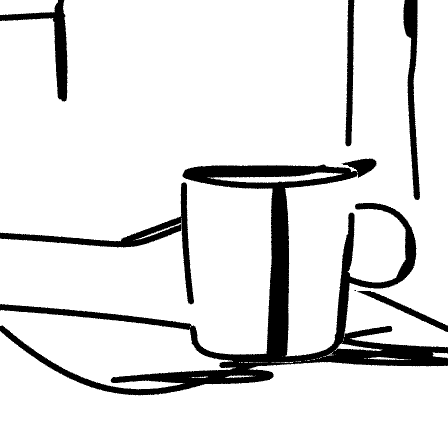}} &
    \frame{\includegraphics[width=0.098\textwidth,height=0.098\textwidth]{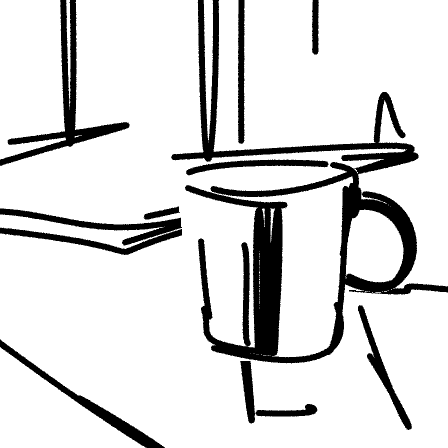}} &
    \frame{\includegraphics[width=0.098\textwidth,height=0.098\textwidth]{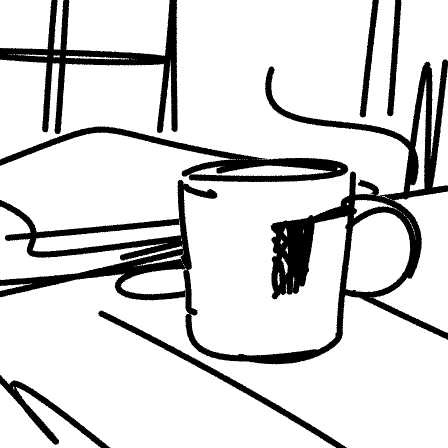}} &
    \frame{\includegraphics[width=0.098\textwidth,height=0.098\textwidth]{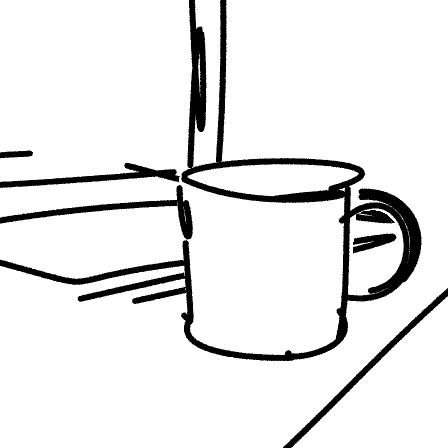}} \\
    
    \frame{\includegraphics[width=0.098\textwidth,height=0.098\textwidth]{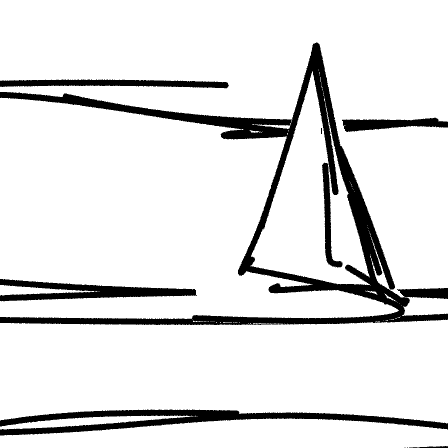}} &
    \frame{\includegraphics[width=0.098\textwidth,height=0.098\textwidth]{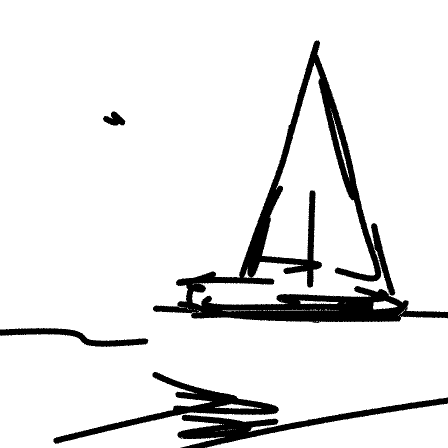}} &
    \frame{\includegraphics[width=0.098\textwidth,height=0.098\textwidth]{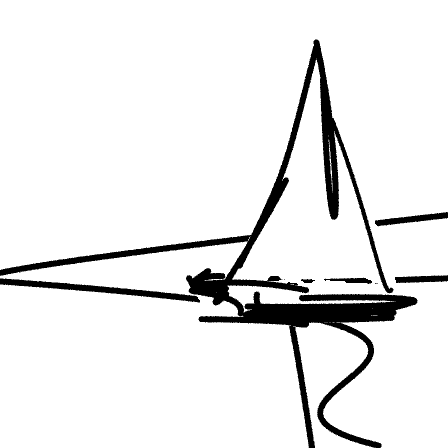}} &
    \frame{\includegraphics[width=0.098\textwidth,height=0.098\textwidth]{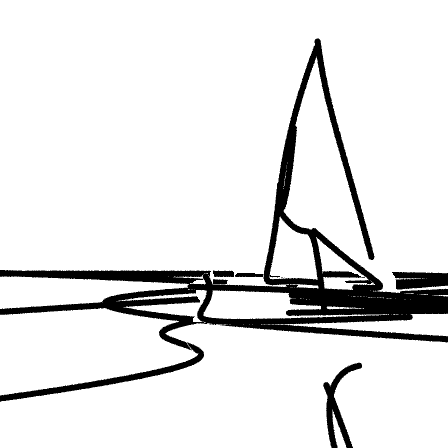}} &
    \hspace{0.5cm}
    \frame{\includegraphics[width=0.098\textwidth,height=0.098\textwidth]{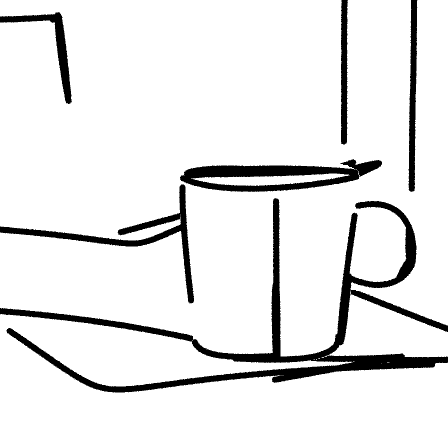}} &
    \frame{\includegraphics[width=0.098\textwidth,height=0.098\textwidth]{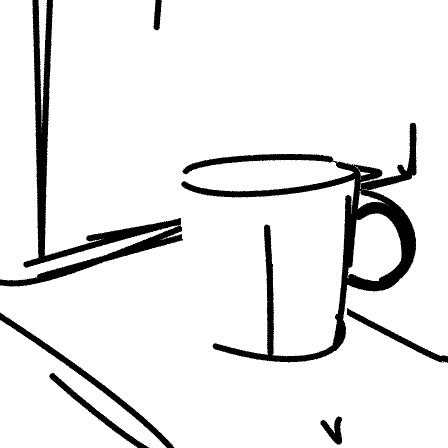}} &
    \frame{\includegraphics[width=0.098\textwidth,height=0.098\textwidth]{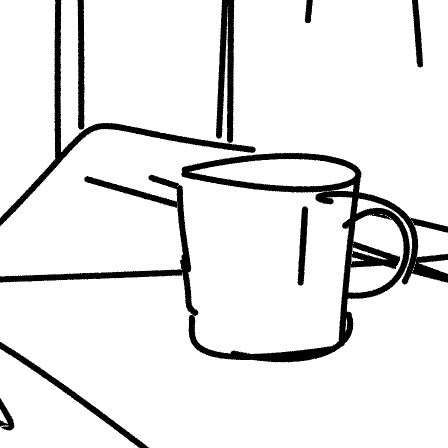}} &
    \frame{\includegraphics[width=0.098\textwidth,height=0.098\textwidth]{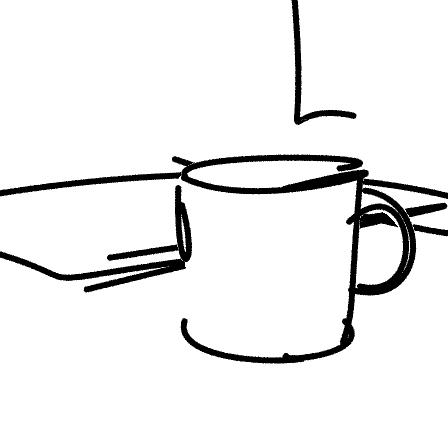}} \\

    \end{tabular}
    \caption{The $4\times4$ matrix of sketches produced by our method. Columns from left to right illustrate the change in fidelity, from precise to loose, and rows from top to bottom illustrate the visual simplification.}
    
    \label{fig:matrix4}
\end{figure*}

\begin{figure*}
    \centering
    
    \begin{tabular}{c c c c c c c c}

    \includegraphics[width=0.098\textwidth,height=0.098\textwidth]{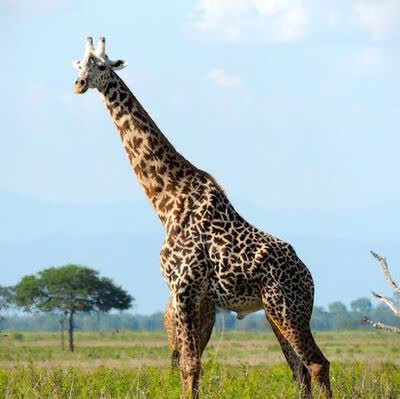} & & & &
    \hspace{0.5cm}
    \includegraphics[width=0.098\textwidth,height=0.098\textwidth]{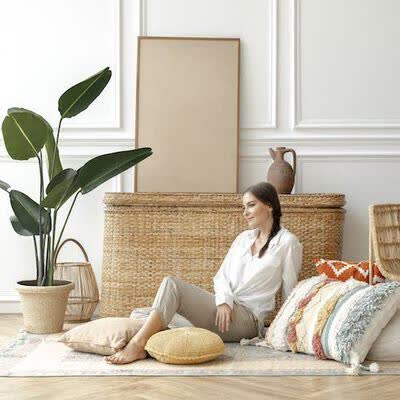} & & & \\
    
    \frame{\includegraphics[width=0.098\textwidth,height=0.098\textwidth]{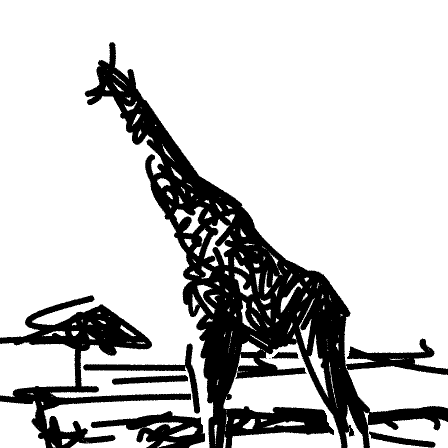}} &
    \frame{\includegraphics[width=0.098\textwidth,height=0.098\textwidth]{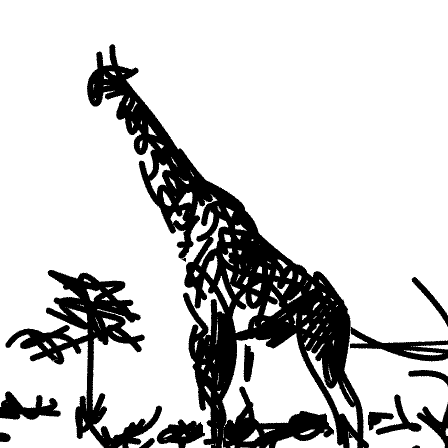}} &
    \frame{\includegraphics[width=0.098\textwidth,height=0.098\textwidth]{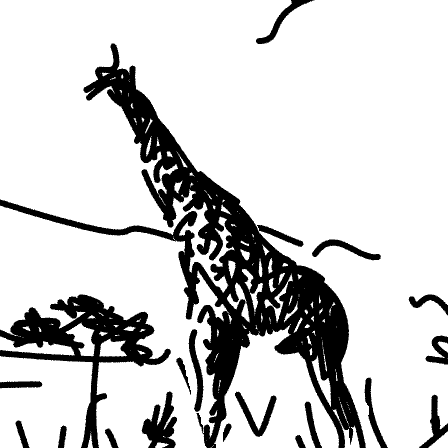}} &
    \frame{\includegraphics[width=0.098\textwidth,height=0.098\textwidth]{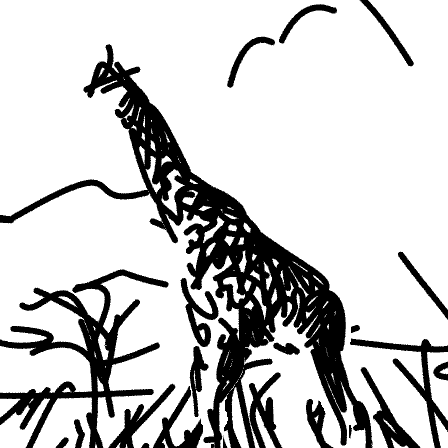}} &
    \hspace{0.5cm}
    \frame{\includegraphics[width=0.098\textwidth,height=0.098\textwidth]{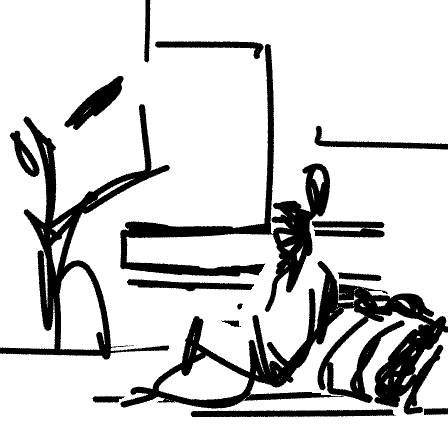}} &
    \frame{\includegraphics[width=0.098\textwidth,height=0.098\textwidth]{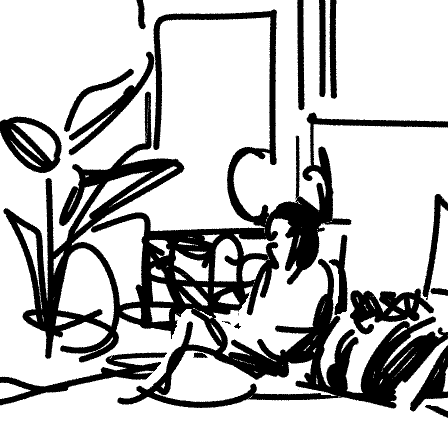}} &
    \frame{\includegraphics[width=0.098\textwidth,height=0.098\textwidth]{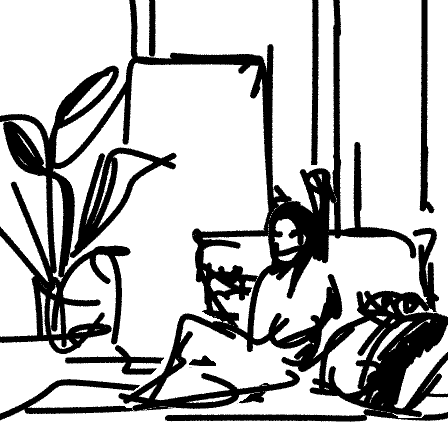}} &
    \frame{\includegraphics[width=0.098\textwidth,height=0.098\textwidth]{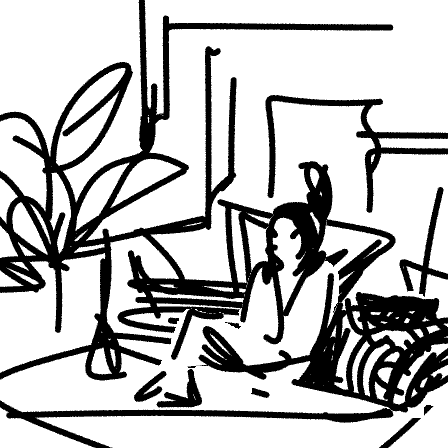}} \\
    
    \frame{\includegraphics[width=0.098\textwidth,height=0.098\textwidth]{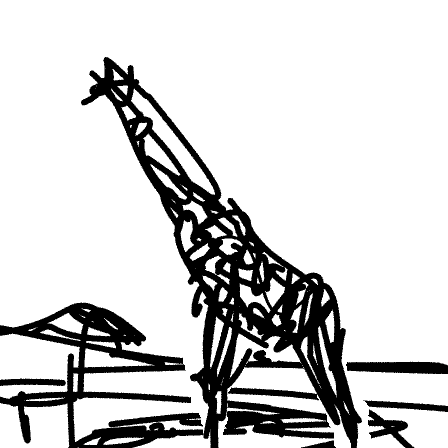}} &
    \frame{\includegraphics[width=0.098\textwidth,height=0.098\textwidth]{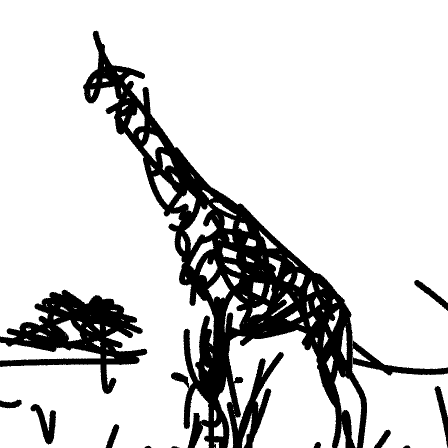}} &
    \frame{\includegraphics[width=0.098\textwidth,height=0.098\textwidth]{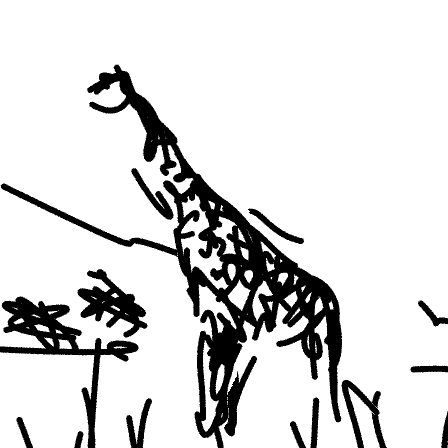}} &
    \frame{\includegraphics[width=0.098\textwidth,height=0.098\textwidth]{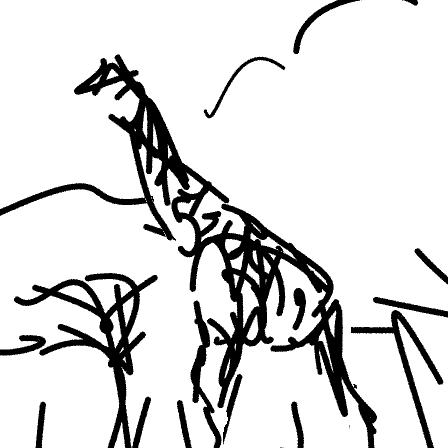}} &
    \hspace{0.5cm}
    \frame{\includegraphics[width=0.098\textwidth,height=0.098\textwidth]{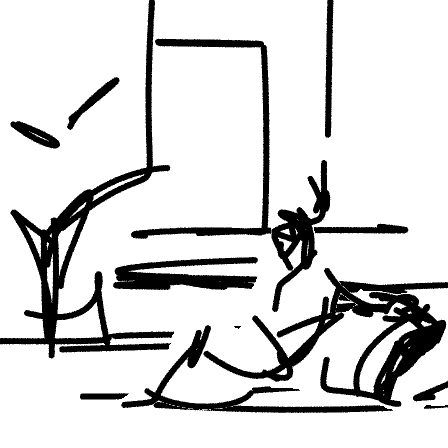}} &
    \frame{\includegraphics[width=0.098\textwidth,height=0.098\textwidth]{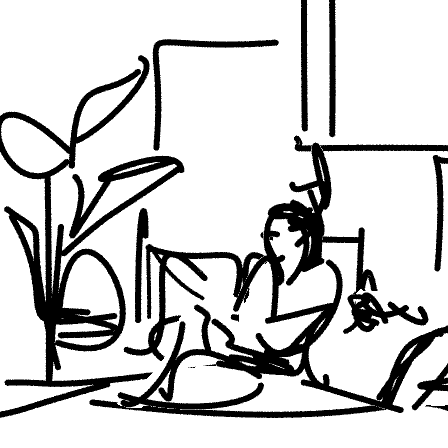}} &
    \frame{\includegraphics[width=0.098\textwidth,height=0.098\textwidth]{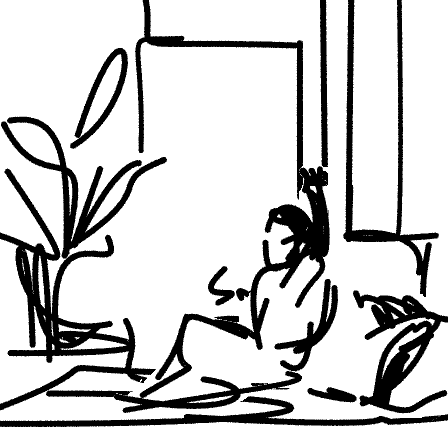}} &
    \frame{\includegraphics[width=0.098\textwidth,height=0.098\textwidth]{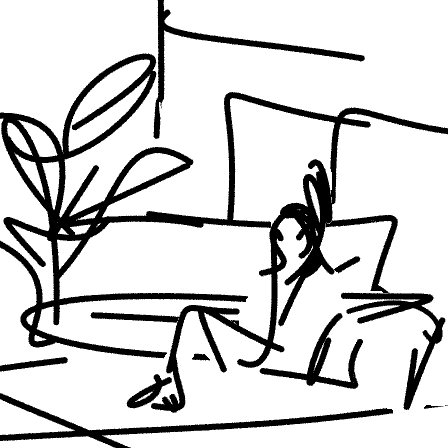}} \\
    
    \frame{\includegraphics[width=0.098\textwidth,height=0.098\textwidth]{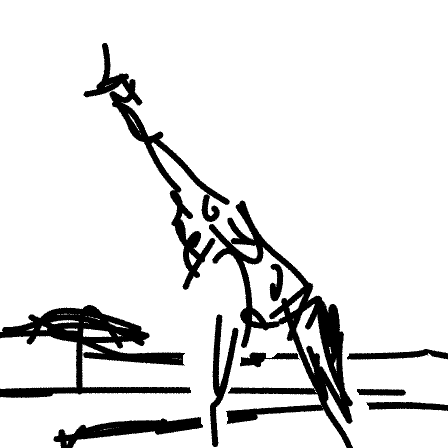}} &
    \frame{\includegraphics[width=0.098\textwidth,height=0.098\textwidth]{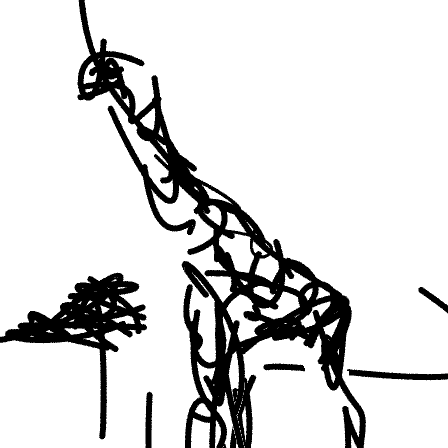}} &
    \frame{\includegraphics[width=0.098\textwidth,height=0.098\textwidth]{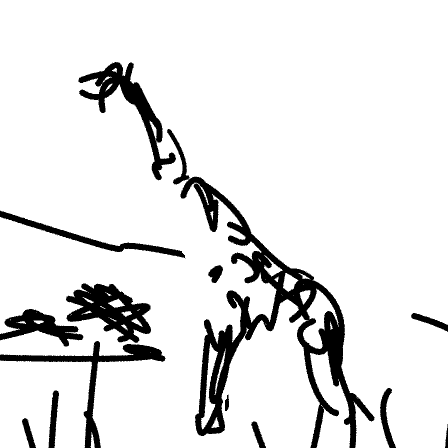}} &
    \frame{\includegraphics[width=0.098\textwidth,height=0.098\textwidth]{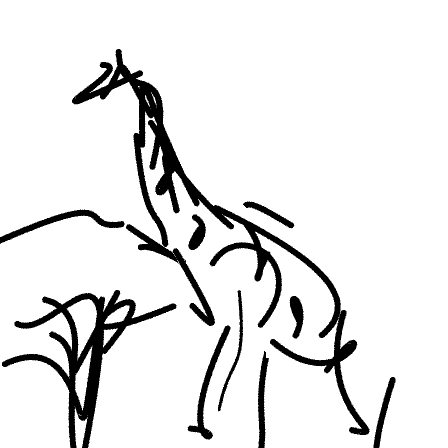}} &
    \hspace{0.5cm}
    \frame{\includegraphics[width=0.098\textwidth,height=0.098\textwidth]{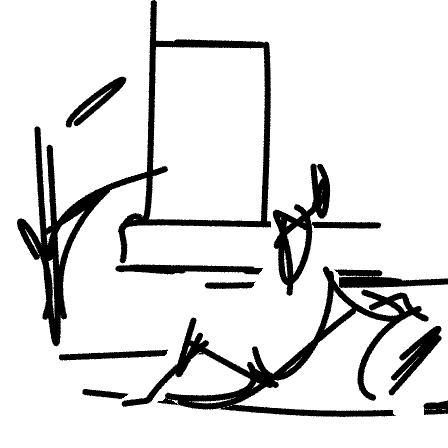}} &
    \frame{\includegraphics[width=0.098\textwidth,height=0.098\textwidth]{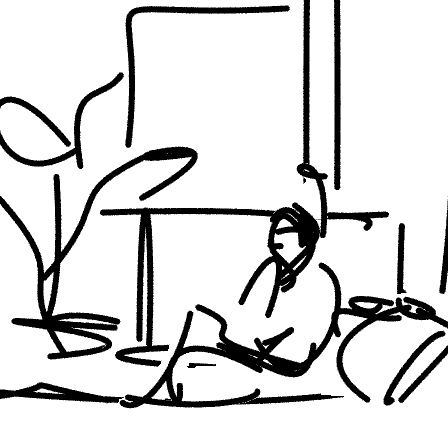}} &
    \frame{\includegraphics[width=0.098\textwidth,height=0.098\textwidth]{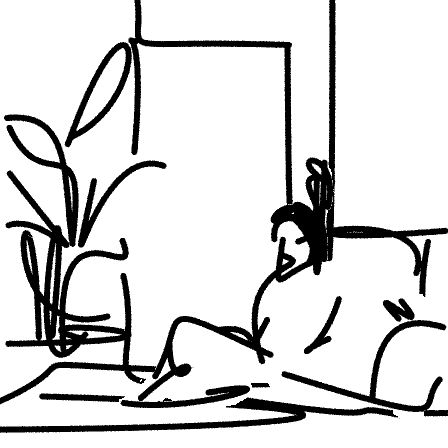}} &
    \frame{\includegraphics[width=0.098\textwidth,height=0.098\textwidth]{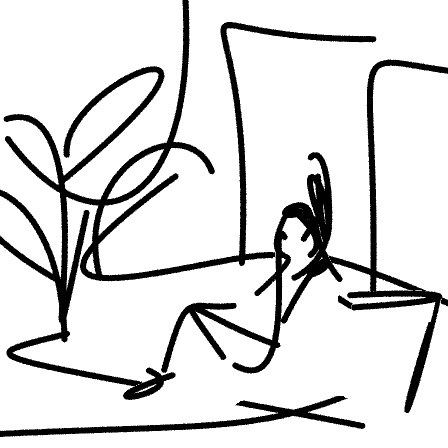}} \\
    
    \frame{\includegraphics[width=0.098\textwidth,height=0.098\textwidth]{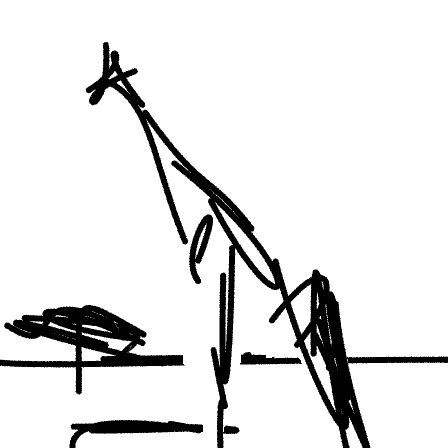}} &
    \frame{\includegraphics[width=0.098\textwidth,height=0.098\textwidth]{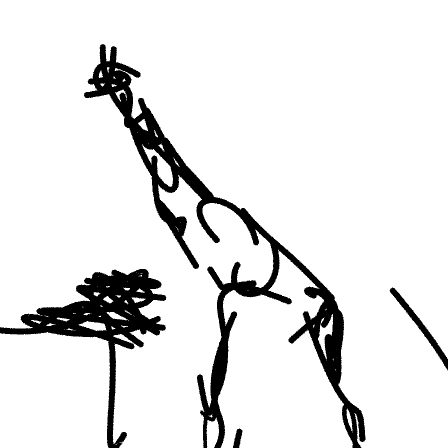}} &
    \frame{\includegraphics[width=0.098\textwidth,height=0.098\textwidth]{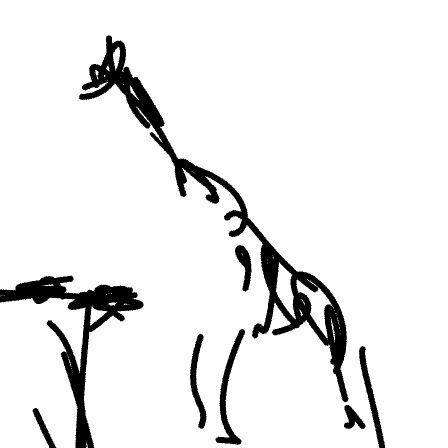}} &
    \frame{\includegraphics[width=0.098\textwidth,height=0.098\textwidth]{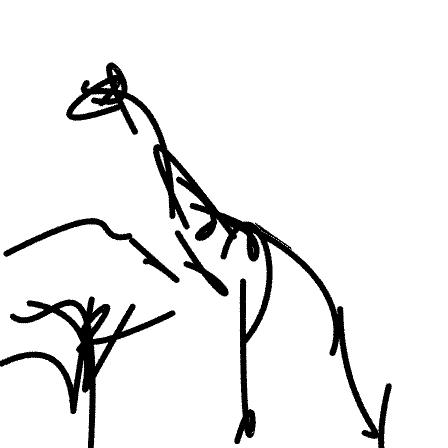}} &
    \hspace{0.5cm}
    \frame{\includegraphics[width=0.098\textwidth,height=0.098\textwidth]{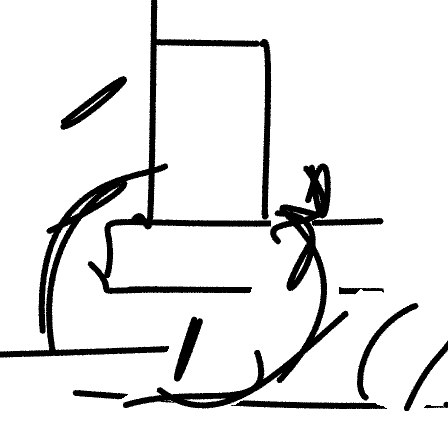}} &
    \frame{\includegraphics[width=0.098\textwidth,height=0.098\textwidth]{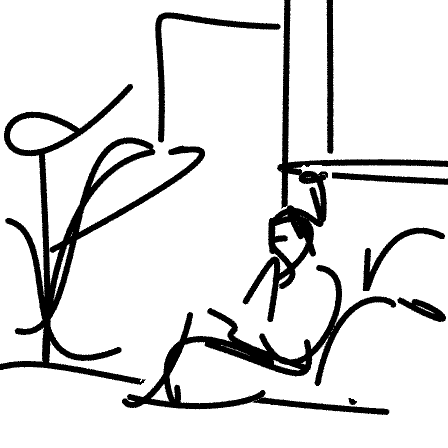}} &
    \frame{\includegraphics[width=0.098\textwidth,height=0.098\textwidth]{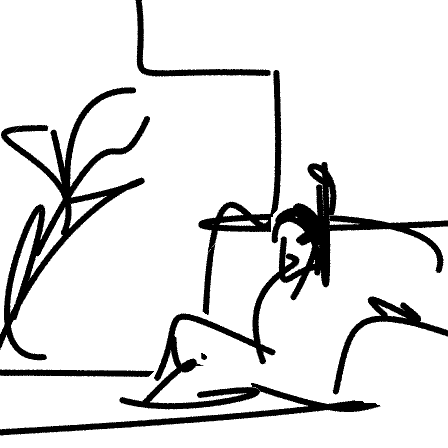}} &
    \frame{\includegraphics[width=0.098\textwidth,height=0.098\textwidth]{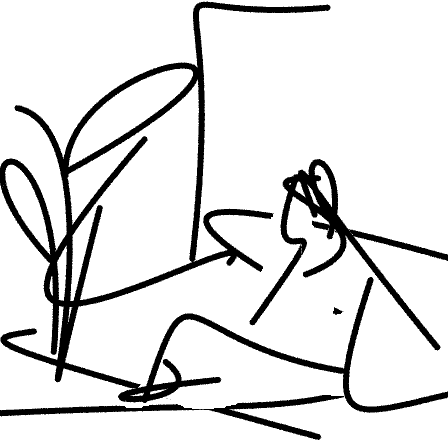}} \\

    \\
    \\
    
    \includegraphics[width=0.098\textwidth,height=0.098\textwidth]{clipascene/figs/inputs/cat_c.jpg} & & & &
    \hspace{0.5cm}
    \includegraphics[width=0.098\textwidth,height=0.098\textwidth]{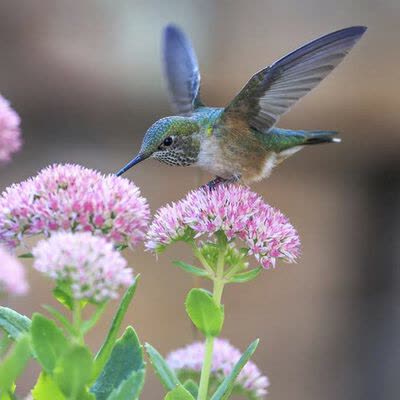} & & & \\

    \frame{\includegraphics[width=0.098\textwidth,height=0.098\textwidth]{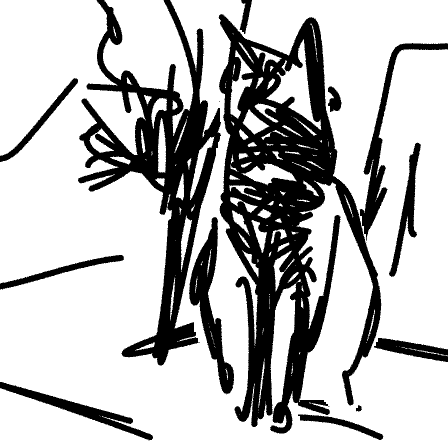}} &
    \frame{\includegraphics[width=0.098\textwidth,height=0.098\textwidth]{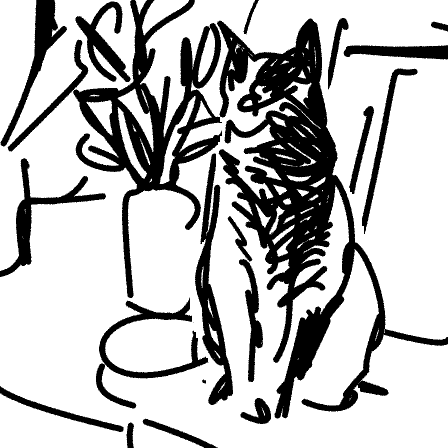}} &
    \frame{\includegraphics[width=0.098\textwidth,height=0.098\textwidth]{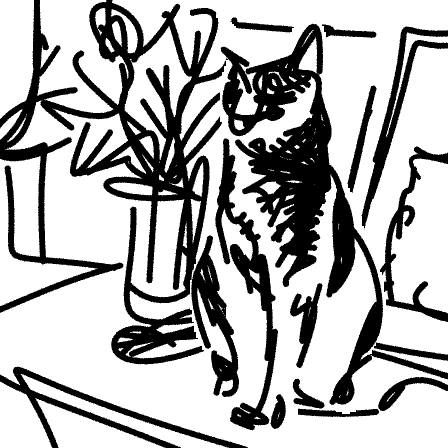}} &
    \frame{\includegraphics[width=0.098\textwidth,height=0.098\textwidth]{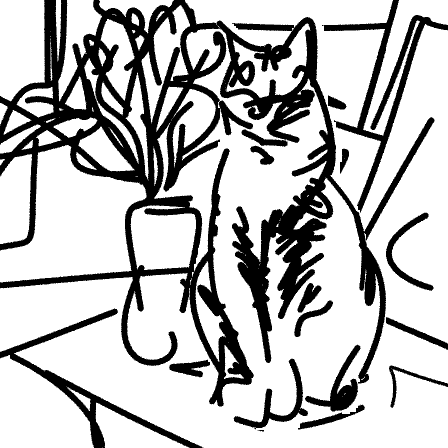}} &
    \hspace{0.5cm}
    \frame{\includegraphics[width=0.098\textwidth,height=0.098\textwidth]{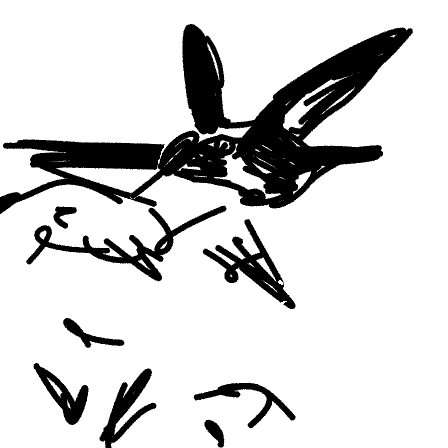}} &
    \frame{\includegraphics[width=0.098\textwidth,height=0.098\textwidth]{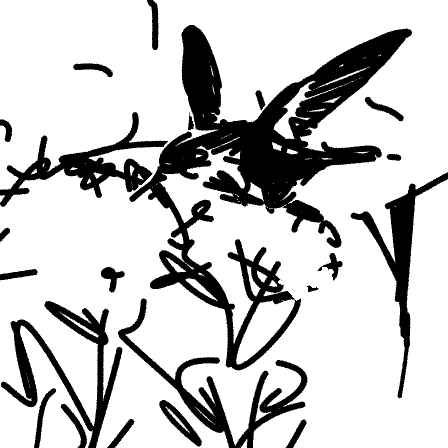}} &
    \frame{\includegraphics[width=0.098\textwidth,height=0.098\textwidth]{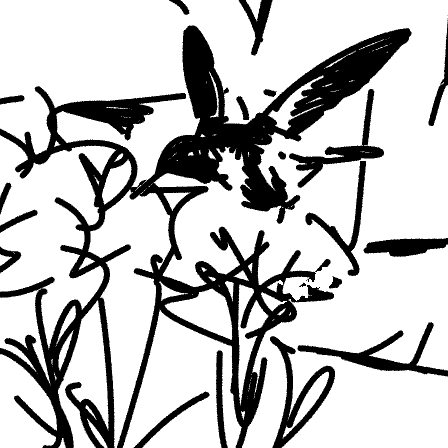}} &
    \frame{\includegraphics[width=0.098\textwidth,height=0.098\textwidth]{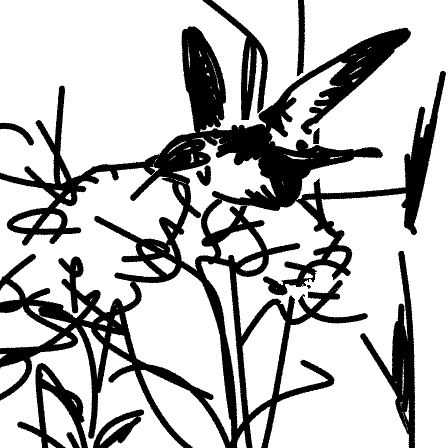}} \\
    
    \frame{\includegraphics[width=0.098\textwidth,height=0.098\textwidth]{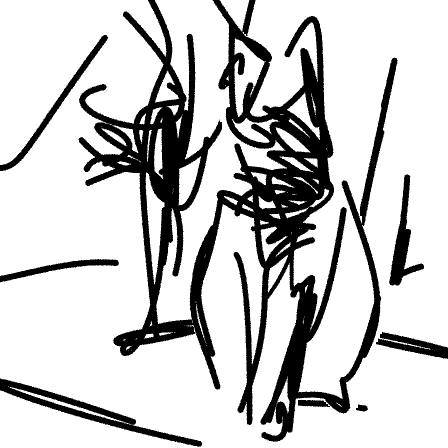}} &
    \frame{\includegraphics[width=0.098\textwidth,height=0.098\textwidth]{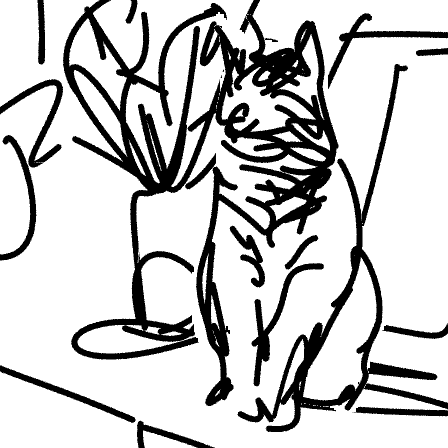}} &
    \frame{\includegraphics[width=0.098\textwidth,height=0.098\textwidth]{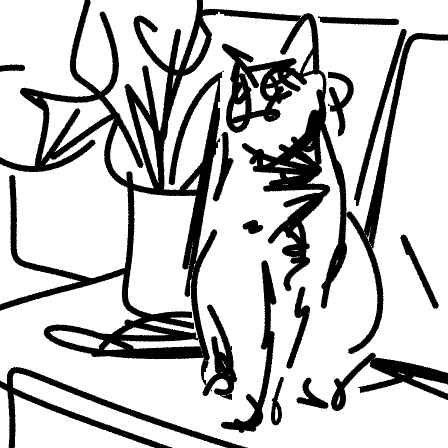}} &
    \frame{\includegraphics[width=0.098\textwidth,height=0.098\textwidth]{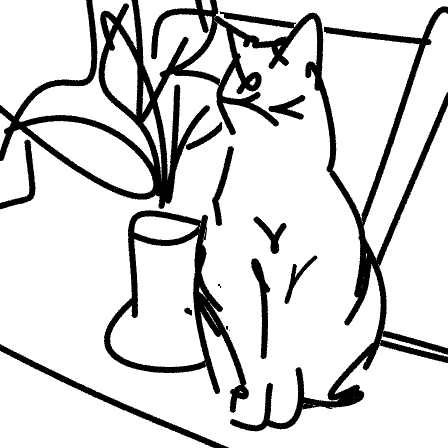}} &
    \hspace{0.5cm}
    \frame{\includegraphics[width=0.098\textwidth,height=0.098\textwidth]{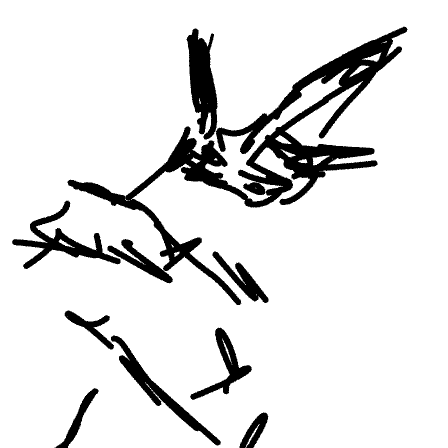}} &
    \frame{\includegraphics[width=0.098\textwidth,height=0.098\textwidth]{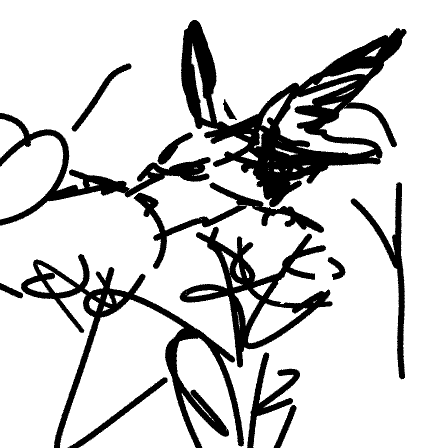}} &
    \frame{\includegraphics[width=0.098\textwidth,height=0.098\textwidth]{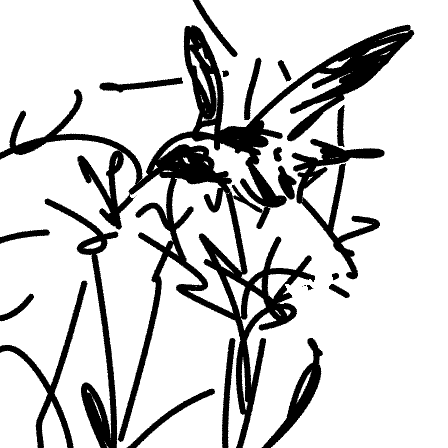}} &
    \frame{\includegraphics[width=0.098\textwidth,height=0.098\textwidth]{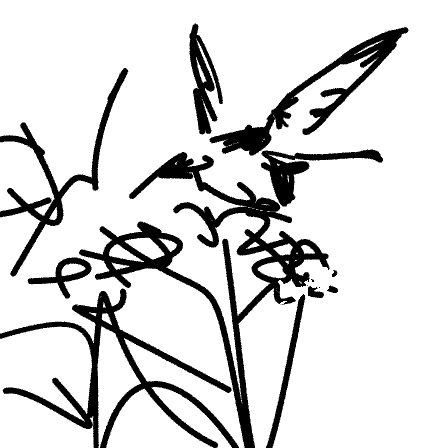}} \\
    
    \frame{\includegraphics[width=0.098\textwidth,height=0.098\textwidth]{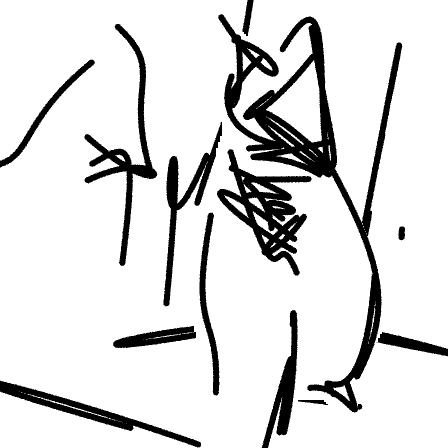}} &
    \frame{\includegraphics[width=0.098\textwidth,height=0.098\textwidth]{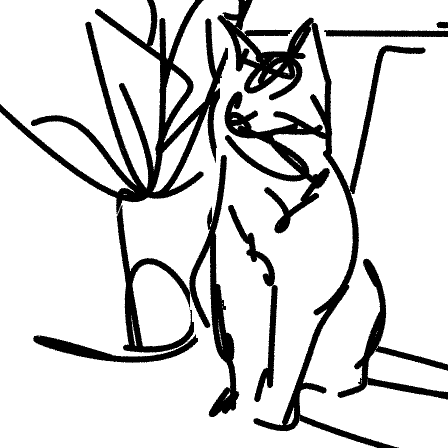}} &
    \frame{\includegraphics[width=0.098\textwidth,height=0.098\textwidth]{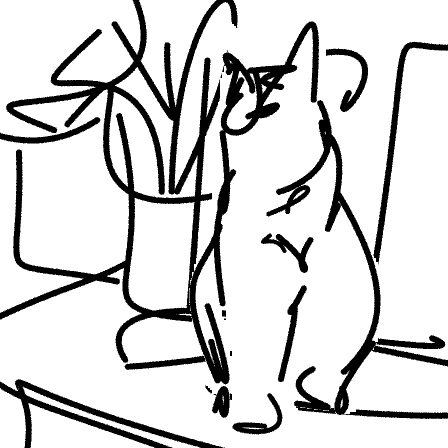}} &
    \frame{\includegraphics[width=0.098\textwidth,height=0.098\textwidth]{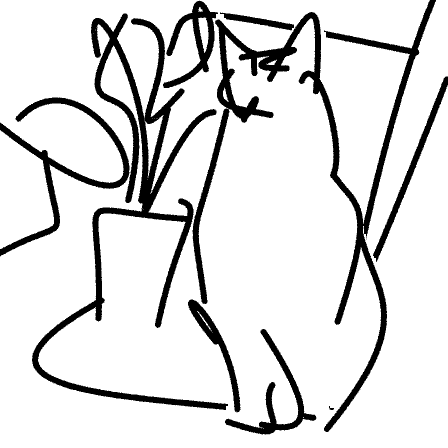}} &
    \hspace{0.5cm}
    \frame{\includegraphics[width=0.098\textwidth,height=0.098\textwidth]{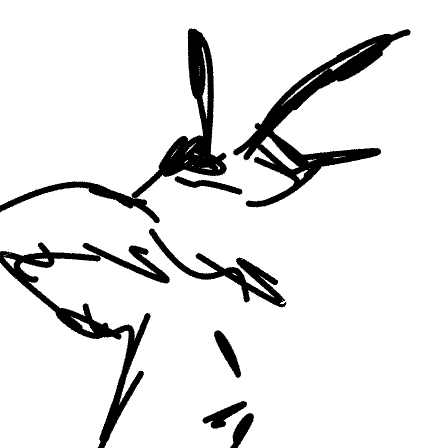}} &
    \frame{\includegraphics[width=0.098\textwidth,height=0.098\textwidth]{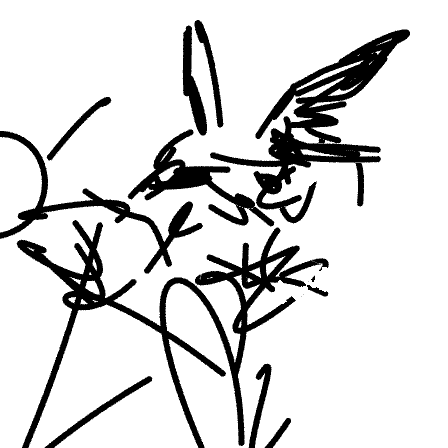}} &
    \frame{\includegraphics[width=0.098\textwidth,height=0.098\textwidth]{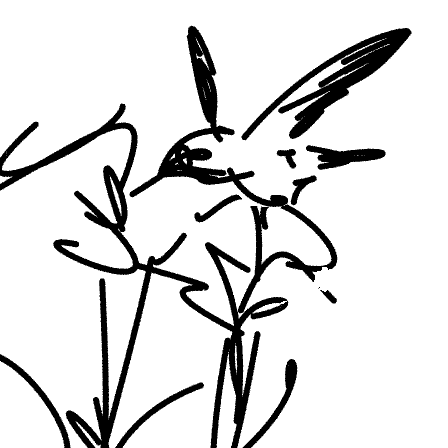}} &
    \frame{\includegraphics[width=0.098\textwidth,height=0.098\textwidth]{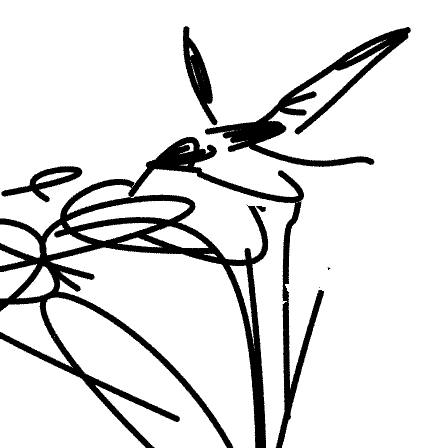}} \\
    
    \frame{\includegraphics[width=0.098\textwidth,height=0.098\textwidth]{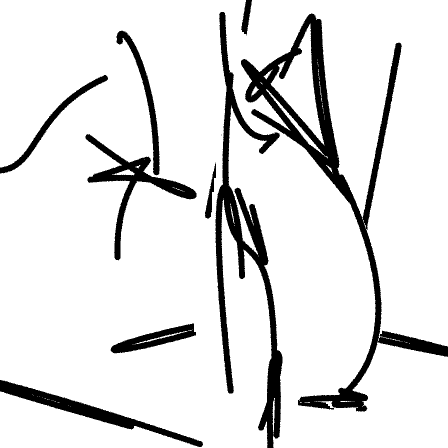}} &
    \frame{\includegraphics[width=0.098\textwidth,height=0.098\textwidth]{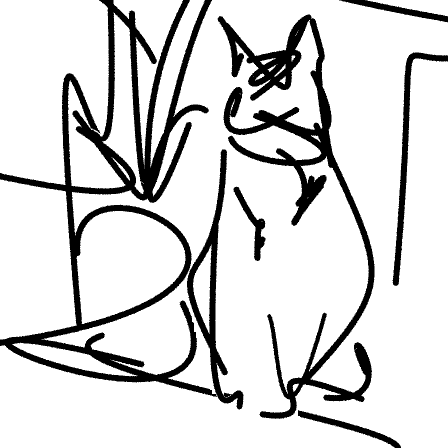}} &
    \frame{\includegraphics[width=0.098\textwidth,height=0.098\textwidth]{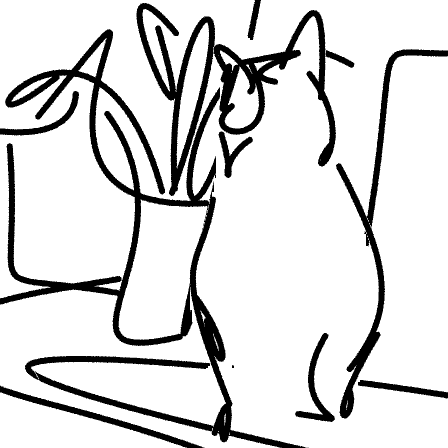}} &
    \frame{\includegraphics[width=0.098\textwidth,height=0.098\textwidth]{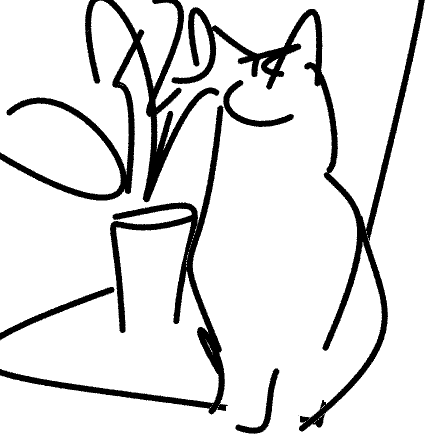}} &
    \hspace{0.5cm}
    \frame{\includegraphics[width=0.098\textwidth,height=0.098\textwidth]{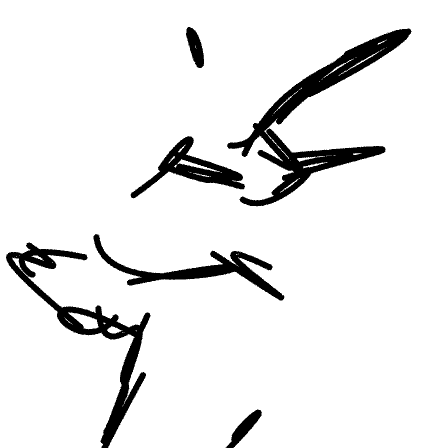}} &
    \frame{\includegraphics[width=0.098\textwidth,height=0.098\textwidth]{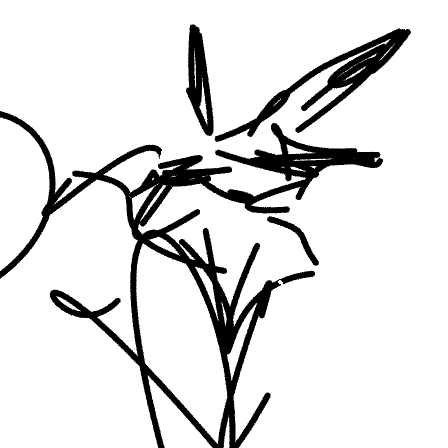}} &
    \frame{\includegraphics[width=0.098\textwidth,height=0.098\textwidth]{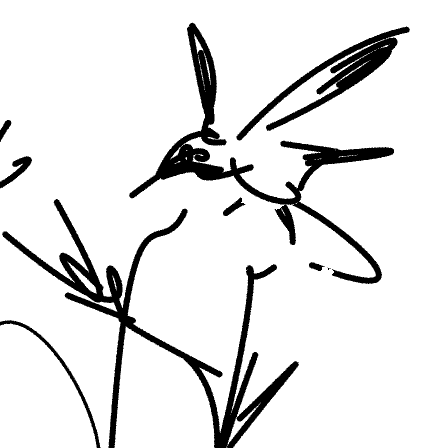}} &
    \frame{\includegraphics[width=0.098\textwidth,height=0.098\textwidth]{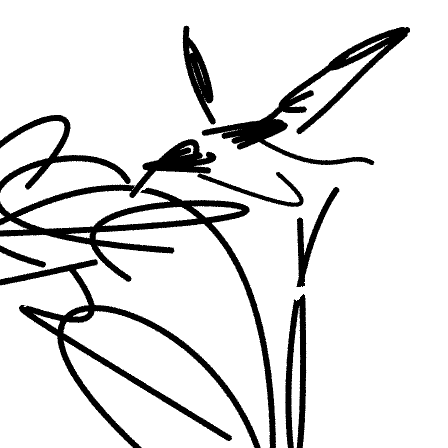}} \\

    \end{tabular}
    \caption{The $4\times4$ matrix of sketches produced by our method. Columns from left to right illustrate the change in fidelity, from precise to loose, and rows from top to bottom illustrate the visual simplification.}
    
    \label{fig:matrix5}
\end{figure*}

\part{Communicative Illustrations in Typography}
\label{part:two}
\chapter{Word-as-Image for Semantic Typography}
\label{chap:word-as-image}
\begin{figure*}[h]
    \centering
    \includegraphics[width=1\linewidth]{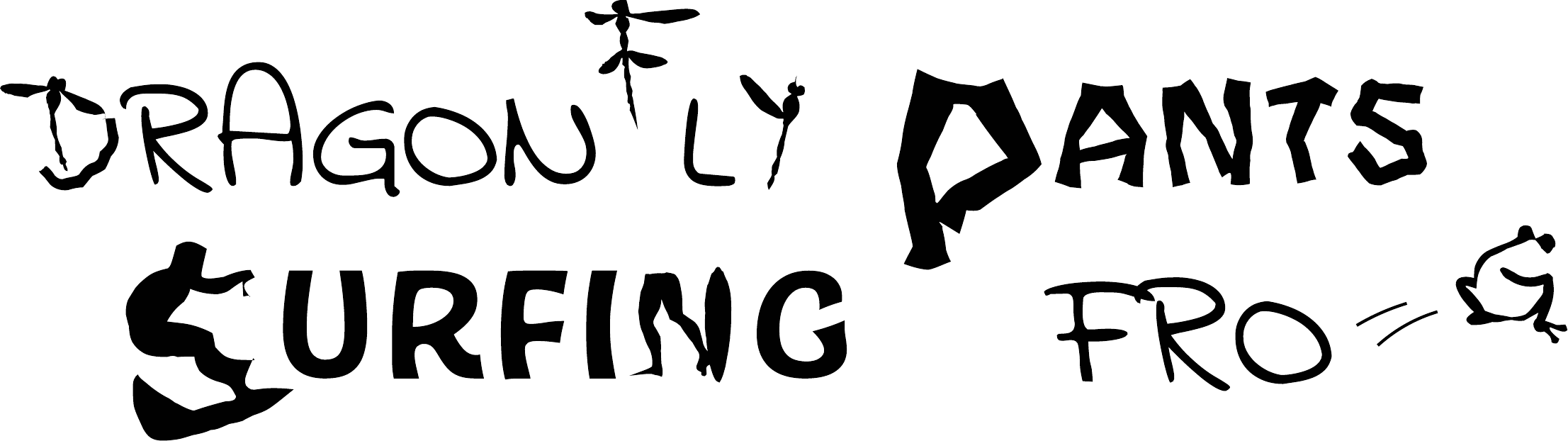} 
    \caption[]{\small A few examples of our word-as-image illustrations in various fonts and for different textual concept. The semantically adjusted letters are created completely automatically using our method, and can then be used for further creative design as we illustrate here.\footnotemark} \label{fig:teaser_wordasim}
\end{figure*}

\footnotetext{Project page: \url{https://wordasimage.github.io/Word-As-Image-Page/}}

Semantic typography is the practice of using typography to visually reinforce the meaning of text. This can be achieved through the choice of typefaces, font sizes, font styles, and other typographic elements. A more elaborate and engaging technique for semantic typography is presented by word-as-image illustrations, where the semantics of a given word are illustrated using only the graphical elements of its letters. 
Such illustrations provide a visual representation of the meaning of the word, while also preserving the readability of the word as a whole.

\begin{figure}[t]
    \centering
    \includegraphics[width=0.8\linewidth]{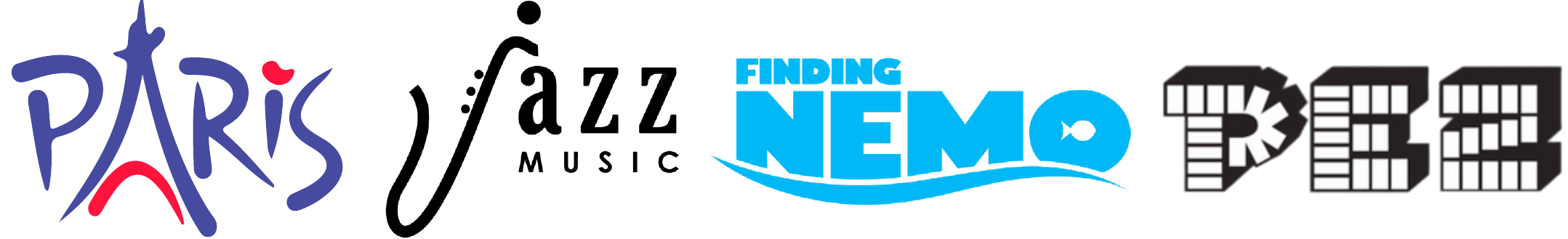}
    \caption{Manually created word-as-image illustrations}
    \label{fig:logos_etc}
\end{figure}

The task of creating a word-as-image is highly challenging, as it requires the ability to understand and depict the visual characteristics of the given concept, and to convey them in a concise, aesthetic, and comprehensible manner without harming legibility. It requires a great deal of creativity and design skills to integrate the chosen visual concept into the letter's shape \cite{WordAsImage}. 
In Figure~\ref{fig:logos_etc} we show some word-as-image examples created manually. For example, to create the ``jazz'' depiction, the designer had to first choose the visual concept that would best fit the semantics of the text (a saxophone), consider the desired font characteristics, and then choose the most suitable letter to be replaced. Finding the right visual element to illustrate a concept is ill-defined as there are countless ways to illustrate any given concept. In addition, one cannot simply copy a selected visual element onto the word -- there is a need to find subtle modifications of the letters shape. 

Because of these complexities, the task of automatic creation of word-as-image illustrations was practically impossible to achieve using computers until recently.
In this paper, we define an algorithm for automatic creation of word-as-image illustrations based on recent advances in deep-learning and the availability of huge foundational models that combine language and visual understanding. 
Our resulting illustrations (see Figure~\ref{fig:teaser_wordasim}) could be used for logo design, for signs, in greeting cards and invitations, and simply for fun. They can be used as-is, or as inspiration for further refinement of the design.

Existing methods in the field of text stylization often rely on raster textures \cite{Yang_2018_Context}, place a manually created style on top of the strokes segmentation \cite{BerioStrokestyles2022}, or deform the text into a pre-defined target shape \cite{zouLegibleCompactCalligrams2016} (see Figure~\ref{fig:prev_work}). 
Only a few works \cite{tendulkarTrickTReATThematic2019, zhangSynthesizingOrnamentalTypefaces2017} deal with \textit{semantic} typography, and they often operate in the raster domain and use existing icons for replacement (see Figure~\ref{fig:prev_work}E).

Our word-as-image illustrations concentrate on changing only the \emph{geometry} of the letters to convey the meaning. We deliberately do not change color or texture and do not use embellishments. This allows simple, concise, black-and-white designs that convey the semantics clearly. In addition, since we preserve the vector-based representation of the letters, this allows smooth rasterization in any size, as well as applying additional style manipulations to the illustration using colors and texture, if desired.

Given an input word, our method is applied separately for each letter, allowing the user to later choose the most likeable combination for replacement.
We represent each letter as a closed vectorized shape, and optimize its parameters to reflect the \emph{meaning} of the word, while still preserving its original style and design.

We rely on the prior of a pretrained Stable Diffusion model ~\cite{rombach2022highresolution} to connect between text and images, and utilize the Score Distillation Sampling approach \cite{poole2022dreamfusion} to encourage the appearance of the letter to reflect the provided textual concept. 
Since the Stable Diffusion model is trained on raster images, we use a differentiable rasterizer \cite{diffvg} that allows to backpropagate gradients from a raster-based loss to the shape's parameters.

To preserve the shape of the original letter and ensure legibility of the word, we utilize two additional loss functions. The first loss regulates the shape modification by constraining the deformation to be as-conformal-as-possible over a triangulation of the letter's shape. The second loss preserves the local tone and structure of the letter by comparing the low-pass filter of the resulting rasterized letter to the original one. 

We compare to several baselines, and present many results using various typefaces and a large number of concepts. Our word-as-image illustrations convey the intended concept while maintaining legibility and preserving the appearance of the font, demonstrating visual creativity.

\begin{figure}[h]
\centering
\includegraphics[width=0.8\linewidth]{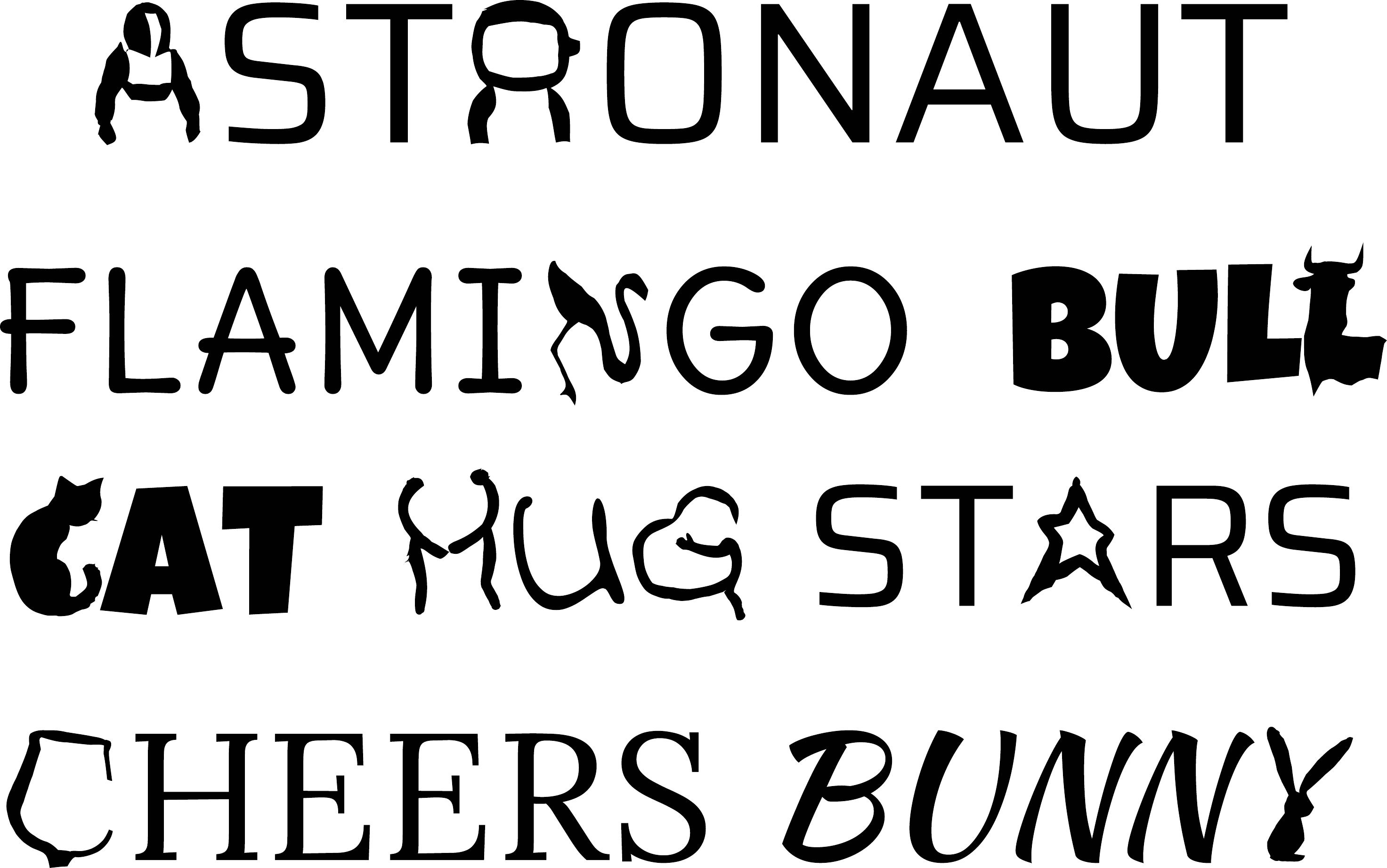}
    \caption{\small More word-as-images produced by our method. Note how styles of different fonts are preserved by the semantic modification.}
    \label{fig:intro_res}
\end{figure}

\section{Related Work}
\label{sec:related}

\begin{figure}[t]
    \centering
    \includegraphics[width=0.7\linewidth]{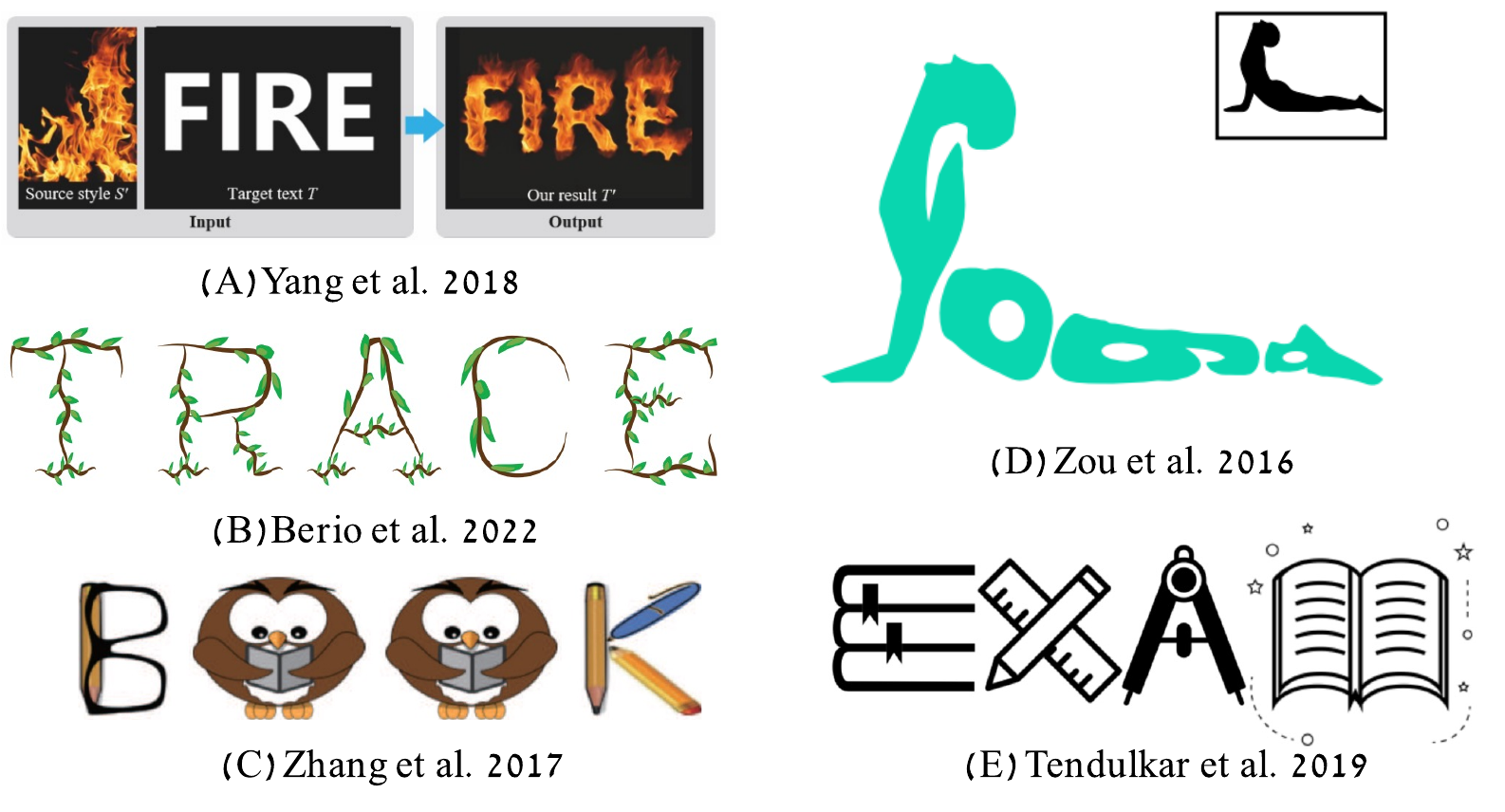}
    \caption{\small Examples of previous text stylization works -- (A) Yang et al. \cite{Yang_2018_Context}, (B) Berio et al. \cite{BerioStrokestyles2022}, (C) Zhang et al. \cite{zhangSynthesizingOrnamentalTypefaces2017}, (D) Zou et al. \cite{zouLegibleCompactCalligrams2016}, and (E) Tendulkar et al. \cite{tendulkarTrickTReATThematic2019}. Most use color and texture or copy icons onto the letters. Our work concentrates on subtle \emph{geometric} shape deformations of the letters to convey the \emph{semantic meaning} without color or texture (that can be added later).}
    \vspace{-0.2cm}
    \label{fig:prev_work}
\end{figure}

\paragraph{Text Stylization}
One approach to text stylization is artistic text style transfer, where the style from a given source image is migrated into the desired text (such as in Figure \ref{fig:prev_work}A).
To tackle this task, existing works incorporate patch-based texture synthesis \cite{Yang_2017_CVPR, SketchPatch2020} as well as variants of GANs \cite{Azadi_2018_CVPR, Wang_2019_CVPR, Jiang_SC_2019, YangGAN2022, Mao_2022_Intelligent}.
These works operate within the raster domain, a format that is undesirable for typographers since fonts must be scalable. In contrast, we operate on the \textit{parametric} outlines of the letters, and our glyph manipulation is guided by the semantic meaning of the word, rather than a pre-defined style image.

A number of works \cite{ha2017neural,Lopes_2019_ICCV,wangDeepVecFontSynthesizingHighquality2021} tackle the task of font generation and stylization in the vector domain. 
Commonly, a latent feature space of font's outlines is constructed, represented as outline samples \cite{Campbell2014, Balashova2018} or parametric curve segments \cite{ha2017neural,Lopes_2019_ICCV,wangDeepVecFontSynthesizingHighquality2021}. These approaches are often limited to mild deviations from the input data. Other methods rely on
templates \cite{Suveeranont2010, lian2018easyfont} or on user guided \cite{Phan2015} and automatic \cite{BerioStrokestyles2022} stroke segmentation to produce letter stylization (such as in Figure \ref{fig:prev_work}B). However, they rely on a manually defined style, while we rely on the expressiveness of Stable Diffusion to guide the modification of the letters' shape, to convey the \textit{meaning} of the provided word.
In the task of calligram generation \cite{zouLegibleCompactCalligrams2016,xuCalligraphicPacking2007} the entire word is deformed into a \textit{given} target shape. 
This task prioritises shape over the readability of the word (see Figure \ref{fig:prev_work}D), and is inherently different from ours, as we use the \textit{semantics} of the word to derive the deformation of individual letters.

Most related to our goal, are works that perform semantic stylization of text. \cite{tendulkarTrickTReATThematic2019} replace letters in a given word with clip-art icons describing a given theme (see Figure \ref{fig:prev_work}E). To choose the most suitable icon for replacement, an autoencoder is used to measure the distance between the letter and icons from the desired class. 
Similarly, \cite{zhangSynthesizingOrnamentalTypefaces2017} replace stroke-like parts of one or more letters with instances of clip art to generate ornamental stylizations. An example is shown in Figure \ref{fig:prev_work}C.
These approaches operate in the raster domain, and replace letters with existing icons, which limits them to a predefined set of classes present in the dataset. Our method, however, operates in the vector domain, and incorporates the expressiveness of large pretrained image-language models to create a new illustration that conveys the desired concept.

\paragraph{Large Language-Vision Models}
With the recent advancement of language-vision models \cite{Radfordclip} and diffusion models ~\cite{ramesh2022hierarchical, nichol2021glide, rombach2022highresolution}, the field of image generation and editing has undergone unprecedented evolution. 
Having been trained on millions of images and text pairs, these models have proven effective for performing challenging vision related tasks such as image segmentation \cite{SegDiff}, domain adaptation \cite{DomainAdaptationSong}, image editing \cite{Avrahami_2022_CVPR, hertz2022prompt, Plug-and-Play}, personalization \cite{ruiz2023dreambooth, gal2022textual}, and explainability \cite{Chefer_2021_ICCV}.
Despite being trained on raster images, the strong visual and semantic priors have also been shown to be successfully applied to other domains, such as motion \cite{tevet2022motionclip}, meshes \cite{text2mesh}, point cloud \cite{PointCLIP}, and vector graphics. 
CLIPDraw \cite{CLIPDraw} uses a differentiable rasterizer \cite{diffvg} to optimize a set of colorful curves w.r.t. a given text prompt, guided by CLIP's image-text similarity metric.
Tian and Ha \cite{TianEvolution2021} use evolutionary algorithms combined with CLIP guidance to create abstract visual concepts based on text. Other works \cite{vinker2022clipasso, clipascene} utilize the image encoder of CLIP to generate abstract vector sketches from images. 

Diffusion models have been used for the task of text guided image-to-image translation \cite{ILVR, Plug-and-Play}.
In SDEdit \cite{meng2022sdedit}, an adequate amount of noise is added to a reference image, such that its overall structure is preserved, and then the image is denoised in a reverse process with a guiding text. 
Pretrained diffusion models have also been used to generate 3D objects \cite{poole2022dreamfusion,Latent-NeRF}, or vector art \cite{jain2022vectorfusion} conditioned on text.

In our work we also utilize the strong visual and semantic prior induced by a pretrained Stable Diffusion model \cite{rombach2022highresolution}, however, for the task of \textit{semantic typography}. For that purpose we add new components to the optimization process to preserve the font's style and text legibility.

\begin{figure}[t]
    \centering
    \includegraphics[width=0.7\linewidth]{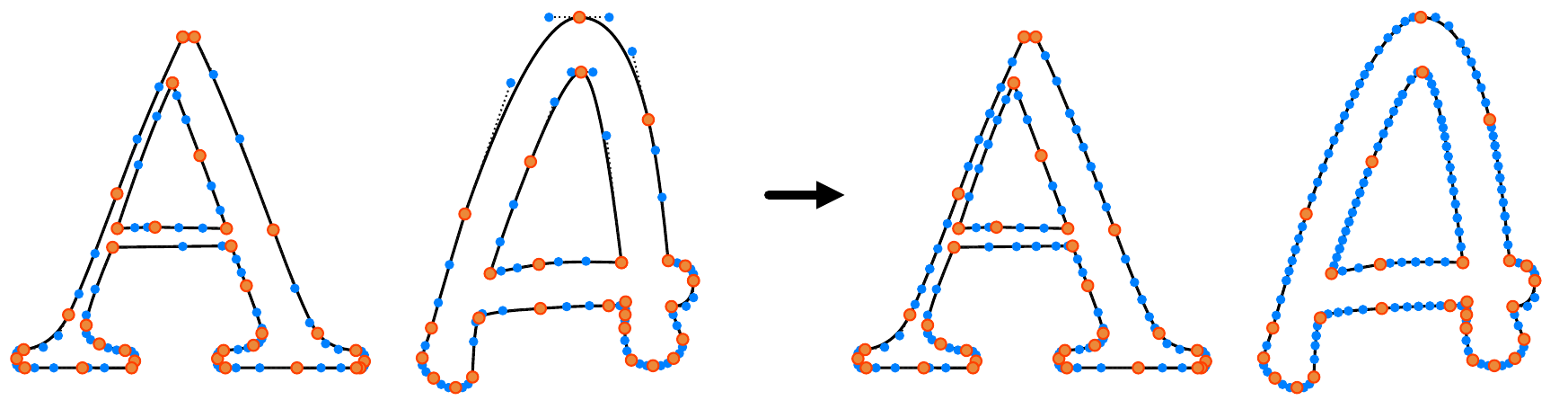}
    \caption{\small Illustration of the letter's outline and control points before (left) and after (right) the subdivision process. The orange dots are the initial \Bezier curve segment endpoints. The blue dots are the remaining control points respectively before and after subdivision. }
    \label{fig:initCC}
\end{figure}

\section{Method}
\label{sec:method1}
Given a word $W$ represented as a string with $n$ letters $\{l_1, ... l_n\}$, our method is applied to every letter $l_i$ separately to produce a semantic visual depiction of the letter. The user can then choose which of the letters to replace and which to keep in their original form. The final placement of any replaced letter is done by scaling it to fit the bounding box of the original letter, while aligning their centers and preserving the original spacing between all the word’s letters.

\begin{figure*}
    \centering
    \includegraphics[width=1\linewidth]{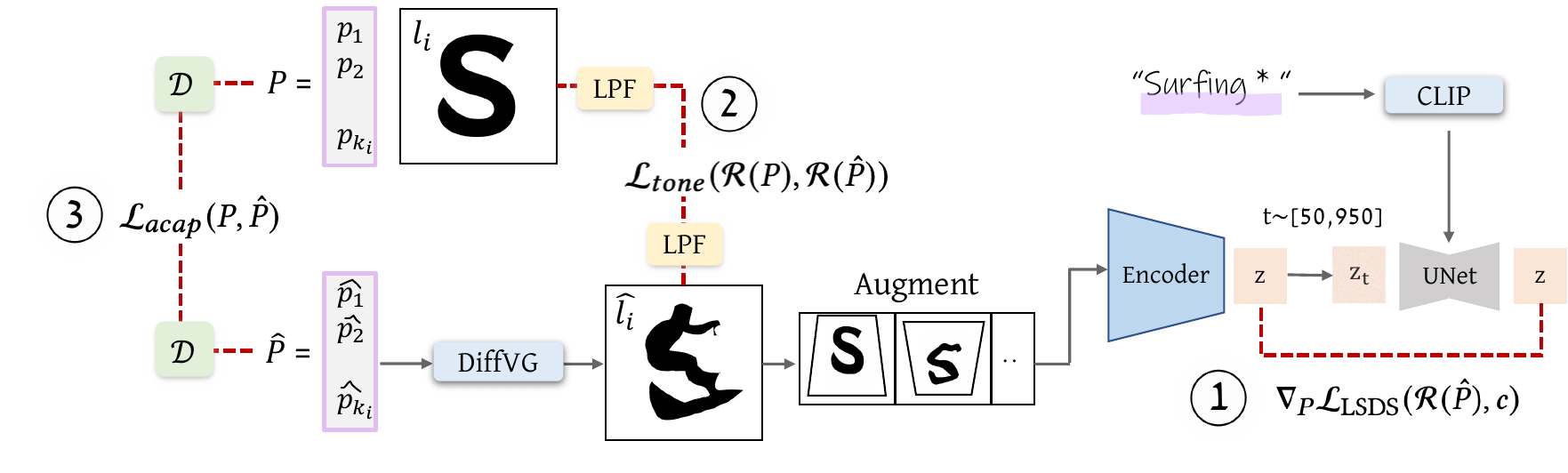} 
    \caption{\small An overview of our method. Given an input letter $l_i$ represented by a set of control points $P$, and a concept (shown in purple), we optimize the new positions $\hat{P}$ of the deformed letter $\hat{l_i}$ iteratively using three loss functions together (marked as 1-3). At each iteration, the set $\hat{P}$ is fed into a differentiable rasterizer (DiffVG marked in blue) that outputs the rasterized deformed letter $\hat{l_i}$. $\hat{l_i}$ is then augmented and passed into a pretrained frozen Stable Diffusion model, that drives the letter shape to convey the semantic concept using the $\nabla_{\hat{P}} \mathcal{L}_\text{LSDS}$ loss (1). $l_i$ and $\hat{l_i}$ are also passed through a low pass filter (LPF marked in yellow) to compute $\mathcal{L}_{tone}$ loss (2), which encourages the preservation of the overall tone of the font style and also the local letter shape. Additionally, the sets $P$ and $\hat{P}$ are passed through a  Delaunay triangulation operator ($\mathcal{D}$ marked in green), defining $\mathcal{L}_{acap}$ loss (3) which encourages the preservation of the initial shape.}
    \label{fig:method}
\end{figure*}

\subsection{Letter Representation}
\label{subsec:letter_representation}
We begin by defining the parametric representation of the letters in $W$.
We use the FreeType font library \cite{freetype} to extract the outline of each letter. We then translate each outline into a set of cubic \Bezier curves, to have a consistent representation across different fonts and letters, and to facilitate the use of diffvg \cite{diffvg} for differentiable rasterization. 

Depending on the letter's complexity and the style of the font, the extracted outlines are defined by a different number of control points.
We have found that the number of control points defining the letter shape affects the final appearance significantly: as the number of control points increases, there is more freedom for visual changes to occur.
Therefore, we apply subdivision to letters that initially contain a small number of control points as follows. 
We define a desired number of control points to each letter of the alphabet (shared across different fonts), and then iteratively subdivide the \Bezier segments until reaching at least this number. 
At each iteration, we compute the maximum arc length among all \Bezier segments and split each segment with this length into two (see Figure~\ref{fig:initCC}). We analyse the effect of the number of control points in Section \ref{sec:ablation}.

This procedure defines a set of $k_i$ control points $P_i = \{p_j\}_{j=1}^{k_i}$ representing the shape of each letter $l_i$.

\begin{figure}[t]
    \centering
    \includegraphics[width=0.65\linewidth]{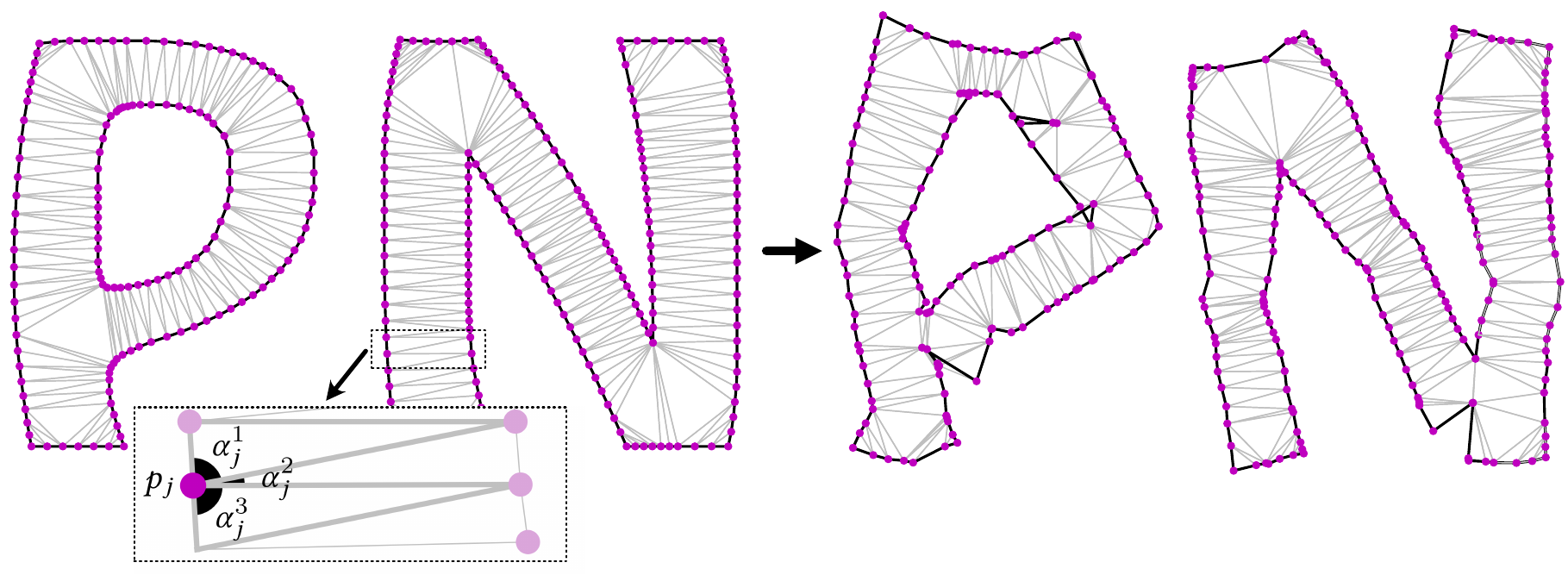}
    \caption{\small Visual illustration of the constraint Delaunay triangulation applied to the initial shapes (left) and the resulting ones (right), for the word ``pants''. The ACAP loss maintains the structure of the letter after the deformation. The zoomed rectangle shows the angles for a given control point $p_j$. }
    \label{fig:triangulation}
    \vspace{-0.2cm}
\end{figure}

\subsection{Optimization}
The pipeline of our method is provided in Figure \ref{fig:method}. Since we are optimizing each letter $l_i$ separately, for brevity, we will omit the letter index $i$ in the following text and define the set of control points for the input letter as $P$. 

Given $P$ and the desired textual concept $c$ (both marked in purple in Figure \ref{fig:method}), our goal is to produce a new set of control points, $\hat{P}$, defining an adjusted letter $\hat{l}$ that conveys the given concept, while maintaining the overall structure and characteristics of the initial letter $l$.
We use three loss functions together to guide the optimization process while changing their weight through the optimization process (Section~\ref{sec:weighting}).

We initialize the learned set of control points $\hat{P}$ with $P$, and pass it through a differentiable rasterizer $\mathcal{R}$ \cite{diffvg} (marked in blue), which outputs the rasterized letter $\mathcal{R}(\hat{P})$.
The rasterized letter is then randomly augmented and passed into a pretrained Stable Diffusion \cite{rombach2022highresolution} model, conditioned on the CLIP's embedding of the given text $c$.
The SDS loss $\nabla_{\hat{P}} \mathcal{L}_\text{LSDS}$ is then used to encourage $\mathcal{R}(\hat{P})$ to convey the given text prompt.

To preserve the shape of each individual letter and ensure the legibility of the word as a whole, we use two additional loss functions to guide the optimization process. The first loss limits the overall shape change by defining as-conformal-as-possible constraint on the shape deformation. The second loss preserves the overall shape and style of the font by constraining the tone (i.e. amount of dark vs. light areas in local parts of the shape) of the modified letter not to diverge too much from the original letter (see Section \ref{subsec:losses}). 

The gradients obtained from all the losses are then backpropagated, to update the parameters $\hat{P}$. We repeat this process for 500 steps, which takes $\sim 5$ minutes to produce a single letter illustration on RTX2080 GPU. %

\subsection{Loss Functions}
\label{subsec:losses}
Our primary objective of encouraging the resulting shape to convey the intended semantic concept, is utilized by $\nabla_{\hat{P}} \mathcal{L}_\text{LSDS}$ loss (described in \Cref{chap:background}).
We observe that using $\nabla_{\hat{P}} \mathcal{L}_\text{LSDS}$ solely can cause large deviations from the initial letter appearance, which is undesired. Hence, our additional goal is to maintain the shape and legibility of the letter $\mathcal{R}(\hat{P})$, as well as to keep the original font's characteristics. For that purpose we use two additional losses.

\paragraph{As-Conformal-As-Possible Deformation Loss}
To prevent the final letter shape from diverging too much from the initial shape, we triangulate the inner part of the letter and constrain the deformation of the letter to be as conformal as possible (ACAP) \cite{hormann2000mips}. 
We use constrained Delaunay triangulation \cite{delaunay1934sphere, barber1995qhull} on the set of control points defining the glyph. It is known that Delaunay triangulation can be used to produce the skeleton of an outline \cite{prasad1997morphological,zou2001shape}, so the ACAP loss also implicitly captures a skeletal representation of the letter form.

The Delaunay triangulation $\mathcal{D}(P)$ splits the glyph represented by $P$ into a set of triangles. This defines a set of size $m_j$ of corresponding angles for each control point $p_j$ (see Figure \ref{fig:triangulation}). We denote this set of angles as $\{\alpha_{j}^i\}_{i=1}^{m_j}$. 
The ACAP loss encourages the induced angles of the optimized shape $\hat{P}$ not to deviate much from the angles of the original shape $P$, and is defined as the L2 distance between the corresponding angles:
\begin{equation}
\label{eqn:acap_loss}
    \mathcal{L}_{acap}(P, \hat{P}) = \frac{1}{k} \sum_{j=1}^{k} \left( \sum_{i=1}^{m_j} \big{(} \alpha_j^i - \hat{\alpha}_j^i \big{)}^2 \right)
\end{equation}
where $k=|P|$ and $\hat{\alpha}$ are the angles induced by $\mathcal{D}(\hat{P})$.

\begin{figure}
    \centering
    \includegraphics[width=0.6\linewidth]{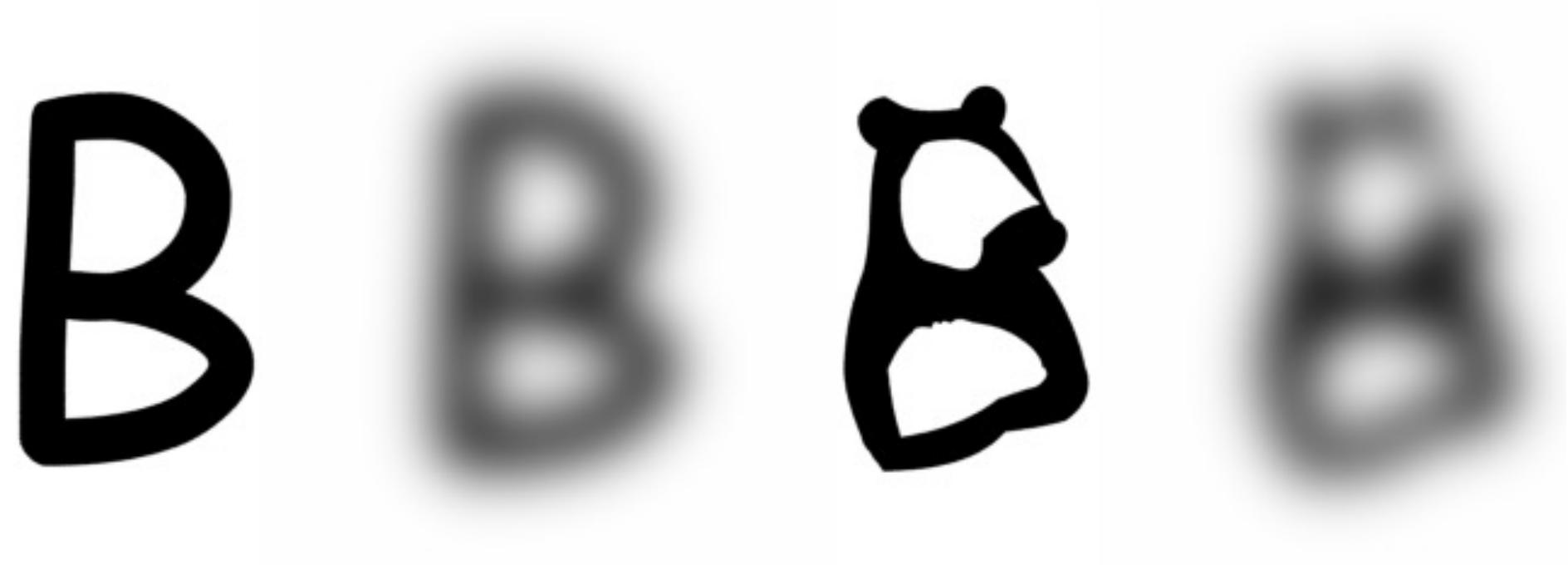}
    \caption{\small Our tone-preserving loss preserves the local tone of the font by comparing the low-pass filter of the letters images before (left) and after deformation (right). It constrains the adjusted letter not to deviate too much from the original.
    This example is of the letter B and the word ``Bear''.}
    \label{fig:LPF}
\end{figure}

\paragraph{Tone Preservation Loss}
To preserve the style of the font as well as the structure of the letter we add a local-tone preservation loss term. This term constrains the tone (amount of black vs. white in all regions of the shape) of the adjusted letter not to deviate too much from tone of the original font's letter. Towards this end, we apply a low pass filter (LPF) to the rasterized letter (before and after deformation) and compute the L2 distance between the resulting blurred letters:
\begin{equation}
\label{eqn:tone_loss}
    \mathcal{L}_{tone}= \big{\|} LPF(\mathcal{R}(P)) - LPF(\mathcal{R}(\hat{P})) \big{\|}_2^2
\end{equation}
An example of the blurred letters is shown in Figure \ref{fig:LPF}, as can be seen, we use a high value of standard deviation $\sigma$ in the blurring kernel to blur out small details such as the ears of bear.

Our final objective is then defined by the weighted average of the three terms:
\begin{equation}
\begin{aligned}
\label{eqn:final_loss}
    \min_{\hat{P}} \nabla_{\hat{P}} \mathcal{L}_\text{LSDS}(\mathcal{R}(\hat{P}), c) + \alpha\cdot\mathcal{L}_{acap}(P,\hat{P}) \\
    + \beta_t\cdot\mathcal{L}_{tone}(\mathcal{R}(P),\mathcal{R}(\hat{P}))    
\end{aligned}
\end{equation}
where $\alpha=0.5$ and $\beta_t$ depends on the step $t$ as described next.

\begin{figure}[t]
\centering
\includegraphics[width=0.6\linewidth]{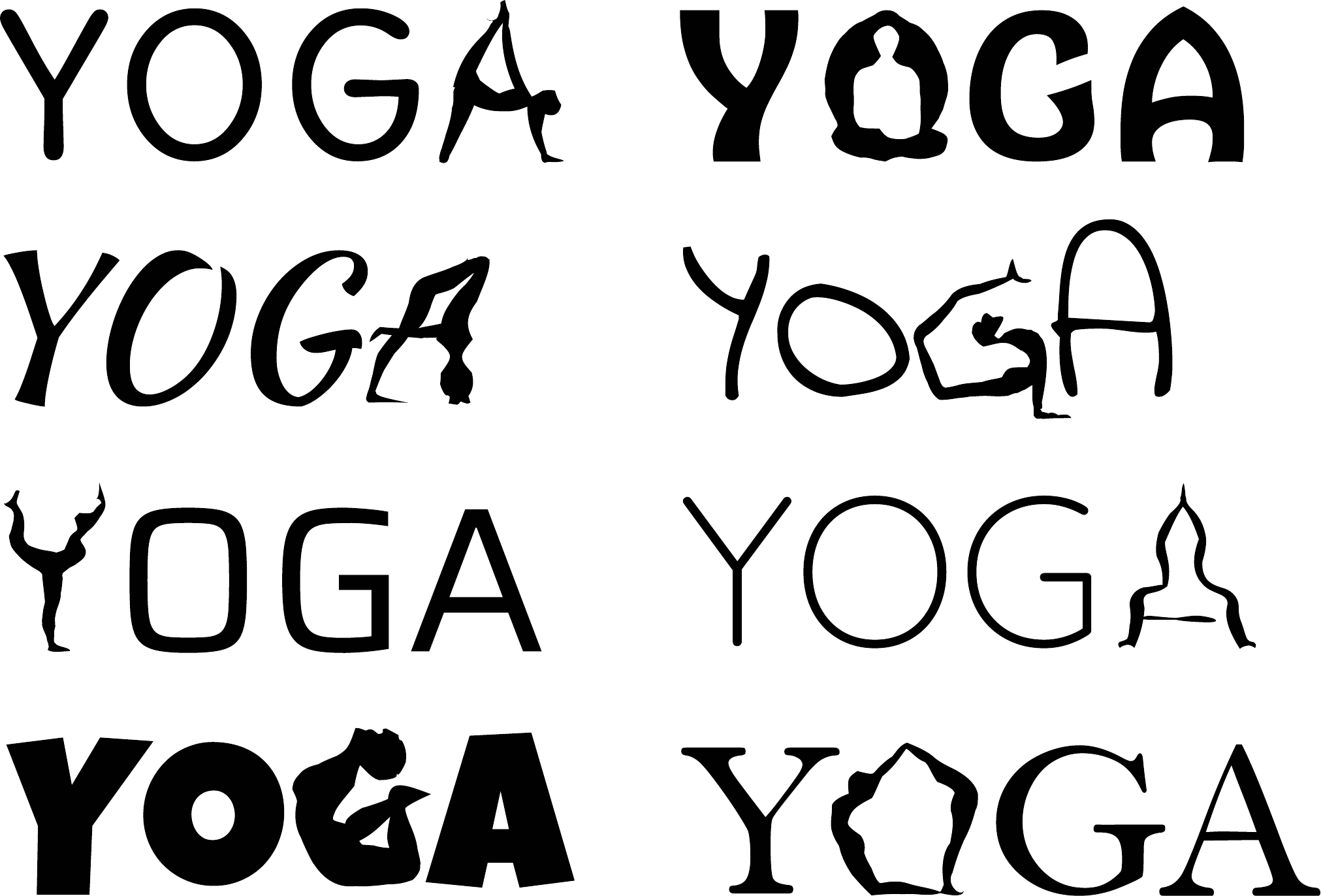}
    \caption{\small Word-as-images produced by our method for the word ``YOGA'', using eight different fonts.}
    \label{fig:many_fonts}
\end{figure}

\subsection{Weighting}
\label{sec:weighting}
Choosing the relative weights of the three losses presented above is crucial to the appearance of the final letter. 
While the $\nabla_{\hat{P}} \mathcal{L}_\text{LSDS}$ loss encourages the shape to deviate from its original appearance to better fit the semantic concept, the two terms $\mathcal{L}_{tone}$ and $\mathcal{L}_{acap}$ are responsible for maintaining the original shape.
Hence, we have two competing parts in the formula, and would like to find a balance between them to maintain the legibility of the letter while allowing the desired semantic shape to change.

We find that $\mathcal{L}_{tone}$ can be very dominant. In some cases, if it is used from the beginning, no semantic deformation is performed.
Therefore, we adjust the weight of $\mathcal{L}_{tone}$ to kick-in only after some semantic deformation has occurred. We define $\beta_t$ as follows:
\begin{equation}
    \beta_t = a \cdot \exp \big{(}-\frac{(t-b)^2}{2c^2}\big{)}
\end{equation}
with $a = 100, b = 300, c = 30$.
We analyse the affect of various weighting in Section \ref{sec:ablation}.
Note that the same hyper-parameter choice works for various words, letters, and fonts. 

\subsection{Implementation Details}
\label{sec:implementation}
Our method is based on the pre-trained $v1-5$ Stable Diffusion model \cite{rombach2022highresolution}, which we use through the diffusers \cite{von-platen-etal-2022-diffusers} Python package.
We optimize only the control points' coordinates (i.e. we do not modify the color, width, and other parameters of the shape). We use the Adam optimizer with $\beta_{1} = 0.9$, $\beta_{2} = 0.9$, $\epsilon = 10^{-6}$. We use learning rate warm-up from $0.1$ to $0.8$ over $100$ iterations and exponential decay from $0.8$ to $0.4$ over the rest $400$ iterations, $500$ iteration in total.
The optimization process requires at least 10GB memory and approximately 5 minutes to produce a single letter illustration on RTX2080 GPU.

Before we feed the rasterized $600x600$ letter image into the Stable Diffusion model, we apply random augmentations as proposed in CLIPDraw \cite{CLIPDraw}. Specifically, perspective transform with a distortion scale of $0.5$, with probability $0.7$, and a random $512x512$ crop. 
We add the suffix "a {[\textit{word}]}. minimal flat 2d vector. lineal color. trending on artstation." to the target word $W$, before feeding it into the text encoder of a pretrained CLIP model.

\begin{figure}
\centering
    \includegraphics[width=0.8\linewidth]{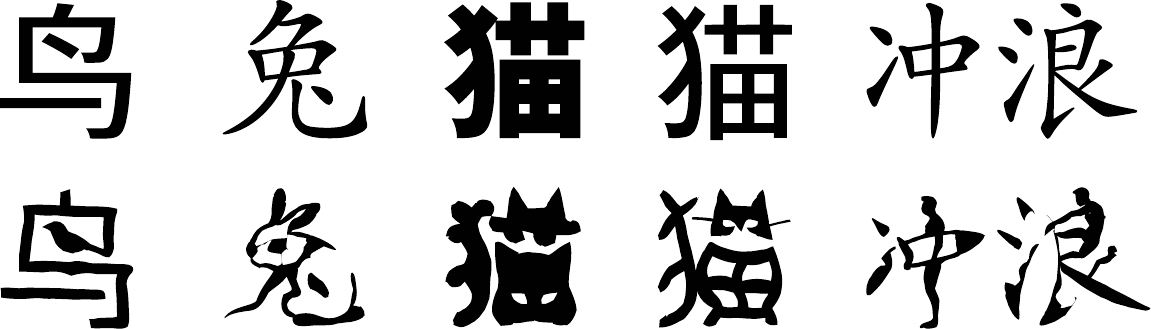}
    \caption{\small Additional examples of word-as-image applied on Chinese characters. In Chinese, a whole word can be represented by one character. From left: bird, rabbit, cat and surfing (two last characters together). The complexity of characters imposes an additional challenge for our method. This could be alleviated in the future, for example, by dividing the characters to radicals and applying the method only on parts of the character.}
    \label{fig:china}
\end{figure}

\begin{figure*}
    \centering
    \setlength{\tabcolsep}{1.5pt}
    {\small
    \begin{tabular}{l@{\hspace{0.2cm}} c@{\hspace{0.2cm}} | @{\hspace{0.2cm}}c c@{\hspace{0.2cm}} | c c @{\hspace{0.2cm}} | c @{\hspace{0.2cm}} | @{\hspace{0.2cm}}c l}

        \raisebox{0.5cm}{\makecell[l]{BIRD, \\ letter R}} &
        \hspace{0.1cm}
        \raisebox{0.25cm}{\includegraphics[height=0.042\textwidth]{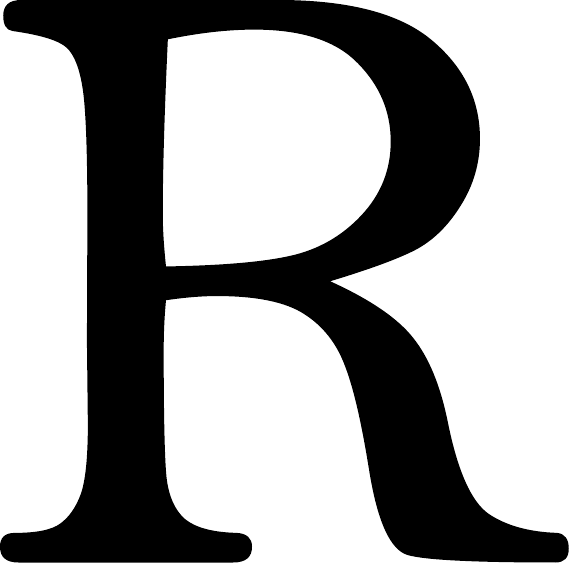}} &
        \hspace{0.1cm}
        \includegraphics[height=0.07\textwidth]{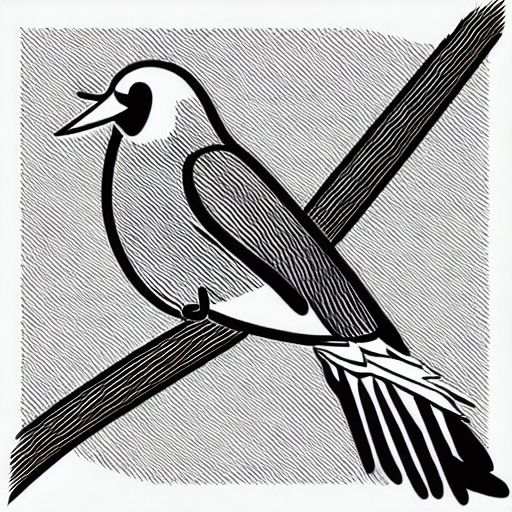} &
        \includegraphics[height=0.07\textwidth]{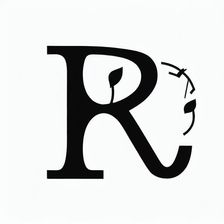} &
        \hspace{0.1cm}
        \includegraphics[height=0.07\textwidth]{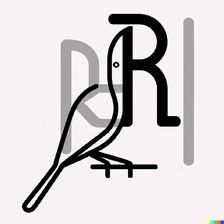} &
        \includegraphics[height=0.07\textwidth]{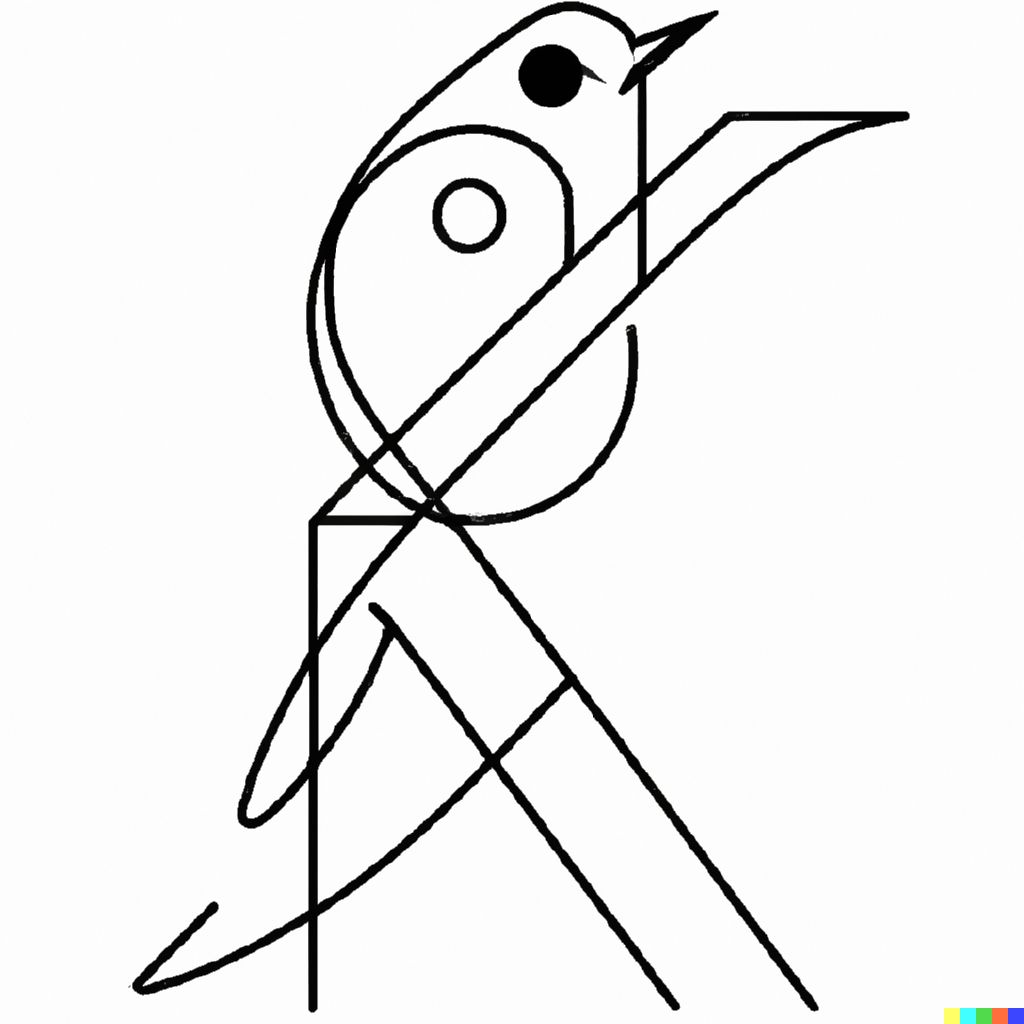} &
        \hspace{0.1cm}
        \includegraphics[height=0.07\textwidth]{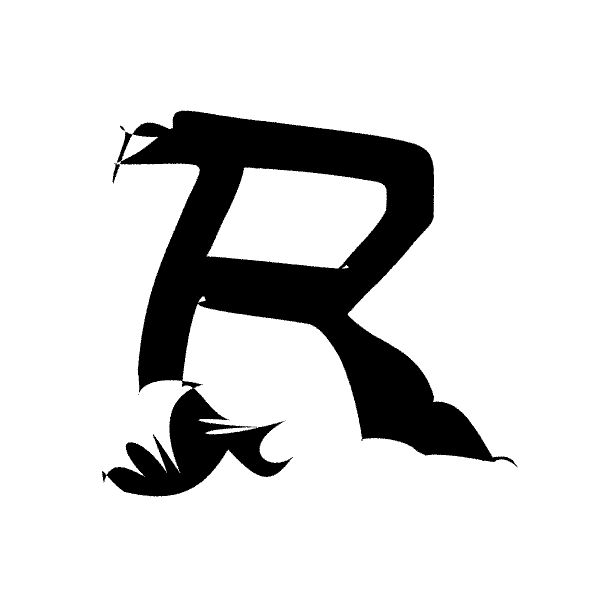} &
        \raisebox{0.23cm}{\includegraphics[height=0.048\textwidth]{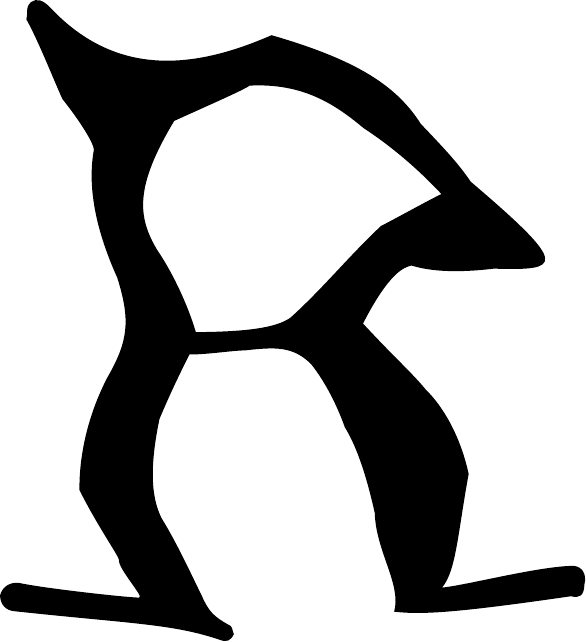}} &
        \hspace{0.1cm}
        \raisebox{0.23cm}{\includegraphics[height=0.048\textwidth]{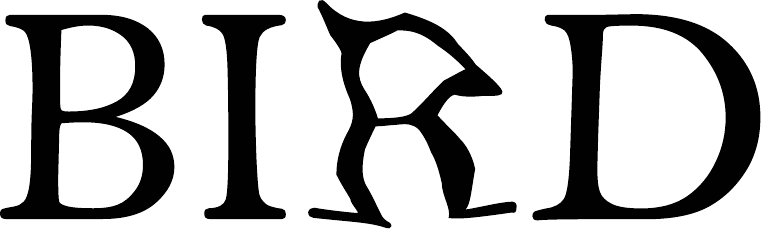}} \\

        \raisebox{0.5cm}{\makecell[l]{DRESS, \\ letter E}} &
        \hspace{0.1cm}
        \raisebox{0.25cm}{\includegraphics[height=0.042\textwidth]{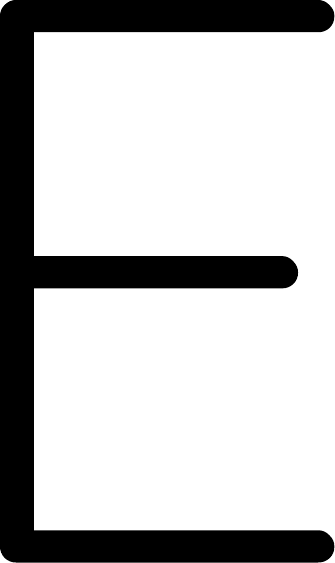}} &
        \hspace{0.1cm}
        \includegraphics[height=0.07\textwidth]{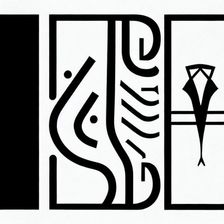} &
        \includegraphics[height=0.07\textwidth]{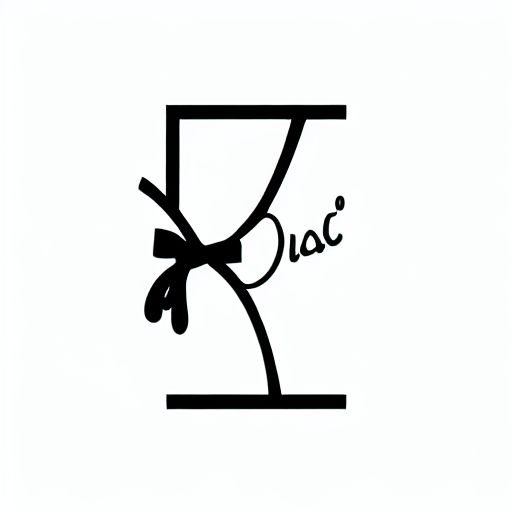} &
        \hspace{0.1cm}
        \includegraphics[height=0.07\textwidth]{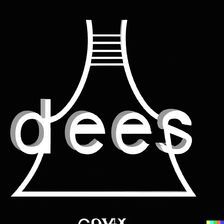} &
        \includegraphics[height=0.07\textwidth]{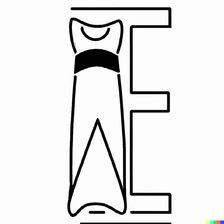} &
        \hspace{0.1cm}
        \includegraphics[height=0.07\textwidth]{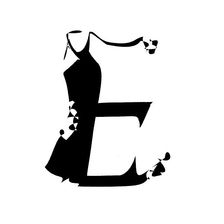} &
        \raisebox{0.2cm}{\includegraphics[height=0.048\textwidth]{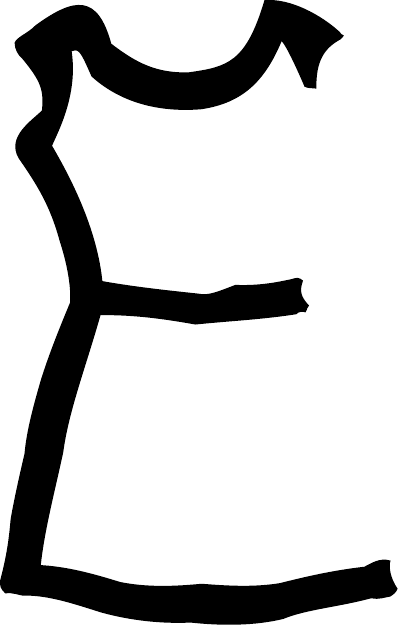}} &
        \hspace{0.1cm}
        \raisebox{0.2cm}{\includegraphics[height=0.048\textwidth]{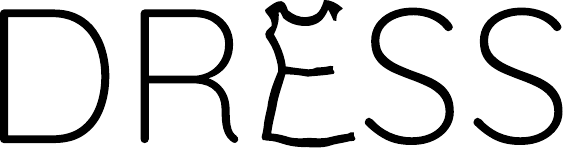}} \\

        \raisebox{0.5cm}{\makecell[l]{TULIP, \\ letter U}} &
        \hspace{0.1cm}\includegraphics[height=0.06\textwidth]{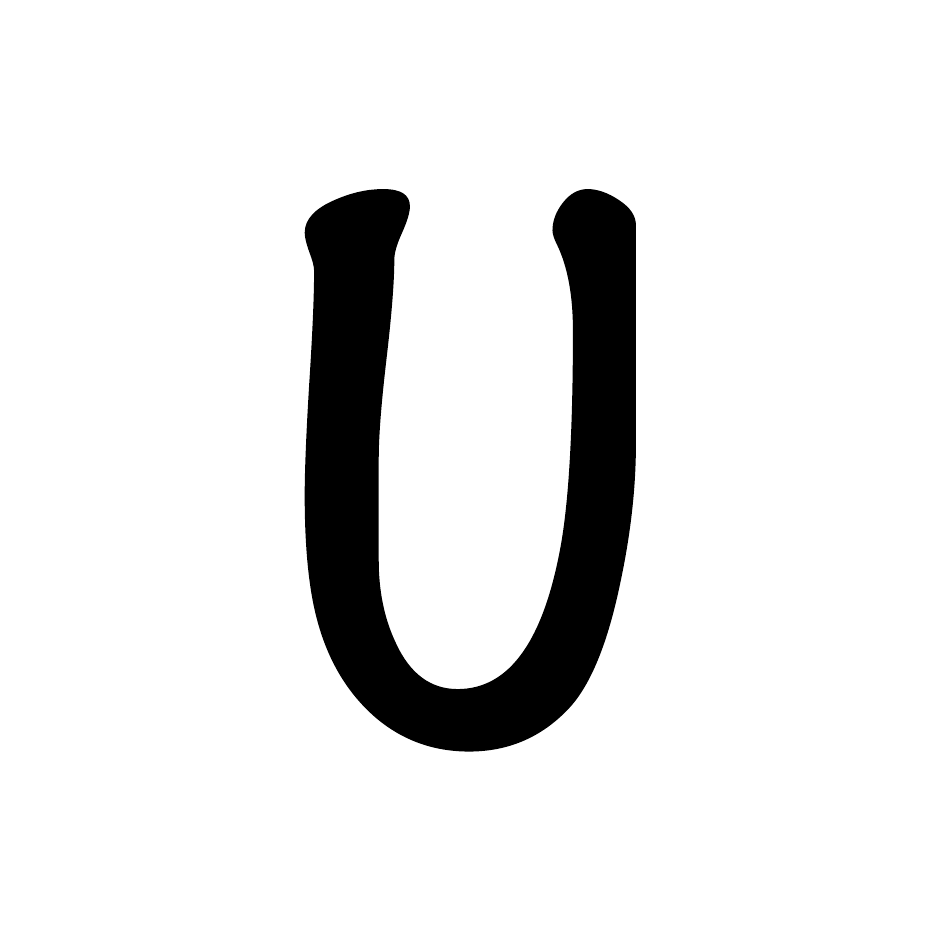} &
        \hspace{0.1cm}
        \includegraphics[height=0.07\textwidth]{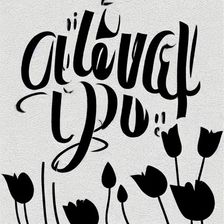} &
        \includegraphics[height=0.07\textwidth]{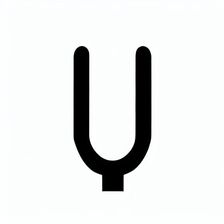} &
        \hspace{0.1cm}
        \includegraphics[height=0.07\textwidth]{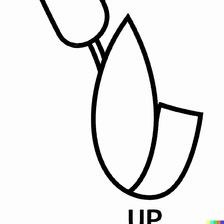} &
        \includegraphics[height=0.07\textwidth]{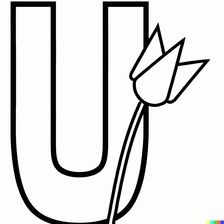} &
        \hspace{0.1cm}
        \includegraphics[height=0.07\textwidth]{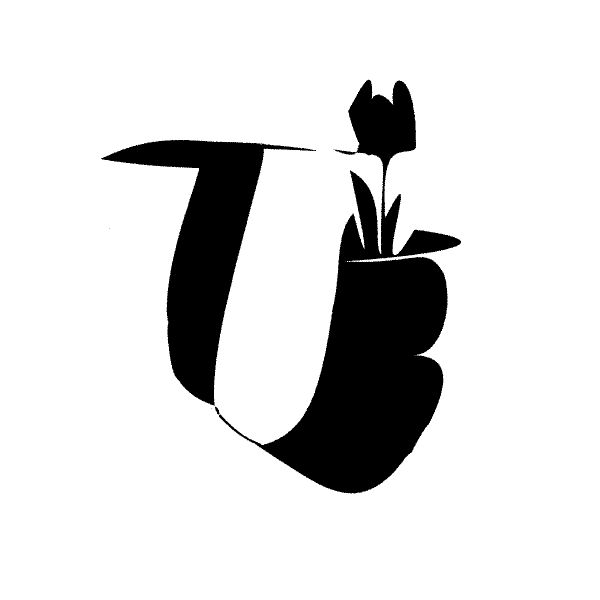} &
        \raisebox{0.16cm}{\includegraphics[height=0.048\textwidth]{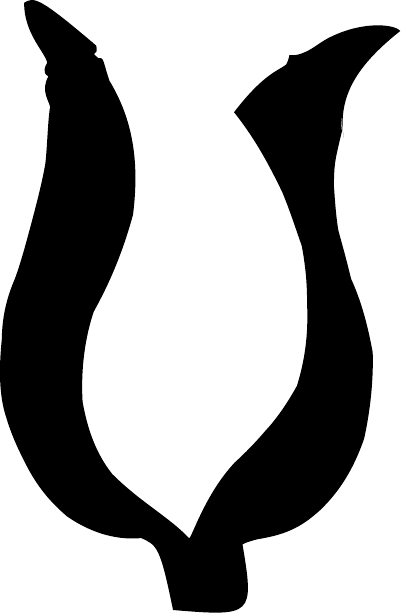}} &
        \hspace{0.1cm}
        \raisebox{0.16cm}{\includegraphics[height=0.048\textwidth]{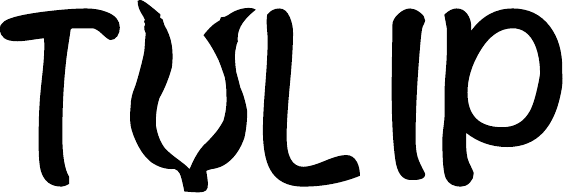}} \\

        \multicolumn{2}{c@{\hspace{0.2cm}}}{Input}&\hspace{0.1cm} SD & SDEdit &\hspace{0.1cm} DallE2 & \begin{tabular}[c]{@{}c@{}} DallE2 \\  +letter \end{tabular} &\hspace{0.1cm} CLIPDraw & \multicolumn{2}{c}{Ours}

    \end{tabular}
    }
    \caption{\small Comparison to alternative methods based on large scale text-to-image models. On the left are the letters used as input (only for SDEdit, CLIPDraw, and ours), as well as the desired object of interest. The results from left to right obtained using Stable Diffusion \cite{rombach2022highresolution}, SDEdit \cite{meng2022sdedit}, DallE2 \cite{ramesh2022hierarchical}, DallE2 with a letter specific prompt, CLIPDraw \cite{CLIPDraw}, and our single-letter results, as well as the final word-as-image.}
    \label{fig:comp_diffusion}
\end{figure*}

\section{Results}
The robustness of our approach means it should be capable of handling a wide range of input concepts as well as supporting different font designs.
Figures \ref{fig:teaser_wordasim}, \ref{fig:intro_res}, \ref{fig:all_res1}, \ref{fig:mayne_res}, and more results in the supplemental file demonstrate that our approach can handle inputs from many different categories and various fonts, and that the generated results are legible and creative.
Figure~\ref{fig:many_fonts} demonstrate how the illustrations created by our method for the same word follow the characteristics of different fonts, while Figure~\ref{fig:china} shows examples in a different language.
Although the perceived aesthetics of a word-as-image illustration can be subjective, we define three objectives for an effective result: (1) it should visually capture the given semantic concept, (2) it should maintain readability, and (3) it should preserve the original font's characteristics.

We evaluate the performance of our method on a randomly selected set of inputs.
We select five common concept classes - animals, fruits, plants, sports, and professions. 
Using ChatGPT, we sample ten random instances for each class, resulting in 50 words in total.
Next, we select four fonts that have distinct visual characteristics, namely Quicksand, Bell MT, Noteworthy-Bold, and HobeauxRococeaux-Sherman.
For each word, we randomly sampled one of the four fonts, and applied our method to each letter.
For each word with $n$ letters we can generate $2^n$ possible word-as-images, which are all possible combinations of replacements of illustrated letters.
A selected subset of these results is presented in Figure \ref{fig:all_res1}.
The results of all letters and words are presented in the supplementary material.

As can be seen, the resulting word-as-image illustrations successfully convey the given semantic concept in most cases while still remaining legible.
In addition, our method successfully captures the font characteristics. For example, in Figure \ref{fig:all_res1}, the replacements for the ``DRESS'' and ``LION'' are thin and fit well with the rest of the word. In addition, observe the serifs of the letter A used for the fin of the shark in the ``SHARK'' example. We further use human evaluation to validate this as described below.

\subsection{Quantitative}
\label{subsec:quant}
We conduct a perceptual study to quantitatively assess the three objectives of our resulting word-as-images. 
We randomly select two instances from each of the resulting word-as-image illustrations for the five classes described above, and visually select one letter from each word, resulting in 10 letters in total. 
In each question we show an isolated letter illustration, without the context of the word.
To evaluate the ability of our method to visually depict the desired concept, we present four label options from the same class, and ask participants to choose the one that describes the letter illustration best.
To evaluate the legibility of the results, we ask participants to choose the most suitable letter from a random list of four letters.
To asses the preservation of the font style, we present the four fonts and ask participants to choose the most suitable font for the illustration.
We gathered answers from 40 participants, and the results are shown in Table \ref{tb:user_study}. As can be seen, the level of concept recognizability and letter legibility are very high, and the $51\%$ of style matching of the letter illustration to the original font is well above random, which is $25\%$. We also test our algorithm without the two additional structure and style preserving losses ($\mathcal{L}_{acap}$ and $\mathcal{L}_{tone}$) on the same words and letters (``Only SDS'' in the table). As expected, without the additional constraints, the letter deforms significantly resulting in higher concept recognizability but lower legibility and font style preservation. More details and examples are provided in the supplementary material.

\begin{table}
    \small
    \centering
    \setlength{\tabcolsep}{2pt}
    \caption{\small Perceptual study results - average level of accuracy (and standard deviation in parentheses). The level of concept recognizability and letter legibility are very high, and style matching of the font is well above random. The ``Only SDS'' results are created by removing our structure and style preserving losses.} 
    \begin{tabular}{l c c c} 
    \toprule
    Method & \begin{tabular}{c} Semantics \end{tabular} & \begin{tabular}{c} Legibility \end{tabular} & \begin{tabular}{c} Font \end{tabular} \\
    \midrule
    Ours    & 0.8 (0.191) & 0.9 (0.170) & 0.51 (0.241) \\
    Only SDS  & 0.88 (0.151) & 0.53 (0.336) & 0.33 (0.211) \\
    \bottomrule
    \end{tabular}
    \vspace{-0.2cm}
    \label{tb:user_study}
\end{table}
 
\subsection{Comparison}
\label{subsec:comparisons}
In the absence of a relevant baseline for comparison, we define baselines based on large popular text-to-image models. Specifically, we use \textbf{(1) SD} Stable Diffusion \cite{rombach2022highresolution}, \textbf{(2) SDEdit} \cite{meng2022sdedit}, \textbf{(3) DallE2} \cite{ramesh2022hierarchical} illustrating the word, \textbf{(4) DallE2+letter} illustrating only the letter, and \textbf{(5) CLIPDraw} \cite{CLIPDraw}. We applied the methods above (details can be found in supplemental material) to three representative words -- ``bird'', ``dress'', and ``tulip'', with the fonts Bell MT, Quicksand, and Noteworthy-Bold, respectively. The results can be seen in Figure \ref{fig:comp_diffusion}.

In some cases Stable Diffusion (SD) did not manage to produce text at all (such as for the bird) and when text is produced, it is often not legible.
The results obtained by SDEdit preserve the font's characteristics and the letter's legibility, but often fail to reflect the desired concept, such as in the case of the bird and the dress. Additionally, it operates in the raster domain and tends to \textit{add} details on top of the letter, while our method operates directly on the vector representation of the letters with the objective of modifying their \textit{shape}.
DallE2 manages to reflect the visual concept, however it often fails to produce legible text. When applied with a dedicated prompt to produce the word-as-image of only one letter (fifth column), it manages to produce a legible letter, but there is less control over the output -- it is impossible to specify the desired font or to control the size, position, and shape of the generated letter. Therefore, it is not clear how to combine these output illustrations into the entire word to create a word-as-image.

CLIPDraw produces reasonable results conveying the semantics of the input word. However, the results are non-smooth and the characteristics of the font are not preserved (for example observe how the letter "E" differs from the input letter). We further examine CLIPDraw with our shape preservation losses in the next Section.

\begin{figure}[h]
    \centering
    \setlength{\tabcolsep}{5pt}
    \renewcommand{\arraystretch}{1.5} 
    \begin{tabular}{c c | c c c}
        \raisebox{0.4cm}{"Ballet"} &
        \includegraphics[height=0.1\linewidth]{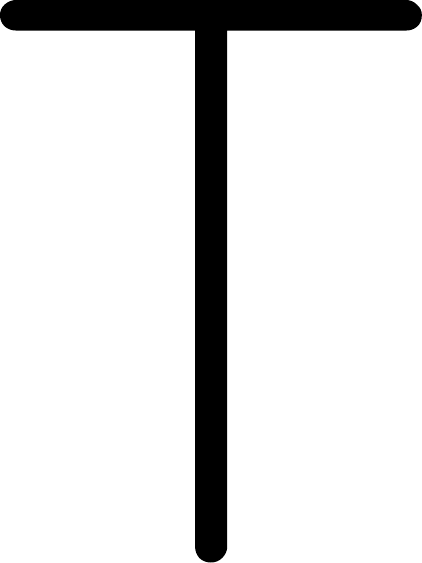} &
        \hspace{0.1cm}
        \includegraphics[height=0.1\linewidth]{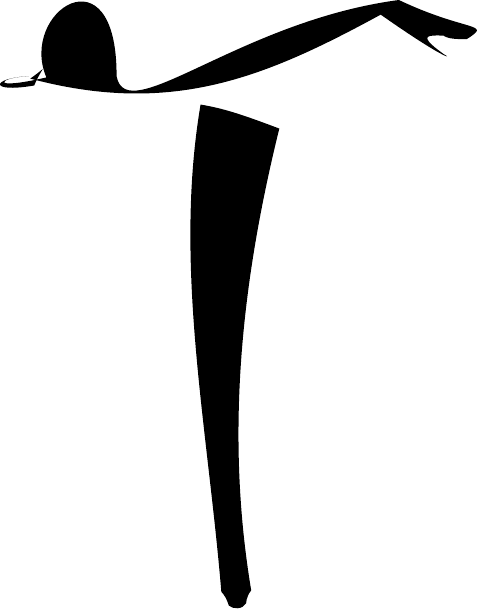} &
        \includegraphics[height=0.1\linewidth]{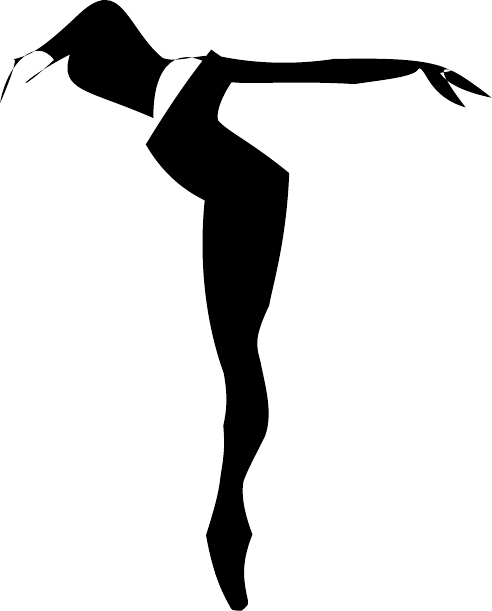} &
        \includegraphics[height=0.1\linewidth]{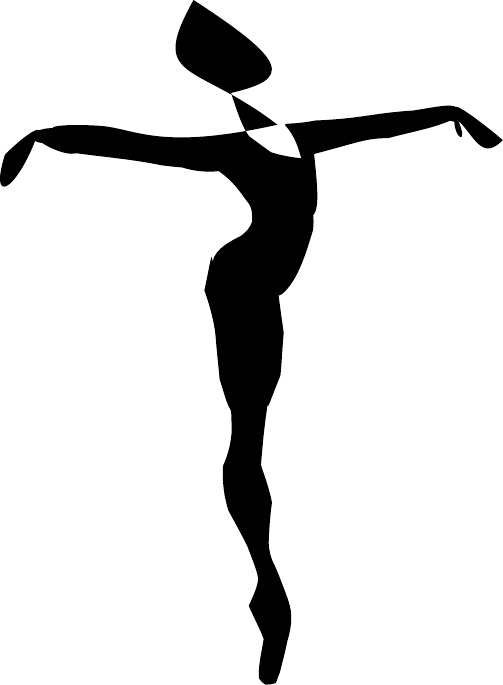} \\
        
        \raisebox{0.4cm}{"Gorilla"} &
        \includegraphics[height=0.1\linewidth]{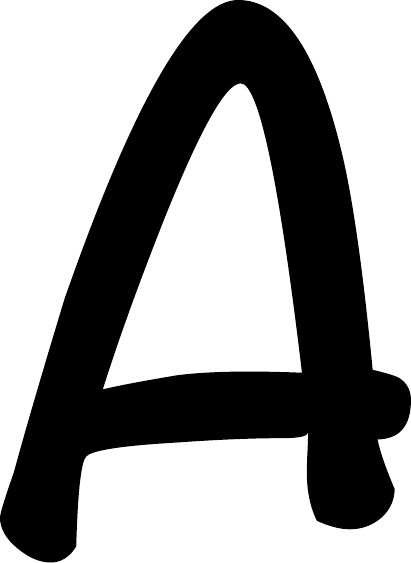} &
        \hspace{0.1cm}
        \includegraphics[height=0.1\linewidth]{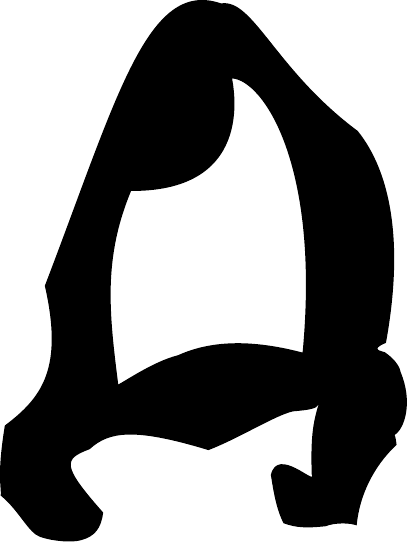} &
        \includegraphics[height=0.1\linewidth]{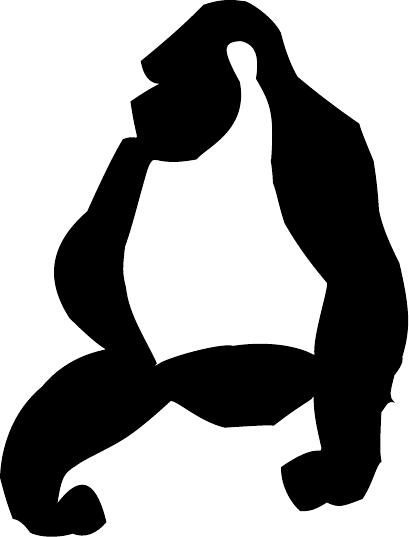} &
        \includegraphics[height=0.1\linewidth]{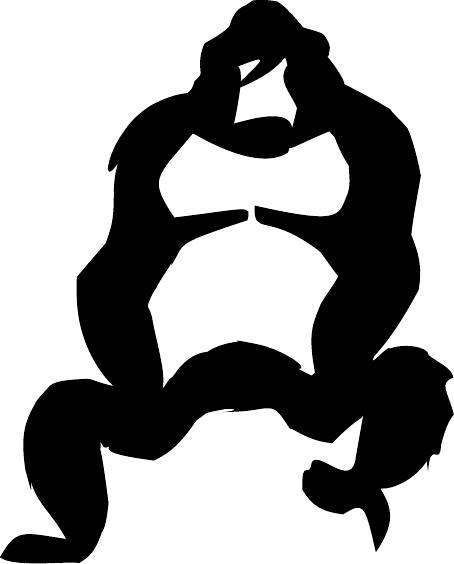} \\

        \raisebox{0.4cm}{"Gym"} &
        \includegraphics[height=0.1\linewidth]{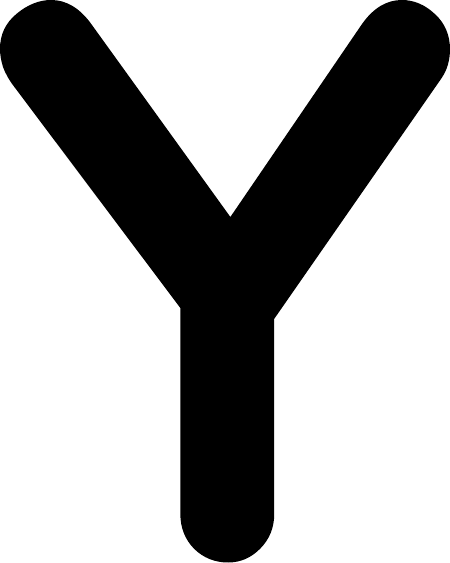} &
        \hspace{0.1cm}
        \includegraphics[height=0.1\linewidth]{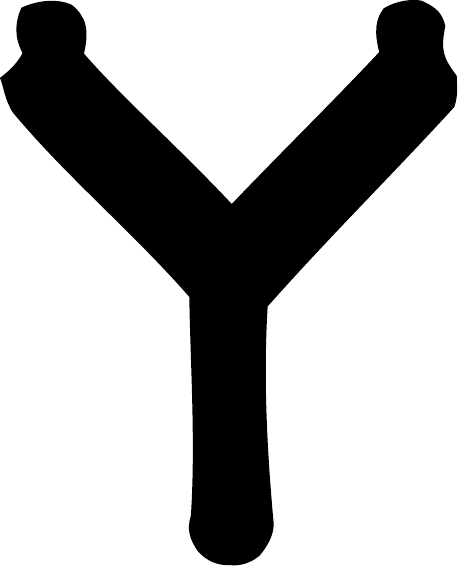} &
        \includegraphics[height=0.1\linewidth]{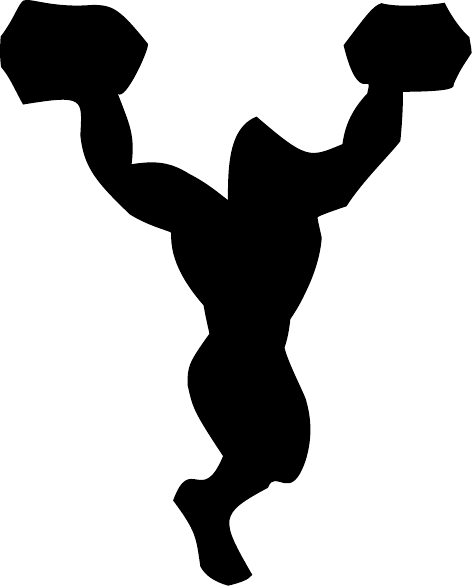} &
        \includegraphics[height=0.1\linewidth]{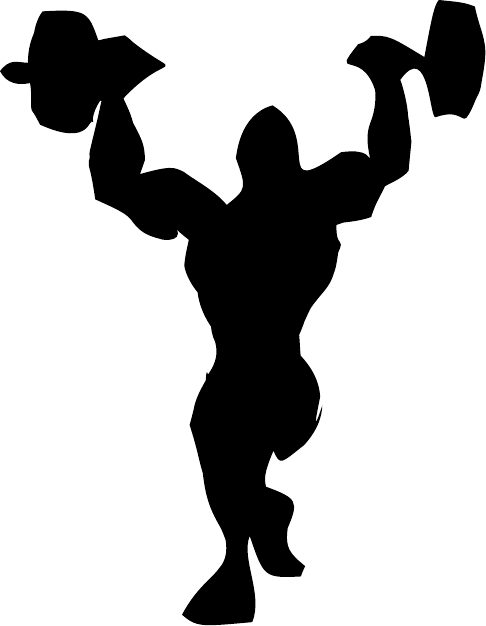} \\
        
        \multicolumn{2}{c}{Input} & $P_o$ & $P$ & $2\times{P}$\\
    \end{tabular}
    \vspace{-0.2cm}
    \caption{\small The effect of the initial number of control points on outputs. On the left are the input letters and the target concepts used to generate the results on the right. $P_o$ indicates the original number of control points as extracted from the font, $P$ is the input letter with our chosen hyperparameters, and for $2\times{P}$ we increase the number of control points in $P$ by two.}
    \label{fig:cc_effect}
\end{figure}

\begin{figure}
\centering
    \setlength{\tabcolsep}{5pt}
    \renewcommand{\arraystretch}{1}
    \begin{tabular}{c c c c c c}
        
        \raisebox{0.4cm}{\makecell[l]{Input \\ Letter}} & 
        \includegraphics[height=0.07\linewidth]{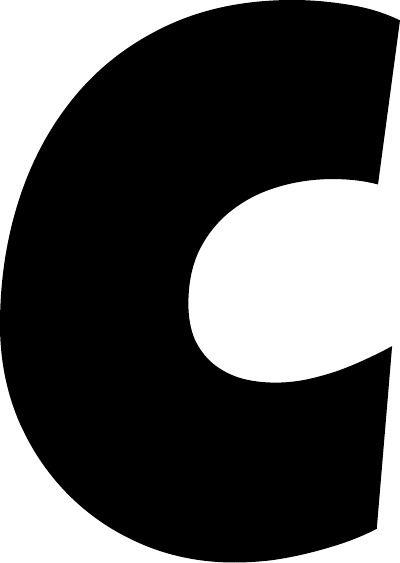} &
        \includegraphics[height=0.07\linewidth]{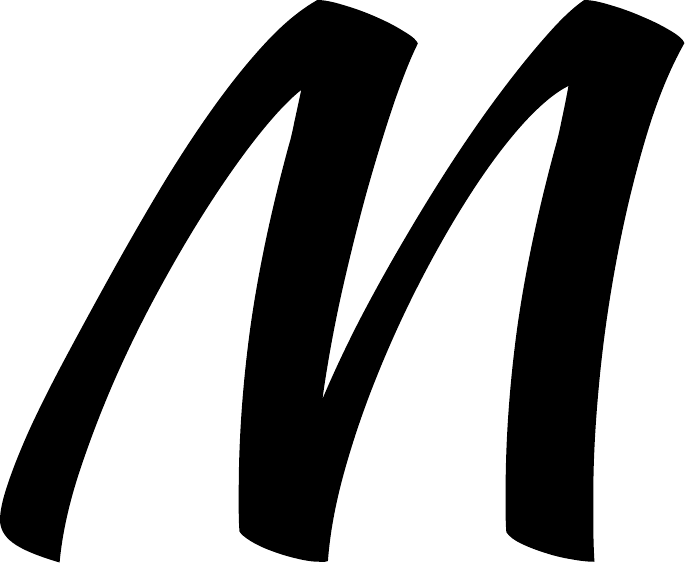} &
        \includegraphics[height=0.07\linewidth]{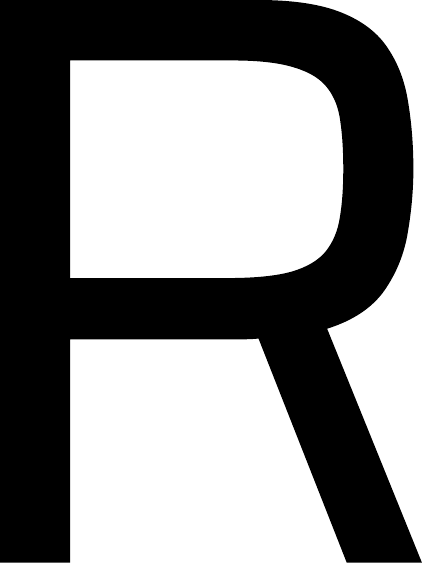} &
        \includegraphics[height=0.07\linewidth]{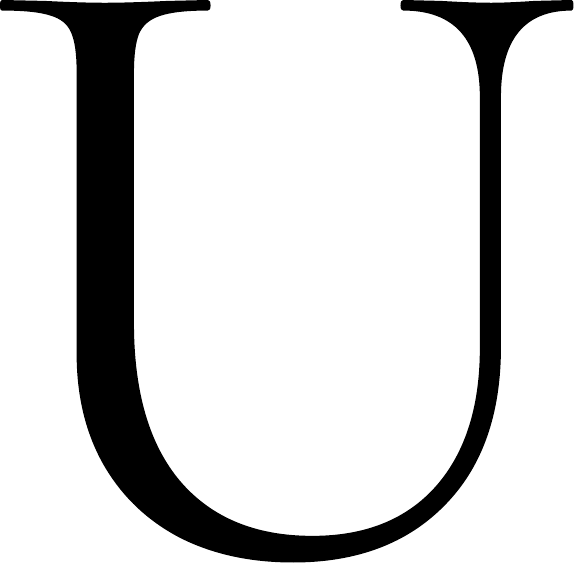} &
        \includegraphics[height=0.08\linewidth]{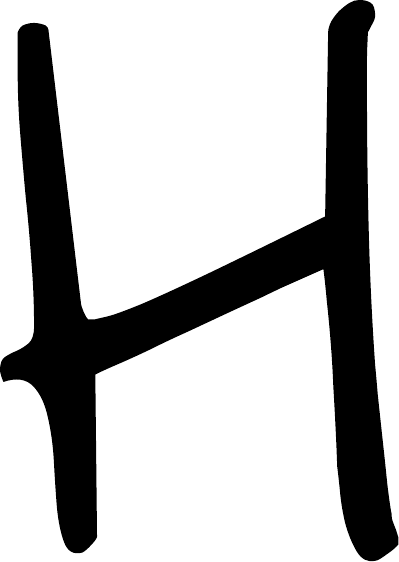} \\
        \midrule
        \raisebox{0.4cm}{Ours} & 
        \includegraphics[height=0.08\linewidth]{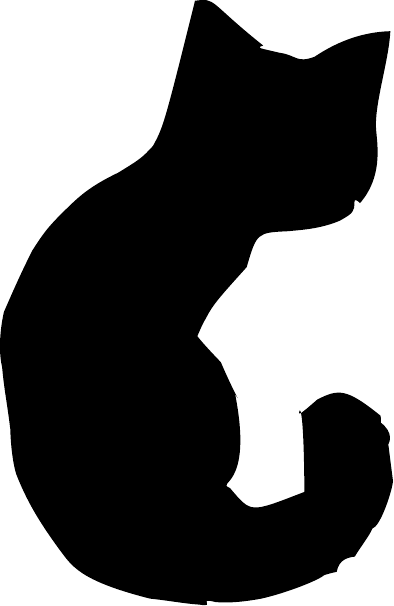} & 
        \includegraphics[height=0.08\linewidth]{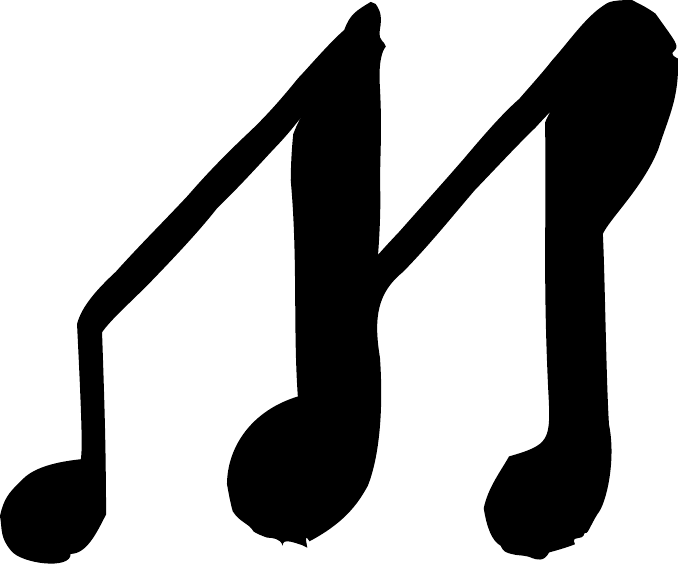} &
        \includegraphics[height=0.08\linewidth]{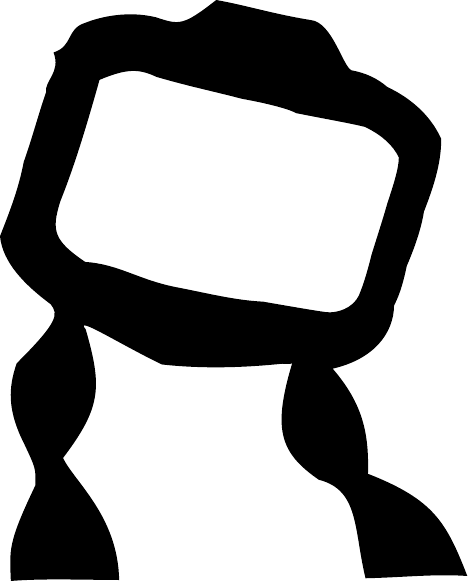} &
        \includegraphics[height=0.08\linewidth]{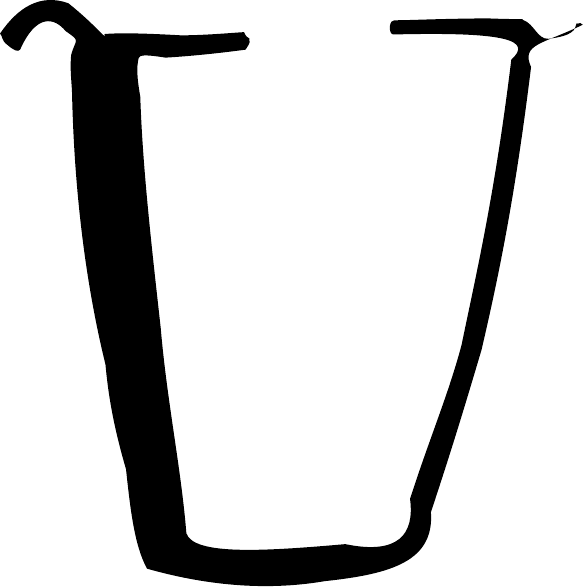} &
        \includegraphics[height=0.08\linewidth]{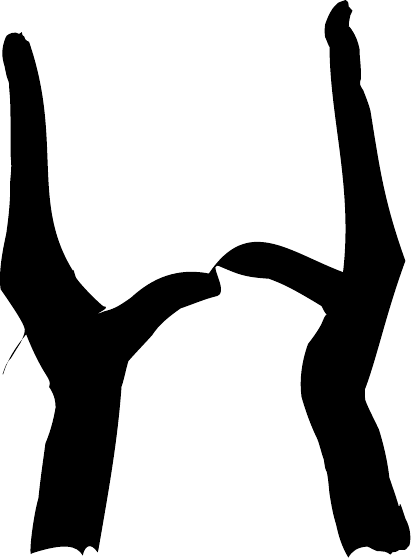} \\
        \raisebox{0.4cm}{\makecell[l]{Only \\ SDS}}  & 
        \includegraphics[height=0.08\linewidth]{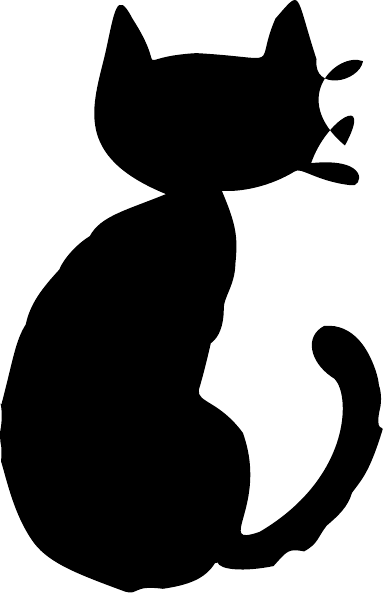} & 
        \includegraphics[height=0.08\linewidth]{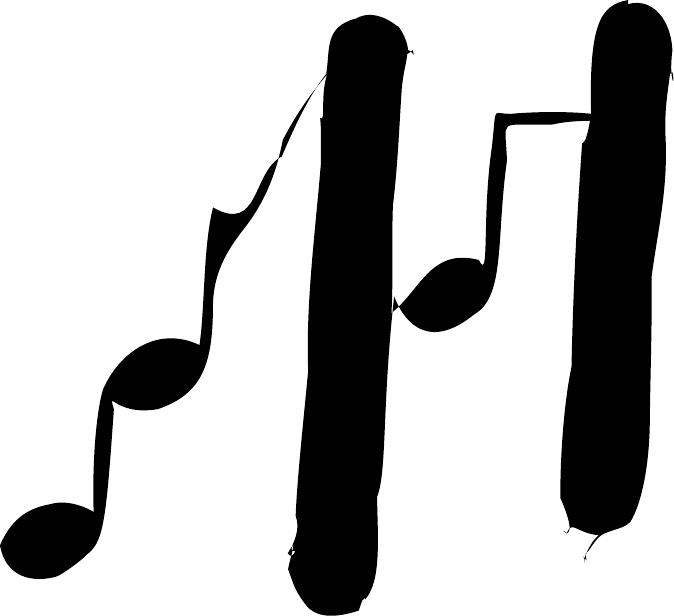} & 
        \includegraphics[height=0.08\linewidth]{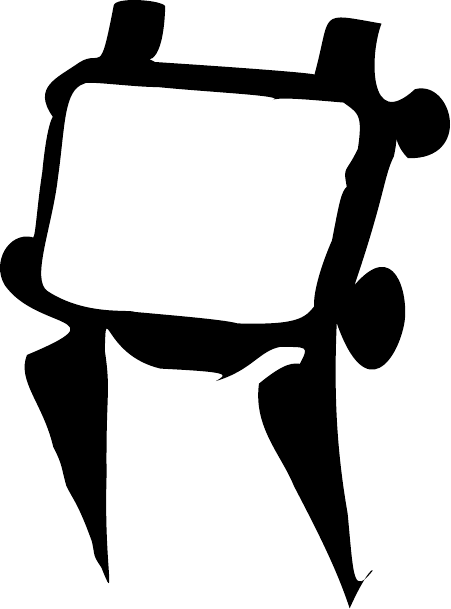} &
        \includegraphics[height=0.08\linewidth]{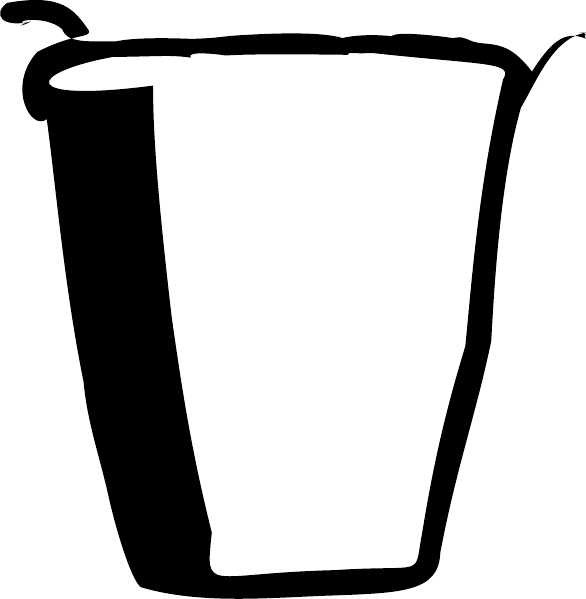} &
        \includegraphics[height=0.08\linewidth]{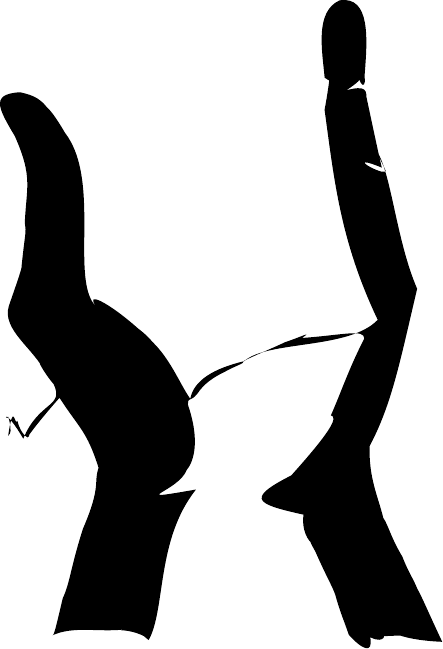} \\
        & "Cat" & "Music" & "Robot" & "Cup" & "Hands" \\
    \end{tabular}
    \caption{\small The effect of using only the SDS loss: note how the third row simply looks like icon illustrations, while the second row still resembles legible letters. }
    \label{fig:sds_lr}
\end{figure}

\begin{figure}
\centering
\setlength{\tabcolsep}{5pt}
\renewcommand{\arraystretch}{1} 
\begin{tabular}{c c | c c c c c}
    \raisebox{0.4cm}{"Bear"} &
    \raisebox{0.01cm}{\includegraphics[height=0.075\linewidth]{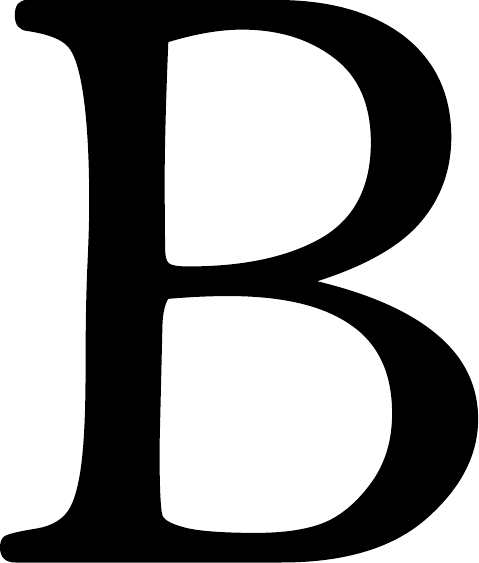}} &
    \hspace{0.2cm}
    \includegraphics[height=0.08\linewidth]{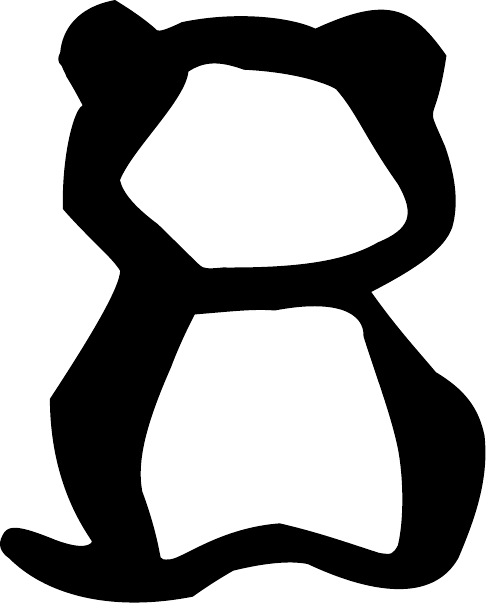} &
    \includegraphics[height=0.08\linewidth]{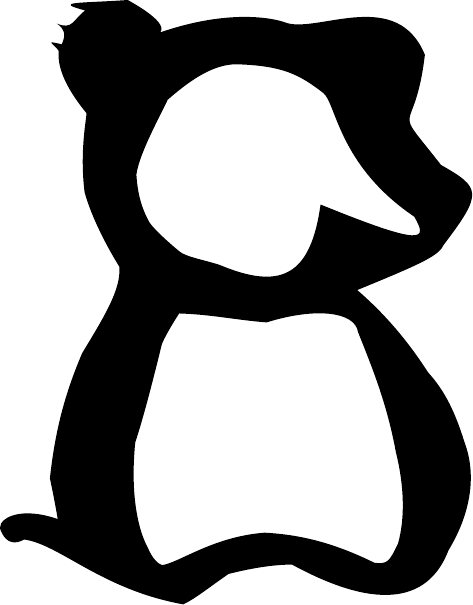} &
    \includegraphics[height=0.08\linewidth]{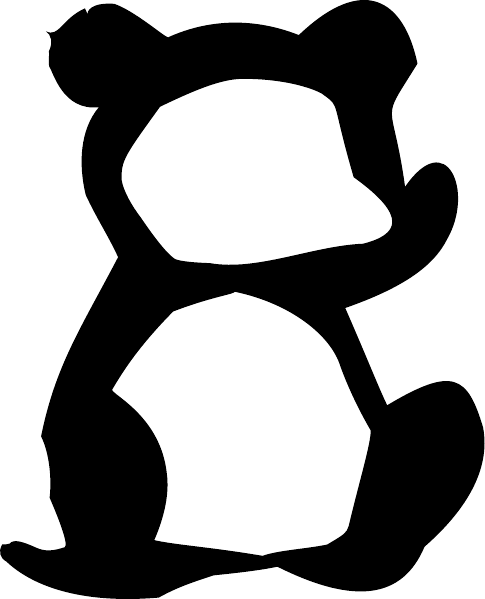} &
    \includegraphics[height=0.08\linewidth]{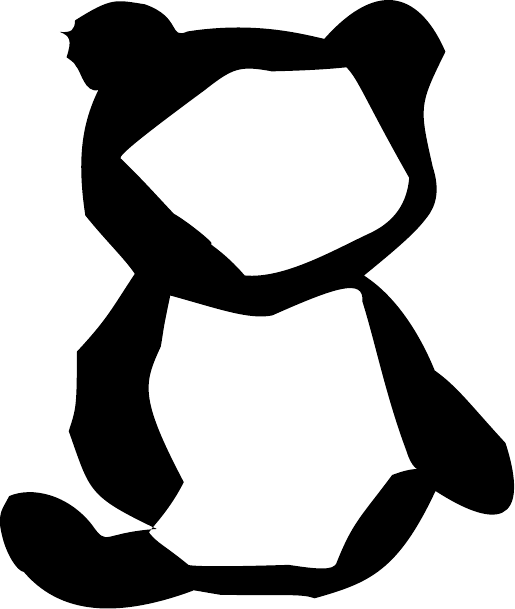} &
    \includegraphics[height=0.08\linewidth]{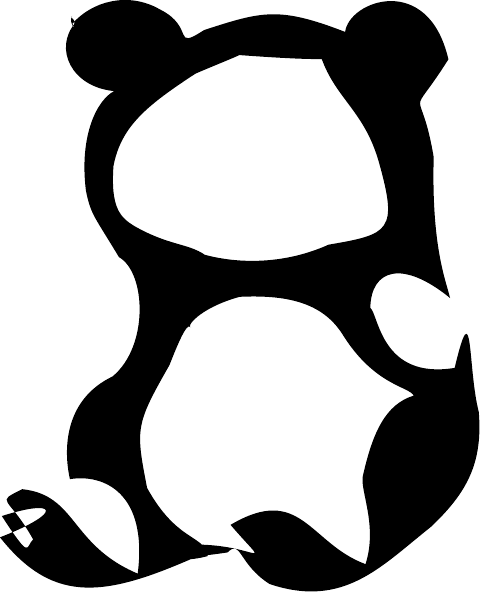} \\

    \raisebox{0.4cm}{"Singer"} &
    \includegraphics[height=0.08\linewidth]{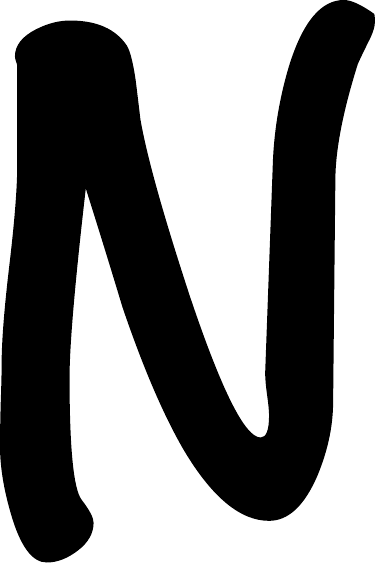} &
    \hspace{0.2cm}
    \includegraphics[height=0.08\linewidth]{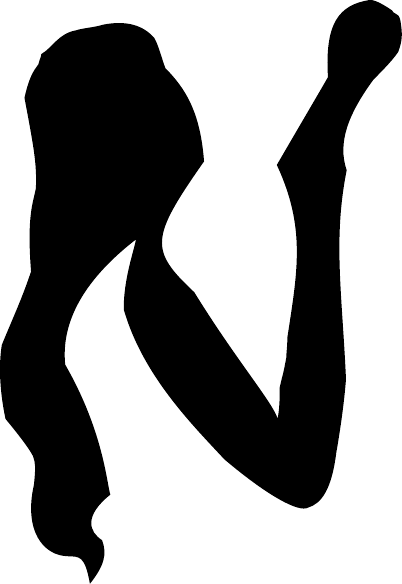} &
    \includegraphics[height=0.08\linewidth]{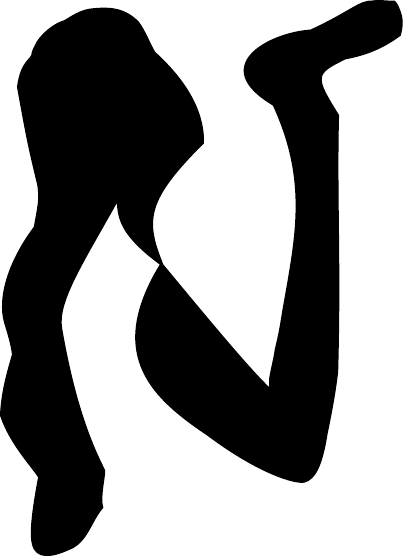} &
    \includegraphics[height=0.08\linewidth]{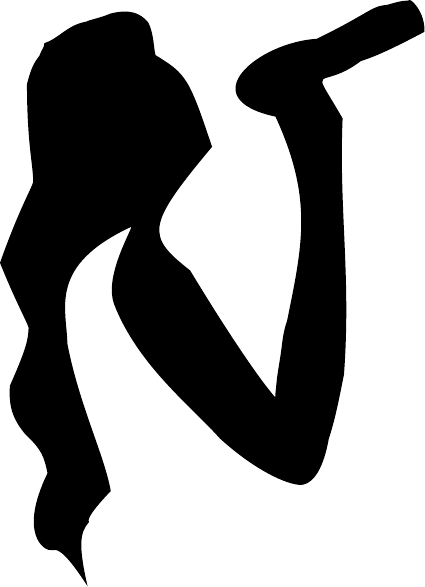} &
    \includegraphics[height=0.08\linewidth]{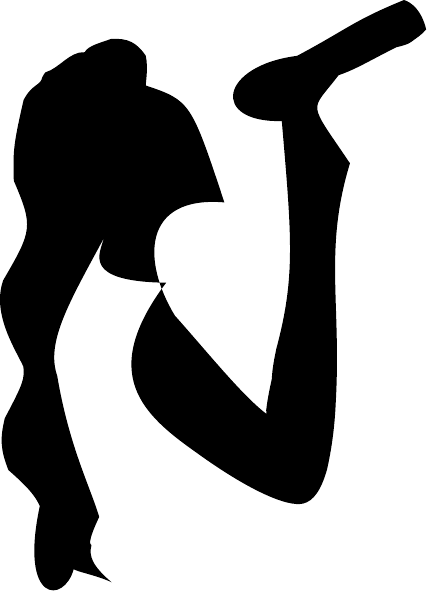} &
    \includegraphics[height=0.08\linewidth]{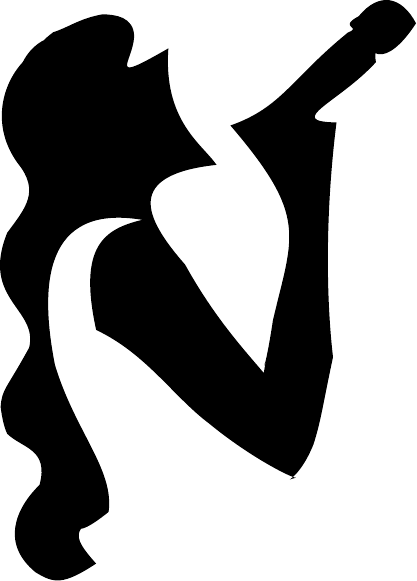} \\

    \raisebox{0.4cm}{"Giraffe"} &
    \includegraphics[height=0.075\linewidth]{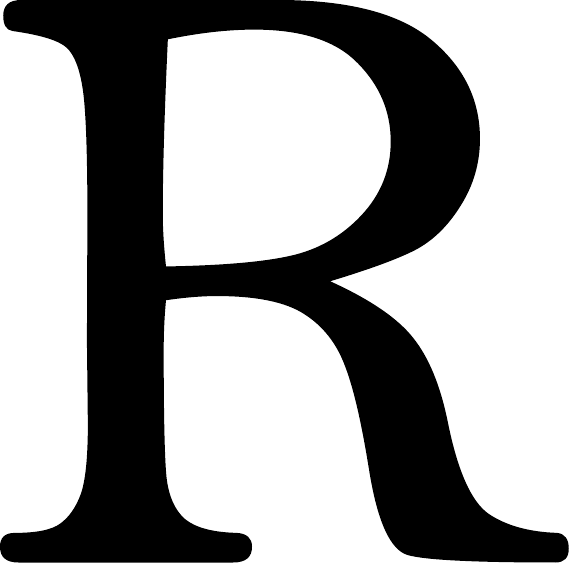} &
    \hspace{0.2cm}
    \includegraphics[height=0.08\linewidth]{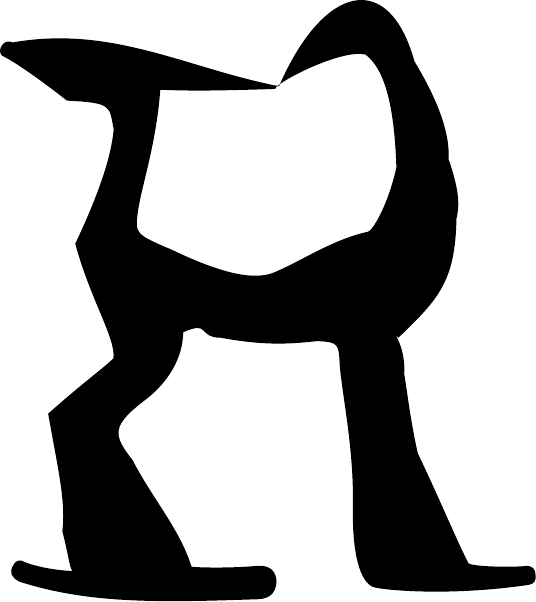} &
    \includegraphics[height=0.08\linewidth]{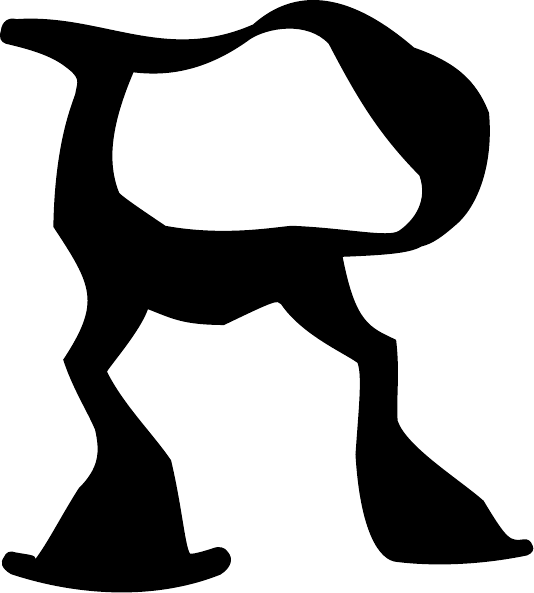} &
    \includegraphics[height=0.08\linewidth]{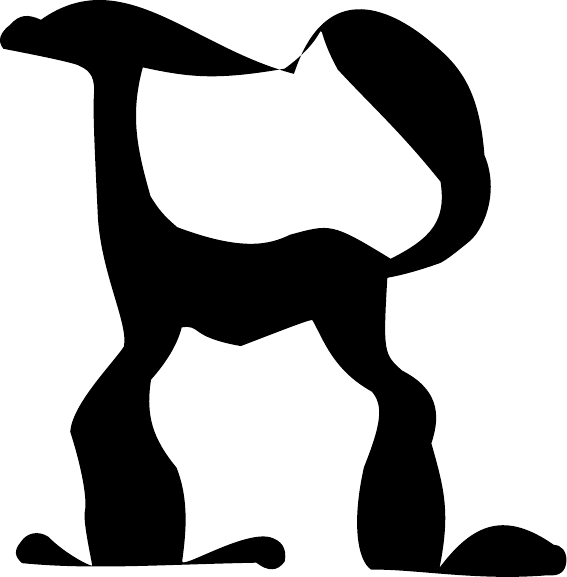} &
    \includegraphics[height=0.08\linewidth]{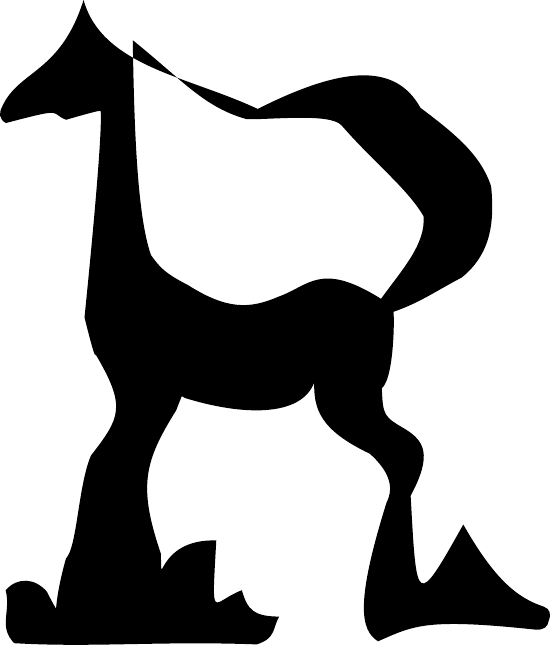} &
    \includegraphics[height=0.08\linewidth]{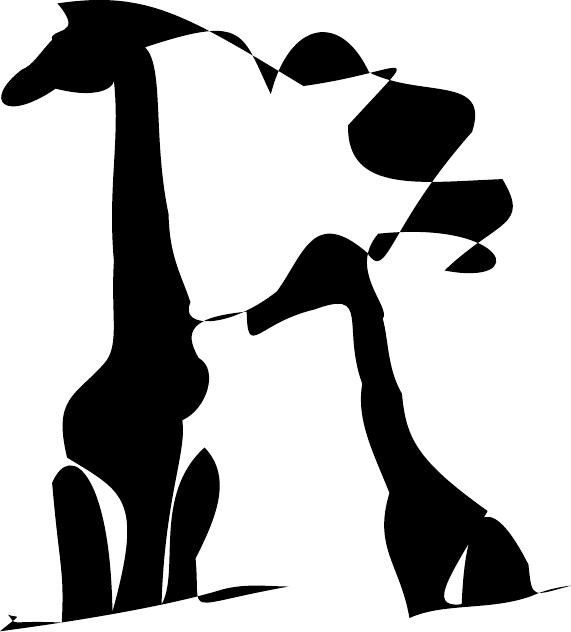} \\
    \multicolumn{2}{c}{Input} & 1 & 0.75 & 0.5 & 0.25 & \makecell[c]{Without \\ $\mathcal{L}_{acap}$} \\ 
\end{tabular}
\caption{\small Altering the weight $\alpha$ of the $\mathcal{L}_{acap}$ loss. On the leftmost column are the original letters and concepts used, then from left to right are the results obtained when using $\alpha\in\{1, 0.75, 0.5, 0.25, 0\}$.}
\label{fig:weights_conformal}
\end{figure}

\begin{figure}[h]
\centering
\setlength{\tabcolsep}{5pt}
\renewcommand{\arraystretch}{1} 
\begin{tabular}{c c | c c c c c}
    \raisebox{0.4cm}{"Bear"} &
    \raisebox{0.01cm}{\includegraphics[height=0.09\linewidth]{word-as-image/images/weight_effect/Bell_MT_B_scaled.pdf}} &
    \hspace{0.2cm}
    \includegraphics[height=0.1\linewidth]{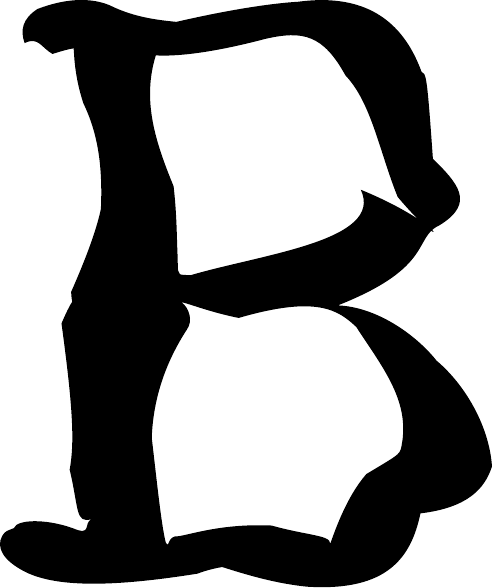} &
    \includegraphics[height=0.1\linewidth]{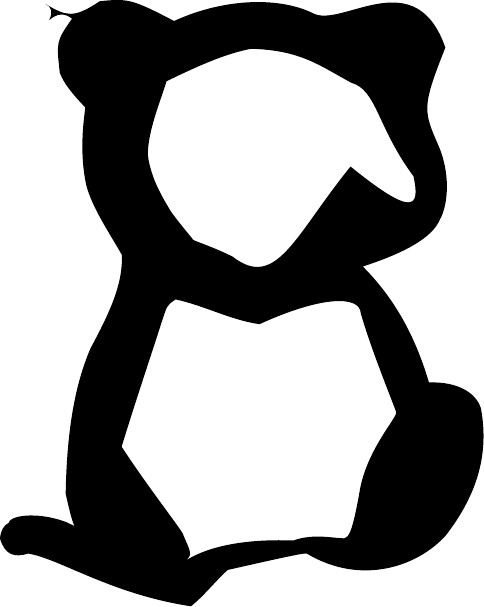} &
    \includegraphics[height=0.1\linewidth]{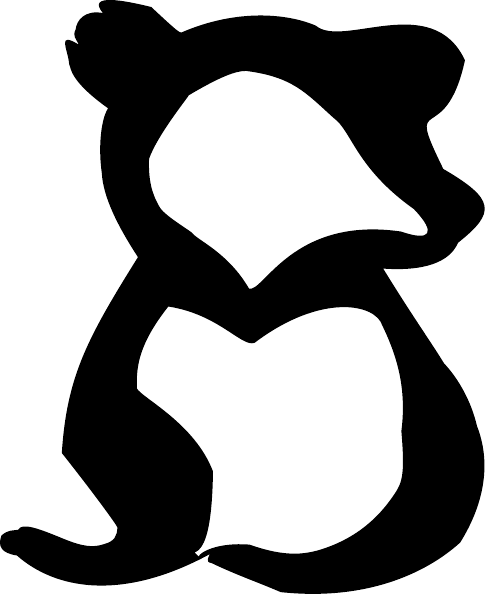} &
    \includegraphics[height=0.1\linewidth]{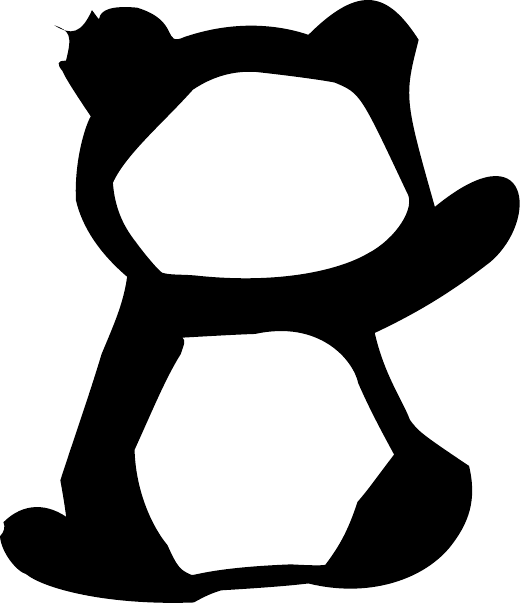} &
    \includegraphics[height=0.1\linewidth]{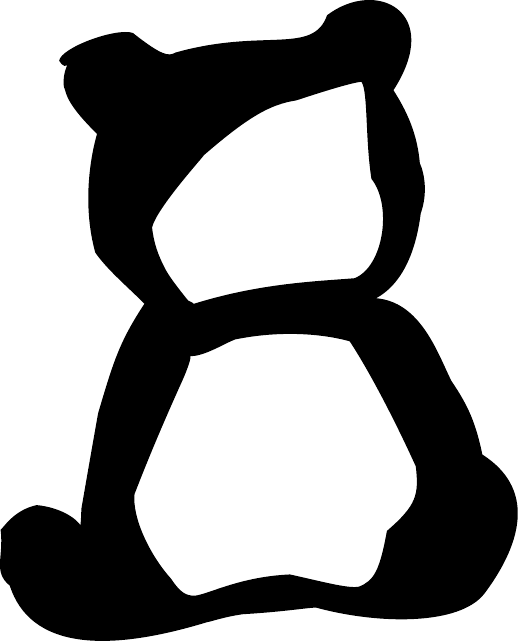}\\
    
    \raisebox{0.4cm}{"Singer"} &
    \includegraphics[height=0.09\linewidth]{word-as-image/images/weight_effect/Noteworthy-Bold_N_scaled.pdf} &
    \hspace{0.2cm}
    \includegraphics[height=0.1\linewidth]{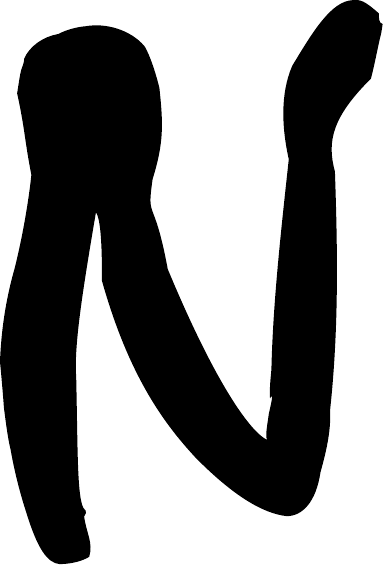} &
    \includegraphics[height=0.1\linewidth]{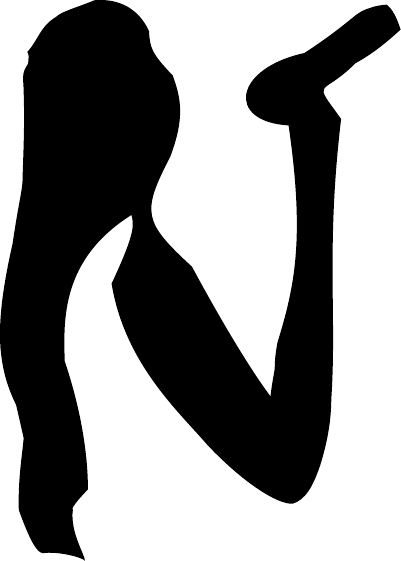} &
    \includegraphics[height=0.1\linewidth]{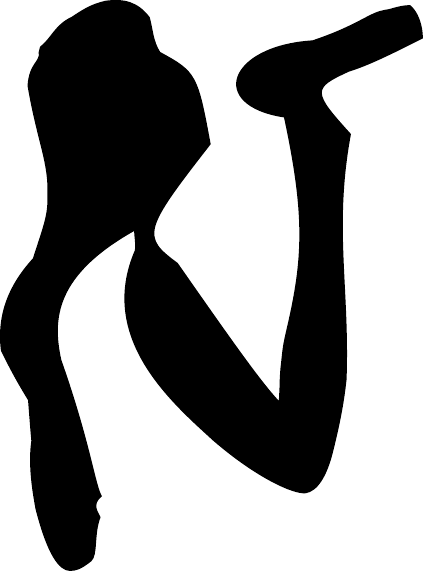} &
    \includegraphics[height=0.1\linewidth]{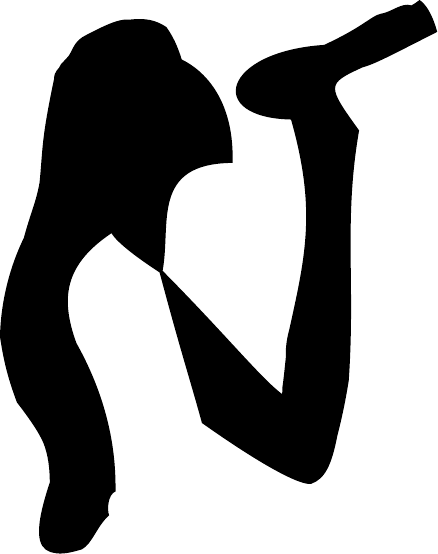} &
    \includegraphics[height=0.1\linewidth]{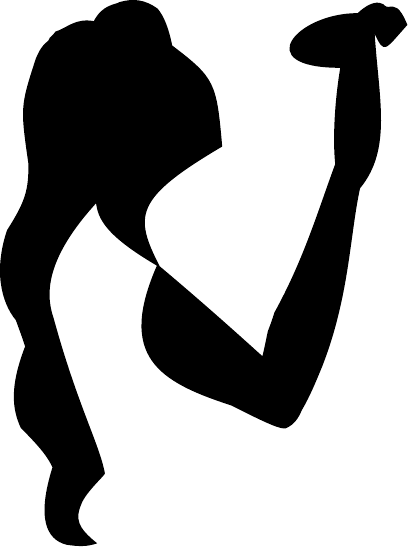} \\

    \raisebox{0.4cm}{"Giraffe"} &
    \includegraphics[height=0.09\linewidth]{word-as-image/images/weight_effect/Bell_MT_R_scaled.pdf} &
    \hspace{0.2cm}
    \includegraphics[height=0.1\linewidth]{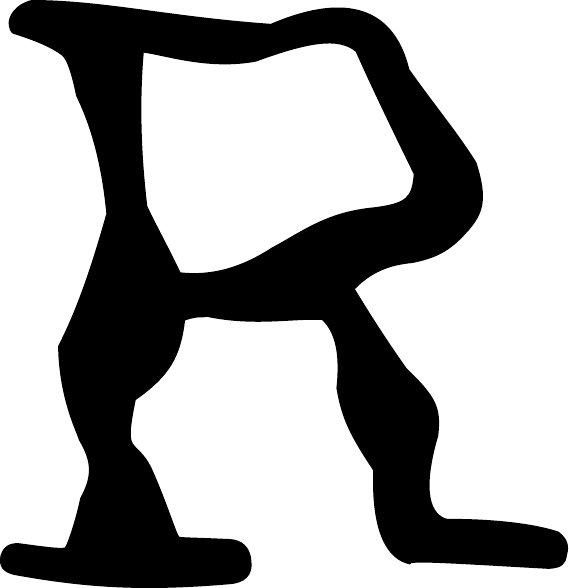} &
    \includegraphics[height=0.1\linewidth]{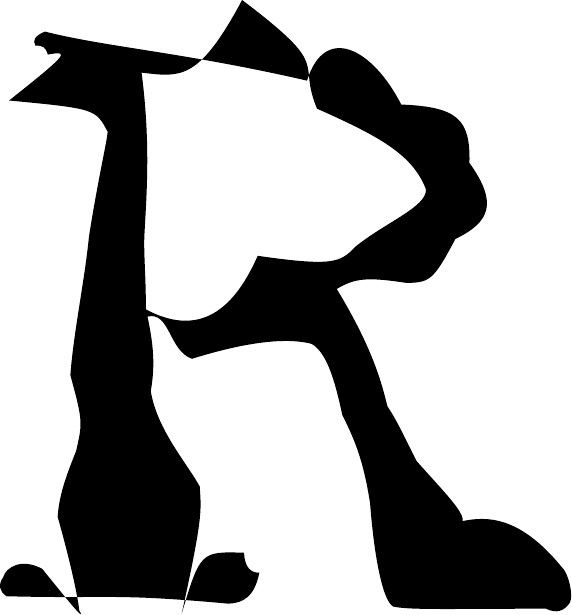} &
    \includegraphics[height=0.1\linewidth]{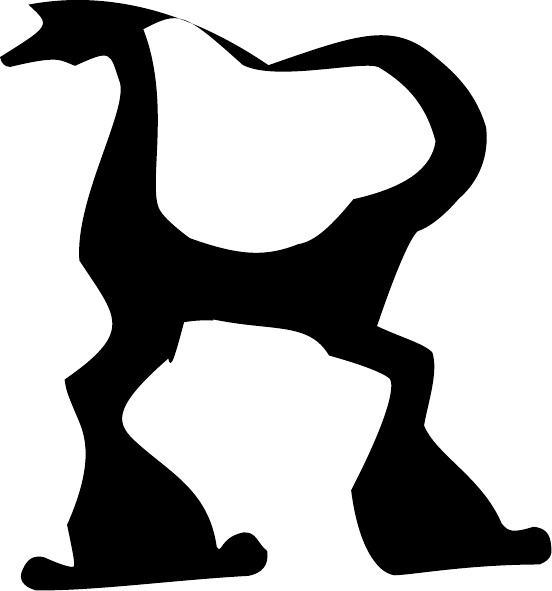} &
    \includegraphics[height=0.1\linewidth]{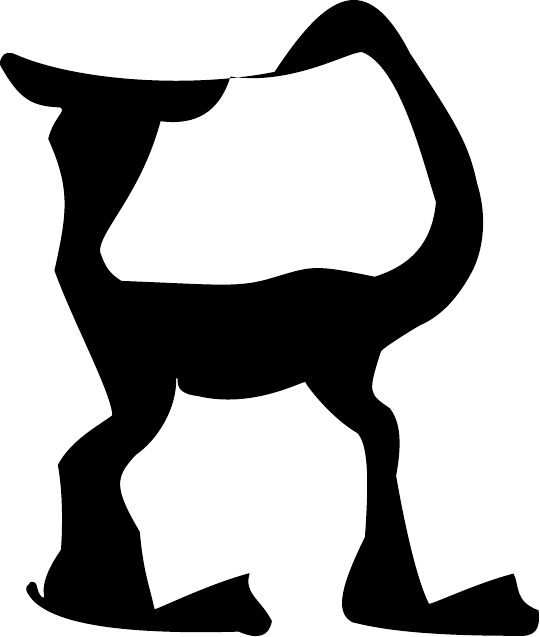} &
    \includegraphics[height=0.1\linewidth]{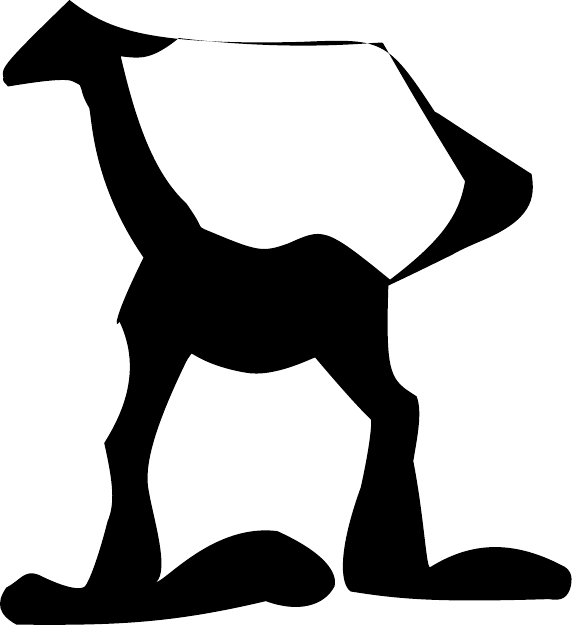} \\

    \multicolumn{2}{c}{Input} & 1 & 5 & 30 & 200 & \makecell[c]{Without \\ $\mathcal{L}_{tone}$}\\
        
\end{tabular}
\caption{\small Altering the $\sigma$ parameter of the low pass filter using in the $\mathcal{L}_{tone}$ loss. On the leftmost column are the original letters and concepts used, then from left to right are the results obtained when using $\sigma\in\{1, 5, 30, 200\}$, and without $\mathcal{L}_{tone}$.}
\label{fig:sigma_L2}
\end{figure}

\begin{figure}[hb]
    \centering
    \setlength{\tabcolsep}{5pt}
    \renewcommand{\arraystretch}{1}
    \begin{tabular}{l@{\hspace{0.2cm}} | @{\hspace{0.2cm}}c c c c cl}

        \raisebox{0.4cm}{\makecell[l]{Input \\ Letter }} &
        \hspace{0.2cm}
        \raisebox{0.25cm}{\includegraphics[height=0.08\linewidth]{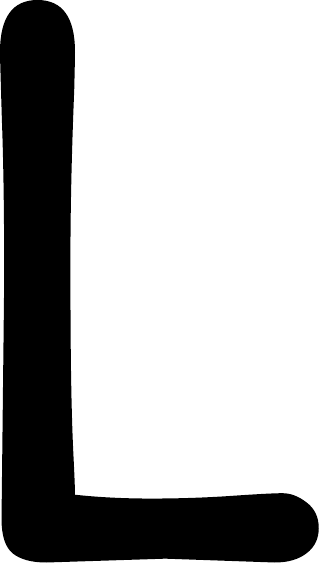}} &
        \raisebox{0.25cm}{\includegraphics[height=0.08\linewidth]{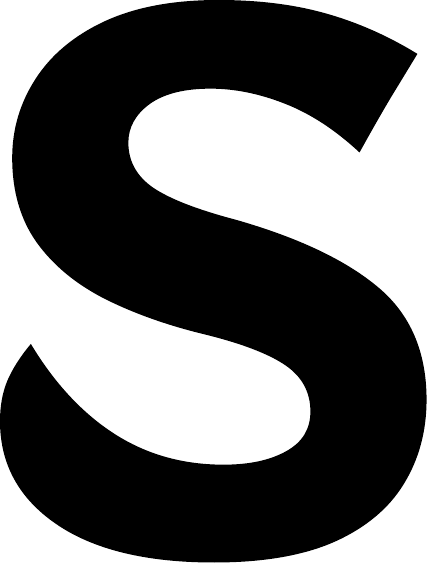}} &
        \raisebox{0.25cm}{\includegraphics[height=0.08\linewidth]{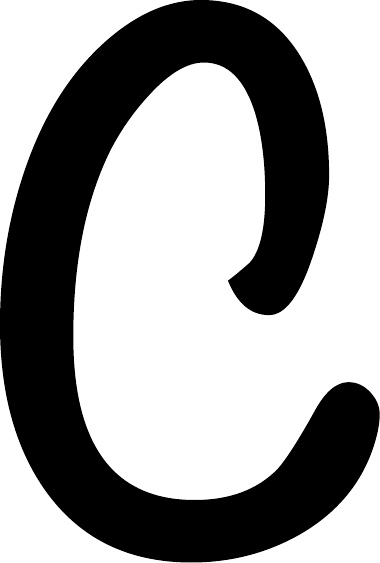}} &
        \raisebox{0.25cm}{\includegraphics[height=0.08\linewidth]{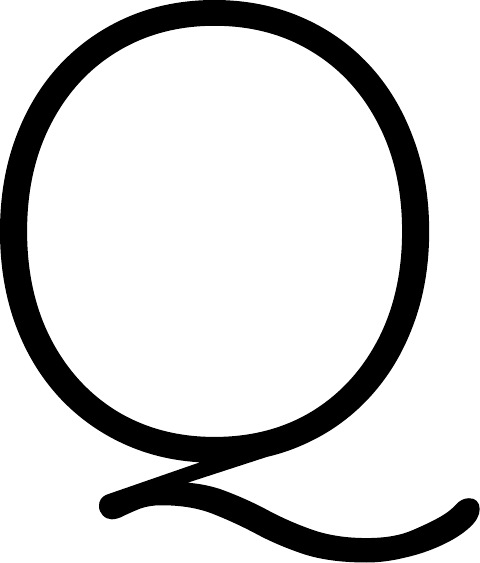}} &
        \raisebox{0.25cm}{\includegraphics[height=0.08\linewidth]{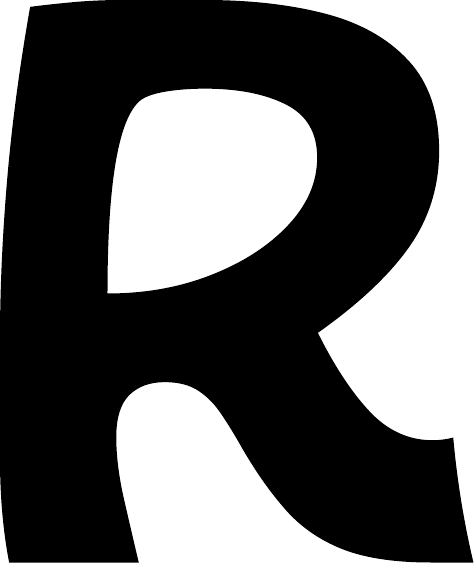}} \\

        \raisebox{0.4cm}{\makecell[l]{CLIP \\ loss}} &
        \hspace{0.2cm}
        \raisebox{0.25cm}{\includegraphics[height=0.08\linewidth]{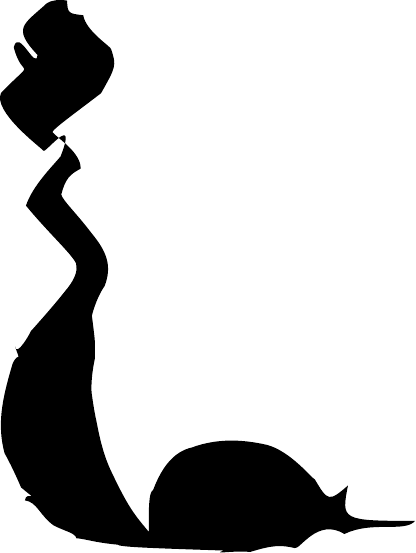}} &
        \raisebox{0.25cm}{\includegraphics[height=0.08\linewidth]{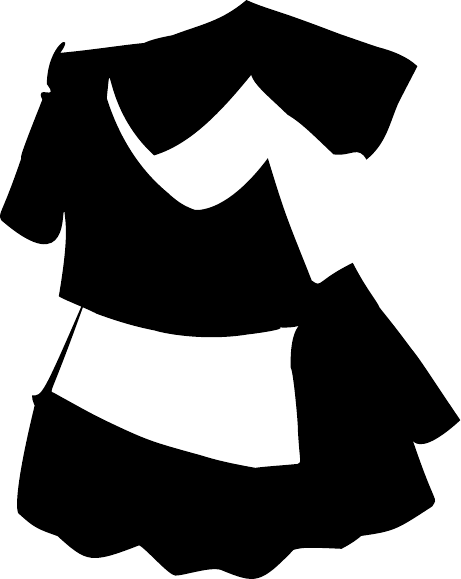}} &
        \raisebox{0.25cm}{\includegraphics[height=0.08\linewidth]{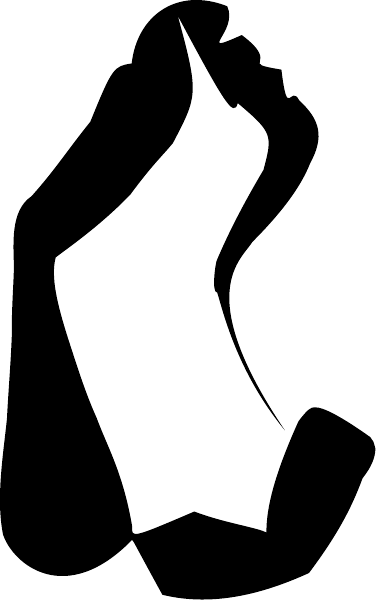}} &
        \raisebox{0.25cm}{\includegraphics[height=0.08\linewidth]{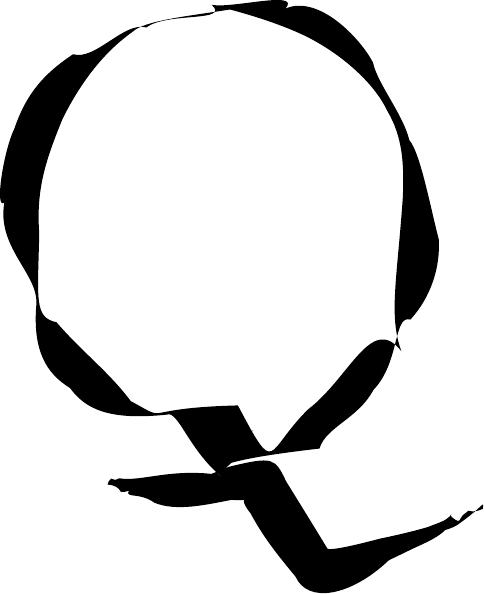}} &
        \raisebox{0.25cm}{\includegraphics[height=0.08\linewidth]{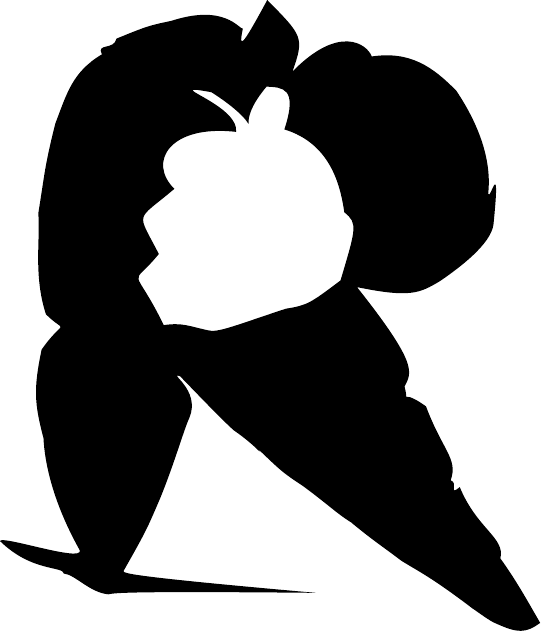}} \\

        \raisebox{0.4cm}{\makecell[l]{SDS \\ loss}} &
        \hspace{0.2cm}
        \raisebox{0.25cm}{\includegraphics[height=0.08\linewidth]{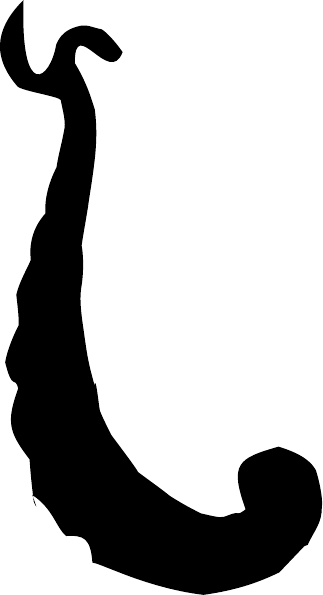}} &
        \raisebox{0.25cm}{\includegraphics[height=0.08\linewidth]{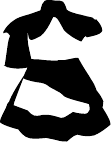}} &
        \raisebox{0.25cm}{\includegraphics[height=0.08\linewidth]{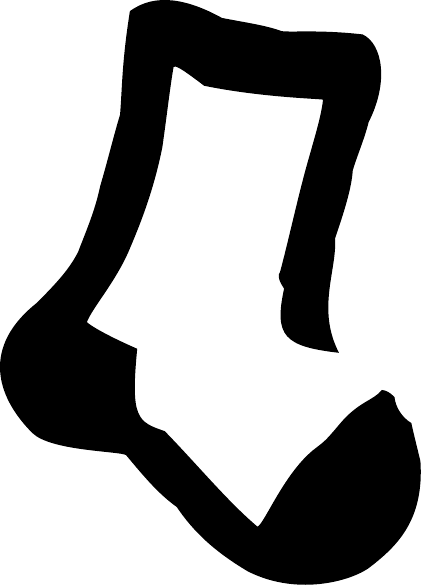}} &
        \raisebox{0.25cm}{\includegraphics[height=0.08\linewidth]{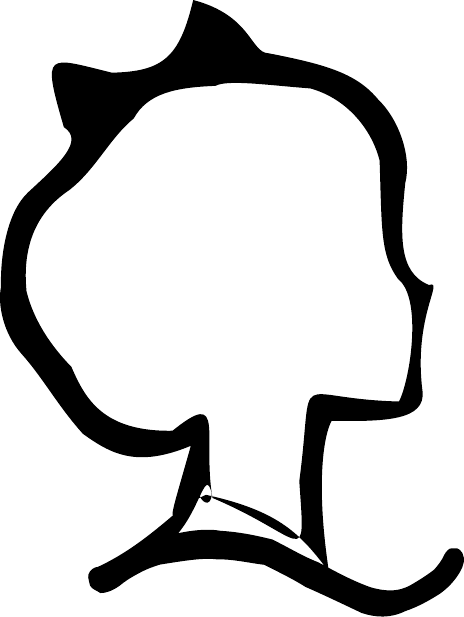}} &
        \raisebox{0.25cm}{\includegraphics[height=0.08\linewidth]{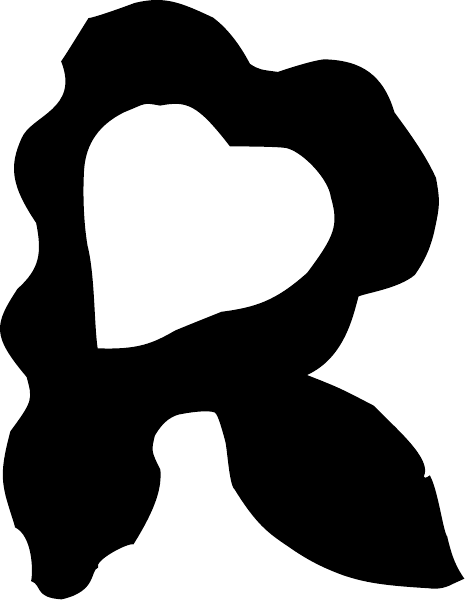}} \\

        &\hspace{0.1cm} 
        "Snail" & "Skirt" & "Socks" & "Queen" & "Strawberry" \\
    \end{tabular}
    \caption{\small Replacing the SDS loss with a CLIP-based loss.}
    \label{fig:comp_clip}
\end{figure}

\subsection{Ablation}
\label{sec:ablation}
Figure \ref{fig:cc_effect} illustrates the impact of the letter's initial number of control points. 
When less control points are used ($P_o$ is the original number of control points), we may get insufficient variations, such as for the gorilla. However, this can also result in more abstract depictions, such as the ballerina.
As we add control points, we get more graphic results, with the tradeoff that it often deviate from the original letter. 
In Figure \ref{fig:sds_lr} we show the results of using only the $\nabla_{\hat{P}} \mathcal{L}_\text{LSDS}$ loss.
As can be seen, in that case the illustrations strongly convey the semantic concept, however at the cost of legibility.
In Figure \ref{fig:weights_conformal} we analyze the effect of the weight $\alpha$ applied to $\mathcal{L}_{acap}$. Ranging from $1$ to $0$.
When $\mathcal{L}_{acap}$ is too dominant, the results may not enough reflect the semantic concept, while the opposite case harms legibility.
Figure \ref{fig:sigma_L2} illustrates a change in the $\sigma$ parameter of the low pass filter. When $\sigma=1$ almost no blur is applied, resulting in a shape constraint that is too strong.

In Figure \ref{fig:comp_clip} we show the results of replacing the $\nabla_{\hat{P}} \mathcal{L}_\text{LSDS}$ loss with a CLIP based loss, while using our proposed shape preservation terms. Although the results obtained with CLIP often depict the desired visual concept, we find that using Stable Diffusion leads to smoother illustrations, that capture a wider range of semantic concepts.

By using the hyperparameters described in the paper, we are able to achieve a reasonable balance between semantics and legibility. The parameters were determined manually based on visual assessments, but can be adjusted as needed based on the user's personal taste and goals.

\begin{figure}[h!]
\centering
\includegraphics[width=0.95\linewidth]{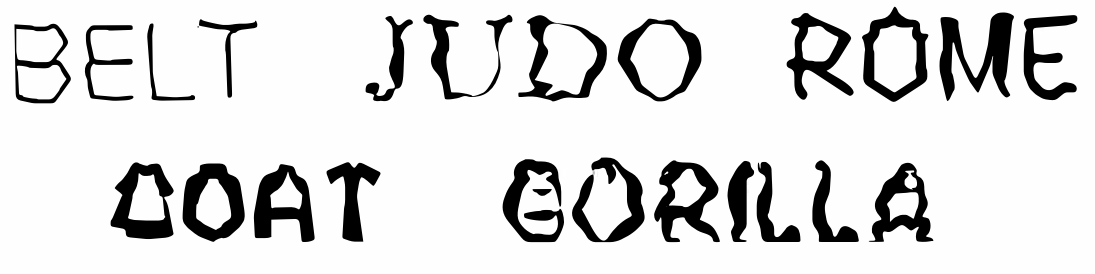}
    \caption{\small Failure cases: letters that do not convey the word's concept (top), or lose their legibility (bottom -- `C' in `Coat' and 'G' in 'Gorilla') }
    \label{fig:failure}
\end{figure}
\section{Limitations and Future Work}
There are some limitations to our method.
In some cases, the deformed letters may not convey the intended semantic concept. 
For example, see the words ``Belt'', ``Judo'', and ``Rome'' in Figure~\ref{fig:failure}, top.
A possible explanation for this limitation is that the $\nabla_{\hat{P}} \mathcal{L}_\text{LSDS}$ loss fails to capture the desired concept, or that the shape regularization terms are too strict.

On the other hand, there may be cases where the letters convey the semantic concept well, but the word is not legible. This can be seen in the `C' of the coat, and the `G' of the Gorilla in Figure~\ref{fig:failure}, bottom.
In addition, we chose to operate on each letter individually, and therefore, our method cannot be applied to deform the entire word at once. This would possibly demand adding more loss functions to preserve the word structure and is left for future research.

Our approach works best on concrete visual concepts, and may fail with more abstract ones. This can be alleviated by optimizing the shape of letters using different concepts than the word itself. Lastly, we do not change the layout of letters algorithmically, but this can be done for example, using methods such as \cite{Wang_2022_CVPR}.

\section{Conclusions}
We presented a method for the automatic creation of vector-format word-as-image illustrations. Our method can handle a large variety of semantic concepts and be used on any font, while preserving the legibility of the text and the font's style.

Our word-as-image illustrations demonstrate visual creativity and open the possibility for the use of large vision-language models for semantic typography, possibly also adding human-in-the-loop to arrive at more synergistic design methods of ML models and humans.

\begin{figure*}
\centering
    \includegraphics[width=1\textwidth]{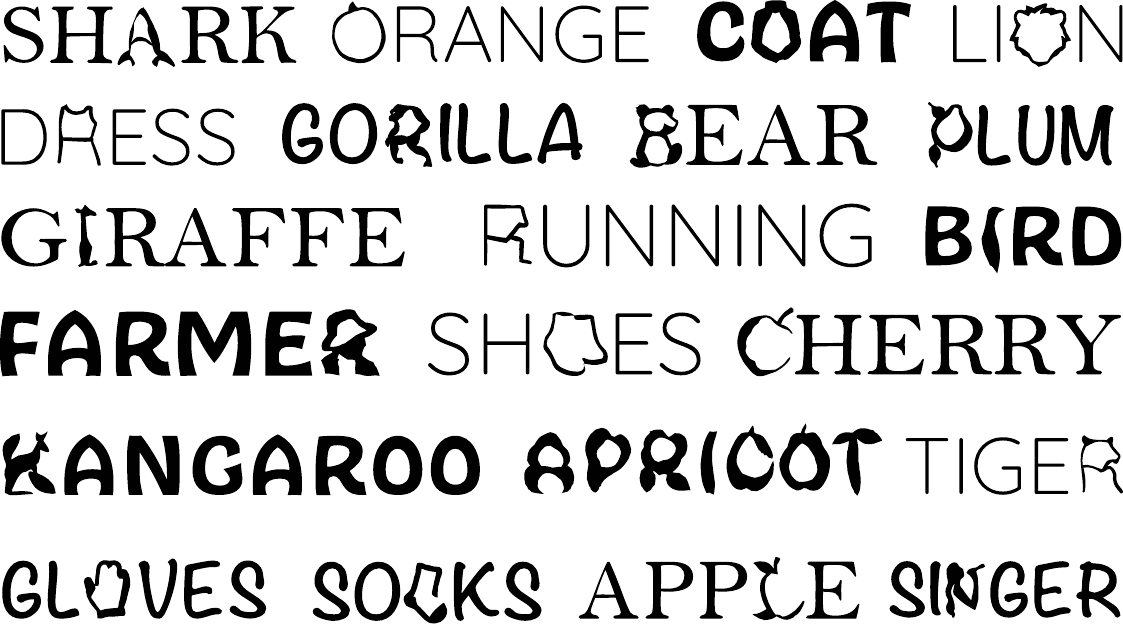}
    \caption{\small Word-as-images produced by our method. This subset was chosen from the random set of words.}
    \label{fig:all_res1}
\end{figure*}

\begin{figure*}
\centering
    \includegraphics[width=1\textwidth]{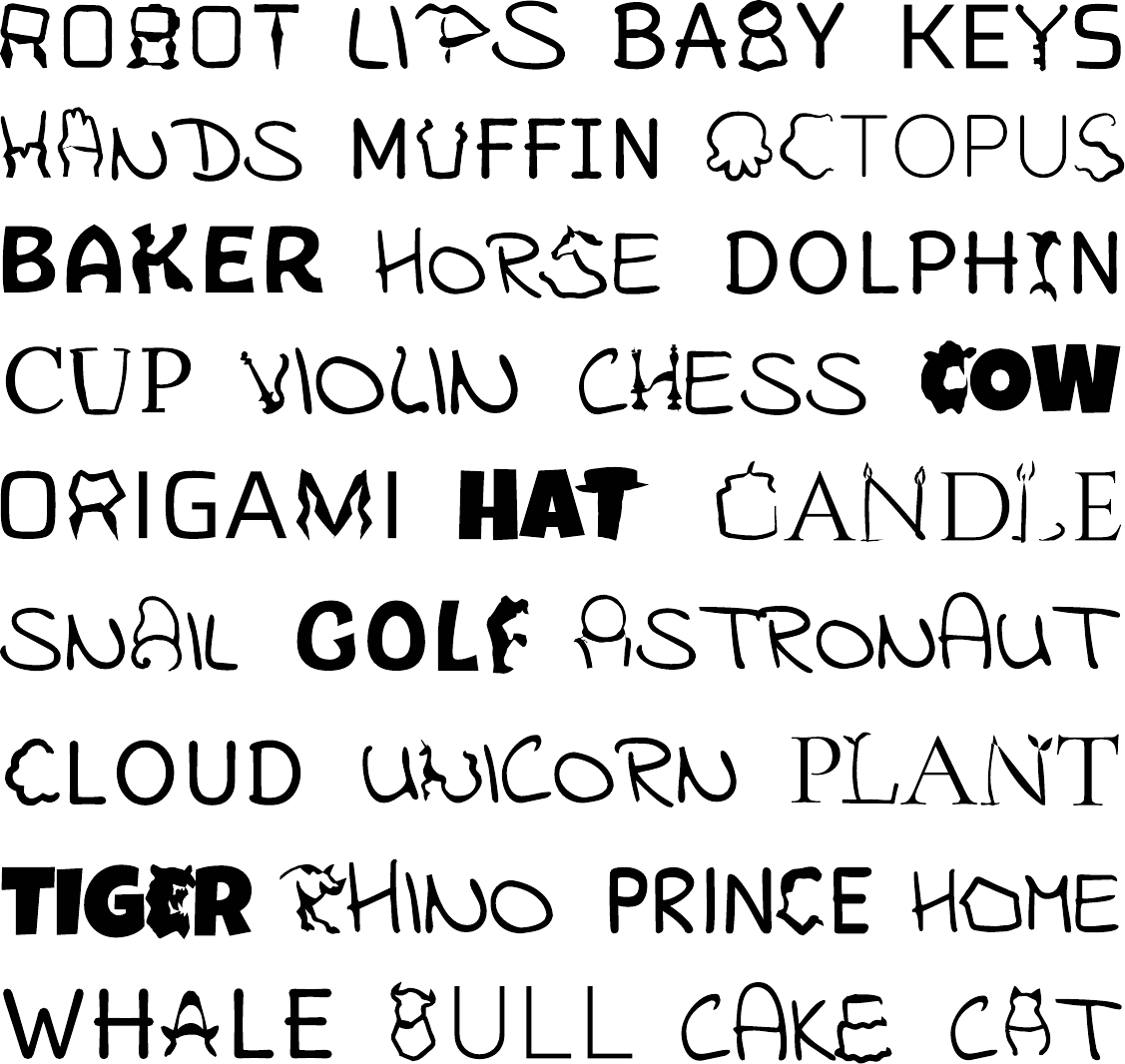}
    \caption{\small Additional results produced by our method.}
    \label{fig:mayne_res}
\end{figure*}

\part{Text Guided Generation of Short Animations}
\label{part:three}
\chapter{Breathing Life Into Sketches Using Text-to-Video Priors}
\label{chap:videosketch}

\begin{figure*}[h]
    \centering
    \includegraphics[width=1\linewidth]{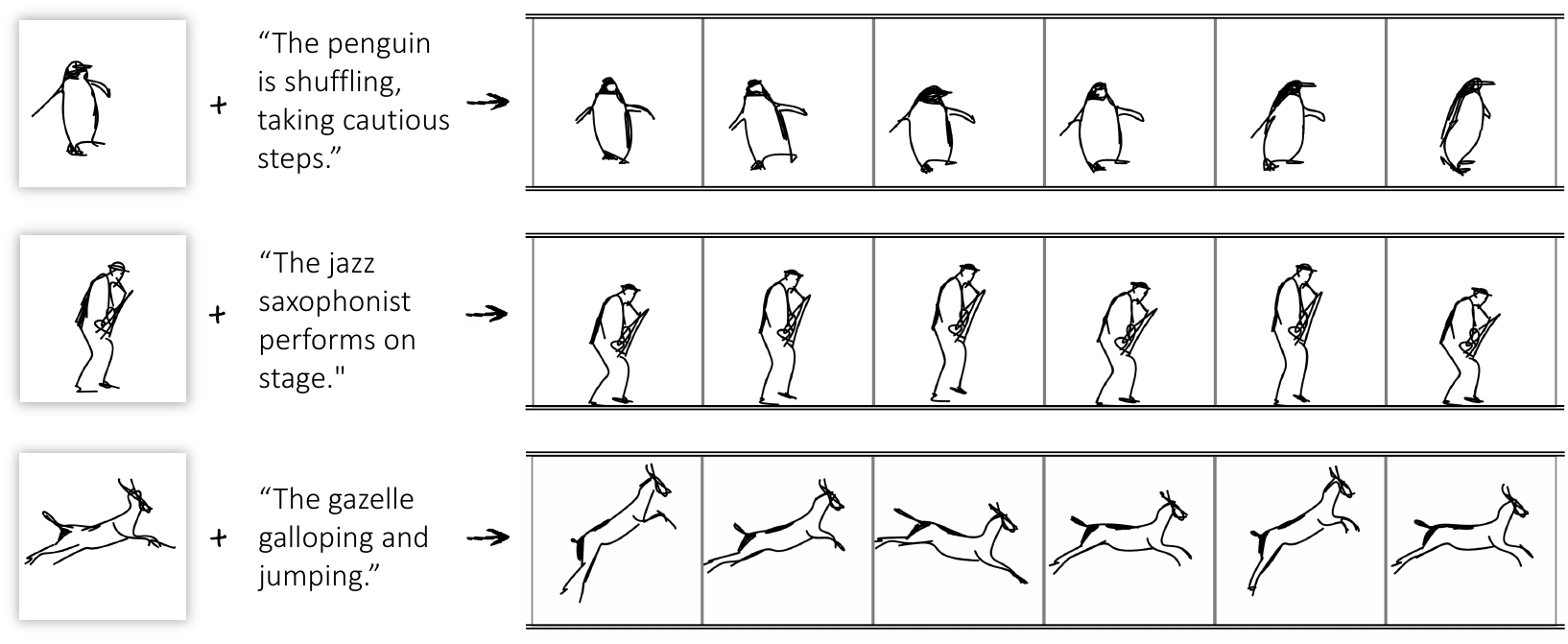}
    \caption[]{\small Given a still sketch in vector format and a text prompt describing a desired action, our method automatically animates the drawing with respect to the prompt.\footnotemark}    
    \label{fig:teaser_videosketch}
\end{figure*}

\footnotetext{Project page: \url{https://livesketch.github.io/}}
Sketches serve as a fundamental and intuitive tool for visual expression and communication \cite{Fan2023DrawingAA,aubert2014pleistocene,gombrich1995story}.
Sketches capture the essence of visual entities with a few strokes, allowing humans to communicate abstract visual ideas.
In this paper, we propose a method to \ap{breathe life} into a static sketch by generating semantically meaningful short videos from it.
Such animations can be useful for storytelling, illustrations, websites, presentations, and just for fun. 

Animating sketches using conventional tools (such as Adobe Animate and Toon Boom) is challenging even for experienced designers~\cite{su2018live}, requiring specific artistic expertise.
Hence, long-standing research efforts in computer graphics sought to develop automatic tools to simplify this process. 
However, these tools face multiple hurdles, such as a need to identify the semantic component of the sketch, or learning to create motion that appears natural. As such, existing methods commonly rely on user-annotated skeletal key points \cite{smith2023method,dvorovznak2018toonsynth} or user-provided reference motions that align with the sketch semantics \cite{su2018live,bregler2002turning,weng2019photo}.

In this work, we propose to bring a given static sketch to life, based on a textual prompt, without the need for any human annotations or explicit reference motions. We do so by leveraging a pretrained text-to-video diffusion model~\cite{khachatryan2023text2video}.
Several recent works propose using the prior of such models to bring life to a static \textit{image}~\cite{videocomposer2023,xing2023dynamicrafter,ni2023conditional}. 
However, sketches pose distinct challenges, which existing methods fail to tackle as they are not designed with this domain in mind. 
Our method takes the recent advancement in text-to-video models into this new realm, aiming to tackle the challenging task of sketch animation. For this purpose, we propose specific design choices considering the delicate characteristics of this abstract domain.

In line with prior sketch generation approaches \cite{vinker2022clipasso, clipascene}, we use a vector representation of sketches, defining a sketch as a set of strokes (cubic \Bezier curves) parameterized by their control points. 
Vector representations are popular among designers as they offer several advantages compared to pixel-based images. They are resolution-independent, \ie can be scaled without losing quality. 
Moreover, they are easily editable: one can modify the sketch's appearance by choosing different stroke styles or change its shape by dragging control points. Additionally, their sparsity promotes smooth motion while preventing pixelization and blurring.

To bring a static sketch to life, we train a network to modify the stroke parameters for each video frame with respect to a given text prompt. Our method is optimization-based and requires no data or fine-tuning of large models.
In addition, our method is general and can easily adapt to different text-to-video models, facilitating the use of future advancements in this field.

We train the network using a score-distillation sampling (SDS) loss \cite{poole2022dreamfusion}. This loss was designed to leverage pretrained text-to-\textit{image} diffusion models for the optimization of non-pixel representations (e.g., NeRFs~\cite{mildenhall2021nerf,metzer2022latent} or SVGs~\cite{jain2022vectorfusion,iluz2023wordasimage}) to meet given text-based constraints. 
We use this loss to extract motion priors from pretrained text-to-\textit{video} diffusion models \cite{ho2022imagen,videocomposer2023,singer2023text}. Importantly, this allows us to inherit the internet-scale knowledge embedded in such models, enabling animation for a wide range of subjects across multiple categories.

We seperate the object movement into two components: local motion and global motion. Local motion aims to capture isolated, local effects (a saxophone player bending their knee). Conversely, global motion affects the object shape as a whole and is modeled through a per-frame transformation matrix. 
It can thus capture rigid motion (a penguin hobbling across the frame), or coordinate effects (the same penguin growing in size as it approaches the camera).
We find that this separation is crucial in generating motion that is both locally smooth and globally significant while remaining faithful to the original characteristics of the subject.

We animate sketches from various domains and demonstrate the effectiveness of our approach in producing smooth and natural motion that conveys the intention of the control text while better preserving the shape and appearance of the input sketch. 

We compare our results with recent pixel-based approaches highlighting the advantage of vector-based animation in the sketch domain. Our work allows anyone to breath life into their sketch in a simple and intuitive manner.

\section{Previous Work}
\label{sec:prevwork}

\paragraph{Sketches}
Free-hand sketching is a valuable tool for expressing ideas, concepts, and actions~\cite{Fan2018CommonOR, Hertzmann2020WhyDL, Fan2023DrawingAA}.
Extensive research has been conducted on the automatic generation of sketches \cite{xu2020deep}. 
Some works utilize pixel representation \cite{li2019photosketching, song2018learning, human-like-sketches, xie2015holistically}, 
while others employ vector representation \cite{SketchRNN, Lin2020SketchBERTLS, Bhunia2020PixelorAC, Ribeiro2020SketchformerTR, Chen2017Sketchpix2seqAM, mo2021virtualsketching, Deformable_Stroke, bhunia2021doodleformer, mihai2021learning}. 
Several works propose a unified algorithm to produce sketches with a variety of styles \cite{chan2022learning, yi2020unpaired, liu2021neural} or at varying levels of abstraction \cite{Berger2013,Deep-Sketch-Abstraction,vinker2022clipasso,clipascene}. 
Traditional methods for sketch generation commonly rely on human-drawn sketch datasets. More recently, some works~\cite{CLIPDraw, vinker2022clipasso, clipascene} incorporated the prior of large pretrained language-vision models to eliminate the dependency on such datasets.
We also rely on such priors, and use a vector-based approach to depict our sketches, as it is a more natural representation for sketches and finds widespread use in character animation.

\vspace{-0.15cm}
\paragraph{Sketch-based animation}
A long-standing area of interest in computer graphics aims to develop intuitive tools for creating life-like animations from still inputs.
In character animation, motion is often represented as a temporal sequence of poses. These poses are commonly represented via user provided annotations, such as stick figures~\cite{davis2006sketching}, skeletons~\cite{Pan2011,ArtiSketch2013}, or partial bone lines~\cite{Oztireli13Differential}. 
An alternative line of work represents motion through user provided 2D paths~\cite{DavisKSketch2008,IgarashiPath98,Gleicher2001,thorne2004motion}, or through space-time curves~\cite{guay2015space}. However, these approaches still require some expertise and manual work to adjust different keyframes.
Some methods assist animation by interactively predicting what users will draw next~\cite{WangVideoTooning2004,Agarwala2004,Xing2015}. However, they still require manual sketching operations for each keyframe. 
Rather than relying on user-created motion, some works propose to extract motion from real videos by statistical analysis of datasets \cite{Min2009}, or by applying dynamic deformations extracted from a driving video~\cite{su2018live}. 
Others turn to physically-based motion effects~\cite{DRACOKazi2014, XingEnergyBrushes2016}, or learn to synthesize animations of hand-drawn 2D characters using a set of images depicting the character in various poses~\cite{dvorovznak2018toonsynth,poursaeed2020neural,hinz2022charactergan}. Zhang~\etal~\cite{Zhang_2023} presented a method for transferring motion between vector-graphics videos using a \ap{motion vectorization} pipeline.

Drawing on 3D literature, some works aim to "wake up" a photo or a painting, extracting a textured human mesh from the image and moving it using pre-defined animations~\cite{Weng2018PhotoW3,levon2020texanmesh,Hornung2007}.
More recently, given a hand-drawn sketch of a human figure, Smith~\etal~\cite{smith2023method} construct a character rig onto which they re-target motion capture data. Their approach is similarly limited to human figures and a predefined set of movements. Moreover, it commonly requires direct human intervention to fix skeleton joint estimations. 

In contrast to these methods, our method requires only a single sketch and no skeletons or explicit references. Instead, it leverages the strong prior of text-to-video generative models and generalizes across a wide range of animations described by free-form prompts.

\vspace{-0.2cm}
\paragraph{Text-to-video generation}
Early works explored expanding the capabilities of recurrent neural networks~\cite{babaeizadeh2017stochastic,denton2018stochastic,castrejon2019improved}, GANs~\cite{kim2020tivgan,tian2021a,zhu2023motionvideogan,pan2017create, li2018video}, and auto-regressive transformers~\cite{weissenborn2019scaling, yan2021videogpt,wu2021godiva,wu2022nuwa} from image generation to video generation. However, these works primarily focused on generating videos within limited domains.
More recent research extends the capabilities of powerful text-to-image diffusion models to video generation by incorporating additional temporal attention modules into existing models or by temporally aligning an image decoder~\cite{singer2022make,blattmann2023videoldm,videofusion2023,wang2023lavie}. Commonly, such alignment is performed in a latent space~\cite{zhou2023magicvideo,li2023videogen,an2023latentshift,videofusion2023,esser2023structure,wang2023modelscope}.
Others train cascaded diffusion models~\cite{ho2022imagen}, or learn to directly generate videos within a lower-dimensional 3D latent space~\cite{he2022lvdm}.
We propose to extract the motion prior from such models~\cite{singer2023text} and apply it to a vector sketch representation.

\vspace{-0.2cm}
\paragraph{Image-to-video generation}
A closely related research area is image-to-video generation, where the goal is to animate an input image. Make-It-Move~\cite{hu2022make} train an encoder-decoder architecture to generate video sequences conditioned on an input image and a driving text prompt. Latent Motion Diffusion~\cite{hu2023lamd} learn the optical flow between pairs of video frames and use a 3D-UNet-based diffusion model to generate the resulting video sequence. CoDi~\cite{tang2023anytoany} align multiple modalities (text, image, audio, and video) to a shared conditioning space and output space. ModelScope~\cite{wang2023modelscope} train a latent video diffusion model, conditioned on an image input. Others first caption an image, then use the caption to condition a text-to-video model~\cite{videofusion2023}. VideoCrafter~\cite{chen2023videocrafter1} train a model conditioned on both text and image, with a special focus preserving the content, structure, and style of this image. Gen-2~\cite{gen2runway} also operate in this domain, though their model's details are not public. 
While showing impressive results in the pixel domain, these methods struggle to generalize to sketches. Our method is designed for sketches, constraining the outputs to vector representations that better preserve both the domain, and the characteristics of the input sketch.

\section{Method}
\label{sec:method_videosketch}

Our method begins with two inputs: a user-provided static sketch in vector format, and a text prompt describing the desired motion. Our goal is to generate a short video, in the same vector format, which depicts the sketched subject acting in a manner consistent with the prompt.
We therefore define three objectives that our approach should strive to meet: (1) the output video should match the text prompt, (2) the characteristics of the original sketch should be preserved, and (3) the generated motion should appear natural and smooth.
Below, we outline the design choices we use to meet each of these objectives.

\begin{figure}
    \centering
     \includegraphics[width=0.5\linewidth]{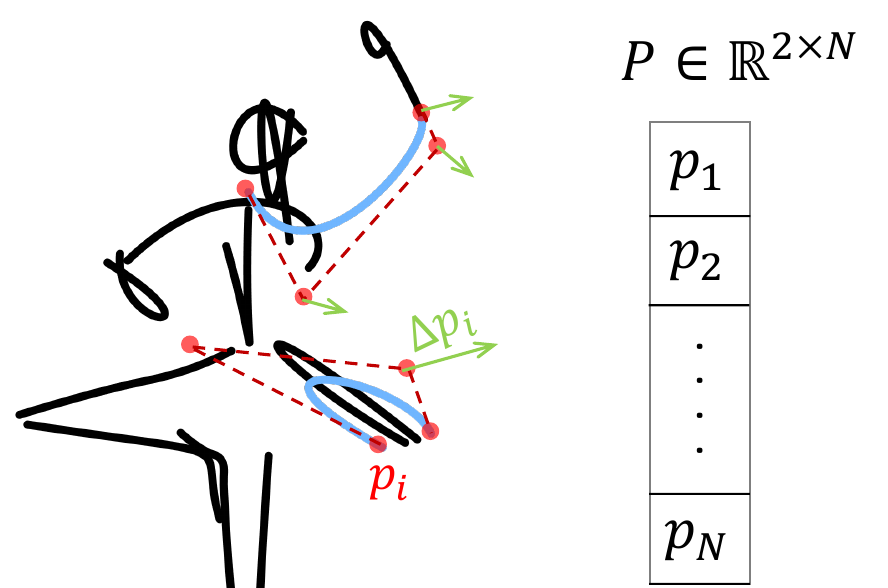}
    \caption{\small Data representation. Each curve (black or blue) is a cubic \Bezier curve with $4$ control points (red, shown for the blue curves). The total number of control points in the given sketch is denoted by $N$. For each frame and control point $p_i$, we learn a displacement $\Delta p_i$ (green).
    }
    \label{fig:data_rep}
\end{figure}

\subsection{Representation}
The input vector image is represented as a set of strokes placed over a white background, where each stroke is a two-dimensional \Bezier curve with four control points. Each control point is represented by its coordinates: $p = (x, y) \in \mathbb{R}^2$. We denote the set of control points in a single frame with $P=\{p_1, .. p_N\} \in \mathbb{R}^{N \times 2}$, where $N$ denotes the total number of points in the input sketch (see \Cref{fig:data_rep}). This number will remain fixed across all generated frames.
We define a video with $k$ frames as a sequence of $k$ such sets of control points, and denote it by $Z = \{P^j\}_{j=1}^k \in \mathbb{R}^{N \cdot k \times 2}$. 

Let $P^{init}$ denote the set of points in the initial sketch. We duplicate $P^{init}$ $k$ times to create the initial set of frames $Z^{init}$.
Our goal is to convert such a static sequence of frames into a sequence of frames animating the subject according to the motion described in the text prompt.
We formulate this task as learning a set of 2D displacements $\Delta Z = \{\Delta p_i^j\}_{i\in N}^{j\in k}$, indicating the displacement of each point $p_i^j$, for each frame $j$ (\cref{fig:data_rep}, in green).

\begin{figure}
    \centering
    \includegraphics[width=0.55\linewidth]{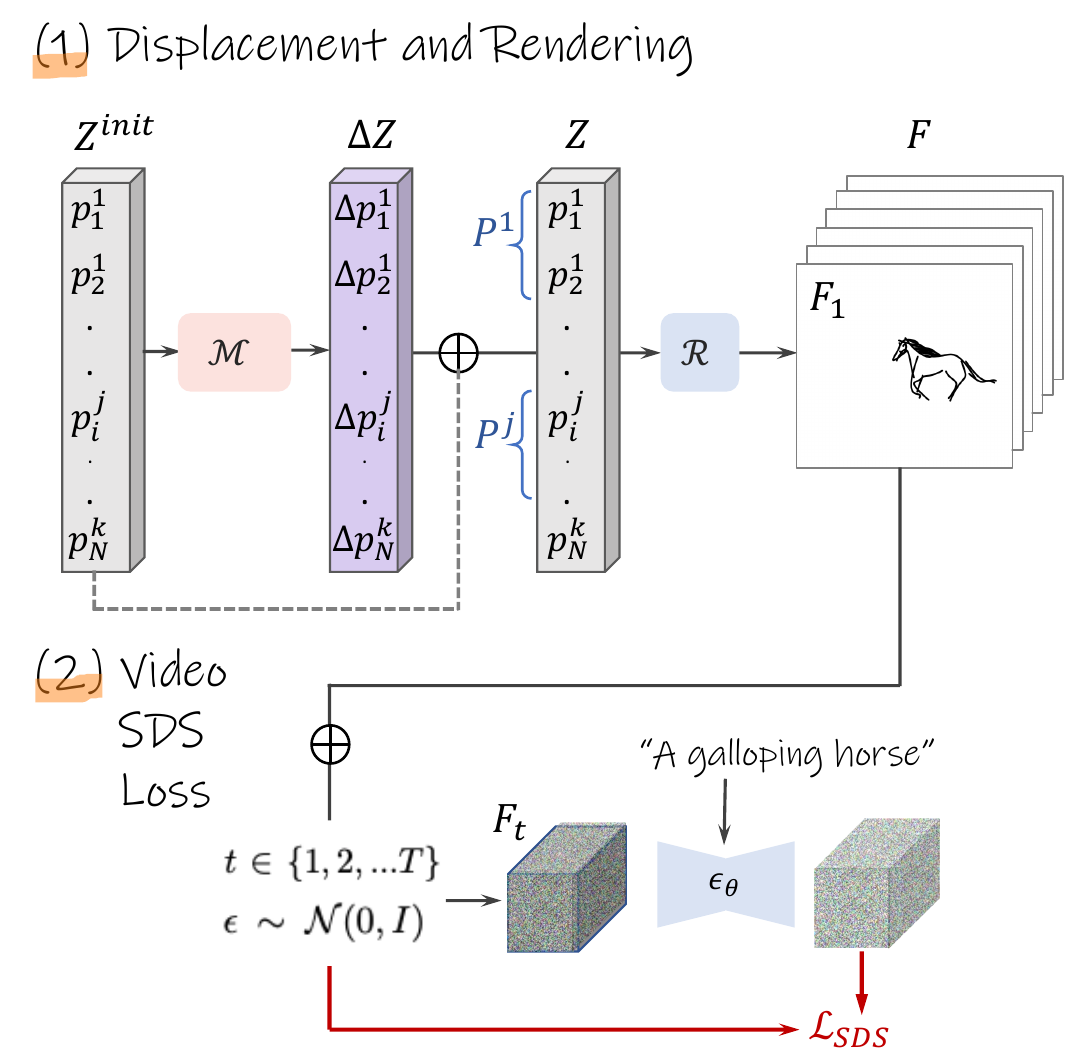}
    \caption{\small Text-driven optimization. At each training iteration: (1) We duplicate the initial control points across $k$ frames and sum them with their predicted offsets. We render each frame and concatenate them to create the output video. (2) We use the SDS loss to extract a signal from a pretrained text-to-video model, which is used to update $\mathcal{M}$, the model that predicts the offsets.}
    \label{fig:pipe_part1}

\end{figure}

\subsection{Text-Driven Optimization}

\label{subsec:optimize_loss}

We begin by addressing our first objective: creating an output animation that aligns with the text prompt. We model the animation using a ``neural displacement field'' (\cref{sec:neural_field}), a small network $\mathcal{M}$ that receives as input the initial point set $Z^{init}$ and predicts their displacements $\mathcal{M}(Z^{init}) = \Delta Z$. To train this network, we distill the motion prior encapsulated in a pretrained text-to-video diffusion model \cite{videocomposer2023}, using the SDS loss of \cref{eq:sds_loss}.

At each training iteration (illustrated in \cref{fig:pipe_part1}), we add the predicted displacement vector $\Delta Z$ (marked in purple) to the initial set of points $Z^{init}$ to form the sequence $Z$. 
We then use a differentiable rasterizer $\mathcal{R}$~\cite{diffvg}, to transfer each set of per-frame points $P^j$ to its corresponding frame in pixel space, denoted as $F^j = \mathcal{R}(P^j)$. The animated sketch is then defined by the concatenation of the rasterized frames, $F = \{F^1, .. F^k\} \in \mathbb{R}^{h\times w \times k}$. 

Next, we sample a diffusion timestep $t$ and noise $\epsilon \sim \mathcal{N}\left(0,1\right)$. We use these to add noise to the rasterized video according to the diffusion schedule, creating $F_t$. This noisy video is then denoised using the pretrained text-to-video diffusion model $\epsilon_{\theta}$, where the diffusion model is conditioned on a prompt describing an animated scene (e.g., ``a galloping horse''). Finally, we use \cref{eq:sds_loss} to update the parameters of $\mathcal{M}$ and repeat the process iteratively.

The SDS loss thus guides $\mathcal{M}$ to learn displacements whose corresponding rasterized animation aligns with the desired text prompt. The extent of this alignment, and hence the intensity of the motion, is determined by optimization hyperparameters such as the diffusion guidance scale and learning rates. 
However, we find that increasing these parameters typically leads to artifacts such as jitter and shape-deformations, compromising both the fidelity of the original sketch and the fluidity of natural motion (see \cref{sec:ablation_videosketch}).
As such, SDS alone fails to address our additional goals: (2) preserving the input sketch characteristics, and (3) creating natural motion. Instead, we tackle these goals through the design of our displacement field, $\mathcal{M}$.

\begin{figure*}[h!]
    \centering
    \includegraphics[width=1\linewidth]{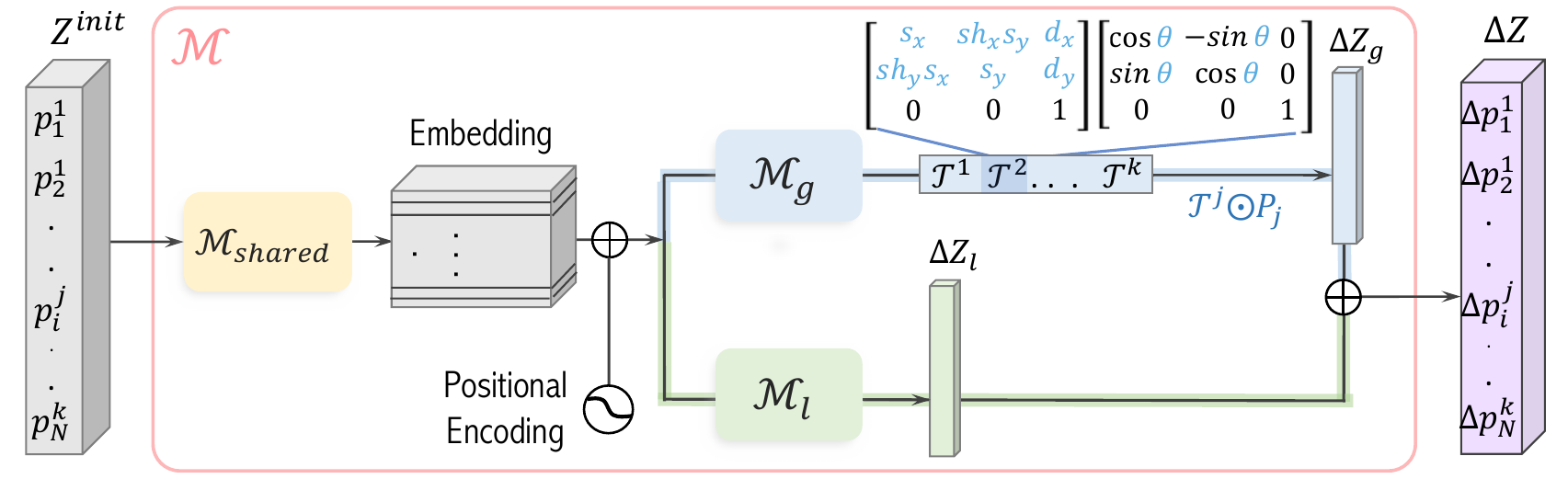}
    \caption{\small Network architecture. The input to the network is the initial set of control points $Z^{init}$ (left, gray), and the output is the set of displacements $\Delta Z$. The network consists of three parts. First, each control point $p_i^j$ is projected with $M_{shared}$ into a latent representation and summed with a positional encoding. These point features are passed to two different branches to predict global and local motion. The local motion predictor $M_l$ (green) is a simple MLP that predicts an offset for each point ($\Delta Z_l$), representing unconstrained local motion. The global motion predictor $M_g$ predicts a per-frame transformation matrix $\gtransform^j$ which applies scaling, shear, rotation, and translation. $\gtransform^j$ is then applied to the points $P_j$ in the corresponding frame to produce $\Delta Z_g$. $\Delta Z$ is given by the sum: $\Delta Z =\Delta Z_g +\Delta Z_l$. }
    \label{fig:architecture}
\end{figure*}
\subsection{Neural Displacement Field}\label{sec:neural_field}

We approach the network design with the intent of producing smoother motion with reduced shape deformations. We hypothesize that the artifacts observed with the unconstrained SDS optimization approach can be attributed in part to two mechanisms: (1) The SDS loss can be minimized by deforming the generated shape into one that better aligns with the text-to-video model's semantic prior (e.g., prompting for a scuttling crab may lead to undesired changes in the shape of the crab itself). (2) Smooth motion requires small displacements at the local scale, and the network struggles to reconcile these with the large changes required for global translations. We propose to tackle both of these challenges by modeling our motion through two components: An unconstrained local-motion path, which models small deformations, and a global path which models affine transformations applied uniformly to an entire frame. This split will allow the network to separately model motion along both scales while restricting semantic changes in the path that controls greater scale movement. Below we outline the specific network design choices, as well as the parametrization that allows us to achieve this split.

\vspace{-0.2cm}
\paragraph{Shared backbone}
Recall that our network, illustrated in \cref{fig:architecture}, aims to map the initial control point set $Z^{init}$ to their per-frame displacements $\mathcal{M}(Z^{init}) = \Delta Z$. Our first step is to create a shared feature backbone which will feed the separate motion paths. This component is built of an embedding step, where the coordinates of each control point are projected using a shared matrix $\mathcal{M}_{shared}$, and then summed with a positional encoding that depends on the frame index, and on the order of the point in the sketch. These point features are then fed into two parallel prediction paths: local, and global (\cref{fig:architecture}, green and blue paths, respectively).

\vspace{-0.2cm}
\paragraph{Local path} The local path is parameterized by $\mathcal{M}_l$, a small MLP that takes the shared features and maps them to an offset $\Delta Z_l$ for every control point in $Z^{init}$. Here, the goal is to allow the network to learn unconstrained motion on its own to best match the given prompt. Indeed, in \cref{sec:results} we show that an unconstrained branch is crucial for the model to create meaningful motion. On the other hand, using this path to create displacements on the scale needed for global changes requires stronger SDS guidance or larger learning rates, leading to jitter and unwanted deformations at the local level. Hence, we delegate these changes to the global motion path. 
We note that similar behavior can be observed when directly optimizing the control points (\ie without a network, following \cite{jain2022vectorfusion,iluz2023wordasimage}, see \cref{sec:ablation_videosketch}).

\begin{figure*}[!t]
    \centering
     \includegraphics[width=1\linewidth]{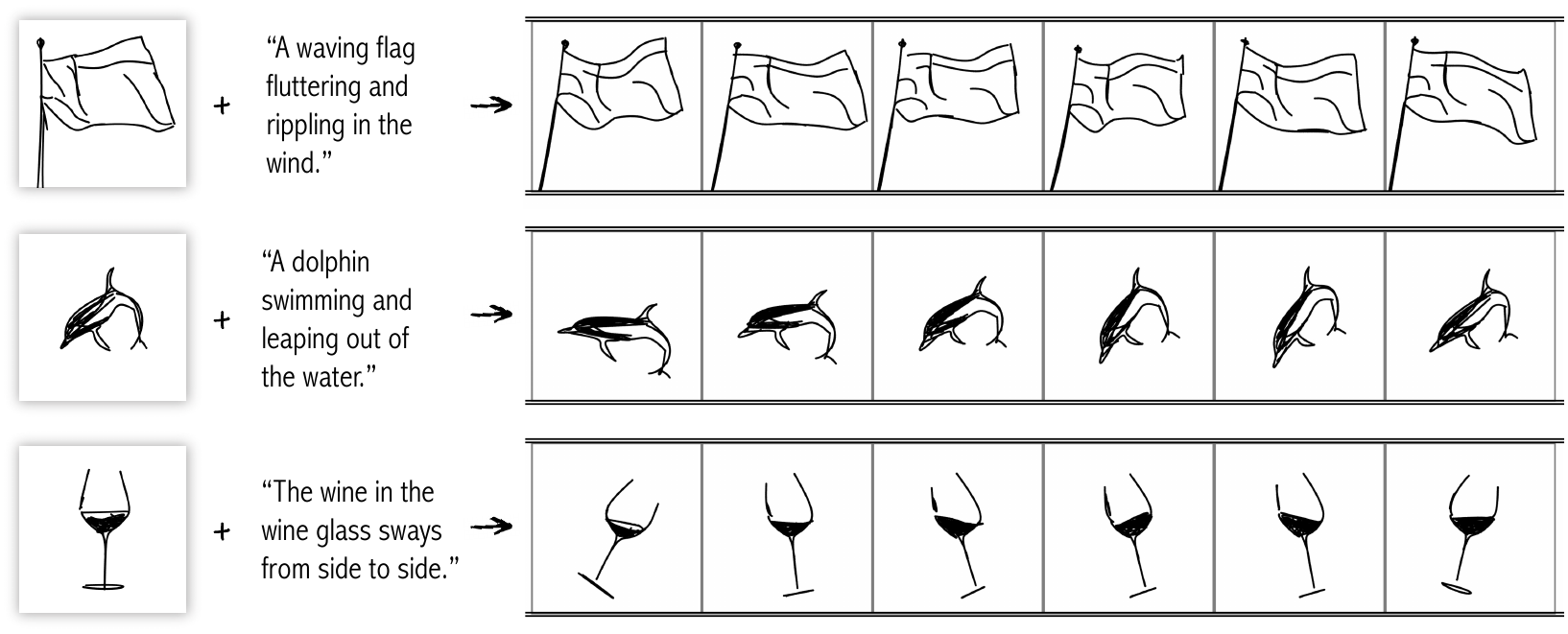}
    \caption{\small Qualitative results. Our model converts an initial sketch and a driving prompt describing some desired motion into a short video depicting the sketch moving according to the prompt. See the supplementary for the full videos and additional results.}
    \label{fig:ours_qualitative}
\end{figure*}

\vspace{-0.2cm}
\paragraph{Global path}
The goal of the global displacement prediction branch is to allow the model to capture meaningful global movements such as center-of-mass translation, rotation, or scaling, while maintaining the object's original shape. This path consists of a neural network, $\mathcal{M}_g$, that predicts a single global transformation matrix for each frame $P^j$. The matrix is then used to transform all control points of that frame, ensuring that the shape remains coherent. Specifically, we model the global motion as the sequential application of scaling, shear, rotation, and translation. These are parameterized using their standard affine matrix form (\cref{fig:architecture}), which contains two parameters each for scale, shear, and translation, and one for rotation.  %
Denoting the successive application of these transforms for frame $j$ by $\gtransform^j$, the global branch displacement for each point in this frame is then given by: $\Delta p^j_{i,global} = \gtransform^j \odot p^{init}_i - p^{init}_i$.

We further extend the user's control over the individual components of the generated motion by adding a scaling parameter for each type of transformation: $\lambda_t, \lambda_r, \lambda_s$ and $\lambda_{sh}$ for translation, rotation, scale, and shear, respectively. For example, let $(d_x^j, d_y^j)$ denote the network's predicted translation parameters. We re-scale them as: $(d_x^j, d_y^j) \rightarrow (\lambda_{t}d_x^j, \lambda_{t}d_y^j)$. This allows us to attenuate specific aspects of motion that are undesired. For example, we can keep a subject roughly stationary by setting $\lambda_t = 0$. By modeling global changes through constrained transformations, applied uniformly to the entire frame, we limit the model's ability to create arbitrary deformations while preserving its ability to create large translations or coordinated effects.

\mbox{} \\
Our final predicted displacements $\Delta Z$ are simply the sum of the two branches: $\Delta Z_l + \Delta Z_g$. The strength of these two terms (governed by the learning rates and guidance scales used to optimize each branch) will affect a tradeoff between our first goal (text-to-video alignment), and the other two goals (preserving the shape of the original sketch and creating smooth and natural motion). As such, a user can use this tradeoff to gain additional control over the generated video. For instance, prioritizing the preservation of sketch appearance by using a low learning rate for the local path, while affording greater freedom to global motion. We further demonstrate this tradeoff in the supplementary.

\subsection{Training Details}

We alternate between optimizing the local path and optimizing the global path. The shared backbone is optimized in both cases. Unless otherwise noted, we set the SDS guidance scale to $30$ for the local path and $40$ for the global path. We use Adam~\cite{Kingma2015AdamAM} with a learning rate of $1\mathrm{e}{-4}$ for the global path and a learning rate of $5\mathrm{e}{-3}$ for the local path. We find it useful to apply augmentations (random crops and perspective transformations) to the rendered videos during training. We further set $\lambda_t = 1.0, \lambda_r = 1\mathrm{e}{-2}, \lambda_s = 5\mathrm{e}{-2}, \lambda_{sh} = 1\mathrm{e}{-1}$. 
For our diffusion backbone, we use ModelScope text-to-video~\cite{wang2023modelscope}, but observe similar results with other backbones (see the supplementary file).
We optimize the networks for $1,000$ steps, taking roughly $30$ minutes per video on a single A100 GPU. In practice, the model often converges after $500$ steps ($15$ minutes). For additional training details, see the supplementary.

\section{Results}
\label{sec:results}
\begin{figure}
    \centering
     \includegraphics[width=0.7\linewidth]{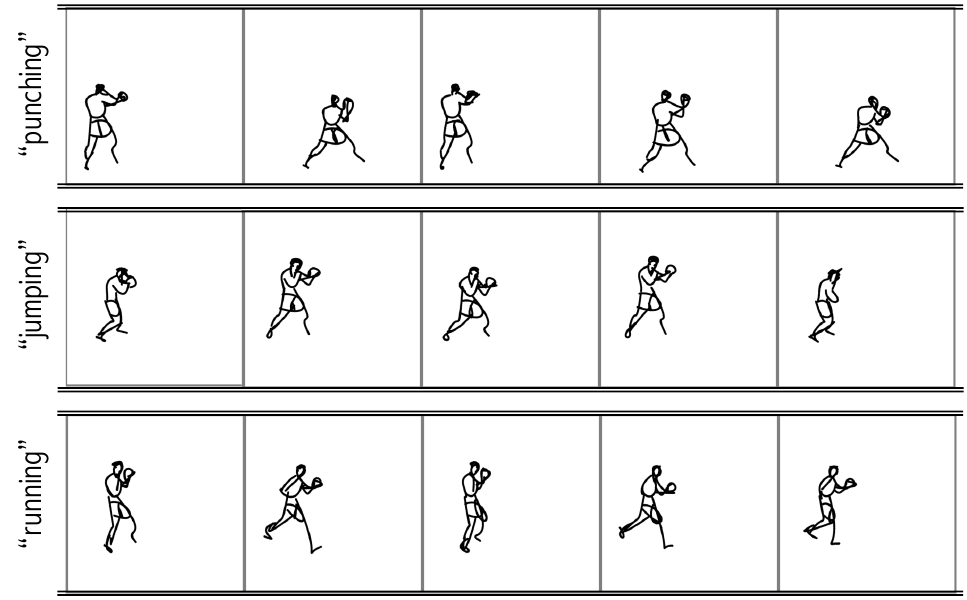}
    \caption{\small Our method can be used to animate the same sketch according to different prompts. These are typically restricted to actions that the portrayed subject would naturally perform. See the supplementary videos for more examples.}
    \label{fig:boxer_prompts}
\end{figure}

We begin by showcasing our method's ability to animate a diverse set of sketches, following an array of text prompts (see \cref{fig:ours_qualitative} and supplementary videos). Our method can capture the delicate swaying of a dolphin in the water, follow a ballerina's dance routine, or mimic the gentle motion of wine swirling in a glass. Notably, it can apply these motions to sketches without any common skeleton or an explicit notation of parts. Moreover, our approach can animate the same sketch using different prompts (see \cref{fig:boxer_prompts}), extending the freedom and diversity of text-to-video models to the more abstract sketch domain. 
Additional examples and full videos can be found in the supplementary materials.

\begin{figure}
    \centering
    \includegraphics[width=0.9\linewidth]{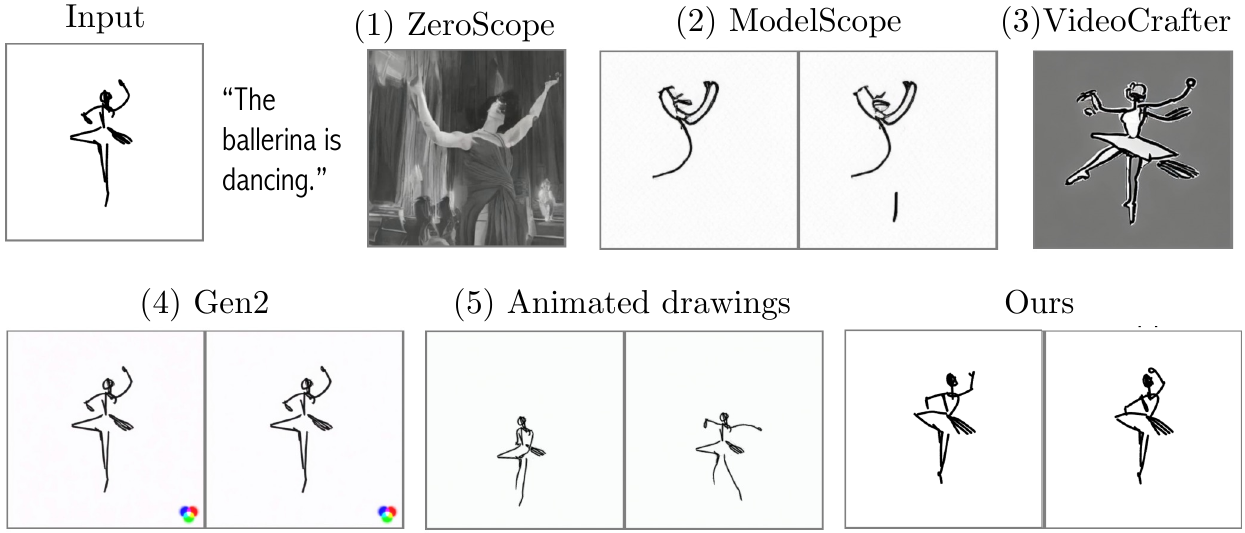}
    \caption{\small Qualitative comparisons. Image-to-video models suffer from artifacts and struggle to preserve the sketch shape (or even remain in a sketch domain). Animated drawings relies on skeletons and pre-captured reference motions. Hence, it cannot generalize to new domains. See the supplementary videos for more examples. }
    \label{fig:image-to-video}
\end{figure}

\subsection{Comparisons}\label{sec:comparison_vidsketch}
As no prior art directly tackles the reference-free sketch animation task, we explore two alternative approaches: pixel-based image-to-video approaches, and skeleton-based methods that build on pre-defined motions.

In the pixel-based scenario, we compare our method with four models: (1) ZeroScope image-to-video~\cite{videofusion2023} which automatically captions the image~\cite{Yu2022CoCaCC} and uses the caption to prompt a text-to-video model. (2) ModelScope~\cite{wang2023modelscope} image-to-video, which is directly conditioned on the image. (3) VideoCrafter~\cite{chen2023videocrafter1} which is conditioned on both the image and the given text prompt, and (4) Gen-2~\cite{gen2runway}, a commercial web-based tool, conditioned on both image and text. 

The results are shown in \cref{fig:image-to-video}. We select representative frames from the output videos. The full videos are available in the supplementary material.

The results of ZeroScope and VideoCrafter show significant artifacts, and commonly fail to even produce a sketch. ModelScope fare batter, but struggle to preserve the shape of the sketch. Gen-2 either struggle to animate the sketch, or transforms it into a real image, depending on the input parameters (see the supplementary videos).

We further compare our approach with a skeleton and reference-based method~\cite{smith2023method} (\cref{fig:image-to-video}, Animated Drawings). This method accounts for the sketch-based nature of our data and can better preserve its shape. However, it requires per-sketch manual annotations and is restricted to a pre-determined set of human motions. Hence, it struggles to animate subjects which cannot be matched to a human skeleton, or whose motion does not align with the presets (see supplementary). In contrast, our method inherits the diversity of the text-to-video model and generalizes to multiple target classes without annotations or explicit references.

We additionally evaluate our method quantitatively. We compare with open methods that require no human intervention and can be evaluated at scale (ZeroScope~\cite{videofusion2023}, ModelScope~\cite{wang2023modelscope}, and Videocrafter~\cite{chen2023videocrafter1}).
We follow CLIPascene \cite{clipascene} and collect sketches spanning three categories: humans, animals, and objects. We asked ChatGPT to randomly select ten instances per category and suggest prompts describing their typical motion.
We used CLIPasso~\cite{vinker2022clipasso} to generate a sketch for each subject. We applied our method and the alternative methods to these sketches and prompts, resulting in 30 animations per method (videos in the supplementary).

\begin{table*}[t]\setlength{\tabcolsep}{3pt}
\footnotesize
\centering 
\begin{tabular}{c c}

\begin{tabular}{lcc} 
    \toprule
    \multirow{2}{*}{Method} & Sketch-to-video & Text-to-Video  \\
     & consistency $\left(\uparrow\right)$ & alignment$\left(\uparrow\right)$ \\
    \midrule 
    ZeroScope & $0.754 \pm 0.009$ & - \\
    ModelScope & $0.779 \pm 0.009$ & - \\
    VideoCrafter & $0.876 \pm 0.007$ & $0.124 \pm 0.005$ \\
    Ours & $\mathbf{0.965} \pm 0.003$ & $\mathbf{0.142} \pm 0.005$ \\
    \bottomrule \\
    \multicolumn{3}{c}{(a) Comparisons to pixel-based approaches}
\end{tabular} &

\begin{tabular}{lcc} 
    \toprule
    \multirow{2}{*}{Setup} 
     & Sketch-to-video & Text-to-Video \\
     & consistency $\left(\uparrow\right)$ & alignment$\left(\uparrow\right)$ \\
    \midrule 
    Full & $0.965 \pm 0.003$ & $0.142 \pm 0.005$ \\
    No Net & $0.926 \pm 0.007$ & $0.142 \pm 0.005$  \\
    No Glob. & $0.936 \pm 0.006$ & $0.140 \pm 0.005$  \\
    No Local & $0.970 \pm 0.002$ & $0.140 \pm 0.004$ \\
    \bottomrule \\
    \multicolumn{3}{c}{(b) Ablation results}
\end{tabular} 

\end{tabular}
\caption{\small Quantitative metrics. \textbf{(a)} CLIP-based consistency and text-video alignment comparisons to open-source image-to-video baselines. \textbf{(b)} The same CLIP-metrics used for an ablation study. }\label{tab:all_quant}
\end{table*}

Following pixel-based methods~\cite{esser2023structure,chen2023videocrafter1}, we use CLIP~\cite{Radfordclip} to measure the \ap{sketch-to-video} consistency, defined as the average cosine similarity between the video's frames and the input sketch used to produce it. 

We further evaluate the alignment between the generated videos and their corresponding prompts (\ap{text-to-video alignment}). We use X-CLIP~\cite{XCLIP}, a model that extends CLIP to video recognition. Here, we compare to the only baseline which is jointly guided by both image and text~\cite{chen2023videocrafter1}. 

All results are provided in \cref{tab:all_quant}a. 
Our method outperforms the baselines on sketch-to-video consistency. In particular, it achieves significant gains over ModelScope whose text-to-video model serves as our prior. Moreover, our approach better aligns with the prompted motion, despite the use of a weaker text-to-video model as a backbone. These results, and in particular the ModelScope scores, demonstrate the importance of the vector representation which assists us in successfully extracting a motion prior without the low quality and artifacts introduced when trying to create sketches in the pixel domain.

\begin{figure}
    \centering
    \includegraphics[width=0.5\linewidth]{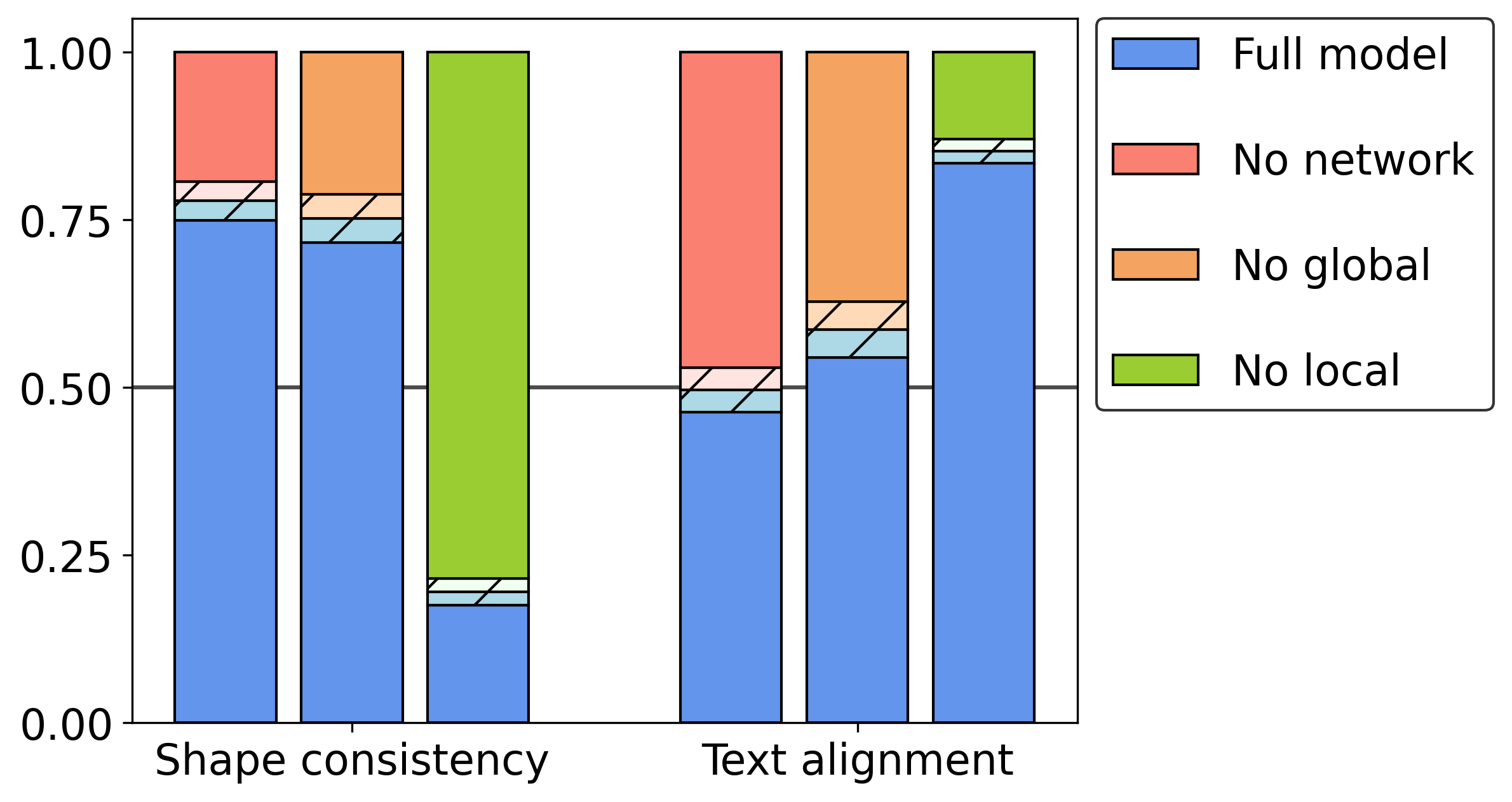}
    \caption{\small User study results. We pit our full model against each ablation setup. The blue bar indicates the percent of responders that preferred our full model over each baseline. Dashed area is one standard error. }
    \label{fig:user_study_sketchvid}
\end{figure}

\begin{figure}
    \centering
    \includegraphics[width=0.9\linewidth]{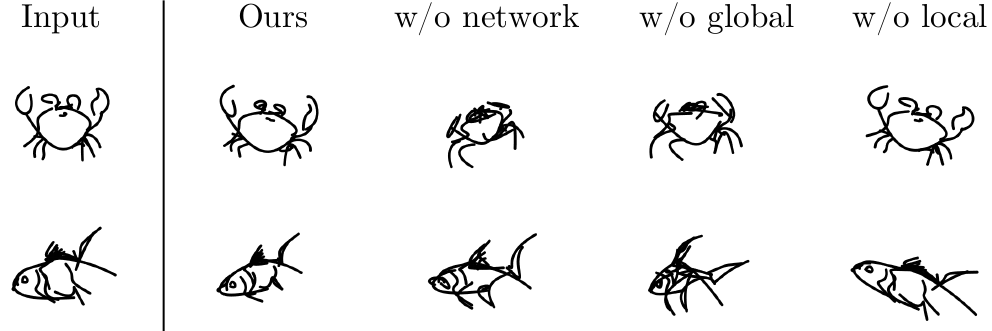}
    \caption{\small Qualitative ablation. Removing the neural network or the global path leads to shape deviations or jittery motion due to the need for higher learning rates (see supplementary videos). Modeling only global movement improves shape consistency, but fails to create realistic motion.}
    \label{fig:ablation_qualitative}
\end{figure}

\subsection{Ablation Study}\label{sec:ablation_videosketch}

We further validate our suggested components through an ablation study. In particular, we evaluate the effect of using the neural prior in place of direct coordinate optimization and the effect of the global-local separation. 

Qualitative results are shown in \cref{fig:ablation_qualitative} (the corresponding videos are provided in the supplementary materials). As can be observed, removing the neural network can lead to increased jitter and harms shape preservation. Removing the global path leads to diminished movement across the frame and less coherent shape transformations. In contrast, removing the local path leads to unrealistic wobbling while keeping the original sketch almost unchanged.

In \cref{tab:all_quant}b, we show quantitative results, following the same protocol as in \cref{sec:comparison_vidsketch}. The sketch-to-video consistency results align with the qualitative observations. However, we observe that the metric for text-to-video alignment \cite{XCLIP} is not sensitive enough to gauge the difference between our ablation setups (standard errors are larger than the gaps). 

We additionally conduct a user study, based on a two-alternative forced-choice setup.
Each user is shown two videos (one output from the full method, and one from a random ablation setup) and asked to select: (1) the video that better preserves the appearance of the initial sketch, and (2) the video that better matches the motion outlined in the prompt. We collected responses from 31 participants over $30$ pairs. The results are provided in \cref{fig:user_study_sketchvid}. 

Users considered the full method's text-to-video alignment to be on-par or better than all ablation setups. When considering sketch-to-video consistency, our method is preferred over both setups that create reasonable motion (no network and no global). Removing the local path leads to higher consistency with respect to the original frame, largely because the sketch remains almost unchanged. Our full method allows for more expressive motion, while still showing remarkable preservation of the input sketch. 

In the supplementary materials, we provide further analysis on the effects of our hyperparameter choices, and highlight an emergent trade-off between shape preservation and the quality of generated motion.

\section{Limitations and Future Work}
\begin{figure}
    \centering
    \includegraphics[width=0.8\linewidth]{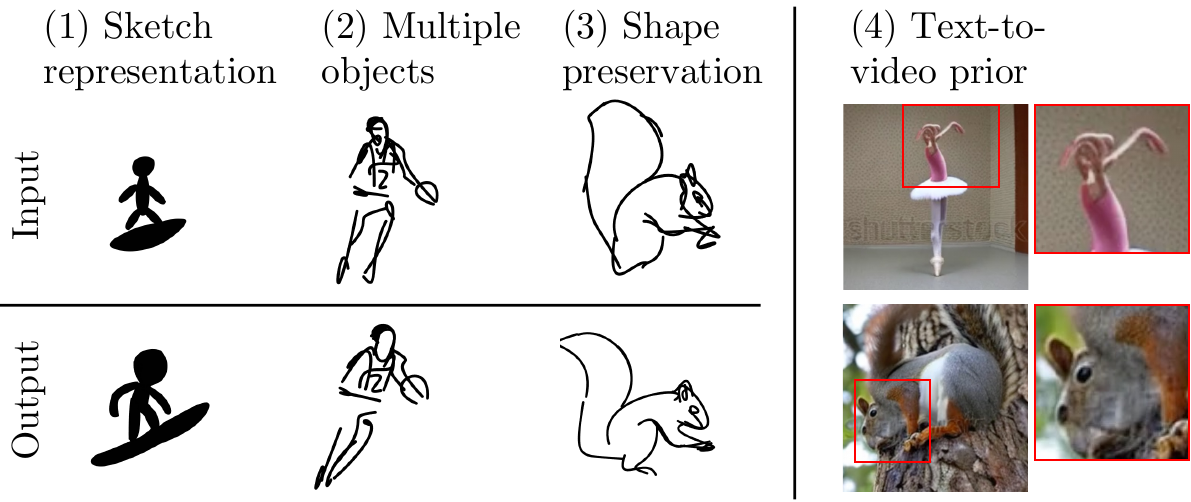}
    \caption{\small Method limitations. The method may struggle with certain sketch representations, fail to tackle multiple objects or complex scenes, or create undesired shape changes. Moreover, it is restricted to motions which the text-to-video prior can create.}
    \label{fig:limitations_vidsketch}
\end{figure}

While our work enables sketch-animation across various classes and prompts, it comes with limitations.
First, we build upon the sketch representation from \cite{vinker2022clipasso}. However, sketches can be represented in many forms with different types of curves and primitive shapes. Using our method with other sketch representations could result in performance degradation. For instance, in \cref{fig:limitations_vidsketch}(1) the surfer's scale has significantly changed. Addressing the diversity of vector sketches requires further development.
Second, our method assumes a single-subject input sketch (a common scenario in character animation techniques). When applied to scene sketches or sketches with multiple objects, we observe reduced result quality due to the design constraints. For example in \cref{fig:limitations_vidsketch}(2), the basketball cannot be separated from the player, contrary to the natural motion of dribbling.

Third, our method faces a trade-off between motion quality and sketch fidelity, and a diligent balance should be achieved between the two. In \cref{fig:limitations_vidsketch}(3), the animated squirrel's appearance differs from the input sketch. This trade-off is further discussed in the supplementary material. 
Potential improvement lies in adopting a mesh-based representation with an approximate rigidity loss~\cite{igarashi2005arapshape}, or by trying to enforce consistency in the diffusion feature space~\cite{tokenflow2023} .

Finally, our approach inherits the limitations of text-to-video priors. Such models are trained on large-scale data, but may be unaware of specific motions, or portray strong biases. For example, as demonstrated in \cref{fig:limitations_vidsketch}, the model we utilize tend to produce significant artefacts when used for text-to-video generation. However, our method is agnostic to the backbone model and hence could likely by used with newer, improved models as they become available, or with personalized models~\cite{gal2022textual} that were augmented with new, unobserved motions.

\section{Conclusions}

We presented a technique to breath life into a given static sketch, following a text prompt. Our method builds on the motion prior captured by powerful text-to-video models. We show that even though these models struggle with generating sketches directly, they can still comprehend such abstract representations in a semantically meaningful way, creating smooth and appealing motions. 
We hope that our work will facilitate further research to provide intuitive and practical tools for sketch animation that incorporate recent advances in text-based video generation.

\part{Generative Models for Visual Inspiration}
\label{part:four}
\chapter{Concept Decomposition for Visual Exploration and Inspiration}
\label{chap:inspiraitontree}
\begin{figure*}[h]
\centering
\includegraphics[width=1\textwidth]{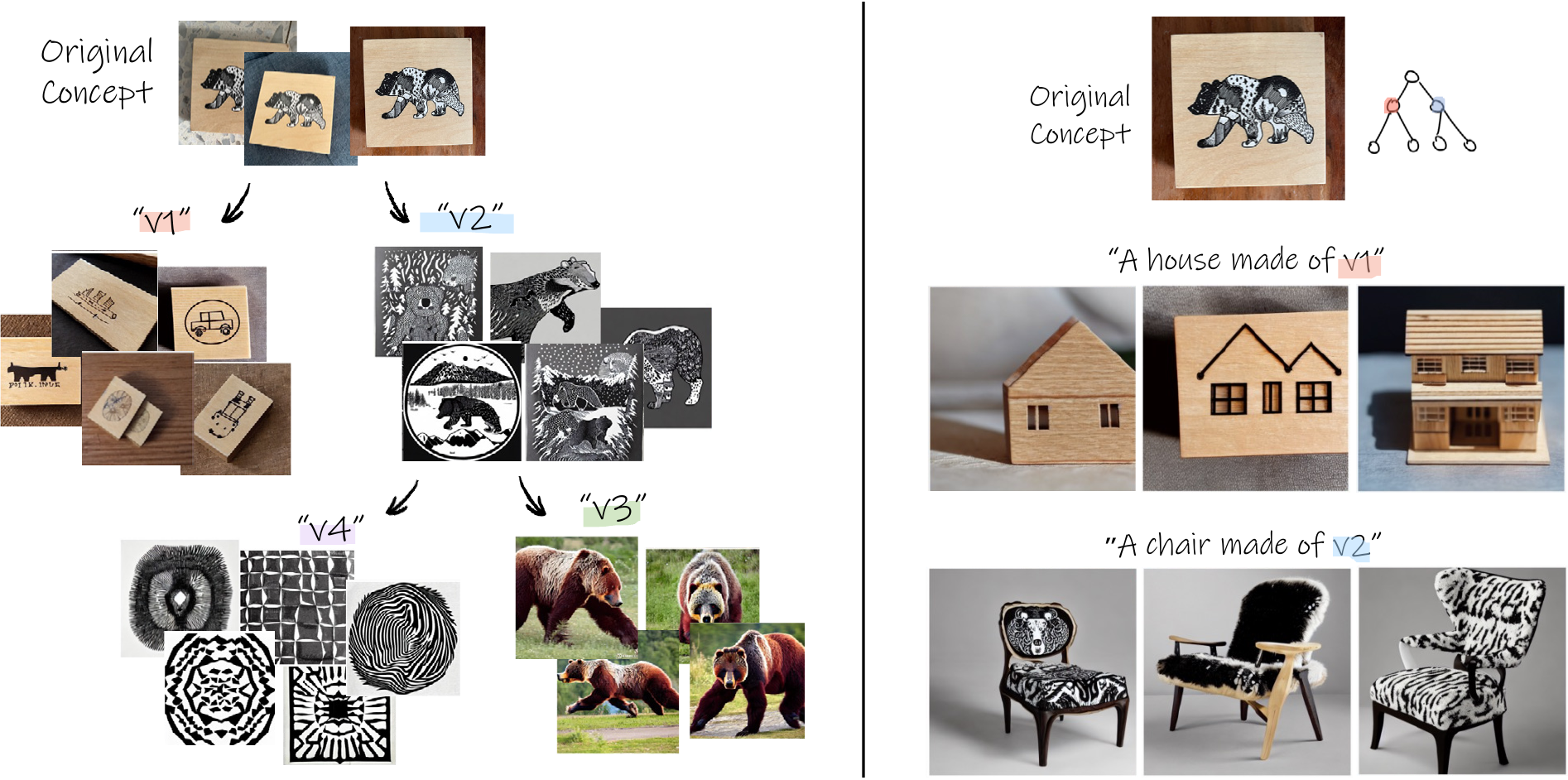}
    \caption[]{\small Left: Our method provides a tree-structured visual exploration space for a given unique concept. The nodes of the tree (\ap{$v_i$}) are newly learned textual vector embeddings, injected to the latent space of a pretrained text-to-image model.
    The nodes encode different \textit{aspects} of the subject of interest. 
    Through examining combinations within and across trees, the different aspects can inspire the creation of new designs and concepts. Right: Combining the learned aspects in natural sentences can be used to produce aspect-based variations.\footnotemark}
    \label{fig:teaser}
\end{figure*}

\begin{wrapfigure}{R}{0.5\linewidth}
    \centering
    \includegraphics[width=1\linewidth]{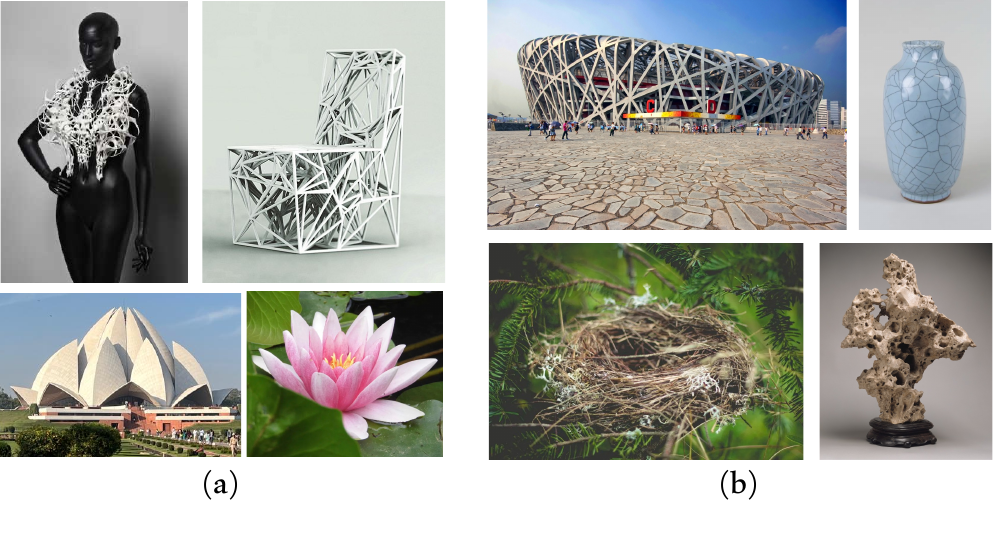}
    \caption{\small Examples of design inspired by visual concepts taken from other concepts.
    (a) top left - fashion design by Iris Van Herpen and Chair by Emmanuel Touraine inspired by nature patterns, bottom left - the Lotus Temple in India, inspired by the lotus flower (b) Beijing National Stadium is inspired by a combination of local Chinese art forms - the crackle glazed pottery that is local to Beijing, and the heavily veined Chinese scholar stones.}
    \vspace{-0.4cm}
    \label{fig:design_inpiration_examples}
\end{wrapfigure}

\footnotetext{Project page: \url{https://inspirationtree.github.io/inspirationtree/}}
Modeling and design are highly creative processes that often require inspiration and exploration~\cite{Gonalves2014WhatID}. 
Designers often draw inspiration from existing visual examples and concepts - either from the real world or using images \cite{henderson1999line, MULLER198912, ECKERT2000523}.
However, rather than simply replicating previous designs, the ability to extract only certain aspects of a given concept is essential to generate original ideas. For example, in \Cref{fig:design_inpiration_examples}a, we illustrate how designers may draw inspiration from patterns and concepts found in nature. 

\newpage
Additionally, by combining multiple aspects from various concepts, designers are often able to create something new. 
For instance, it is described \cite{Beijing_National_Stadium} that the famous Beijing National Stadium, also known as the ``Bird's Nest'', was designed by a group of architects that were inspired by various aspects of different Chinese concepts (see \Cref{fig:design_inpiration_examples}b).
The designers combined aspects of these different concepts -- the shape of a nest, porous Chinese scholar stones, and cracks in glazed pottery art that is local to Beijing, to create an innovative architectural design. 
Such a design process is highly exploratory and often unexpected and surprising.

The questions we tackle in this paper is whether a machine can assist humans in such a highly creative process? Can machines understand different aspects of a given visual concept, and provide inspiration for modeling and design? 
Our work explores the ability of large vision-language models to do just that - express various concepts visually, decompose them into different aspects, and provide almost endless examples that are inspiring and sometimes unexpected. 

We rely on the rich semantic and visual knowledge hidden in large language-vision models.
Recently, these models have been used to perform personalized text-to-image generation \cite{gal2022textual, ruiz2023dreambooth,kumari2022customdiffusion}, demonstrating unprecedented quality of visual concept editing and variation.
We extend the idea presented in \cite{gal2022textual} to allow \textit{aspect-aware} text-to-image generation, which can be used to visually explore new ideas derived from the original visual concept.

Our approach involves (1) decomposing a given visual concept into different aspects, creating a hierarchy of sub-concepts, (2) providing numerous image instances of each learned aspect, and (3) allowing to explore combinations of aspects within the concept and across different concepts.

We model the exploration space using a binary tree, where each node in the tree is a newly learned vector embedding in the textual latent space of a pretrained text-to-image model, representing different aspects of the original visual concept. 
A tree provides an intuitive structure to separate and navigate the different aspects of a given concept. Each level allows to find more aspects of the concepts in the previous level. In addition, each node by itself contains a plethora of samples and can be used for exploration.
For example, in \Cref{fig:teaser}, the original visual concept is first decomposed into its dominant semantic aspects: the wooden saucer in \ap{v1} and the bear drawing in \ap{v2}, next, the bear drawing is further separated into the general concept of a bear in \ap{v3} and its unique texture in \ap{v4}.

\newpage
Given a small set of images depicting the concept of interest as input, we build the tree gradually. For each node, we optimize two child nodes at a time to match the concept depicted in their parent.
We also utilize a CLIP-based \cite{Radfordclip} consistency measurement, to ensure that the concepts depicted in the nodes are coherent and distinct.
The different aspects are learned \textit{implicitly}, without any external constraint regarding the type of separation (such as shape or texture). 
As a result, unexpected concepts can emerge in the process and be used as inspiration for new design ideas.
For example the learned aspects can be integrated into existing concepts by combining them in natural language sentences passed to a pretrained text-to-image model (see \Cref{fig:teaser}, right).
They can also be used to create new concepts by combining different aspects of the same tree (intra-tree combination) or across different trees (inter-tree combination).

We provide many visual results applied to various challenging concepts. We demonstrate the ability of our approach to find different aspects of a given concept, explore and discover new concepts derived from the original one, thereby inspiring the generation of new design ideas.

\section{Related Work}
\paragraph{Design and Modeling Inspiration}
Creativity has been studied in a wide range of fields \cite{Runco2012TheSD,Bonnardel2005TowardsSE,Amabile1996CreativityIC,Elhoseiny2019CreativityIZ,Kantosalo2014FromIT}, and although defining it exactly is difficult, some researchers have suggested that it can be described as the act of evoking and recombinating information from previous knowledge to generate new properties \cite{Bonnardel2005TowardsSE, WILKENFELD200121}.
It is essential, however, to be able to associate ideas in order to generate original ideas rather than just mimicking prior work \cite{Brown2008GUIDINGCD}. Experienced designers and artists are more adept at connecting disparate ideas than novice designers, who need assistance in the evocation process \cite{Bonnardel2005TowardsSE}. By reviewing many exemplars, designers are able to gain a deeper understanding of design spaces and solutions \cite{ECKERT2000523}. 
In the field of human-computer interaction, a number of studies have been conducted to develop tools and software to assist designers in the process of ideation \cite{ImageSense2020,MoodCubes2022,Koch2019MayAD,MetaMap2021,Chilton2019VisiBlendsAF}. 
They are focused on providing better tools for collecting, arranging, and searching visual and textual data, often collected from the web. In contrast, our work focuses on extracting different aspects of a given visual concept and \textit{generating} new images for inspiration.
Our work is close to a line of work on visualizing and exploring design alternatives for geometry \cite{Marks1997DesignGA, Denning2013MeshGitDA, Dobos20123DDA, Matejka2018DreamLE}, including utilizing evolutionary algorithms to inspire users' creativity \cite{CohenOr2015FromIM,Xu2012FitAD,Averkiou2014ShapeSynthPM}.
However, they mostly work in the field of 3D content generation and do not decompose different aspects from existing concepts.

\paragraph{Hierarchical Structure of Images and Language}
Humans are believed to comprehend and interpret intricate visual scenes by breaking them down into hierarchical parts and wholes \cite{Hinton1979SomeDO}.
Substantial research has focused on the hierarchical structure of images, involving capsule networks \cite{Sabour2017DynamicRB, Hinton2021HowTR, Hinton2011}, and-or graphs \cite{Tu2013UnsupervisedSL, Tu2003ImagePU}, as well as scene and object parsing \cite{Chen2014DetectWY, Zhou2017ScenePT, Liang2016SemanticOP, Zhang2016GrowingIP}.
The relationship between the hierarchical nature of language and vision has also been explored in a variety of tasks, including image-text retrieval \cite{Cao2022ImagetextRA, Kiros2014UnifyingVE, Karpathy2014DeepFE}, visual metaconcept learning \cite{Mei2022FALCONFV,Han2020VisualCL}, and visual question answering \cite{Aditya2018ImageUU, Anderson2017BottomUpAT}.
Recently, the field of image generation and editing has undergone unprecedented evolution with the advancement of large language-vision models~\cite{Radfordclip, ramesh2022hierarchical, nichol2021glide, rombach2022highresolution}.
These models have been trained on millions of images and text pairs and have shown to be effective in performing challenging vision related tasks \cite{SegDiff,Avrahami_2022_CVPR,sheynin2022knndiffusion}.
Furthermore, the strong visual and semantic priors of these models have also been demonstrated to be effective for artistic and design tasks \cite{TianEvolution2021,vinker2022clipasso,clipascene,midjourney,Oppenlaender2022TheCO}.
In our work, we hypothesize that the latent space of such models also contain some hierarchical structure and demonstrate how large language-vision models can be used to decompose and transform existing concepts into new ones in order to inspire the development of new ideas.

\paragraph{Personalization}
Personalized text-to-image generation has been introduced recently \cite{gal2022textual,ruiz2023dreambooth,kumari2022customdiffusion,hu2021lora}, with the goal of creating novel scenes based on user provided unique concepts.
In addition to demonstrating unprecedented quality results, these technologies enabled intuitive editing, made design more accessible, and attracted interest even beyond the research community.
We utilize these ideas to facilitate the ideation process of designers and common users, by learning different visual aspects of user-provided concepts.
Current personalization methods either optimize a set of embeddings to describe the concept \cite{gal2022textual}, or modify the denoising network to tie a rarely used word embedding to the new concept \cite{ruiz2023dreambooth}.
While the latter provides more accurate reconstruction and is more robust, it uses much more memory and requires a model for each object.
In this regard, we choose to rely on the approach presented in \cite{gal2022textual}.
It is important to note that our goal is to capture multiple \textit{aspects} of the given concept, and not to improve the accuracy of reconstruction as in \cite{Wei2023ELITEEV,Tewel2023KeyLockedRO,Gal2023DesigningAE,Voynov2023PET,Shi2023InstantBoothPT,Han2023SVDiffCP}.

\section{Method}
Given a small set of images $I^0 = \{I_1^0 ... I_m^0\}$ depicting the desired visual concept, our goal is to construct a rich visual exploration space expressing different aspects of the input concept.
We model the exploration space as a binary tree, whose nodes $V = \{v_1 .. v_n\}$ are learned vector embeddings corresponding to newly discovered words $S = \{s_1 .. s_n\}$ added to the predefined dictionary, representing different aspects of the original concept. 
These newly learned words are used as input to a pretrained text-to-image model \cite{rombach2022highresolution} to generate a rich variety of image examples in each node. 
We find a binary tree to be a suitable choice for our objective, because of the ease of visualization, navigation, and the quality of the sub-concepts depicted in the nodes (see supplemental file for further analysis).

\begin{figure*}
    \centering
    \includegraphics[width=1\textwidth]{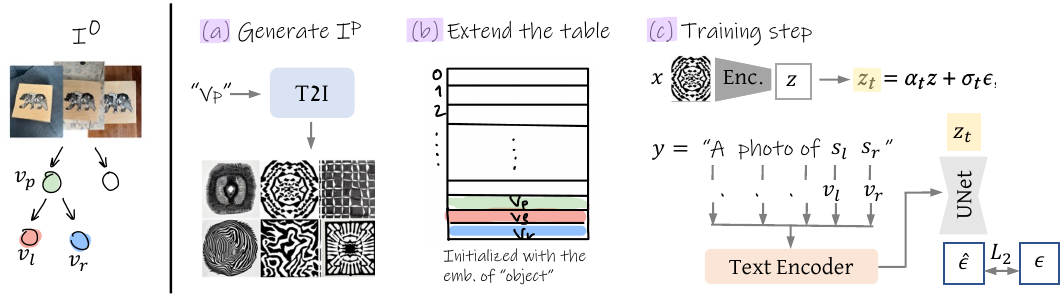}
    \caption{\small High level pipeline of the \ap{binary reconstruction} stage. We optimize two sibling nodes $v_l, v_r$ at a time (marked in red and blue). (a) We first generate a small training set of images $I^p$ depicting the concept in the parent node using a pretrained text-to-image model (T2I). At the root, we use the original set of images $I^0$. (b) We then extend the existing dictionary by adding the two new vectors, initialized with the embedding of the word \ap{object}. (c) Lastly, we optimize $v_l, v_r$ w.r.t. the LDM loss (see details in the text).}
    \label{fig:pipeline1}
\end{figure*}

\subsection{Tree Construction}
\label{sec:tree_construction}
The exploration tree is built gradually as a binary tree from top to bottom, where we iteratively add two new nodes at a time.
To create two child nodes, we optimize new embedding vectors according to the input image-set generated from the concept depicted in the parent node.
During construction, we define two requirements to encourage the learned embeddings to follow the tree structure: (1)~\textbf{Binary Reconstruction} each pair of children nodes together should encapsulate the concept depicted by their parent node, and (2)~\textbf{Coherency} each individual node should depict a coherent concept which is distinct from its sibling. Next, we describe the loss functions and procedures designed to follow these requirements.

\vspace{-0.1cm}
\paragraph{Binary Reconstruction}
We use the reconstruction loss suggested in \cite{gal2022textual}, with some modifications tailored to our goal.
The procedure is illustrated in \Cref{fig:pipeline1} -- in each optimization phase, our goal is to learn two vector embeddings $v_l, v_r$ corresponding to the left and right sibling nodes, whose parent node is marked with $v_p$ (illustrated in \Cref{fig:pipeline1}, left).
We begin with generating a new small training set of images $I^p = \{I_1^p ... I_{10}^p\}$, reflecting the concept depicted by the vector $v_p$ (\Cref{fig:pipeline1}a). At the root, we use the original set of images $I^0$.
Next, we extend the current dictionary by adding two new vector embeddings $v_l, v_r$, corresponding to the right and left children of their parent node $v_p$ (\Cref{fig:pipeline1}b). 
To represent general concepts, the newly added vectors are initialized with the embedding of the word \ap{object}.
At each iteration of optimization (\Cref{fig:pipeline1}c), an image $x$ is sampled from the set $I^p$ and encoded to form the latent image $z = \mathcal{E}(x)$.
A timestep $t$ and a noise $\epsilon$ are also sampled to define the noised latent $z_t = \alpha_t z + \sigma_t \epsilon$ (marked in yellow).
Additionally, a neutral context text $y$ is sampled, containing the new placeholder words in the following form ``A photograph of $s_l$ $s_r$'', to reinforce that optimizing toward such concatenated text prompts ideally will capture naturally different concepts for the left and right nodes.

\begin{figure}[t]
    \centering
    \includegraphics[width=0.7\linewidth]{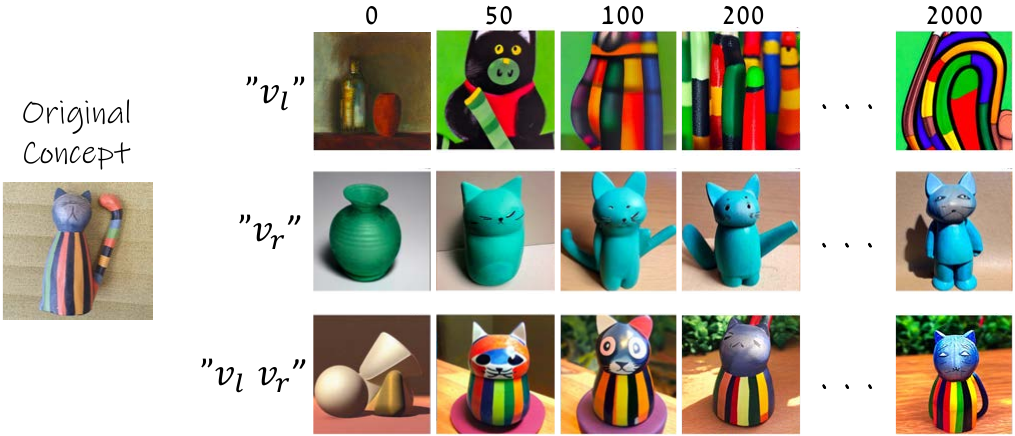}
    \caption{\small Optimization iterations. The embedding of both children nodes $v_l, v_r$ are initilized with the word ``object''. During iterations, they gradually depict two aspects of the original concept. Note that using both embedding together reconstructs the original parent concept.}
    \label{fig:iter_example}
\end{figure}

The noised latent $z_t$ is fed to a pretrained Stable Diffusion UNet model $\epsilon_\theta$, conditioned on the CLIP embedding $c(y)$ of the sampled text, to predict the noise $\epsilon$.
The prediction loss is backpropagated w.r.t. the vector embeddings $v_l, v_r$:
\begin{equation}
   \{v_l, v_r\} = \argmin_v \mathbb{E}_{z\sim\mathcal{E}(x), y, \epsilon \sim \mathcal{N}(0, 1), t }\Big[ \Vert \epsilon - \epsilon_\theta(z_{t},t, c(y)) \Vert_{2}^{2}\Big] .
    \label{eq:v_opt}
\end{equation}
\newpage
This procedure encourages $v_l, v_r$ together to express the visual concept of their parent depicted in the set $I^p$.
\Cref{fig:iter_example} illustrates how the two embeddings begin by representing the word \ap{object}, and gradually converge to depict two aspects of the input concept. 
We hypothesize that the
nodes do not converge to a similar concept because of the model's training on natural sentences, where consecutive identical words are uncommon.

We use the timestep sampling approach proposed in ReVersion \cite{huang2023reversion}, which skews the sampling distribution so that a larger $t$ is assigned a higher probability, according to the following importance sampling function:
\begin{equation}
   f(t) = \frac{1}{T}(1 - \alpha \cos \frac{\pi t}{T}).
\end{equation}
We set $\alpha = 0.5$. We find that this sampling approach improves stability and content separation. This choice is further discussed in the supplementary file.

\begin{figure}
    \centering
    \includegraphics[width=0.7\linewidth]{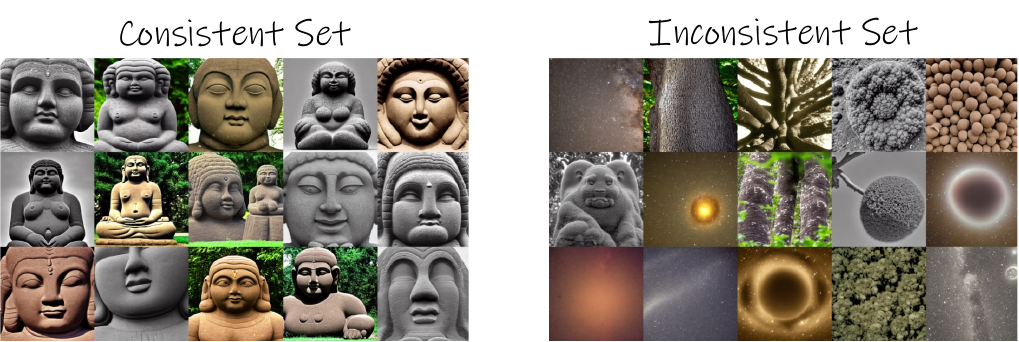}
    \caption{\small We demonstrate two sets of random images generated from two different vector embeddings. An example of a consistent set can be seen on the left, where the concept depicted in the node is clear. We show an inconsistent set on the right, where images appear to depict multiple concepts.}
    \label{fig:consistency_buddha}
\end{figure}

\begin{wrapfigure}{R}{0.5\linewidth}
    \centering
    \includegraphics[width=1\linewidth]{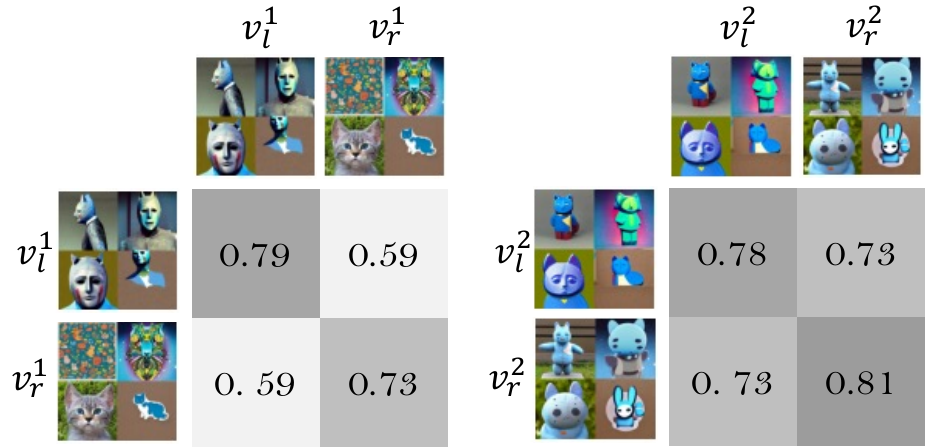}
    \caption{\small Consistency scores matrix between image sample sets of nodes. The seed selection process favors pairs of siblings that have a high consistency score within themselves, and low consistency score between each other. In this example, the left pair is better than the right.}
    \label{fig:coherency_matrix}
\end{wrapfigure}

\paragraph{Coherency}
The resulting pair of embeddings described above together often capture the parent concept depicted in the original images well.
However, the images produced by each embedding individually may not always reflect a logical sub-concept that is coherent to the observer.

We find that such incoherent embeddings are frequently characterized by inconsistent appearance of the images generated from them, i.e., it can be difficult to identify a common concept behind them. For example, in \Cref{fig:consistency_buddha} the concept depicted in the set on the right is not clear, compared to the set of images on the left. 

This issue may be related to the observation that textual inversion often results in vector embedding outside of the distribution of common words in the dictionary, affecting editability as well \cite{Voynov2023PET}.
It is thus possible that embeddings that are highly unusual may not behave as ``real words'', thereby producing incoherent visual concepts.
In addition, textual-inversion based methods are sometimes unstable and depend on the seed and iteration selection.

To overcome this issue we define a consistency test, which allows us to filter out incoherent embeddings. We begin by running the procedure described above to find $v_l, v_r$ using $k$ different seeds in parallel for a sufficient number of steps (in our experiments we found that k=4 and 200 steps are sufficient since at that point the embeddings have already progressed far enough from their initialization word ``object'' as seen in \Cref{fig:iter_example}).

This gives us an initial set of $k$ pairs of vector embeddings $V_{s} = \{v_l^i, v_r^i\}_{i=1}^{k}$.
For each vector $v \in V_s$ we generate a random set $I^v$ of 40 images using our pre-trained text-to-image model.
We then use a pretrained CLIP Image encoder \cite{Radfordclip}, to produce the embedding $CLIP(I_i^v)$ of each image in the set.

We define the consistency of two sets of images $I^a, I^b$ as follows:

\begin{equation} \label{eq:const}
    \mathcal{C}(I^{a}, I^{b}) = mean_{I_i^{a}\in{I^{a}} ,I_j^{b}\in{I^{b}, I_i^{a} \neq I_j^{b}} } (sim(CLIP(I_i^{a}), CLIP(I_j^{b}))).
\end{equation}

Note that $|\mathcal{C}(I^{a}, I^{b})| \leq 1$ because $sim(x,y) = \frac{x\cdot y}{||x||\cdot||y||}$ is the cosine similarity between a pair of CLIP embedding of two different images. This formulation is motivated by the observation that if a set of images depicts a certain semantic concept, their vector embedding in CLIP's latent space should be relatively close to each other. 
Ideally, we are looking for pairs in which each node is coherent by itself, and in addition, two sibling nodes are distinct from each other.
We therefore choose the pair of tokens $\{v_l^*, v_r^*\}\in V_{s}$ as follows:
\begin{equation} \label{eq:consistency}
    \{v_l^*, v_r^*\} = \argmax_{\{v_l^i, v_r^i\} \in V_s} \big[ C_l^i + C_r^i + 
     (min(C_l^i, C_r^i) - \mathcal{C}(I^{v_l^i}, I^{v_r^i})) \big],
\end{equation}
where $ C_l^i = \mathcal{C} (I^{v_l^i}, I^{v_l^i}), C_r^i = \mathcal{C}(I^{v_r^i}, I^{v_r^i})$.
Note that we do not consider the absolute cross consistency score $\mathcal{C}(I^{v_l^i}, I^{v_r^i})$, but we compute its relative difference from the node with the minimum consistency.
We demonstrate this procedure in \Cref{fig:coherency_matrix}.
We optimized two pairs of sibling nodes $\{v_l^1, v_r^1\}, \{v_l^2, v_r^2\}$ using two seeds, w.r.t. the same parent node. Each matrix illustrates the consistency scores $C_l^i, \mathcal{C}(I^{v_l^i}, I^{v_r^i}), C_r^i$ obtained for the sets of images of each seed.
In both cases, the scores on the diagonal are high, which indicates that each set is consistent within itself.
While the sets on the right obtained a higher consistency score within each node, they also obtained a relatively high score across the nodes (0.73), which means they are not distinct enough.

After selecting the optimal seed, we continue the optimization of the chosen vector pair w.r.t. the reconstruction loss in \Cref{eq:v_opt} for $1500$ iterations.

\begin{figure}
    \centering
    \includegraphics[width=0.97\linewidth]{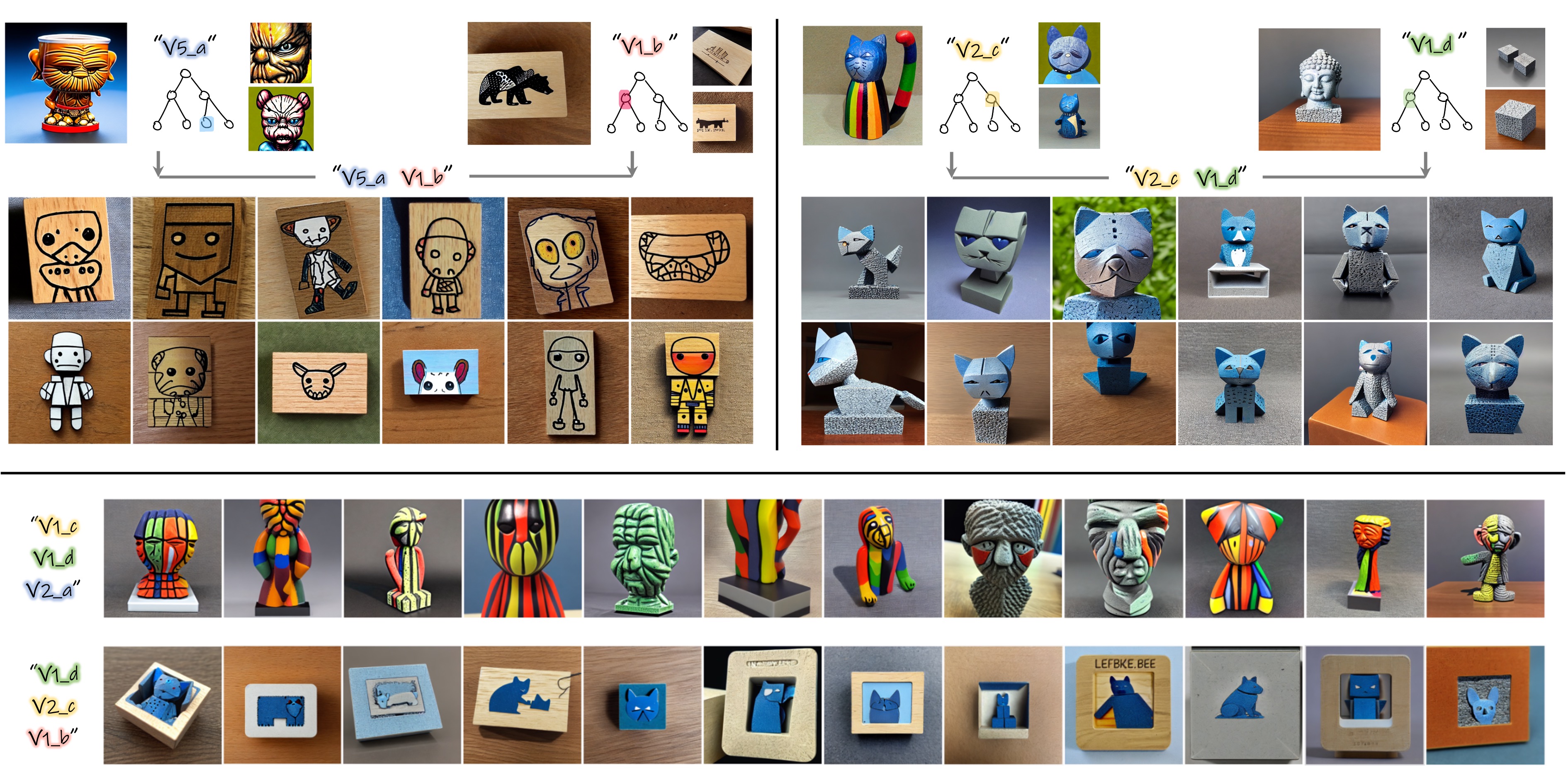}
    \caption{\small Examples of inter-tree combinations. We use our method to produce trees for the four concepts depicted in the first row. We then combine aspects from different trees to generate a set of inter-tree combinations (the chosen aspects are shown next to each concept). We also show combinations of three aspects from different trees at the bottom.}
    \label{fig:mug_bear_comb}
\end{figure}

\section{Results}
In \Cref{fig:teaser,fig:cat_tree,fig:red_teapot}, we show examples of possible trees. 
For each node in the tree, we use its corresponding placeholder word as an input to a pretrained text-to-image model \cite{rombach2022highresolution}, to generate a set of random images.
These images have been generated without any prompt engineering or additional words within the sentence, except for the word itself. For clarity, we use the notion $"v"$ next to each set of images, illustrating that the presented set depicts the concept learned in that node.
As can be seen, the learned embeddings in each node capture different elements of the original concept, such as the concept of a cat and a sculpture, as well as the unique texture in \Cref{fig:cat_tree}. 
The sub-concepts captured in the nodes follow the tree's structure, where the concepts are decomposed gradually, with two sibling nodes decomposing their parent node.
This decomposition is done \textit{implicitly}, without external guidance regarding the split theme.
For many more trees please see our supplementary file. 

\subsection{Applications}
The constructed tree provides a rich visual exploration space for concepts related to the object of interest.
In this section we demonstrate how this space can be used for novel combination and exploration.

\paragraph{Intra-tree combination}
The generated tree is represented via the set of optimized vectors $V = \{v_1 .. v_n\}$. Once this set is learned we can use it to perform further exploration and conceptual editing \textit{within} the object's ``inner world''.
We can explore combinations of different aspects by composing sentences containing different subsets of $V$.
For example, in the bottom left area of \Cref{fig:cat_tree}, we have combined $v_1$ and $v_5$, which resulted in a variation of the original sculpture without the sub-concept relating to the cat (depicted in $v_6$). At the bottom right, we have excluded the sub-concept depicted in $v_5$ (related to a blue sculpture), which resulted in a new representation of a flat cat with the body and texture of the original object.
Such combinations can provide new perspectives on the original concept and inspiration that highlights only specific aspects.

\begin{figure}
    \centering
    \includegraphics[width=0.7\linewidth]{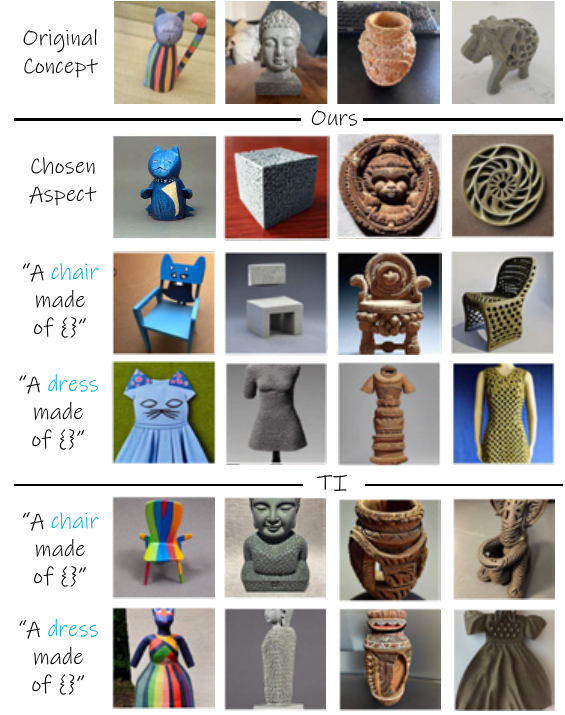}
    \caption{\small Combining the learned aspects in natural sentences to produce aspect-based variations.
    The original concepts are shown at the top. In the third and fourth rows are our text-based generation results applied with the aspects depicted in the second row. Under \ap{TI} we show image generation for the concepts in the first row (without our aspect decomposition approach), produced using \cite{gal2022textual}.}
    \label{fig:text_based_multiple}
\end{figure}

\paragraph{Inter-tree combination}
It is also possible to combine concepts learned across different trees, since we only inject new words into the existing dictionary, and do not fine-tune the model's weights as in other personalization approaches \cite{ruiz2023dreambooth}.
To achieve this, we first build the trees independently for each concept and then visualize the sub-concepts depicted in the nodes to select interesting combinations.
In \Cref{fig:mug_bear_comb} the generated original concepts are shown on top, along with an illustration of the concepts depicted in the relevant nodes.
To combine the concepts across the trees, we simply place the two placeholder words together in a sentence and feed it into the pretrained text-to-image model.
As can be seen, on the left the concept of a \ap{saucer with a drawing} and the \ap{creature} from the mug are combined to create many creative and surprising combinations of the two.
On the right, the blue sculpture of a cat is combined with the stone depicted at the bottom of the Buddha, which together create new sculptures in which the Buddha is replaced with the cat.
\paragraph{Text-based generation}
the placeholder words of the learned embeddings can be composed into natural language sentences to generate various scenes based on the learned aspects.
We illustrate this at the top of \Cref{fig:text_based_multiple}, where we integrate the learned aspects of the original concepts in new designs (in this case of a chair and a dress).
At the bottom of \Cref{fig:text_based_multiple}, we show the effect of using the learned vectors of the original concepts instead of specific aspects. We apply Textual Inversion (TI) \cite{gal2022textual} with the default hyperparameters to fit a new word depicting each concept, and choose a representative result.
The results suggest that without aspect decomposition, generation can be quite limited.
For instance, in the first column, both the dress and the chair are dominated by the texture of the sculpture, whereas the concept of a blue cat is almost ignored. Furthermore, TI may exclude the main object of the sentence (second and third columns), or the results may capture all aspects of the object (fourth column), thereby narrowing the exploration space.

\subsection{Evaluations}
\paragraph{Consistency Score Validation.}
We first show that our consistency test proposed in \Cref{eq:const} aligns well with human perception of consistency.
We conducted a perceptual study with $35$ participants in which we presented $15$ pairs of random image sets depicting sub-concepts of $9$ objects.
We asked participants to determine which of the sets is more consistent within itself in terms of the concept it depicts (an example of such a pair can be seen in \Cref{fig:consistency_buddha}).
We also measured the consistency scores for these sets using our CLIP-based approach, and compared the results. The CLIP-based scores matched the human choices in $82.3\%$ (stdv: $1\%$) of the cases.

\vspace{-0.2cm}
\paragraph{Reconstruction and Separation.} We quantitatively evaluate our method's ability to follow the tree requirements of reconstruction and sub-concept separation. 
We collected a set of $13$ concepts ($9$ from existing personalization datasets \cite{gal2022textual,kumari2022customdiffusion}, and $4$ new concepts from our dataset), and generated $13$ corresponding trees.
Note that we chose concepts that are complex enough and have the potential to be divided into different aspects (we discuss this in the limitations section).
For each pair of sibling nodes $v_l, v_r$ and their parent node $v_p$, we produced their corresponding sets of images -- $I^{v_l}, I^{v_r}, I^{v_p}$ (where for nodes in the first level we used the original set of images $I^0$ as $I^{v_p}$). We additionally produced the set $I^{v_l v_r}$, depicting the joint concept learned by two sibling nodes.

We first compute $\mathcal{C}(I^{v_p}, I^{v_l v_r})$ to measure the quality of reconstruction, i.e., that two sibling nodes together represent the concept depicted in their parent node. The average score obtained for this measurement is $0.8$, which suggests that on average, the concept depicted by the children nodes together is consistent with that of their parent node.
Second, we measure if two sibling nodes depict distinct concepts by using $\mathcal{C}(I^{v_l}, I^{v_r})$. The average score obtained was $0.59$, indicating there is larger separation between siblings, but they are still close.

\paragraph{Aspects Relevancy.}
We assess the ability of our method to encode different aspects connected to the input concept via a perceptual study.
We chose 5 objects from the dataset above, and 3 random aspects for each object.
We presented participants with a random set of images depicting one aspect of one object at a time. We asked the participants to choose the object they believe this aspect originated from, along with the option `none'. In total we collected answers from $35$ participants, and achieved recognition rates of $87.8\%$ (stdv: $1\%$). 
These evaluations demonstrate that our method can indeed separate a concept into \emph{relevant} aspects, where each new sub-concept is \emph{coherent}, and the binary tree structure is valid - i.e., the combination of two children can \emph{reconstruct} the parent concept.

\newpage
\subsection{Ablation}
\paragraph{Binary Tree}
Our choice to use a binary tree stems from two main reasons: (1) complexity, and (2) consistency.

Our method allows to build a tree with more than two children per node, however, this can add redundant complexity to the method. For example, using three children nodes, after only two levels we will get 12 aspects, which may be difficult to visualize and navigate. In addition, the use of more than two children will result in a longer running time in each level since we will have to split more nodes.

\begin{wrapfigure}{R}{0.5\textwidth}
    \centering
    \includegraphics[width=0.9\linewidth]{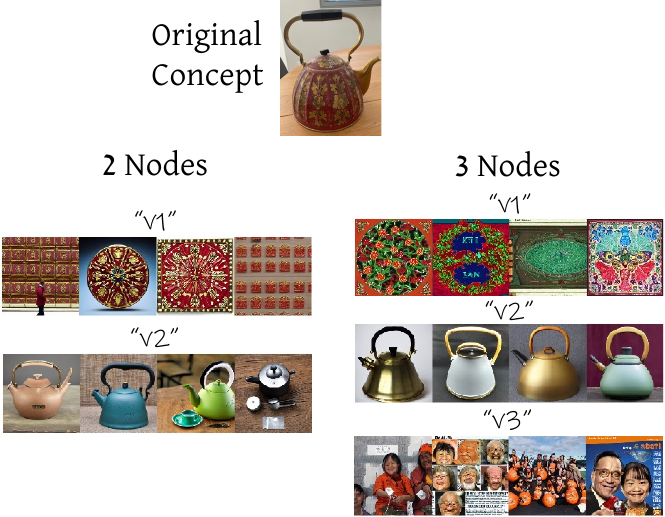}
    \caption{Comparison of optimizing for two child nodes (left) v.s. three child nodes (right). Using three nodes increases the chance of arriving at inconsistent or irrelevant concepts.}
    \label{fig:number_of_nodes}
\end{wrapfigure}

In terms of consistency, we observe that when optimizing more than two nodes at a time, the chance of receiving inconsistent nodes increases.
Often, two nodes will be consistent, and the third node is inconsistent or may depict irrelevant concepts such as background.
We demonstrate this in \Cref{fig:number_of_nodes}, on the \ap{red teapot} object.
We present the aspects obtained from the optimal seed after 200 iterations, for the case of two nodes (left) and for the case of three nodes (right). As shown, the sub-concept in $v_3$ for the 3 nodes optimization does not appear to be consistent or comprehensible, and therefore is not useful in achieving our goal of extracting aspects from the parent concept. 

The following quantitative experiment further confirms this observation.
We obtained 52 trees for our set of 13 objects (using four seeds for each object as described in the main paper). Each tree is a 3-node tree with one level, resulting in a total number of 156 nodes. We measured our CLIP-based consistency test on each node to determine its average consistency score. Next, for each tree, we sorted the 3 nodes $\{v_1, v_2, v_3\}$ according to their consistency score, from the most consistent ($v_1$) to the least consistent ($v_3$).
We then average the scores of $\{v_1, v_2, v_3\}$ across all trees, and received the final scores of: $0.804, 0.742, 0.633$ for $\{v_1, v_2, v_3\}$ correspondingly. The noticeable consistency gap between the top 2 nodes and the third node indicates that, on average, two of the three nodes are consistent, while the last may contain incoherent information.
This experiment correlates well with our visual observation (as demonstrated in \Cref{fig:number_of_nodes}).

\begin{figure}[h]
    \centering
    \includegraphics[width=0.8\linewidth]{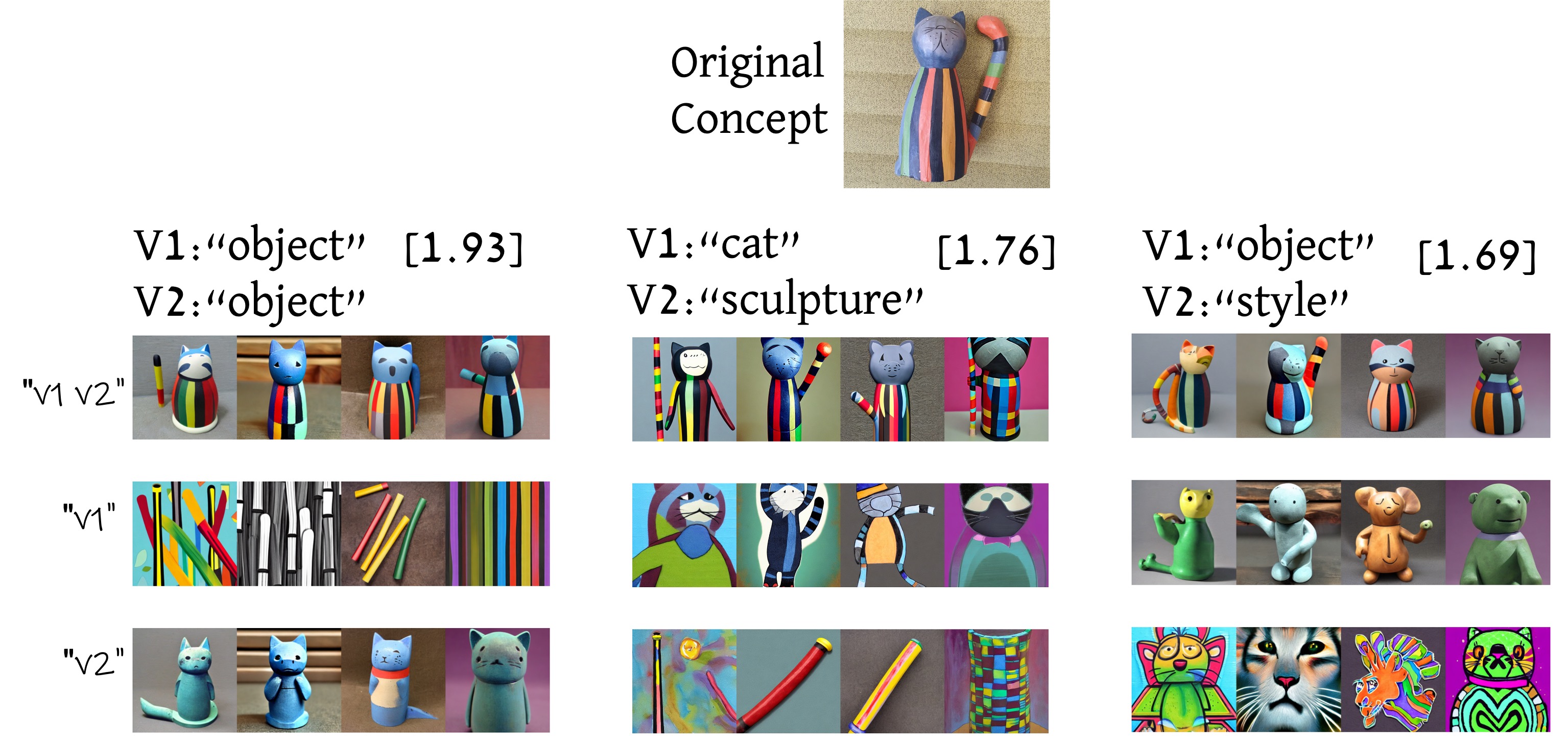}
    \caption{Comparing different initialization approaches. The columns show the different initialization used, and the rows show the results after 200 iterations.}
    \label{fig:init_user_input}
\end{figure}

\subsection{Vectors Initialization}
As mentioned in \Cref{sec:tree_construction}, we use the embedding of the word \ap{object} to initialize the new vectors $v_l, v_r$.
In choosing the word \ap{object} as a generic concept, we eliminate the requirement for user-defined input specific to the concept. Furthermore, the use of a most general concept for initialization allows for more unexpected decompositions to occur.

It is possible, however, for the user to select other words for initialization if they wish to encourage the generation of certain sub-aspects.
In \Cref{fig:init_user_input}, we demonstrate the impact of using different initialization approaches.
The first column illustrates the baseline results, using "object" to initialize both nodes. 
The second column displays the results of initialization based on potential user defined inputs -- a \ap{cat} and a \ap{sculpture} for the left and right nodes, respectively. In the third column, we present an alternative generic initialization option consisting of \ap{object} and \ap{style}.
Each experiment was conducted using four different seeds and $200$ iterations, with similar results.

As can be seen in the first column the results we get are consistent, with a maximum consistency score of $1.93$.
In the second column, we can see that v1 represents the cat aspect and v2 represents a concept that is more akin to a sculpture, with a consistency score of $1.76$.
The third column indicates that using "style" instead of "object" negatively impacted the results, resulting in a consistency score of $1.69$. More examples can be found in the supplemental file.

\section{Limitations and Future Work}
\begin{figure}
    \centering
    \includegraphics[width=0.9\linewidth]{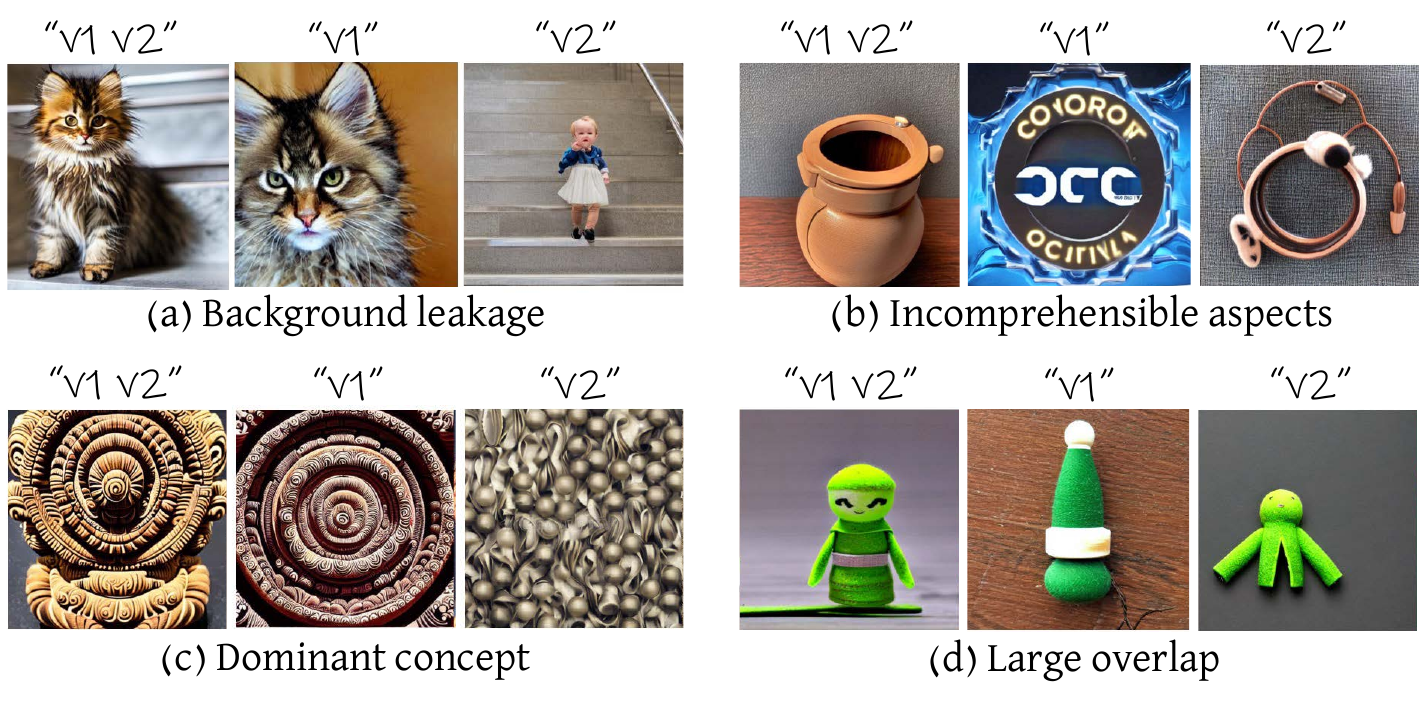}
    \caption{\small We demonstrate four general cases of decomposition failure.}
    \label{fig:limitations1}
\end{figure}

\begin{figure*}
    \centering
    \includegraphics[width=1\textwidth]{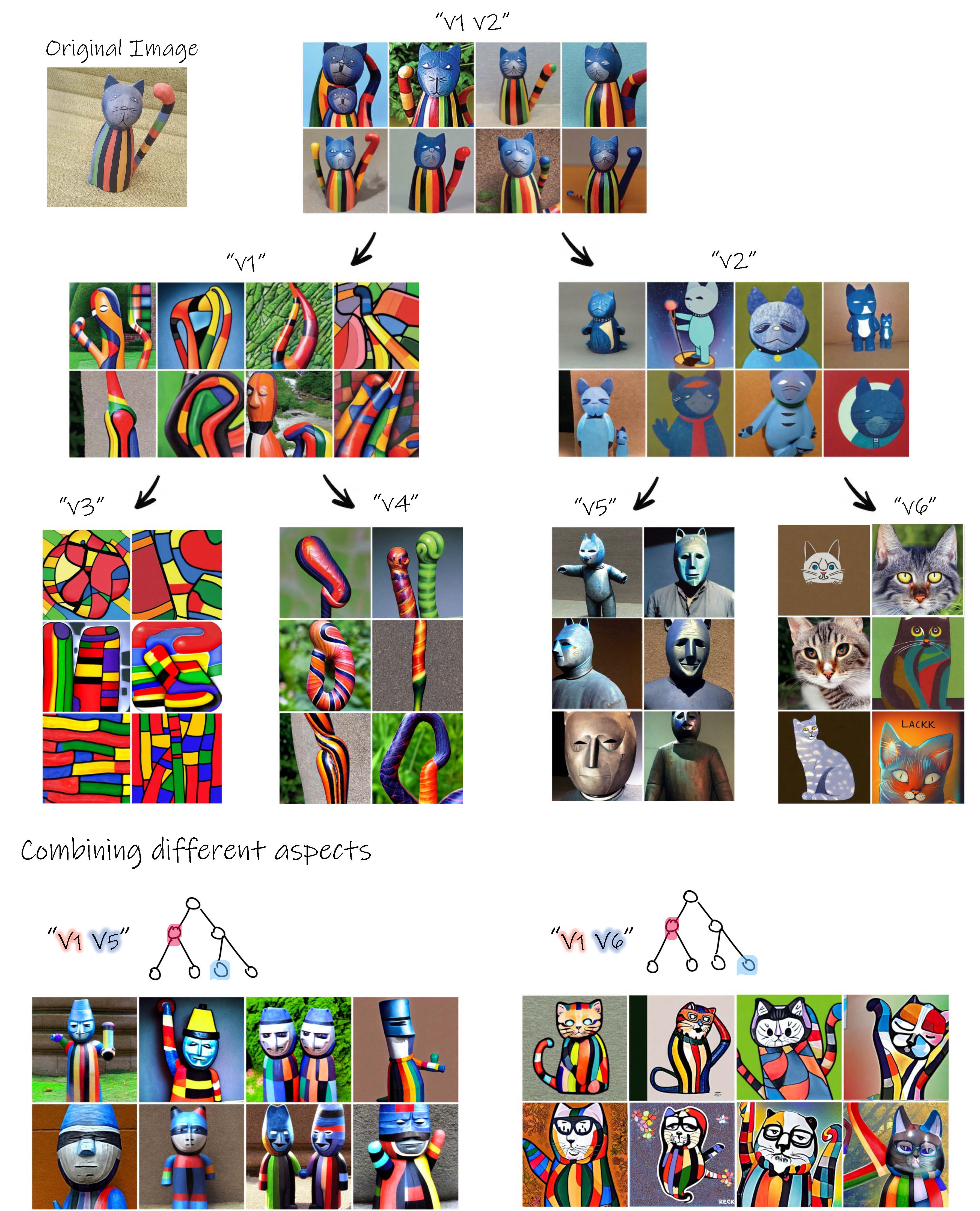}
    \caption{\small Exploration tree for the cat sculpture. At the bottom we show examples of possible intra-tree combinations.}
    \label{fig:cat_tree}
\end{figure*}

\begin{figure*}
    \centering
    \includegraphics[width=1\textwidth]{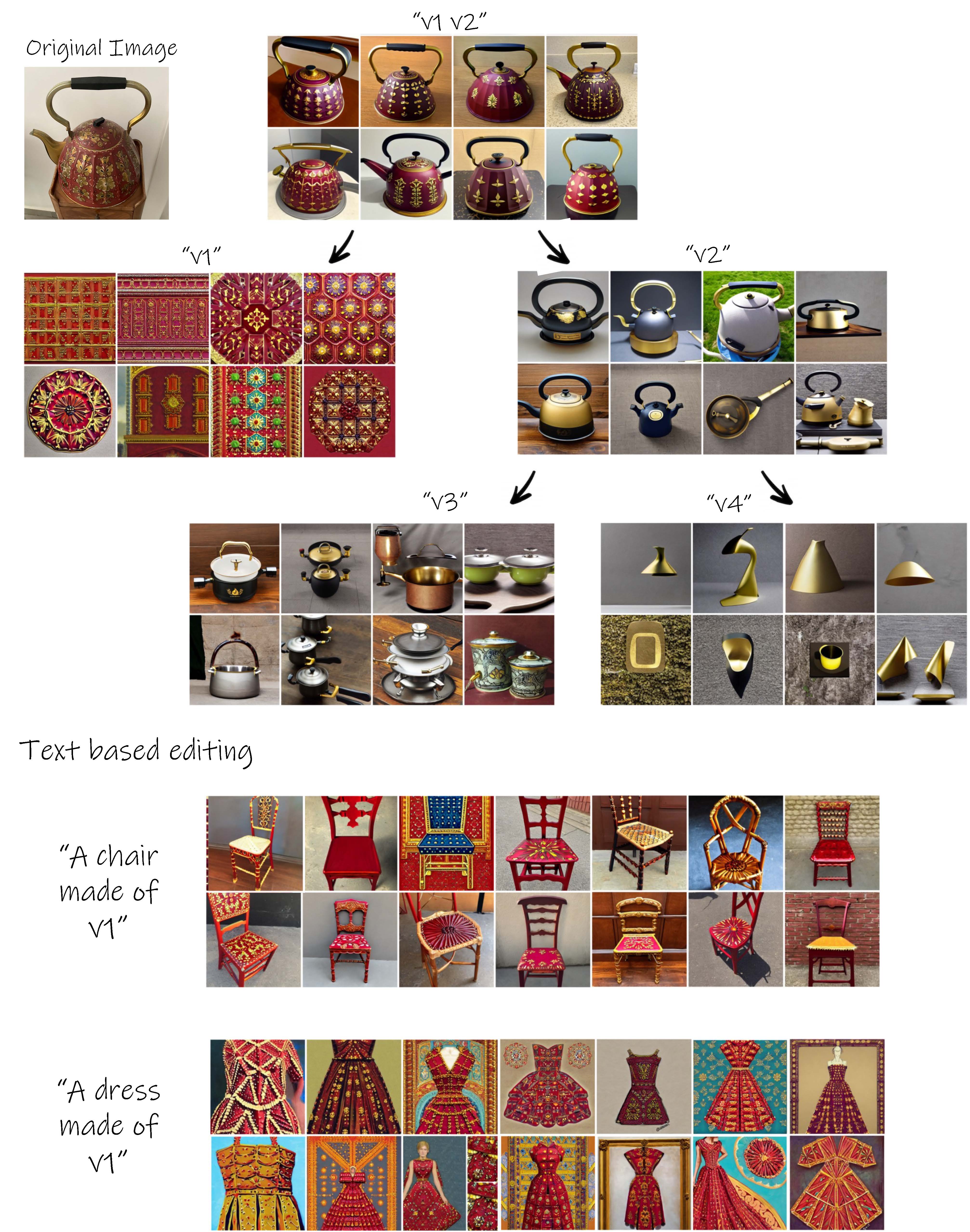}
    \caption{\small Exploration tree for a decorated teapot. At the bottom we show examples of possible text-based generation.}
    \label{fig:red_teapot}
\end{figure*}

Our method may fail to decompose an input concept. We divide the failure cases into four general categories illustrated in \Cref{fig:limitations1}:

\noindent (1) Background leakage - the training images should be taken from different perspectives and with varying backgrounds (this requirement also exists in \cite{gal2022textual}). When images do not meet these criteria, one of the sibling nodes often captures information from the background instead of the object itself. 

\noindent (2) Incomprehensible aspects - some separations may not satisfy clear, interesting, aesthetic, or inspiring aspects, even when the coherency principle holds.

\noindent (3) Dominant sub-concept - we illustrate this in \Cref{fig:limitations1}c, where we show a split on the second level of the concept depicted under \ap{$v_1 v_2$}. As shown, v1 has dominated the information, so even if the coherency term is held, decomposition to two sub-concepts has not really been achieved.

\noindent (4) Large overlap when two aspects share information -- we illustrate this in \Cref{fig:limitations1}d, which is a split of the second level, where both concepts depicted in v1 and v2 appear to share too similar.

We hope that such limitations could be resolved in the future using additional regularization terms in the optimization process or through the development of more robust personalization methods. 

Additionally, our method can have difficulties to create deeper trees. 
Our observations show two main factors influencing whether a node could be further split – the complexity of the concept depicted in the node and its’ coherency. As we go deeper into the tree, the concepts become simpler and more challenging to decompose. 
When a concept reaches one of these conditions we stop the tree growth. This opens interesting avenues for future research to explore how can concept trees be further extended.

Currently the time for decomposing a node can reach up to approximately $40$ minutes on a single A100 GPU. However, as textual inversion optimization techniques will progress, so will our method.

\section{Conclusions}
We presented a method to implicitly decompose a given visual concept into various aspects to construct an inspiring visual exploration space.
Our method can be used to generate numerous representations and variations of a certain subject, to combine aspects across objects, as well as to use these aspects as part of natural language sentences that drive visual generation of novel concepts.

The aspects are learned implicitly, without external guidance regarding the type of separation.
This implicit approach also provides another small step in revealing the rich latent space of large vision-language models, allowing surprising and creative representations to be produced.
We demonstrated the effectiveness of our method on a variety of challenging concepts. We hope our work will open the door to further research aimed at developing and improving existing tools to assist and inspire designers and artists.

\begin{figure*}
    \centering
    \includegraphics[width=1\textwidth]{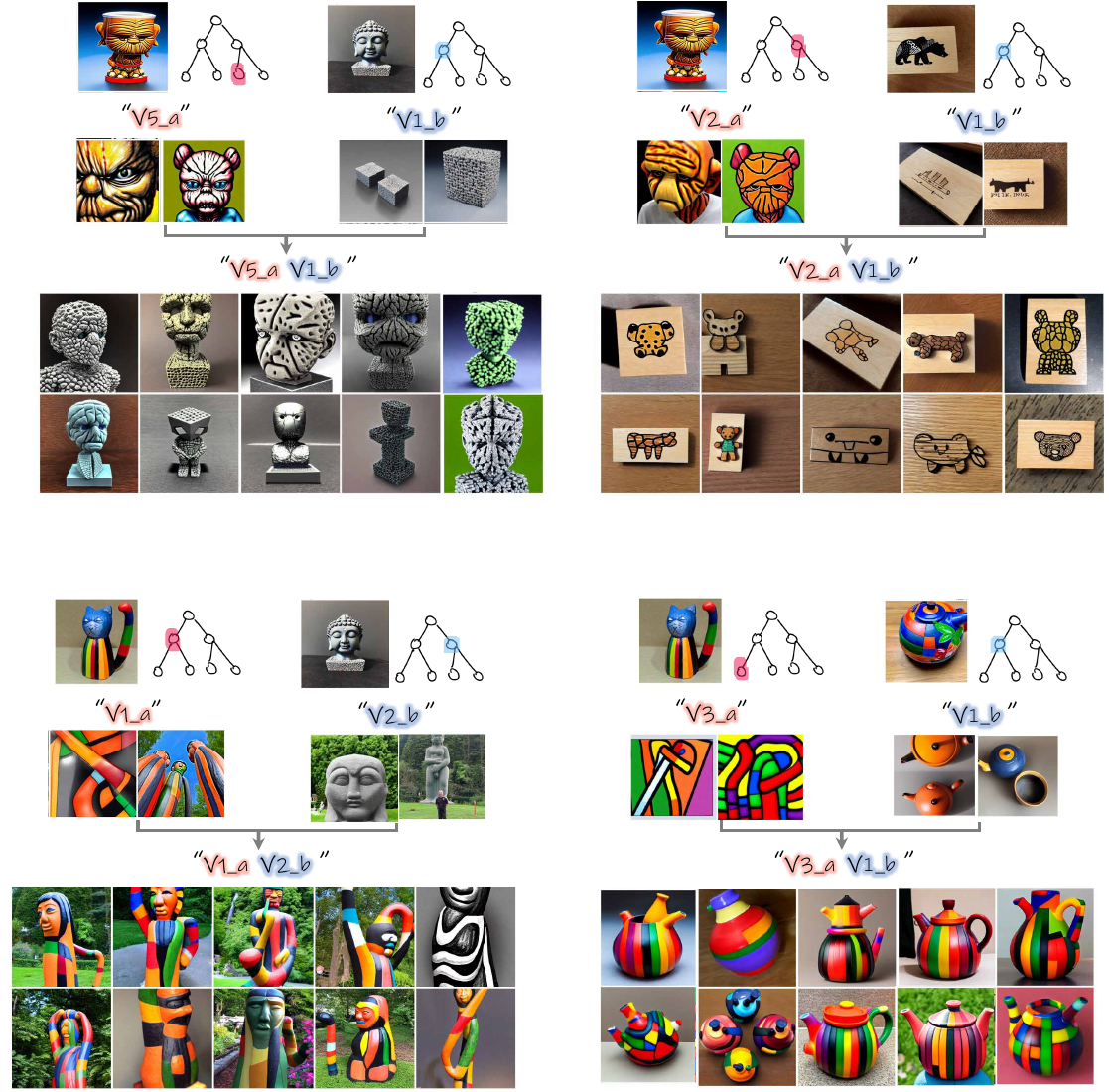}
    \caption{\small More examples of inter-tree combinations.}
    \label{fig:comb1}
\end{figure*}

\begin{figure*}
    \centering
    \includegraphics[width=1\textwidth]{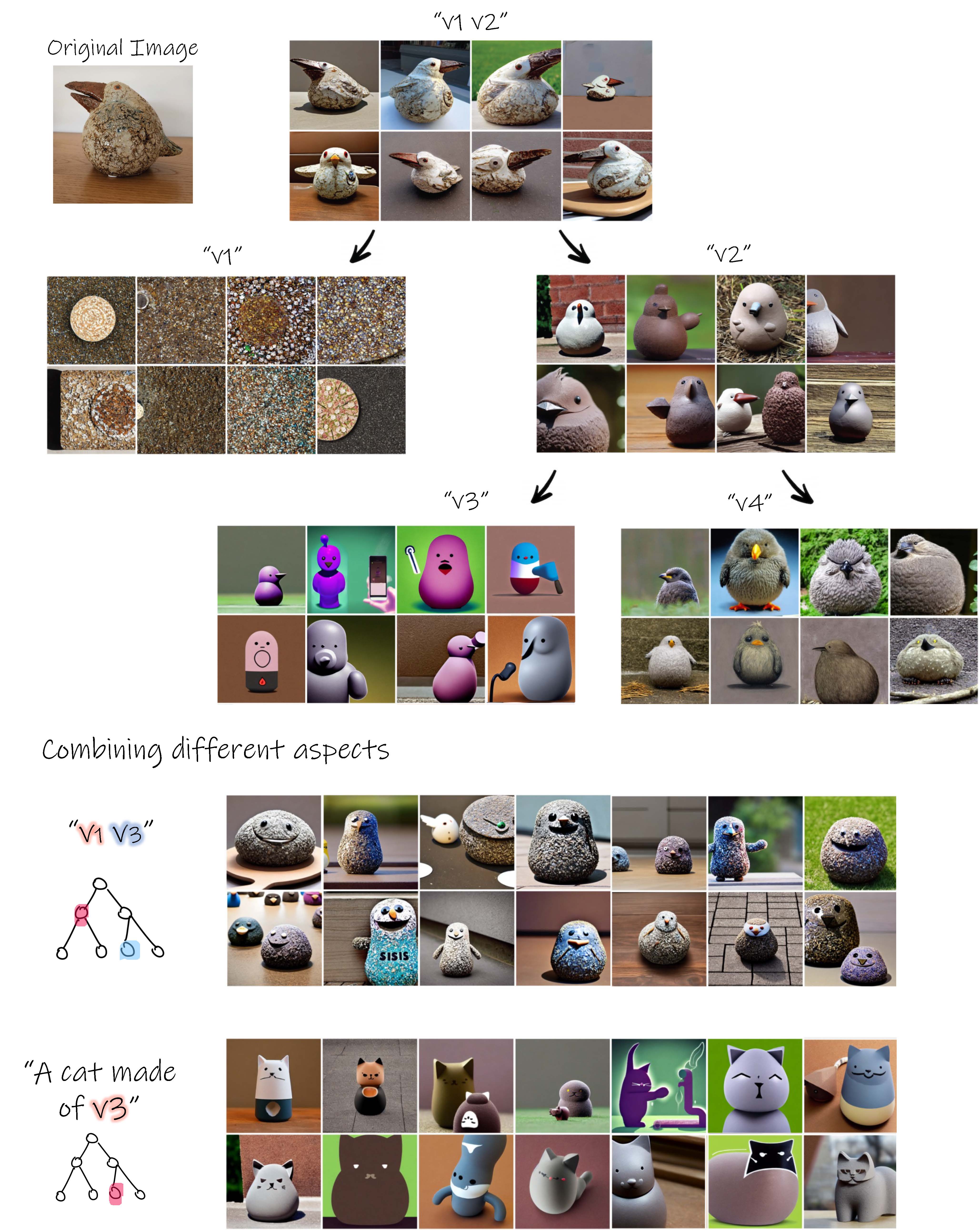}
    \caption{\small Exploration tree for the \ap{round bird} object. At the bottom we show examples of possible intra-tree combinations and text-based generation.}
    \label{fig:round_bird}
\end{figure*}

\begin{figure*}
    \centering
    \includegraphics[width=1\textwidth]{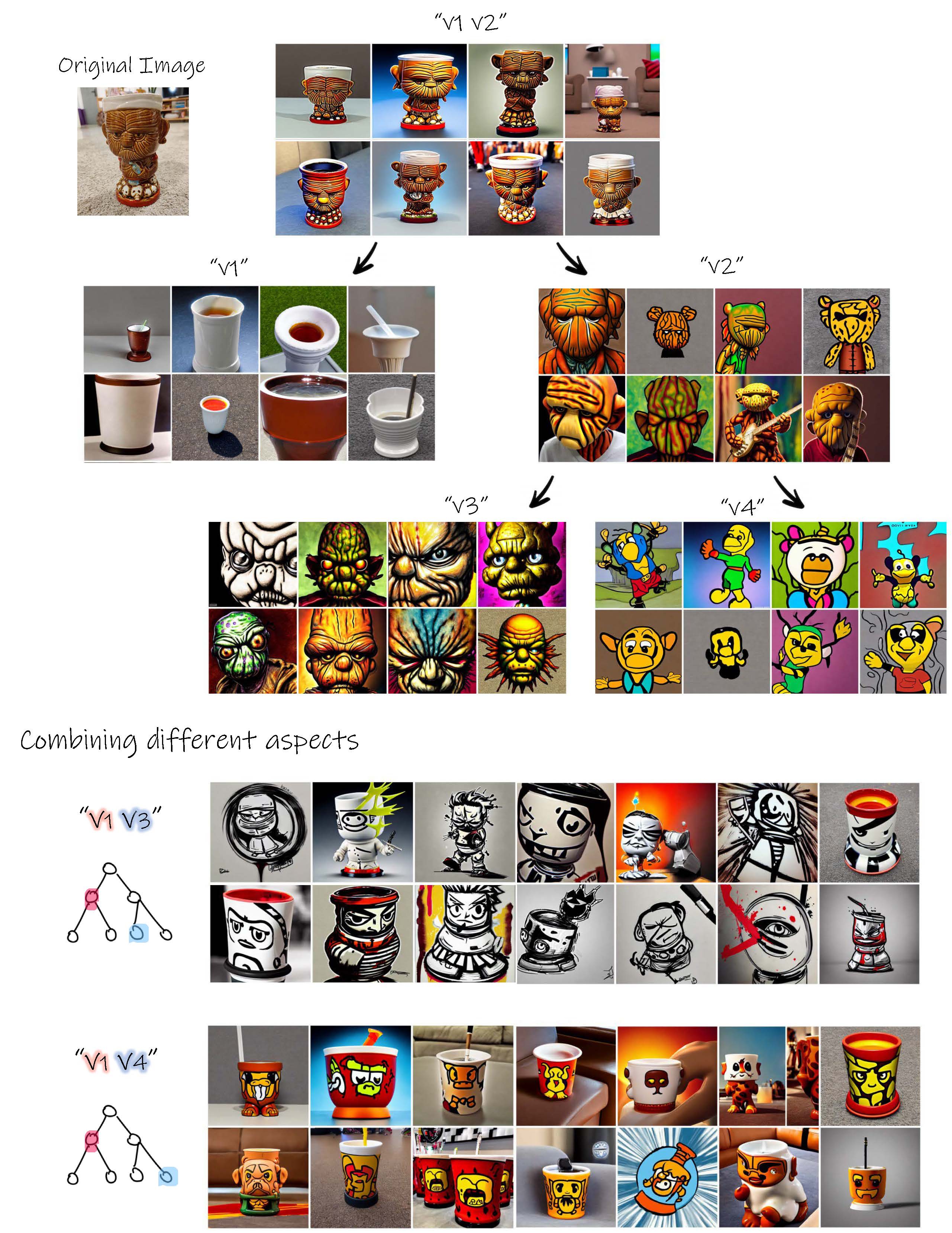}
    \caption{\small Exploration tree for the \ap{scary mug} object. At the bottom we show examples of possible intra-tree combinations.}
    \label{fig:mug_tree}
\end{figure*}

\begin{figure*}
    \centering
    \includegraphics[width=1\textwidth]{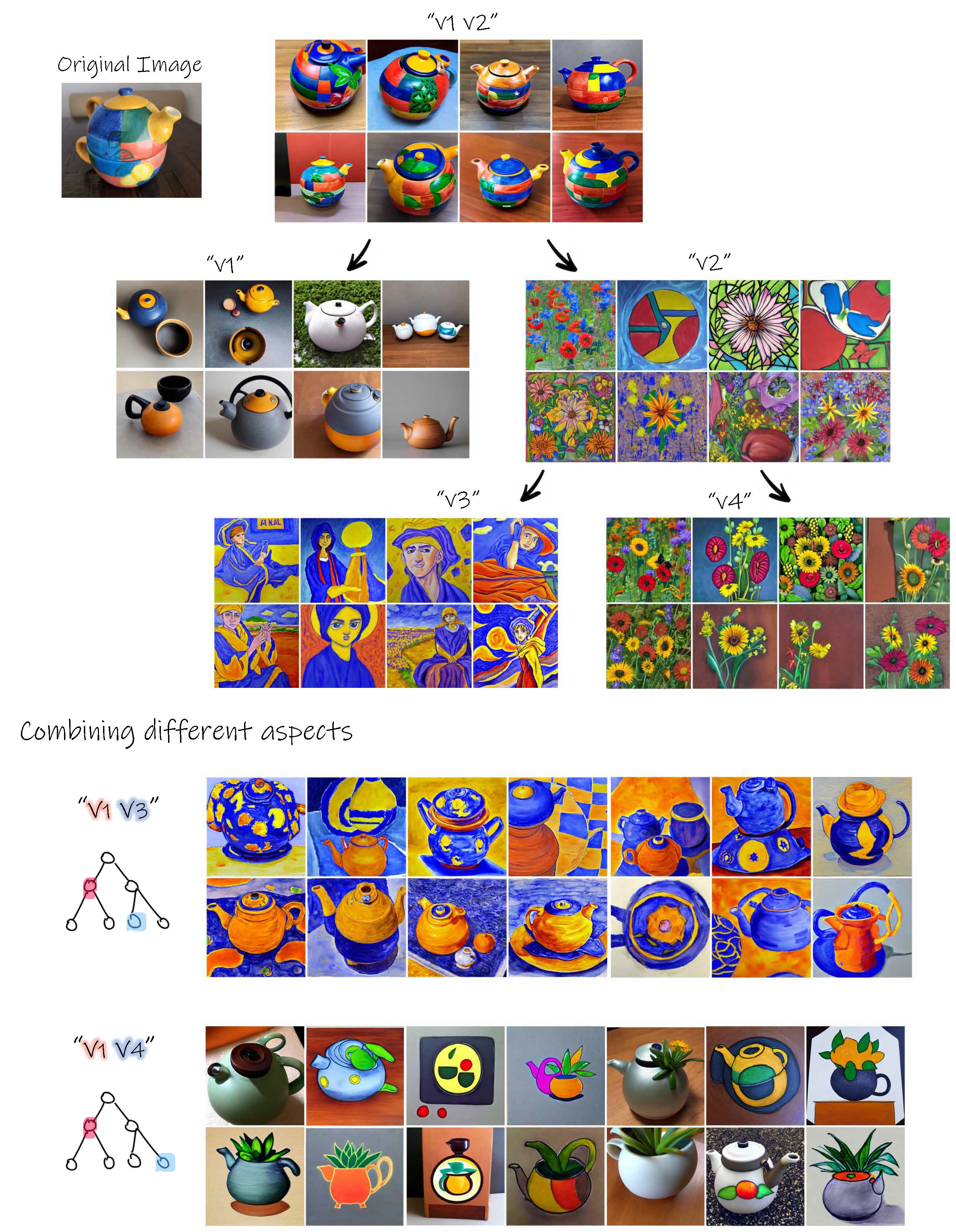}
    \caption{Exploration tree for the \ap{colorful teapot} object. At the bottom we show examples of possible intra-tree combinations.}
    \label{fig:teapot_tree}
\end{figure*}

\begin{figure*}
    \centering
    \includegraphics[width=1\textwidth]{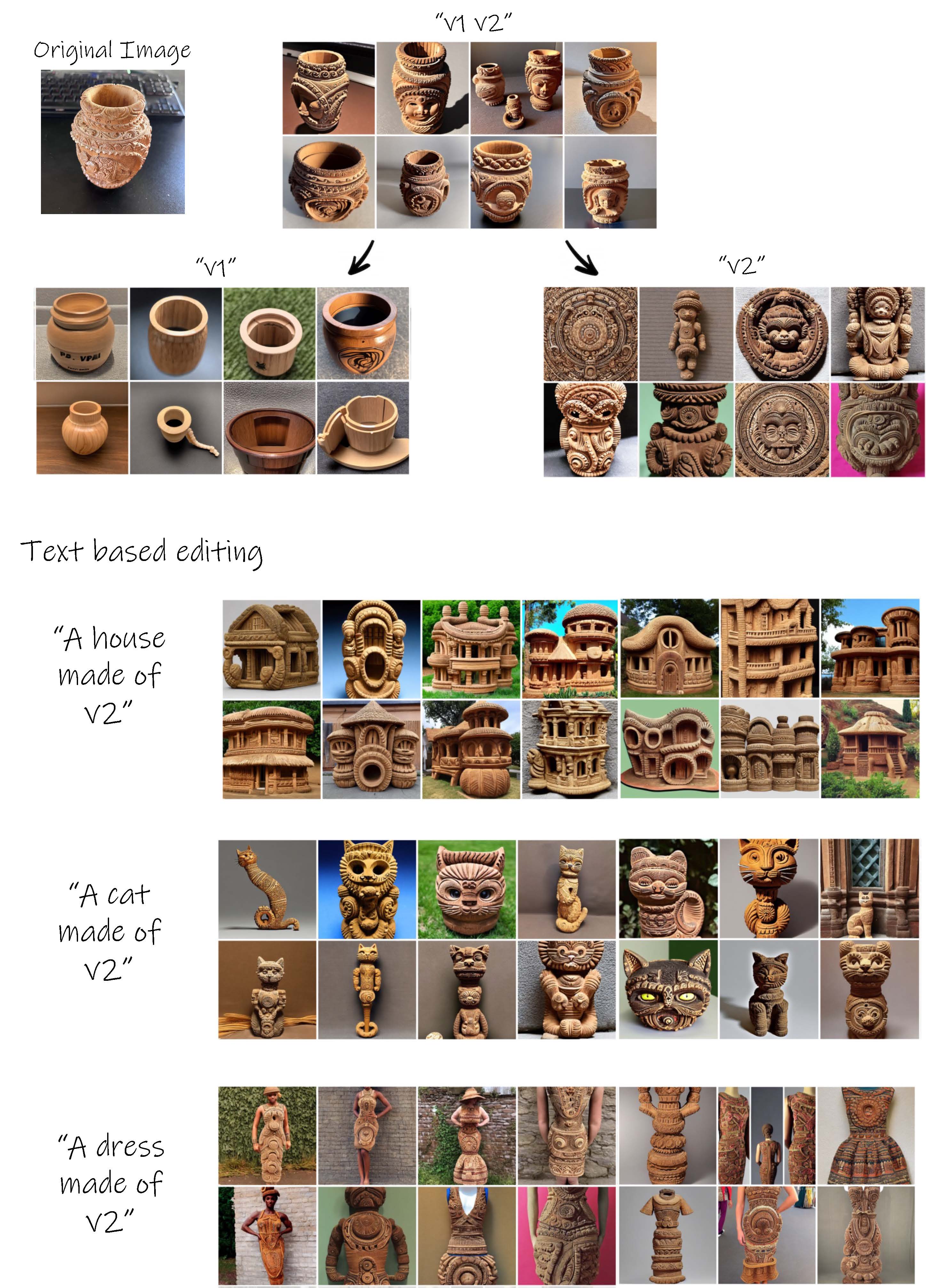}
    \caption{Exploration tree for the \ap{wooden pot} object. At the bottom we show examples of possible text-based generation.}
    \label{fig:wooden_pot}
\end{figure*}

\begin{figure*}
    \centering
    \includegraphics[width=1\textwidth]{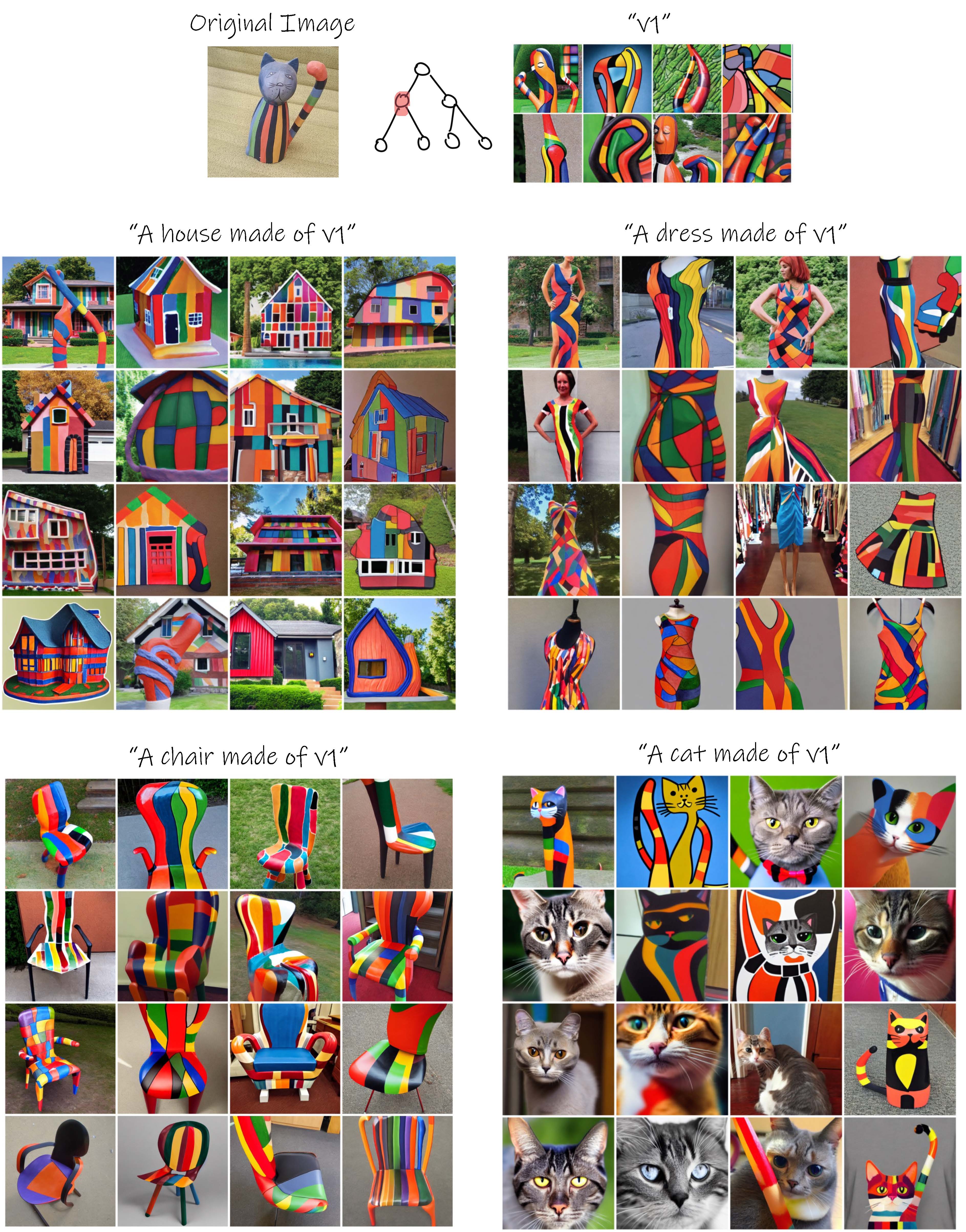}
    \caption{More examples of text based generation for the \ap{cat sculpture} object. The full original tree is shown in the main paper.}
    \label{fig:cat_text_editing1}
\end{figure*}

\begin{figure*}
    \centering
    \includegraphics[width=1\textwidth]{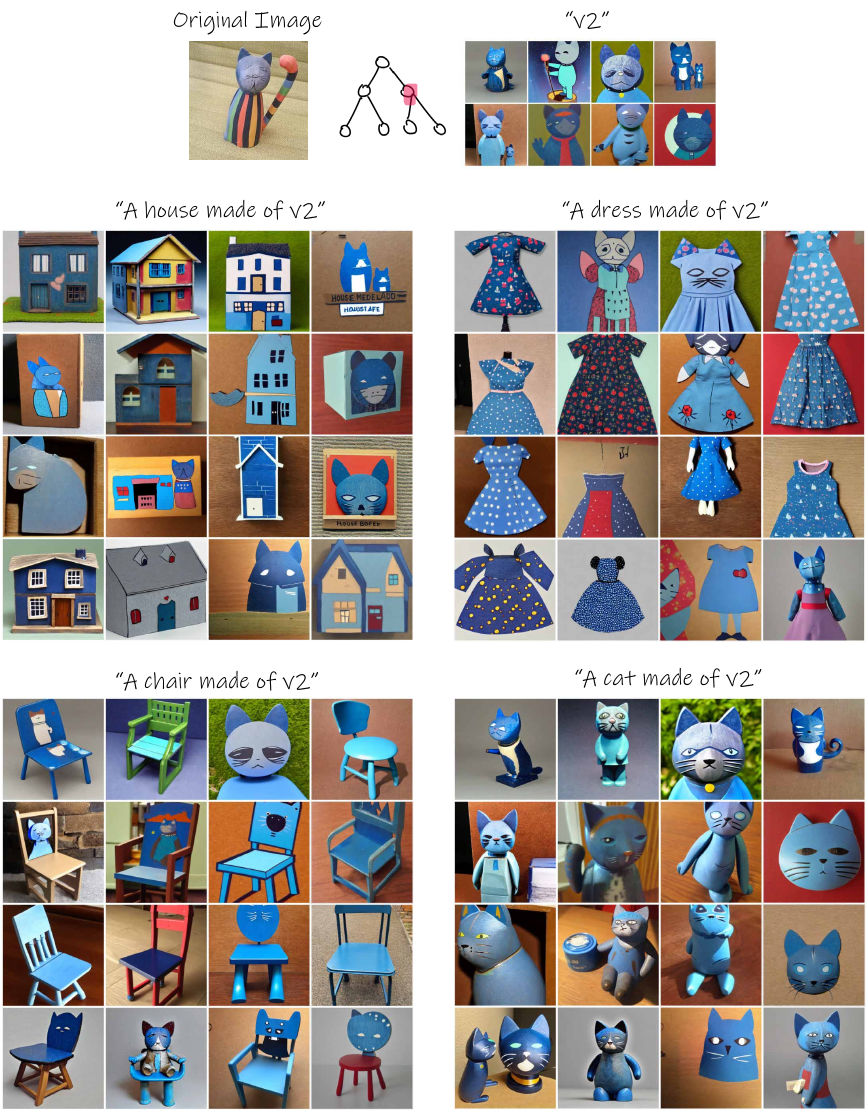}
    \caption{More examples of text based generation for the \ap{cat sculpture} object. The full original tree is shown in the main paper.}
    \label{fig:cat_text_editing2}
\end{figure*}

\begin{figure*}
    \centering
    \includegraphics[width=1\textwidth]{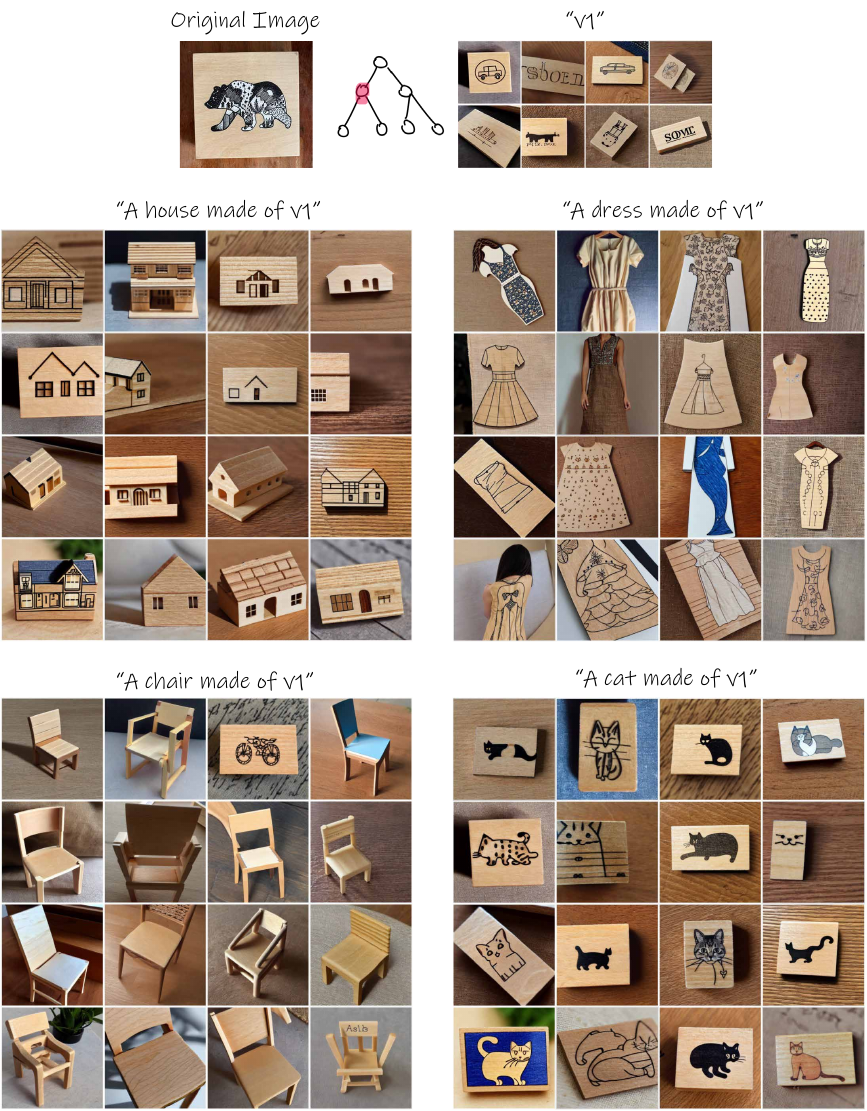}
    \caption{More examples of text based generation for the \ap{wooden saucer bear} object. The full original tree is shown in the main paper.}
    \label{fig:bear_text_editing1}
\end{figure*}

\begin{figure*}
    \centering
    \includegraphics[width=1\textwidth]{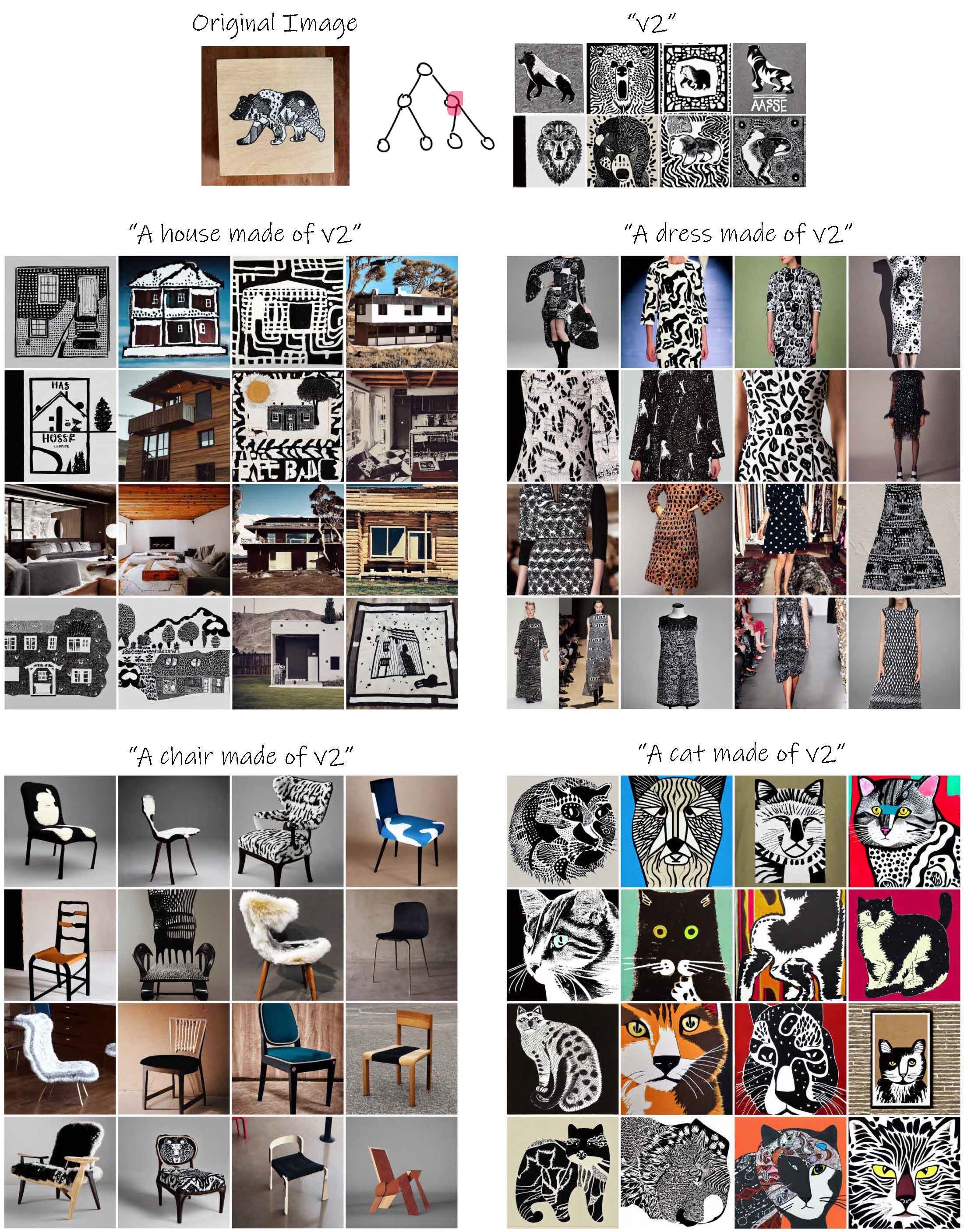}
    \caption{More examples of text based generation for the \ap{wooden saucer bear} object. The full original tree is shown in the main paper.}
    \label{fig:bear_text_editing2}
\end{figure*}

\chapter{Conclusions and Future Work}
Recent advancements in large vision-language models mark an exciting period where generative research meets the world of art and design.

In this dissertation, we introduced generative tools capable of tackling complex design-related tasks by harnessing the visual-semantic knowledge embedded within pretrained VLMs. Our focus was on defining generative tasks that demonstrate the capability of these models to assist in generating effective visual communication designs.

Specifically, in \Cref{part:one}, we presented two methods for automatically generating sketches from images under different settings.
In these works, we demonstrated that CLIP can effectively convey the geometric and semantic essential attributes of a given object through posing the task of generating abstract sketches.
Moreover, unlike previous research in this domain, we highlight the significance of abstraction in the context of sketch generation, proposing hypotheses for defining different types and levels of abstraction. We also employ vector representation to allow for greater artistic control.

In \Cref{part:two,part:three}, we presented two methods that harness the capabilities of pretrained vision-language models to facilitate challenging design-related vector-based tasks, demonstrating how such models can support designers in creating effective visual communication designs. 
The first method entails the automatic creation of word-as-image illustrations in vector format. Our word-as-image illustrations showcase visual creativity and introduce the possibility of employing large vision-language models for semantic typography. In the subsequent method, we introduced a technique to breath life into a given static sketch based on a given text prompt. This approach capitalizes on the motion cues captured by pretrained text-to-video models.

In \Cref{part:four}, we presented a method to implicitly decompose a given visual concept into various aspects to construct an inspiring visual exploration space.
Our method can be used to generate numerous representations and variations of a certain subject, to combine aspects across objects, as well as to use these aspects as part of natural language sentences that drive visual generation of novel concepts.

Our research explored generative text-to-image models from a distinctive angle, one that emphasizes the significance of abstraction and simplified representations in design—a perspective that remains relatively unexplored in most generative research. Additionally, our work is driven by a consideration of the designer's viewpoint, leading us to develop tools that utilize editable and practical representations, such as vector graphics. Moreover, our tasks are initially inspired by observing how designers engage in the design process, including activities like seeking visual inspiration and utilizing sketches.

We hope our work will open the door to further research aimed at developing tools that can generate effective visual communication designs, to assist and inspire designers and artists.
Several potential avenues for future research are highlighted below:

\paragraph{Abstraction beyond sketches} In this dissertation, we focused on sketches as the primary medium for visual abstraction. However, the concept of abstraction can extend beyond sketches into various other domains. A promising avenue for future research involves defining visual abstraction within new contexts that incorporate additional representations such as color, shape, and more. This notion also relates to our exploration of concept decomposition, where concepts are broken down into constituent parts to facilitate ideation. Combining abstraction with concept decomposition could lead to novel insights and approaches in artistic processes. Identifying the essence of visual concepts is crucial for artistic endeavors, and the development of automated tools that offer suggestions in this realm could greatly enhance the ideation process.

\paragraph{Sketch generation} We believe this field is still in its infancy. While we have introduced approaches towards a more robust sketch generation framework, we are still far from matching human-drawn sketches. Possible directions for advancement include enhancing real-time generation speed (as opposed to our optimization-based methods), defining and implementing new types of sketch abstraction inspired by human sketch production, and extending sketch generation to novel task settings beyond image-based or text-based inputs. A promising direction involves developing systems that facilitate sequential sketching. The significance of sequential sketching lies in the assumption that free exploration is core to ideation, where each step influences the subsequent one, leading to unforeseen results. Similarly, a sequential generative model for visual abstraction could yield different outcomes each time, reflecting its unique creation process, akin to the variability observed in human sketching.

\paragraph{Vector representation of data} Many of the ideas presented in this dissertation utilize vector representation to facilitate design related tasks. Vector graphics finds widespread usage in scientific and artistic applications, including architecture, 3D rendering, typography, and graphic design.
Vector graphics often exhibit a compact, scale-free and smooth appearance, making them particularly suitable for icons, emojis, logos, sketches, and more.
A promising future research area is to extend the capabilities of generative models beyond the pixel domain.
Typically, large language-vision models are trained using large datasets of text-image pairs represented as pixels, which makes them well suited to vision-related applications that assume a pixel representation.
Developing strong generative models at large-scale that can model the complex relationship between vector shapes and parameters is a challenging task. However such models could be very useful to develop tools that support designers as vector representation is highly editable and practical.

\printbibliography

\end{document}